%% file: iclr2027_conference.tex
\documentclass{article} %
\usepackage{iclr2027_conference,times}

\input{math_commands.tex}

\usepackage{subcaption} %
\usepackage{hyperref}
\usepackage{url}
\usepackage[capitalize,noabbrev]{cleveref}
\usepackage[T1]{fontenc}
\usepackage{amsmath,amssymb,amsthm,mathtools}
\usepackage{graphicx,booktabs,array,longtable,multirow}
\usepackage{xcolor}
\usepackage{algorithm,algpseudocode}
\usepackage{flafter}
\usepackage{enumitem}
\usepackage{listings}

\crefname{proposition}{Proposition}{Propositions}
\crefname{lemma}{Lemma}{Lemmas}
\crefname{corollary}{Corollary}{Corollaries}

\title{Layer-Informed Fine-Tuning via Three-Stage Functional Segmentation of LLMs}

\author{
Junning Shao$^{1}$, Siwei Wang$^{2}$, and Zhixuan Fang$^{1,3}$\thanks{
Corresponding author: Zhixuan Fang
(\href{mailto:zfang@mail.tsinghua.edu.cn}{\texttt{zfang@mail.tsinghua.edu.cn}}).}\\
$^{1}$Tsinghua University, Beijing, China\\
$^{2}$Microsoft Research Asia, Beijing, China\\
$^{3}$Shanghai Qi Zhi Institute, Shanghai, China
}

\iclrfinalcopy
\hypersetup{
  pdftitle={Layer-Informed Fine-Tuning via Three-Stage Functional Segmentation of LLMs},
  pdfauthor={Junning Shao, Siwei Wang, Zhixuan Fang}
}
\begin{document}

\maketitle
\lhead{Preprint}

\begin{abstract}
In recent years, the performance of large language models (LLMs) on reasoning tasks has been remarkable, even surpassing human capabilities on various benchmarks. However, there remains a lack of clear understanding in the academic community regarding how the structure and internal parameters of LLMs progressively solve complex reasoning problems.  
In this study, we investigate the inference process of LLMs on cross-linguistic materials and propose the hypothesis that LLM layers exhibit a structured division of labor across conceptualization, reasoning, and textualization. 
Based on this hypothesis, we introduce a bottleneck identification mechanism using sensitivity analysis to pinpoint the most critical functional stage for a specific task. Leveraging this insight, we propose a novel approach, Layer-Informed Fine-Tuning (LIFT), which achieves efficient and effective fine-tuning by selectively updating only these functionally critical layers.
We then conduct extensive experiments to show that the LIFT method not only accelerates the training process but also significantly improves model performance.
\end{abstract}

\section{Introduction}

In recent years, large language models (LLMs) \citep{meta2024introducing, bai2023qwen, achiam2023gpt,abdin2024phi,guo2025deepseek} have demonstrated outstanding performance in a wide range of tasks and made significant progress in handling complex problems such as reasoning \citep{wei2022emergent}, mathematics \citep{OpenAI2024,LlamaWebsite2024}, and programming \citep{guo2024deepseek}. 
LLMs, typically built on deep transformer architectures \citep{waswani2017attention}, iteratively refine representations to capture increasingly complex linguistic and logical patterns.
Research on how LLMs internally comprehend problems, perform reasoning, and ultimately generate correct answers has become a highly significant topic. 

\begin{figure*}[t]  %
    \centering
    \includegraphics[width=0.9\textwidth]{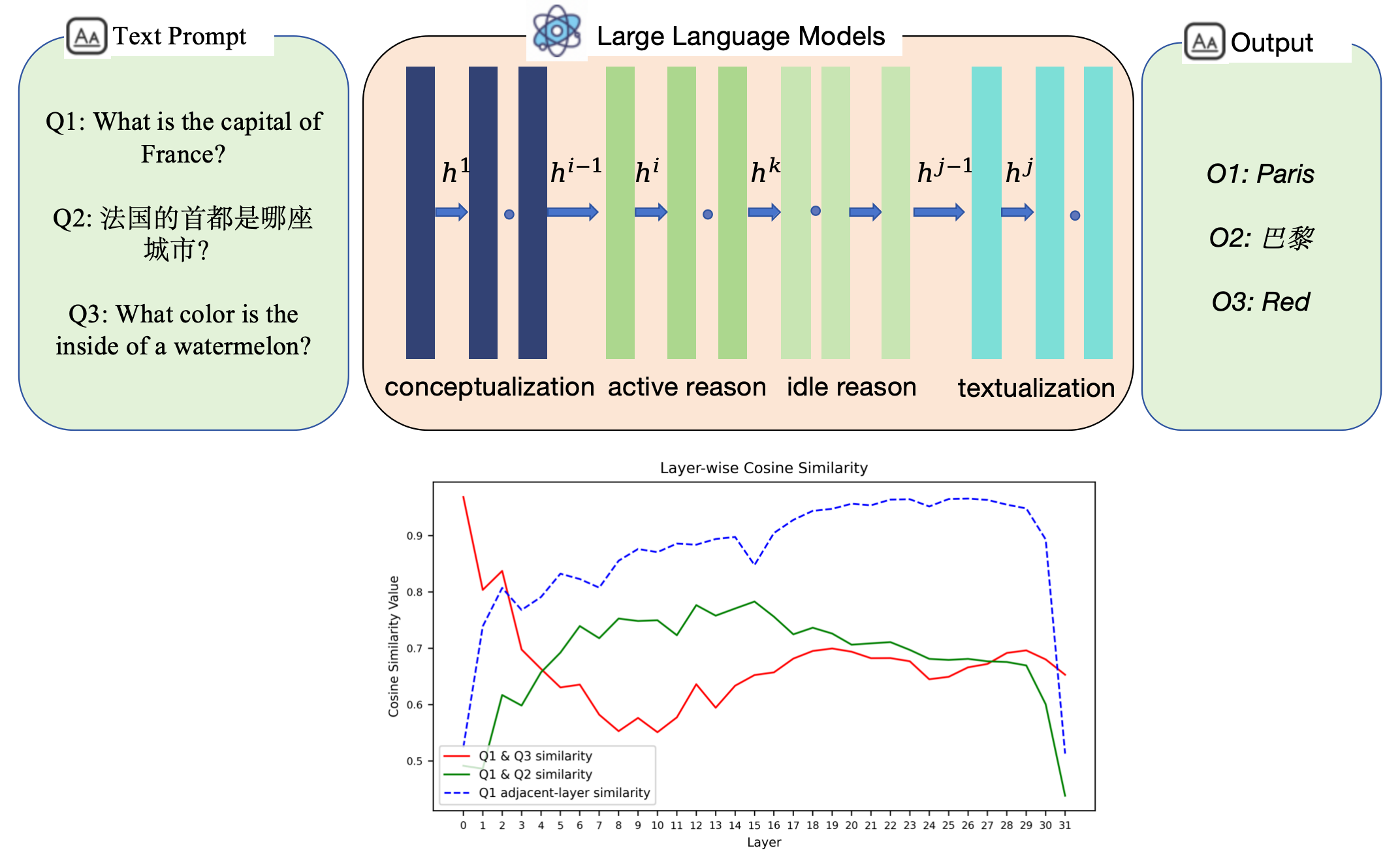}  %
    \caption{The upper part of the figure illustrates how an LLM processes a simple question under the Three-Stage Functional Segmentation hypothesis. The lower part presents a toy experiment validating this hypothesis by comparing hidden state similarities across conceptualization, reasoning, and textualization layers. We have also added a curve showing the cosine similarity between adjacent layers, with higher values indicating greater redundancy in those layers. In the figure, Q2 and O2 are the Chinese expressions of Q1 and O1, respectively.}
    \label{figure1}
\end{figure*}
\vspace{-0.1cm}
Many studies investigate the mechanistic interpretability of LLMs using methods such as the logit lens \citep{nostalgebraist2020logit}, structural probing \citep{hewitt2019structural}, and cosine‐similarity analyses between hidden states \citep{timkey2021all}.
For example, \citet{wendler2024llamas} found that when generating non-English text, the logit-lens-decoded tokens %
initially resemble English before transitioning into the target language in the last few layers of the LLM. 
Another study \citep{li2024safety} focuses on the safety of LLMs and finds that certain consecutive intermediate layers may have the ability to detect harmful questions and decide whether to refuse to answer them. 
In addition, \citet{jin2024exploring} finds that complex problems may require more layers for comprehension and reasoning, whereas simple problems can be processed and resolved within just a few layers.

Inspired by these studies, we investigate the functional division of different layers within LLMs and propose the Three-Stage Functional Segmentation hypothesis.
We posit that LLM layers functionally divide into three contiguous stages: (1) conceptualization layers, where initial layers encode linguistic inputs into abstract, task-relevant representations; (2) reasoning layers, where middle layers perform logical inference to derive latent answers; and (3) textualization layers, where final layers reconstruct these abstract concepts into natural language outputs.  In addition, driven by the observed redundancy (especially in layers closer to the output) in LLM reasoning, we further split the reasoning stage into active and idle layers. Active reasoning layers carry out the main inference work, whereas idle reasoning layers add little or no new information. Following \citet{jin2024exploring}, we also hypothesize that the proportion of idle reasoning layers scales with task complexity, shrinking for harder tasks and growing for easier ones.

Figure \ref{figure1} illustrates the hypothesis using the query ``What is the capital of France?'': the model encodes the intent (conceptualization layers), derives the answer ``Paris'' (reasoning layers), and decodes it into natural language (textualization layers). To validate this, the lower part of Figure \ref{figure1} tracks hidden state similarity. For semantically equivalent cross-lingual queries (Q1, Q2), similarity rises during conceptualization, peaks in the reasoning stage, and drops during textualization as outputs diverge into different languages. Conversely, for distinct queries in the same language (Q1, Q3), similarity declines in conceptualization, remains low during reasoning, but rises in textualization as representations align to the target language. Additionally, a peak in adjacent-layer similarity signals the presence of redundant idle reasoning layers.

Motivated by the Three-Stage Functional Segmentation hypothesis, we posit that efficient model adaptation does not require updating all parameters, but rather targeting the specific functional component that limits task performance. Drawing an analogy to \textbf{Liebig's Law of the Minimum} \citep{von1841organic}—which states that growth is dictated not by total resources but by the scarcest resource—we hypothesize that a model's performance on a specific task is governed by its functional bottleneck. 
To operationalize this, we propose a systematic framework comprising two key steps: (1) \textit{Location}: automatically delineating the boundaries of conceptualization, reasoning (partitioned into active and idle), and textualization layers; and (2) \textit{Diagnosis}: pinpointing the specific stage acting as the bottleneck via a quantization sensitivity analysis. Building on these diagnostics, we introduce \textbf{Layer-Informed Fine-Tuning (LIFT)}, a strategy that maximizes efficiency by selectively fine-tuning only the identified critical layers.

Our contributions can be summarized in three key aspects:  

\begin{itemize}
[leftmargin=*, itemsep=3pt, parsep=0pt, topsep=2pt]
    \item We propose the hypothesis that LLMs can be functionally divided into conceptualization layers, reasoning layers, and textualization layers. Moreover, for each task, we further refine the reasoning layers by partitioning them into active and idle subsets.

    \item We develop a comprehensive diagnostic framework to operationalize this hypothesis. This includes a method for delineating the boundaries of these functional stages and a quantization-based sensitivity analysis to empirically pinpoint the functional bottleneck of the model for a task.
    
    \item We introduce Layer-Informed Fine-Tuning (LIFT), a parameter-efficient strategy that selectively targets the identified critical layers. Our experiments demonstrate that LIFT is both effective and efficient for fine-tuning LLMs.
\end{itemize}
\vspace{-2mm}
\section{Related Works}

\textbf{Internal Structure of LLMs}
Research on the mechanistic interpretability of LLMs often utilizes techniques such as the logit lens \citep{nostalgebraist2020logit}, probing \citep{hewitt2019structural}, and cosine similarity \citep{timkey2021all} to explore their internal structures. 
Building upon fundamental probing techniques, a rich line of research has progressively unveiled the layer-wise functional specialization within language models. Early investigations into masked language models demonstrated that internal representations evolve from capturing surface features to syntactic and semantic properties in a pipeline-like manner \citep{jawahar2019does, tenney2019bert}. As models scaled to large autoregressive architectures, this localized specialization was mapped to specific structural components, identifying feed-forward networks as key-value memories \citep{geva2021transformer} and pinpointing precise layers responsible for storing factual associations \citep{meng2022locating} or encoding function vectors \citep{todd2023function}. 
Recently, the mechanistic interpretability lens has expanded to track complex, multi-step phenomena, tracing where in-context learning emerges \citep{sia2024does}, how layers communicate to compose functions \citep{merullo2024talking, khandelwal2025language}, and why contextualization errors occur \citep{lepori2025racing}. A prominent theme unifying these discoveries is the unique role of intermediate layers. Functionally, middle layers act as a “semantic hub” \citep{wu2024semantic} or a “high-dimensional abstraction phase” \citep{cheng2024emergence} where diverse inputs are mapped to a shared operational space. This is particularly evident in multilingual factual recall \citep{lu2025paths}, where models dynamically map inputs to a latent operational language (often English) for intermediate reasoning before decoding into the target language at the final layers \citep{wendler2024llamas, zhao2024how, zhong2024beyond}. The distinct functional identity of these mid-depth representations is further corroborated by the localization of causally validated “language-selective units” \citep{alkhamissi2024llm} and their superior performance on downstream transfer tasks relative to final-layer outputs \citep{skean2025layer}, with models engaging deeper layers primarily for more complex problems \citep{jin2024exploring}.  These studies provide a valuable ``map'' of the functional properties contained within different layers.

Concurrently, research on model robustness has found that early and final layers are highly sensitive to structural interventions like deletion or swapping, whereas middle layers are remarkably resilient \citep{sun2025transformer, lad2024remarkable}. This resilience is particularly pronounced in the later-middle layers, which have been shown to be more redundant and less impactful on performance when pruned compared to earlier layers \citep{gromov2024unreasonable, men2024shortgpt}, further supporting their distinct functional role.

While these studies provide a valuable ``map'' of the functional properties within LLM layers, our work provides a ``recipe'' for using them. We shift the focus from a static analysis of what is in each layer to the dynamic transitions between these functional stages. This approach allows us to develop targeted and parameter-efficient fine-tuning strategies that improve performance by focusing updates only on functionally-relevant blocks.

\textbf{Parameter-Efficient Fine-Tuning.}
The conventional full-parameter fine-tuning paradigm often incurs high computational costs when fine-tuning LLMs. To address this, parameter-efficient fine-tuning (PEFT) has emerged as a promising alternative. Key PEFT approaches include Adapter Tuning \citep{houlsby2019parameter,pfeiffer2020adapterfusion,karimi2021compacter}, Prompt Tuning \citep{li2023prompt}, and Low-Rank Adaptation (LoRA) \citep{hu2021lora}. Adapter Tuning introduces trainable adapter modules between all model layers, Prompt Tuning incorporates tunable prefix tokens into inputs.The original LoRA \citep{hu2021lora} decomposes the weight updates for various parameter matrices into low-rank adaptation matrices. 
More advanced variants, e.g., AdaLoRA \citep{zhang2023adaptive}, improve upon this by dynamically allocating the parameter budget to more important weight matrices based on their sensitivity during training. 
These methods only need to fine-tune a small subset of parameters, typically less than 1\% of the original, and can significantly reduce computational and memory overhead.  

Compared with the above methods, our LIFT focuses on fine-tuning specific layers, which follows a distinct trajectory from the above methods and can be seamlessly integrated with them. For instance, in this paper, we incorporate LoRA to achieve both high performance and accelerated fine-tuning.

\vspace{-2mm}
\section{Preliminary}
\vspace{-2mm}
\subsection{Large Language Models} \label{LLMS}
Large Language Models, such as Llama-3 \citep{meta2024introducing}, are built on a transformer-based architecture composed of multiple stacked layers. Each layer consists of a multi-head self-attention mechanism that captures contextual dependencies, a feedforward network for further processing, and residual connections with layer normalization to stabilize training. 
In this framework, hidden states represent the evolving internal representations of tokens, which are progressively enriched layer by layer. These states are updated iteratively by adding residuals, where the residual is computed as a function of the hidden states of all preceding tokens.  
In this paper, we employ two LLMs from different model families: Llama-3-8B-Instruct \citep{meta2024introducing}, and Qwen2-7B-Instruct \citep{qwen2}. These two LLMs were developed by different companies, each with distinct internal architectures and training methodologies, making them highly representative of diverse model designs.

\vspace{-2mm}
\subsection{Datasets} \label{dataset}
To train and evaluate the reasoning capabilities of LLMs, we select three representative logical reasoning datasets: ProofWriter \citep{tafjord2020proofwriter}, FOLIO \citep{han2022folio} and the LogicalDeduction dataset from BigBench \citep{srivastava2022beyond}. Our data processing method follows a similar approach to that proposed in \citet{pan2023logic}. These datasets ensures a comprehensive assessment of the model's ability to perform various forms of logical reasoning. In addition to the reasoning dataset, we also conducted experiments on GSM8K \citep{cobbe2021gsm8k}, a mathematical dataset.
\vspace{-2mm}
\section{Three-Stage Functional Segmentation of LLMs}\label{sec:hypo}

In this section, we present our hypothesis, the Three-Stage Functional Segmentation of LLMs, and provide rigorous and comprehensive experiments to support the hypothesis.

According to this hypothesis, an LLM can be functionally divided into three contiguous stages:
\begin{enumerate}[leftmargin=*, itemsep=3pt, parsep=0pt, topsep=2pt]
    \item \textbf{Conceptualization layers} – located near the input, responsible for transforming natural language into abstract representations.
    \item \textbf{Reasoning layers} – situated in the middle, where task-relevant logical processing occurs. These can be further divided into \textbf{active reasoning layers} and \textbf{idle reasoning layers}.
    \item \textbf{Textualization layers} – positioned near the output, where abstract representations are mapped back into natural language.
\end{enumerate}

The conceptualization layers encode linguistic inputs into abstract, task-relevant representations. This initial stage marks a functional transition from processing surface-level language to forming the conceptual foundation required for logical processing.

The reasoning layers then manipulate these abstract representations to perform logical inference. Operating on task-specific logic rather than surface linguistics, we hypothesize that hidden states in this stage will exhibit high similarity for the same task across different languages, yet low similarity for different tasks within the same language. Furthermore, these layers can be subdivided into active reasoning layers, which execute core inference, and idle reasoning layers, which are redundant.

Finally, the textualization layers translate the inferred abstract concepts back into natural language. This final stage reverses the similarity pattern observed during reasoning: representations for different tasks with the same language converge as they are mapped to a common linguistic form, while those for the same task in different languages diverge into their respective linguistic outputs.

To investigate the functional segmentation of layers within the LLM, we design the following experiments during the inference process.
Suppose a large language model has $K$ hidden layers and a task-consistent dataset $\mathcal{D}_{\text{raw}} = \{ d_i \}_{i=1}^{N}$. Without loss of generality, assume that this dataset is an English dataset. We then define its corresponding datasets in other natural languages as $\mathcal{D}_{\text{Non-English}} = \{ d'_i \}_{i=1}^{N}$.  

The Non-English datasets may include any non-English language supported by the LLM, such as Chinese, French, Spanish, or Hindi.  
The identical index $i$ in these datasets indicates the same semantic content, differing only in language.  

When using these two datasets as the input for inference in the LLM, we obtain the hidden state sets from the last position of each layer in the first autoregressive process, denoted as \( S(D_{\text{raw}}) \) and \( S(D_{\text{Non-English}}) \). The specific form of these vector sets is as follows:
\begin{gather*}
S(D_{\text{raw}}) = \left\{ [h_{(d_i)}^1, h_{(d_i)}^2, \dots, h_{(d_i)}^K] \right\}_{i=1}^N, \qquad \\
S(D_{\text{Non-English}}) = \left\{ [h_{(d'_i)}^1, h_{(d'_i)}^2, \dots, h_{(d'_i)}^K] \right\}_{i=1}^N
\end{gather*}

where \( h_{(d_i)}^k \) denotes the hidden state at the last position in layer \( k \) in the first autoregressive process. 
\vspace{-0.15cm}
\subsection{Cosine Similarity between Hidden States in the Same Layer} \label{Cosine_Similarity}
First, we compute the cosine similarity for the following pairs. These results quantify the similarity between hidden representations, offering insights into how information is processed and transformed within the model.
\vspace{-0.1cm}
\begin{figure*}[htb]
    \centering

    \begin{subfigure}{0.24\textwidth}
        \includegraphics[width=\textwidth]{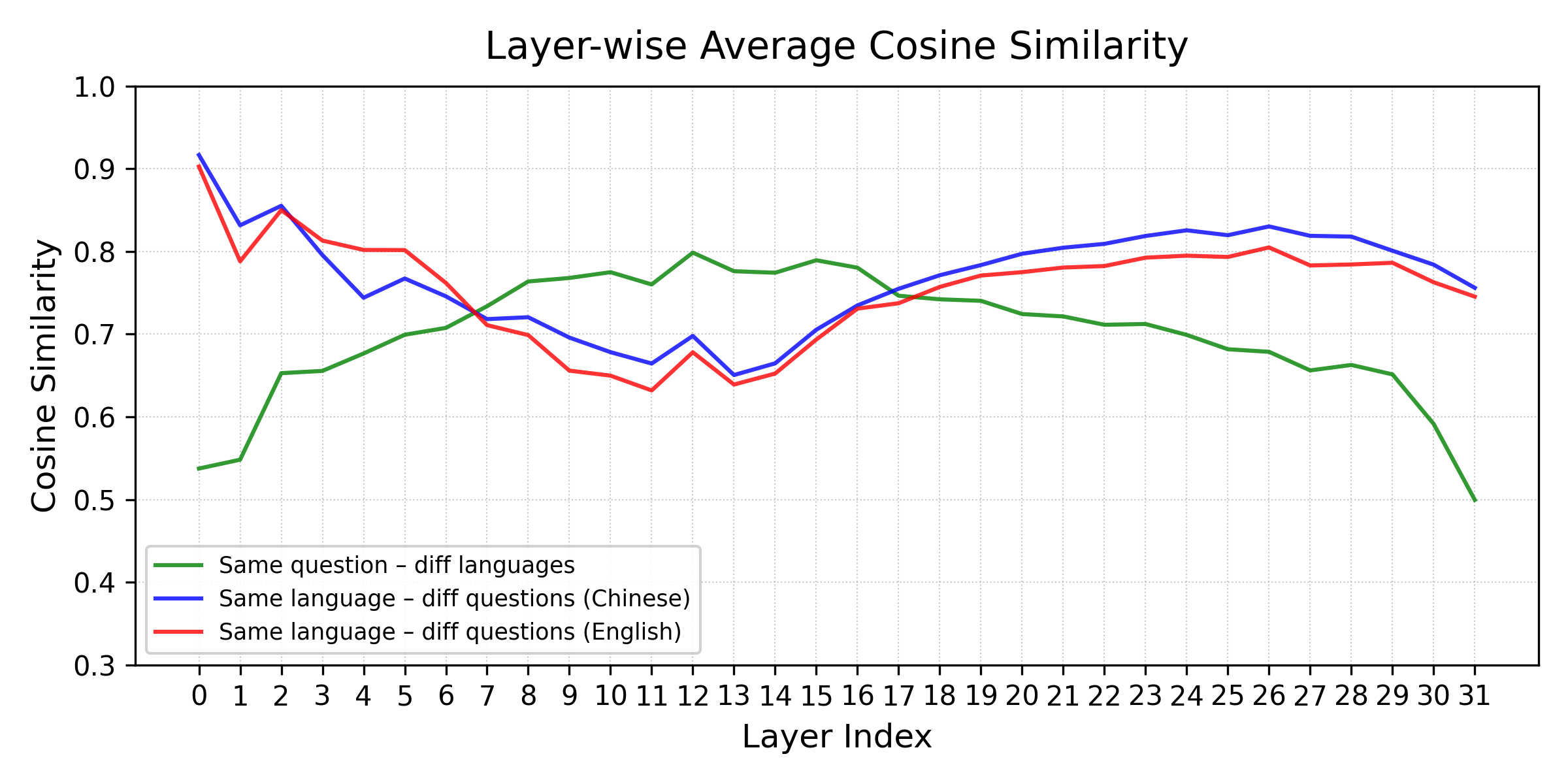}
        \caption{LlaMA-3-8B-Instruct on GSM8K}
    \end{subfigure}
    \begin{subfigure}{0.24\textwidth}
        \includegraphics[width=\textwidth]{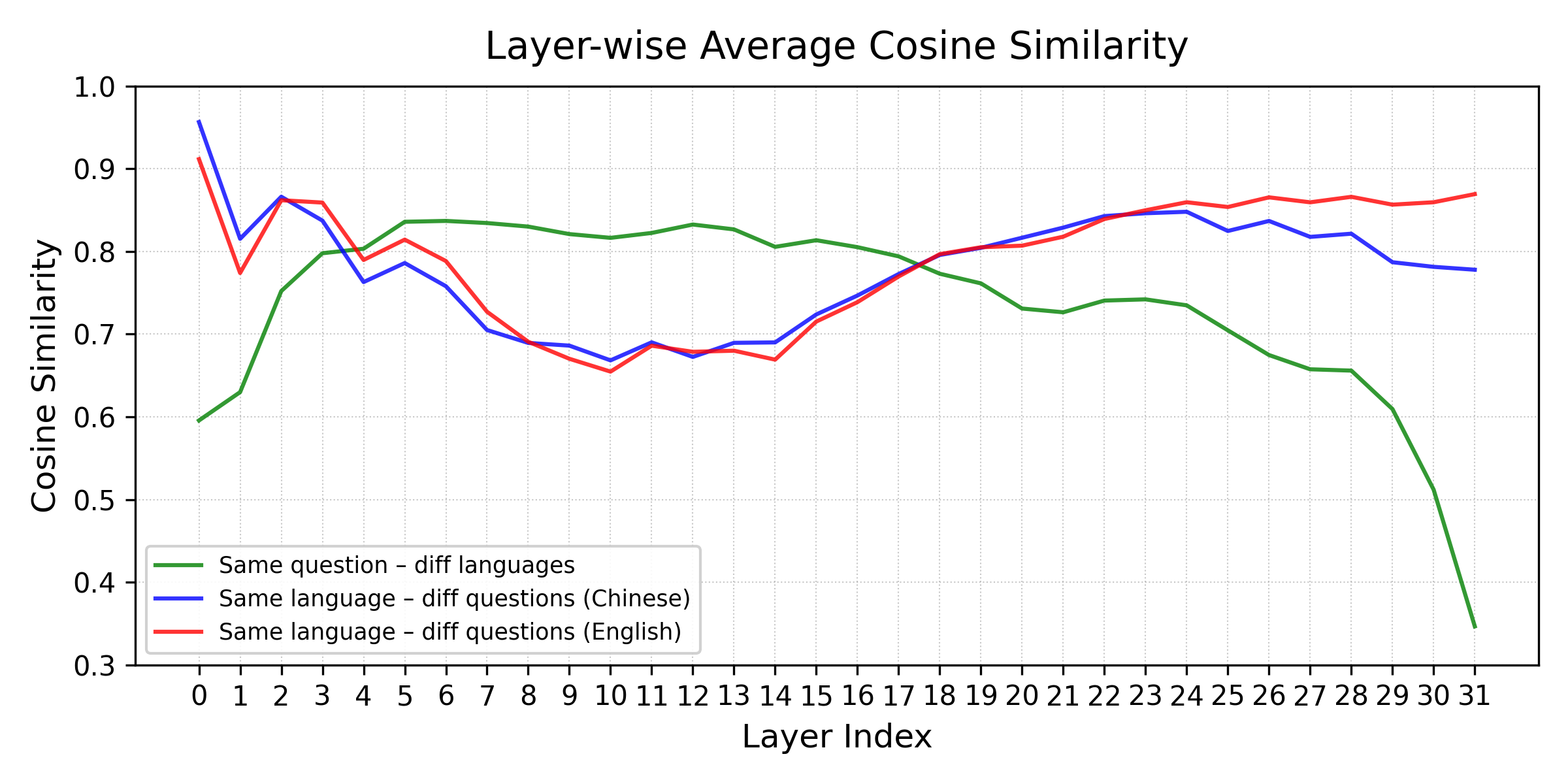}
        \caption{LlaMA-3-8B-Instruct  on LogicalDeduction}
    \end{subfigure}
    \begin{subfigure}{0.24\textwidth}
        \includegraphics[width=\textwidth]{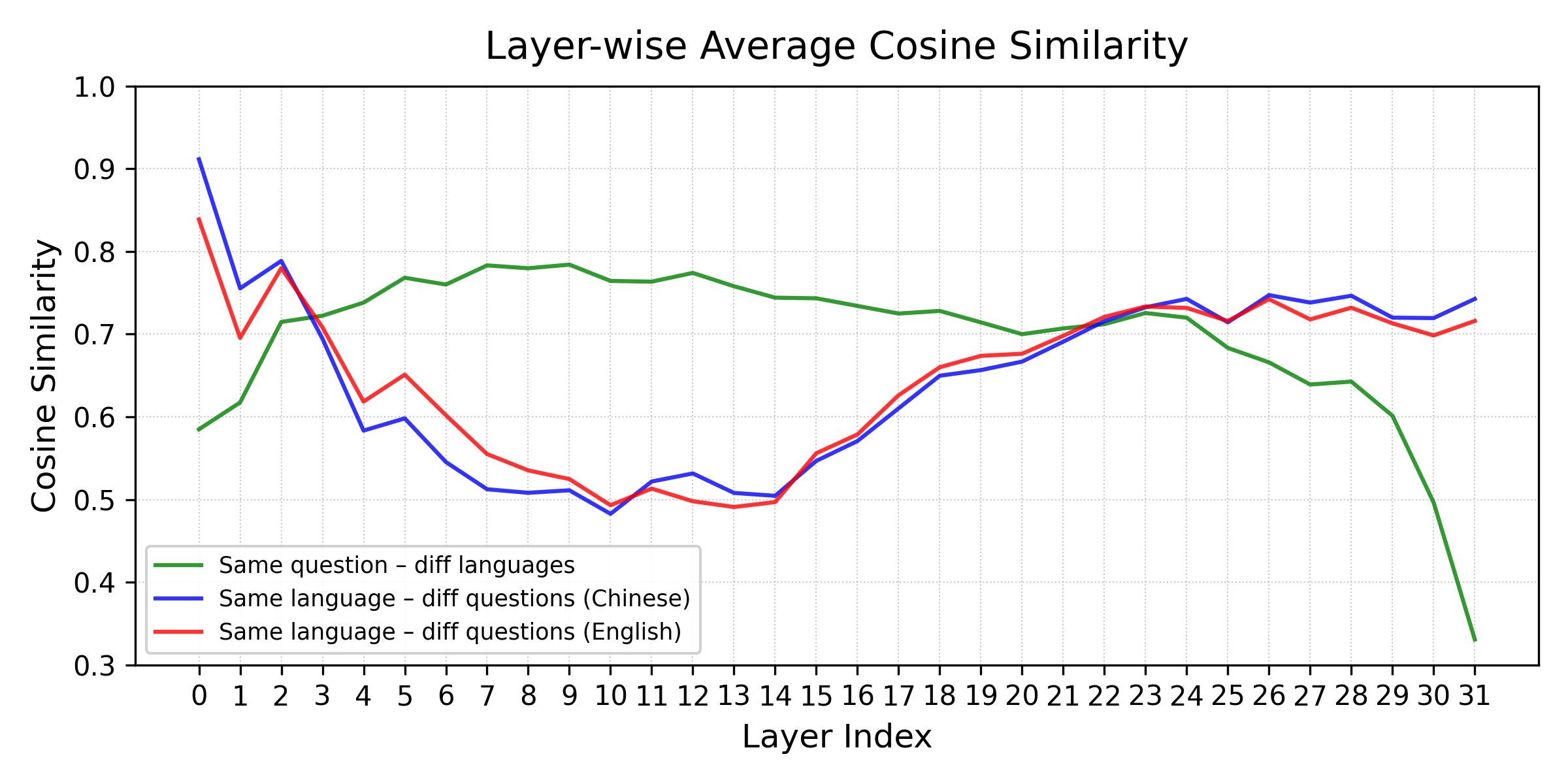}
        \caption{LlaMA-3-8B-Instruct  on FOLIO}
    \end{subfigure}
    \begin{subfigure}{0.24\textwidth}
        \includegraphics[width=\textwidth]{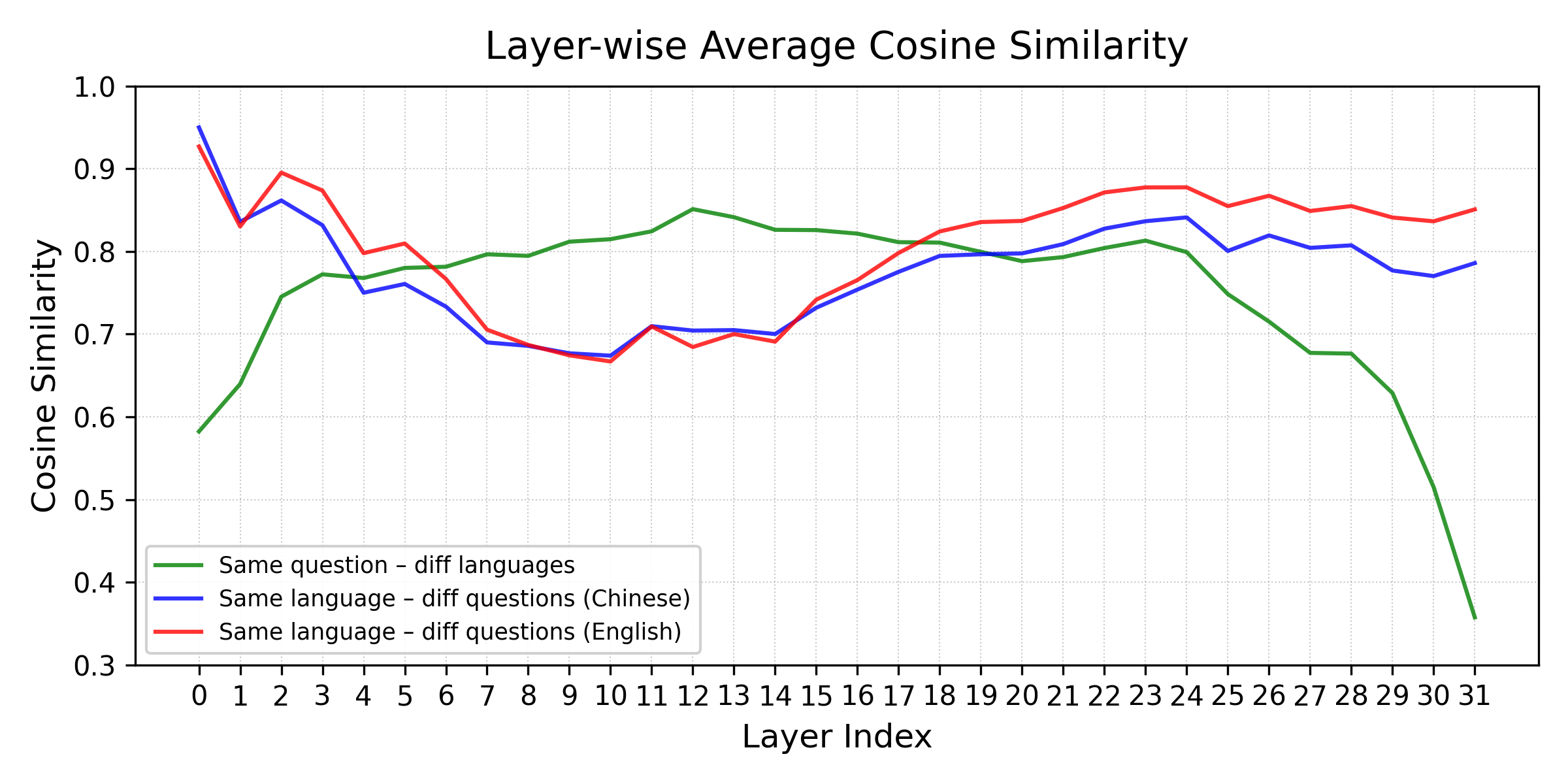}
        \caption{LlaMA-3-8B-Instruct  on ProofWriter}
    \end{subfigure}
    
    \vspace{0.2cm} %

    \begin{subfigure}{0.24\textwidth}
        \includegraphics[width=\textwidth]{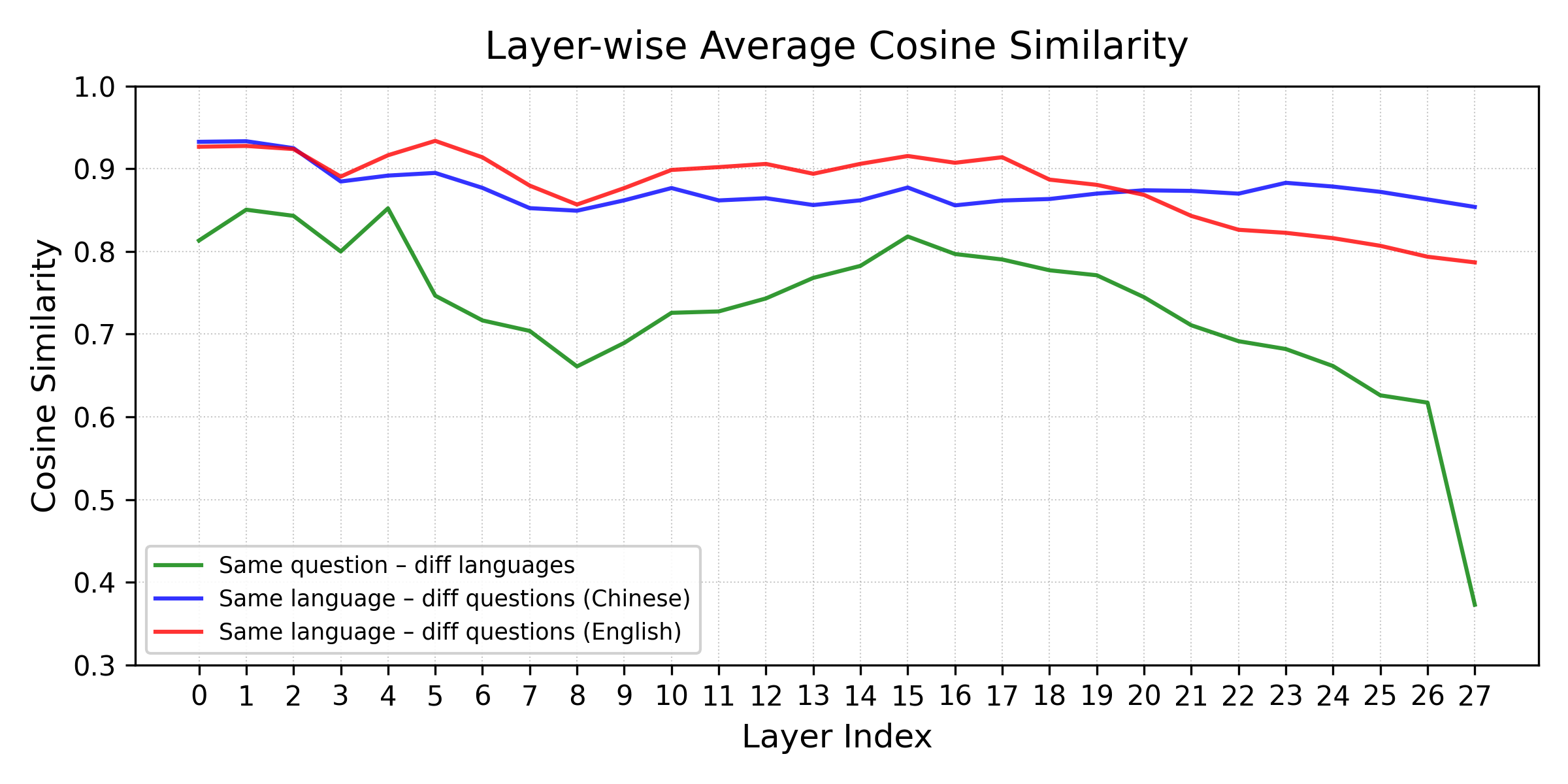}
        \caption{Qwen2-7B-Instruct  on GSM8K}
    \end{subfigure}
    \begin{subfigure}{0.24\textwidth}
        \includegraphics[width=\textwidth]{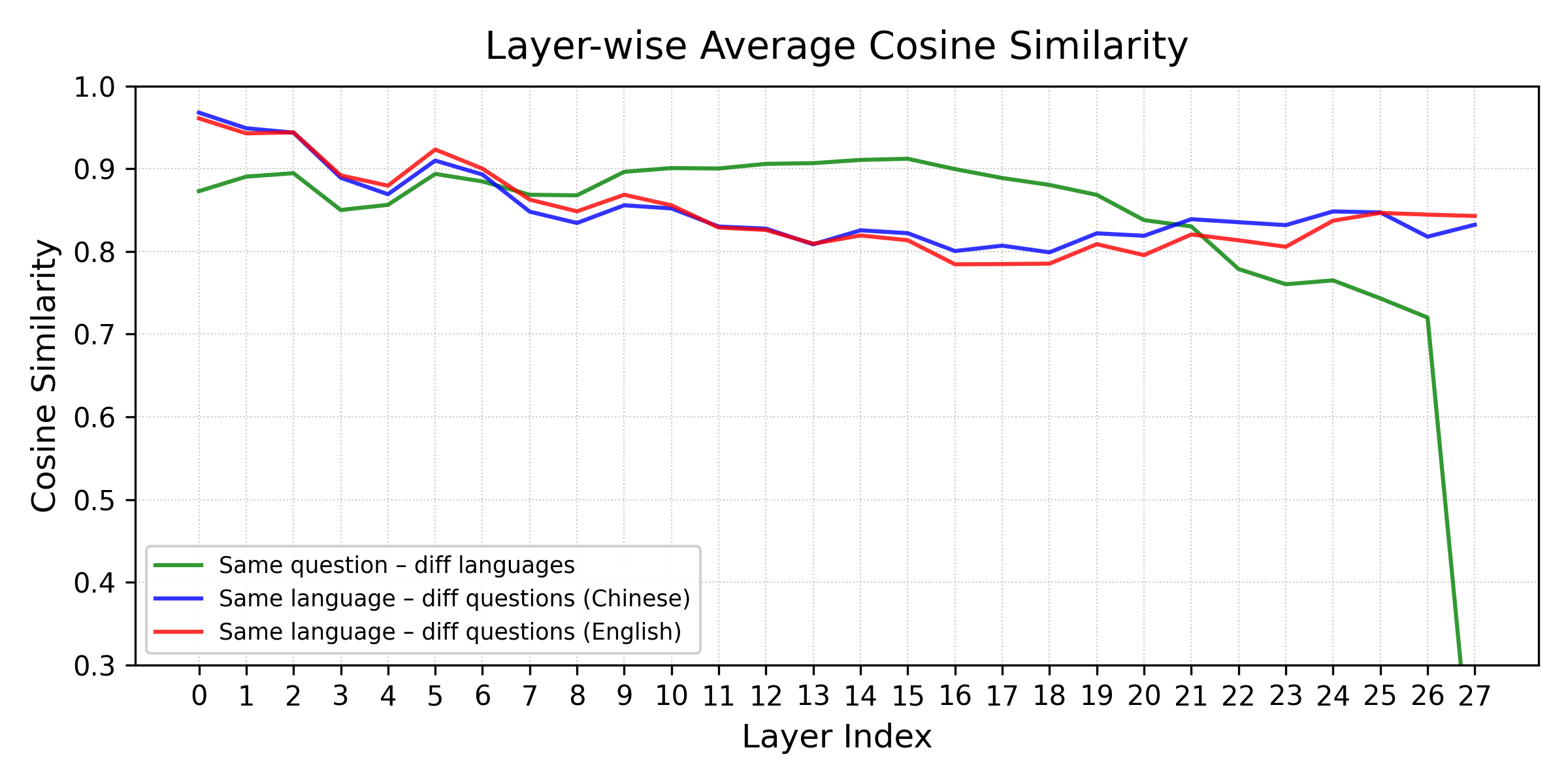}
        \caption{Qwen2-7B-Instruct  on LogicalDeduction}
    \end{subfigure}
    \begin{subfigure}{0.24\textwidth}
        \includegraphics[width=\textwidth]{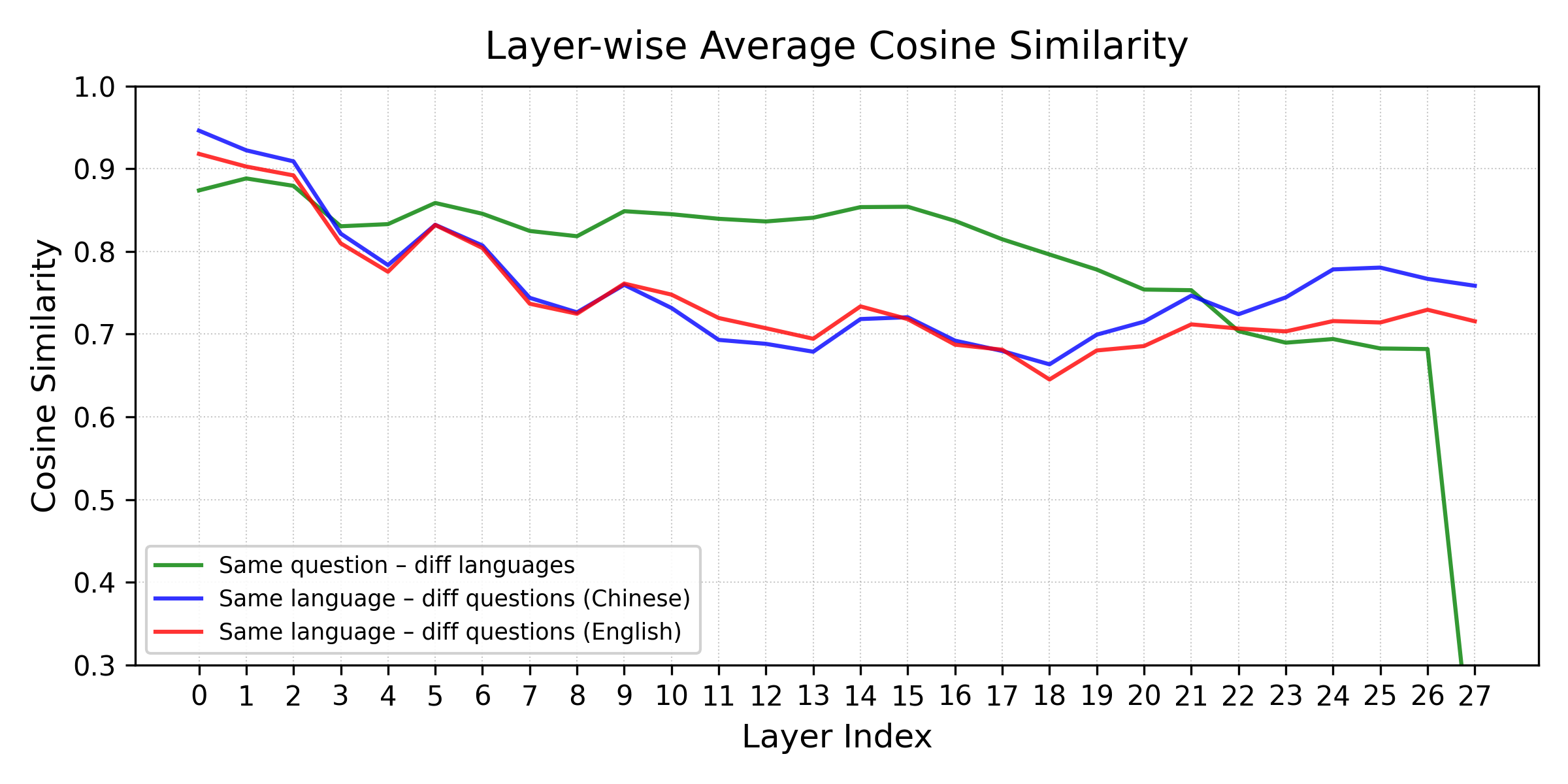}
        \caption{Qwen2-7B-Instruct  on FOLIO}
    \end{subfigure}
    \begin{subfigure}{0.24\textwidth}
        \includegraphics[width=\textwidth]{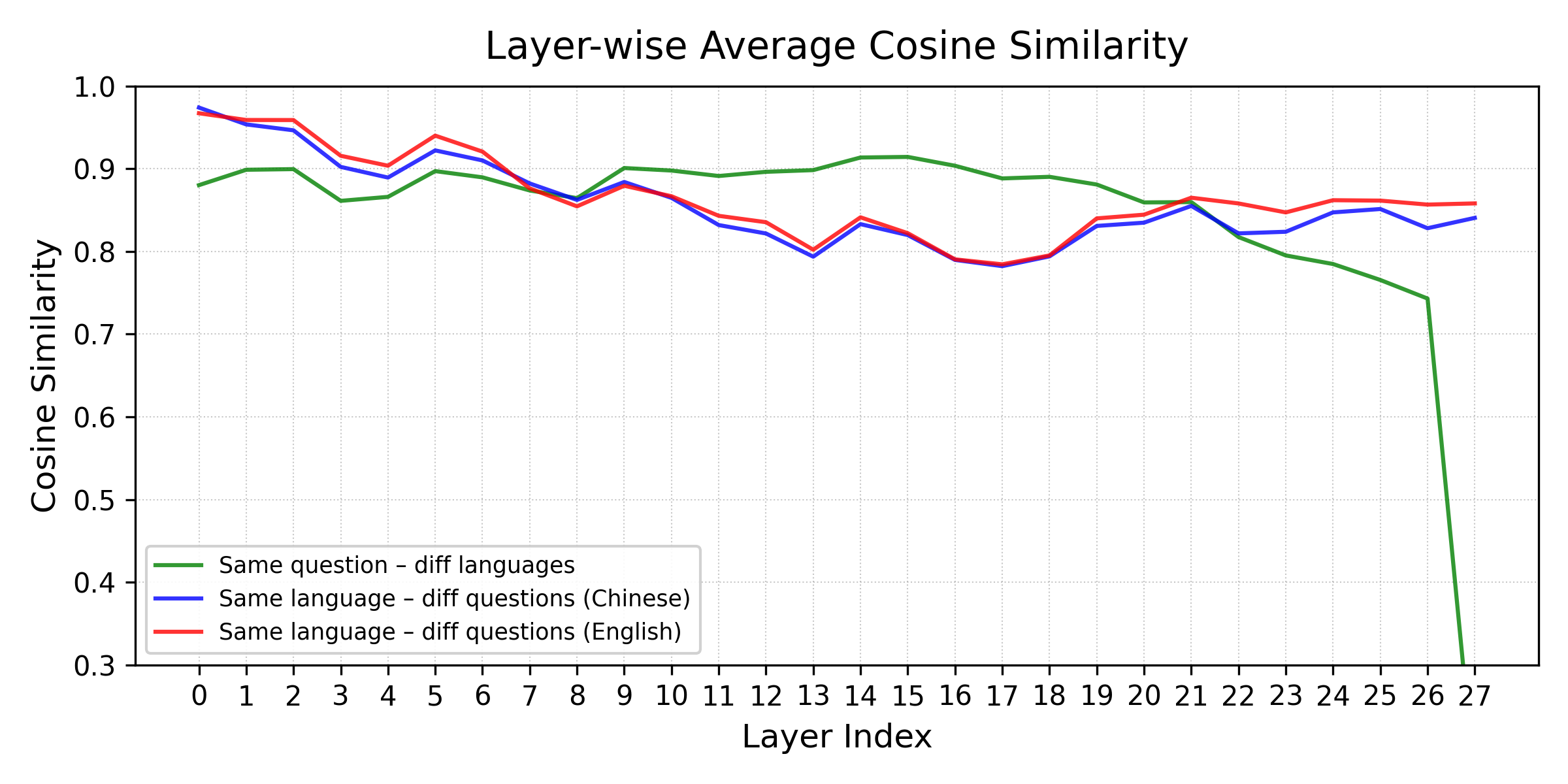}
        \caption{Qwen2-7B-Instruct on ProofWriter}
    \end{subfigure}

    \caption{Cosine similarity between Hidden States in the Same Layer.}
    \label{fig_similar}
\end{figure*}
\vspace{-2mm}

\begin{enumerate}[leftmargin=*, itemsep=3pt, parsep=0pt, topsep=2pt]
    \item \label{same_question} For pairs of the same question in different languages, we randomly select an index \(i\) from 1 to \(N\) and compute the cosine similarity between different language versions of the same question \(i\), denoted as $g_{i}^k = \operatorname{cos}(h_{(d_i)}^k, h_{(d'_i)}^k)$ for $1\le k \le K$. 
    \item For pairs of different questions within the English dataset, we randomly select two indices $i\ne j$ from \( 1 \) to \( N \) and compute the cosine similarity between different question \( i \) and \( j \) in English, denoted as $r_{i,j}^k = \operatorname{cos}(h_{(d_i)}^k, h_{(d_j)}^k)$ for $1\le k \le K$.

    \item For pairs of different questions within the Non-English dataset, we randomly select two indices $i\ne j$  from \( 1 \) to \( N \) and compute the cosine similarity between different question \( i \) and \( j \), denoted as $b_{i,j}^k = \operatorname{cos}(h_{(d'_i)}^k, h_{(d'_j)}^k)$ for $1\le k \le K$.

\end{enumerate}

To substantiate our hypothesis, we conduct extensive experiments across multiple models, languages, and datasets. Specifically, we evaluate two widely used models from different model families: Llama-3-8B-Instruct \citep{meta2024introducing} and Qwen2-7B-Instruct \citep{qwen2}. Furthermore, we assess model behavior across five diverse languages (English, Chinese, French, Spanish, and Hindi) and four reasoning-focused datasets: ProofWriter \citep{tafjord2020proofwriter}, FOLIO \citep{han2022folio},  LogicalDeduction \citep{srivastava2022beyond} and GSM8K \citep{cobbe2021gsm8k}.
By systematically analyzing model behavior across these varied conditions, we establish robust empirical evidence for our proposed hypothesis.
The results are shown in Figure \ref{fig_similar}, demonstrating a high degree of consistency across models, languages, and datasets, indicating their robustness and broad applicability. 
To mitigate the impact of experimental randomness, we randomly repeat the experiments 1000 times and take the average of $g_{i}^k, r_{i,j}^k, b_{i,j}^k$ for the random $i,j$'s. 

As shown in Figure \ref{fig_similar}, the green line is the cosine similarity between the first kind of pairs ($g_{i}^k$'s for different layer $k$), the red line is the cosine similarity between the second kind of pairs ($r_{i,j}^k$'s for different layer $k$), and the blue line is the cosine similarity between the third kind of pairs ($b_{i,j}^k$'s for different layer $k$). 
These results reveal a fundamental divergence between the similarity trends of task-relevant and task-irrelevant pairs. The task-relevant similarity (green line) follows a clear trajectory: it rises to form a distinct plateau throughout the middle layers and subsequently declines in the final layers. Conversely, the task-irrelevant similarity (red and blue lines) exhibits an opposite, U-shaped pattern of decreasing and then increasing.
We note that the initial values can be variable, a phenomenon we hypothesize is an artifact of the early hidden states' sensitivity to tokenization differences. Our core claim, however, is based on these starkly opposing trends, which are consistent across languages and strongly substantiate our three-stage hypothesis.

We also investigate the internal representations of these transformer-based models by visualizing hidden states obtained from different language versions of the same dataset, by employing three widely used dimensionality reduction techniques: Principal component analysis (PCA) \citep{hotelling1933analysis}, t-distributed stochastic neighboring embedding (t-SNE) \citep{van2008visualizing}, and uniform manifold approximation and projection (UMAP) \citep{mcinnes2018umap}.
The visualization results are consistent with the similarity results in Figure~\ref{fig_similar}, and we defer them to Appendix. 

\vspace{-0.1cm}
\subsection{Cosine Similarity between Hidden State in Adjacent Layers} \label{adj_Cosine_Similarity}
 \begin{figure*}[htb]
    \centering
    
    \begin{subfigure}{0.49\textwidth}
        \includegraphics[width=\textwidth]{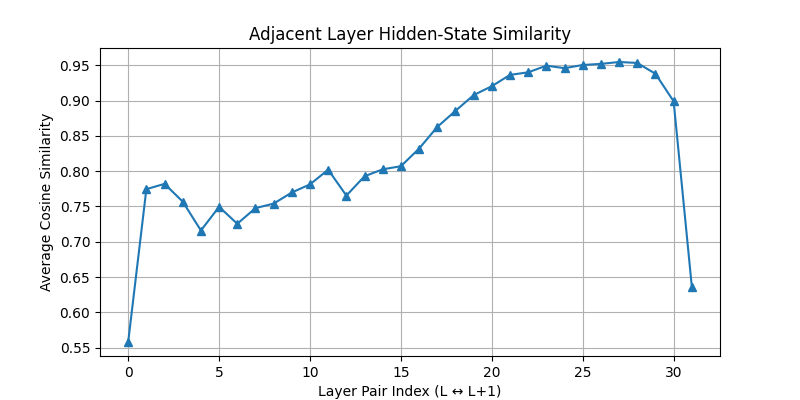}  %
        \caption{LLaMA-3-8B-Instruct on GSM8K    \vspace{0.15in} }\label{fig_similar-2:a}
    \end{subfigure}
    \begin{subfigure}{0.49\textwidth}
        \includegraphics[width=\textwidth]{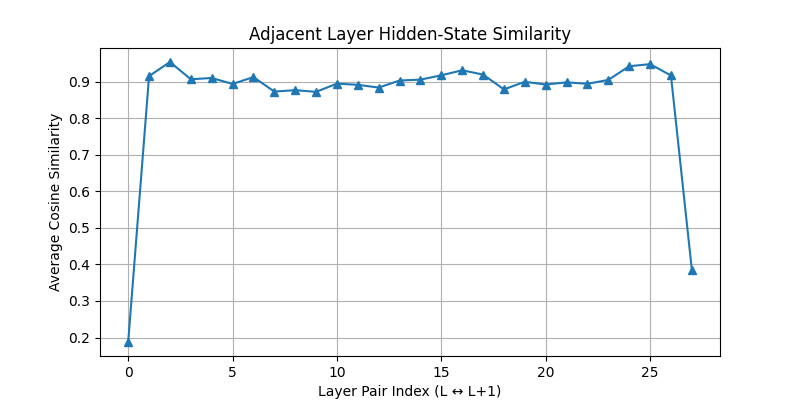}
        \caption{Qwen2-7B-Instruct on LogicalDeduction\vspace{0.15in} }
    \end{subfigure}

    \caption{Cosine Similarity between Hidden Stated in Adjacent Layers}
    \label{fig_similar-2}
\end{figure*}
\vspace{-0.1in}

To achieve a more refined segmentation of the reasoning layers, we adopt the adjacent‐layer cosine similarity analysis similar to \citet{gromov2024unreasonable} and \citet{men2024shortgpt}. For each dataset in Subsection \ref{dataset}, we compute the cosine similarity between every pair of successive layers. Due to the residual architecture of LLMs, a very high similarity between adjacent layers indicates that the latter layer contributes little new information and is therefore highly redundant. Consistent with the findings of \citet{fan2024not}, \citet{gromov2024unreasonable}, and \citet{men2024shortgpt}, from Figure~\ref{fig_similar-2}, we observe that reasoning layers near the textualization stage exhibit similarities exceeding 95\% (e.g., layers 23-28 in Figure~\ref{fig_similar-2:a}. We classify these high-similarity layers as idle layers, while the remaining reasoning layers with lower adjacent-layer similarity are designated as active reasoning layers.

\vspace{-0.15cm}
\subsection{Summary of Experiments}

The observed dynamics of representation similarity systematically align with our proposed functions: an initial shift from linguistic to abstract conceptual encoding (conceptualization), followed by a stage of stable, language-agnostic processing (reasoning), and a final reversion to linguistic formatting for output (textualization). This trajectory strongly validates the functional distinctiveness of each stage. Furthermore, our analysis of layer-wise redundancy reveals the existence of idle reasoning layers where information processing plateaus.

However, acknowledging these distinct stages leads to a critical insight regarding model optimization. While inference requires the integration of all stages, we hypothesize that performance is not limited by total capacity, but rather by a specific functional bottleneck—the stage that constitutes the primary constraint for a given task.
Since different tasks impose varying cognitive demands—some requiring complex logical manipulation while others hinge on precise linguistic understanding—the functional stage that acts as the limiting factor, or bottleneck, may vary across different model-task pairs. 
This implies that there is no static most important layer for all scenarios. Instead, a more effective strategy is to empirically identify the specific functional bottleneck for a given task and target it for refinement. 
This insight lays the groundwork for the methodology proposed in the following section, where we introduce a systematic approach to locate and leverage these critical layers.

\vspace{-2mm}
\section{LIFT: Layer-Informed Fine Tuning}  \label{LIFT}
\vspace{-0.15cm}
Building upon the hypothesis of functional segmentation proposed in the previous section, we introduce a fine-tuning method aimed at achieving higher efficiency and improved performance: Layer-Informed Fine-Tuning (LIFT). 
The LIFT framework consists of two sequential steps: Functional Segmentation and Bottleneck Identification. First, we partition the model into distinct functional stages—conceptualization, active reasoning, idle reasoning, and textualization—based on the analysis of cross-lingual and adjacent-layer representation similarities. Second, we employ a perturbation-based analysis to determine which of these stages acts as the functional bottleneck for a given task. Finally, we apply fine-tuning exclusively to these critical layers.
We denote LIFT[a, b) to represent the fine-tuning process restricted to layers from $a$ to $b$ (excluding layer $b$).

\vspace{-2mm}
\subsection{Layer Partition}
To implement LIFT, it is essential to first identify the location of functional layer segmentation.  We employed a straightforward approach to delineate the layer boundaries. To determine the boundaries between the conceptualization layers and the reasoning layers, we employ the cosine similarity between different language versions of the same question \(g^k\) (as described in Section \ref{sec:hypo}), represented by the green line in Figure \ref{fig_similar}. As previously mentioned in Section \ref{Cosine_Similarity}, the green line reflects task-relevant information, which gradually increases within the conceptualization layers and remains relatively stable in the reasoning layers. Therefore, the inflection point during its growth phase can be considered a marker for the functional transition from conceptualization to reasoning.
For this discrete case, we apply the discrete second derivative method to identify the inflection point \citep{weissteinInflectionPoint}. First, we compute the discrete second derivative of the layer-wise similarity \(g^k\) by
\[
\Delta^2 g^k \;=\; g^{k+1} \;-\; 2\,g^k \;+\; g^{k-1}.
\]
 Second, we identify the first index \( k^* \) where \( \Delta^2 g^k \) changes sign, i.e., where \(\Delta^2 g^{k-1}\,\Delta^2 g^k<0\) as the inflection point marking the transition from conceptualization to reasoning. 
\begin{table*}[h]
  \centering

  \resizebox{\linewidth}{!}{
  \begin{tabular}{|c|c|c|c|c|}
    \hline
    \multicolumn{5}{|c|}{Llama-3-8B-Instruct (Drop)} \\ \hline
    Datasets & conceptualization layers & active reasoning layers & idle reasoning layers & textualization layers \\ \hline
    ProofWriter & 44.04\% ($\pm$2.03\%) & \textbf{42.44\% ($\pm$1.70\%)} & 44.95\% ($\pm$1.54\%) & 44.18\% ($\pm$1.89\%) \\ \hline
    FOLIO & 52.88\% ($\pm$0.97\%) & \textbf{51.33\% ($\pm$0.83\%)} & 52.80\% ($\pm$1.36\%) & 52.48\% ($\pm$1.17\%) \\ \hline
    LogicalDeduction & 47.48\% ($\pm$0.76\%) & \textbf{43.18\% ($\pm$1.45\%)} & 48.86\% ($\pm$1.43\%) & 48.58\% ($\pm$0.89\%) \\ \hline
    gsm8k & 60.30\% ($\pm$1.73\%) & \textbf{58.95\% ($\pm$2.48\%)} & 61.96\% ($\pm$1.00\%) & 61.39\% ($\pm$1.50\%) \\ \hline
  \end{tabular}
  }
  \vspace{0.5cm} %
  \resizebox{\linewidth}{!}{
  \begin{tabular}{|c|c|c|c|c|}
    \hline
    \multicolumn{5}{|c|}{Llama-3-8B-Instruct (Recovery)} \\ \hline
    Datasets & conceptualization layers & active reasoning layers & idle reasoning layers & textualization layers \\ \hline
    ProofWriter & 42.56\% ($\pm$1.76\%) & \textbf{43.99\% ($\pm$1.51\%)} & 42.61\% ($\pm$1.82\%) & 42.33\% ($\pm$1.44\%) \\ \hline
    FOLIO & 52.10\% ($\pm$1.35\%) & \textbf{53.46\% ($\pm$0.97\%)} & 51.20\% ($\pm$0.68\%) & 51.24\% ($\pm$0.93\%) \\ \hline
    LogicalDeduction & 43.78\% ($\pm$1.40\%) & \textbf{47.55\% ($\pm$0.77\%)} & 43.26\% ($\pm$1.41\%) & 43.50\% ($\pm$1.50\%) \\ \hline
    gsm8k & \textbf{63.43\% ($\pm$2.47\%)} & 60.94\% ($\pm$1.53\%) & 62.16\% ($\pm$2.57\%) & 63.30\% ($\pm$1.36\%) \\ \hline
  \end{tabular}
  }
  \vspace{-0.1cm}
  \caption{Performance of quantization probing (Drop and Recovery) over Various Layer Ranges of Llama-3-8B-Instruct (Averaged over 8 Seeds).}
  \label{tab:ablation_combined}
\end{table*}
Next, to partition the reasoning block into active layers and idle layers, we utilize the cosine similarity between hidden states in adjacent layers described in Section \ref{adj_Cosine_Similarity}. A higher cosine similarity between adjacent layers indicates greater redundancy. 
Therefore, we set a threshold value of $\beta$ and classify consecutive layers exceeding this threshold as idle reasoning layers. We select $\beta = 0.94$ based on empirical observations across our benchmark tasks: 
We consistently observed that the adjacent layer similarity in the idle stage saturates at a peak level of approximately 0.95, and thus chose 0.94 as a robust boundary since this value is slightly lower than the saturation peak and ensures that we capture the high-redundancy plateau without encroaching on the active reasoning layers.
Consequently, if such a high-similarity region exists, the boundary between the idle reasoning stage and the textualization stage is naturally defined by the point where the similarity drops back below $\beta$. All layers succeeding this idle block are then categorized as textualization layers. Conversely, if no layers surpass the threshold $\beta$---as we observed in certain datasets for Qwen2-7B-Instruct---we conclude that the model does not exhibit an idle reasoning phase for that specific task. In such cases, the active reasoning block transitions directly into the textualization layers.
Based on these methods, we can obtain the specific functional layer segmentation for different tasks across various models. Due to space limitations, the detailed segmentation results are provided in the Appendix.

\vspace{-0.1cm}
\subsection{Identifying Functionally Critical Layers}

The inference process of an LLM can be viewed as a sequential dependency chain: input is first conceptualized, then reasoned over, and finally textualized. Analogous to the ``Liebig’s Law of the Minimum," \citep{von1841organic} the overall performance of the model is often constrained by its least effective functional component—or conversely, performance is maximized when the most ``load-bearing" component is optimized. Therefore, after establishing the boundaries between conceptualization, reasoning, and textualization, the next imperative is %
to identify and target the specific functional block that acts as the primary determinant of task success.

To empirically isolate these critical layers, we employ a dual-faceted sensitivity analysis using weight-only quantization (specifically AWQ \citep{lin2023awq}). We hypothesize that the layers most vital to a task’s functionality will exhibit the highest sensitivity to precision loss and the greatest capacity for performance restoration. We design two complementary diagnostic tests:

\begin{enumerate}[leftmargin=*, itemsep=3pt, parsep=0pt, topsep=2pt]
    \item Degradation Analysis (Drop): We selectively quantize a specific functional block  to low precision (INT4) while maintaining the remainder of the model in high precision (BF16). The magnitude of the performance decline indicates the model's reliance on that specific block; a larger drop suggests a critical bottleneck.
    \item Recovery Analysis (Recovery): This acts as the inverse of the Degradation Analysis. Here, we maintain a specific functional block in high precision (BF16) while quantizing the rest of the model to low precision (INT4). The extent to which performance is recovered relative to a fully quantized model highlights the functional importance of that block.
\end{enumerate}

\begin{table*}[htbp]
  \centering
  \renewcommand{\arraystretch}{1.2}
  \setlength{\tabcolsep}{4pt} %
  \resizebox{\linewidth}{!}{
  \begin{tabular}{c|cccc|cccc}
    \toprule
    & \multicolumn{4}{c|}{Llama-3-8B-Instruct} 
    & \multicolumn{4}{c}{Qwen2-7B-Instruct} \\
    \midrule
    ProofWriter & Base model & FullFT & AdaLoRA & LIFT[2,20) & Base model & FullFT & AdaLoRA & LIFT[3,26) \\
    \midrule
    Accuracy & 45.67\% & 56.12\% (±2.02\%) & 47.83\% (±0.49\%) & \textbf{57.67\% (±0.36\%)} & 48.00\% & 53.17\% (±3.10\%) & 48.25\% (±1.68\%) & \textbf{53.46\% (±2.36\%)} \\
    Training Time(s) & - & 1383 & 1923 & \textbf{1215} & - & 1288 & 1839 & \textbf{1180} \\
    \midrule
    FOLIO & Base model & FullFT & AdaLoRA & LIFT[2,20) & Base model & FullFT & AdaLoRA & LIFT[3,26) \\
    \midrule
    Accuracy & 55.39\% & 52.08\% (±1.67\%) & 57.11\% (±2.21\%) & \textbf{58.21\% (±2.35\%)} & 34.31\% & \textbf{44.98\% (±0.74\%)} & 18.01\% (±1.57\%) & 44.73\% (±1.98\%) \\
    Training Time(s) & - & 332 & 516 & \textbf{290} & - & 310 & 496 & \textbf{284} \\
    \midrule
    LogicalDeduction & Base model & FullFT & AdaLoRA & LIFT[2,22) & Base model & FullFT & AdaLoRA & LIFT[3,24) \\
    \midrule
    Accuracy & 46.33\% & 53.25\% (±1.89\%) & 49.42\% (±2.27\%) & \textbf{53.58\% (±2.47\%)} & 50.33\% & 53.83\% (±1.23\%) & 51.42\% (±1.17\%) & \textbf{57.08\% (±3.29\%)} \\
    Training Time(s) & - & 516 & 737 & \textbf{460} & - & 477 & 680 & \textbf{428} \\
    \midrule
    GSM8K  & Base model & FullFT & AdaLoRA & LIFT[0,2) & Base model & FullFT & AdaLoRA & LIFT[3,26) \\
    \midrule
    Accuracy & 49.96\% & 67.76\% (±0.57\%) & 72.35\% (±1.07\%) & \textbf{73.96\% (±0.98\%)} & \textbf{78.85\%} & 74.53\% (±0.34\%) & 77.50\% (±0.41\%) & 75.78\% (±0.24\%) \\
    Training Time(s) & - & 3347 & 4611 & \textbf{2667} & - & 3500 & 4832 & \textbf{3208} \\
    \bottomrule
  \end{tabular}
  }
  \caption{Performance and Training Time Comparison of Fine-Tuning Methods over 4 Seeds}
  \label{tab:Performance_LoRA}
\end{table*}

As shown in Table \ref{tab:ablation_combined}, our experimental results generally show a strong consistency between these two metrics. For each task, the functional block causing the greatest performance degradation when quantized is typically the same block that yields the highest performance recovery when preserved.

In rare instances where the two metrics diverge—likely due to the stochastic nature of the generation process—we select the target layers based on a composite importance score. We define the optimal functional block as the one that maximizes the differential between the accuracy in the Recovery setting and the Degradation setting ($Acc_{\text{recovery}} - Acc_{\text{drop}}$). This approach ensures we select the layers that offer the highest robustness and contribution to the task.

Applying this methodology to Llama-3-8B-Instruct (Table \ref{tab:ablation_combined}), we observe distinct functional dependencies. For the ProofWriter, FOLIO, and LogicalDeduction datasets, the active reasoning layers are identified as the most critical. This aligns with the intuition that, for this specific model architecture, these tasks necessitate intensive manipulation of abstract logic. Conversely, for GSM8K, the conceptualization layers exhibit the highest sensitivity and recovery potential, suggesting that the initial translation of the problem statement into an abstract representation is the dominant bottleneck for Llama-3-8B-Instruct on mathematical tasks.

Crucially, however, we do not posit these findings as universal laws for these datasets. As indicated by our broader analysis (see Appendix), the location of the functional bottleneck is intrinsic to the model-task pairing. Different LLMs may allocate computational effort differently for the same task—one model may struggle with conceptualization while another is limited by reasoning capacity. This heterogeneity underscores the necessity of our diagnostic approach, as it dynamically identifies the specific limiting layers for each unique model-task combination rather than relying on static, dataset-level assumptions. 

Due to space constraints, results for Qwen2-7B-Instruct are detailed in the Appendix.

\vspace{-2mm}
\subsection{Finetuning Setting}

We employ the LoRA framework to accelerate fine-tuning. For our LIFT method, we apply LoRA exclusively to the identified bottleneck layers, whereas the FullFT baseline applies LoRA across all transformer layers. To rigorously evaluate parameter efficiency, we compare against AdaLoRA \citep{zhang2023adaptive}, a dynamic-rank baseline. We strictly align the total trainable parameter budgets between the two methods by dynamically constraining AdaLoRA's target rank based on the specific number of layers LIFT updates for a given task. Detailed budget alignment configurations are provided in Appendix I.

\vspace{-0.1cm}
\subsection{Experimental Results}  
Table \ref{tab:Performance_LoRA} presents the performance and training time of the original base models, full Transformer layer fine-tuning (FullFT), the dynamic-rank baseline (AdaLoRA), and our proposed LIFT method. The results demonstrate the superiority of LIFT in both effectiveness and computational efficiency.
We observe that the LIFT approach demonstrates a noticeable improvement in performance compared to FullFT, while achieving an average reduction of 11.4\% in time cost. This efficiency gain stems from LIFT updating drastically fewer trainable parameters (e.g., only 10.5M vs. 167.8M for Llama-3 on GSM8K), which substantially reduces the memory footprint. Furthermore, when compared to AdaLoRA under a strictly aligned parameter budget, LIFT demonstrates superior stability and effectiveness. While AdaLoRA dynamically allocates parameters across all layers, it suffers from massive computational overhead due to its iterative singular value decomposition (SVD) and pruning mechanisms, resulting in the longest training times across all evaluated settings. In contrast, LIFT not only trains significantly faster but also achieves higher accuracy on the majority of logical reasoning tasks, notably preventing catastrophic performance drops such as AdaLoRA's collapse to 18.01\% on the Qwen2 FOLIO benchmark. These experimental results indicate that by selectively fine-tuning only those layers most relevant to a given task’s functionality, LIFT can simultaneously enhance model accuracy, reduce training time, and lower memory requirements.
Additional ablations that fine-tune alternative functional regions are reported in Appendix J and further support the benefit of targeting the LIFT-selected layers.
\vspace{-0.15cm}
\section{Conclusion}
\vspace{-0.15cm}

In this work, we proposed a three-stage functional view of LLMs, consisting of conceptualization, reasoning, and textualization stages, with reasoning further divided into active and idle layers. Based on this structure, we introduced a quantization-based diagnostic to identify task-specific functional bottlenecks and proposed LIFT to selectively fine-tune the corresponding layers. Experiments across two model families and four reasoning datasets show that LIFT can improve task performance while reducing fine-tuning cost. Our results suggest that task-specific layer structure can provide a useful basis for efficient model adaptation.

\section*{AI Use Statement}
In this work, we used GPT-4 to translate the original English benchmark datasets into Chinese, French, Spanish, and Hindi for our multilingual analysis. We additionally used generative AI tools to edit the manuscript for language and readability. We take full responsibility for the final content of the paper and all AI-assisted artifacts.

\section*{Reproducibility Statement}
We have made every effort to ensure the reproducibility of our research. All models used in this study (Llama-3-8B-Instruct, Qwen2-7B-Instruct) are publicly available, as described in Section 3.1. The datasets are standard benchmarks, detailed in Section 3.2, with further examples provided in Appendix B. Our multilingual datasets and the code used for our cosine similarity analysis are available in the supplementary materials, as noted in Appendix B.3. The methodology for identifying functional layer boundaries is detailed in Section 5.1 and Appendix E. Furthermore, Appendix H provides a comprehensive description of our experimental environment, including hardware specifications, software versions (PyTorch, CUDA), and the complete set of hyperparameters used for LoRA fine-tuning. The supplementary materials also contain the necessary code to replicate our main experiments and generate the figures presented in the paper.
\section*{Ethics Statement}
This paper presents work whose goal is to advance the field of machine learning, specifically focusing on the mechanistic interpretability of Large Language Models (LLMs) and parameter-efficient fine-tuning.

Our proposed method, LIFT, significantly reduces the computational resources required for model adaptation by targeting functionally critical layers. This improvement in training efficiency contributes to the broader goal of Green AI, helping to lower the energy consumption and carbon footprint associated with LLM deployment. Furthermore, by identifying the functional segmentation of LLMs, our work provides insights into the internal mechanisms of these models. We believe that such improvements in interpretability are essential for developing safer, more transparent, and more trustworthy AI systems.

Regarding the datasets used in our experiments (ProofWriter, FOLIO, LogicalDeduction, and GSM8K), they are established public benchmarks containing logical and mathematical problems. They do not contain any personally identifiable information or sensitive content. We do not foresee any direct negative societal consequences from this specific research.

\bibliography{iclr2027_conference}
\bibliographystyle{iclr2027_conference}

\appendix
\newpage
\section{LLM structure}
\begin{table}[htbp]
    \centering
    \renewcommand{\arraystretch}{1.2}
    \setlength{\tabcolsep}{8pt}
    \begin{tabular}{c|c|c|c}
        \toprule
        Model & Layers & Model & Layers \\
        \midrule

        Llama-3-8B-Instruct & 32 & Qwen2-7B-Instruct & 28 \\
        \bottomrule
    \end{tabular}
    \caption{Number of layers in each LLM}
    \label{tab:num_layers}
\end{table}

\section{Datasets}

\subsection{Dataset Details}
\begin{itemize}[leftmargin=*, itemsep=3pt, parsep=0pt, topsep=2pt]
    \item \textbf{ProofWriter} \citep{tafjord2020proofwriter} is a widely used dataset for deductive logical reasoning. In our study, we utilize the open-world assumption subset. This dataset is categorized into five classes based on the depth of reasoning required. To ensure a rigorous evaluation, we select the most challenging subset, depth-5.

    \item \textbf{FOLIO} \citep{han2022folio} is a challenging dataset for first-order logic reasoning, featuring expert-written problems that require complex logical deductions. The questions are presented in natural language and are closely aligned with real-world knowledge.

    \item \textbf{LogicalDeduction}  is a logical reasoning task from the BigBench benchmark \citep{srivastava2022beyond}, focusing on constraint satisfaction problems. The task primarily involves deducing the order of a sequence of objects given a minimal set of conditions.

    \item \textbf{GSM8K} \citep{cobbe2021gsm8k}: This dataset contains 8,500 grade school-level math problems that cover a wide range of topics, including arithmetic, algebra, and geometry. 

\end{itemize}

\subsection{Examples of Each Dataset}

\subsubsection{ProofWriter}
\lstset{
    basicstyle=\ttfamily\small,   %
    breaklines=true,              %
    frame=single,                 %
    numberstyle=\tiny,            %
    xleftmargin=2em,              %
    framexleftmargin=1.5em        %
}
\begin{lstlisting}
"instruction": "The cow is blue. The cow is round. The cow likes the lion. The cow visits the tiger. The lion is cold. The lion is nice. The lion likes the squirrel. The squirrel is round. The squirrel sees the lion. The squirrel visits the cow. The tiger likes the cow. The tiger likes the squirrel. If something is cold then it visits the tiger. If something visits the tiger then it is nice. If something sees the tiger and it is young then it is blue. If something is nice then it sees the tiger. If something likes the squirrel and it likes the cow then it visits the tiger. If something is nice and it sees the tiger then it is young. If the cow is cold and the cow visits the lion then the lion sees the squirrel.Based on the above information, is the following statement true, false, or unknown? The tiger is not young.options: ['A) True', 'B) False', 'C) Unknown']",

"input": "",

"output": "B"
\end{lstlisting}

\subsubsection{FOLIO}
\begin{lstlisting}
"instruction": "All people who regularly drink coffee are dependent on caffeine. People either regularly drink coffee or joke about being addicted to caffeine. No one who jokes about being addicted to caffeine is unaware that caffeine is a drug. Rina is either a student and unaware that caffeine is a drug, or neither a student nor unaware that caffeine is a drug. If Rina is not a person dependent on caffeine and a student, then Rina is either a person dependent on caffeine and a student, or neither a person dependent on caffeine nor a student.Based on the above information, is the following statement true, false, or uncertain? Rina is a person who jokes about being addicted to caffeine or unaware that caffeine is a drug.options: ['A) True', 'B) False', 'C) Uncertain']",

"input": "",

"output": "A"

\end{lstlisting}
\subsubsection{LogicalDeduction}
\begin{lstlisting}
"instruction": "The following paragraphs each describe a set of five objects arranged in a fixed order. The statements are logically consistent within each paragraph.On a branch, there are five birds: a quail, an owl, a raven, a falcon, and a robin. The owl is the leftmost. The robin is to the left of the raven. The quail is the rightmost. The raven is the third from the left.Which of the following is true?options: ['A) The quail is the rightmost.', 'B) The owl is the rightmost.', 'C) The raven is the rightmost.', 'D) The falcon is the rightmost.', 'E) The robin is the rightmost.']",

"input": "",

"output": "A"

\end{lstlisting}
\subsubsection{GSM8K}

\begin{lstlisting}
"instruction": "Natalia sold clips to 48 of her friends in April, and then she sold half as many clips in May. How many clips did Natalia sell altogether in April and May?",
"input": "",
"output": "Natalia sold 48/2 = <<48/2=24>>24 clips in May.\nNatalia sold 48+24 = <<48+24=72>>72 clips altogether in April and May.\n#### 72"
\end{lstlisting}

\subsection{Multilingual Dataset and Availability}
To facilitate our multilingual analysis, the original English datasets were translated using the GPT-4 language model. This process generated versions of the datasets in four additional languages: Chinese, French, Spanish, and Hindi. All translated data used in this study is publicly available for review in the supplementary material, located at the following path: Codes/Cosine Similarity/
\section{Details of Cosine similarities of Hidden States}
To make the Cosine Similarity more pronounced, we adjusted the order of the questions in the dataset, presenting the options first followed by the question statement.
We present here the complete set of cosine similarities of hidden states from Section 4. 

\begin{figure}[htbp]
    \centering
    \begin{subfigure}[b]{0.24\textwidth}
        \centering
        \includegraphics[width=\textwidth]{similarity/Similarity/Llama-3_Chinese_ProofWriter_train_modified_similarity.png}
        \caption{Chinese - ProofWriter}
    \end{subfigure}
    \begin{subfigure}[b]{0.24\textwidth}
        \centering
        \includegraphics[width=\textwidth]{similarity/Similarity/Llama-3_Chinese_FOLIO_train_modified_similarity.png}
        \caption{Chinese - FOLIO}
    \end{subfigure}
    \begin{subfigure}[b]{0.24\textwidth}
        \centering
        \includegraphics[width=\textwidth]{similarity/Similarity/Llama-3_Chinese_gsm8k_alpaca_train_similarity.png}
        \caption{Chinese GSM8K}
    \end{subfigure}
    \begin{subfigure}[b]{0.24\textwidth}
        \centering
        \includegraphics[width=\textwidth]{similarity/Similarity/Llama-3_Chinese_LogicalDeduction_train_modified_similarity.png}
        \caption{Chinese-LogicalDeduction}
    \end{subfigure}

    \begin{subfigure}[b]{0.24\textwidth}
        \centering
        \includegraphics[width=\textwidth]{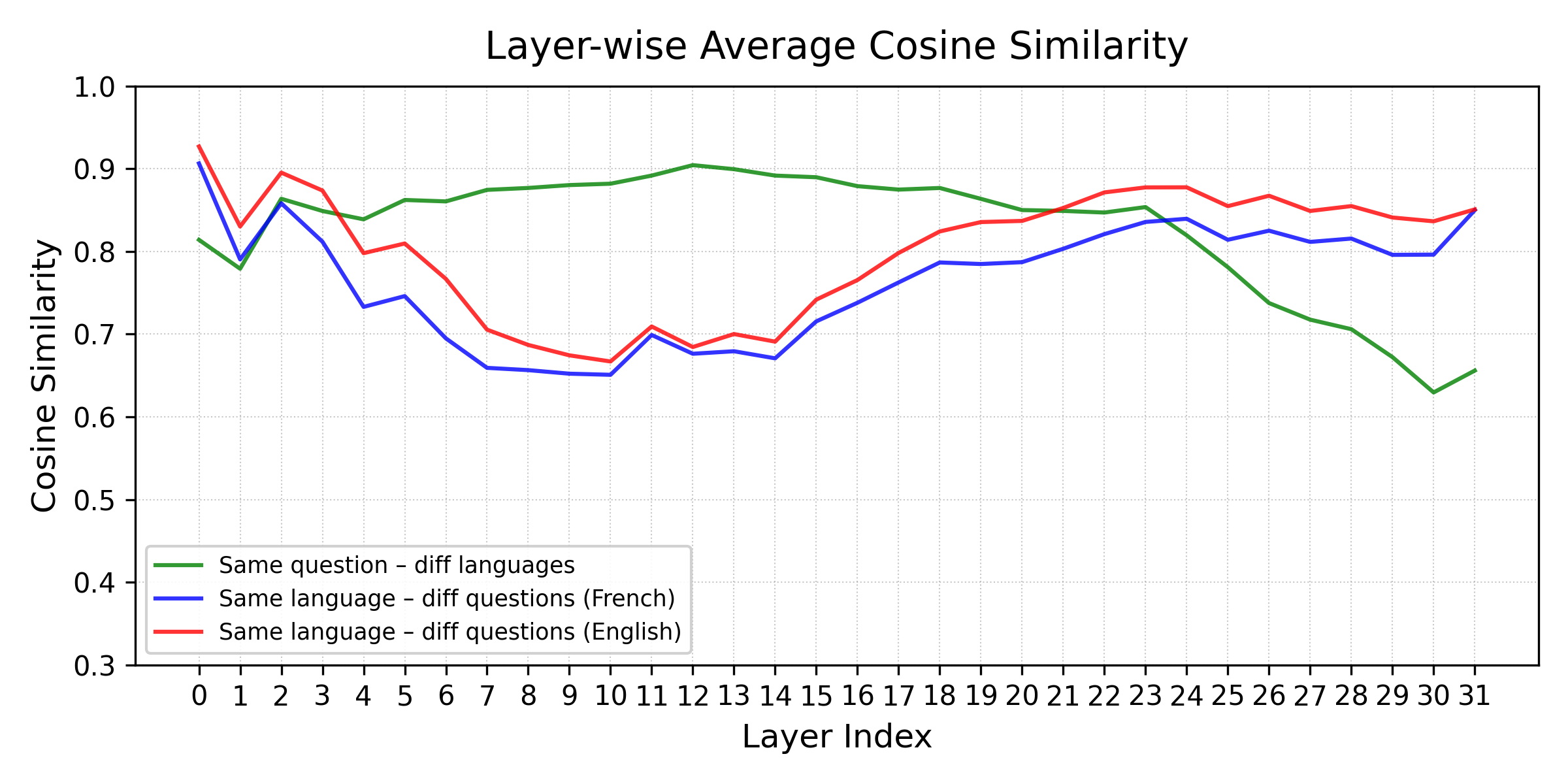}
        \caption{French - ProofWriter}
    \end{subfigure}
    \begin{subfigure}[b]{0.24\textwidth}
        \centering
        \includegraphics[width=\textwidth]{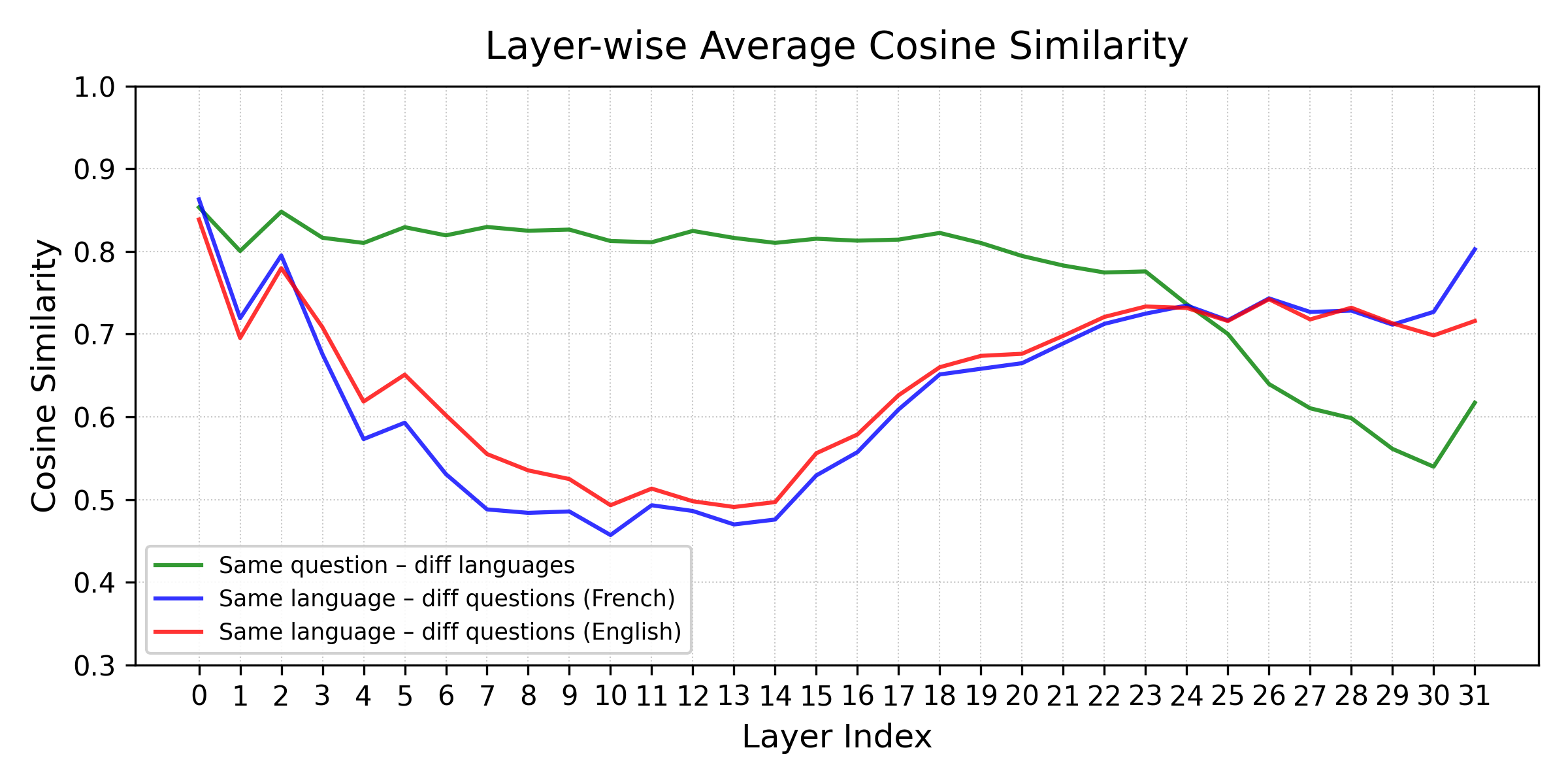}
        \caption{French - FOLIO}
    \end{subfigure}
    \begin{subfigure}[b]{0.24\textwidth}
        \centering
        \includegraphics[width=\textwidth]{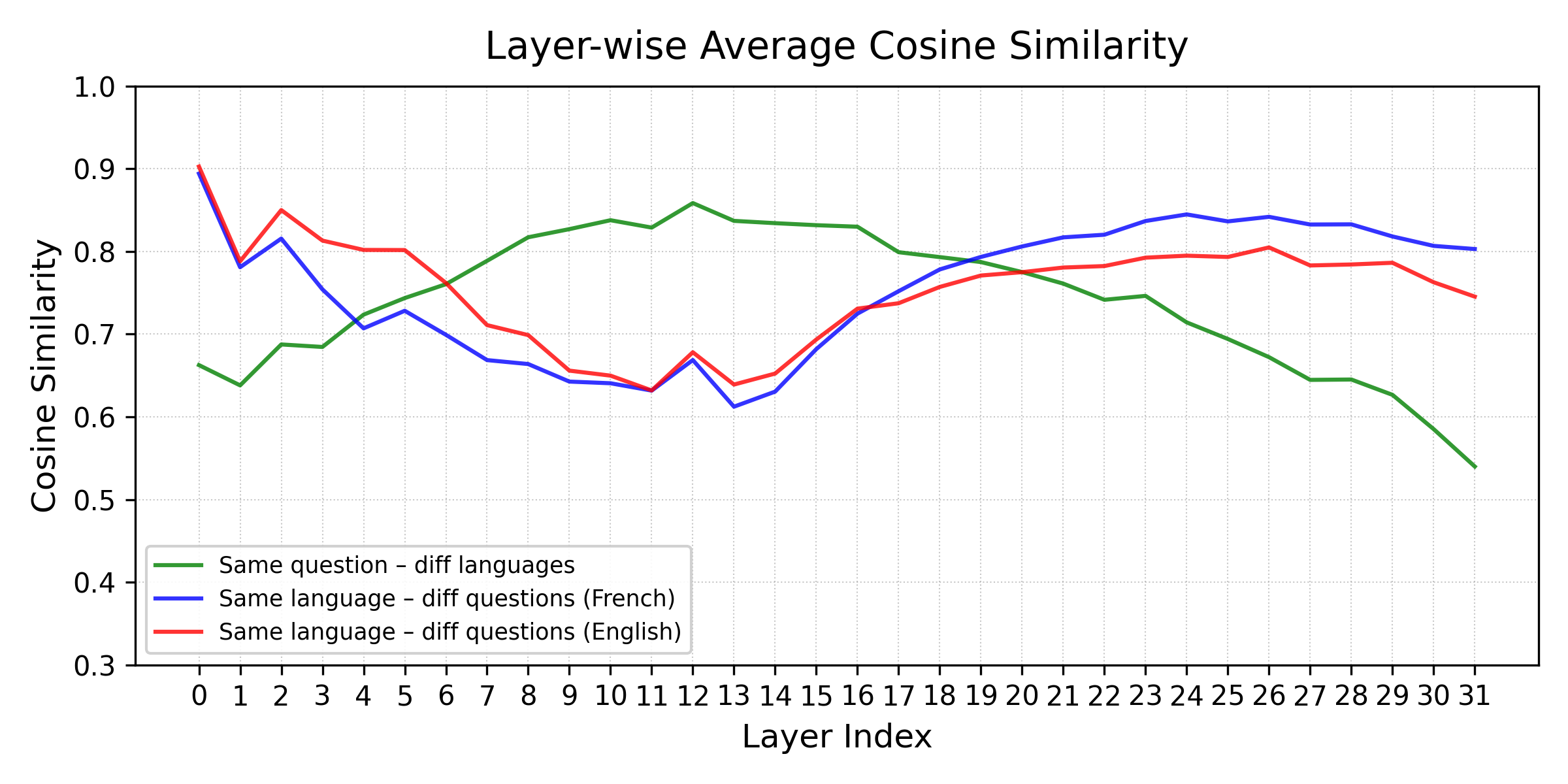}
        \caption{French GSM8K}
    \end{subfigure}
    \begin{subfigure}[b]{0.24\textwidth}
        \centering
        \includegraphics[width=\textwidth]{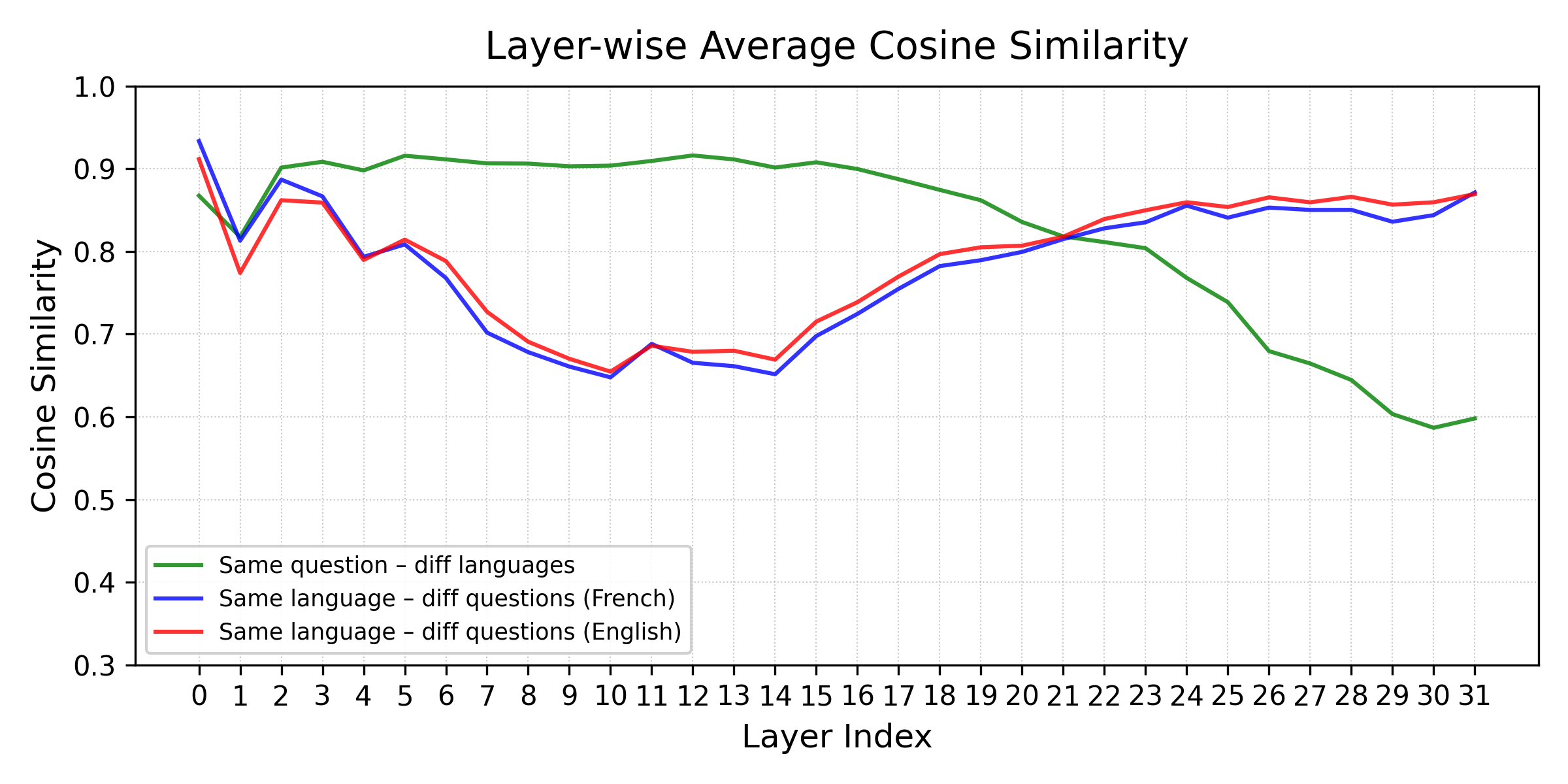}
        \caption{French - LogicalDeduction}
    \end{subfigure}

    \begin{subfigure}[b]{0.24\textwidth}
        \centering
        \includegraphics[width=\textwidth]{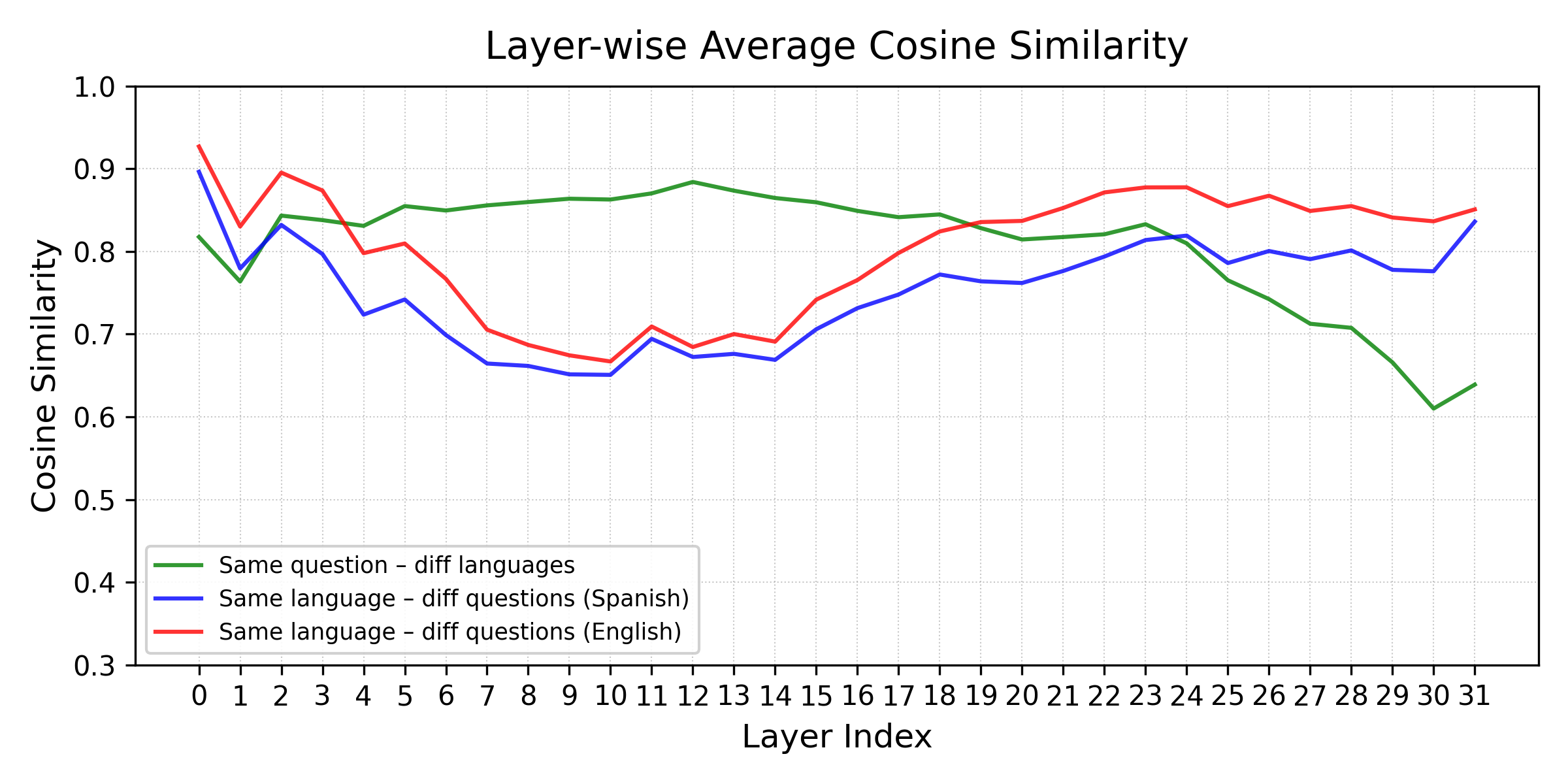}
        \caption{Spanish - ProofWriter}
    \end{subfigure}
    \begin{subfigure}[b]{0.24\textwidth}
        \centering
        \includegraphics[width=\textwidth]{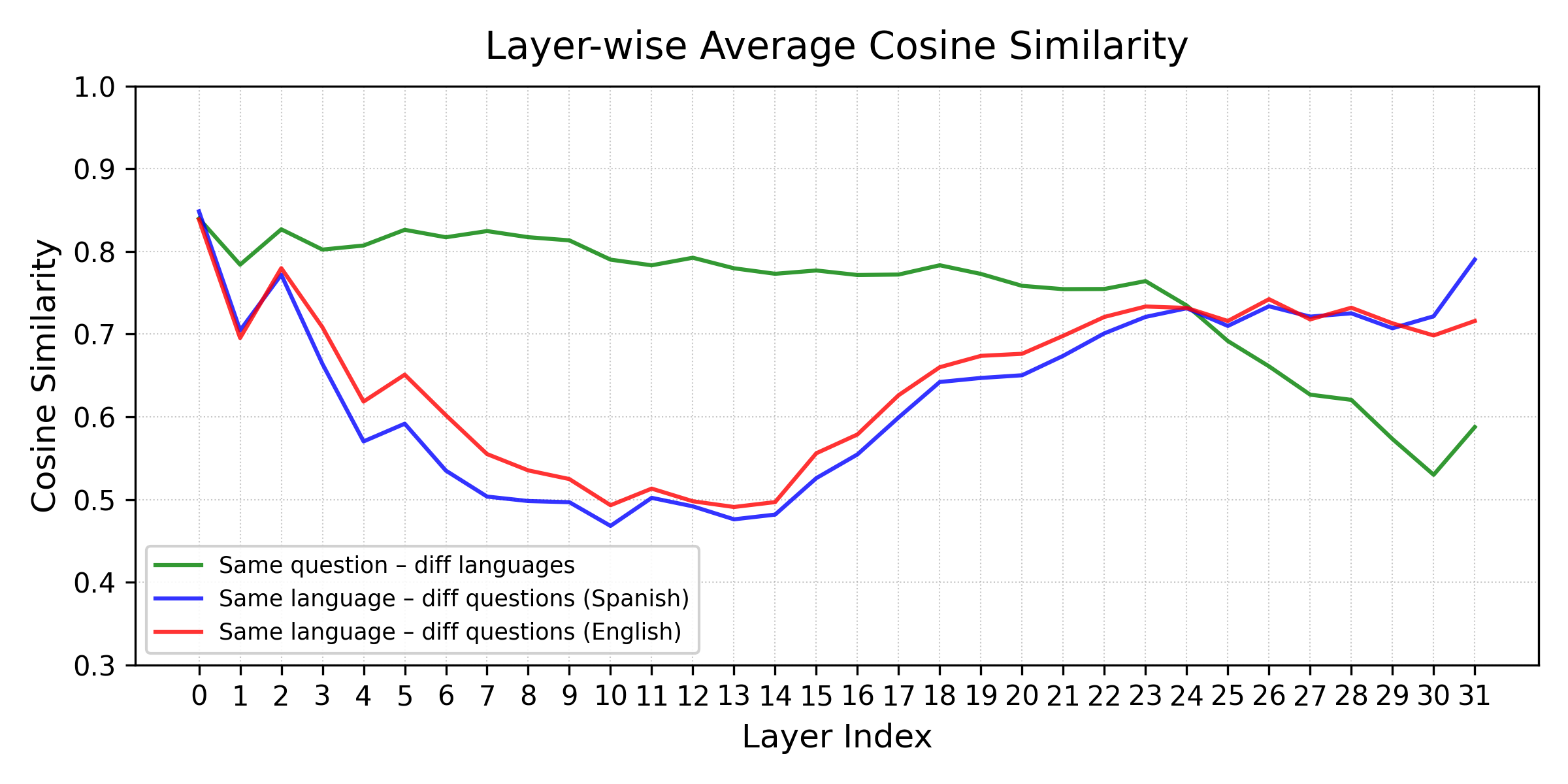}
        \caption{Spanish - FOLIO}
    \end{subfigure}
    \begin{subfigure}[b]{0.24\textwidth}
        \centering
        \includegraphics[width=\textwidth]{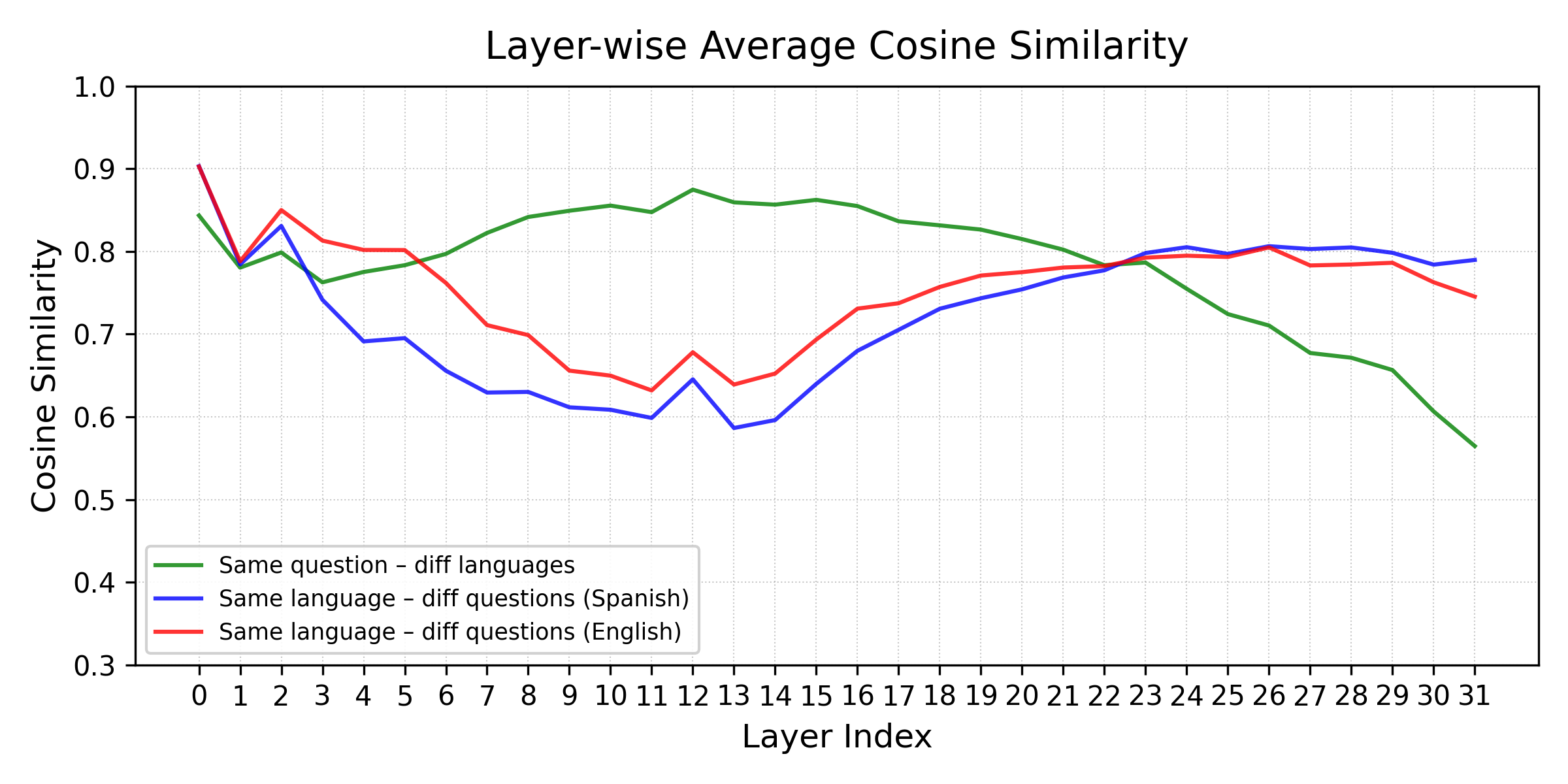}
        \caption{Spanish GSM8K}
    \end{subfigure}
    \begin{subfigure}[b]{0.24\textwidth}
        \centering
        \includegraphics[width=\textwidth]{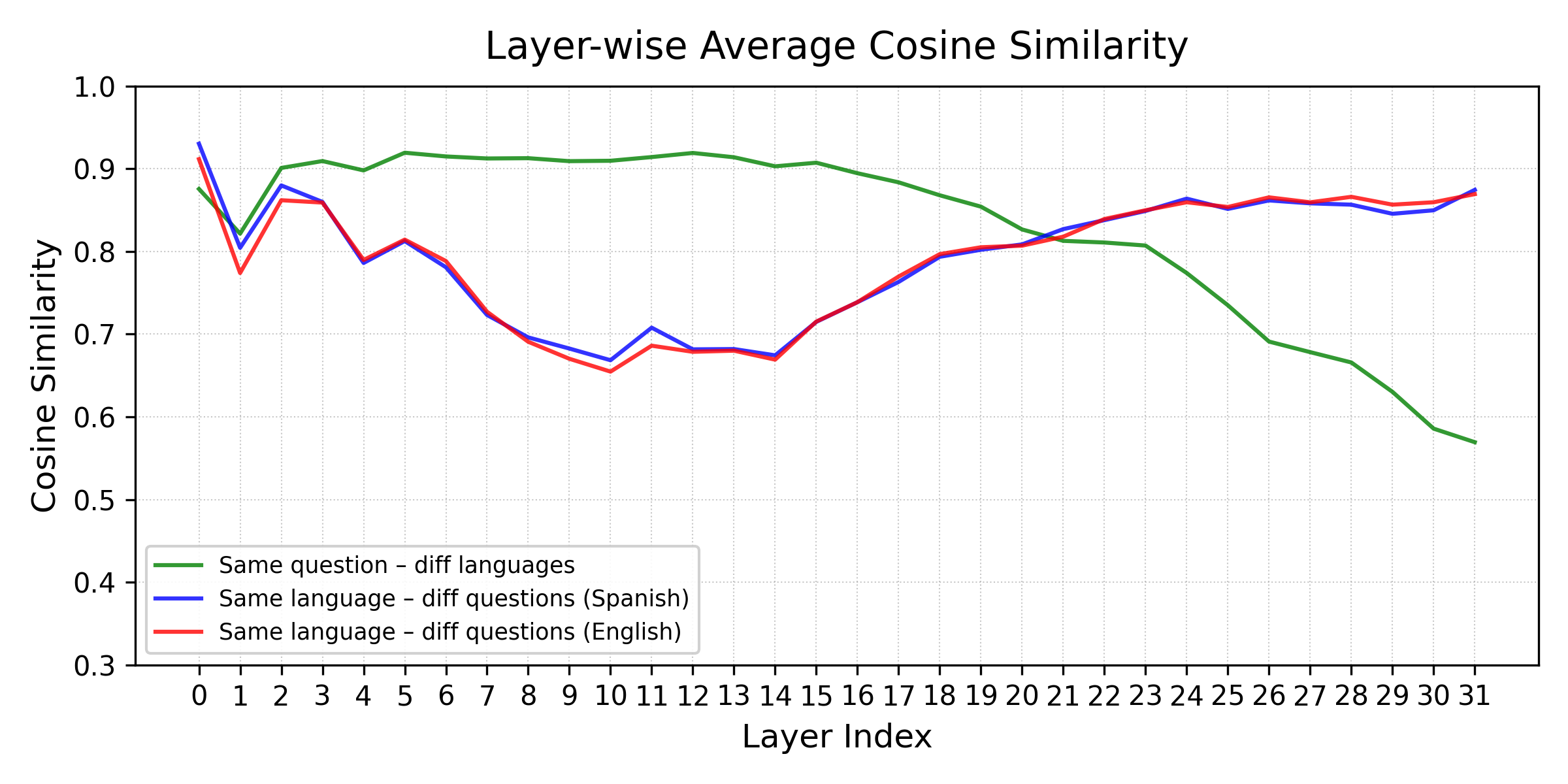}
        \caption{Spanish - LogicalDeduction}
    \end{subfigure}

    \begin{subfigure}[b]{0.24\textwidth}
        \centering
        \includegraphics[width=\textwidth]{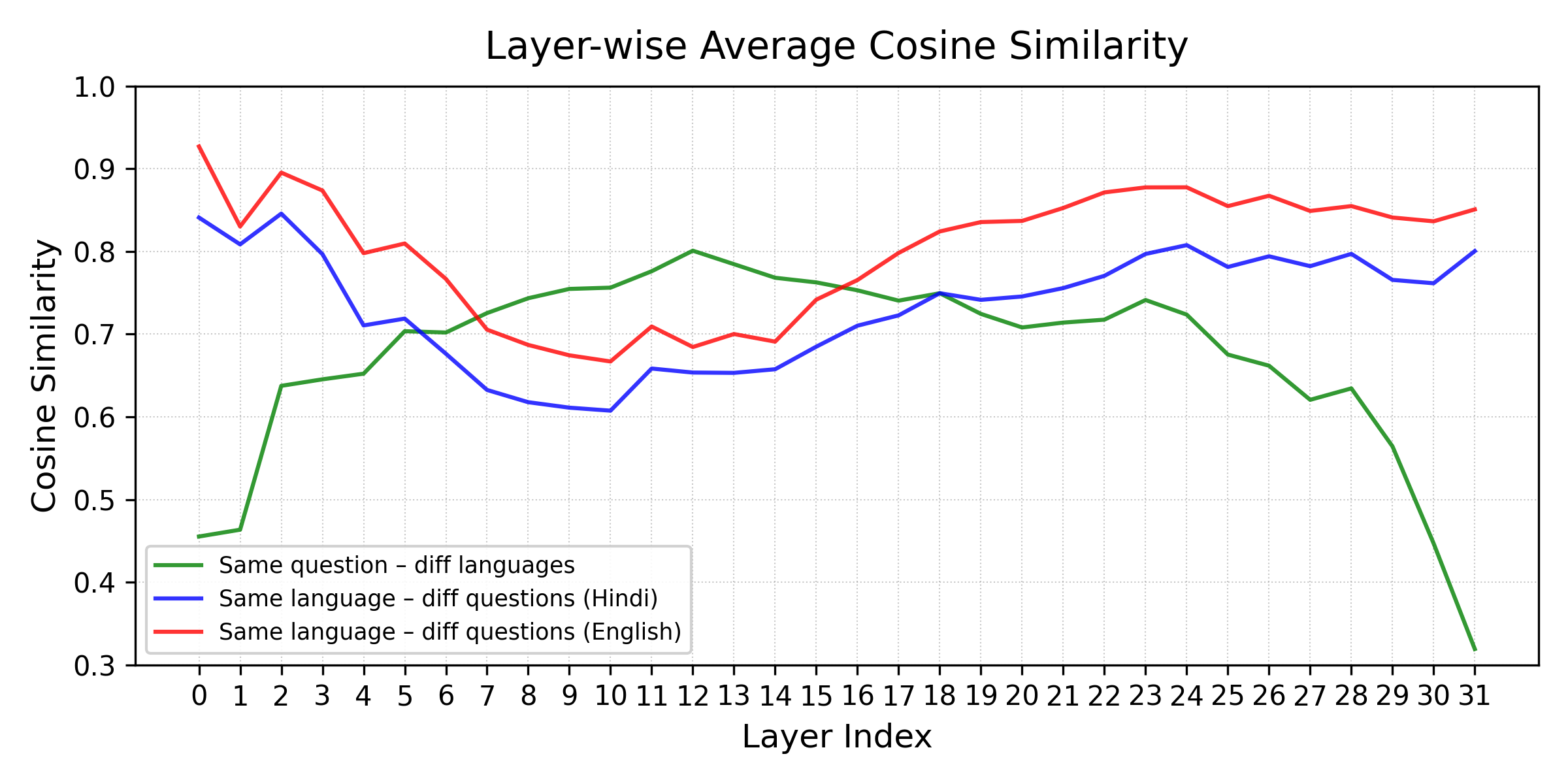}
        \caption{Hindi - ProofWriter}
    \end{subfigure}
    \begin{subfigure}[b]{0.24\textwidth}
        \centering
        \includegraphics[width=\textwidth]{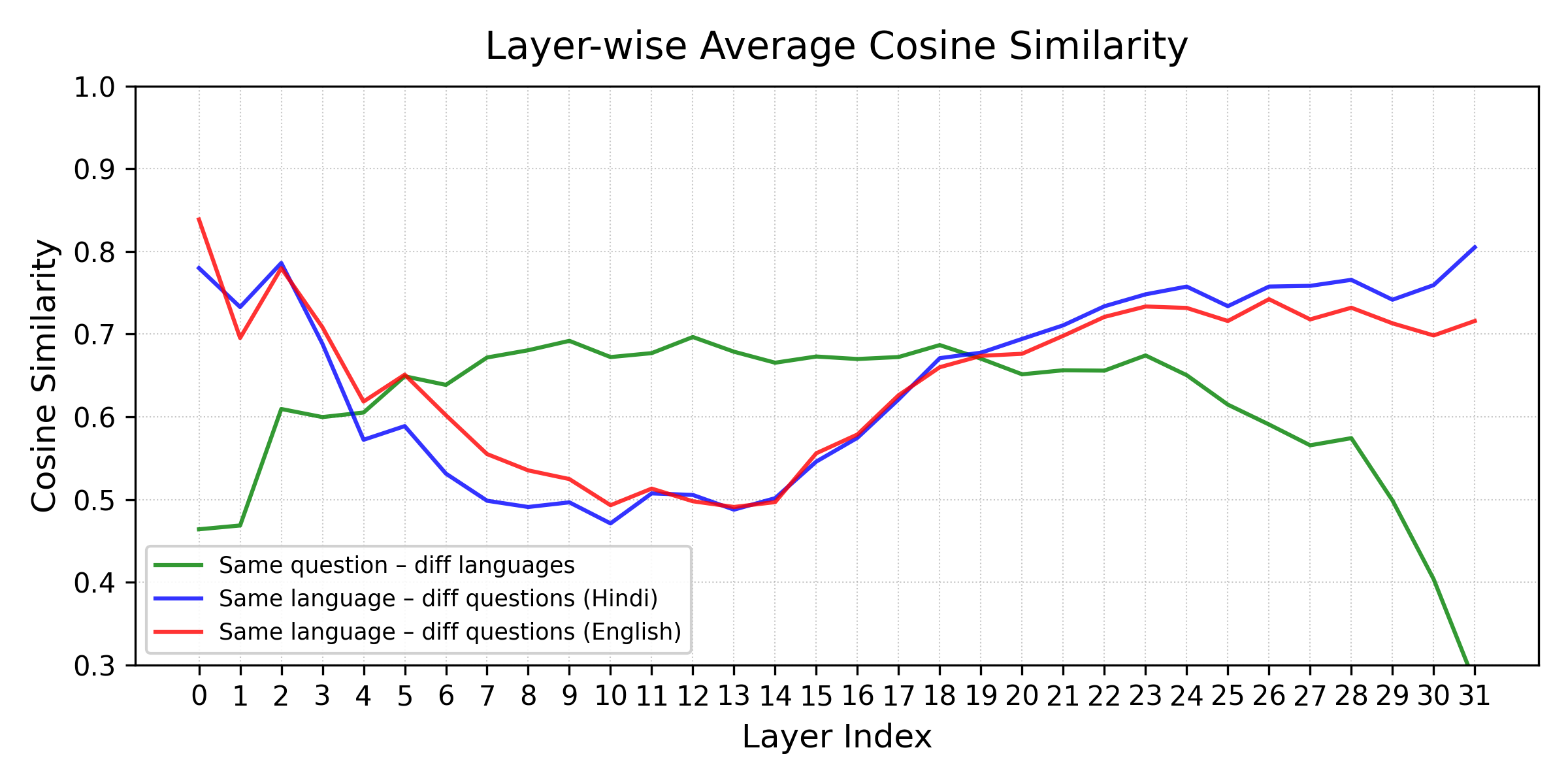}
        \caption{Hindi - FOLIO}
    \end{subfigure}
    \begin{subfigure}[b]{0.24\textwidth}
        \centering
        \includegraphics[width=\textwidth]{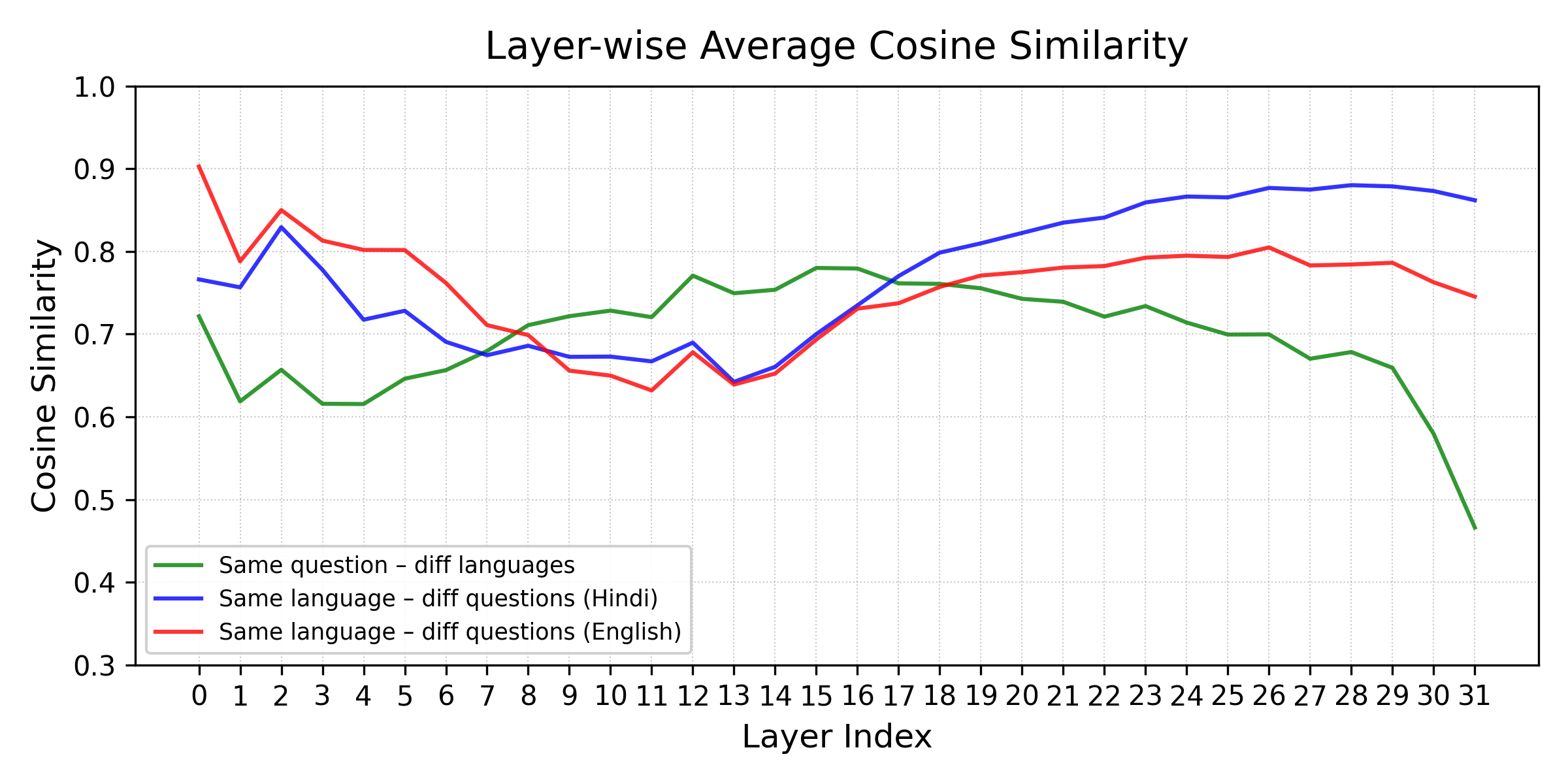}
        \caption{Hindi GSM8K}
    \end{subfigure}
    \begin{subfigure}[b]{0.24\textwidth}
        \centering
        \includegraphics[width=\textwidth]{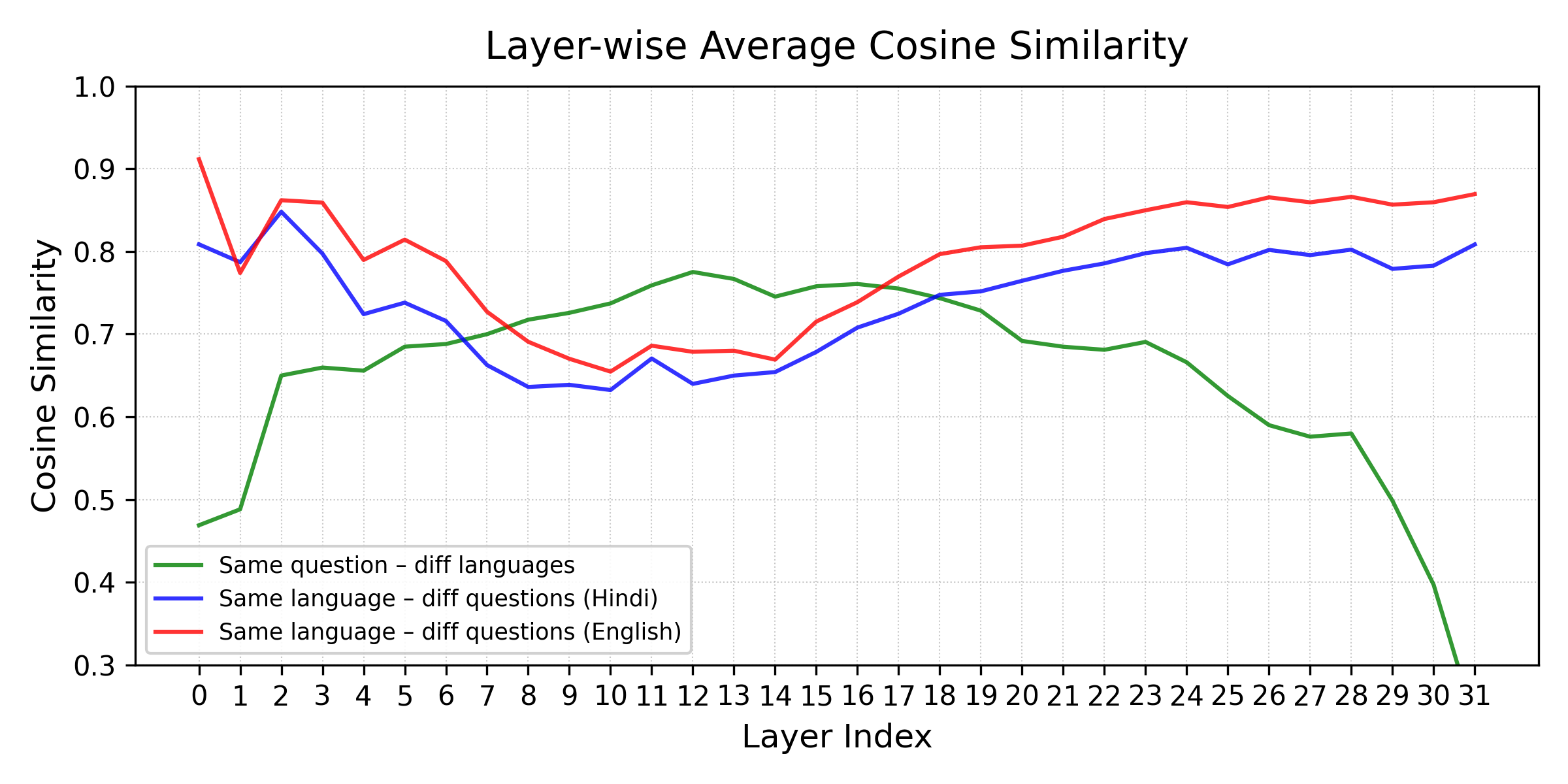}
        \caption{Hindi - LogicalDeduction}
    \end{subfigure}

    \caption{Performance of Llama-3-8B-Instruct across languages and datasets.}
    \label{fig:Llama3_results}
\end{figure}

\begin{figure}[htbp]
    \centering
    \begin{subfigure}[b]{0.24\textwidth}
        \centering
        \includegraphics[width=\textwidth]{similarity/Similarity/Qwen_Qwen2-7B-Instruct_Chinese_ProofWriter_train_modified_similarity.png}
        \caption{Chinese - ProofWriter}
    \end{subfigure}
    \begin{subfigure}[b]{0.24\textwidth}
        \centering
        \includegraphics[width=\textwidth]{similarity/Similarity/Qwen_Qwen2-7B-Instruct_Chinese_FOLIO_train_modified_similarity.png}
        \caption{Chinese - FOLIO}
    \end{subfigure}
    \begin{subfigure}[b]{0.24\textwidth}
        \centering
        \includegraphics[width=\textwidth]{similarity/Similarity/Qwen_Qwen2-7B-Instruct_Chinese_gsm8k_alpaca_train_similarity.png}
        \caption{Chinese GSM8K}
    \end{subfigure}
    \begin{subfigure}[b]{0.24\textwidth}
        \centering
        \includegraphics[width=\textwidth]{similarity/Similarity/Qwen_Qwen2-7B-Instruct_Chinese_LogicalDeduction_train_modified_similarity.png}
        \caption{Chinese-LogicalDeduction}
    \end{subfigure}

    \begin{subfigure}[b]{0.24\textwidth}
        \centering
        \includegraphics[width=\textwidth]{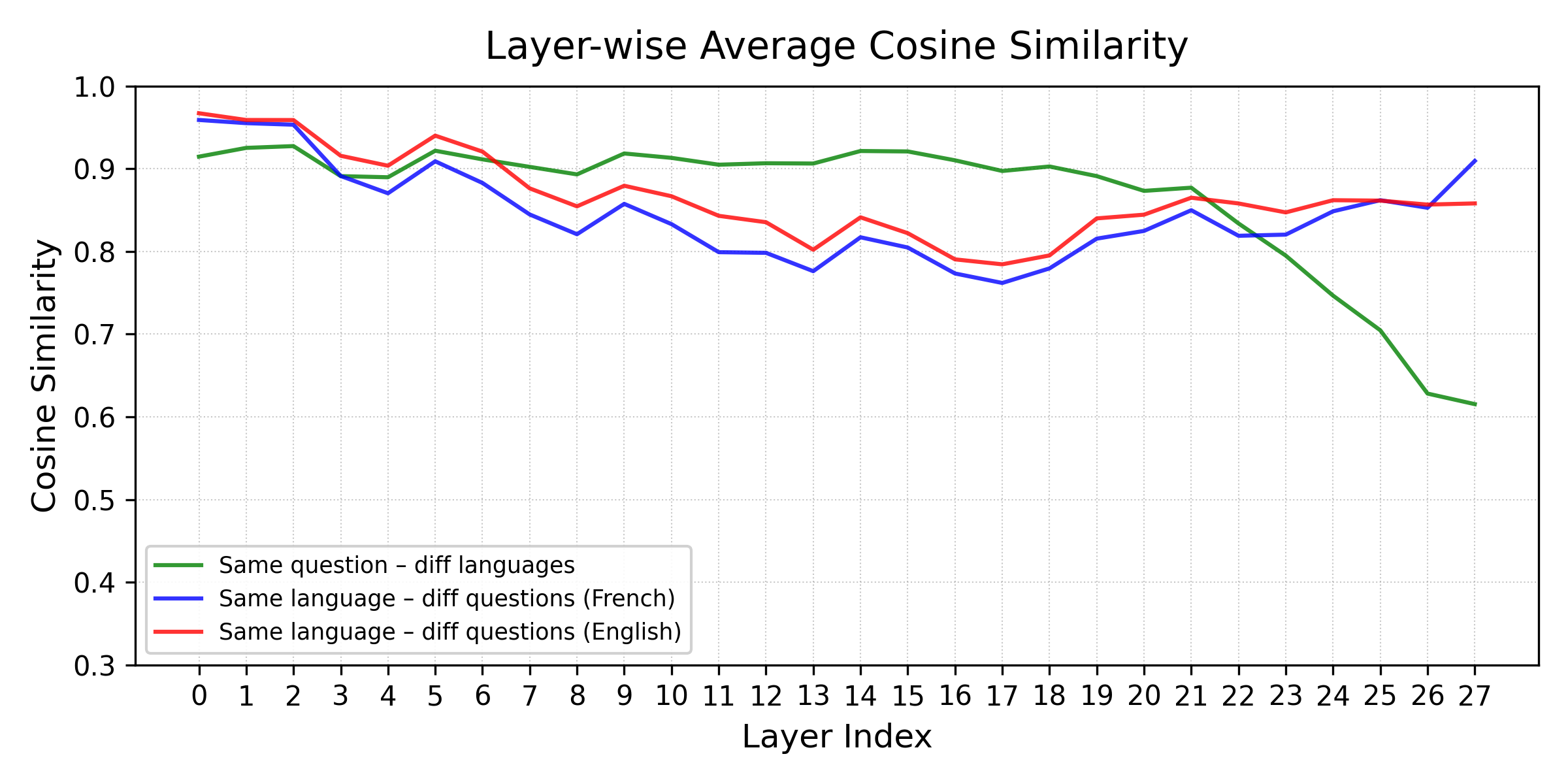}
        \caption{French - ProofWriter}
    \end{subfigure}
    \begin{subfigure}[b]{0.24\textwidth}
        \centering
        \includegraphics[width=\textwidth]{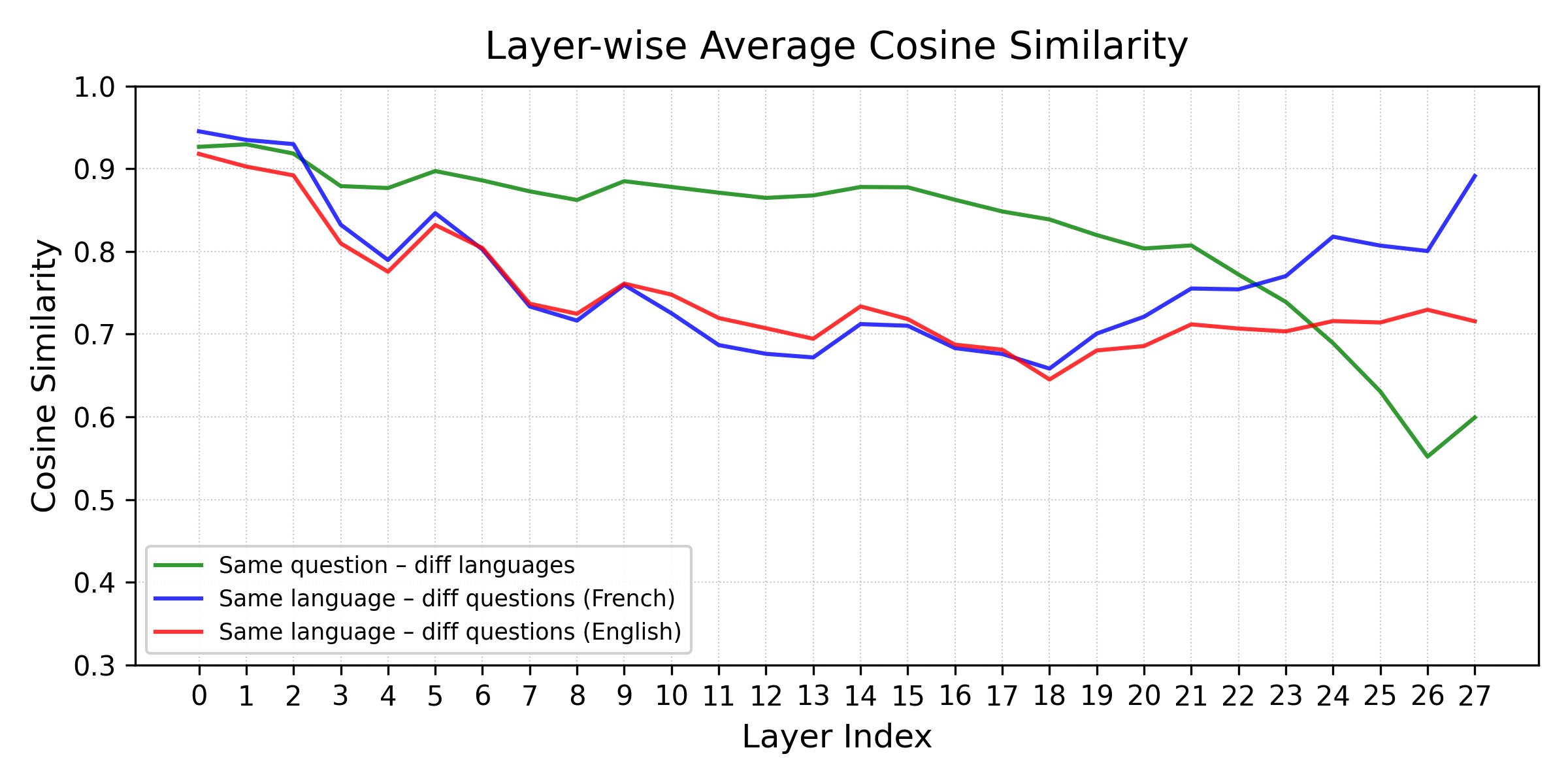}
        \caption{French - FOLIO}
    \end{subfigure}
    \begin{subfigure}[b]{0.24\textwidth}
        \centering
        \includegraphics[width=\textwidth]{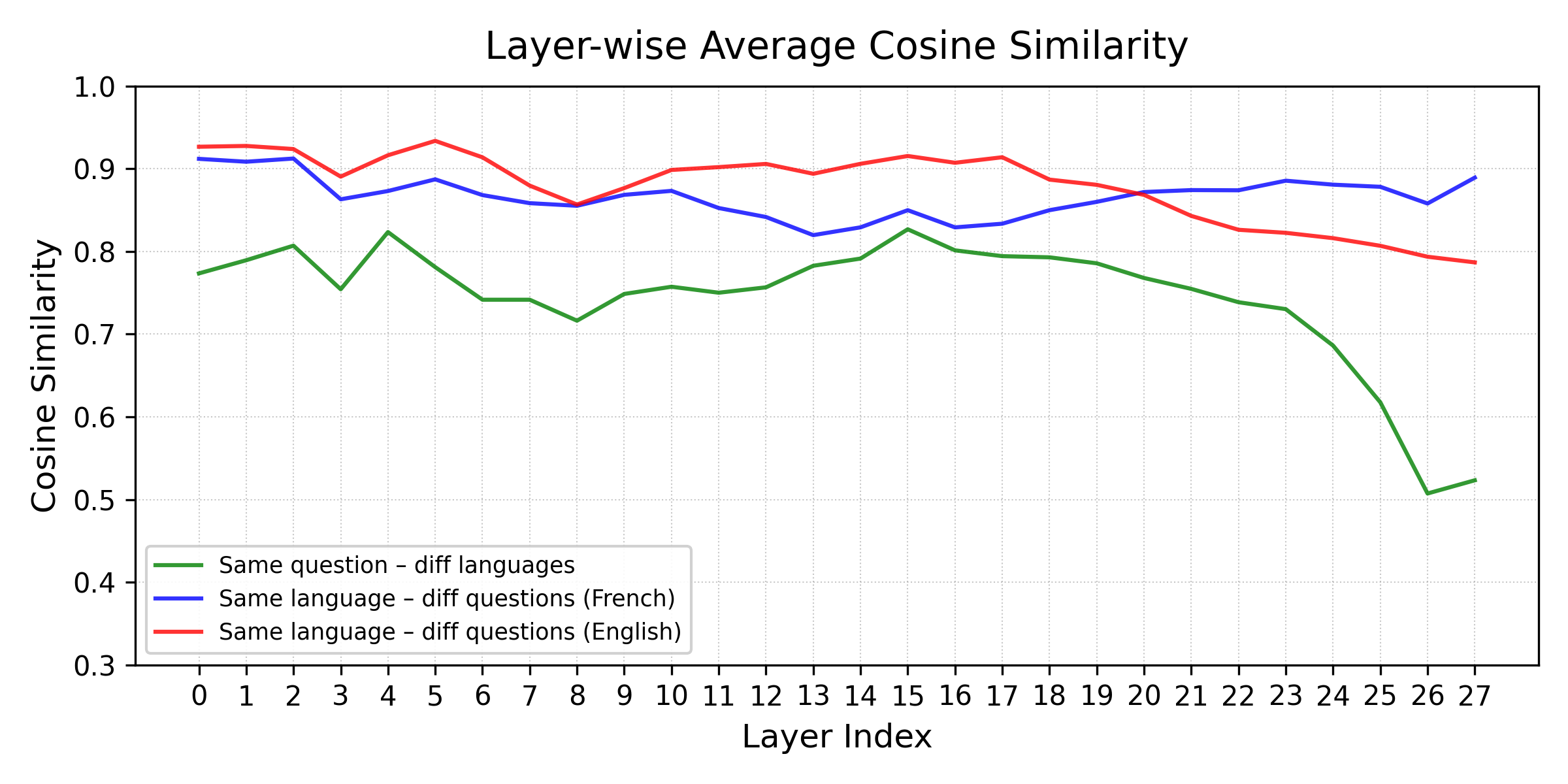}
        \caption{French GSM8K}
    \end{subfigure}
    \begin{subfigure}[b]{0.24\textwidth}
        \centering
        \includegraphics[width=\textwidth]{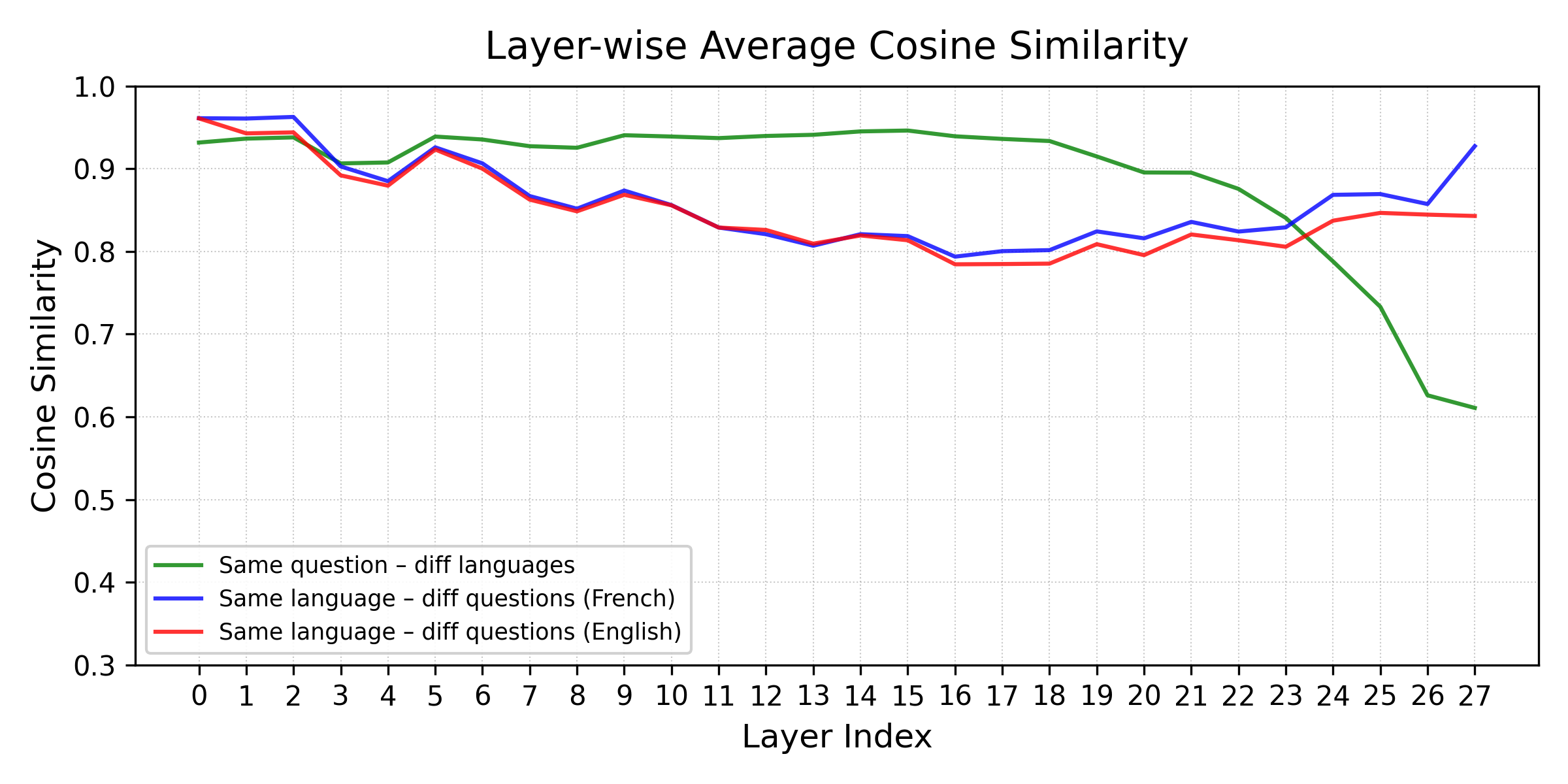}
        \caption{French - LogicalDeduction}
    \end{subfigure}

    \begin{subfigure}[b]{0.24\textwidth}
        \centering
        \includegraphics[width=\textwidth]{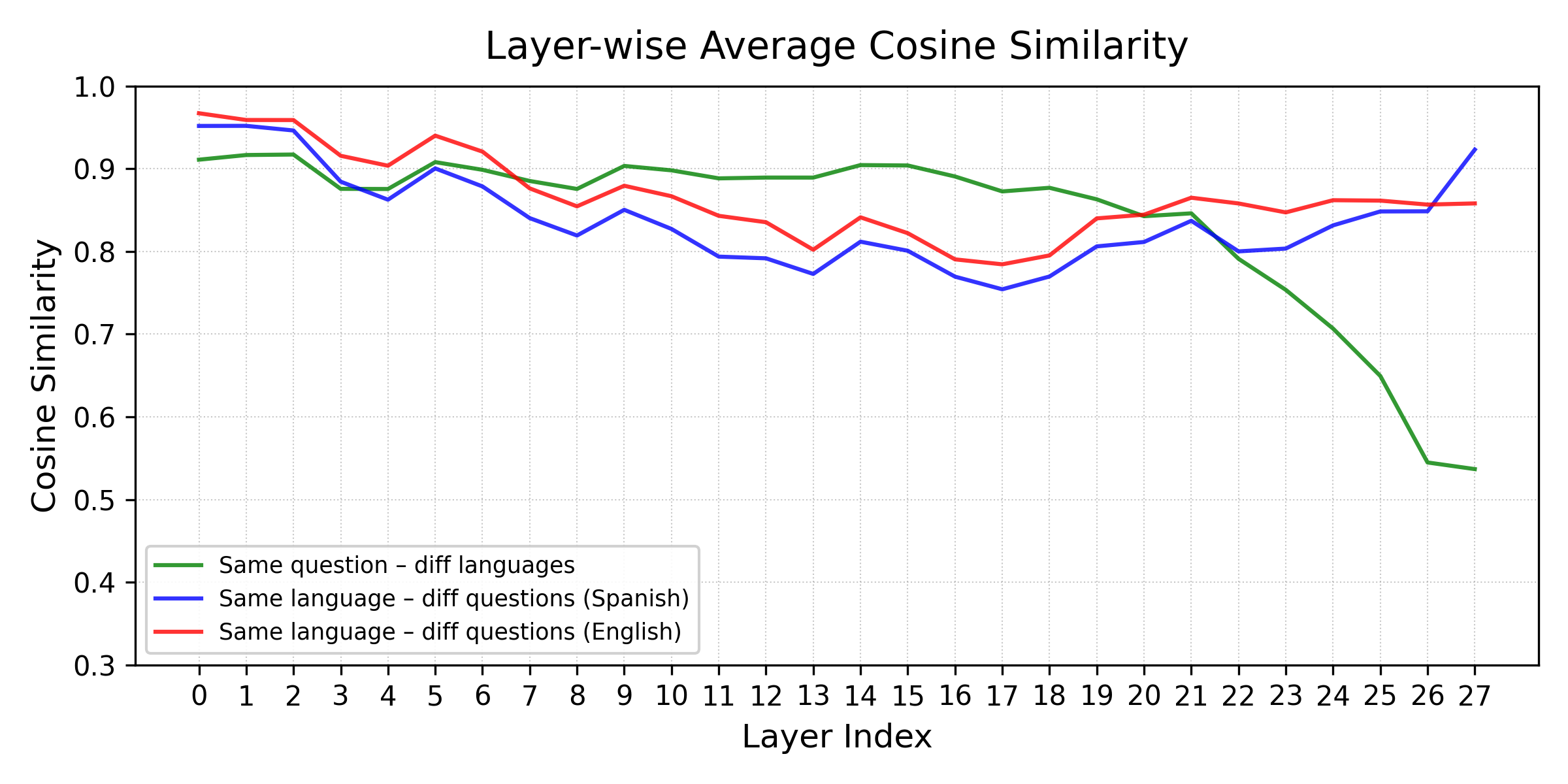}
        \caption{Spanish - ProofWriter}
    \end{subfigure}
    \begin{subfigure}[b]{0.24\textwidth}
        \centering
        \includegraphics[width=\textwidth]{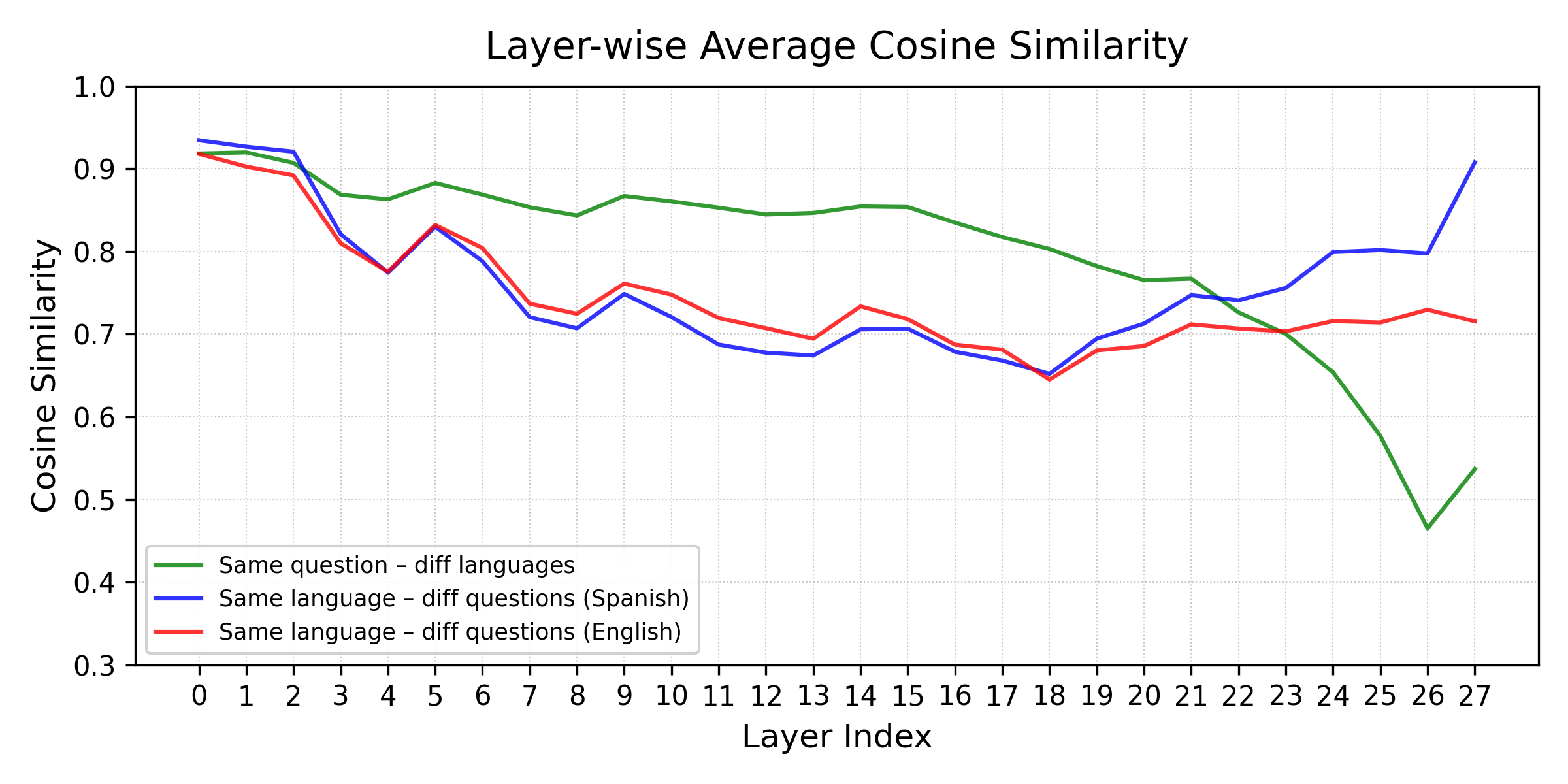}
        \caption{Spanish - FOLIO}
    \end{subfigure}
    \begin{subfigure}[b]{0.24\textwidth}
        \centering
        \includegraphics[width=\textwidth]{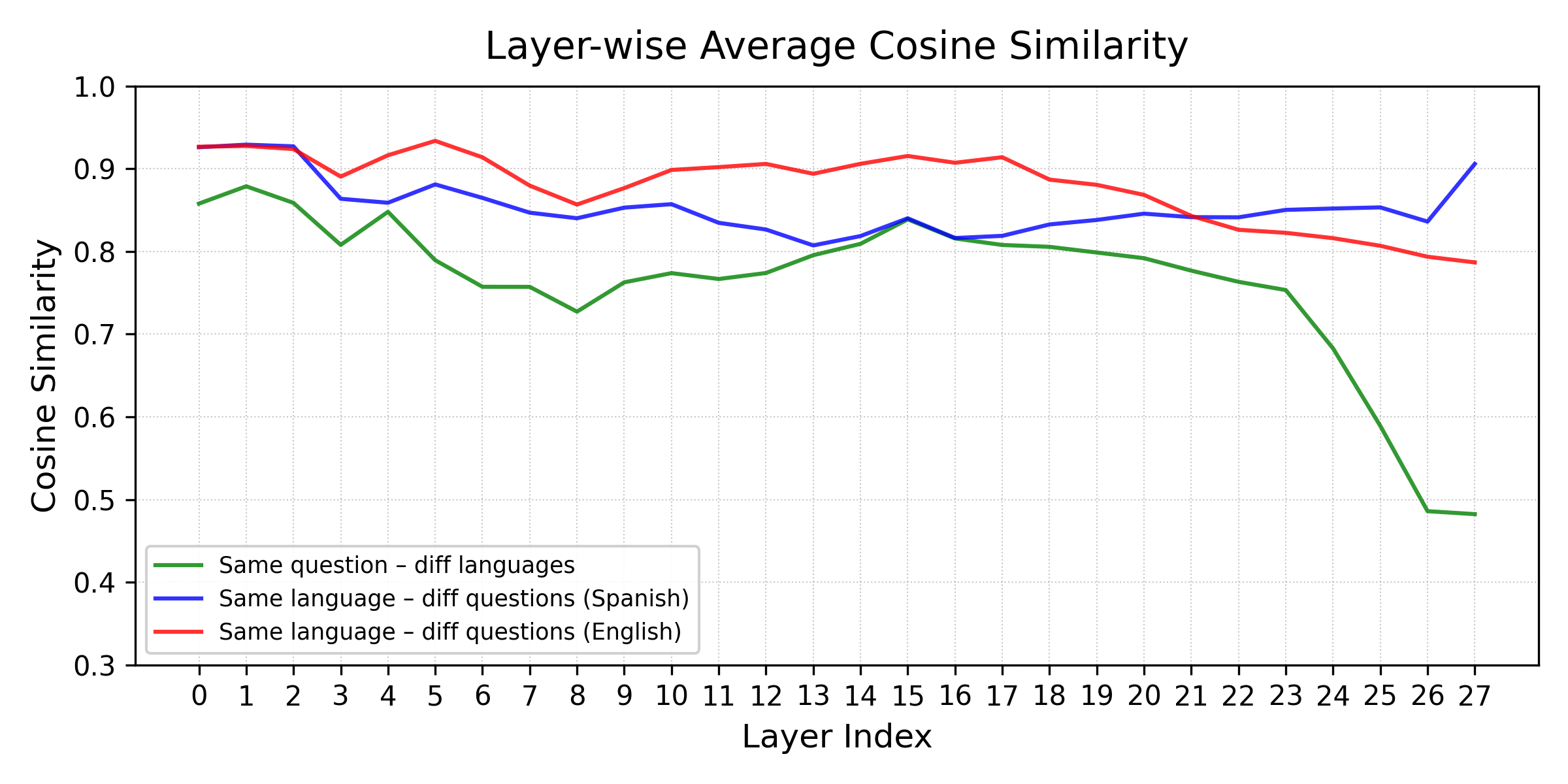}
        \caption{Spanish GSM8K}
    \end{subfigure}
    \begin{subfigure}[b]{0.24\textwidth}
        \centering
        \includegraphics[width=\textwidth]{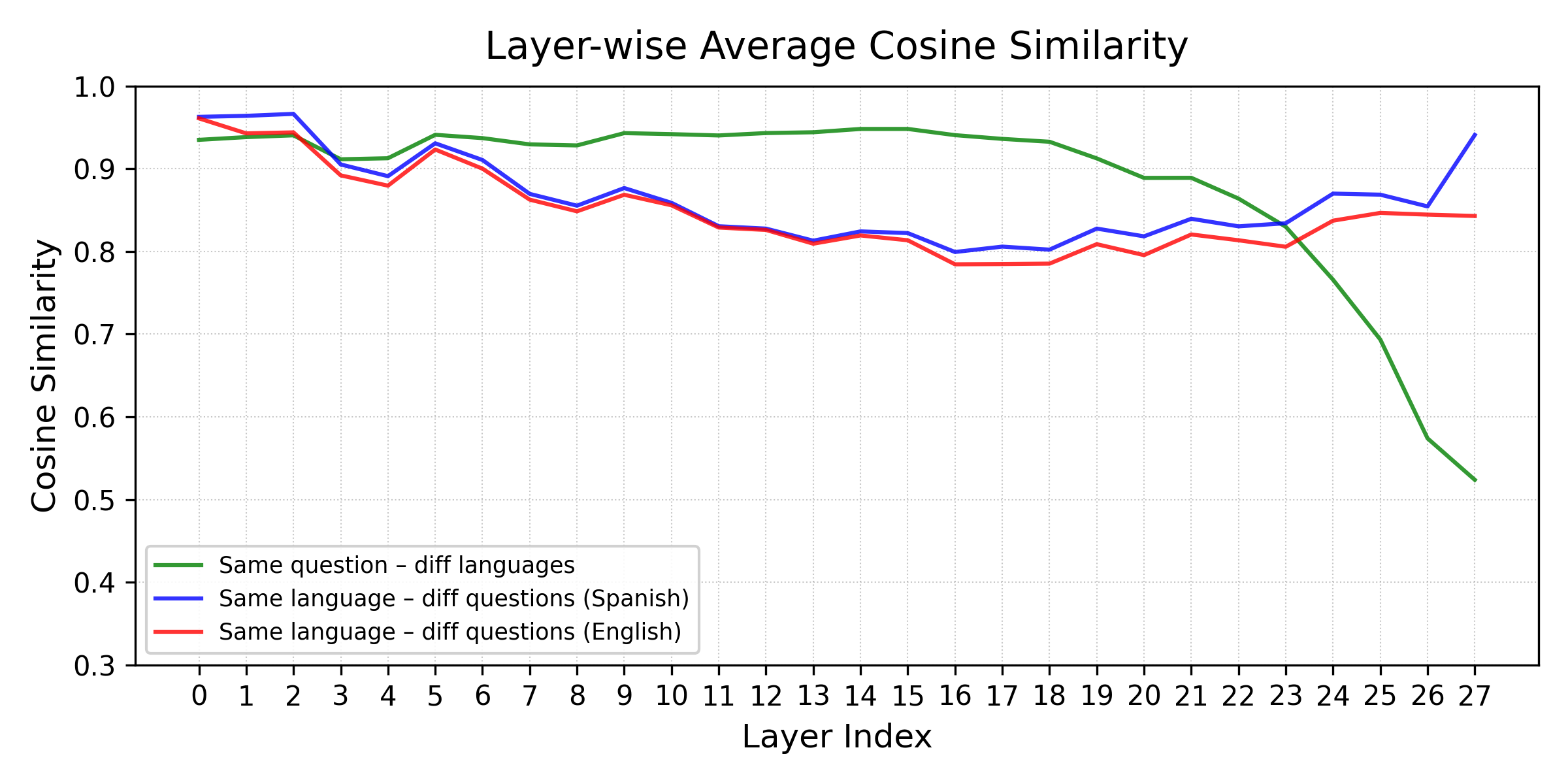}
        \caption{Spanish - LogicalDeduction}
    \end{subfigure}

    \begin{subfigure}[b]{0.24\textwidth}
        \centering
        \includegraphics[width=\textwidth]{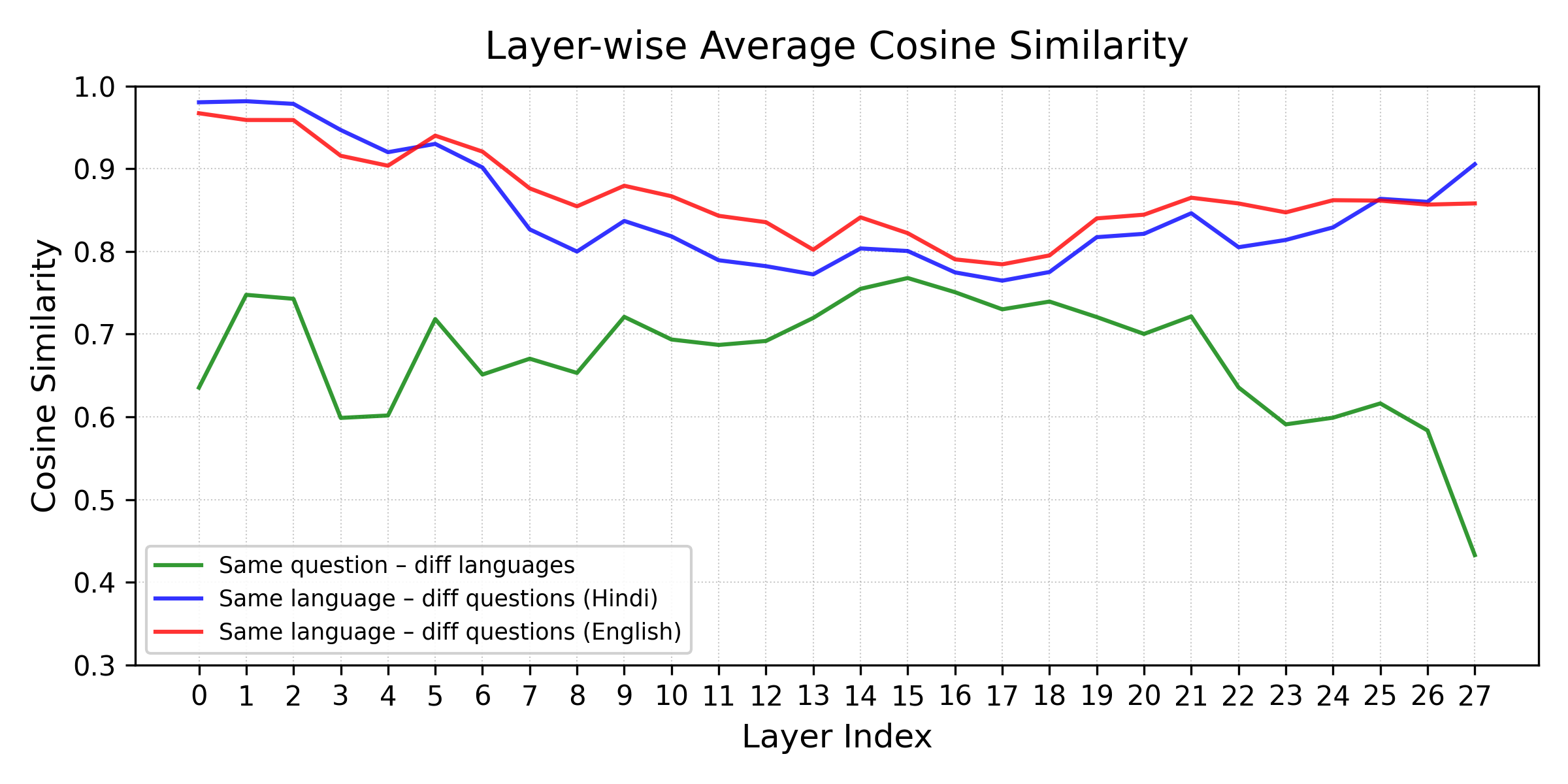}
        \caption{Hindi - ProofWriter}
    \end{subfigure}
    \begin{subfigure}[b]{0.24\textwidth}
        \centering
        \includegraphics[width=\textwidth]{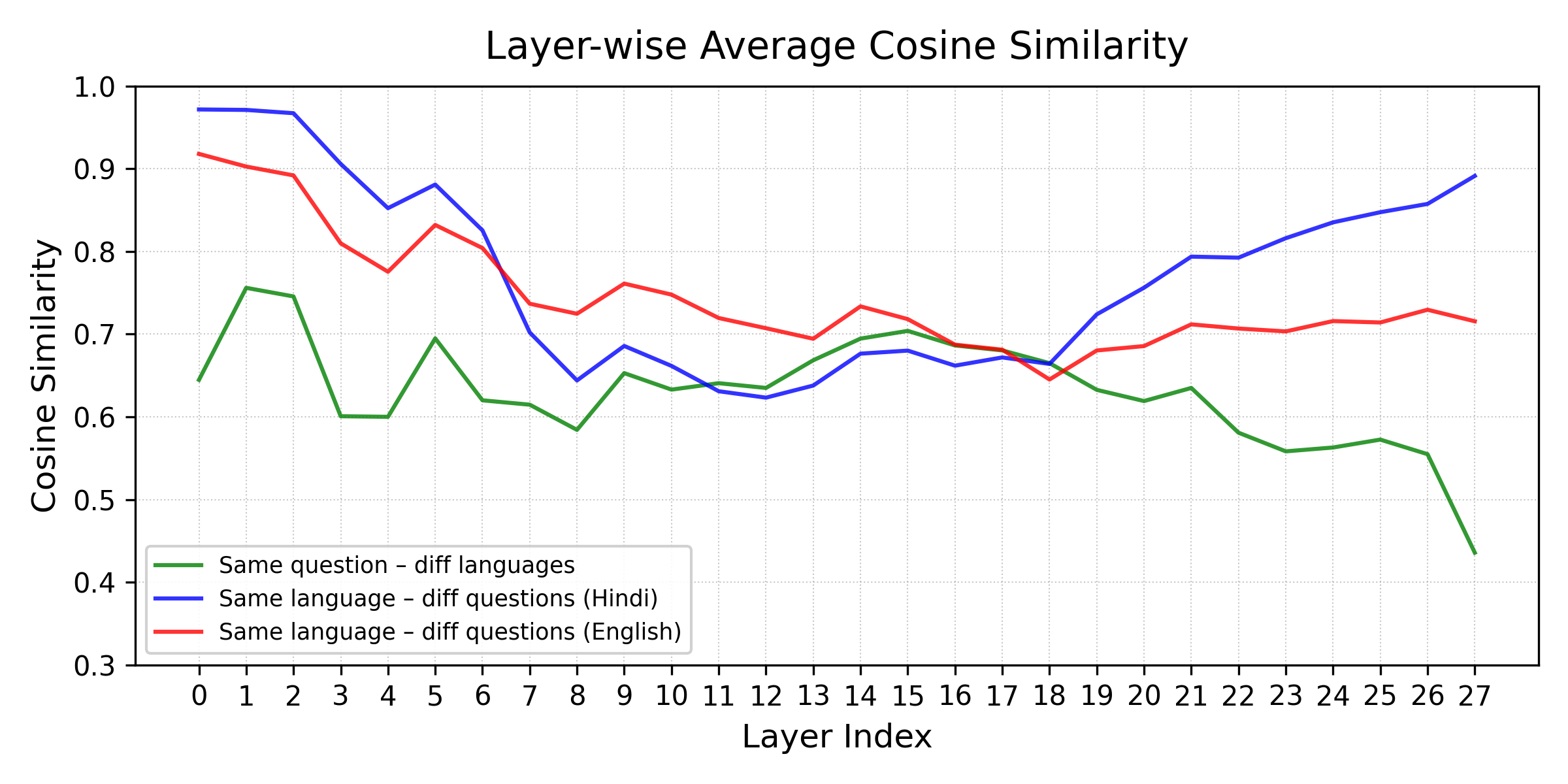}
        \caption{Hindi - FOLIO}
    \end{subfigure}
    \begin{subfigure}[b]{0.24\textwidth}
        \centering
        \includegraphics[width=\textwidth]{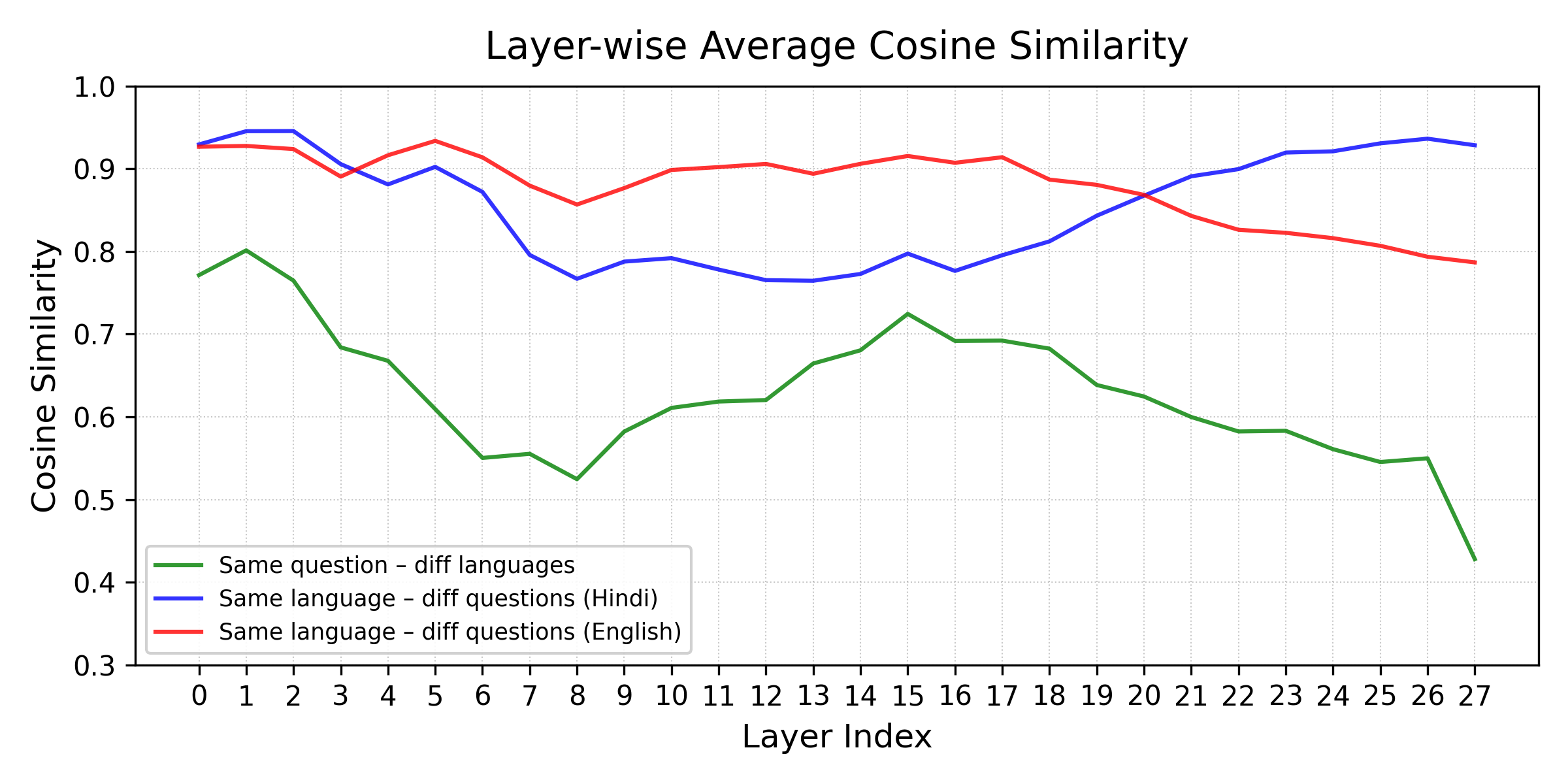}
        \caption{Hindi GSM8K}
    \end{subfigure}
    \begin{subfigure}[b]{0.24\textwidth}
        \centering
        \includegraphics[width=\textwidth]{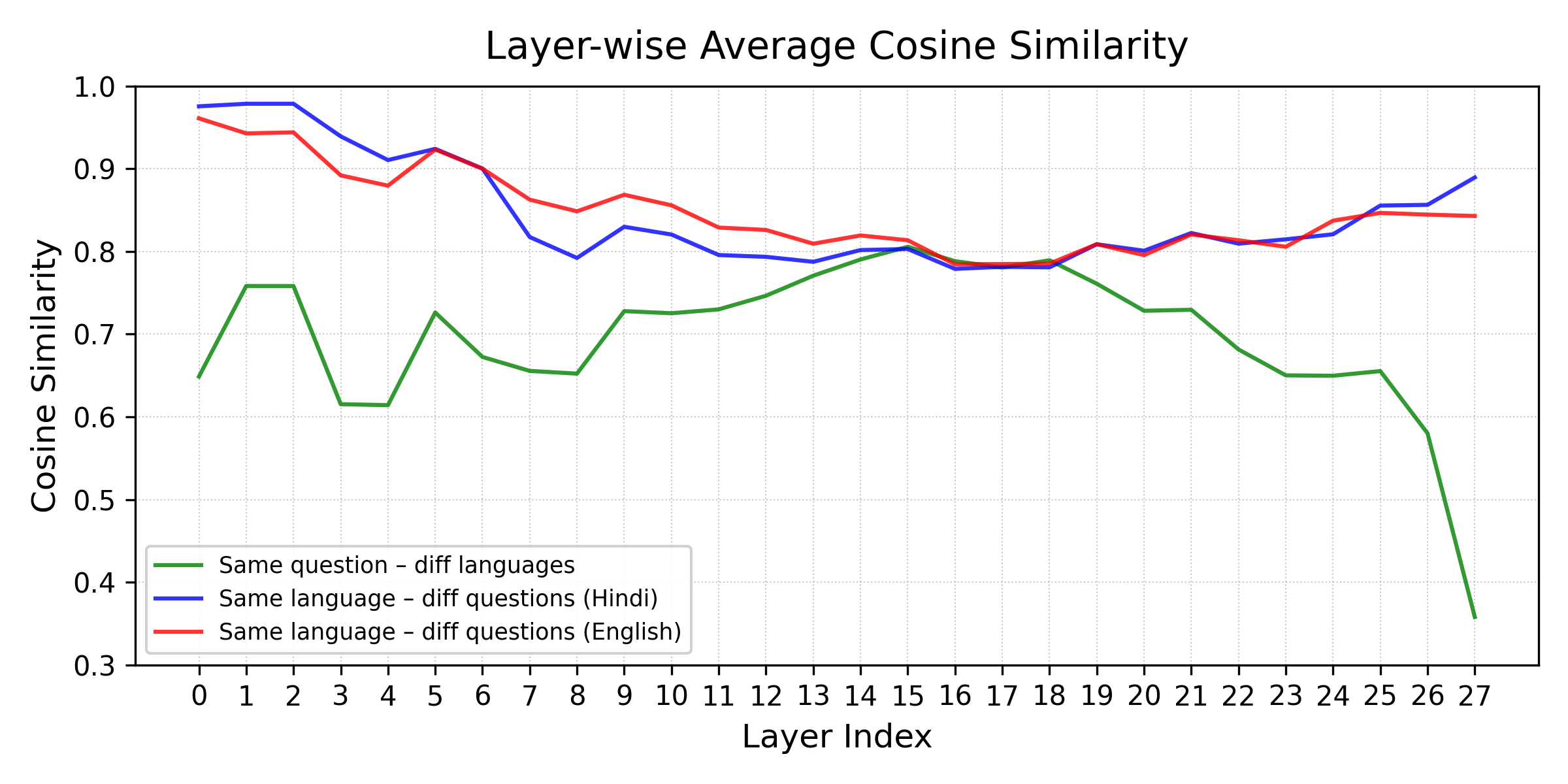}
        \caption{Hindi - LogicalDeduction}
    \end{subfigure}

    \caption{Performance of Qwen2-7B-Instruct across languages and datasets.}
    \label{fig:Qwen_results}
\end{figure}
\clearpage

\section{Details of Cosine Similarity between Hidden Stated in Adjacent Layers}

In this section, we present the full details of the cosine similarity between hidden states in adjacent layers. We conduct experiments on LLMs in Subsection \ref{LLMS} across four datasets in Subsection \ref{dataset}, which directly facilitate the identification of idle reasoning layers.

 \begin{figure}[htbp]
    \centering
    \begin{subfigure}[b]{0.24\textwidth}
    \centering
        \includegraphics[width=\textwidth]{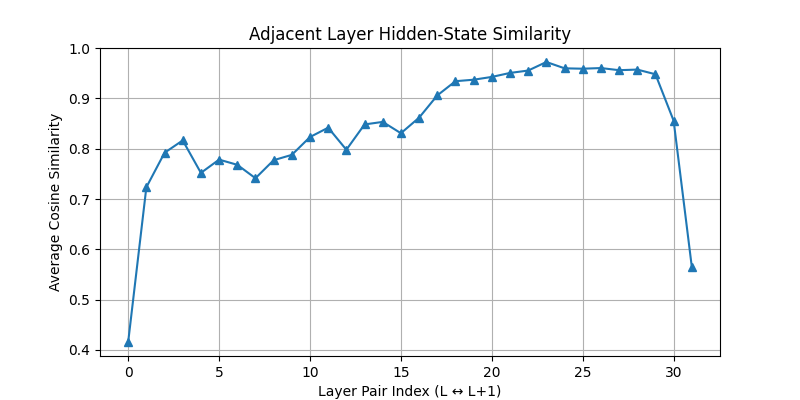}
        \caption{ProofWriter}
    \end{subfigure}
    \begin{subfigure}[b]{0.24\textwidth}
    \centering
        \includegraphics[width=\textwidth]{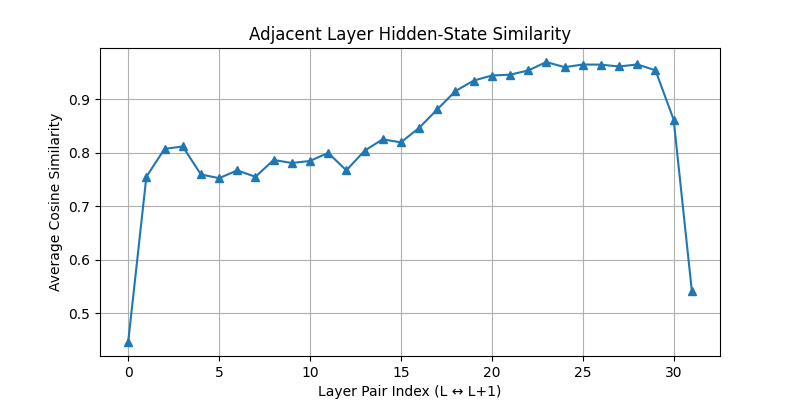}
        \caption{FOLIO}
    \end{subfigure}
    \begin{subfigure}[b]{0.24\textwidth}
    \centering
        \includegraphics[width=\textwidth]{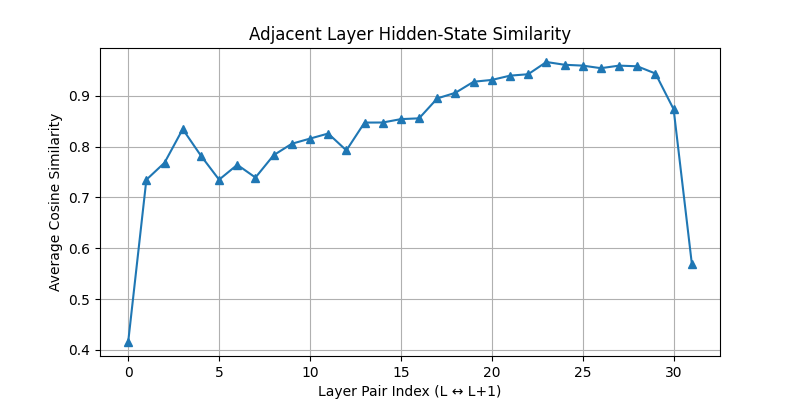}
        \caption{LogicalDeduction}
    \end{subfigure}
    \begin{subfigure}[b]{0.24\textwidth}
    \centering
        \includegraphics[width=\textwidth]{visualization/Visual_Llama-3/gsm8k_alpaca_train_last/English_adjacent_layer_similarity.png}  %
        \caption{GSM8K}
    \end{subfigure}
    \caption{Cosine Similarity between Hidden Stated in Adjacent Layers of Llama-3-8B-Instruct }
    \label{fig_full_adj}
\end{figure}

 \begin{figure}[htbp]
    \centering
    \begin{subfigure}[b]{0.24\textwidth}
    \centering
        \includegraphics[width=\textwidth]{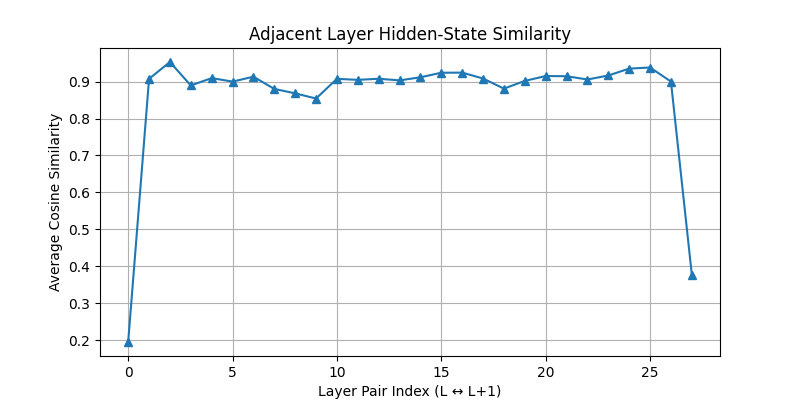}
        \caption{ProofWriter}
    \end{subfigure}
    \begin{subfigure}[b]{0.24\textwidth}
    \centering
        \includegraphics[width=\textwidth]{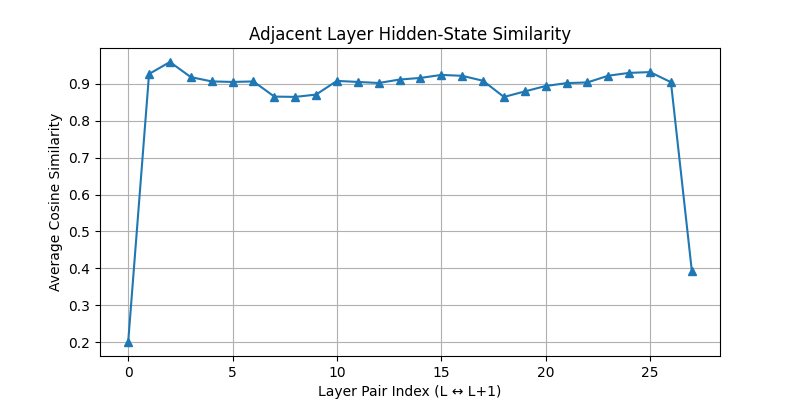}
        \caption{FOLIO}
    \end{subfigure}
    \begin{subfigure}[b]{0.24\textwidth}
    \centering
        \includegraphics[width=\textwidth]{visualization/Visual_Qwen_Qwen2-7B-Instruct/LogicalDeduction_train_modified_last/English_adjacent_layer_similarity.png}
        \caption{LogicalDeduction}
    \end{subfigure}
    \begin{subfigure}[b]{0.24\textwidth}
    \centering
        \includegraphics[width=\textwidth]{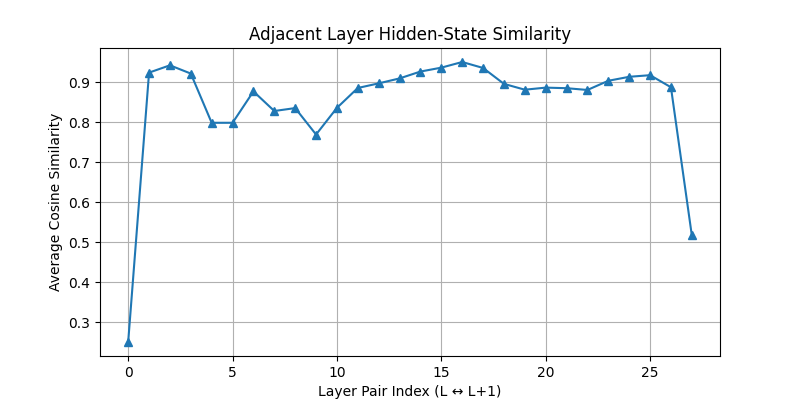}  %
        \caption{GSM8K}
    \end{subfigure}
    \caption{Cosine Similarity between Hidden Stated in Adjacent Layers of Qwen2-7B-Instruct }
\end{figure}
\clearpage

\section{Results of the Method for Functional Layer Segmentation}

In this Section, we provide a detailed explanation of the method used to approximate a reasonable location in Section \ref{LIFT}, and introduce the values we ultimately selected for the functional layer segmentation.

Our method identifies the boundary between conceptualization and reasoning layers by finding the first significant inflection point in the cross-lingual task similarity curve ($g^k$). We locate this by calculating the discrete second derivative, $\Delta^2g^k = g^{k+1} - 2g^k + g^{k-1}$, and identifying the first layer $k^*$ where the sign of $\Delta^2g^k$ changes. To determine the boundary of idle reasoning layers, we simply select layers where the adjacent cosine similarity exceeds a predefined threshold $\beta$.

For instance, for the Llama-3-8B-Instruct model on the ProofWriter dataset (comparing Chinese and English versions), the layer-wise mean cosine similarities ($g^k$) are:
[0.582, 0.640, 0.745, 0.772, 0.768, 0.780, 0.781, 0.797, 0.795, 0.812, 
 0.815, 0.824, 0.851, 0.841, 0.826, 0.826, 0.821, 0.811, 0.811, 0.799, 
 0.788, 0.793, 0.804, 0.813, 0.799, 0.748, 0.715, 0.677, 0.676, 0.629, 
 0.515, 0.357]

Calculating the discrete second derivative for this sequence, the first few values are:
\begin{itemize}
    \item $\Delta^2g^1 = g^2 - 2g^1 + g^0 \approx 0.745 - 2(0.640) + 0.582 = \mathbf{+0.047}$
    \item $\Delta^2g^2 = g^3 - 2g^2 + g^1 \approx 0.772 - 2(0.745) + 0.640 = \mathbf{-0.078}$
\end{itemize}
The first sign change occurs at $k^*=2$, which we identify as the boundary. We have performed this analysis across all our experimental settings and found this result to be remarkably consistent for this model. For the vast majority of tasks and languages, the first inflection point for Llama-3 is identified at \textbf{layer 2}. Based on this robust finding, we select layer 2 as the conceptualization-reasoning boundary for this model. In contrast, the boundary between active and idle reasoning layers is task-dependent; generally, more challenging tasks require a larger number of active reasoning layers. The specific segmentation results are summarized in the table below.

During the experiments, we observed that for certain datasets, the adjacent cosine similarity of Qwen2-7B-Instruct failed to exceed the threshold $\beta$. In such cases, we conclude that the task does not contain idle reasoning layers.

\begin{table}[h]  
    \centering  
    \resizebox{\linewidth}{!}{
    \begin{tabular}{|c|c|c|c|c|}  
        \hline  
        \multicolumn{5}{|c|}{Llama-3-8B-Instruct} \\ \hline  
        Datasets & conceptualization layers & active reasoning layers & idle reasoning layers & textualization layers\\ \hline  
        ProofWriter & [0,2) & [2,20) & [20,30) &[30,32)\\ \hline  
        FOLIO & [0,2) &[2,20) & [20,30) &[30,32)\\ \hline  
        LogicalDeduction & [0,2) & [2,22) & [22,30)  &[30,32)\\ \hline  
        GSM8K & [0,2) & [2,22) & [22,30) &[30,32)\\ \hline  
    \end{tabular}
    }
    \caption{Functional Layer Segmentation of Llama-3-8B-Instruct}  
    \label{tab:example1}  
\end{table}

\begin{table}[h]  
    \centering  
    \resizebox{\linewidth}{!}{
    \begin{tabular}{|c|c|c|c|c|}  
        \hline  
        \multicolumn{5}{|c|}{Qwen2-7B-Instruct} \\ \hline  
        Datasets & conceptualization layers & active reasoning layers & idle reasoning layers & textualization layers\\ \hline  
        ProofWriter & [0,3) & [3,26) & - &[26,28)\\ \hline  
        FOLIO & [0,3) &[3,26) & - &[26,28)\\ \hline  
        LogicalDeduction & [0,3) & [3,24) & [24,26)  &[26,28)\\ \hline  
        GSM8K & [0,3) & [3,26) & - &[26,28)\\ \hline  
    \end{tabular}  
    }
    \caption{Functional Layer Segmentation of Qwen2-7B-Instruct}  
    \label{tab:example3}  
\end{table}  

\clearpage
\section{Appendix: Layer Sensitivity Probing via Simulated Quantization}
\label{sec:appendix_sensitivity}

To identify the layers most susceptible to quantization noise, we developed a sensitivity probing framework based on \textbf{Activation-aware Weight Quantization (AWQ)}. This approach allows us to simulate the effects of low-bit precision on specific model components without the need for specialized inference kernels.

\subsection{Simulated Quantization Mechanism}
Instead of physically packing weights into lower-bit formats (e.g., INT4) which requires specific hardware kernels for inference, we employ a \textit{Simulated Quantization} (or Pseudo-Quantization) strategy. 
The quantization process proceeds in two steps:
\begin{enumerate}
    \item \textbf{Parameter Search:} We first compute the optimal scaling factors for each linear layer using the standard AWQ algorithm, which minimizes the reconstruction error of the layer's output activations.
    \item \textbf{Noise Injection:} We then apply a quantize-dequantize (QDQ) operation to the weights. The weights are quantized to the target precision (INT4) and immediately dequantized back to floating-point format (\texttt{bfloat16}). 
\end{enumerate}

\subsection{Experimental Setup and Data Protocols}
Our sensitivity analysis involves determining the impact of quantizing specific blocks of layers while keeping others in full precision. The data protocols for calibration and evaluation are designed as follows:

\paragraph{Calibration:}
To ensure the quantization parameters capture the activation distribution of the target domain, we perform calibration using the \textbf{training dataset} corresponding to each specific task. We sample a small subset of sequences from the training data to calculate the AWQ scaling factors, ensuring the quantization is tailored to the relevant data distribution.

\paragraph{Evaluation:}
To efficiently measure the sensitivity of different layers, we conduct the evaluation directly on the training dataset. Specifically, to accelerate the probing process and enable rapid iteration across multiple layer configurations, we utilize the \textbf{first 1,000 samples} of the training set for testing. While standard evaluation typically uses a held-out test set, using the training subset for sensitivity probing provides a sufficient signal to distinguish between robust and sensitive layers, as we are primarily measuring the relative degradation of the model's inherent knowledge representation under quantization noise.

\paragraph{Probing Strategy:}
We adopt a ``Drop'' and ``Recovery'' methodology to isolate layer importance. In the \textit{Drop} mode, a target block of layers is simulated in INT4 while the rest remain in BF16, measuring the performance drop. Conversely, the \textit{Recovery} mode keeps the target block in BF16 while quantizing the rest, measuring the performance recovery.

\section{Results of Identifying Functionally Critical Layers}

\subsection{Identification of Target Layers in Qwen2-7B-Instruct}
\label{sec:appendix_qwen_analysis}

To validate the generalizability of our functional segmentation across different architectures, we performed the ``Drop'' and ``Recovery'' ablation studies on Qwen2-7B-Instruct. The results, presented in Table~\ref{tab:ablation_qwen}, consistently identify the \textbf{active reasoning layers} as the functional bottleneck and the optimal target for fine-tuning.

\textbf{High Sensitivity in Drop Mode:} 
As shown in the first section of Table~\ref{tab:ablation_qwen}, omitting the active reasoning layers results in the most severe performance degradation across all evaluated benchmarks. For instance, on the ProofWriter dataset, accuracy drops to its lowest point at 43.15\%, significantly underperforming the configurations where Conceptualization (48.45\%) or Textualization (46.68\%) layers are dropped. This indicates that the active reasoning layers house the most critical parameters for logical inference.

\textbf{Superior Efficiency in Recovery Mode:} 
Conversely, in the Recovery experiments, fine-tuning exclusively the active reasoning layers yields the highest performance restoration. Notably, on GSM8K, targeting these layers achieves an accuracy of 84.39\%, surpassing other layer groups. This confirms that focusing computational resources on this specific block maximizes the marginal gain of fine-tuning.

\textbf{Conclusion:} 
Based on these observations, we confirm that the active reasoning layers act as the primary performance determinant for Qwen2-7B-Instruct, justifying their selection as the target layers in our proposed method.

\begin{table}[htbp]
  \centering

  \resizebox{\linewidth}{!}{
  \begin{tabular}{|c|c|c|c|c|}
    \hline
    \multicolumn{5}{|c|}{Qwen2-7B-Instruct (Drop)} \\ \hline
    Datasets & conceptualization layers & active reasoning layers & idle reasoning layers & textualization layers \\ \hline
    ProofWriter & 48.45\% ($\pm$1.51\%) & \textbf{43.15\% ($\pm$2.05\%)} & - & 46.68\% ($\pm$1.83\%) \\ \hline
    FOLIO & 42.18\% ($\pm$1.45\%) & \textbf{41.71\% ($\pm$3.67\%)} &  - & 42.16\% ($\pm$1.17\%) \\ \hline
    LogicalDeduction & 50.50\% ($\pm$1.54\%) & \textbf{49.51\% ($\pm$0.86\%)} & 50.31\% ($\pm$1.56\%) & 49.56\% ($\pm$0.78\%) \\ \hline
    gsm8k & 84.16\% ($\pm$0.78\%) & \textbf{82.79\% ($\pm$0.54\%)} &  & 84.03\% ($\pm$0.76\%) \\ \hline
  \end{tabular}
  }
  
  \vspace{0.5cm} %

  \resizebox{\linewidth}{!}{
  \begin{tabular}{|c|c|c|c|c|}
    \hline
    \multicolumn{5}{|c|}{Qwen2-7B-Instruct (Recovery)} \\ \hline
    Datasets & conceptualization layers & active reasoning layers & idle reasoning layers & textualization layers \\ \hline
    ProofWriter & 39.86\% ($\pm$2.06\%) & \textbf{45.75\% ($\pm$1.05\%)} &  & 41.30\% ($\pm$1.56\%) \\ \hline
    FOLIO & 40.36\% ($\pm$2.82\%) & \textbf{43.46\% ($\pm$2.17\%)} &  & 41.81\% ($\pm$2.39\%) \\ \hline
    LogicalDeduction & 47.45\% ($\pm$1.10\%) & \textbf{49.68\% ($\pm$0.82\%)} & 47.80\% ($\pm$1.14\%) & 48.84\% ($\pm$1.75\%) \\ \hline
    gsm8k & 81.91\% ($\pm$0.86\%) & \textbf{84.39\% ($\pm$0.92\%)} &  & 82.39\% ($\pm$1.16\%) \\ \hline
  \end{tabular}
  }
  
  \caption{Performance of quantization probing (Drop and Recovery) over Various Layer Ranges of Qwen2-7B-Instruct (Averaged over 8 Seeds)}
  \label{tab:ablation_qwen}
\end{table}

\clearpage
\section{Experimental details}

\subsection{Experimental Environment}
Our experiments were conducted on a server equipped with \textbf{four NVIDIA A40 GPUs}, each with 46 GB of memory. All training and evaluation tasks were performed on a \textbf{single A40 GPU}. The server is powered by an \textbf{AMD EPYC 9654 96-Core Processor} with \textbf{755 GiB of RAM} and \textbf{16 TB of storage}.

We used \textbf{Ubuntu 20.04.6 LTS} as the operating system, with \textbf{Python 3.10.14} as the main programming environment. All deep learning experiments were implemented using \textbf{PyTorch 2.5.1+cu124}, leveraging CUDA for GPU acceleration.

\subsection{Software and Libraries}
\begin{itemize}
    \item \textbf{CUDA Version}: 12.5  
    \item \textbf{Driver Version}: 555.58.02  
    \item \textbf{Deep Learning Framework}: PyTorch 2.5.1 with CUDA 12.4 support  
    \item \textbf{Python Version}: 3.10.14  
    \item \textbf{Operating System}: Ubuntu 20.04.6 LTS  
\end{itemize}

\subsection{Training Configuration}
We fine-tuned these models using the following hyperparameters:
\begin{itemize}
    \item \textbf{Learning Rate}: 1e-5
    \item \textbf{LoRA Rank}: 64
    \item \textbf{LoRA Alpha}: 128
    \item \textbf{LoRA Dropout}: 0.05
    \item \textbf{Batch Size per Device}: 4
    \item \textbf{Gradient Accumulation Steps}: 8
    \item \textbf{Max Sequence Length}: 512
    \item \textbf{Random Seeds}: 0, 1, 2, 3
    \item \textbf{Warmup Ratio}: 0.03
\end{itemize}

\subsection{Execution Strategy}
All training and evaluation were executed on a single A40 GPU to maintain consistency. Multi-GPU capabilities were available but not utilized for this experiment. Distributed strategies like FSDP or DeepSpeed were not applied in this setup.

\section{AdaLoRA Baseline Configuration}
\label{appendix:adalora_config}

To strictly align the global trainable parameter budget between LIFT and AdaLoRA, we dynamically adjusted AdaLoRA's rank settings based on the size of the functional bottleneck identified by LIFT. Specifically, for a given dataset where LIFT targets a functional bottleneck of $k$ layers with a fixed rank of $r=64$, we constrain AdaLoRA's target average rank across all $K=32$ layers of Llama-3 to:
$$r_{target} = 64 \times \frac{k}{K} = 2k$$

To afford AdaLoRA sufficient search space for its pruning schedule, the initial rank is set to $r_{init} = r_{target} + 8$. Other pruning hyperparameters are maintained at their standard settings: $t_{init\_ratio} = 0.25$, $t_{final\_ratio} = 0.8$, and $\Delta T = 100$. This configuration guarantees that AdaLoRA strictly adheres to the exact same parameter budget as LIFT for every evaluated task.

\section{Results of Ablation Experiments}
In this section, we present the results of our ablation experiments. Tables~\ref{tab:ablation1}, and \ref{tab:ablation3} report the performance of fine-tuning various functional regions of different large language models (LLMs), including the conceptualization layers, active reasoning layers, idle reasoning layers, and textualization layers. 

From these experiments, it is evident that our LIFT algorithm significantly outperforms competing approaches in most cases. The fine-tuning strategy, which focuses explicitly on functionally relevant layers, yields substantial improvements in task performance. These results highlight the critical role of selectively targeting specific functional layers within LLMs for effective task-specific optimization.

\begin{table}[h]
  \centering
  \resizebox{\linewidth}{!}{
  \begin{tabular}{|c|c|c|c|c|}
    \hline
    \multicolumn{5}{|c|}{Llama-3-8B-Instruct} \\ \hline
    Datasets & conceptualization layers & active reasoning layers & idle reasoning layers & textualization layers \\ \hline
    ProofWriter & 37.25\% (±1.04\%) & 57.67\% (±0.36\%) & 48.08\% (±0.75\%) & 46.38\% (±0.89\%) \\ \hline
    FOLIO & 53.43\% (±1.65\%) & 58.21\% (±2.35\%) & 56.74\% (±1.29\%) & 55.51\% (±2.31\%) \\ \hline
    LogicalDeduction & 45.67\% (±0.72\%) & 53.58\% (±2.47\%) & 46.83\% (±1.67\%) & 45.25\% (±0.88\%) \\ \hline
    GSM8K & 73.96\% (±0.98\%) & 67.87\% (±0.64\%) & 65.22\% (±0.53\%) & 56.48\% (±0.58\%) \\ \hline
  \end{tabular}
  }
  \caption{Performance of Fine-Tuning over Various Layer Ranges of Llama-3-8B-Instruct}
  \label{tab:ablation1}
\end{table}

\begin{table}[h]
  \centering
  \resizebox{\linewidth}{!}{
  \begin{tabular}{|c|c|c|c|c|}
    \hline
    \multicolumn{5}{|c|}{Qwen2-7B-Instruct} \\ \hline
    Datasets & conceptualization layers & active reasoning layers & idle reasoning layers & textualization layers\\ \hline
    ProofWriter & 45.50\% (±0.93\%) & 53.46\% (±2.36\%) & - & 47.46\% (±0.85\%) \\ \hline
    FOLIO & 27.08\% (±4.19\%) & 44.73\% (±1.98\%) & -  & 28.55\% (±1.62\%) \\ \hline
    LogicalDeduction & 48.08\% (±0.83\%) & 57.08\% (±3.29\%) & 49.00\% (±0.47\%)        & 52.58\% (±2.15\%) \\ \hline
    GSM8K & 75.34\% (±0.66\%) & 75.78\% (±0.24\%) & - & 72.29\% (±0.72\%) \\ \hline
  \end{tabular}
  }
  \caption{Performance of Fine-Tuning over Various Layer Ranges of Qwen2-7B-Instruct}
  \label{tab:ablation3}
\end{table}

\section{Limitation}
This paper examines only two relatively small-scale open-source models with fewer layers. Although these models have a certain degree of representativeness due to belonging to different model families, further research is required to determine whether the proposed method is applicable to larger-scale models, models with different architectures, and even closed-source models.

\section{Results of Visualization of Hidden States}

In this section, we examine the internal representations of transformer-based models by visualizing the hidden states derived from different language versions of the same dataset. To facilitate this analysis, we employ three widely used dimensionality reduction techniques: Principal Component Analysis (PCA) \citep{hotelling1933analysis}, t-distributed Stochastic Neighbor Embedding (t-SNE) \citep{van2008visualizing}, and Uniform Manifold Approximation and Projection (UMAP) \citep{mcinnes2018umap}. For a controlled comparison, we utilize a single dataset with multiple languages, ensuring that the semantic content remains consistent across all versions. The model processes these multilingual inputs and generates corresponding hidden state vectors.

The visualizations produced by PCA, t-SNE, and UMAP (shown in the following figure) provide a robust basis for analyzing the hidden state representations across different languages. Our results reveal a consistent pattern in the evolution of hidden states across the layers of the language model. Specifically, in the input-adjacent and output-adjacent layers, internal representations tend to form clusters based on language. In contrast, in the intermediate layers, hidden states from different languages become increasingly intermixed, exhibiting a clear phenomenon of language blending. This suggests that while early and late layers are more language-specific, intermediate layers abstract away from surface linguistic forms and encode more universal semantic representations.

\captionsetup[subfigure]{labelformat=empty}

\begin{figure*}[htbp]
\centering
\begin{subfigure}{0.18\textwidth}
\includegraphics[width=\textwidth]{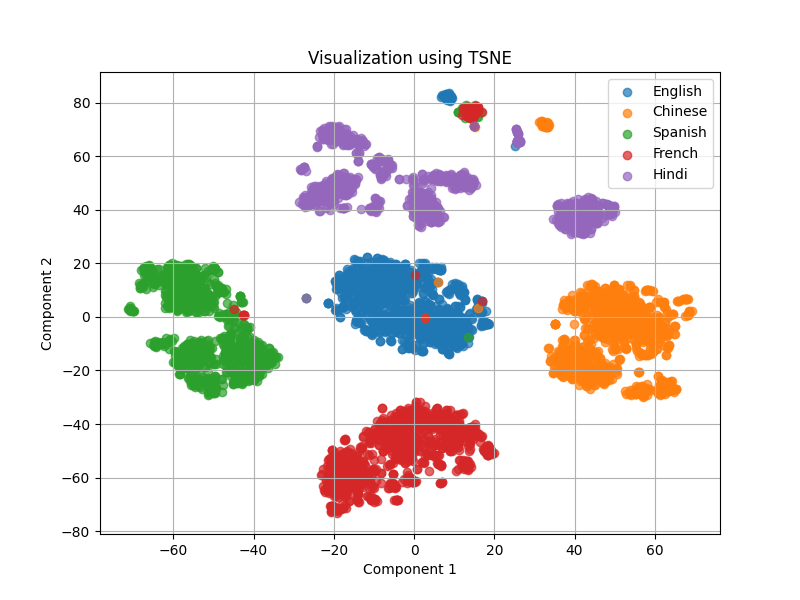}
\caption{T-SNE, Layer 1}
\end{subfigure}
\hfill
\begin{subfigure}{0.18\textwidth}
\includegraphics[width=\textwidth]{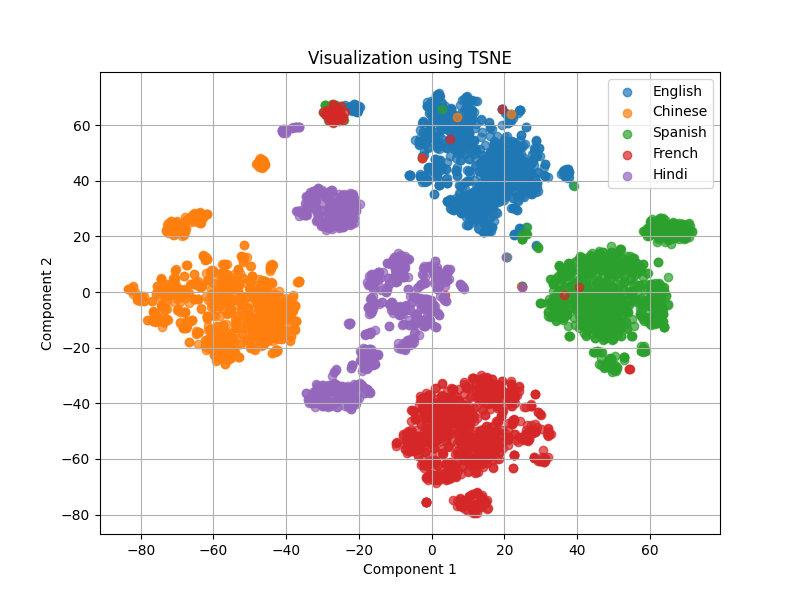}
\caption{T-SNE, Layer 2}

\end{subfigure}
\hfill
\begin{subfigure}{0.18\textwidth}
\includegraphics[width=\textwidth]{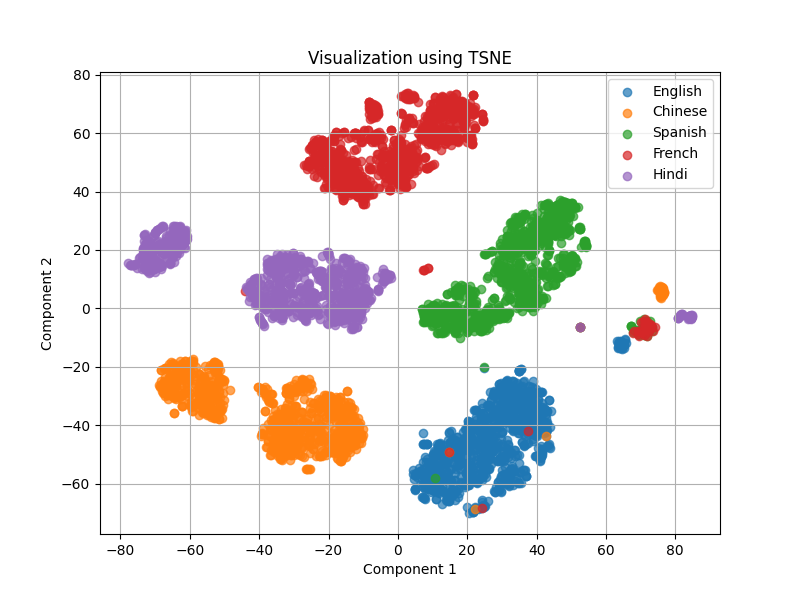}
\caption{T-SNE, Layer 3}

\end{subfigure}
\hfill
\begin{subfigure}{0.18\textwidth}
\includegraphics[width=\textwidth]{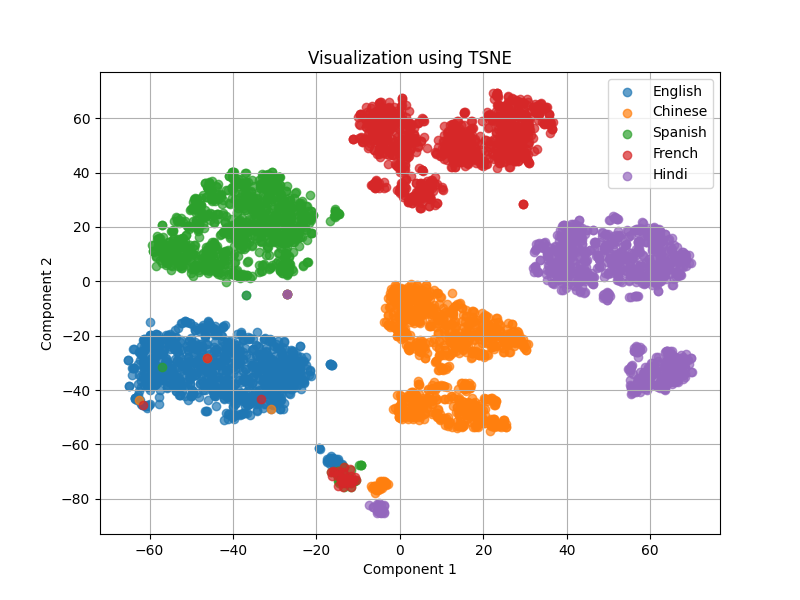}
\caption{T-SNE, Layer 4}

\end{subfigure}
\hfill
\begin{subfigure}{0.18\textwidth}
\includegraphics[width=\textwidth]{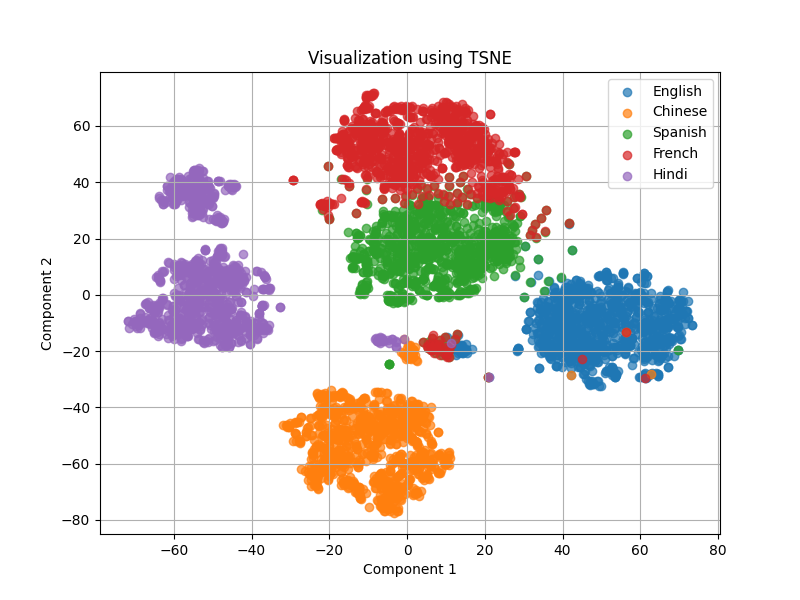}
\caption{T-SNE, Layer 5}

\end{subfigure}
\vspace{0.2in} %
\begin{subfigure}{0.18\textwidth}      \includegraphics[width=\textwidth]{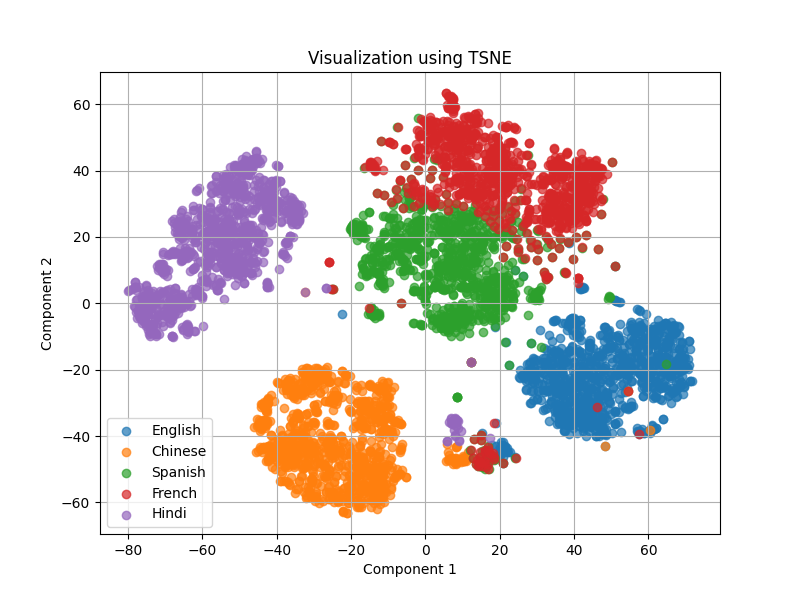}      \caption{T-SNE, Layer 6}        \end{subfigure}  \hfill  \begin{subfigure}{0.18\textwidth}      \includegraphics[width=\textwidth]{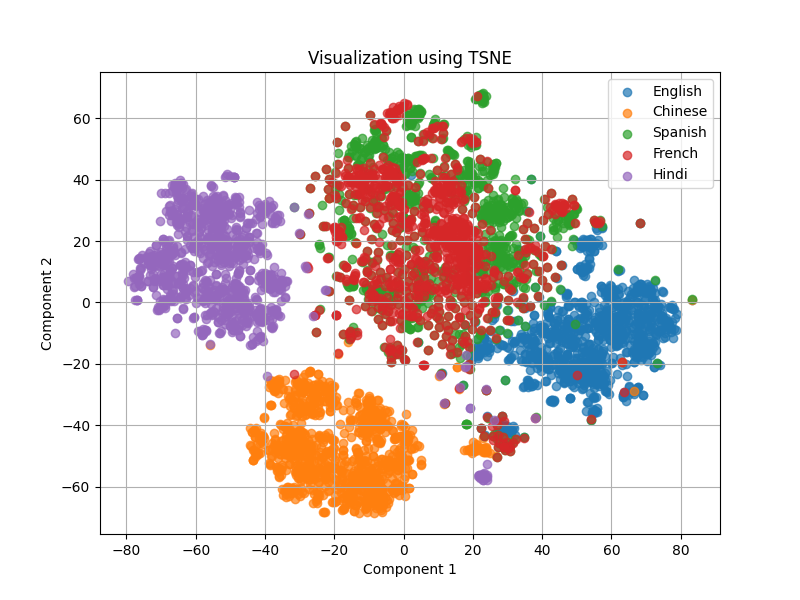}      \caption{T-SNE, Layer 7}        \end{subfigure}  \hfill  \begin{subfigure}{0.18\textwidth}      \includegraphics[width=\textwidth]{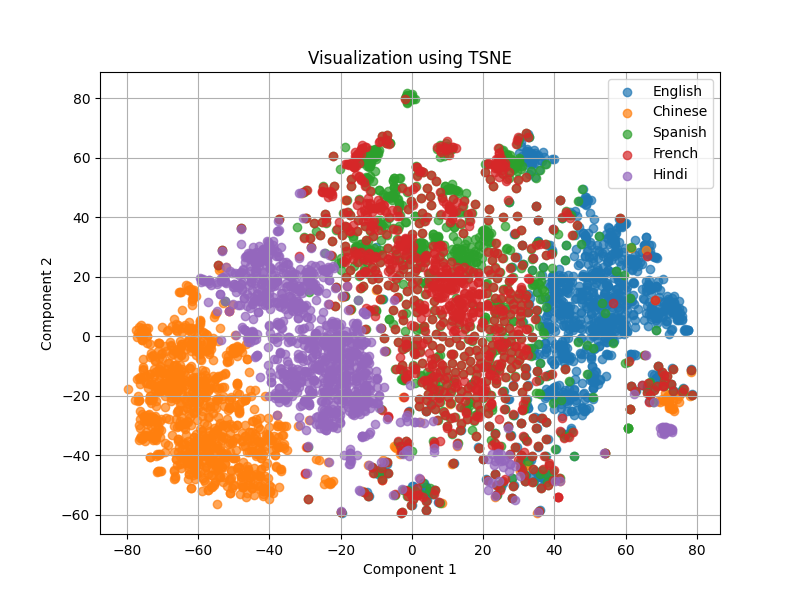}      \caption{T-SNE, Layer 8}        \end{subfigure}  \hfill  \begin{subfigure}{0.18\textwidth}      \includegraphics[width=\textwidth]{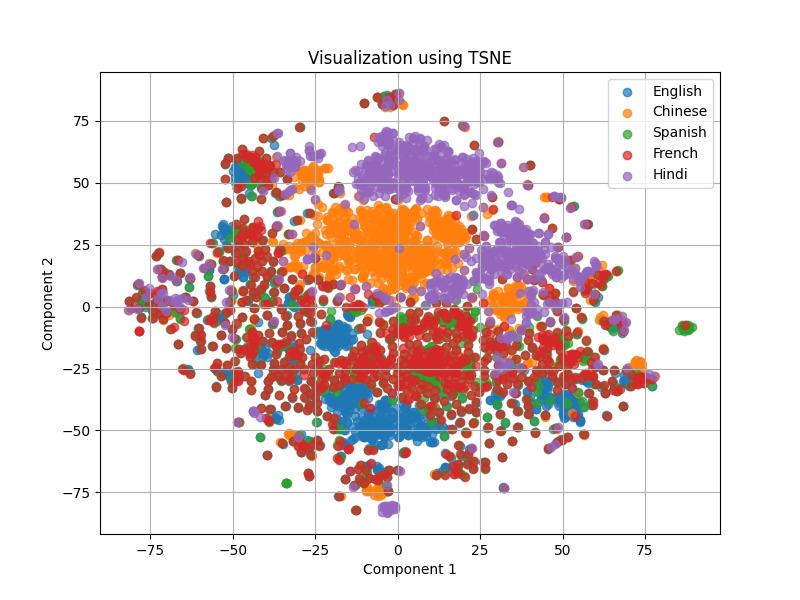}      \caption{T-SNE, Layer 9}        \end{subfigure}  \hfill  \begin{subfigure}{0.18\textwidth}      \includegraphics[width=\textwidth]{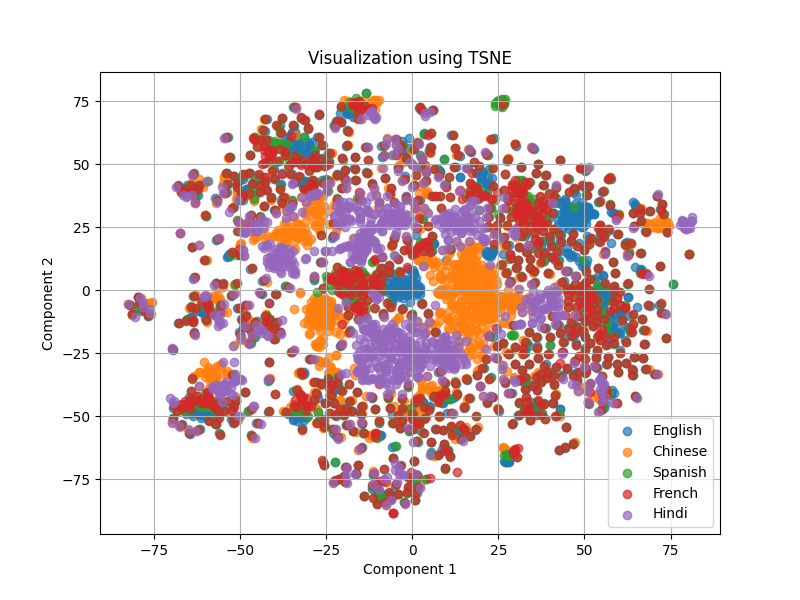}      \caption{T-SNE, Layer 10}        \end{subfigure}    \vspace{0.2in}    %
\begin{subfigure}{0.18\textwidth}      \includegraphics[width=\textwidth]{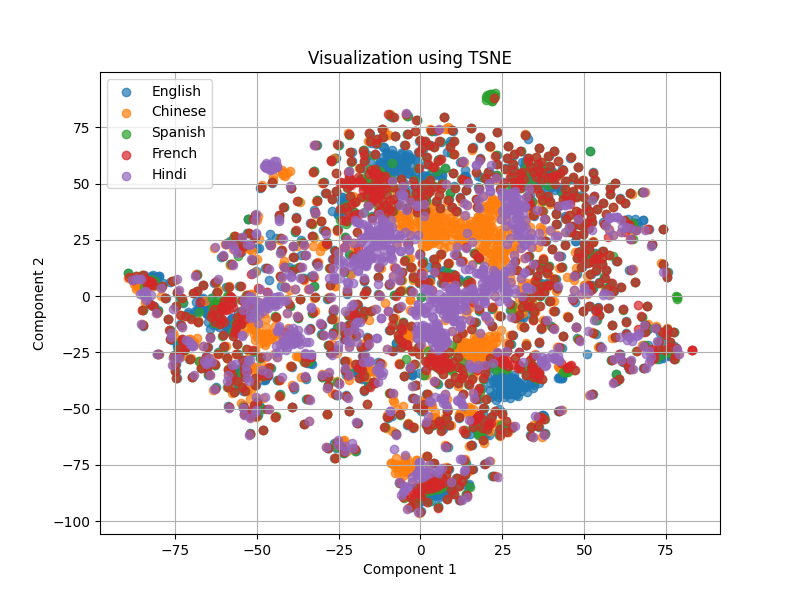}      \caption{T-SNE, Layer 11}        \end{subfigure}  \hfill  \begin{subfigure}{0.18\textwidth}      \includegraphics[width=\textwidth]{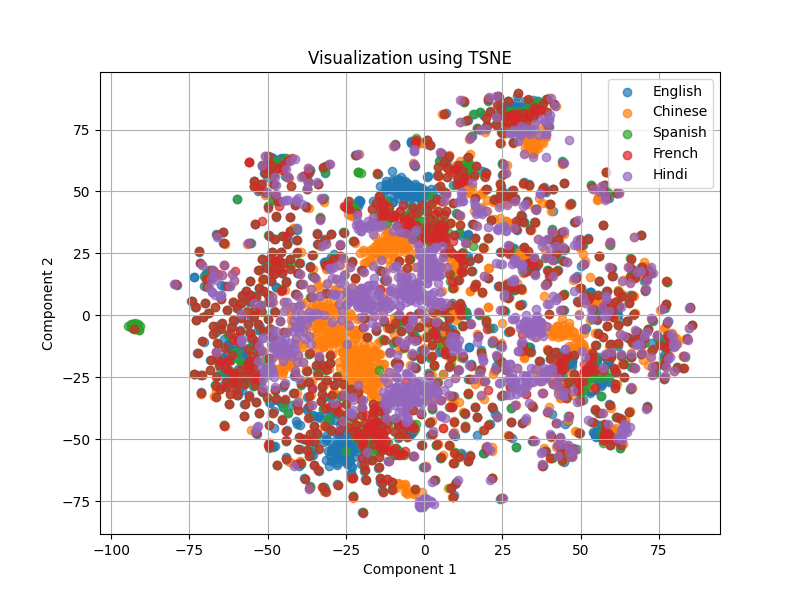}      \caption{T-SNE, Layer 12}        \end{subfigure}  \hfill  \begin{subfigure}{0.18\textwidth}      \includegraphics[width=\textwidth]{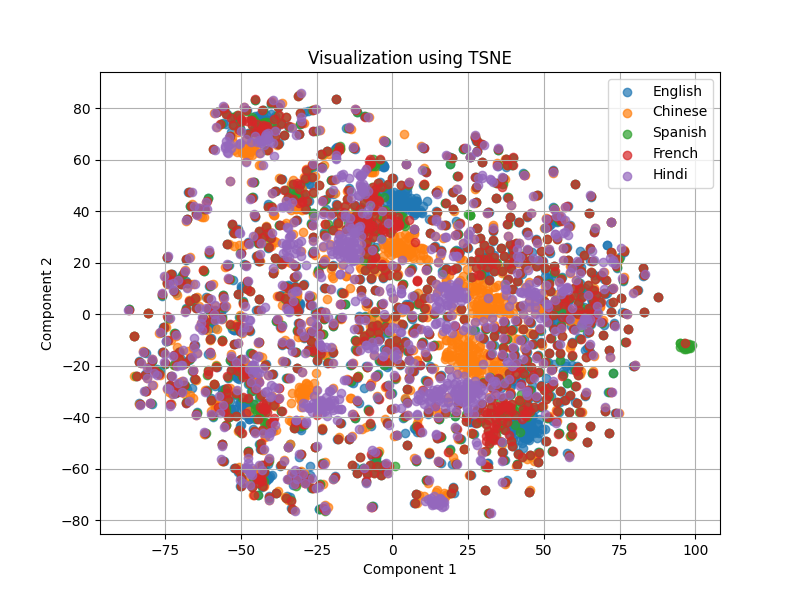}      \caption{T-SNE, Layer 13}        \end{subfigure}  \hfill  \begin{subfigure}{0.18\textwidth}      \includegraphics[width=\textwidth]{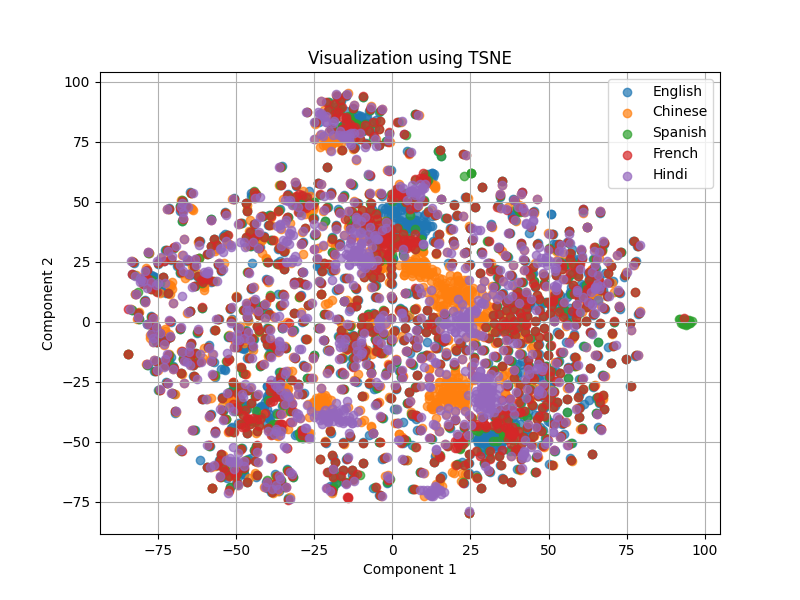}      \caption{T-SNE, Layer 14}        \end{subfigure}  \hfill  \begin{subfigure}{0.18\textwidth}      \includegraphics[width=\textwidth]{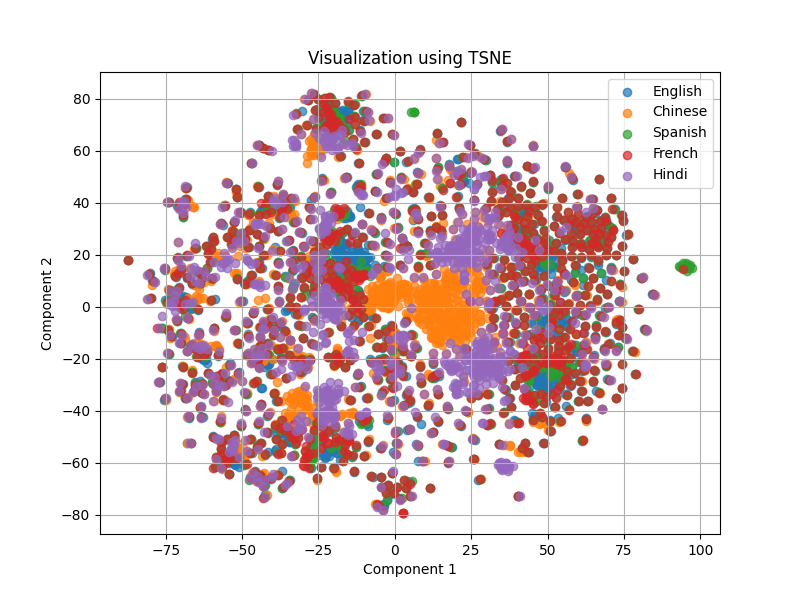}      \caption{T-SNE, Layer 15}        \end{subfigure}    \vspace{0.2in}    %
\begin{subfigure}{0.18\textwidth}      \includegraphics[width=\textwidth]{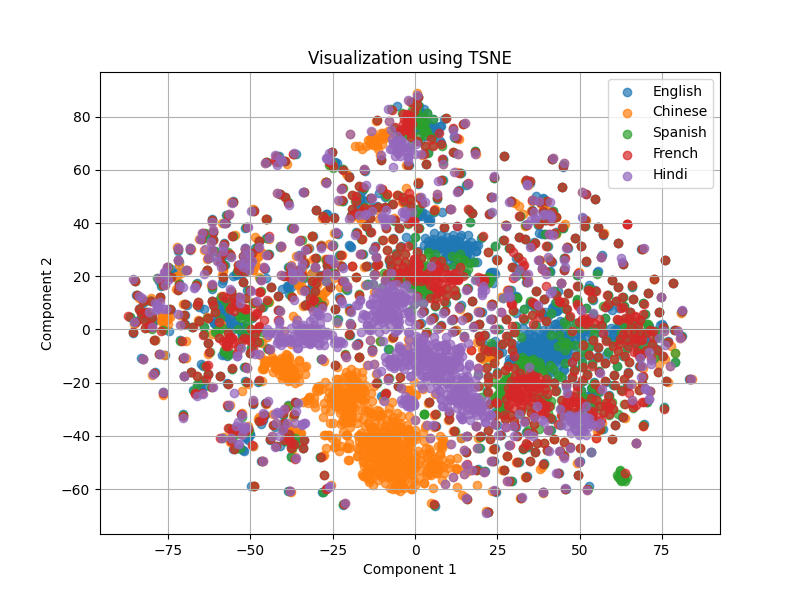}      \caption{T-SNE, Layer 16}        \end{subfigure}  \hfill  \begin{subfigure}{0.18\textwidth}      \includegraphics[width=\textwidth]{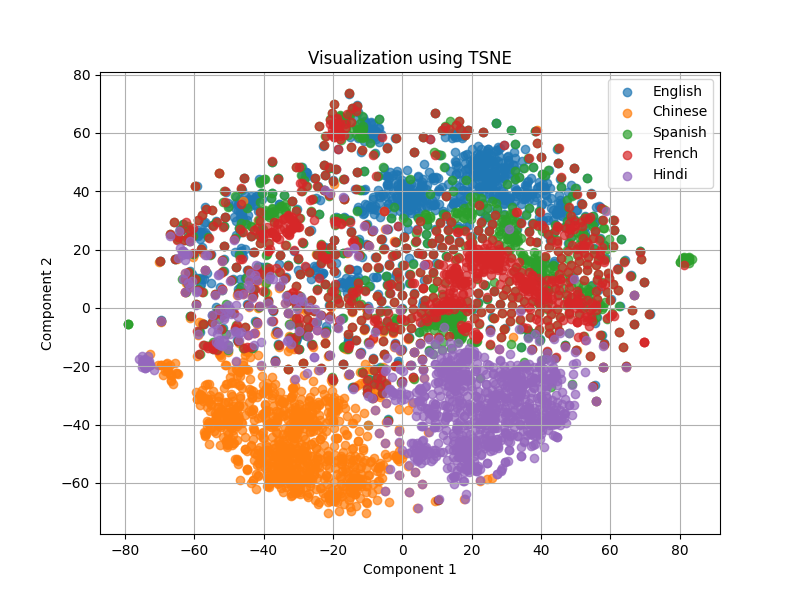}      \caption{T-SNE, Layer 17}        \end{subfigure}  \hfill  \begin{subfigure}{0.18\textwidth}      \includegraphics[width=\textwidth]{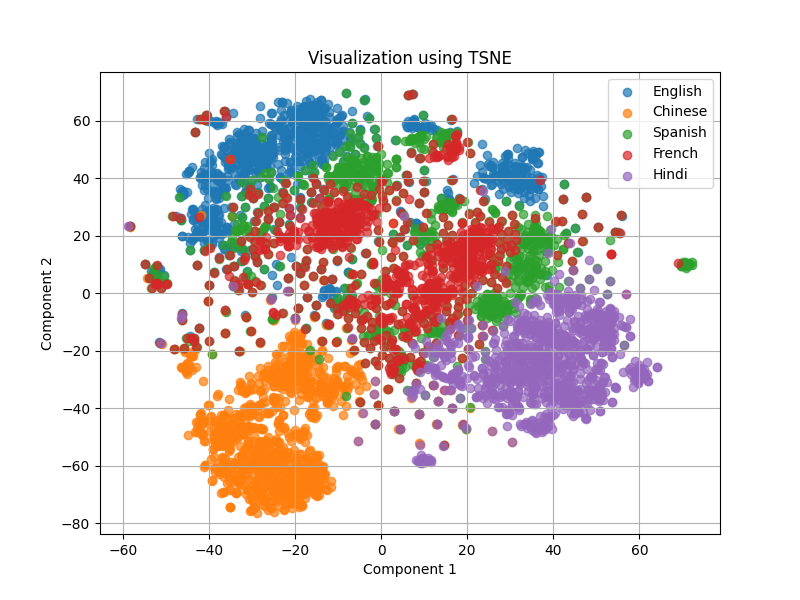}      \caption{T-SNE, Layer 18}        \end{subfigure}  \hfill  \begin{subfigure}{0.18\textwidth}      \includegraphics[width=\textwidth]{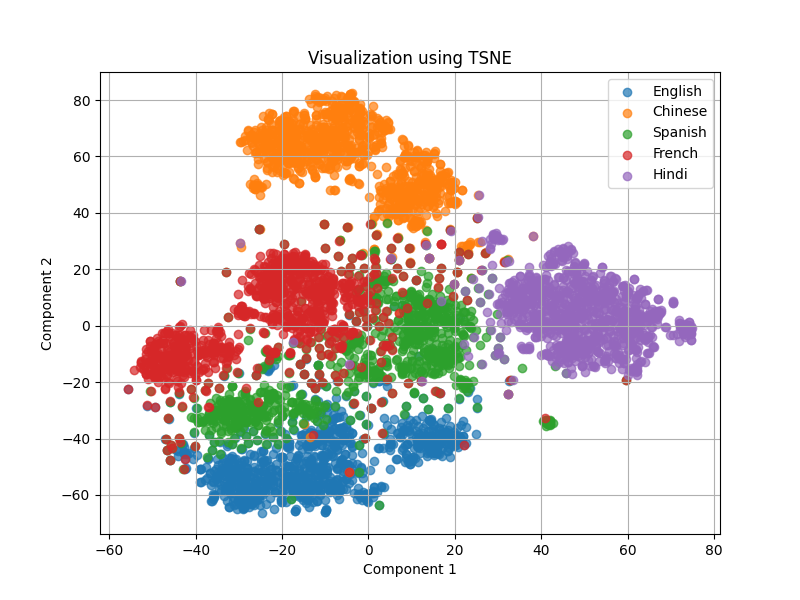}      \caption{T-SNE, Layer 19}        \end{subfigure}  \hfill  \begin{subfigure}{0.18\textwidth}      \includegraphics[width=\textwidth]{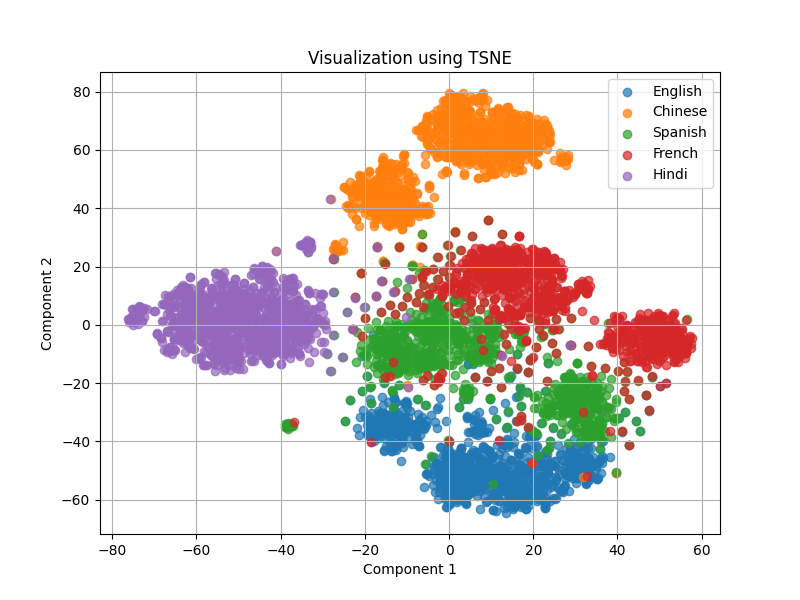}      \caption{T-SNE, Layer 20}        \end{subfigure}    \vspace{0.2in}    %
\begin{subfigure}{0.18\textwidth}      \includegraphics[width=\textwidth]{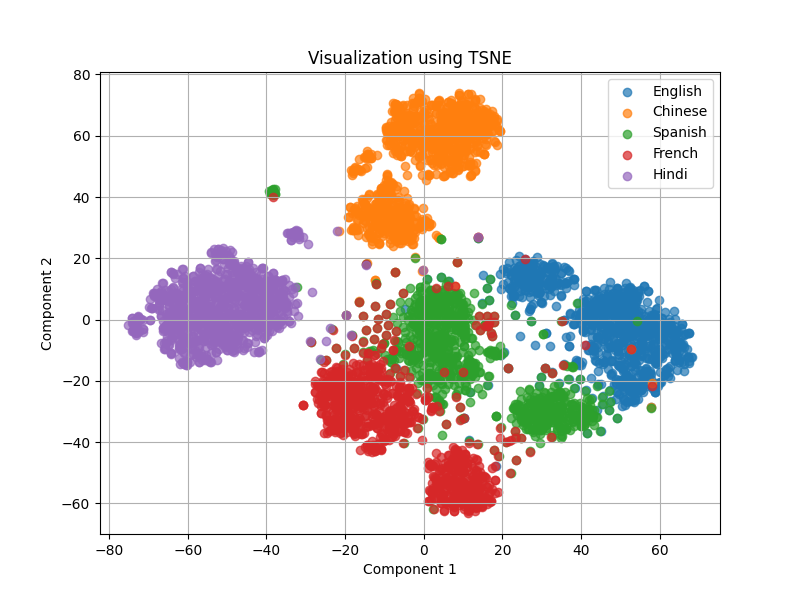}      \caption{T-SNE, Layer 21}        \end{subfigure}  \hfill  \begin{subfigure}{0.18\textwidth}      \includegraphics[width=\textwidth]{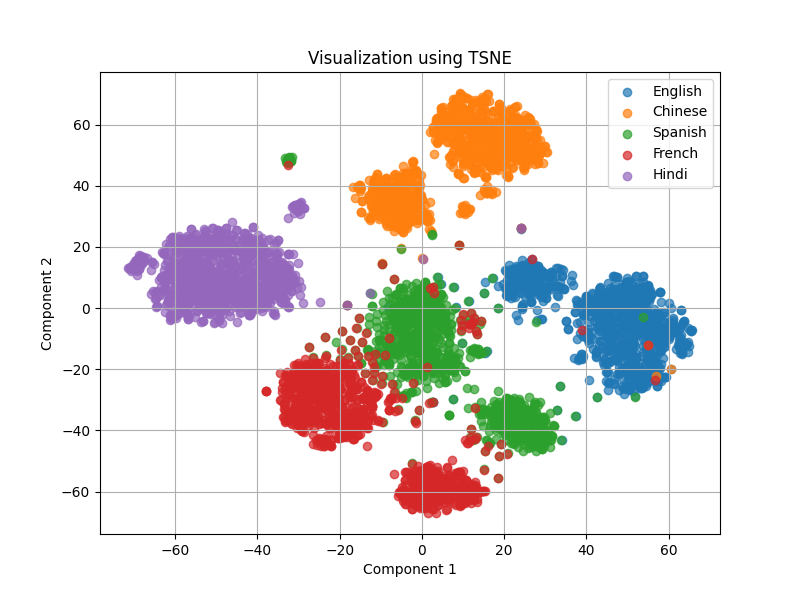}      \caption{T-SNE, Layer 22}        \end{subfigure}  \hfill  \begin{subfigure}{0.18\textwidth}      \includegraphics[width=\textwidth]{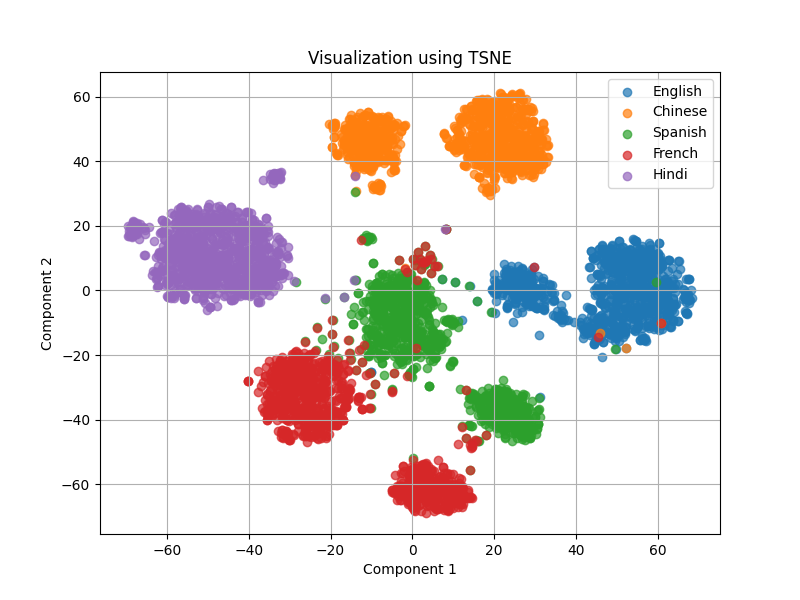}      \caption{T-SNE, Layer 23}        \end{subfigure}  \hfill  \begin{subfigure}{0.18\textwidth}      \includegraphics[width=\textwidth]{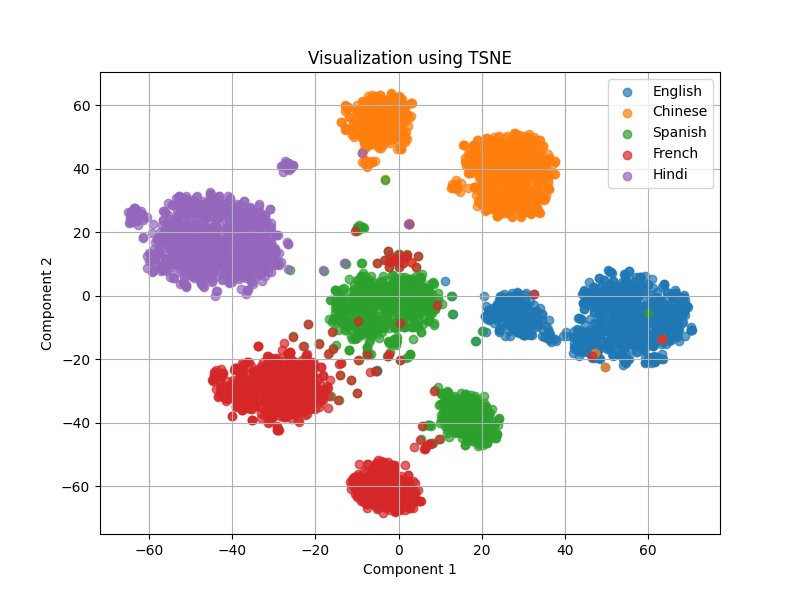}      \caption{T-SNE, Layer 24}        \end{subfigure}  \hfill  \begin{subfigure}{0.18\textwidth}      \includegraphics[width=\textwidth]{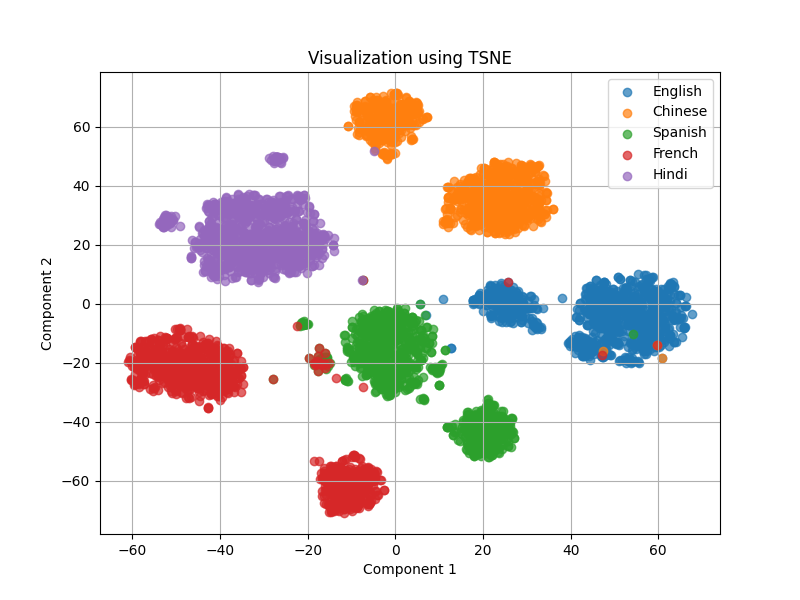}      \caption{T-SNE, Layer 25}        \end{subfigure}    \vspace{0.2in}    %
\begin{subfigure}{0.18\textwidth}      \includegraphics[width=\textwidth]{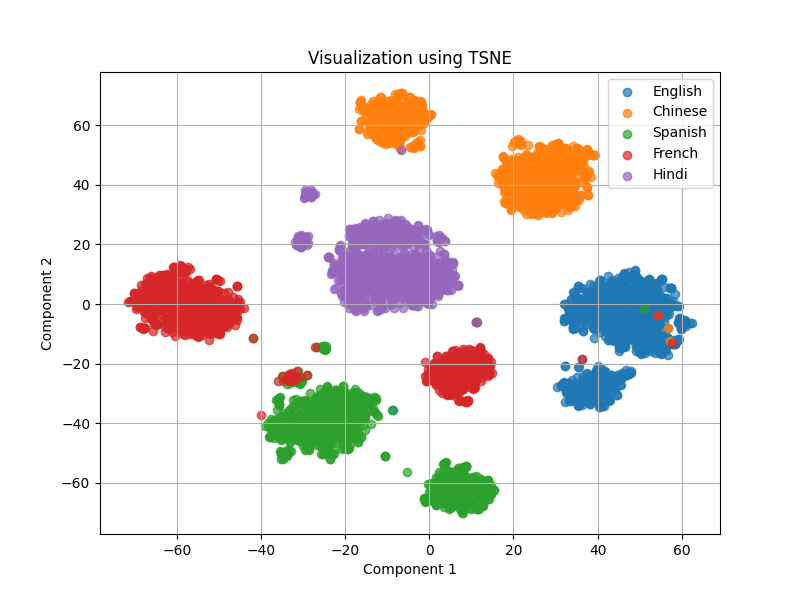}      \caption{T-SNE, Layer 26}        \end{subfigure}  \hfill  \begin{subfigure}{0.18\textwidth}      \includegraphics[width=\textwidth]{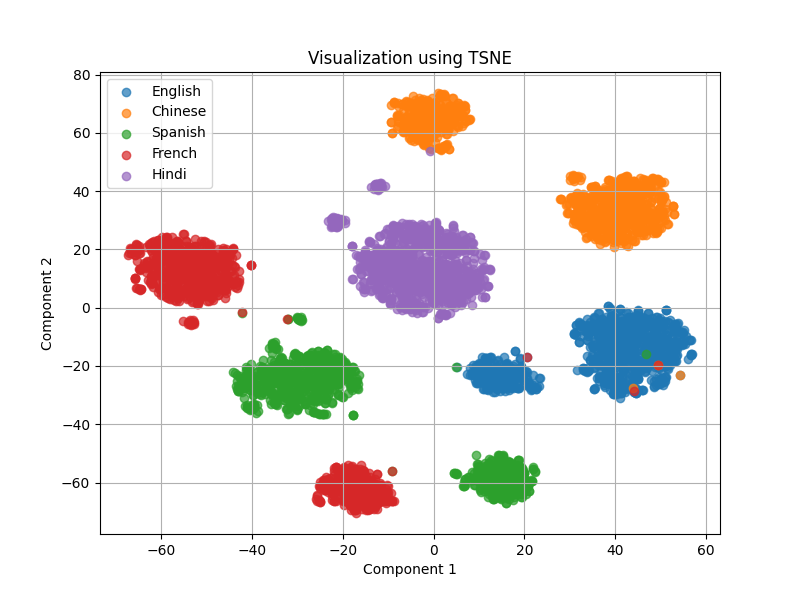}      \caption{T-SNE, Layer 27}        \end{subfigure}  \hfill  \begin{subfigure}{0.18\textwidth}      \includegraphics[width=\textwidth]{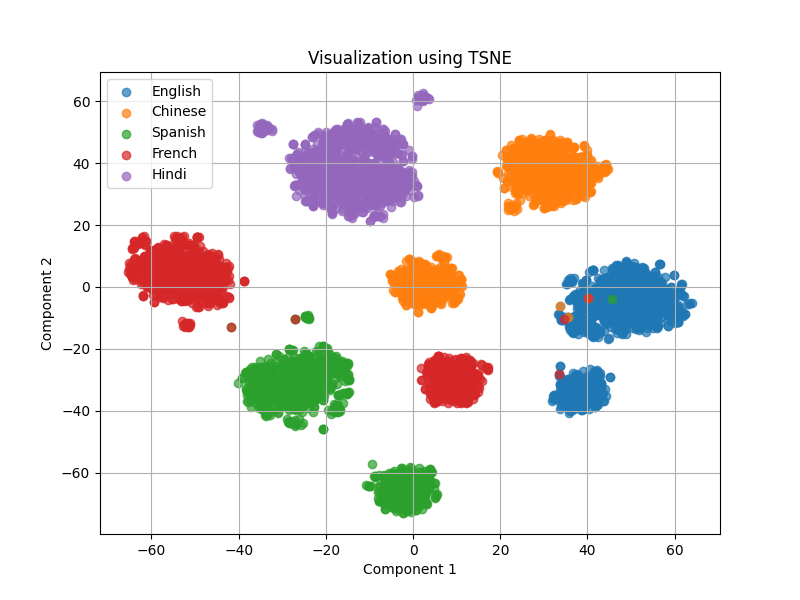}      \caption{T-SNE, Layer 28}        \end{subfigure}  \hfill  \begin{subfigure}{0.18\textwidth}      \includegraphics[width=\textwidth]{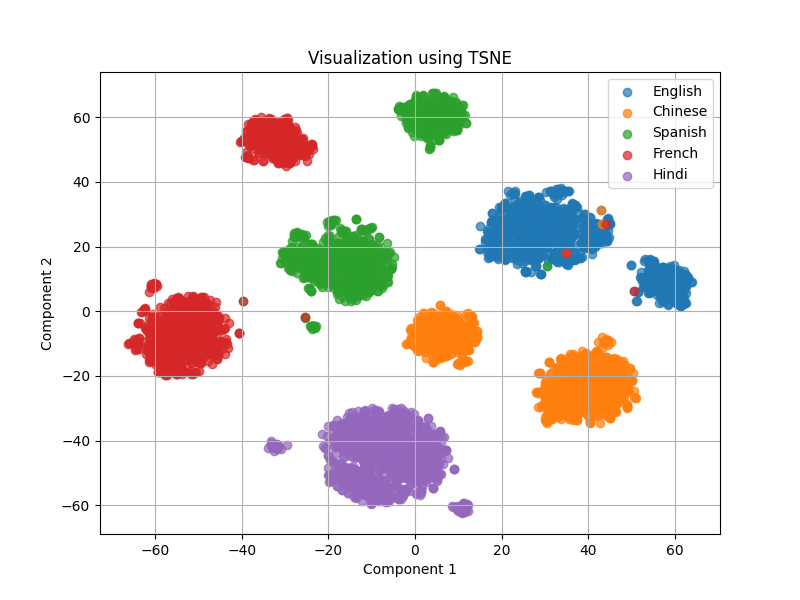}      \caption{T-SNE, Layer 29}        \end{subfigure}  \hfill  \begin{subfigure}{0.18\textwidth}      \includegraphics[width=\textwidth]{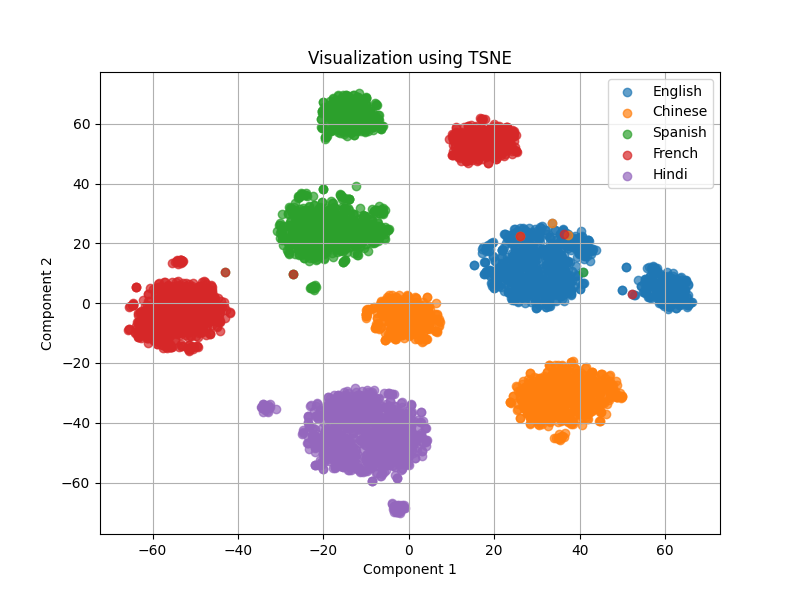}      \caption{T-SNE, Layer 30}        \end{subfigure}    \vspace{0.2in}    %
\begin{subfigure}{0.18\textwidth}      \includegraphics[width=\textwidth]{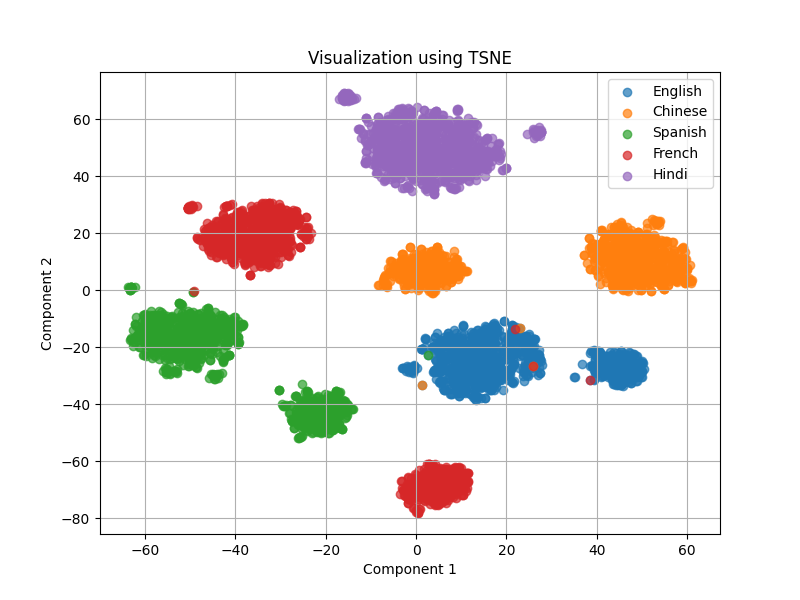}      \caption{T-SNE, Layer 31}        \end{subfigure}  \hfill  \begin{subfigure}{0.18\textwidth}      \includegraphics[width=\textwidth]{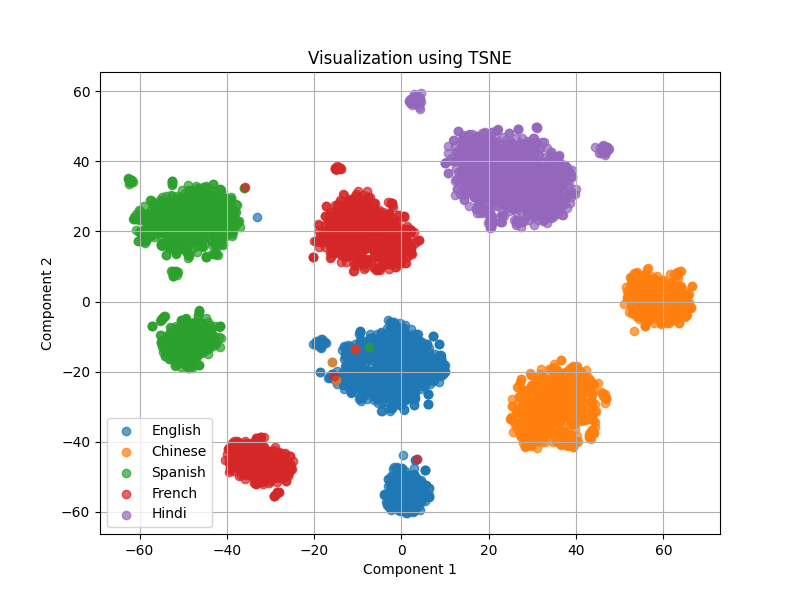}      \caption{T-SNE, Layer 32}        \end{subfigure}    \caption{T-SNE visualizations for layers 1-32 of Llama-3-8B-Instruct on the GSM8K dataset.}  
\end{figure*}

\begin{figure*}[htbp]
\centering
\begin{subfigure}{0.18\textwidth}
\includegraphics[width=\textwidth]{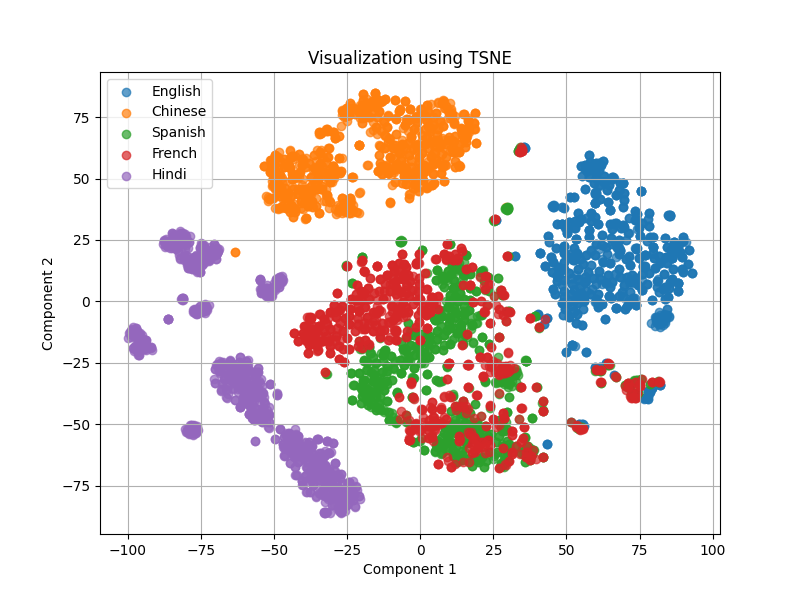}
\caption{T-SNE, Layer 1}
\end{subfigure}
\hfill
\begin{subfigure}{0.18\textwidth}
\includegraphics[width=\textwidth]{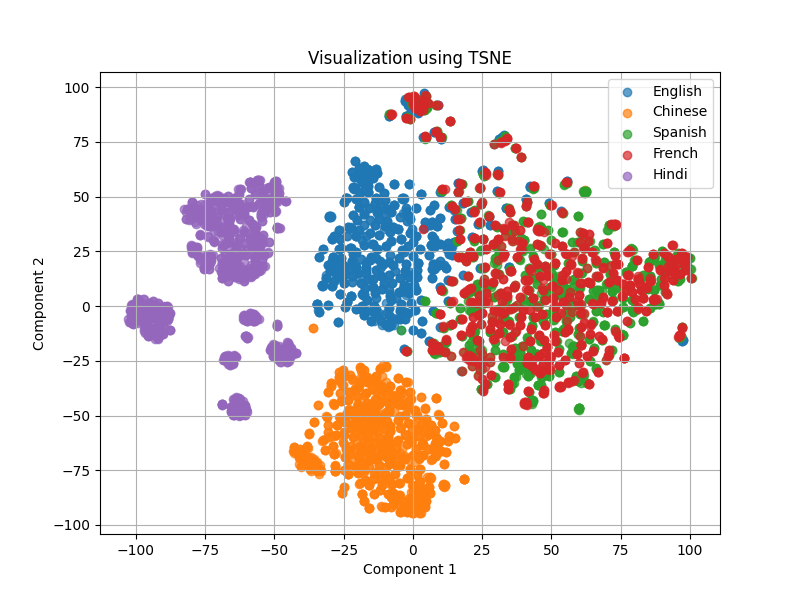}
\caption{T-SNE, Layer 2}

\end{subfigure}
\hfill
\begin{subfigure}{0.18\textwidth}
\includegraphics[width=\textwidth]{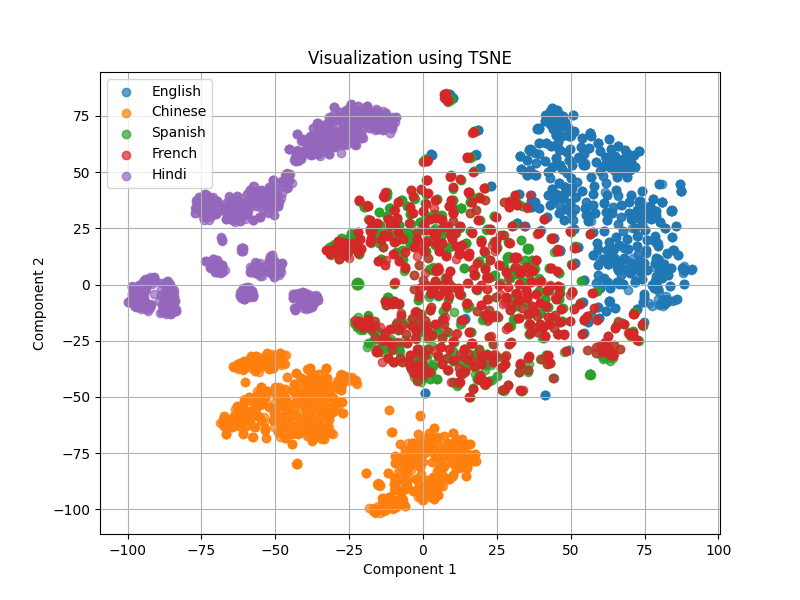}
\caption{T-SNE, Layer 3}

\end{subfigure}
\hfill
\begin{subfigure}{0.18\textwidth}
\includegraphics[width=\textwidth]{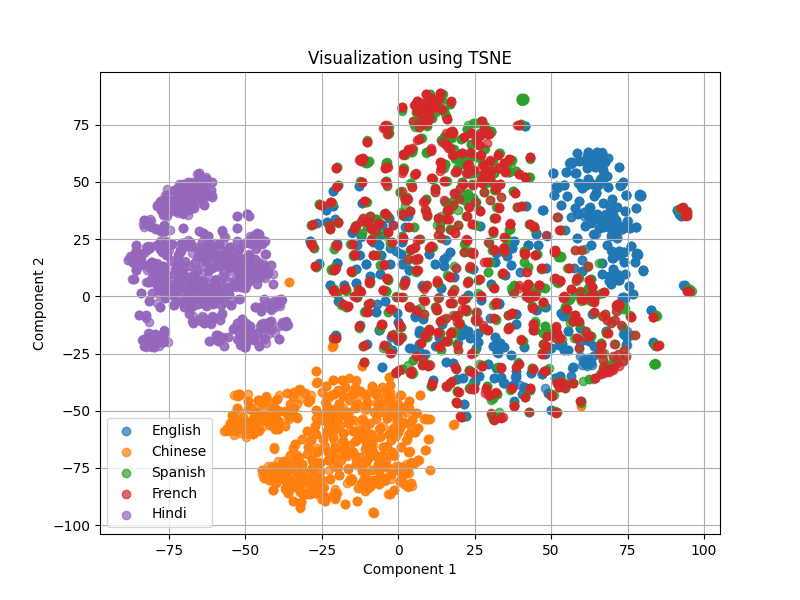}
\caption{T-SNE, Layer 4}

\end{subfigure}
\hfill
\begin{subfigure}{0.18\textwidth}
\includegraphics[width=\textwidth]{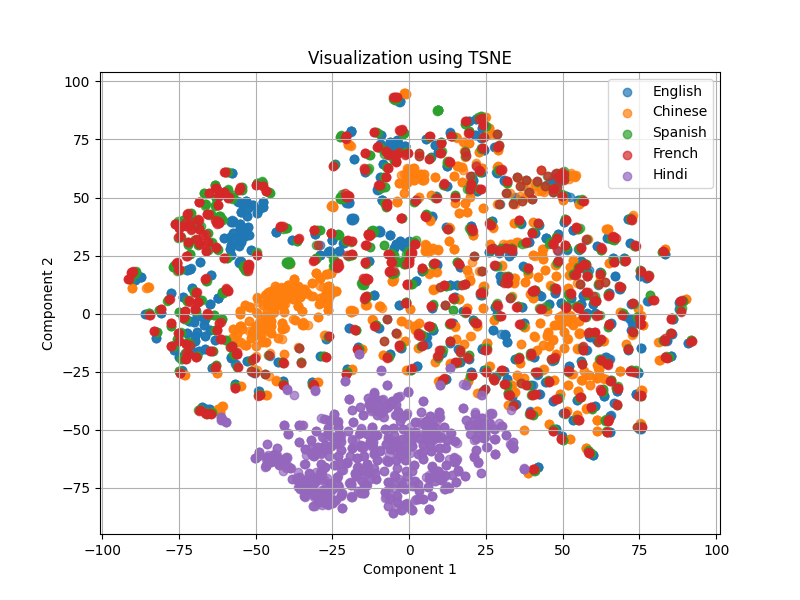}
\caption{T-SNE, Layer 5}

\end{subfigure}
\vspace{0.2in} %
\begin{subfigure}{0.18\textwidth}      \includegraphics[width=\textwidth]{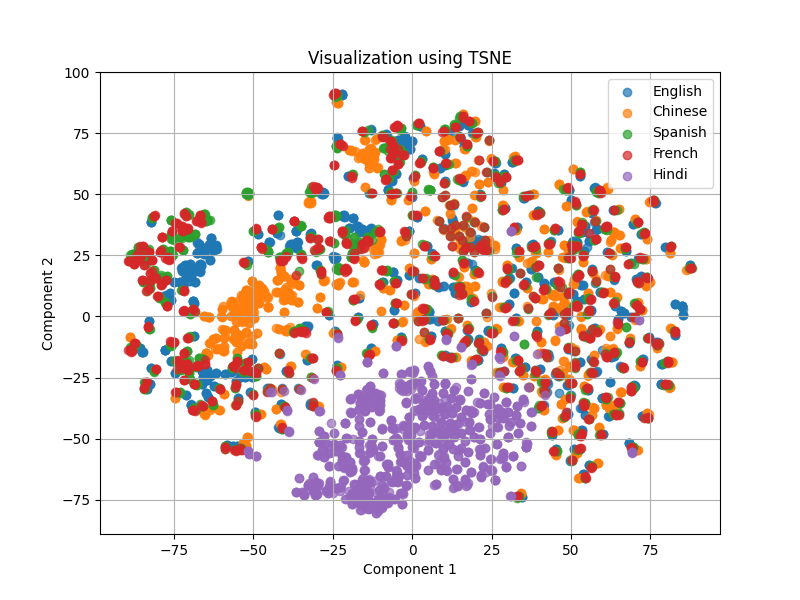}      \caption{T-SNE, Layer 6}        \end{subfigure}  \hfill  \begin{subfigure}{0.18\textwidth}      \includegraphics[width=\textwidth]{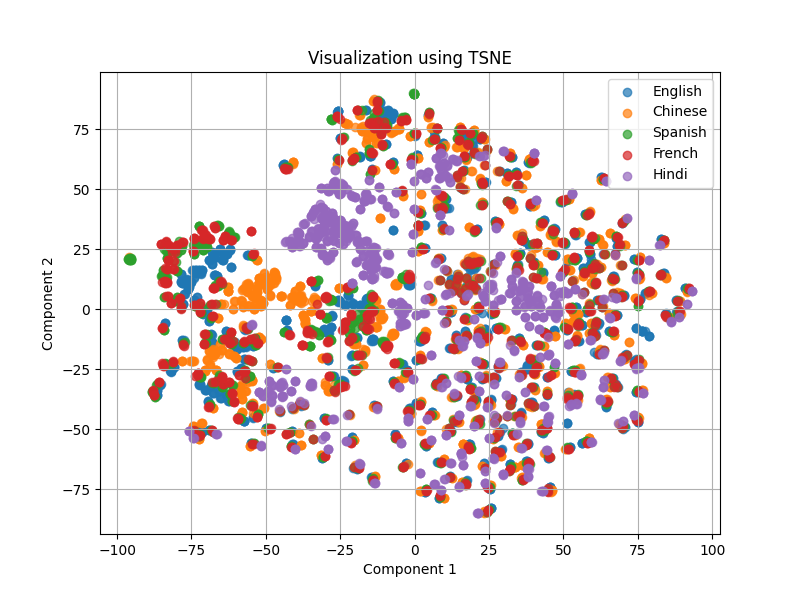}      \caption{T-SNE, Layer 7}        \end{subfigure}  \hfill  \begin{subfigure}{0.18\textwidth}      \includegraphics[width=\textwidth]{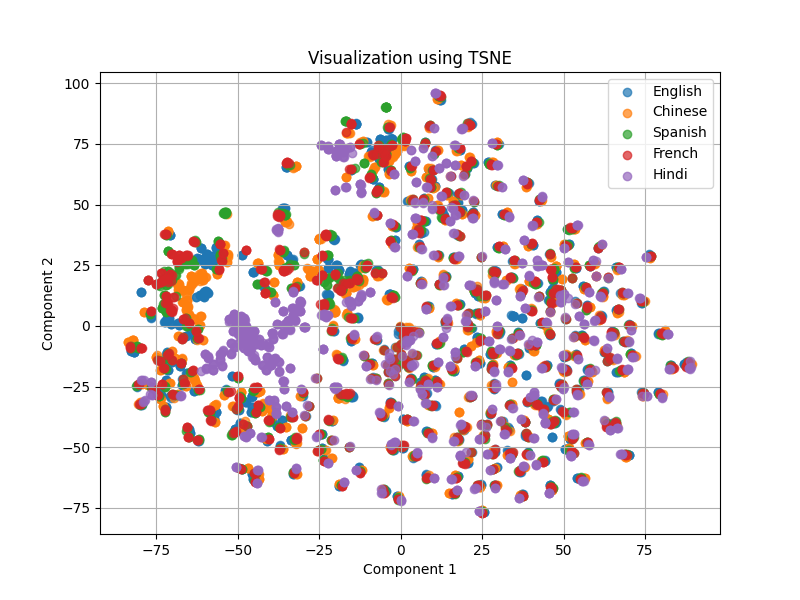}      \caption{T-SNE, Layer 8}        \end{subfigure}  \hfill  \begin{subfigure}{0.18\textwidth}      \includegraphics[width=\textwidth]{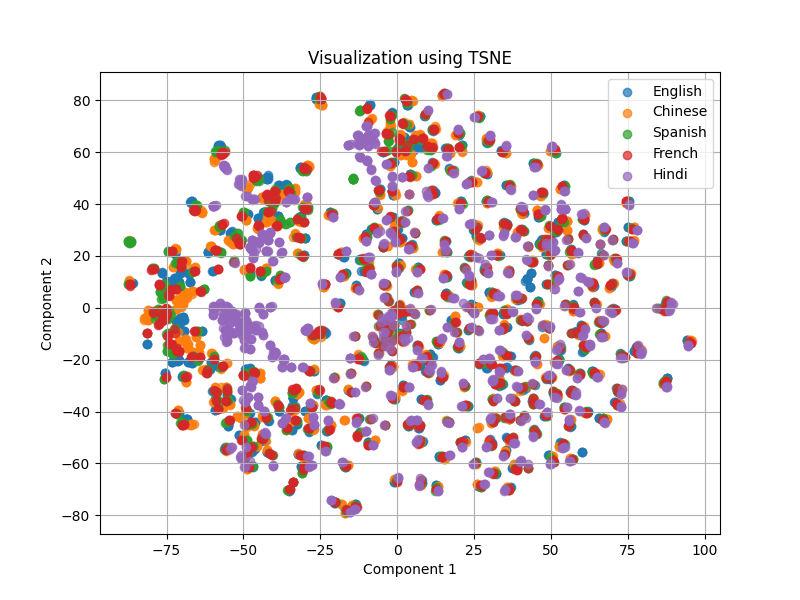}      \caption{T-SNE, Layer 9}        \end{subfigure}  \hfill  \begin{subfigure}{0.18\textwidth}      \includegraphics[width=\textwidth]{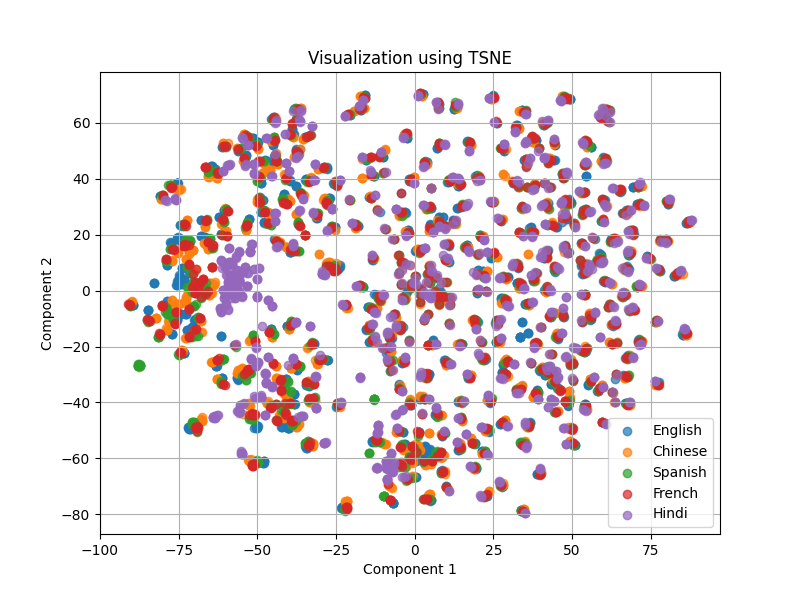}      \caption{T-SNE, Layer 10}        \end{subfigure}    \vspace{0.2in}    %
\begin{subfigure}{0.18\textwidth}      \includegraphics[width=\textwidth]{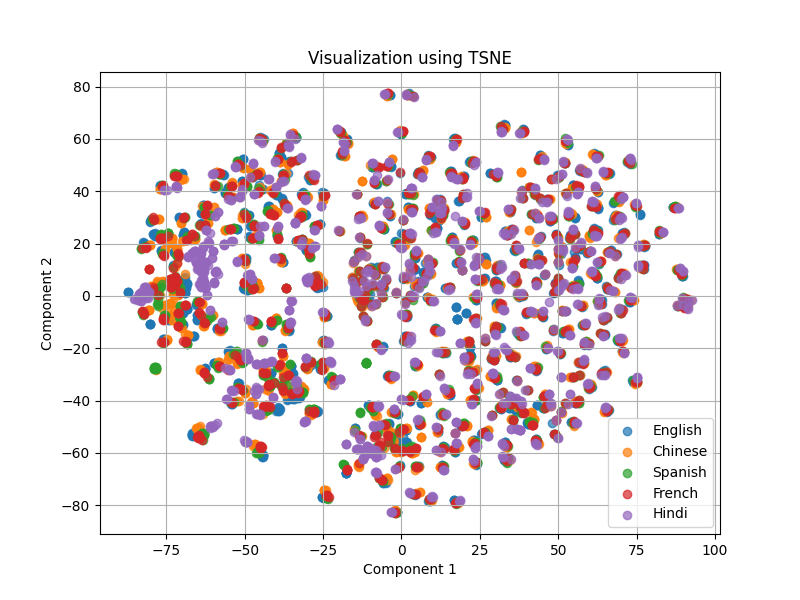}      \caption{T-SNE, Layer 11}        \end{subfigure}  \hfill  \begin{subfigure}{0.18\textwidth}      \includegraphics[width=\textwidth]{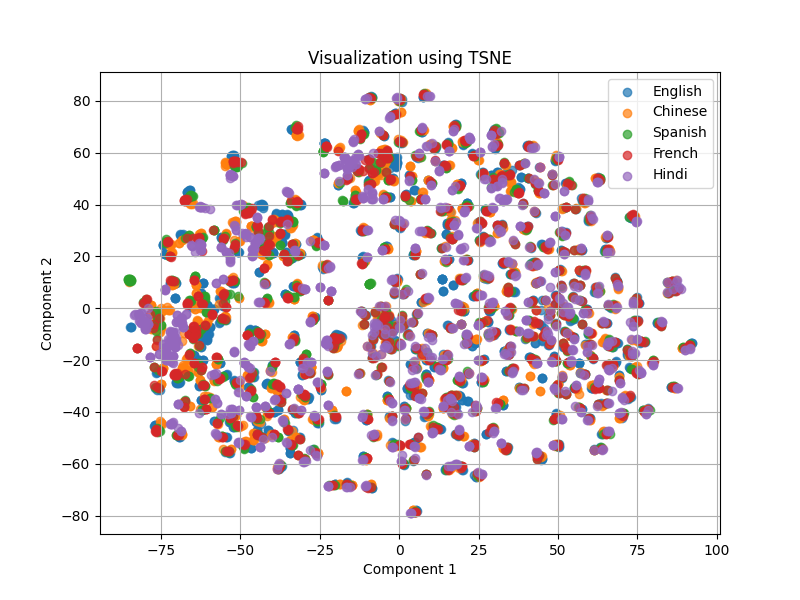}      \caption{T-SNE, Layer 12}        \end{subfigure}  \hfill  \begin{subfigure}{0.18\textwidth}      \includegraphics[width=\textwidth]{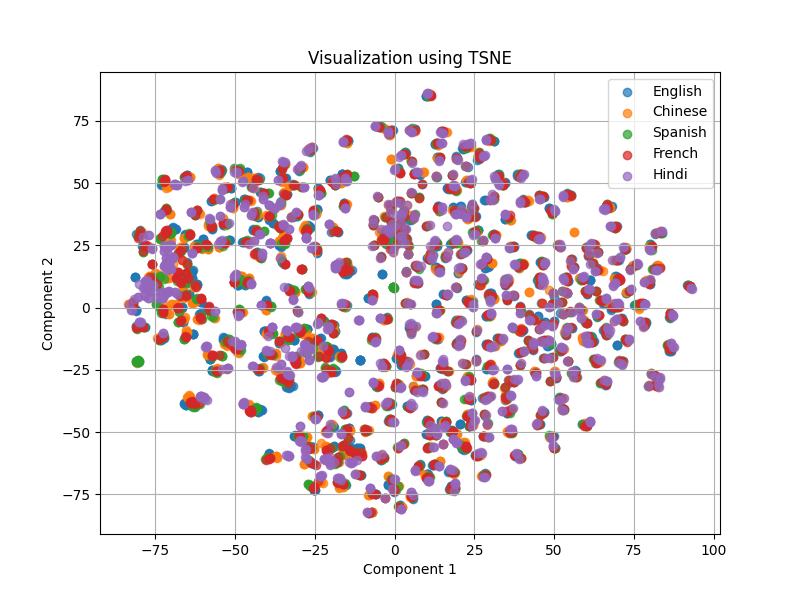}      \caption{T-SNE, Layer 13}        \end{subfigure}  \hfill  \begin{subfigure}{0.18\textwidth}      \includegraphics[width=\textwidth]{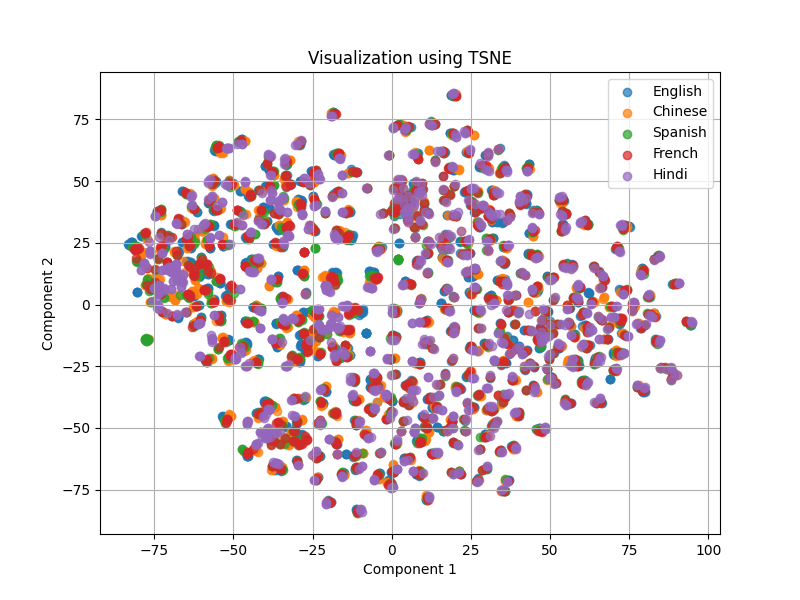}      \caption{T-SNE, Layer 14}        \end{subfigure}  \hfill  \begin{subfigure}{0.18\textwidth}      \includegraphics[width=\textwidth]{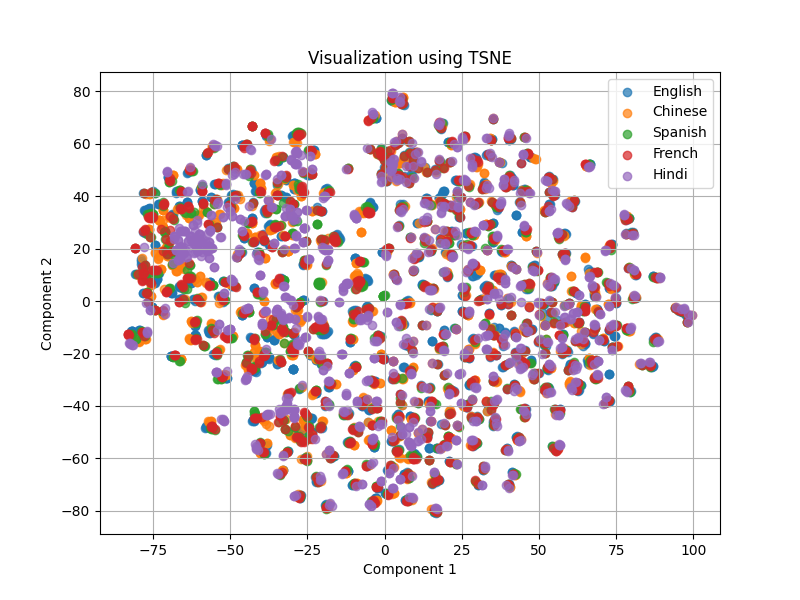}      \caption{T-SNE, Layer 15}        \end{subfigure}    \vspace{0.2in}    %
\begin{subfigure}{0.18\textwidth}      \includegraphics[width=\textwidth]{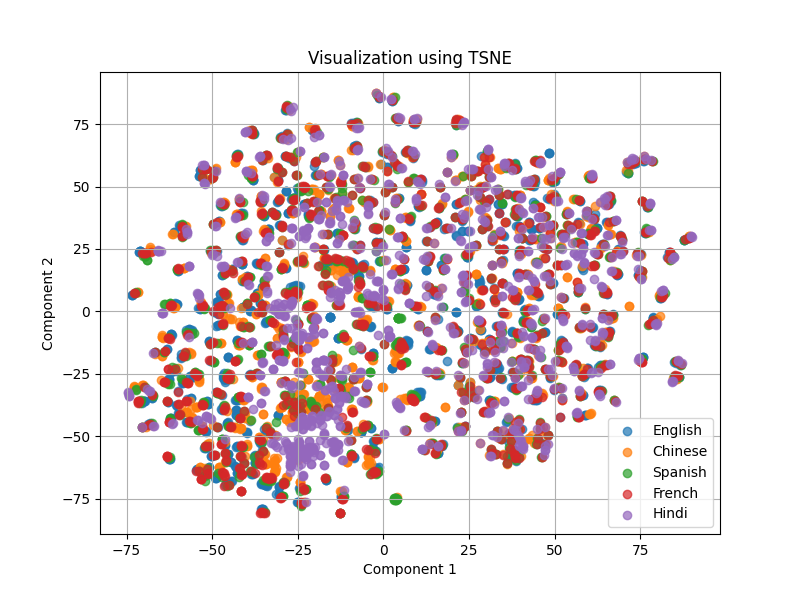}      \caption{T-SNE, Layer 16}        \end{subfigure}  \hfill  \begin{subfigure}{0.18\textwidth}      \includegraphics[width=\textwidth]{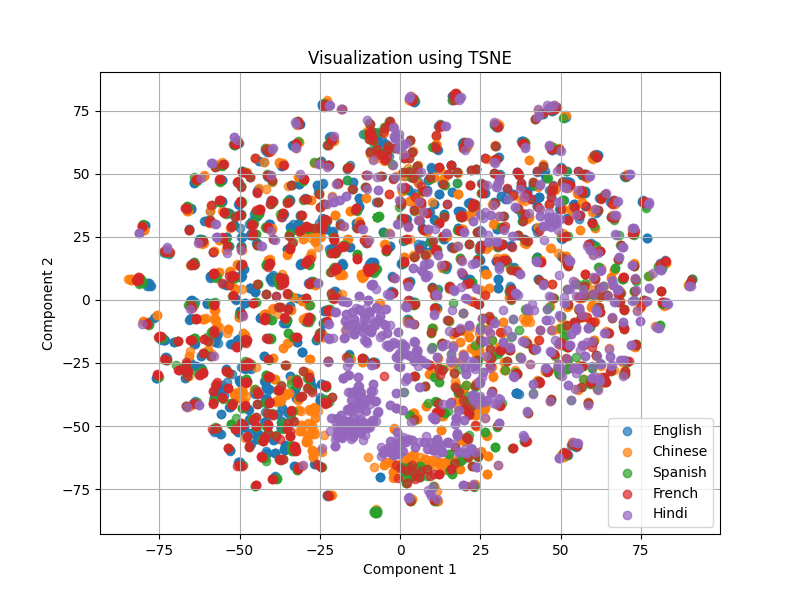}      \caption{T-SNE, Layer 17}        \end{subfigure}  \hfill  \begin{subfigure}{0.18\textwidth}      \includegraphics[width=\textwidth]{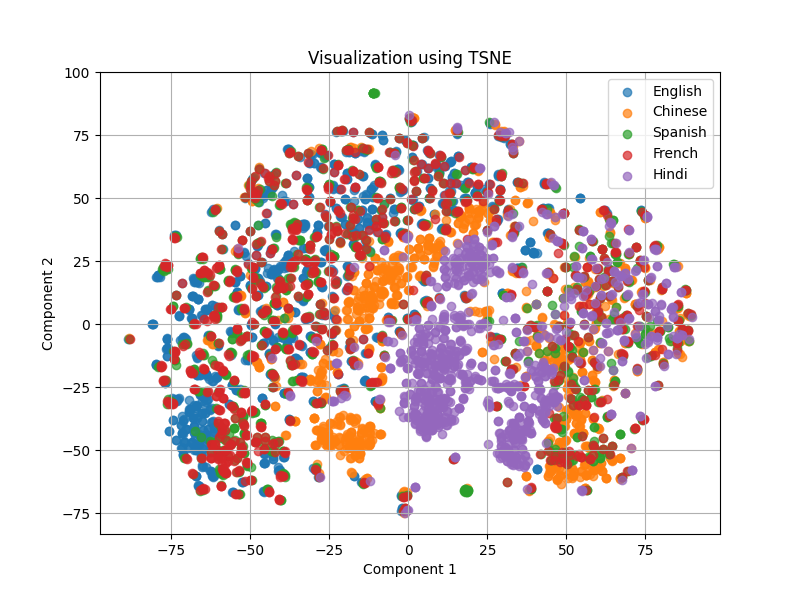}      \caption{T-SNE, Layer 18}        \end{subfigure}  \hfill  \begin{subfigure}{0.18\textwidth}      \includegraphics[width=\textwidth]{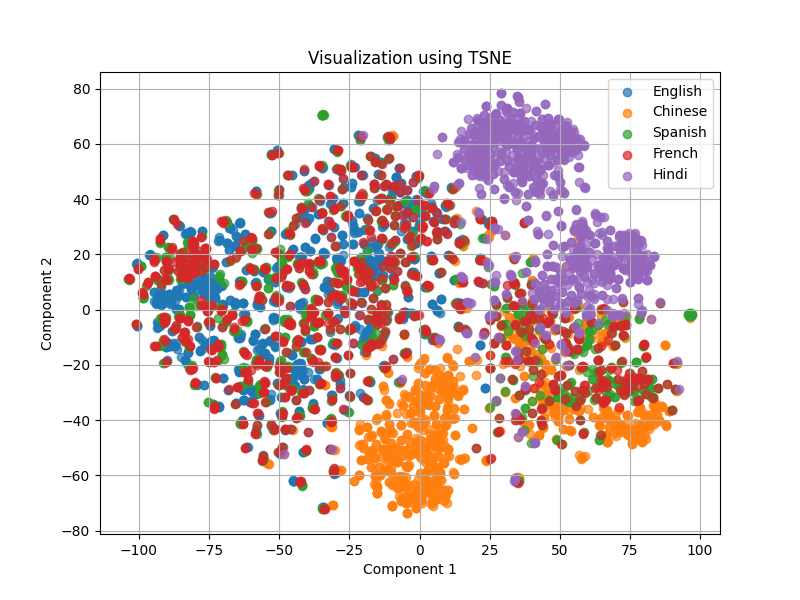}      \caption{T-SNE, Layer 19}        \end{subfigure}  \hfill  \begin{subfigure}{0.18\textwidth}      \includegraphics[width=\textwidth]{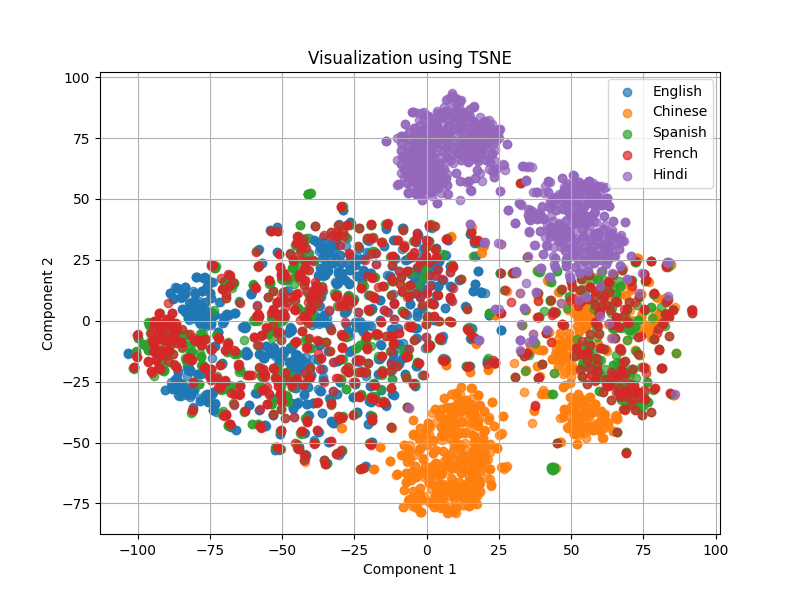}      \caption{T-SNE, Layer 20}        \end{subfigure}    \vspace{0.2in}    %
\begin{subfigure}{0.18\textwidth}      \includegraphics[width=\textwidth]{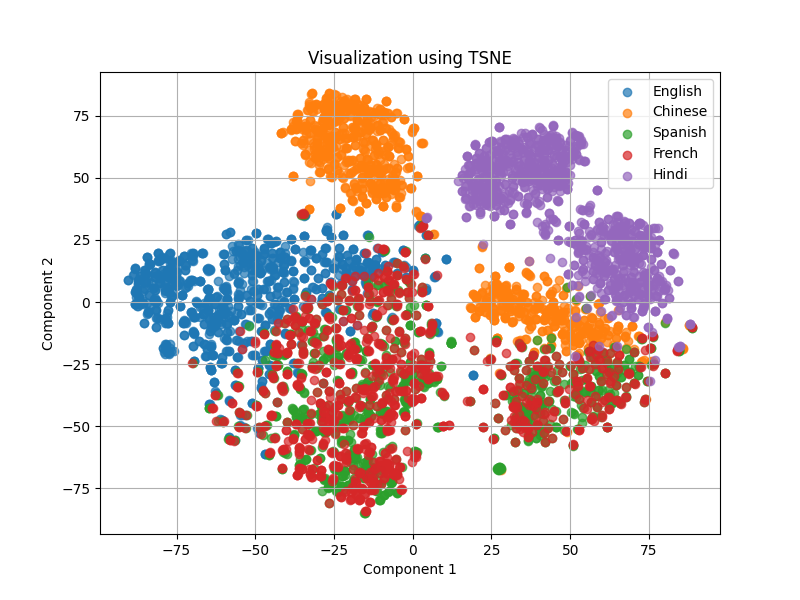}      \caption{T-SNE, Layer 21}        \end{subfigure}  \hfill  \begin{subfigure}{0.18\textwidth}      \includegraphics[width=\textwidth]{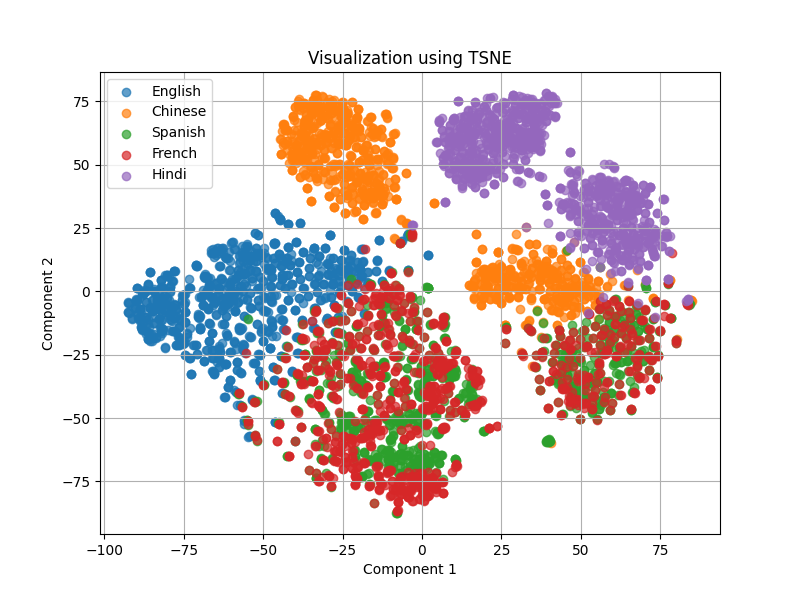}      \caption{T-SNE, Layer 22}        \end{subfigure}  \hfill  \begin{subfigure}{0.18\textwidth}      \includegraphics[width=\textwidth]{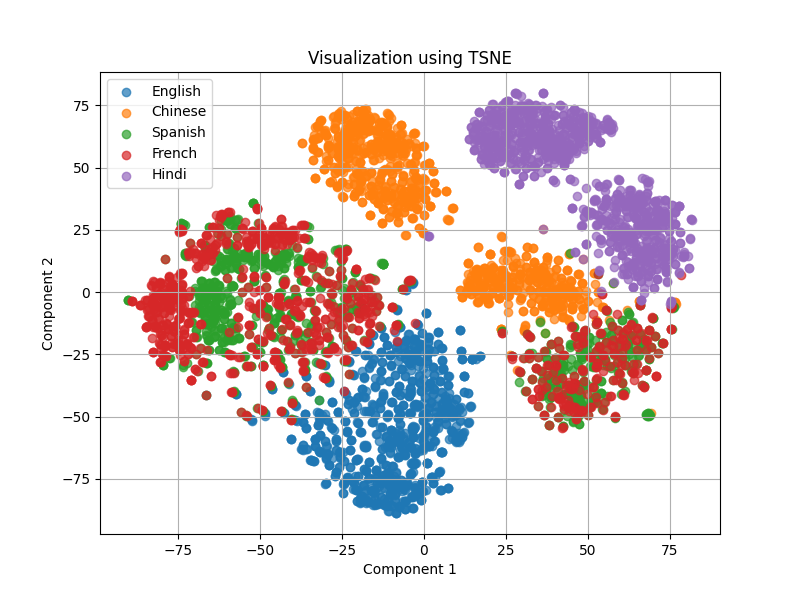}      \caption{T-SNE, Layer 23}        \end{subfigure}  \hfill  \begin{subfigure}{0.18\textwidth}      \includegraphics[width=\textwidth]{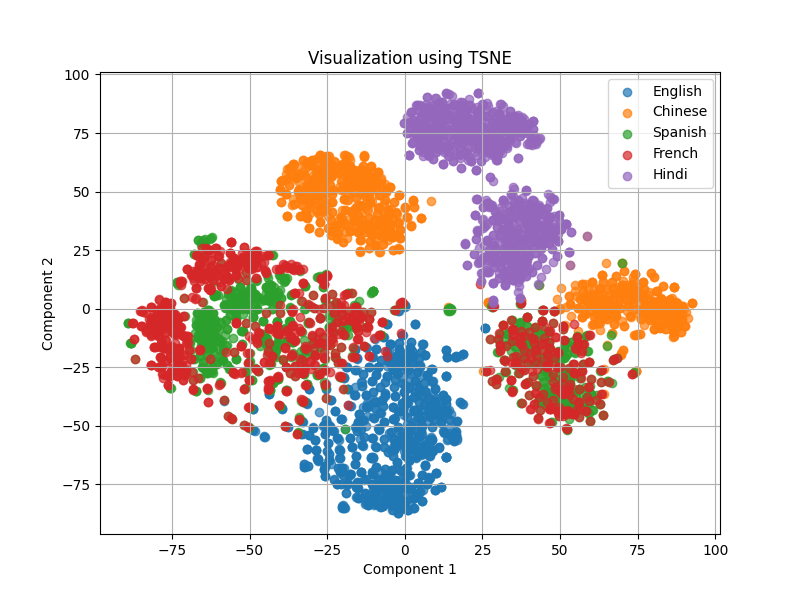}      \caption{T-SNE, Layer 24}        \end{subfigure}  \hfill  \begin{subfigure}{0.18\textwidth}      \includegraphics[width=\textwidth]{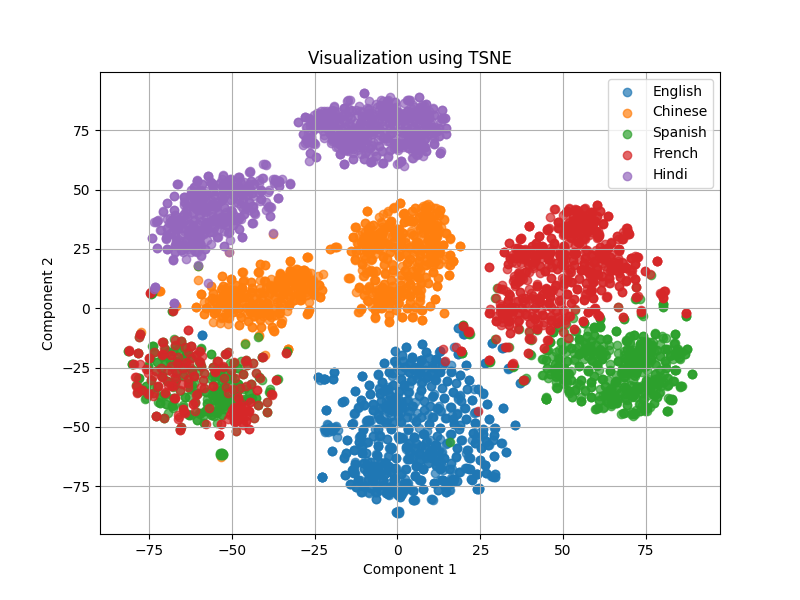}      \caption{T-SNE, Layer 25}        \end{subfigure}    \vspace{0.2in}    %
\begin{subfigure}{0.18\textwidth}      \includegraphics[width=\textwidth]{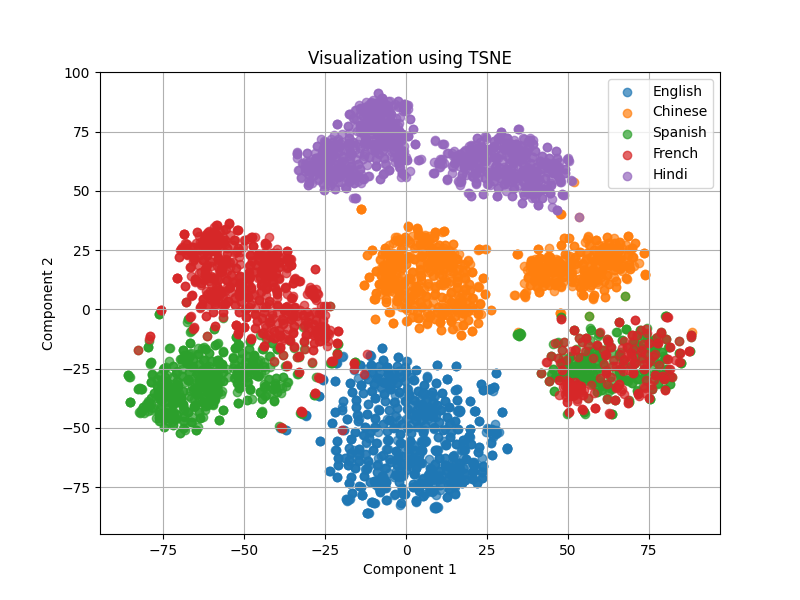}      \caption{T-SNE, Layer 26}        \end{subfigure}  \hfill  \begin{subfigure}{0.18\textwidth}      \includegraphics[width=\textwidth]{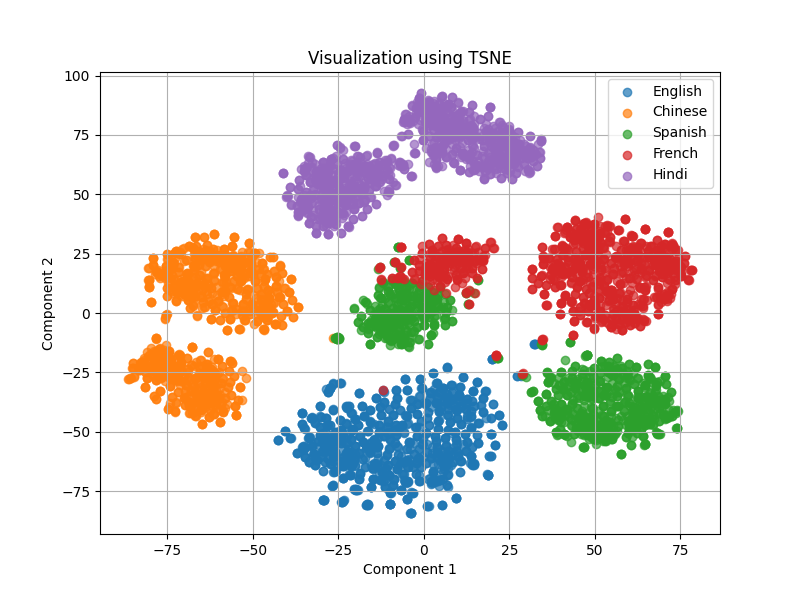}      \caption{T-SNE, Layer 27}        \end{subfigure}  \hfill  \begin{subfigure}{0.18\textwidth}      \includegraphics[width=\textwidth]{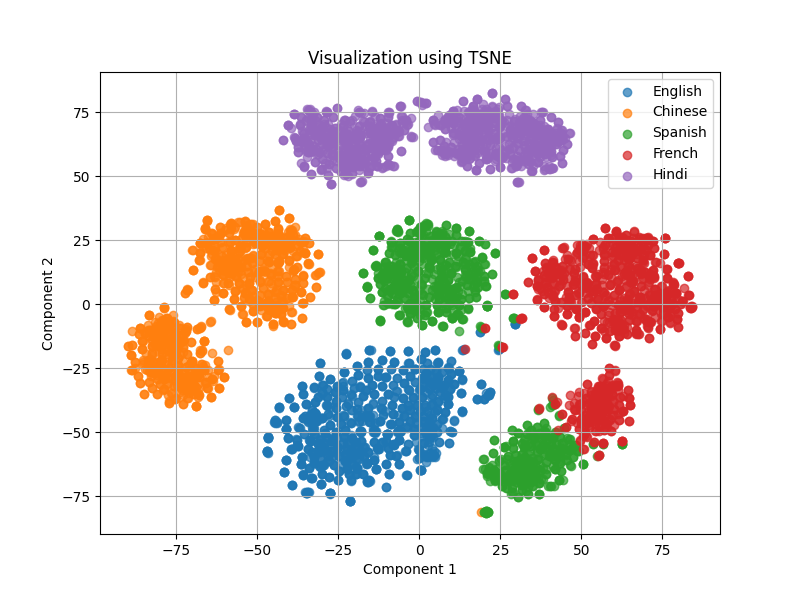}      \caption{T-SNE, Layer 28}        \end{subfigure}  \hfill  \begin{subfigure}{0.18\textwidth}      \includegraphics[width=\textwidth]{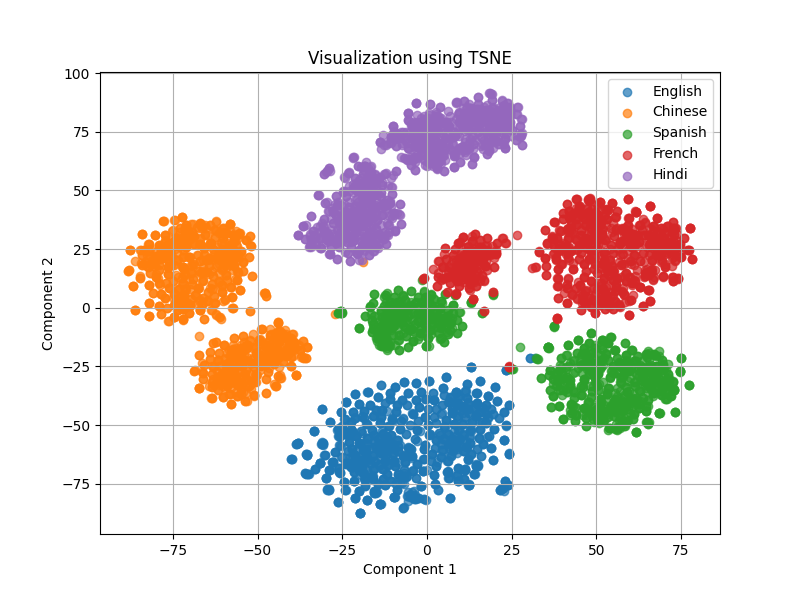}      \caption{T-SNE, Layer 29}        \end{subfigure}  \hfill  \begin{subfigure}{0.18\textwidth}      \includegraphics[width=\textwidth]{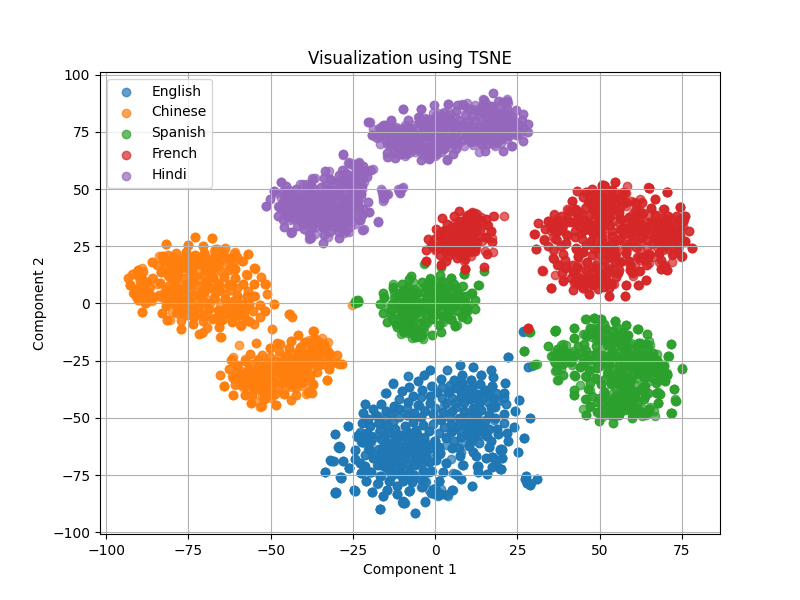}      \caption{T-SNE, Layer 30}        \end{subfigure}    \vspace{0.2in}    %
\begin{subfigure}{0.18\textwidth}      \includegraphics[width=\textwidth]{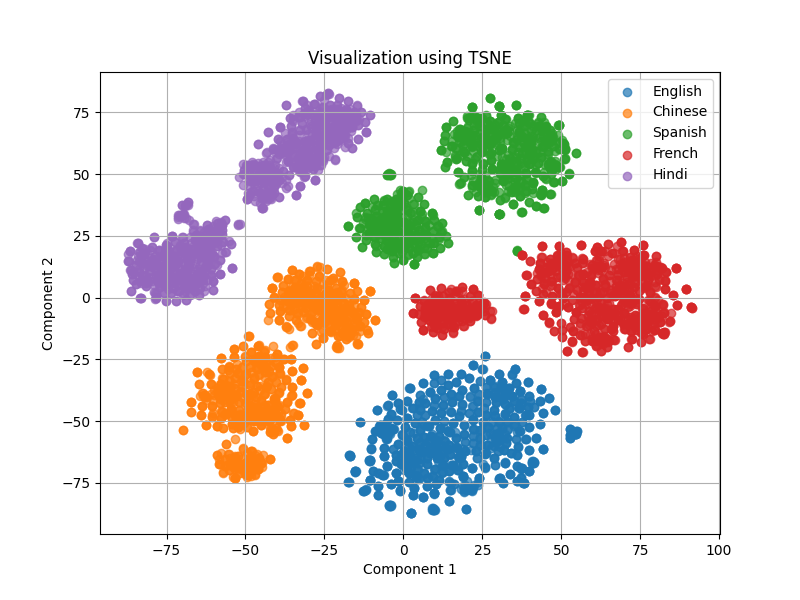}      \caption{T-SNE, Layer 31}        \end{subfigure}  \hfill  \begin{subfigure}{0.18\textwidth}      \includegraphics[width=\textwidth]{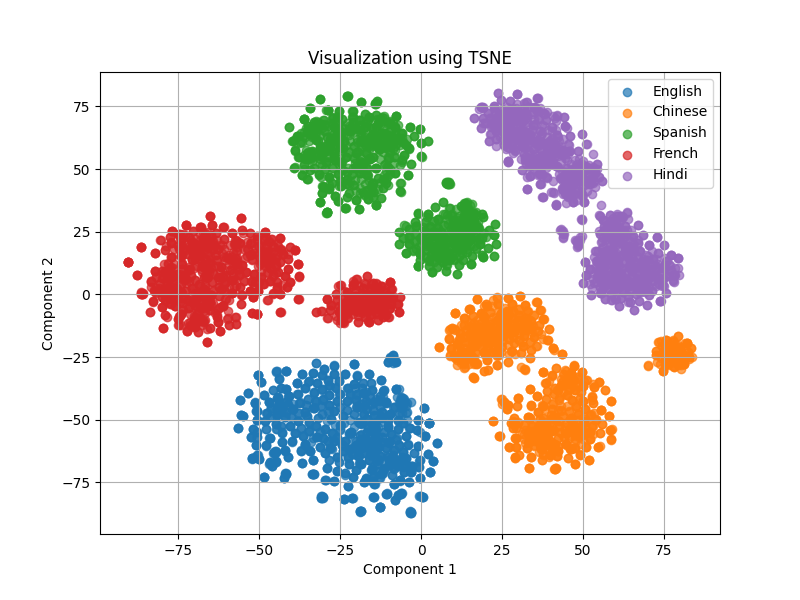}      \caption{T-SNE, Layer 32}        \end{subfigure}    \caption{T-SNE visualizations for layers 1-32 of Llama-3-8B-Instruct on the FOLIO dataset.}  
\end{figure*}

\begin{figure*}[htbp]
\centering
\begin{subfigure}{0.18\textwidth}
\includegraphics[width=\textwidth]{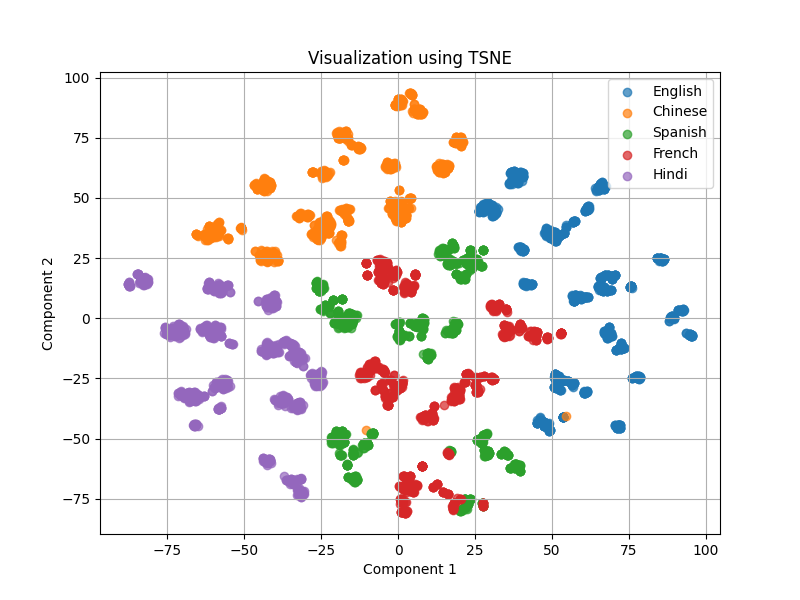}
\caption{T-SNE, Layer 1}
\end{subfigure}
\hfill
\begin{subfigure}{0.18\textwidth}
\includegraphics[width=\textwidth]{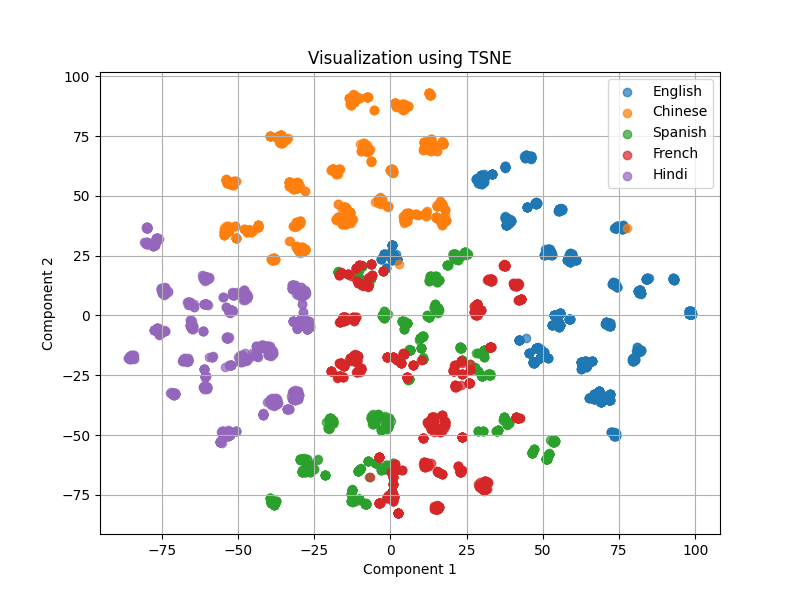}
\caption{T-SNE, Layer 2}

\end{subfigure}
\hfill
\begin{subfigure}{0.18\textwidth}
\includegraphics[width=\textwidth]{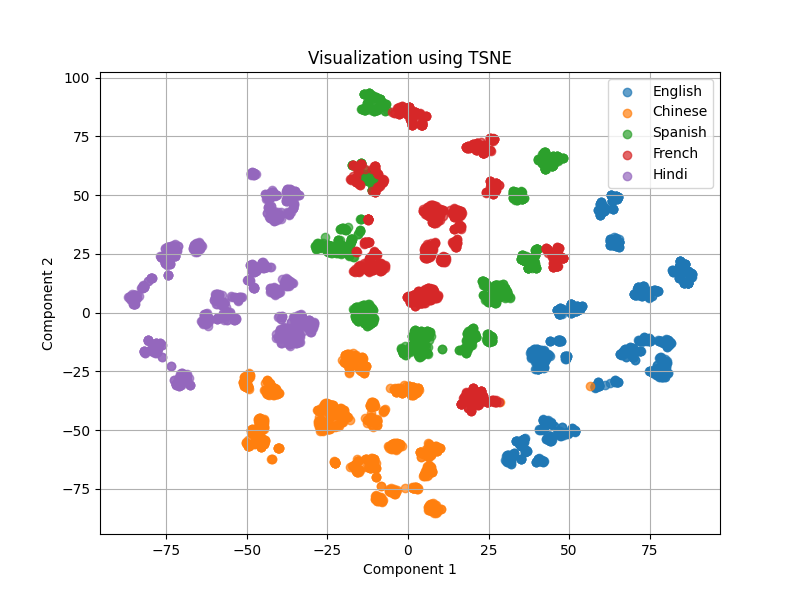}
\caption{T-SNE, Layer 3}

\end{subfigure}
\hfill
\begin{subfigure}{0.18\textwidth}
\includegraphics[width=\textwidth]{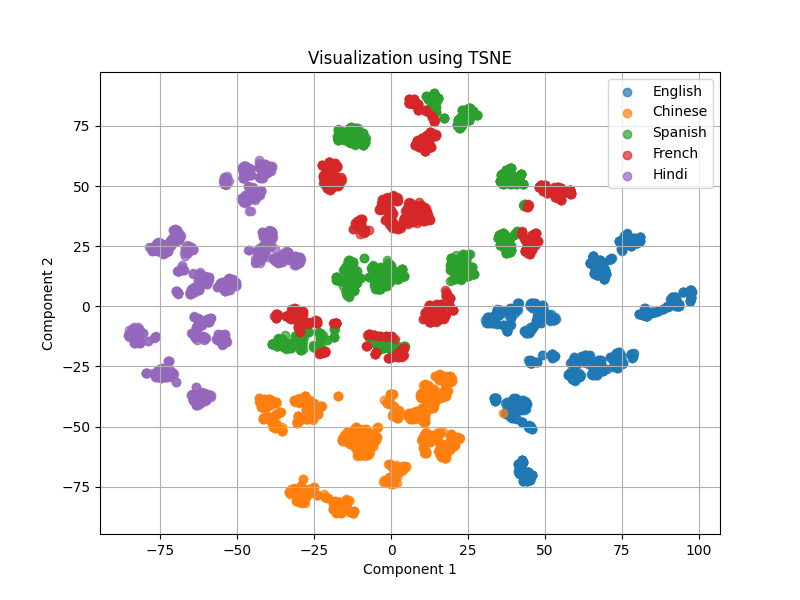}
\caption{T-SNE, Layer 4}

\end{subfigure}
\hfill
\begin{subfigure}{0.18\textwidth}
\includegraphics[width=\textwidth]{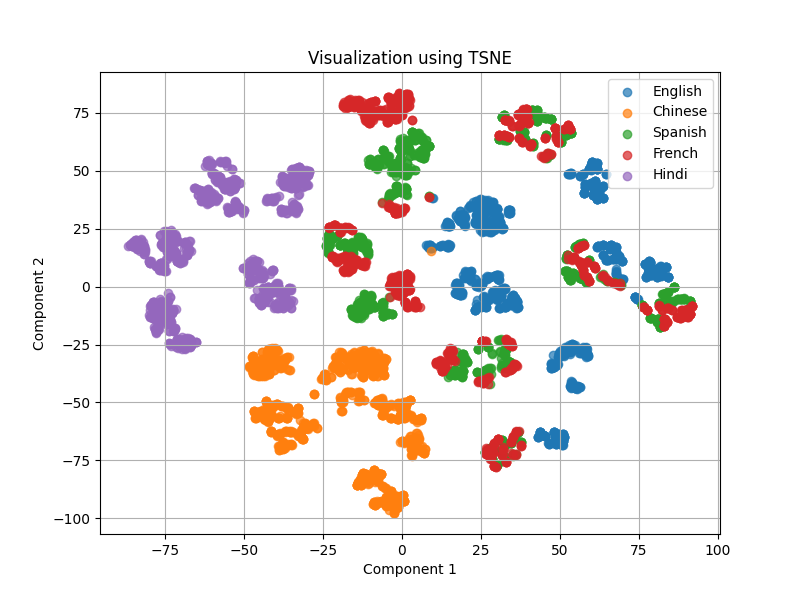}
\caption{T-SNE, Layer 5}

\end{subfigure}
\vspace{0.2in} %
\begin{subfigure}{0.18\textwidth}      \includegraphics[width=\textwidth]{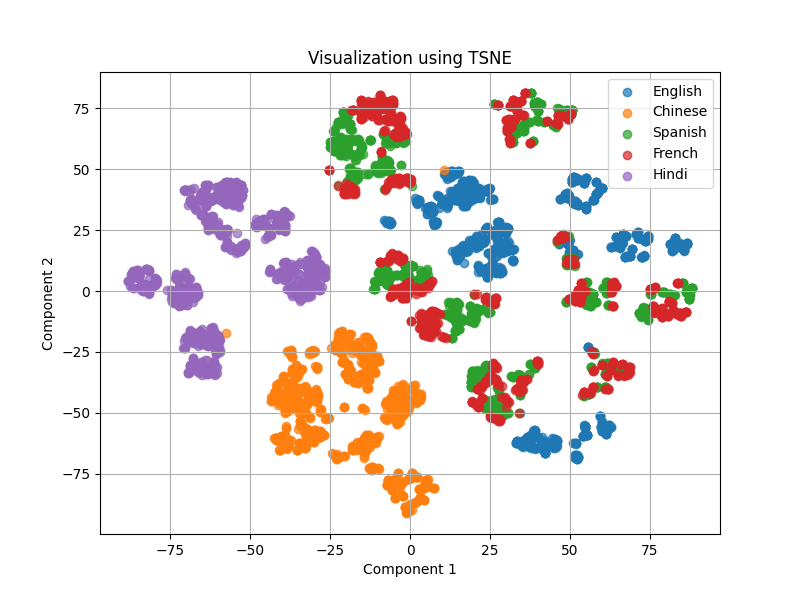}      \caption{T-SNE, Layer 6}        \end{subfigure}  \hfill  \begin{subfigure}{0.18\textwidth}      \includegraphics[width=\textwidth]{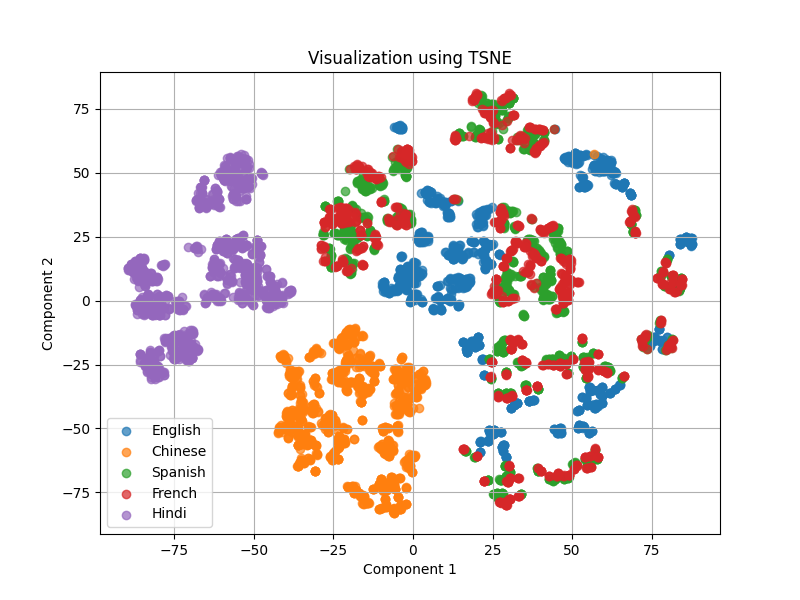}      \caption{T-SNE, Layer 7}        \end{subfigure}  \hfill  \begin{subfigure}{0.18\textwidth}      \includegraphics[width=\textwidth]{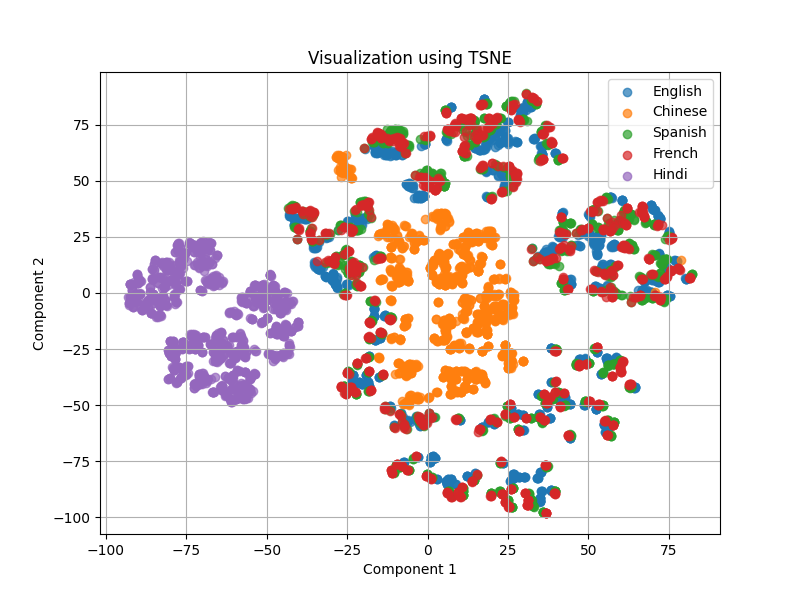}      \caption{T-SNE, Layer 8}        \end{subfigure}  \hfill  \begin{subfigure}{0.18\textwidth}      \includegraphics[width=\textwidth]{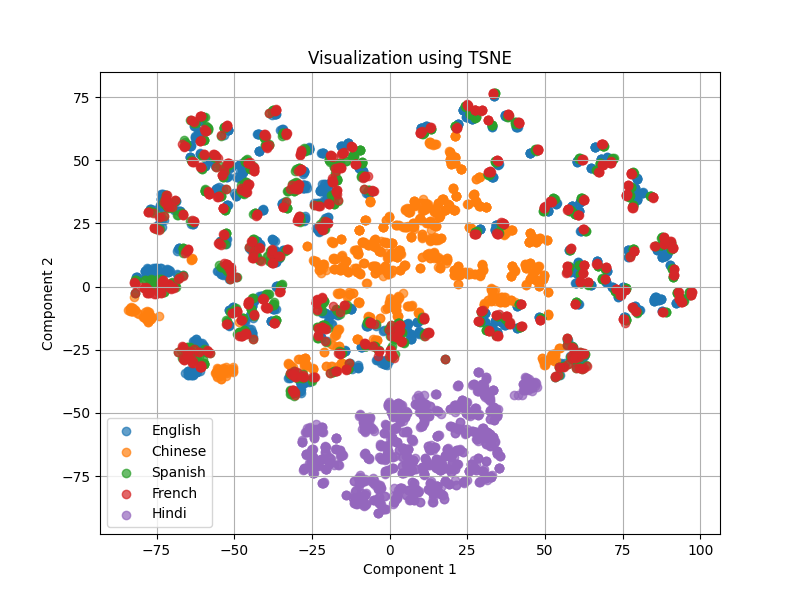}      \caption{T-SNE, Layer 9}        \end{subfigure}  \hfill  \begin{subfigure}{0.18\textwidth}      \includegraphics[width=\textwidth]{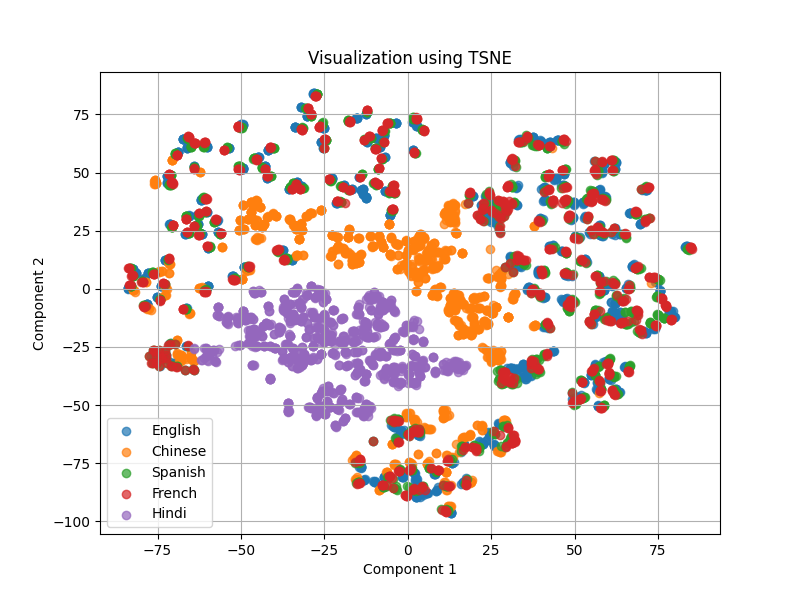}      \caption{T-SNE, Layer 10}        \end{subfigure}    \vspace{0.2in}    %
\begin{subfigure}{0.18\textwidth}      \includegraphics[width=\textwidth]{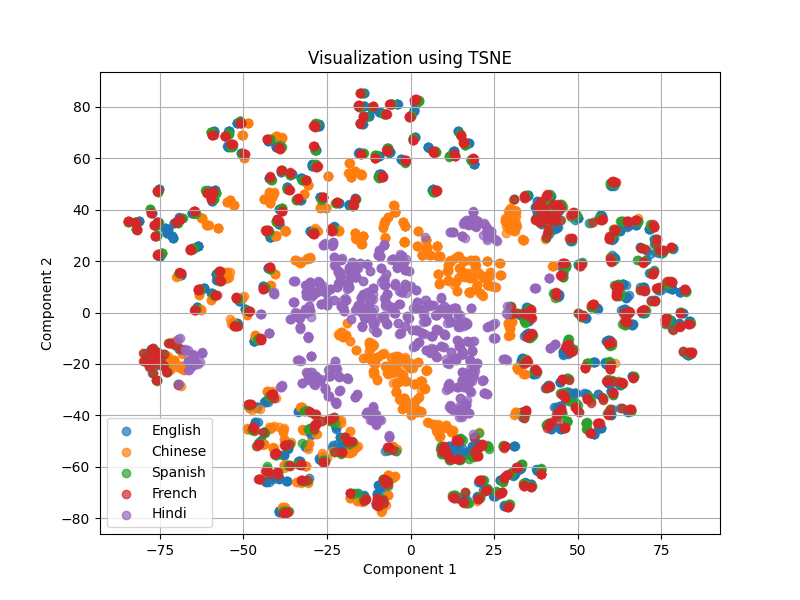}      \caption{T-SNE, Layer 11}        \end{subfigure}  \hfill  \begin{subfigure}{0.18\textwidth}      \includegraphics[width=\textwidth]{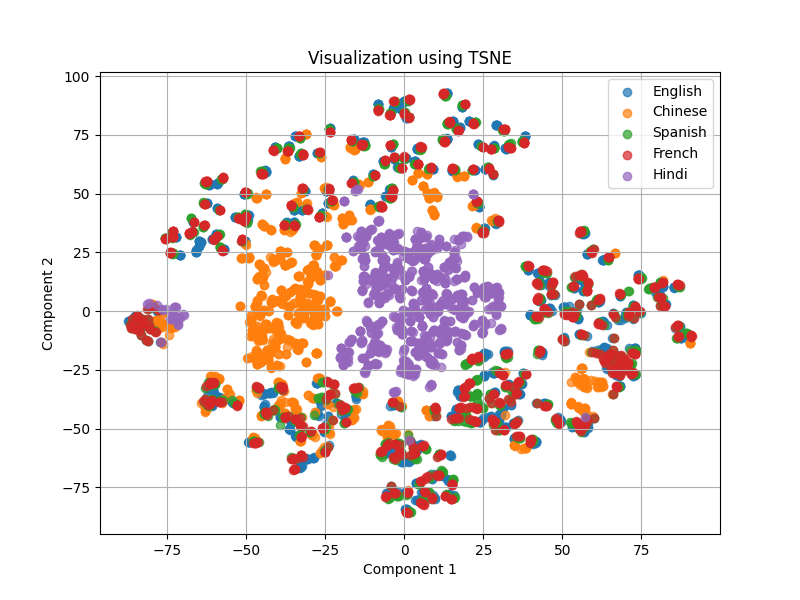}      \caption{T-SNE, Layer 12}        \end{subfigure}  \hfill  \begin{subfigure}{0.18\textwidth}      \includegraphics[width=\textwidth]{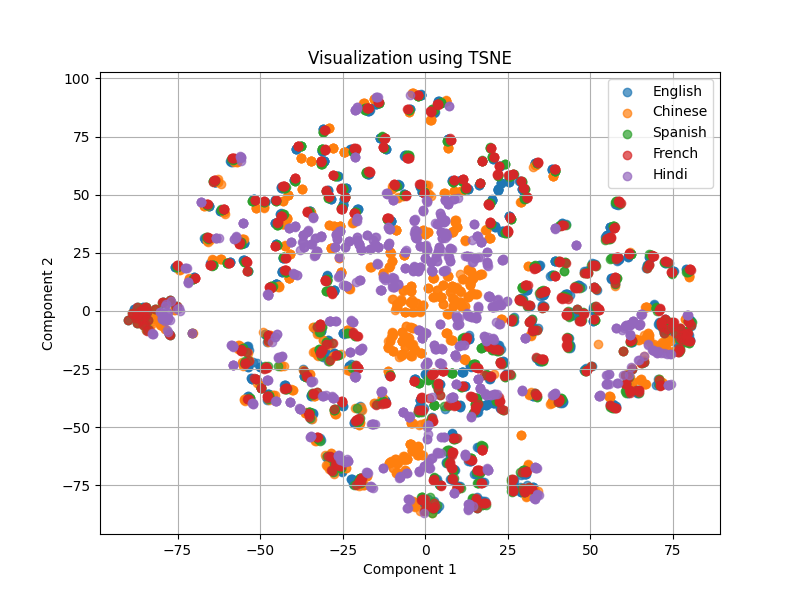}      \caption{T-SNE, Layer 13}        \end{subfigure}  \hfill  \begin{subfigure}{0.18\textwidth}      \includegraphics[width=\textwidth]{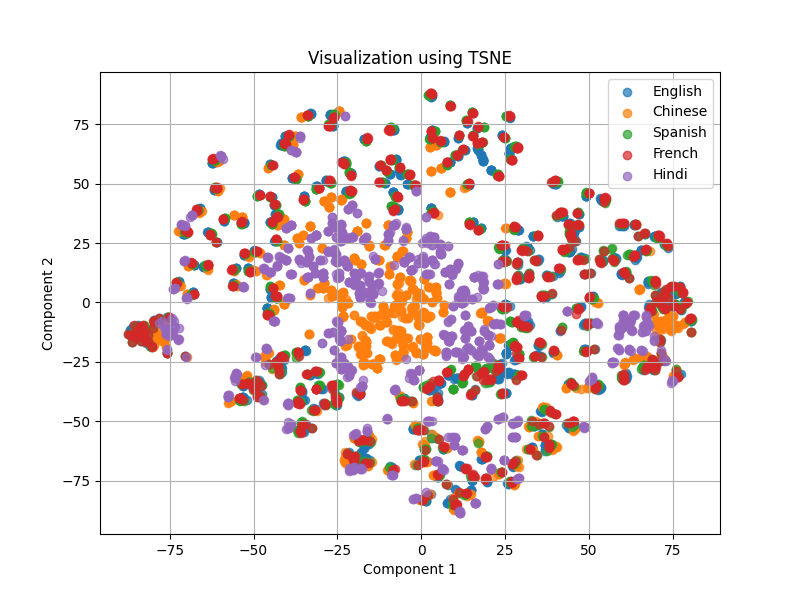}      \caption{T-SNE, Layer 14}        \end{subfigure}  \hfill  \begin{subfigure}{0.18\textwidth}      \includegraphics[width=\textwidth]{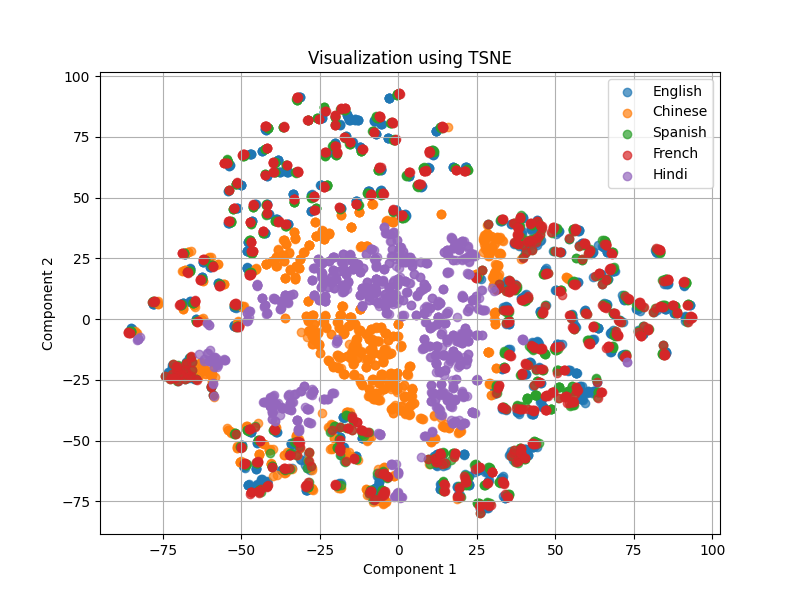}      \caption{T-SNE, Layer 15}        \end{subfigure}    \vspace{0.2in}    %
\begin{subfigure}{0.18\textwidth}      \includegraphics[width=\textwidth]{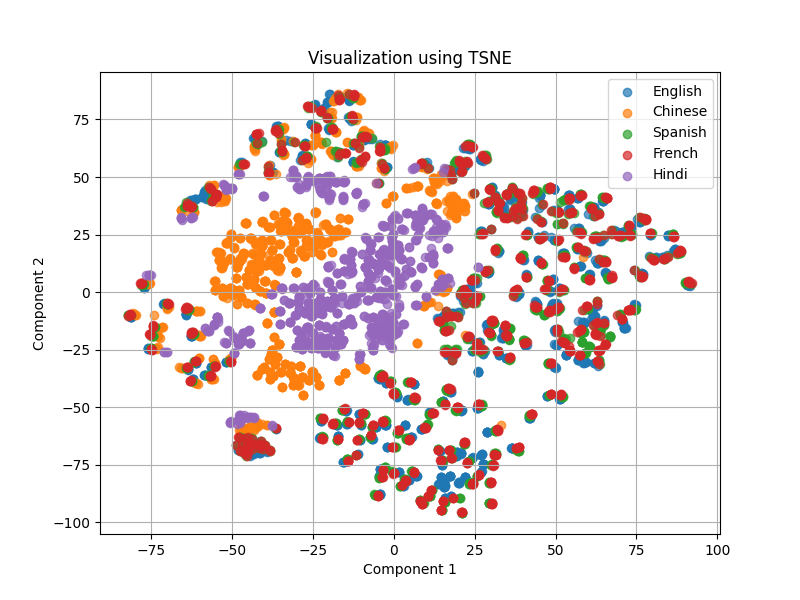}      \caption{T-SNE, Layer 16}        \end{subfigure}  \hfill  \begin{subfigure}{0.18\textwidth}      \includegraphics[width=\textwidth]{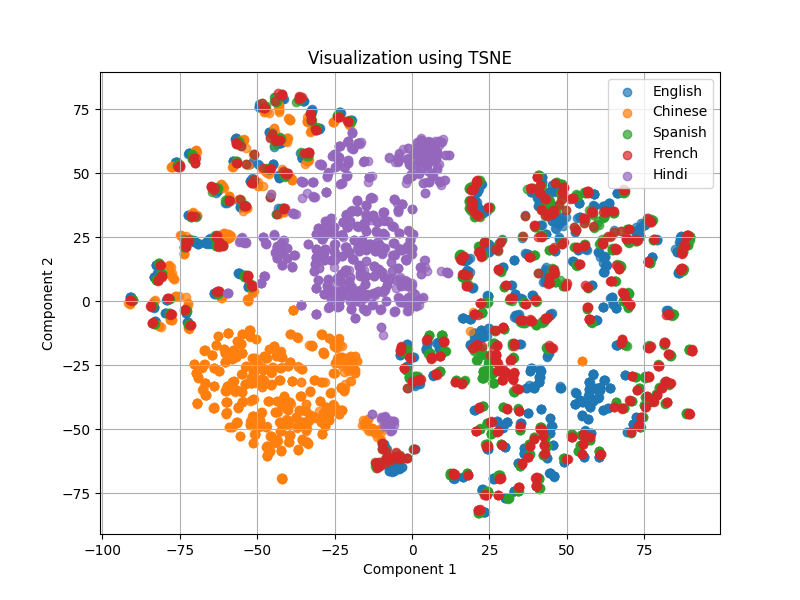}      \caption{T-SNE, Layer 17}        \end{subfigure}  \hfill  \begin{subfigure}{0.18\textwidth}      \includegraphics[width=\textwidth]{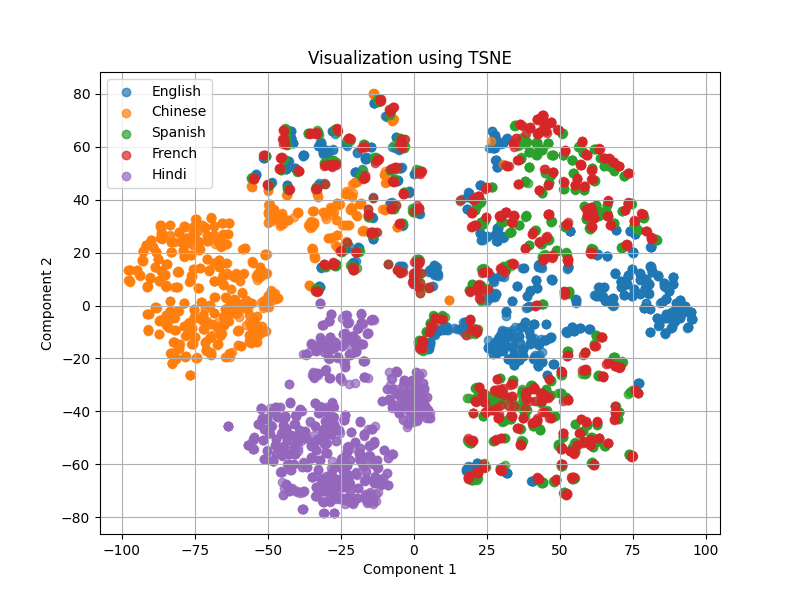}      \caption{T-SNE, Layer 18}        \end{subfigure}  \hfill  \begin{subfigure}{0.18\textwidth}      \includegraphics[width=\textwidth]{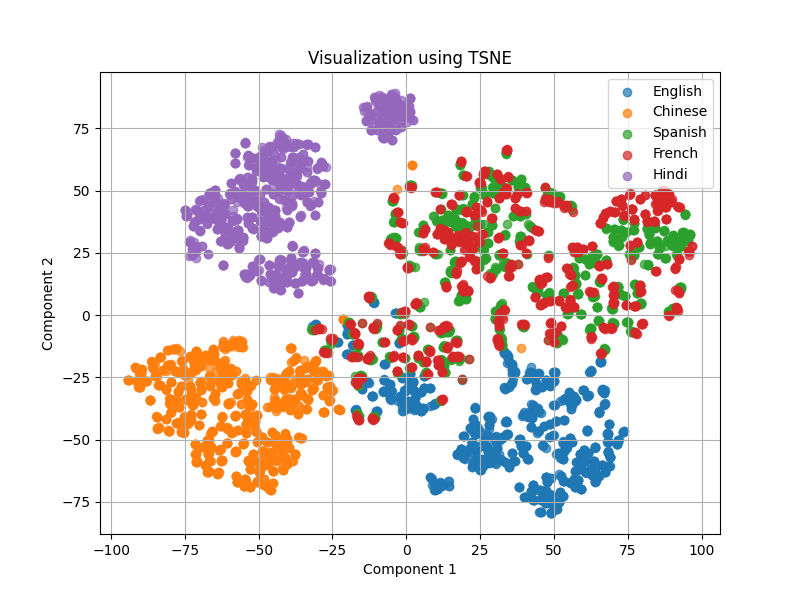}      \caption{T-SNE, Layer 19}        \end{subfigure}  \hfill  \begin{subfigure}{0.18\textwidth}      \includegraphics[width=\textwidth]{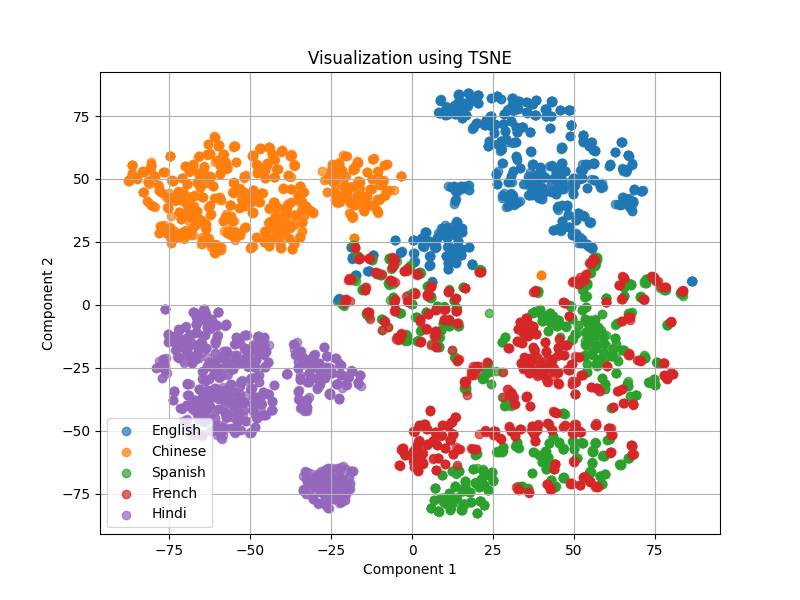}      \caption{T-SNE, Layer 20}        \end{subfigure}    \vspace{0.2in}    %
\begin{subfigure}{0.18\textwidth}      \includegraphics[width=\textwidth]{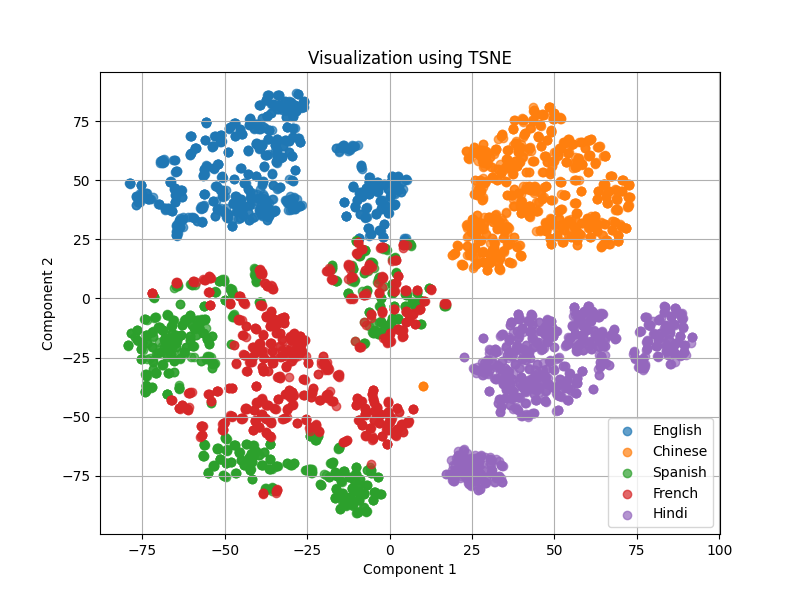}      \caption{T-SNE, Layer 21}        \end{subfigure}  \hfill  \begin{subfigure}{0.18\textwidth}      \includegraphics[width=\textwidth]{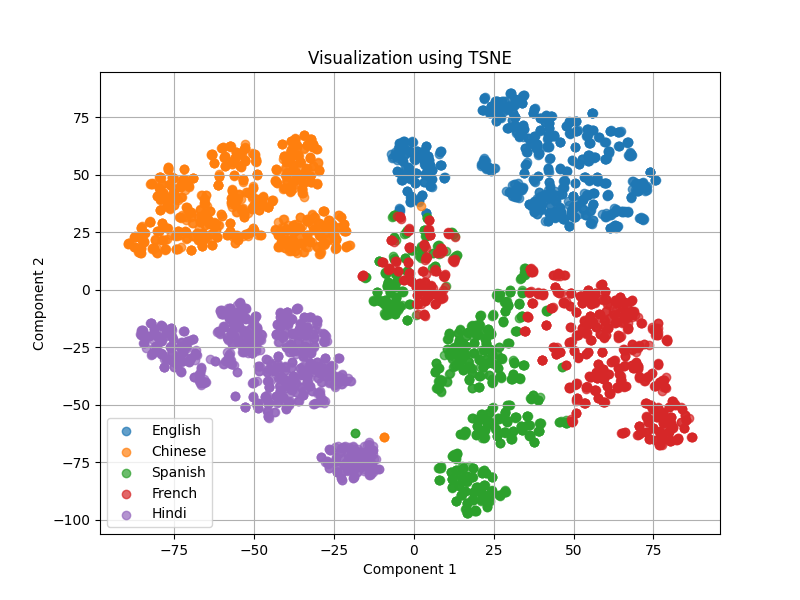}      \caption{T-SNE, Layer 22}        \end{subfigure}  \hfill  \begin{subfigure}{0.18\textwidth}      \includegraphics[width=\textwidth]{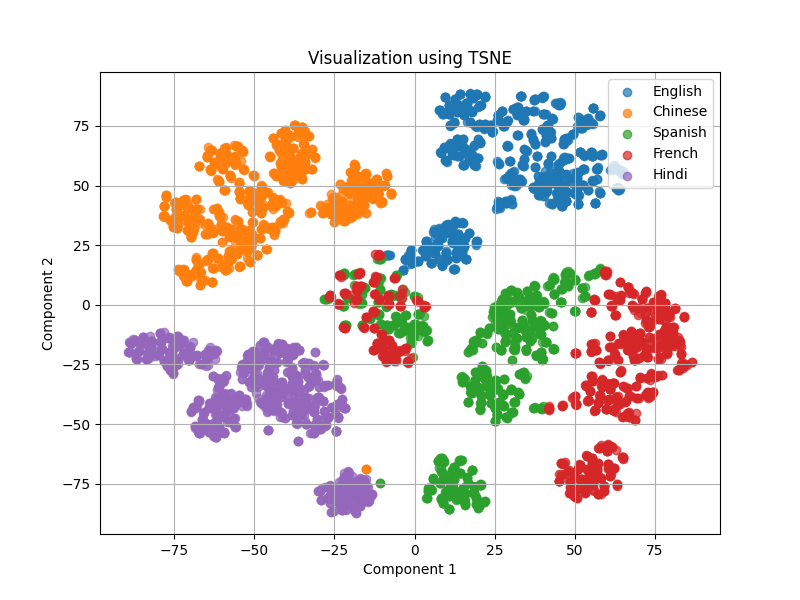}      \caption{T-SNE, Layer 23}        \end{subfigure}  \hfill  \begin{subfigure}{0.18\textwidth}      \includegraphics[width=\textwidth]{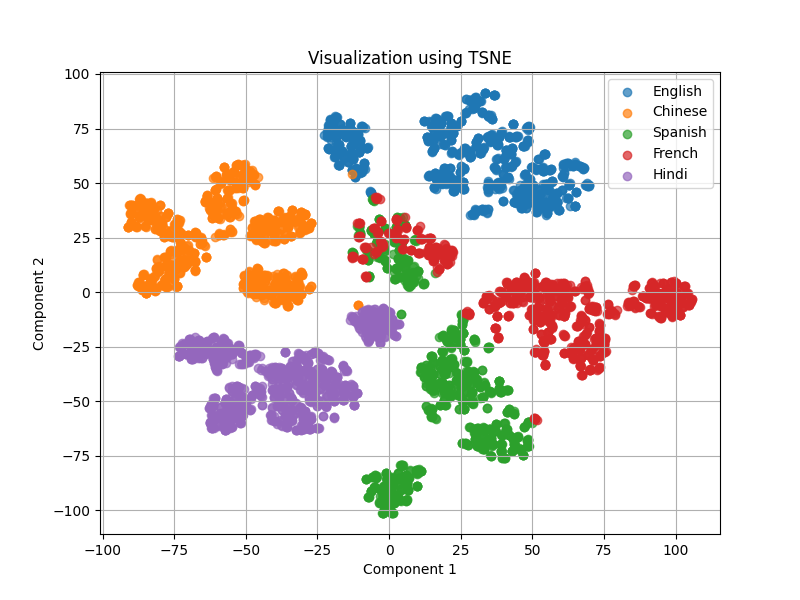}      \caption{T-SNE, Layer 24}        \end{subfigure}  \hfill  \begin{subfigure}{0.18\textwidth}      \includegraphics[width=\textwidth]{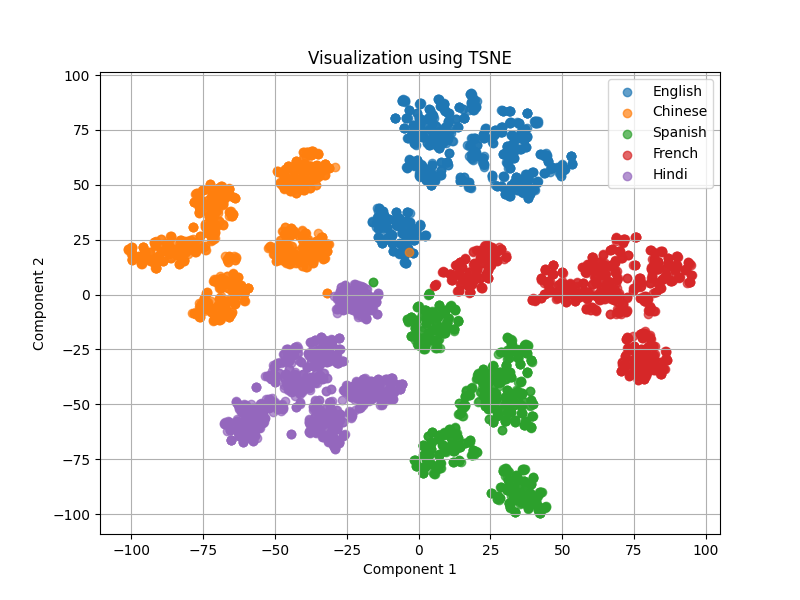}      \caption{T-SNE, Layer 25}        \end{subfigure}    \vspace{0.2in}    %
\begin{subfigure}{0.18\textwidth}      \includegraphics[width=\textwidth]{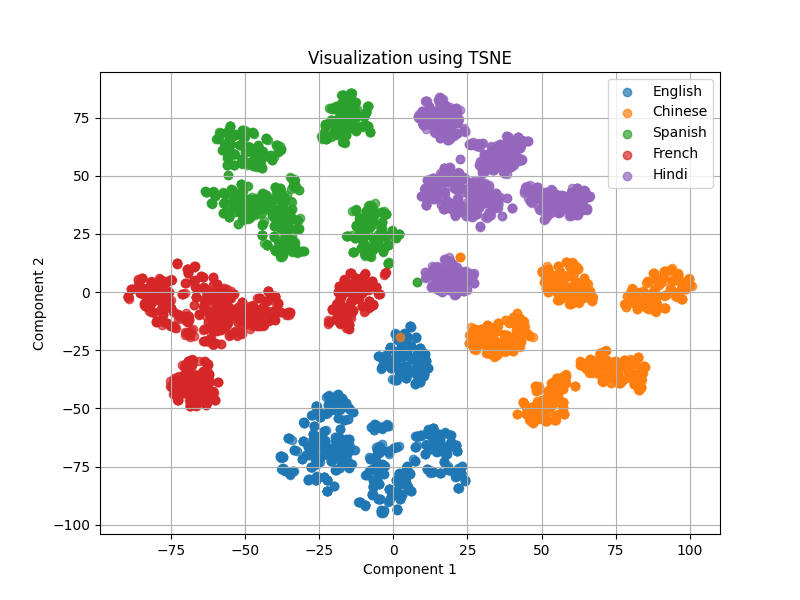}      \caption{T-SNE, Layer 26}        \end{subfigure}  \hfill  \begin{subfigure}{0.18\textwidth}      \includegraphics[width=\textwidth]{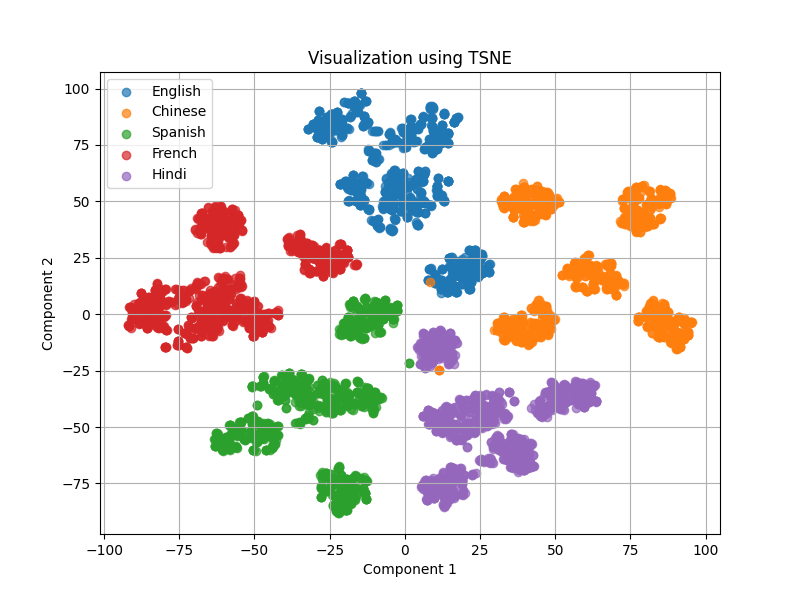}      \caption{T-SNE, Layer 27}        \end{subfigure}  \hfill  \begin{subfigure}{0.18\textwidth}      \includegraphics[width=\textwidth]{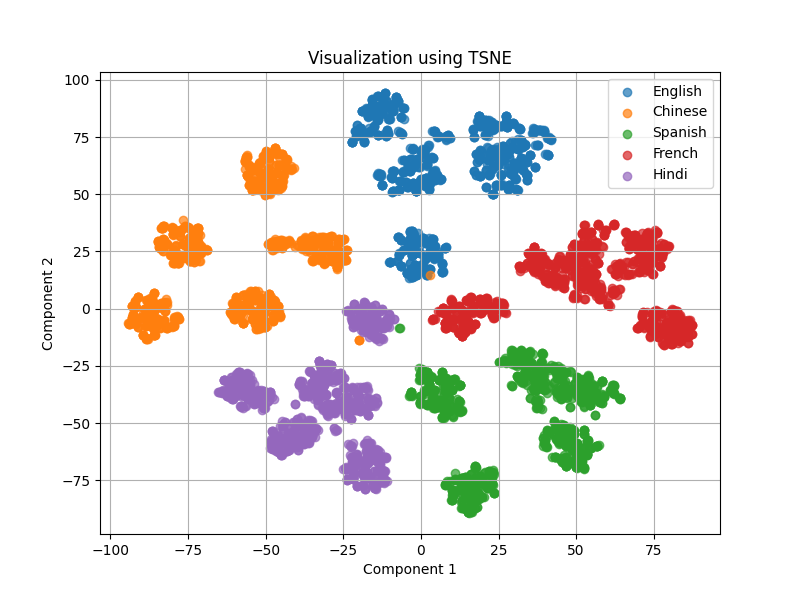}      \caption{T-SNE, Layer 28}        \end{subfigure}  \hfill  \begin{subfigure}{0.18\textwidth}      \includegraphics[width=\textwidth]{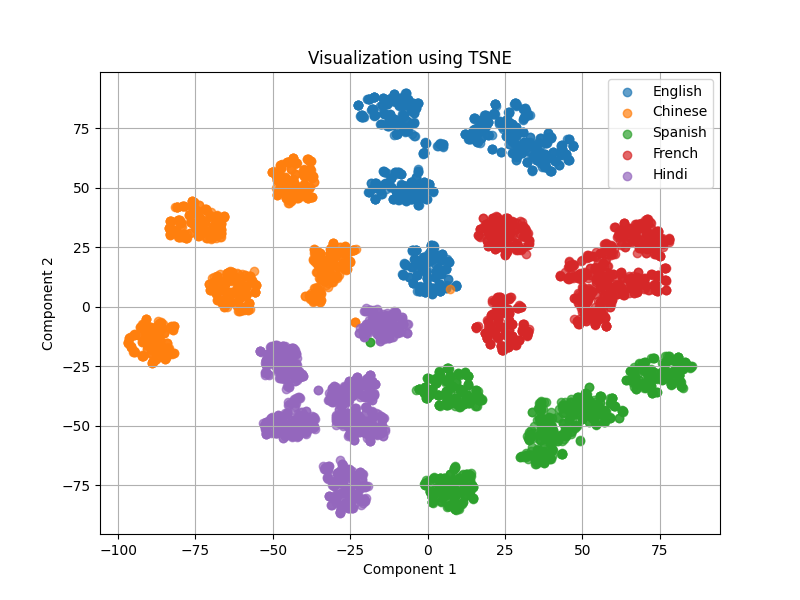}      \caption{T-SNE, Layer 29}        \end{subfigure}  \hfill  \begin{subfigure}{0.18\textwidth}      \includegraphics[width=\textwidth]{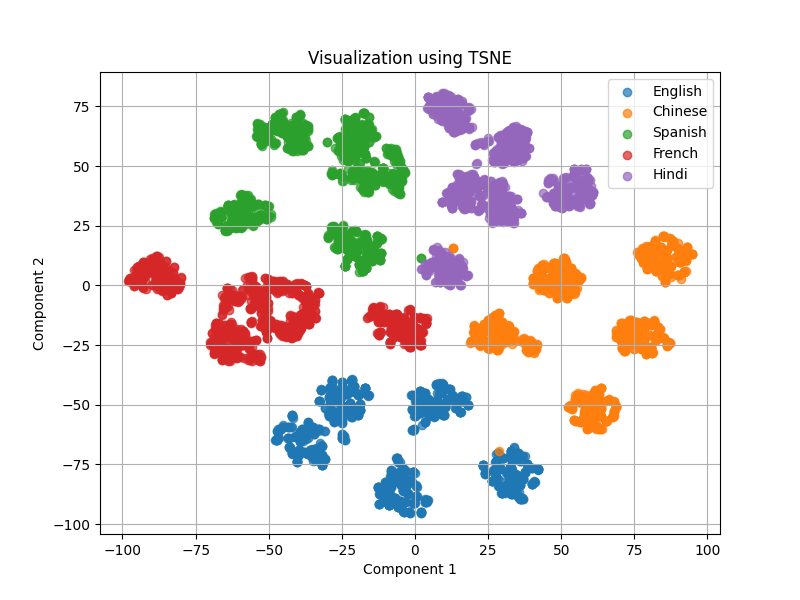}      \caption{T-SNE, Layer 30}        \end{subfigure}    \vspace{0.2in}    %
\begin{subfigure}{0.18\textwidth}      \includegraphics[width=\textwidth]{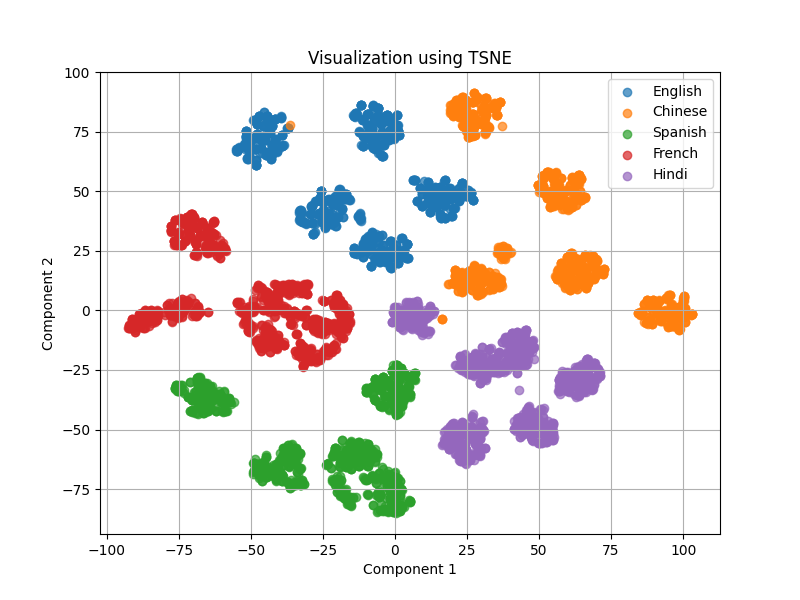}      \caption{T-SNE, Layer 31}        \end{subfigure}  \hfill  \begin{subfigure}{0.18\textwidth}      \includegraphics[width=\textwidth]{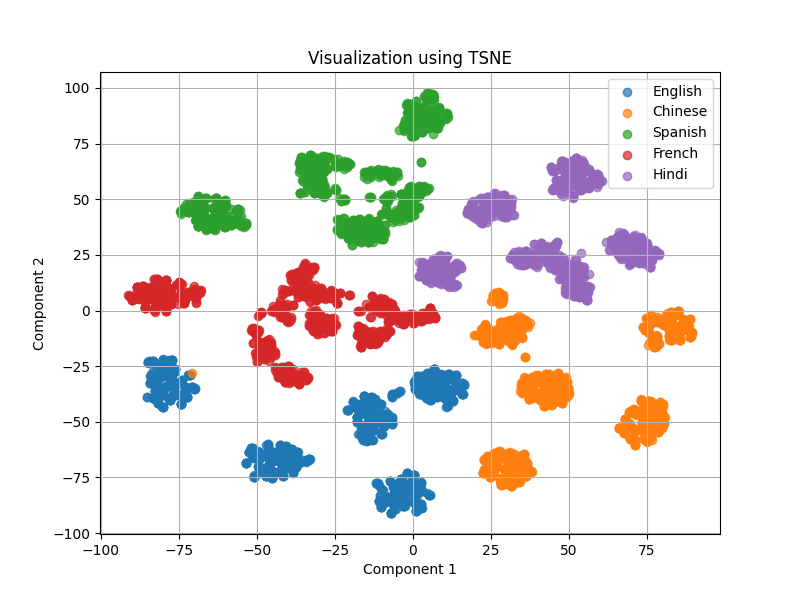}      \caption{T-SNE, Layer 32}        \end{subfigure}    \caption{T-SNE visualizations for layers 1-32 of Llama-3-8B-Instruct on the LogicalDeduction dataset.}  
\end{figure*}

\begin{figure*}[htbp]
\centering
\begin{subfigure}{0.18\textwidth}
\includegraphics[width=\textwidth]{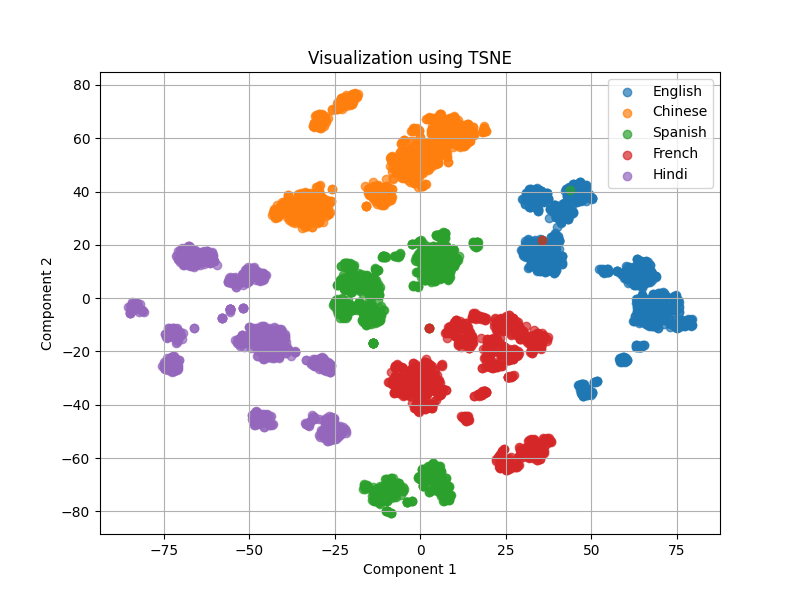}
\caption{T-SNE, Layer 1}
\end{subfigure}
\hfill
\begin{subfigure}{0.18\textwidth}
\includegraphics[width=\textwidth]{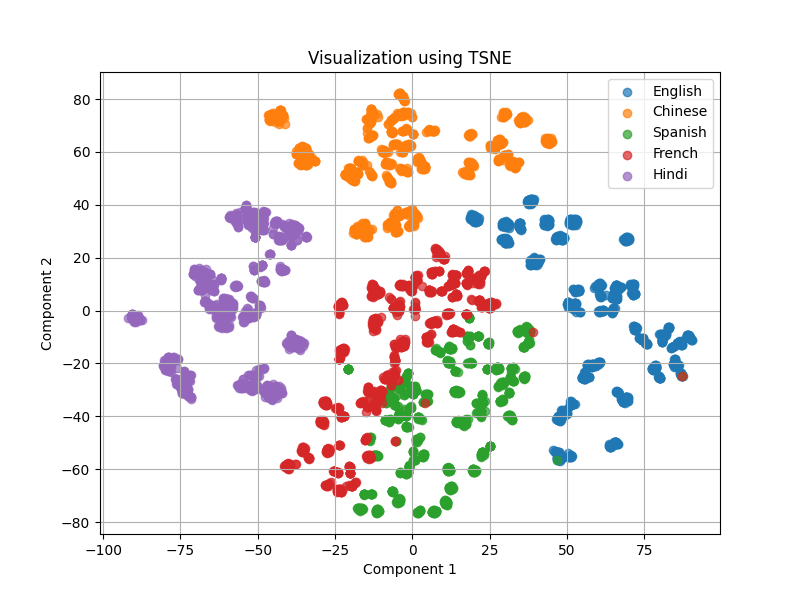}
\caption{T-SNE, Layer 2}

\end{subfigure}
\hfill
\begin{subfigure}{0.18\textwidth}
\includegraphics[width=\textwidth]{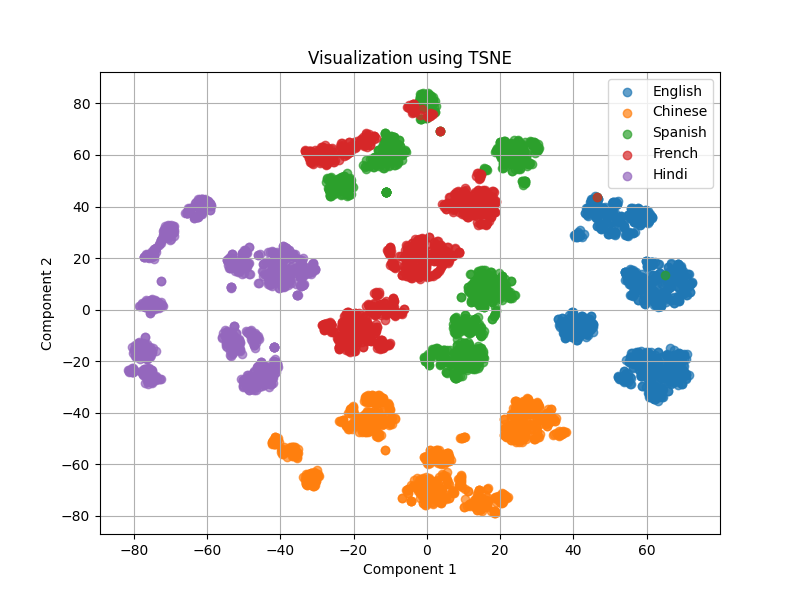}
\caption{T-SNE, Layer 3}

\end{subfigure}
\hfill
\begin{subfigure}{0.18\textwidth}
\includegraphics[width=\textwidth]{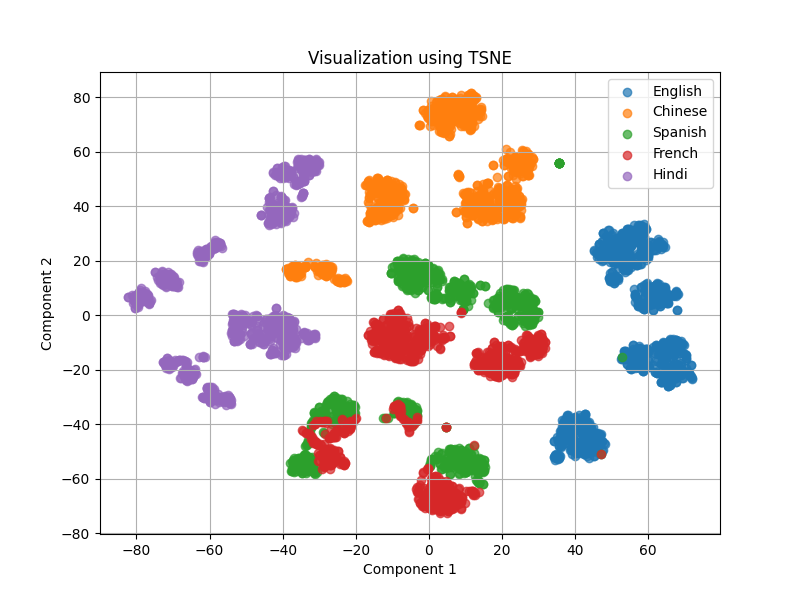}
\caption{T-SNE, Layer 4}

\end{subfigure}
\hfill
\begin{subfigure}{0.18\textwidth}
\includegraphics[width=\textwidth]{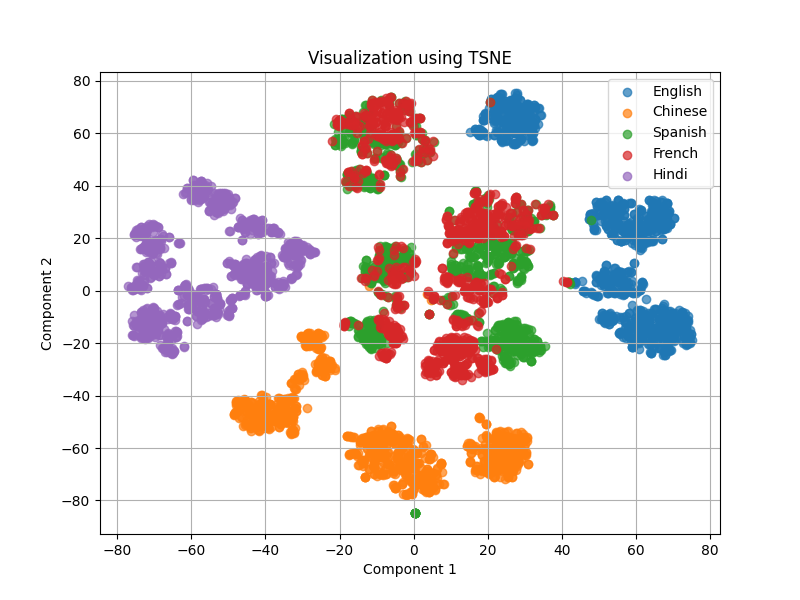}
\caption{T-SNE, Layer 5}

\end{subfigure}
\vspace{0.2in} %
\begin{subfigure}{0.18\textwidth}      \includegraphics[width=\textwidth]{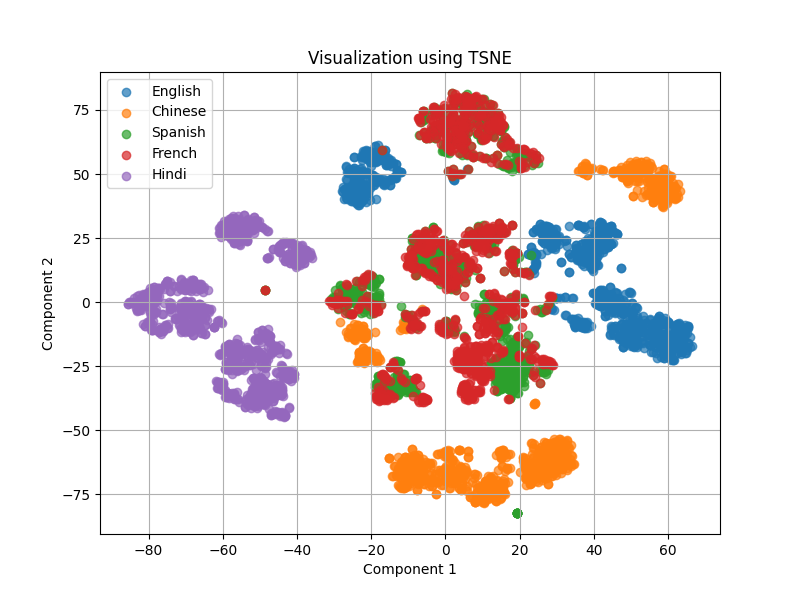}      \caption{T-SNE, Layer 6}        \end{subfigure}  \hfill  \begin{subfigure}{0.18\textwidth}      \includegraphics[width=\textwidth]{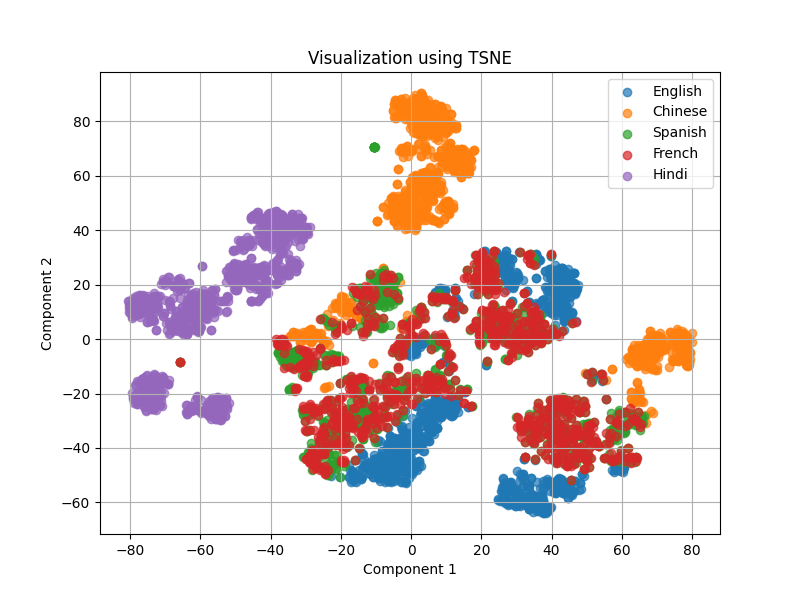}      \caption{T-SNE, Layer 7}        \end{subfigure}  \hfill  \begin{subfigure}{0.18\textwidth}      \includegraphics[width=\textwidth]{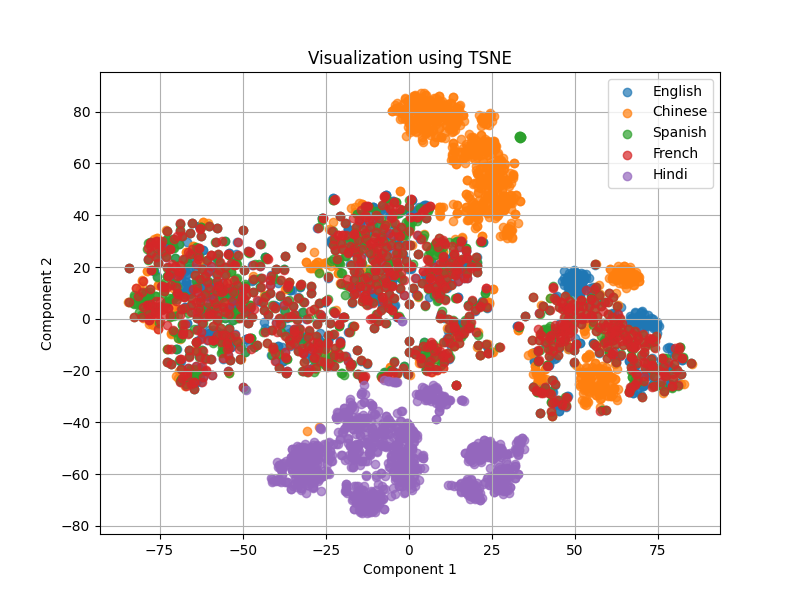}      \caption{T-SNE, Layer 8}        \end{subfigure}  \hfill  \begin{subfigure}{0.18\textwidth}      \includegraphics[width=\textwidth]{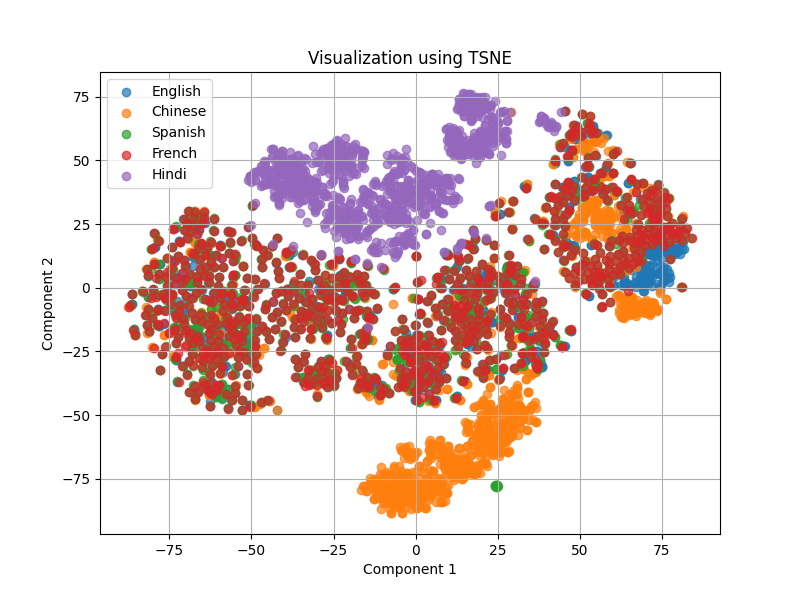}      \caption{T-SNE, Layer 9}        \end{subfigure}  \hfill  \begin{subfigure}{0.18\textwidth}      \includegraphics[width=\textwidth]{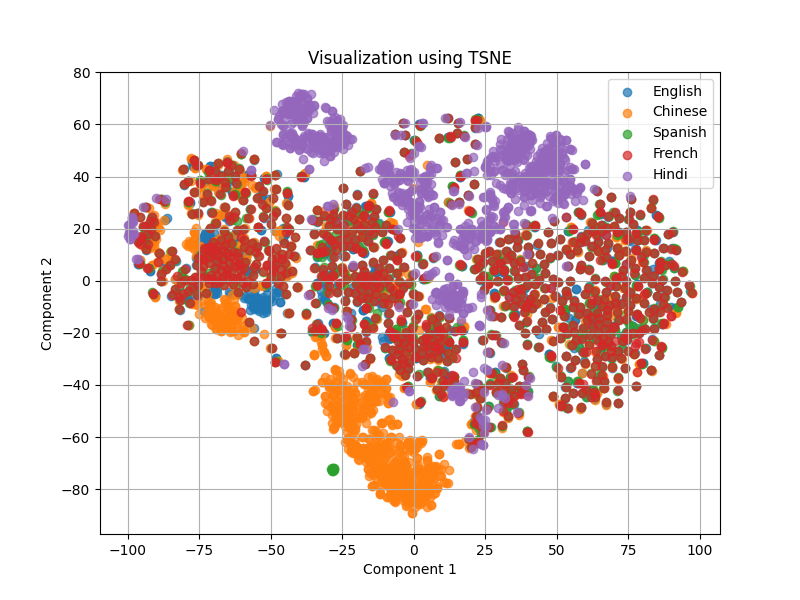}      \caption{T-SNE, Layer 10}        \end{subfigure}    \vspace{0.2in}    %
\begin{subfigure}{0.18\textwidth}      \includegraphics[width=\textwidth]{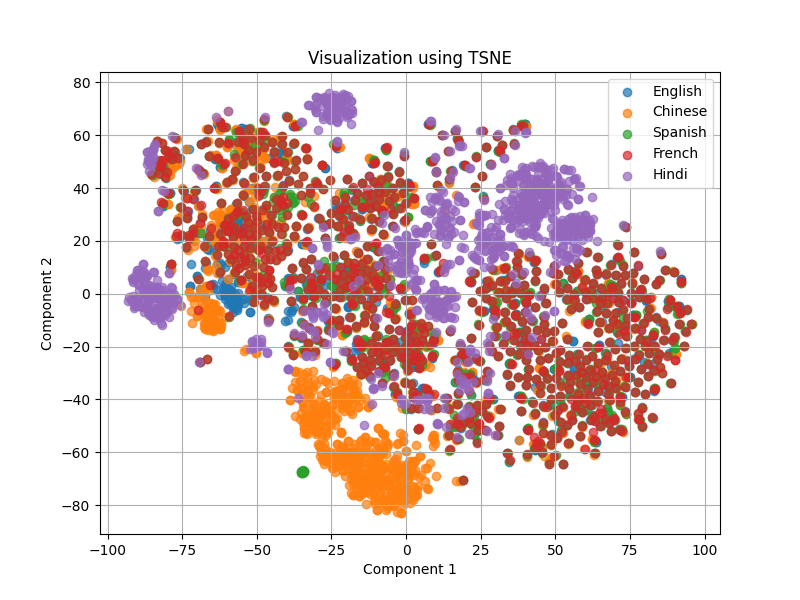}      \caption{T-SNE, Layer 11}        \end{subfigure}  \hfill  \begin{subfigure}{0.18\textwidth}      \includegraphics[width=\textwidth]{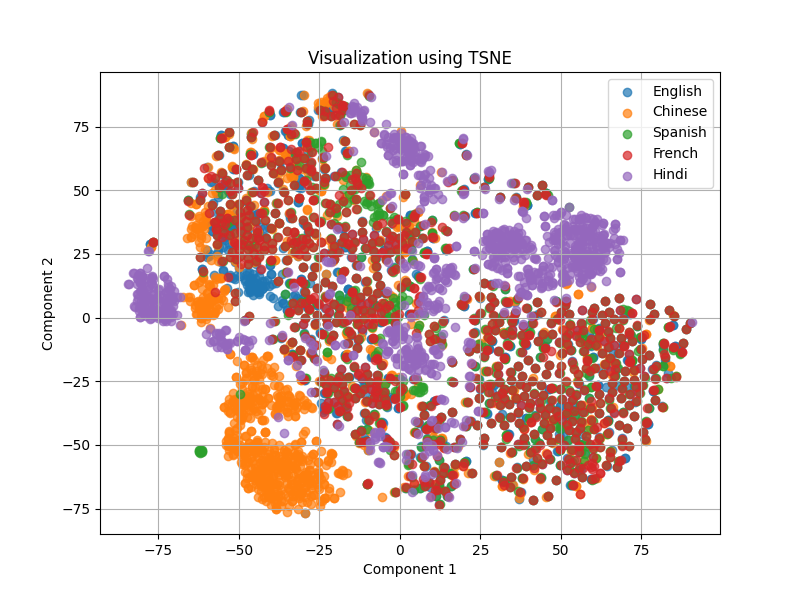}      \caption{T-SNE, Layer 12}        \end{subfigure}  \hfill  \begin{subfigure}{0.18\textwidth}      \includegraphics[width=\textwidth]{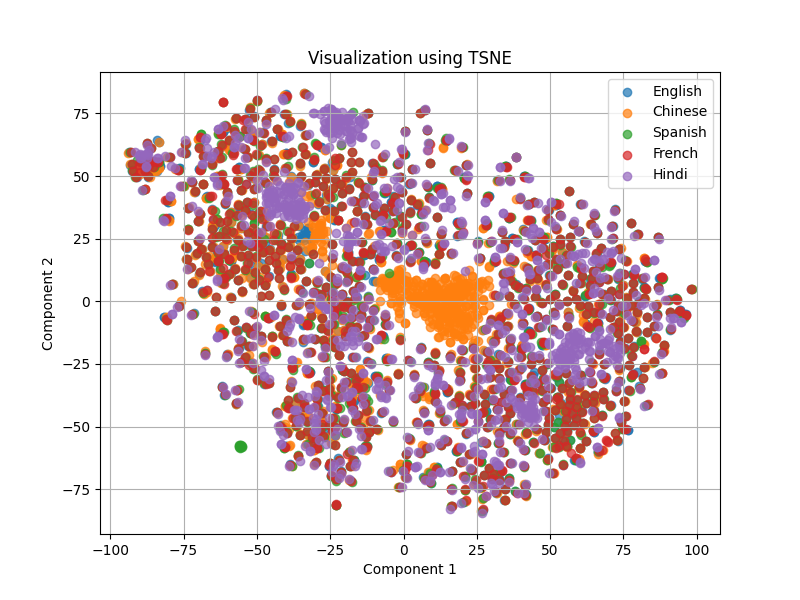}      \caption{T-SNE, Layer 13}        \end{subfigure}  \hfill  \begin{subfigure}{0.18\textwidth}      \includegraphics[width=\textwidth]{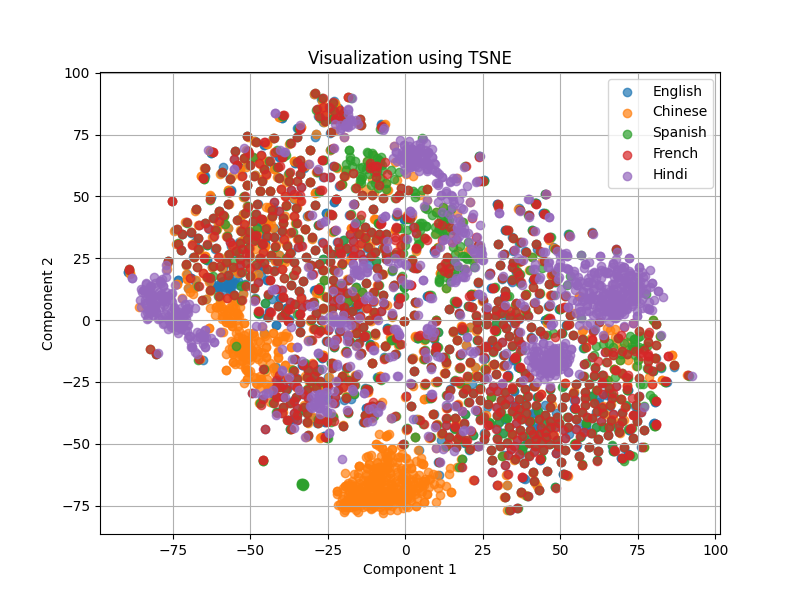}      \caption{T-SNE, Layer 14}        \end{subfigure}  \hfill  \begin{subfigure}{0.18\textwidth}      \includegraphics[width=\textwidth]{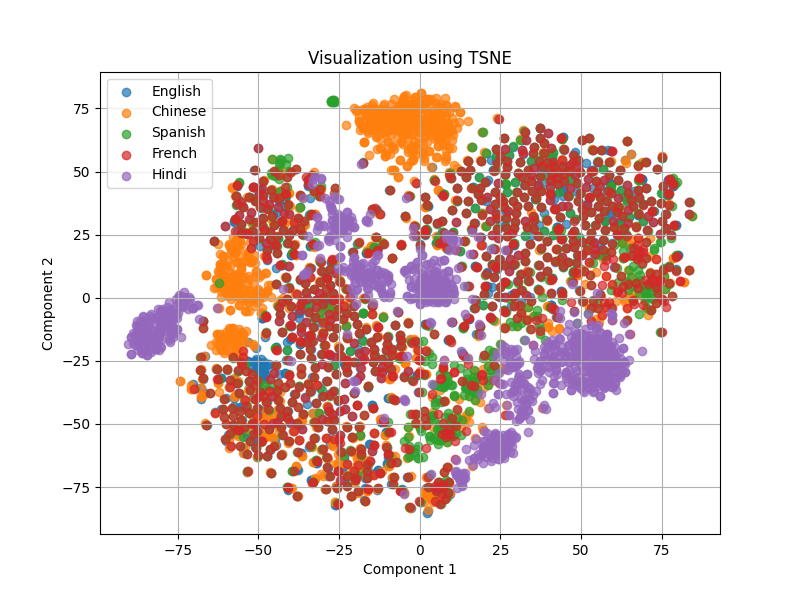}      \caption{T-SNE, Layer 15}        \end{subfigure}    \vspace{0.2in}    %
\begin{subfigure}{0.18\textwidth}      \includegraphics[width=\textwidth]{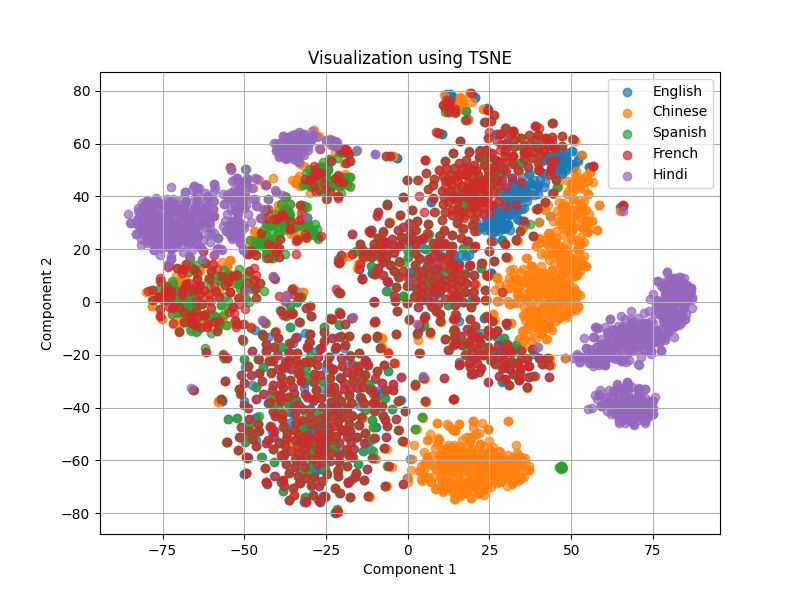}      \caption{T-SNE, Layer 16}        \end{subfigure}  \hfill  \begin{subfigure}{0.18\textwidth}      \includegraphics[width=\textwidth]{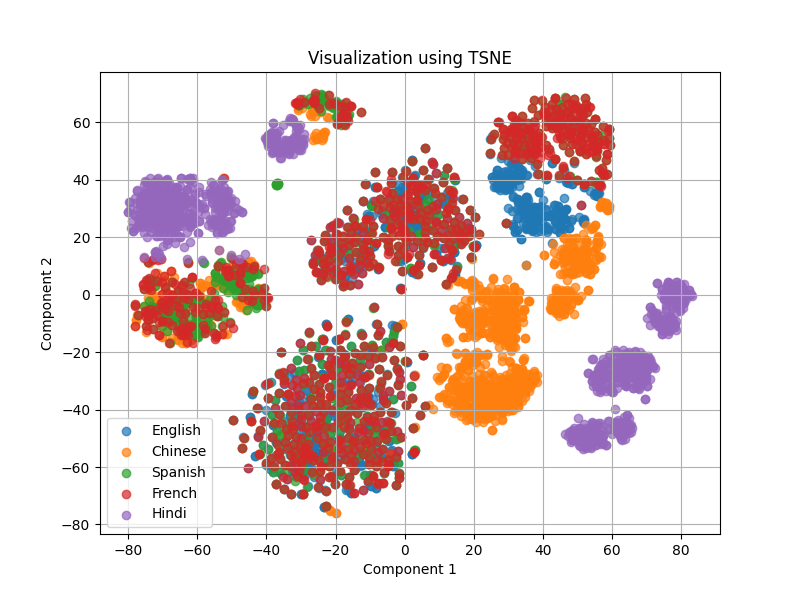}      \caption{T-SNE, Layer 17}        \end{subfigure}  \hfill  \begin{subfigure}{0.18\textwidth}      \includegraphics[width=\textwidth]{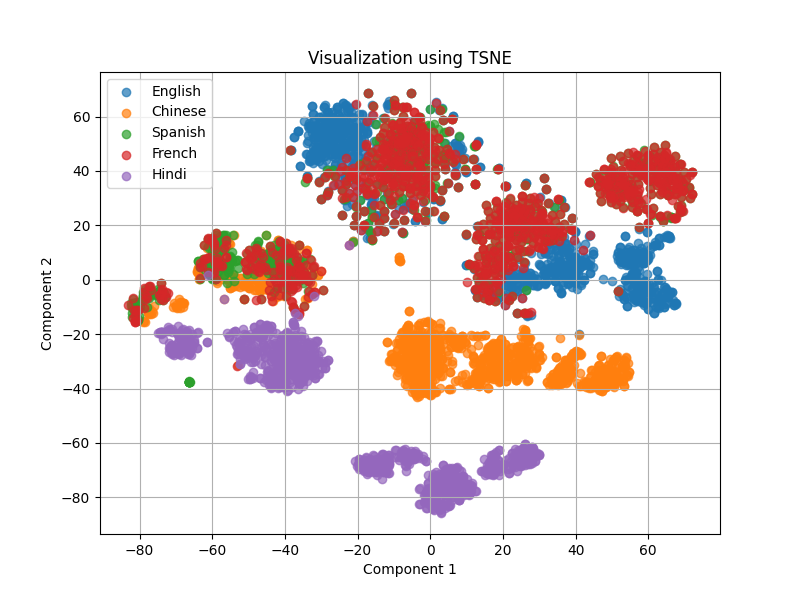}      \caption{T-SNE, Layer 18}        \end{subfigure}  \hfill  \begin{subfigure}{0.18\textwidth}      \includegraphics[width=\textwidth]{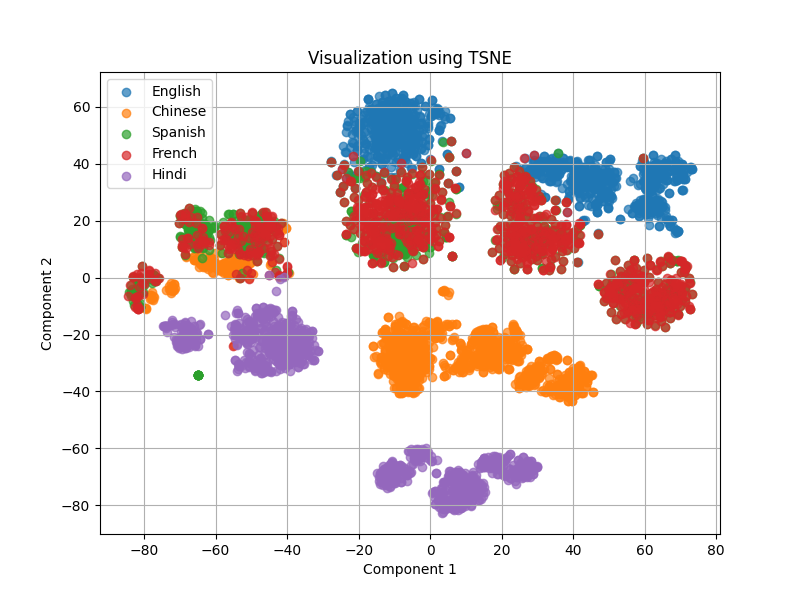}      \caption{T-SNE, Layer 19}        \end{subfigure}  \hfill  \begin{subfigure}{0.18\textwidth}      \includegraphics[width=\textwidth]{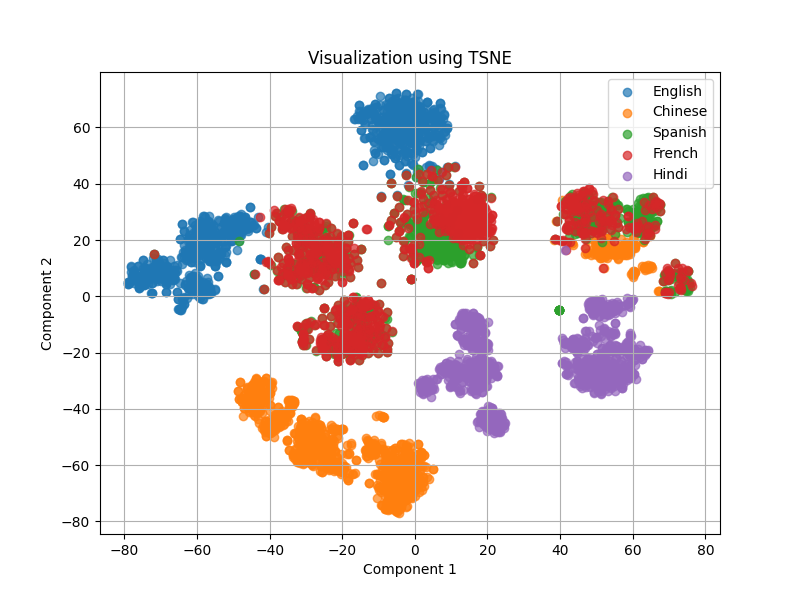}      \caption{T-SNE, Layer 20}        \end{subfigure}    \vspace{0.2in}    %
\begin{subfigure}{0.18\textwidth}      \includegraphics[width=\textwidth]{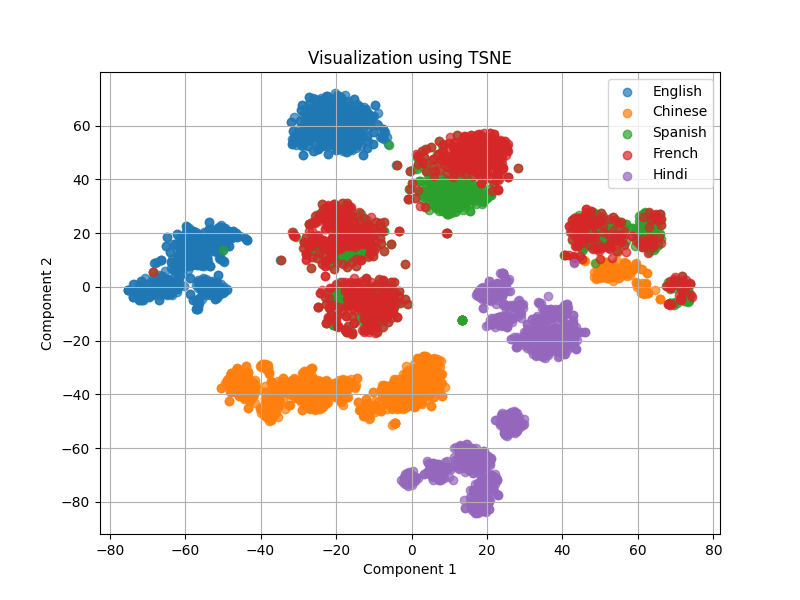}      \caption{T-SNE, Layer 21}        \end{subfigure}  \hfill  \begin{subfigure}{0.18\textwidth}      \includegraphics[width=\textwidth]{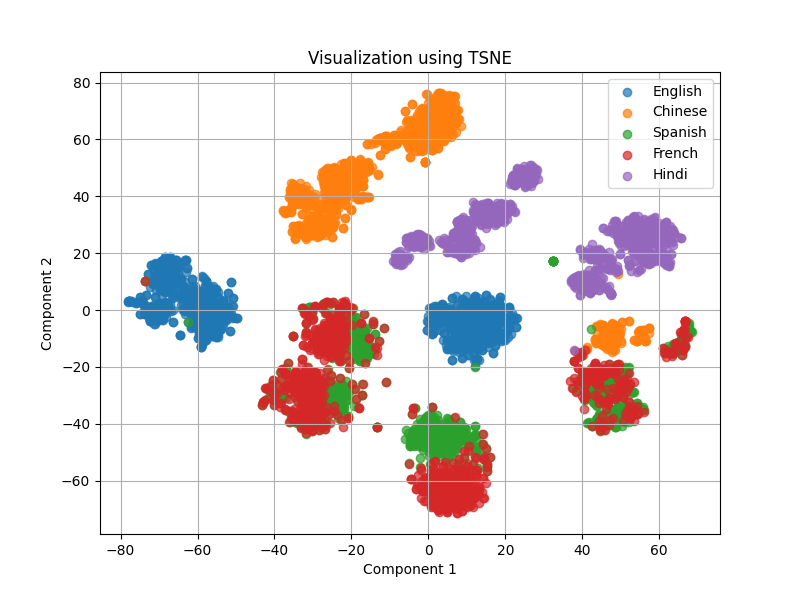}      \caption{T-SNE, Layer 22}        \end{subfigure}  \hfill  \begin{subfigure}{0.18\textwidth}      \includegraphics[width=\textwidth]{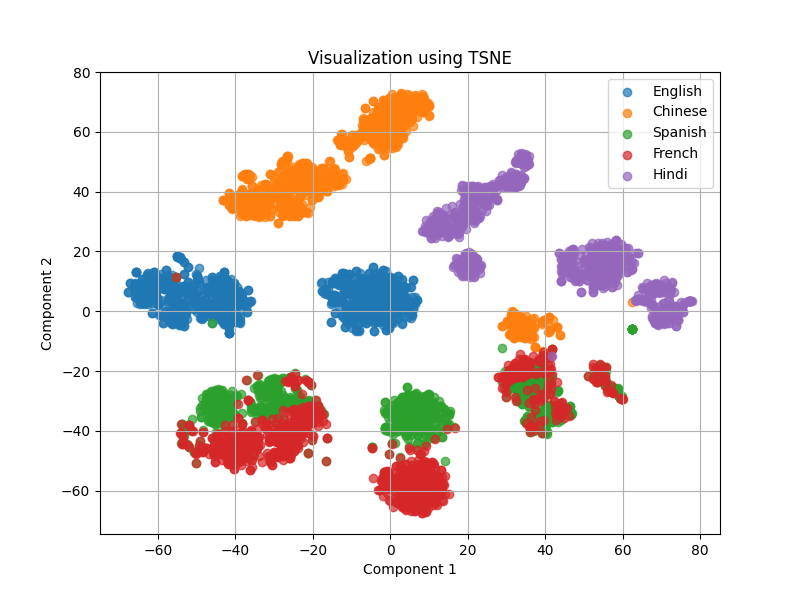}      \caption{T-SNE, Layer 23}        \end{subfigure}  \hfill  \begin{subfigure}{0.18\textwidth}      \includegraphics[width=\textwidth]{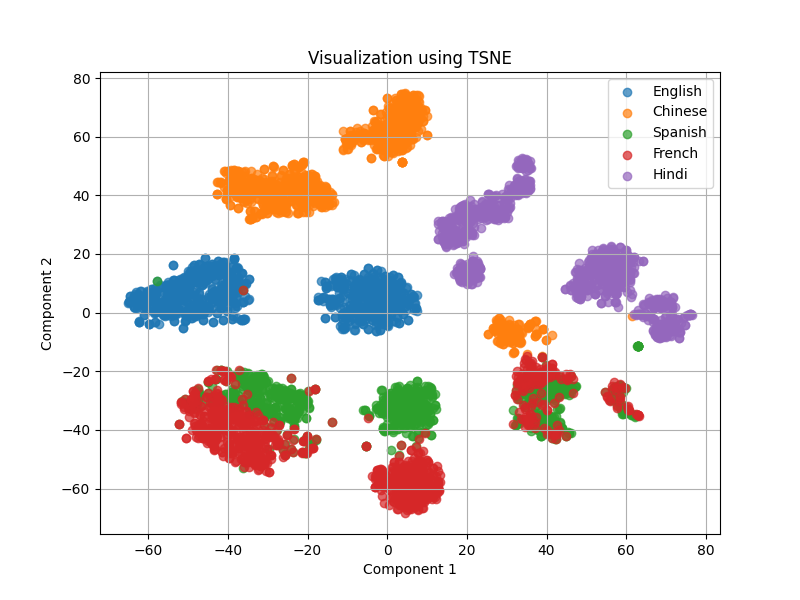}      \caption{T-SNE, Layer 24}        \end{subfigure}  \hfill  \begin{subfigure}{0.18\textwidth}      \includegraphics[width=\textwidth]{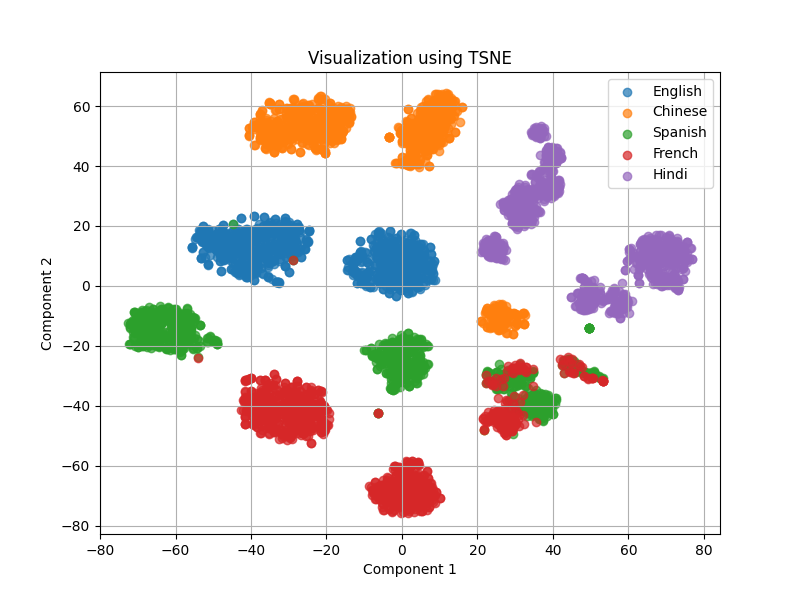}      \caption{T-SNE, Layer 25}        \end{subfigure}    \vspace{0.2in}    %
\begin{subfigure}{0.18\textwidth}      \includegraphics[width=\textwidth]{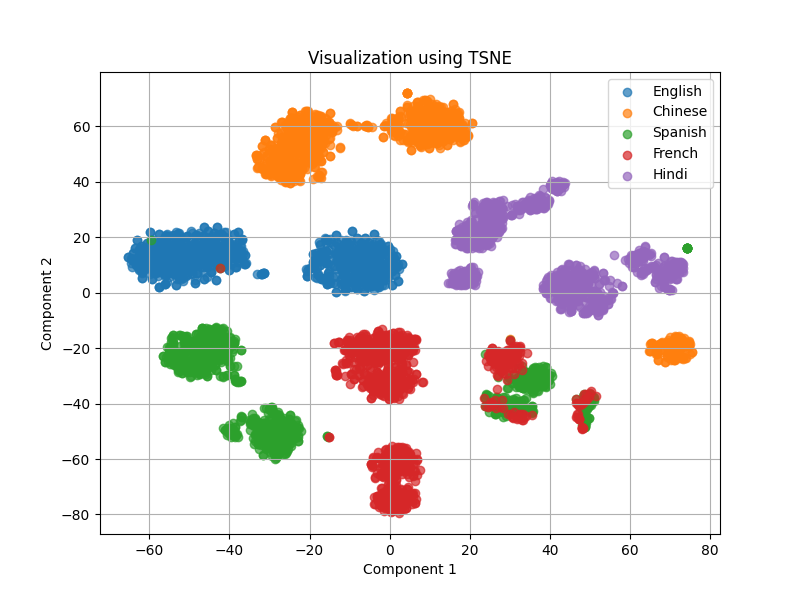}      \caption{T-SNE, Layer 26}        \end{subfigure}  \hfill  \begin{subfigure}{0.18\textwidth}      \includegraphics[width=\textwidth]{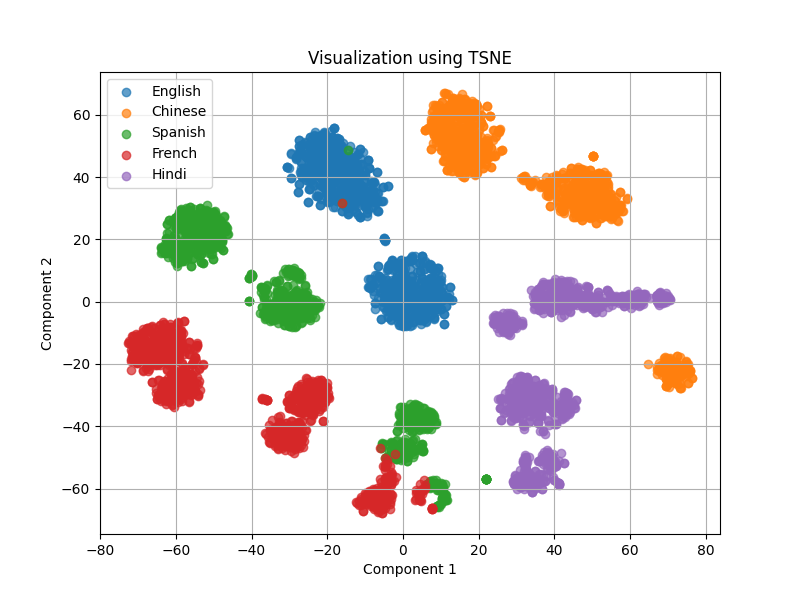}      \caption{T-SNE, Layer 27}        \end{subfigure}  \hfill  \begin{subfigure}{0.18\textwidth}      \includegraphics[width=\textwidth]{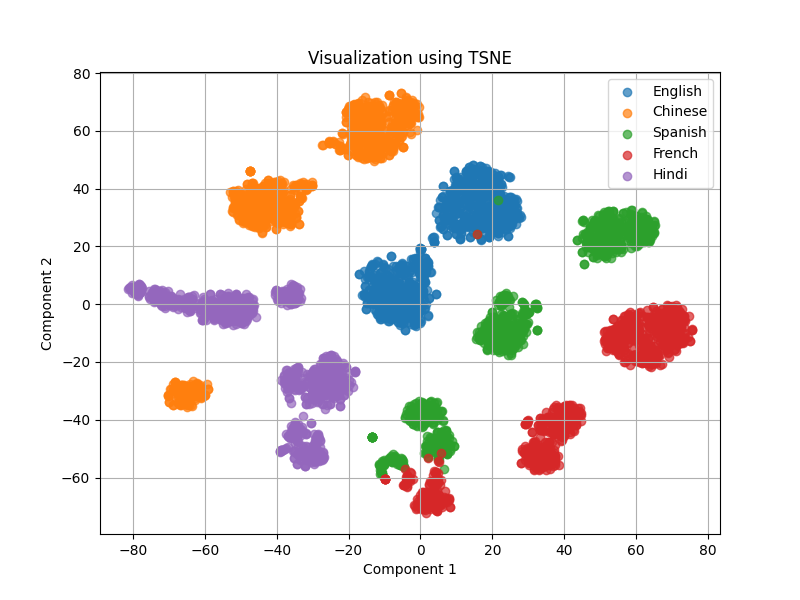}      \caption{T-SNE, Layer 28}        \end{subfigure}  \hfill  \begin{subfigure}{0.18\textwidth}      \includegraphics[width=\textwidth]{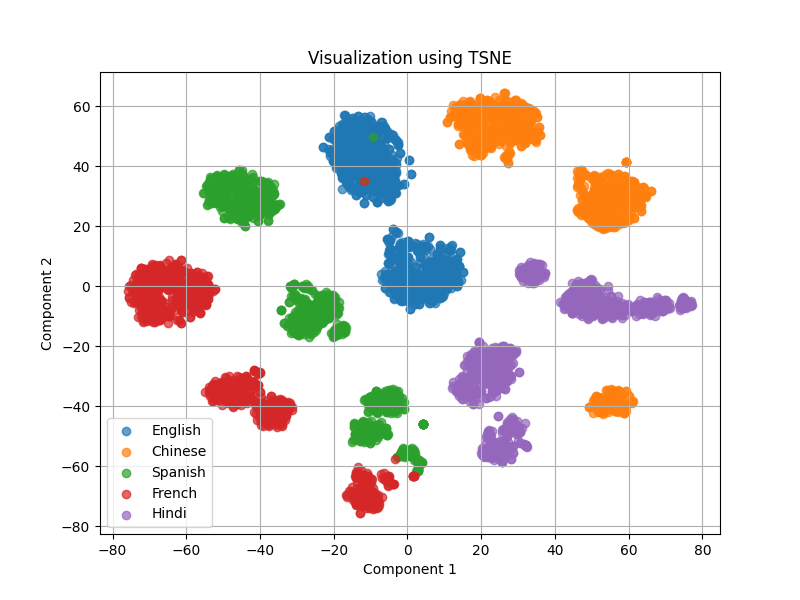}      \caption{T-SNE, Layer 29}        \end{subfigure}  \hfill  \begin{subfigure}{0.18\textwidth}      \includegraphics[width=\textwidth]{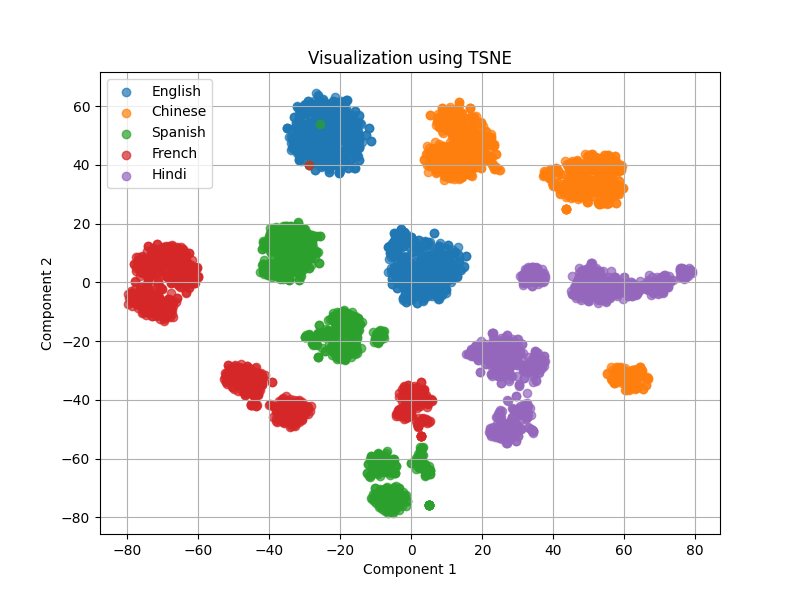}      \caption{T-SNE, Layer 30}        \end{subfigure}    \vspace{0.2in}    %
\begin{subfigure}{0.18\textwidth}      \includegraphics[width=\textwidth]{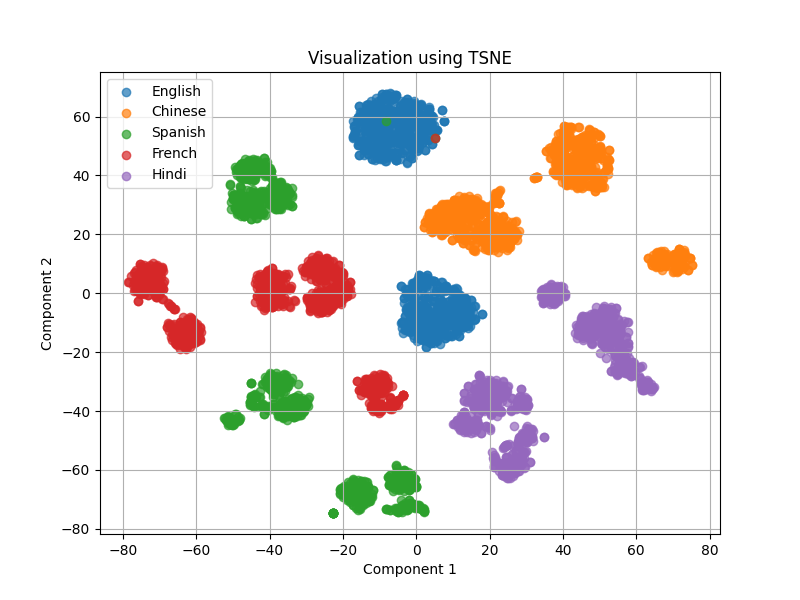}      \caption{T-SNE, Layer 31}        \end{subfigure}  \hfill  \begin{subfigure}{0.18\textwidth}      \includegraphics[width=\textwidth]{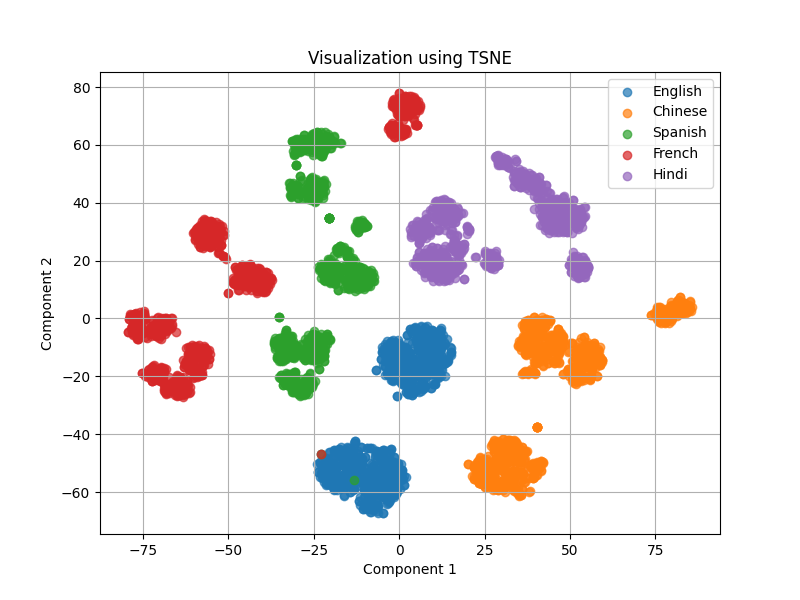}      \caption{T-SNE, Layer 32}        \end{subfigure}    \caption{T-SNE visualizations for layers 1-32 of Llama-3-8B-Instruct on the ProofWriter dataset.}  
\end{figure*}

\begin{figure*}[htbp]
\centering
\begin{subfigure}{0.18\textwidth}
\includegraphics[width=\textwidth]{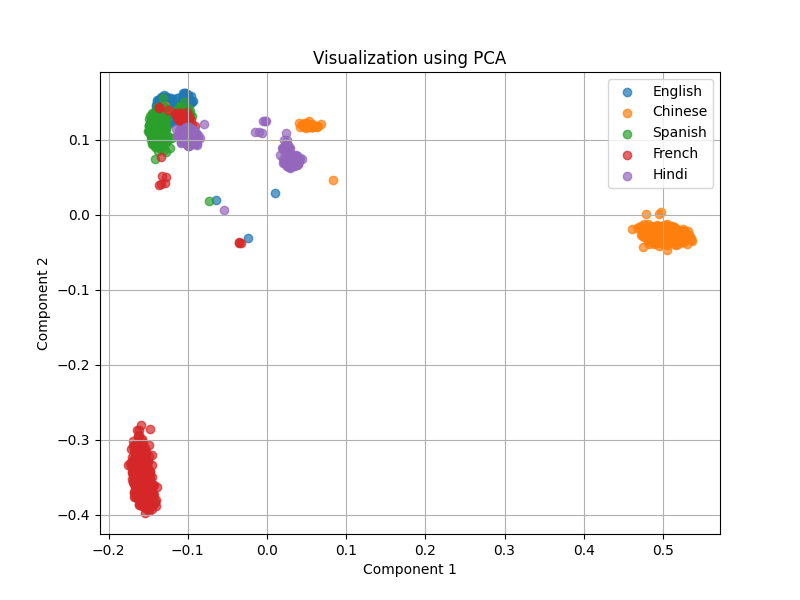}
\caption{PCA, Layer 1}
\end{subfigure}
\hfill
\begin{subfigure}{0.18\textwidth}
\includegraphics[width=\textwidth]{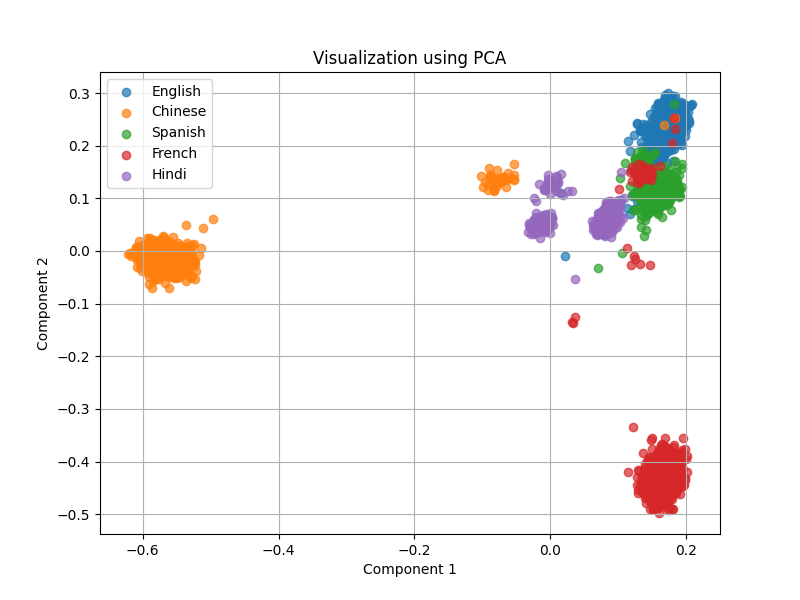}
\caption{PCA, Layer 2}

\end{subfigure}
\hfill
\begin{subfigure}{0.18\textwidth}
\includegraphics[width=\textwidth]{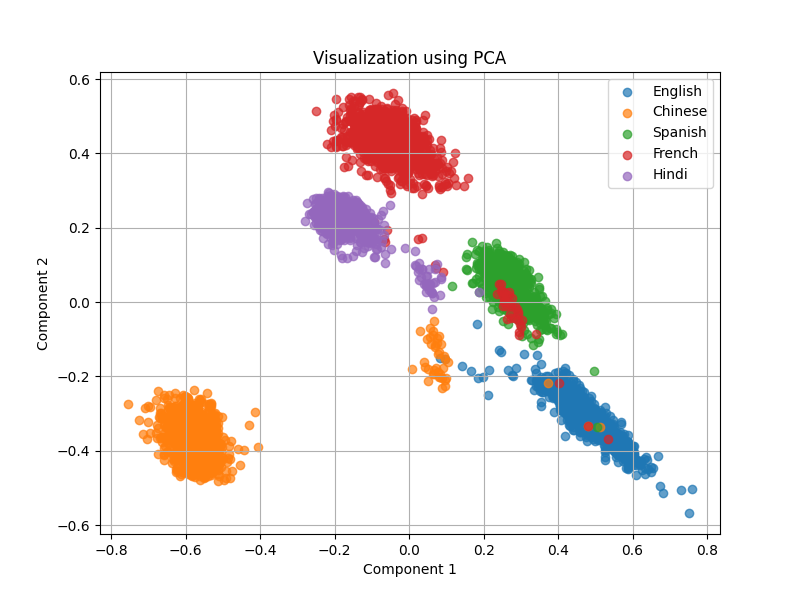}
\caption{PCA, Layer 3}

\end{subfigure}
\hfill
\begin{subfigure}{0.18\textwidth}
\includegraphics[width=\textwidth]{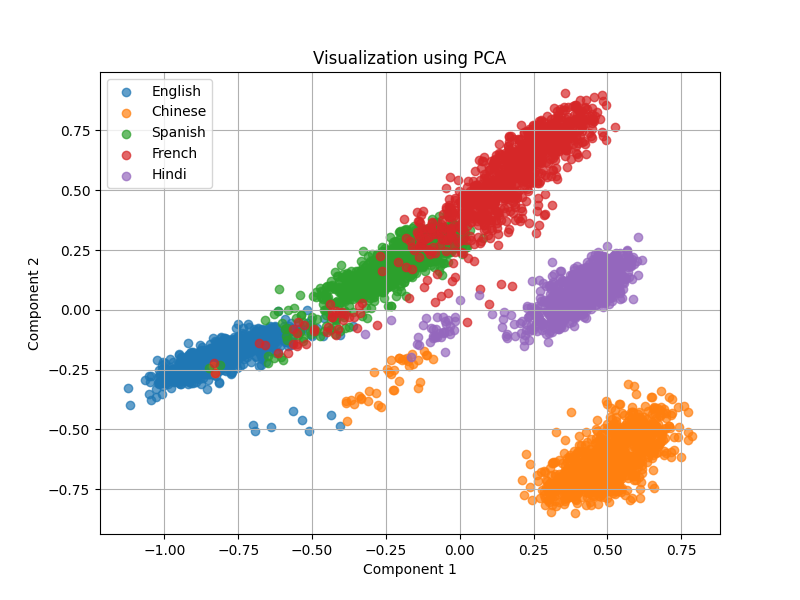}
\caption{PCA, Layer 4}

\end{subfigure}
\hfill
\begin{subfigure}{0.18\textwidth}
\includegraphics[width=\textwidth]{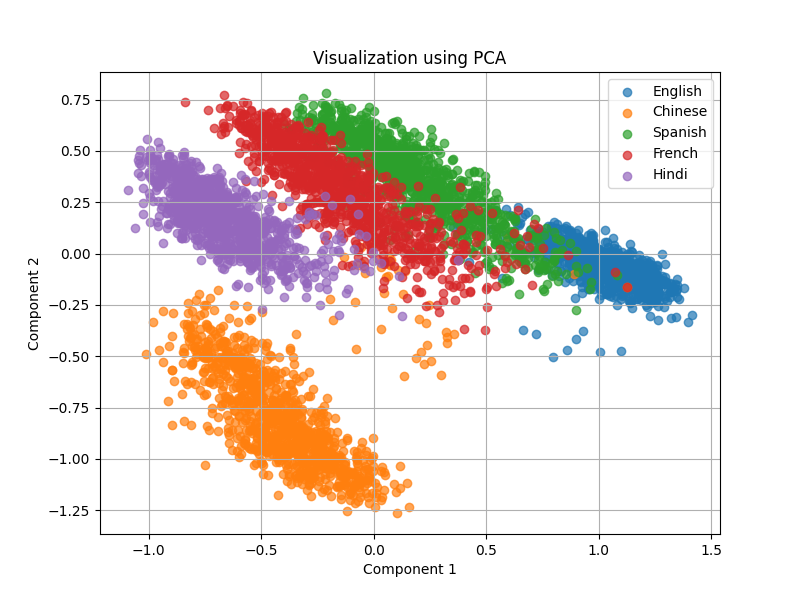}
\caption{PCA, Layer 5}

\end{subfigure}
\vspace{0.2in} %
\begin{subfigure}{0.18\textwidth}      \includegraphics[width=\textwidth]{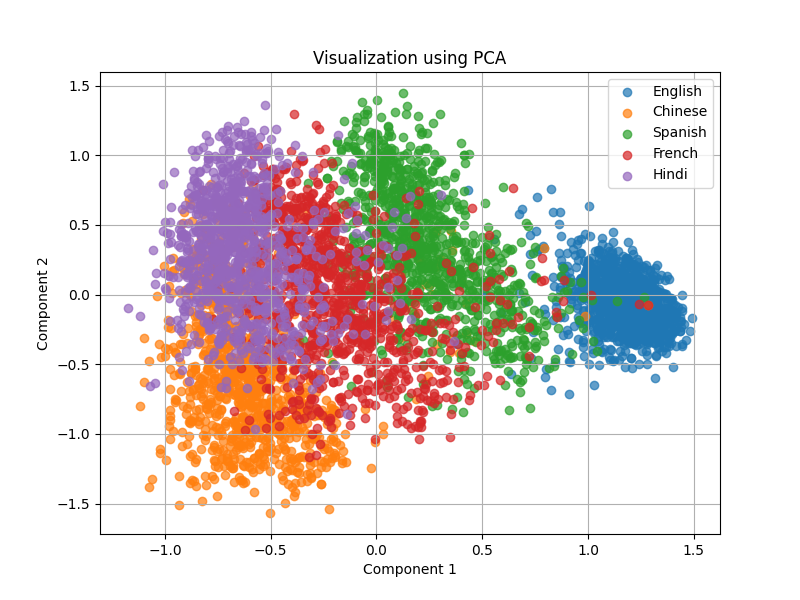}      \caption{PCA, Layer 6}        \end{subfigure}  \hfill  \begin{subfigure}{0.18\textwidth}      \includegraphics[width=\textwidth]{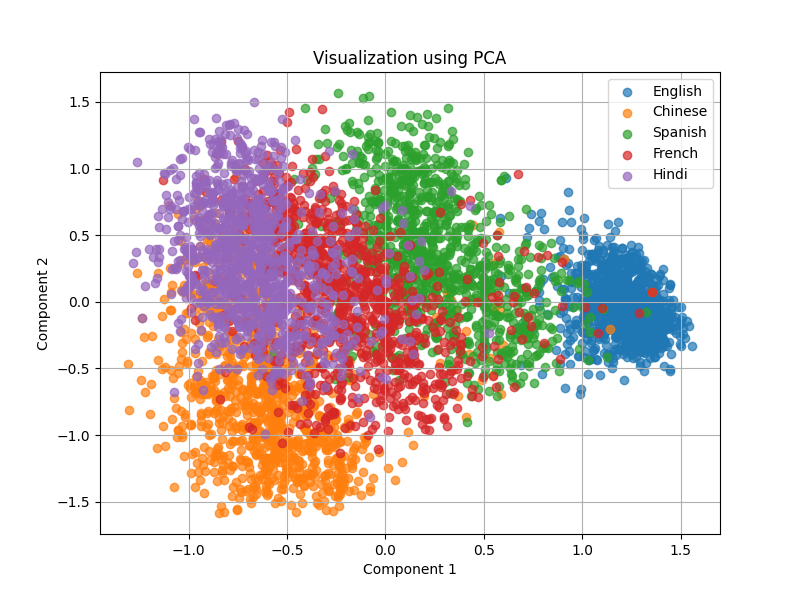}      \caption{PCA, Layer 7}        \end{subfigure}  \hfill  \begin{subfigure}{0.18\textwidth}      \includegraphics[width=\textwidth]{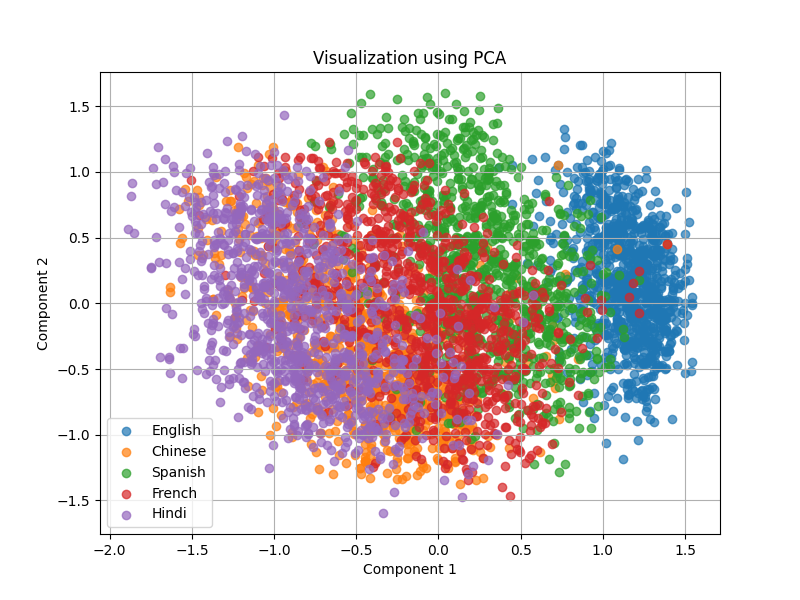}      \caption{PCA, Layer 8}        \end{subfigure}  \hfill  \begin{subfigure}{0.18\textwidth}      \includegraphics[width=\textwidth]{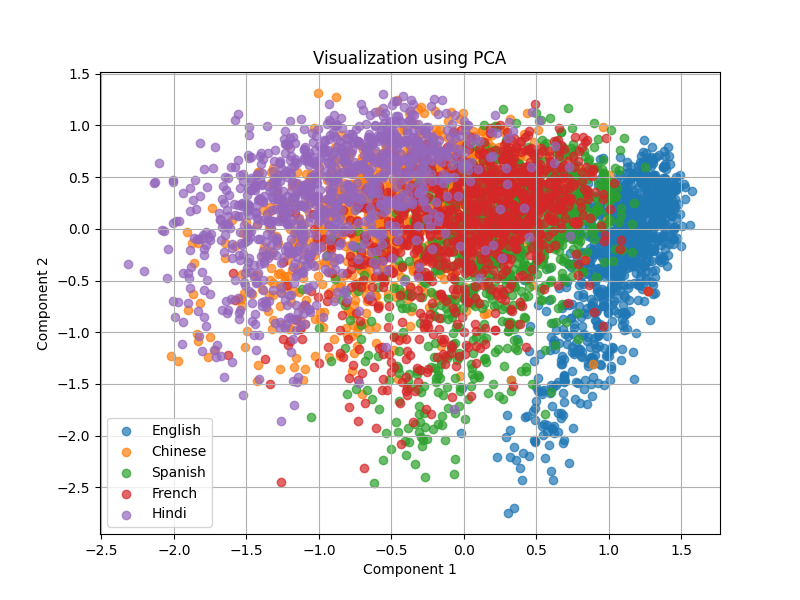}      \caption{PCA, Layer 9}        \end{subfigure}  \hfill  \begin{subfigure}{0.18\textwidth}      \includegraphics[width=\textwidth]{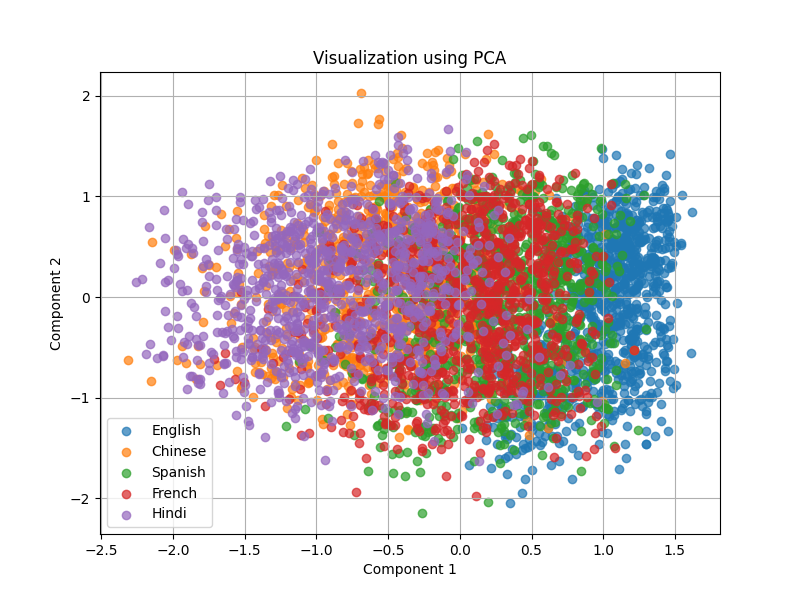}      \caption{PCA, Layer 10}        \end{subfigure}    \vspace{0.2in}    %
\begin{subfigure}{0.18\textwidth}      \includegraphics[width=\textwidth]{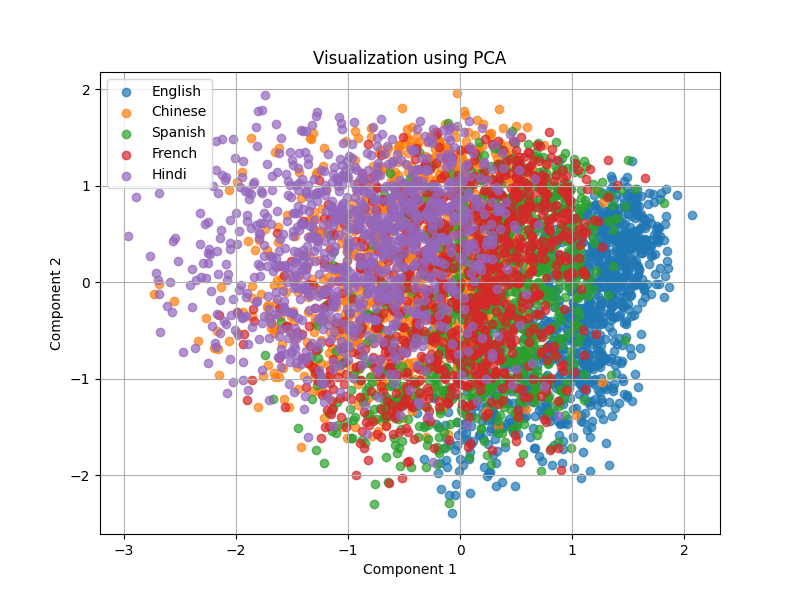}      \caption{PCA, Layer 11}        \end{subfigure}  \hfill  \begin{subfigure}{0.18\textwidth}      \includegraphics[width=\textwidth]{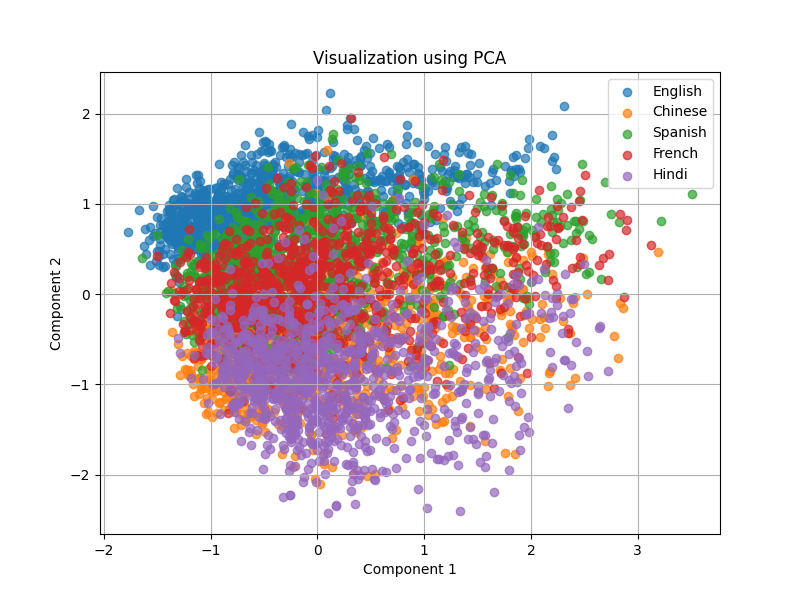}      \caption{PCA, Layer 12}        \end{subfigure}  \hfill  \begin{subfigure}{0.18\textwidth}      \includegraphics[width=\textwidth]{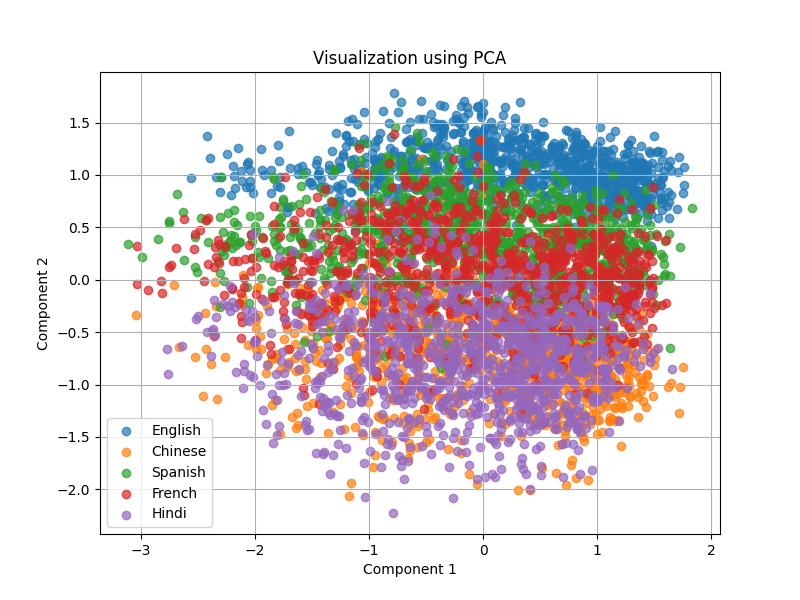}      \caption{PCA, Layer 13}        \end{subfigure}  \hfill  \begin{subfigure}{0.18\textwidth}      \includegraphics[width=\textwidth]{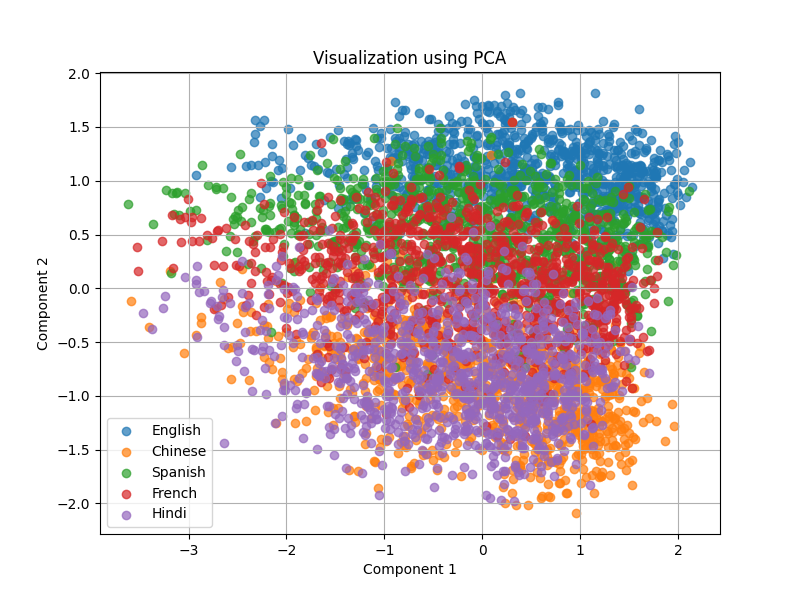}      \caption{PCA, Layer 14}        \end{subfigure}  \hfill  \begin{subfigure}{0.18\textwidth}      \includegraphics[width=\textwidth]{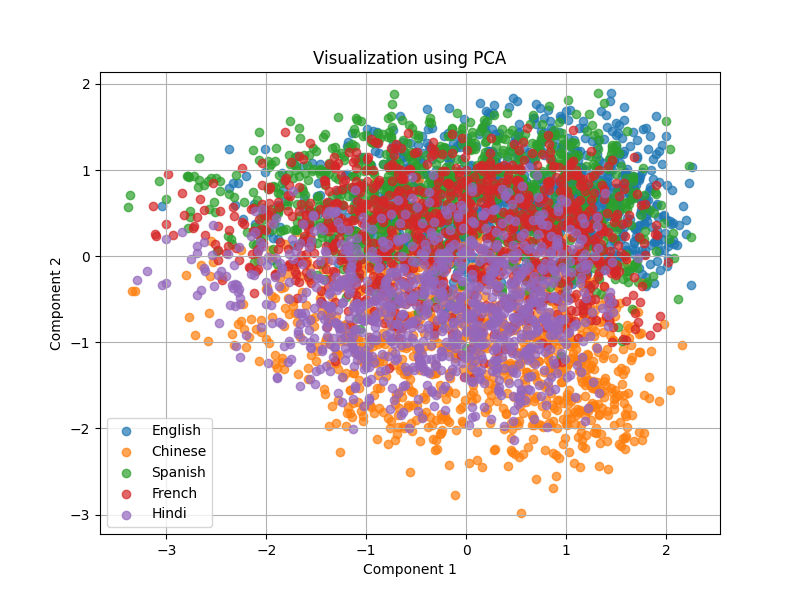}      \caption{PCA, Layer 15}        \end{subfigure}    \vspace{0.2in}    %
\begin{subfigure}{0.18\textwidth}      \includegraphics[width=\textwidth]{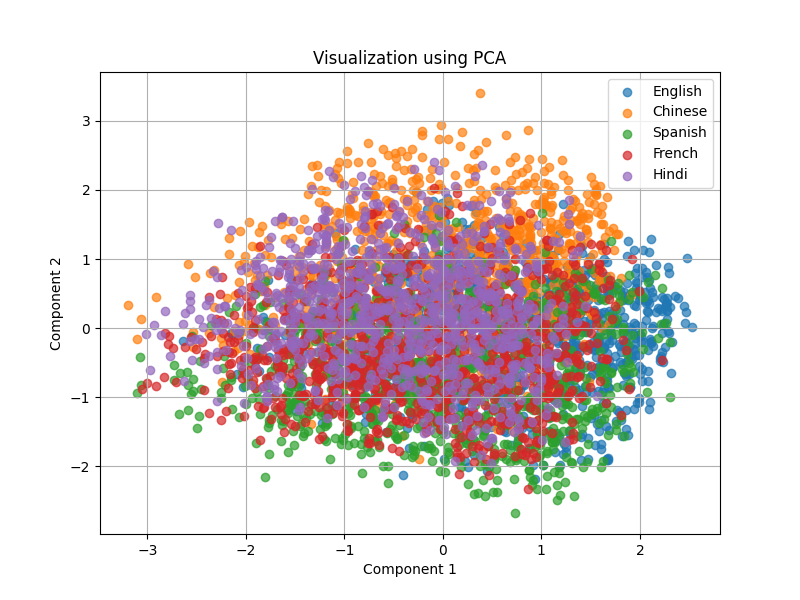}      \caption{PCA, Layer 16}        \end{subfigure}  \hfill  \begin{subfigure}{0.18\textwidth}      \includegraphics[width=\textwidth]{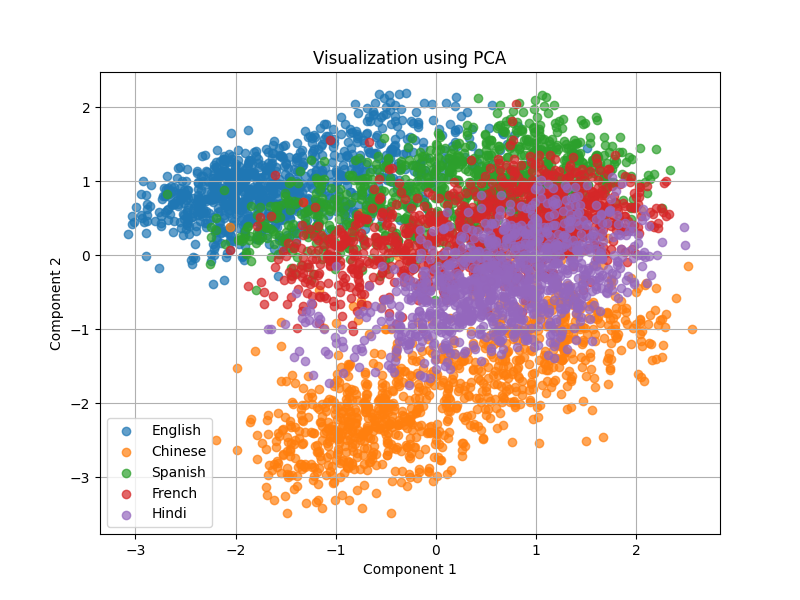}      \caption{PCA, Layer 17}        \end{subfigure}  \hfill  \begin{subfigure}{0.18\textwidth}      \includegraphics[width=\textwidth]{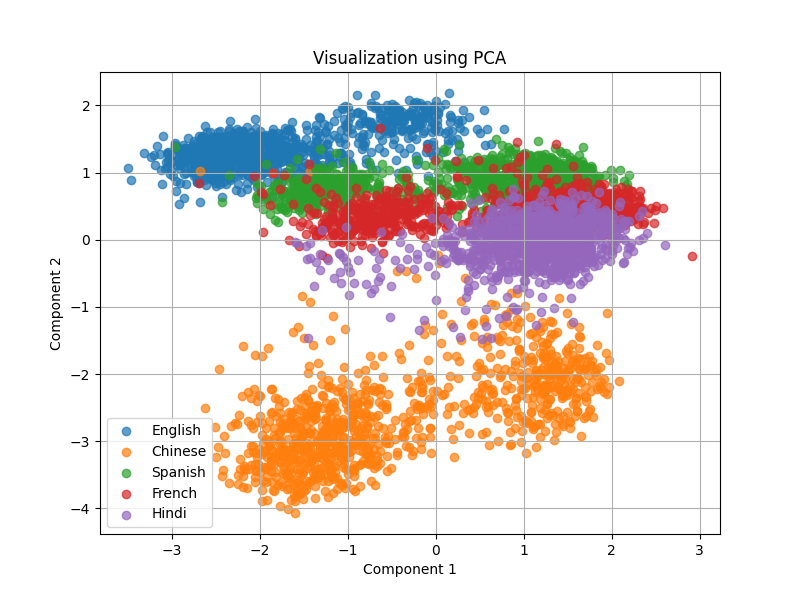}      \caption{PCA, Layer 18}        \end{subfigure}  \hfill  \begin{subfigure}{0.18\textwidth}      \includegraphics[width=\textwidth]{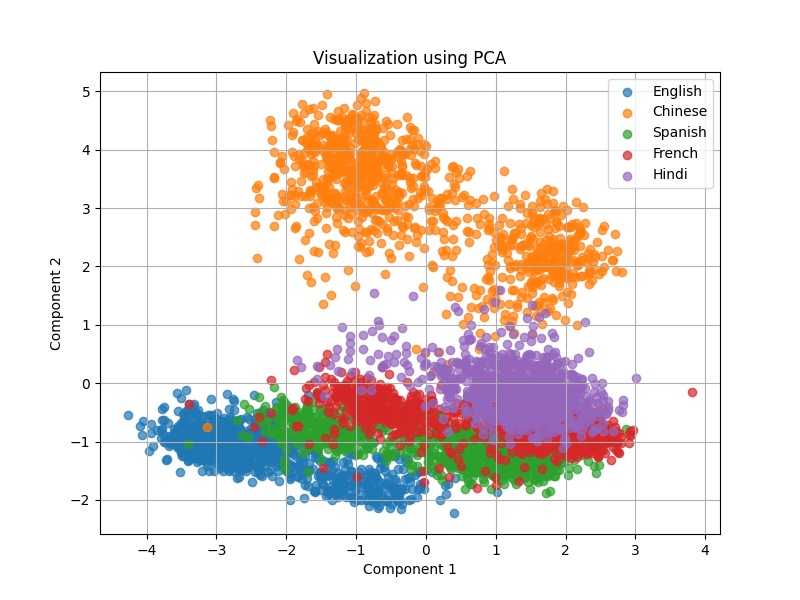}      \caption{PCA, Layer 19}        \end{subfigure}  \hfill  \begin{subfigure}{0.18\textwidth}      \includegraphics[width=\textwidth]{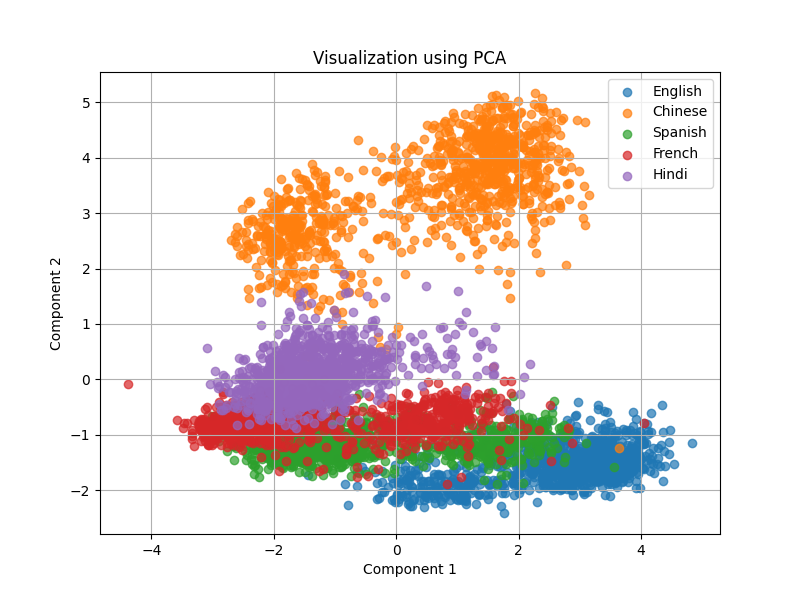}      \caption{PCA, Layer 20}        \end{subfigure}    \vspace{0.2in}    %
\begin{subfigure}{0.18\textwidth}      \includegraphics[width=\textwidth]{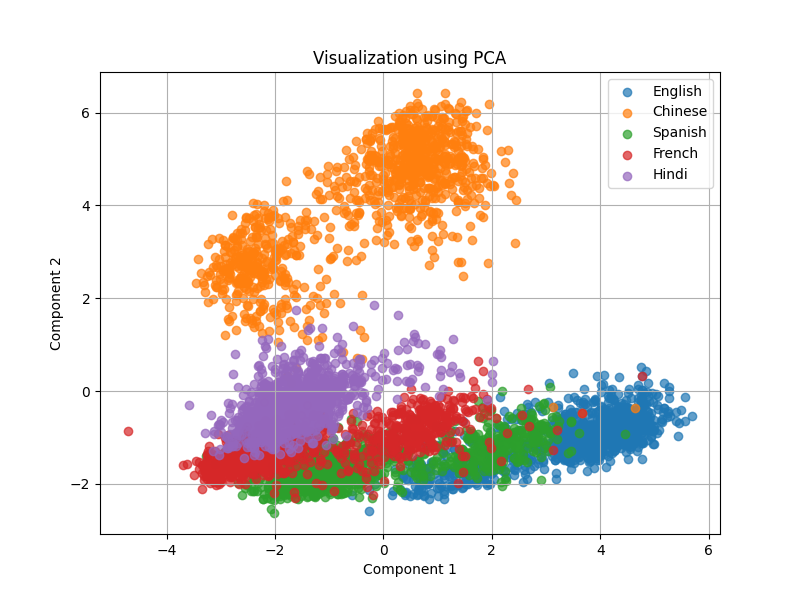}      \caption{PCA, Layer 21}        \end{subfigure}  \hfill  \begin{subfigure}{0.18\textwidth}      \includegraphics[width=\textwidth]{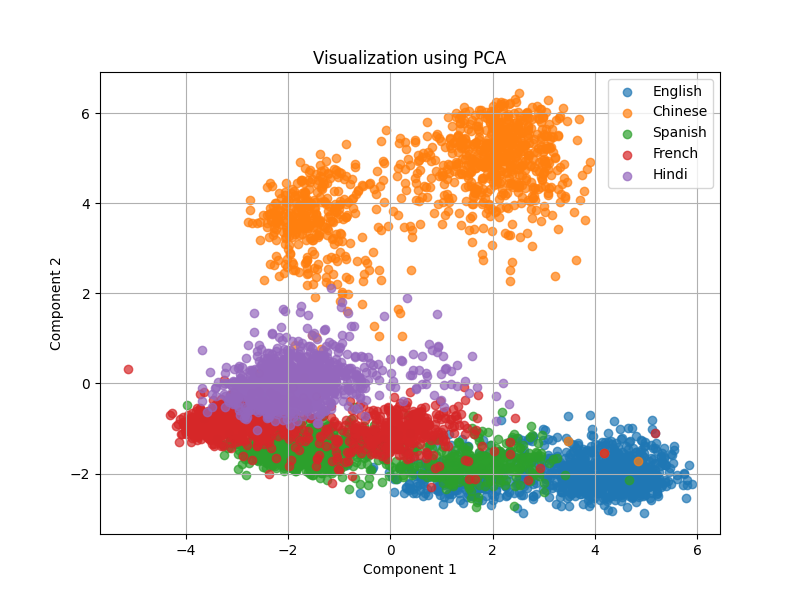}      \caption{PCA, Layer 22}        \end{subfigure}  \hfill  \begin{subfigure}{0.18\textwidth}      \includegraphics[width=\textwidth]{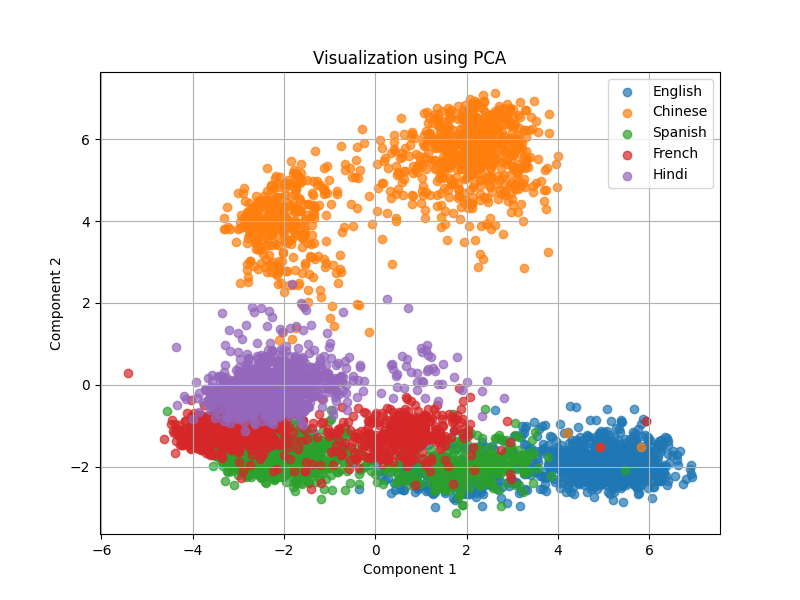}      \caption{PCA, Layer 23}        \end{subfigure}  \hfill  \begin{subfigure}{0.18\textwidth}      \includegraphics[width=\textwidth]{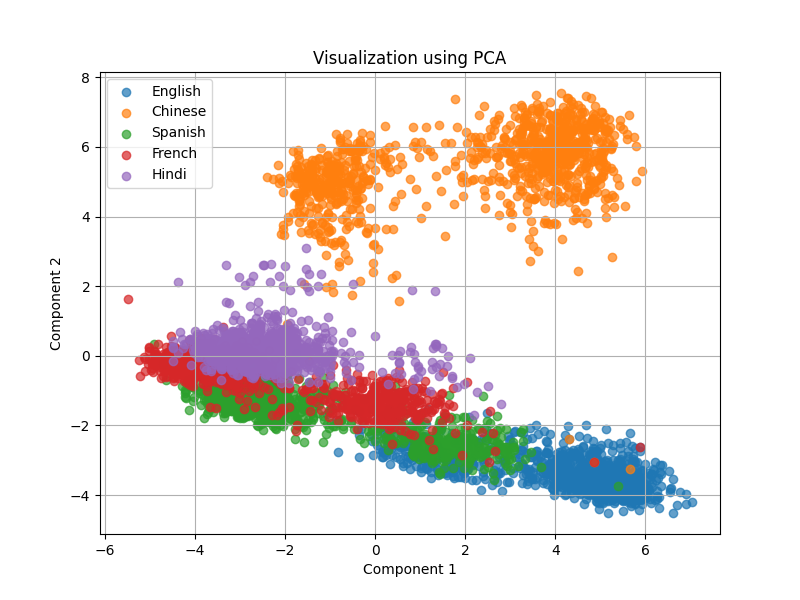}      \caption{PCA, Layer 24}        \end{subfigure}  \hfill  \begin{subfigure}{0.18\textwidth}      \includegraphics[width=\textwidth]{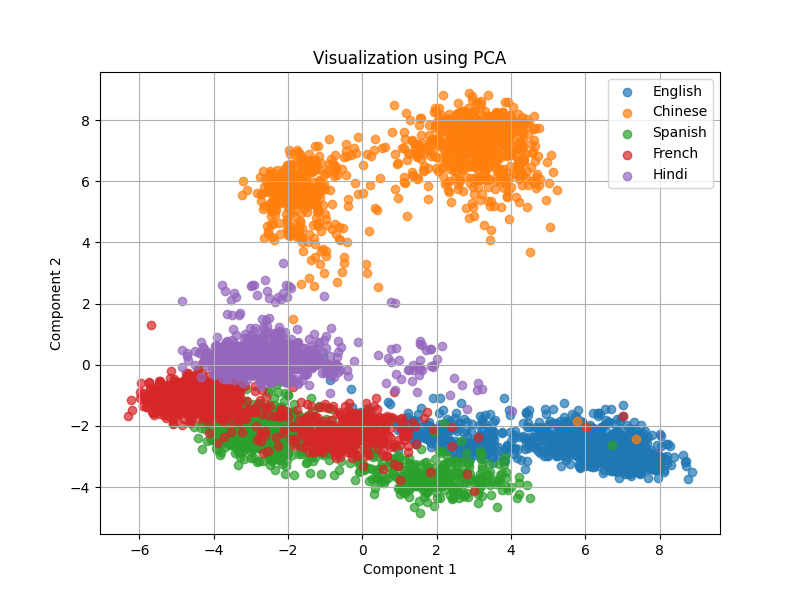}      \caption{PCA, Layer 25}        \end{subfigure}    \vspace{0.2in}    %
\begin{subfigure}{0.18\textwidth}      \includegraphics[width=\textwidth]{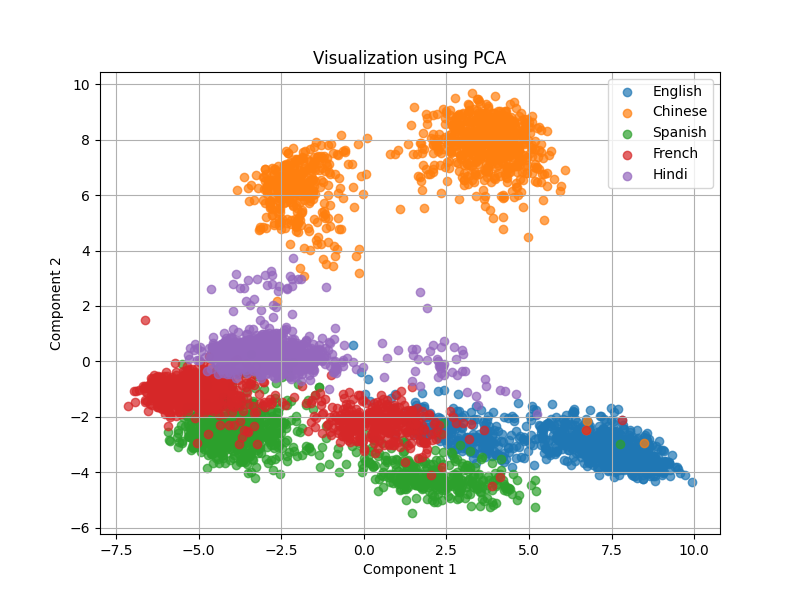}      \caption{PCA, Layer 26}        \end{subfigure}  \hfill  \begin{subfigure}{0.18\textwidth}      \includegraphics[width=\textwidth]{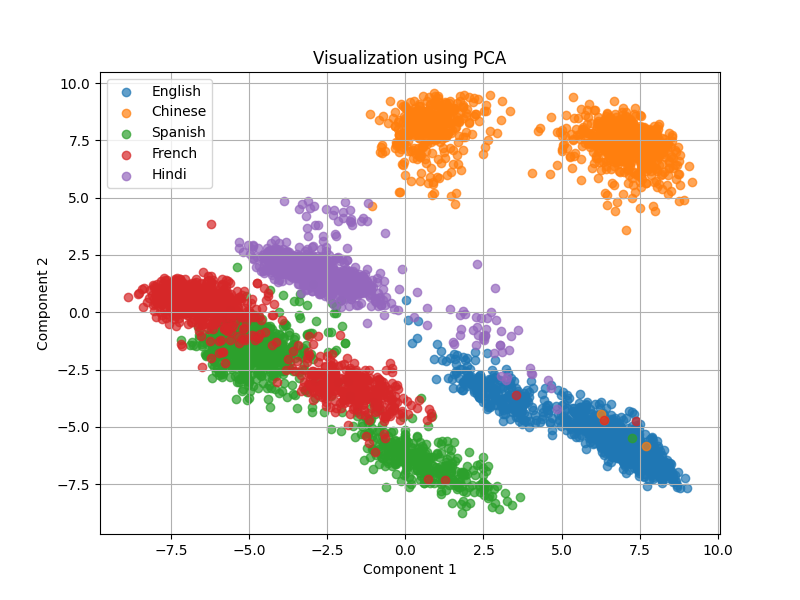}      \caption{PCA, Layer 27}        \end{subfigure}  \hfill  \begin{subfigure}{0.18\textwidth}      \includegraphics[width=\textwidth]{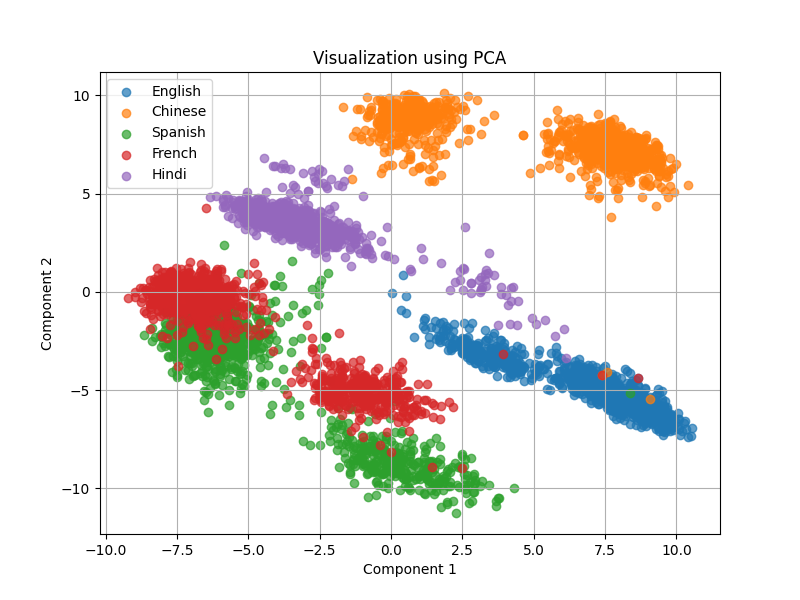}      \caption{PCA, Layer 28}        \end{subfigure}  \hfill  \begin{subfigure}{0.18\textwidth}      \includegraphics[width=\textwidth]{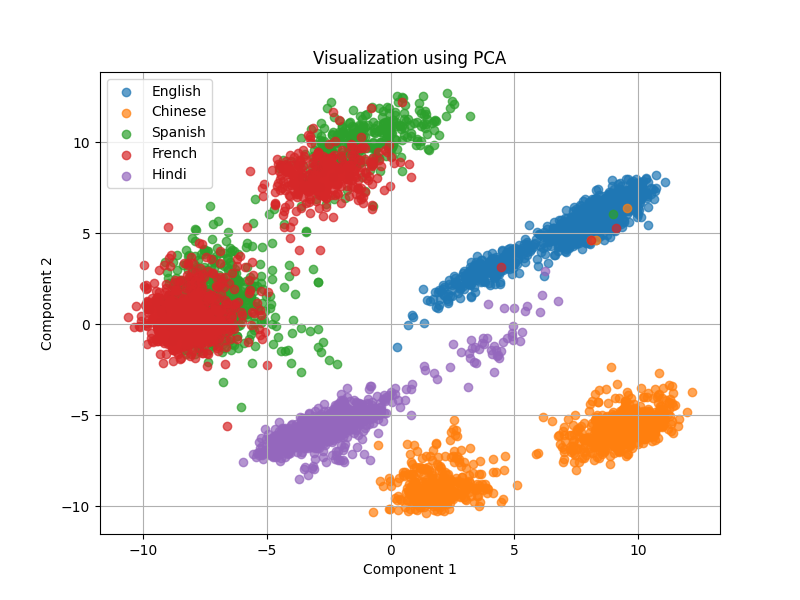}      \caption{PCA, Layer 29}        \end{subfigure}  \hfill  \begin{subfigure}{0.18\textwidth}      \includegraphics[width=\textwidth]{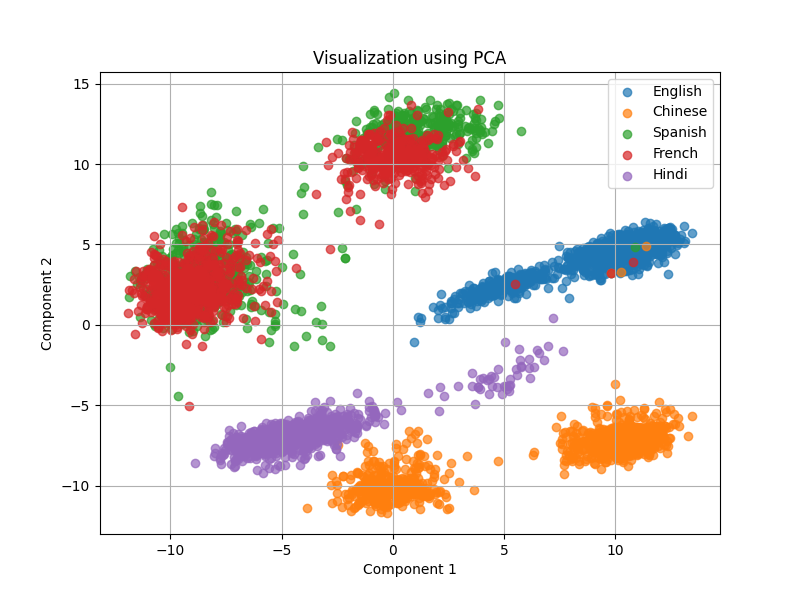}      \caption{PCA, Layer 30}        \end{subfigure}    \vspace{0.2in}    %
\begin{subfigure}{0.18\textwidth}      \includegraphics[width=\textwidth]{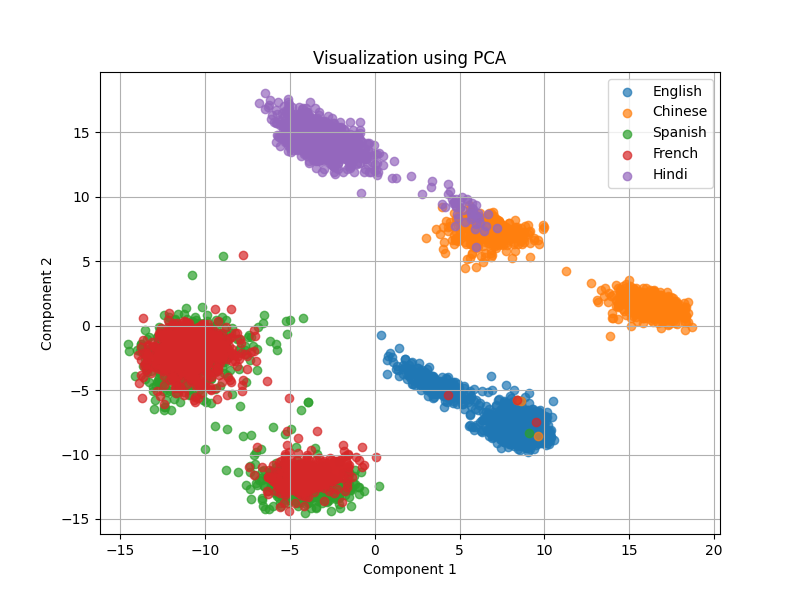}      \caption{PCA, Layer 31}        \end{subfigure}  \hfill  \begin{subfigure}{0.18\textwidth}      \includegraphics[width=\textwidth]{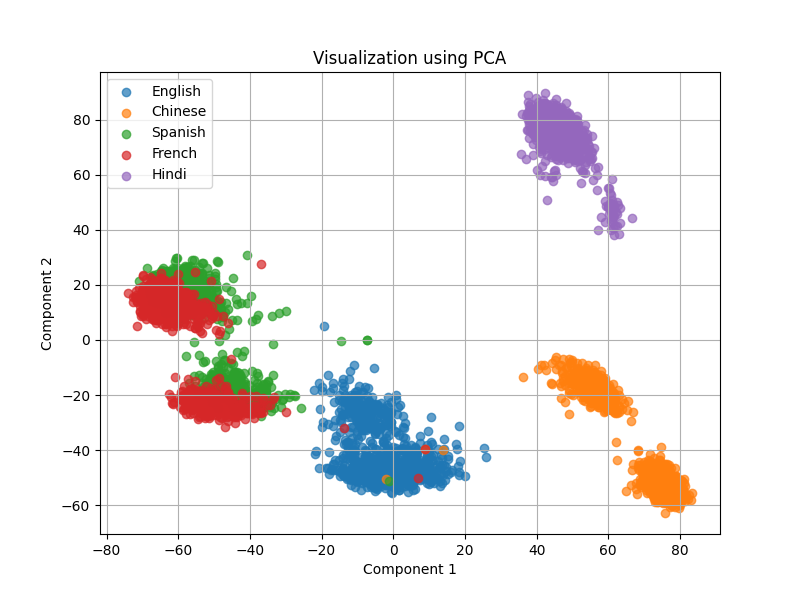}      \caption{PCA, Layer 32}        \end{subfigure}    \caption{PCA visualizations for layers 1-32 of Llama-3-8B-Instruct on the GSM8K dataset.}  
\end{figure*}

\begin{figure*}[htbp]
\centering
\begin{subfigure}{0.18\textwidth}
\includegraphics[width=\textwidth]{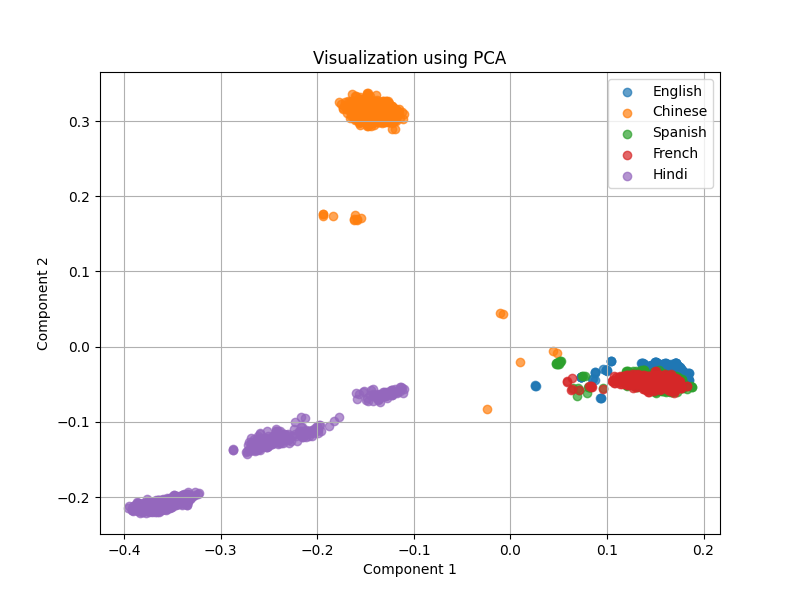}
\caption{PCA, Layer 1}
\end{subfigure}
\hfill
\begin{subfigure}{0.18\textwidth}
\includegraphics[width=\textwidth]{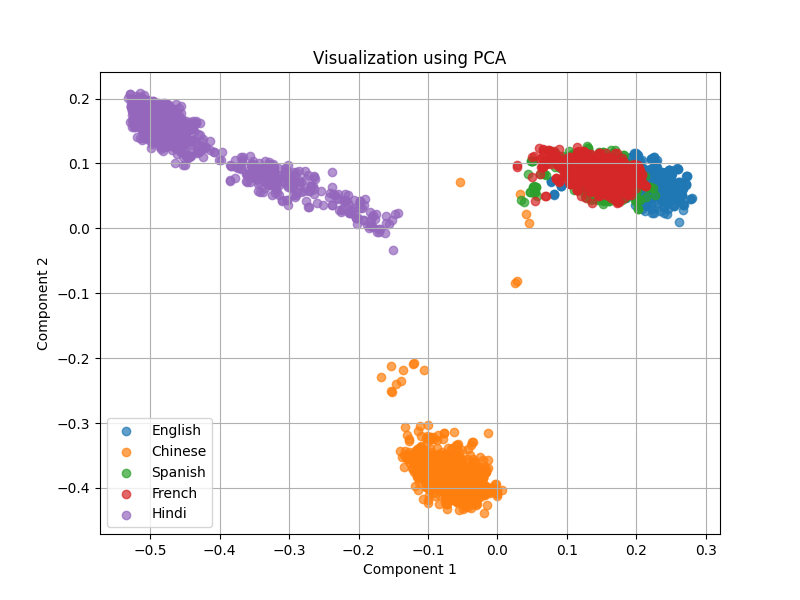}
\caption{PCA, Layer 2}

\end{subfigure}
\hfill
\begin{subfigure}{0.18\textwidth}
\includegraphics[width=\textwidth]{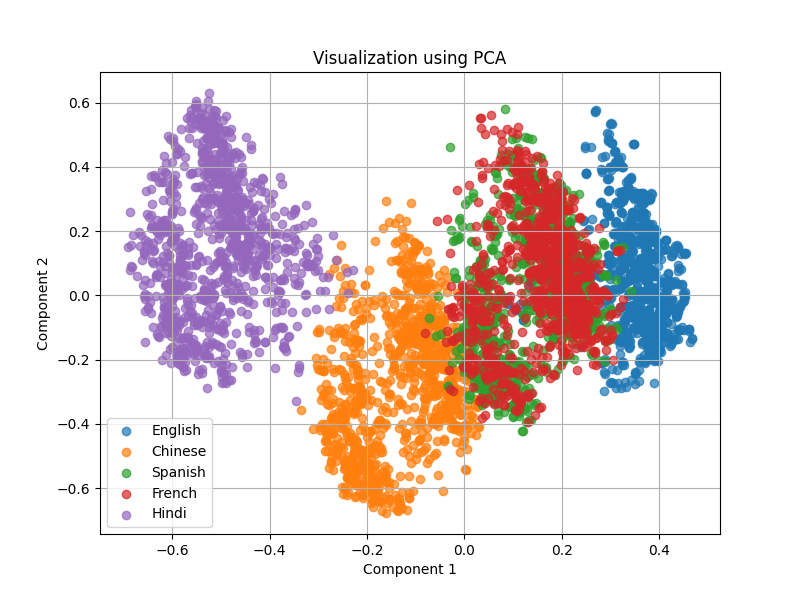}
\caption{PCA, Layer 3}

\end{subfigure}
\hfill
\begin{subfigure}{0.18\textwidth}
\includegraphics[width=\textwidth]{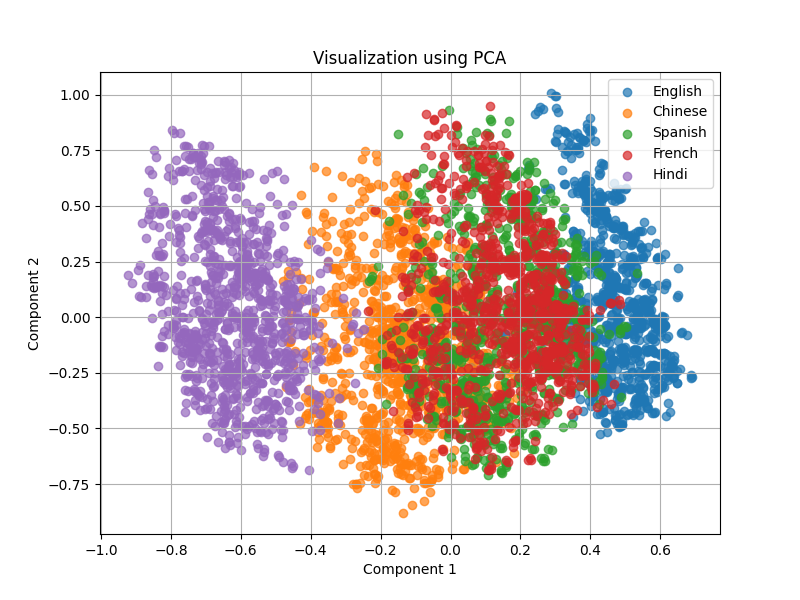}
\caption{PCA, Layer 4}

\end{subfigure}
\hfill
\begin{subfigure}{0.18\textwidth}
\includegraphics[width=\textwidth]{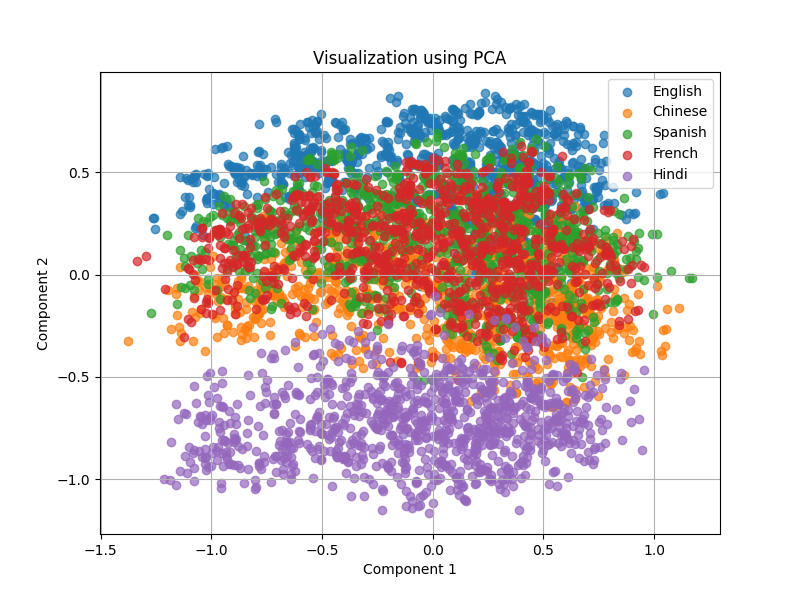}
\caption{PCA, Layer 5}

\end{subfigure}
\vspace{0.2in} %
\begin{subfigure}{0.18\textwidth}      \includegraphics[width=\textwidth]{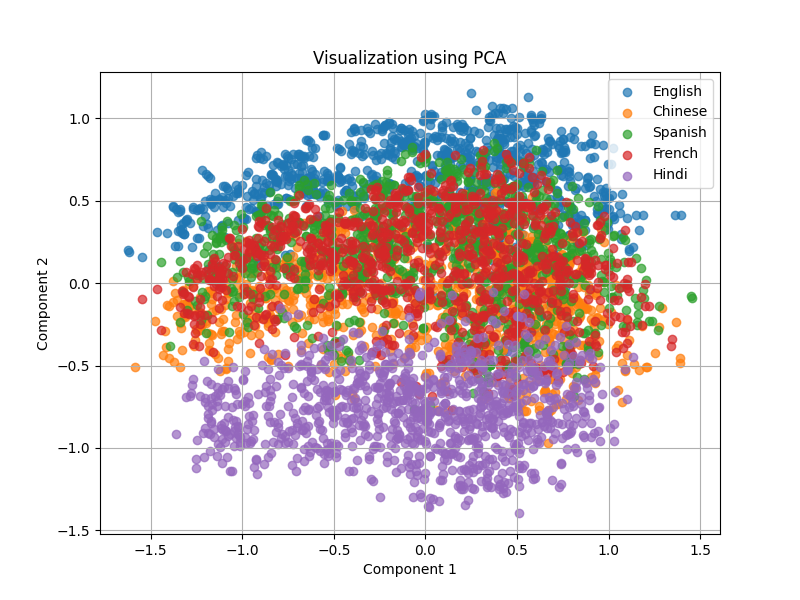}      \caption{PCA, Layer 6}        \end{subfigure}  \hfill  \begin{subfigure}{0.18\textwidth}      \includegraphics[width=\textwidth]{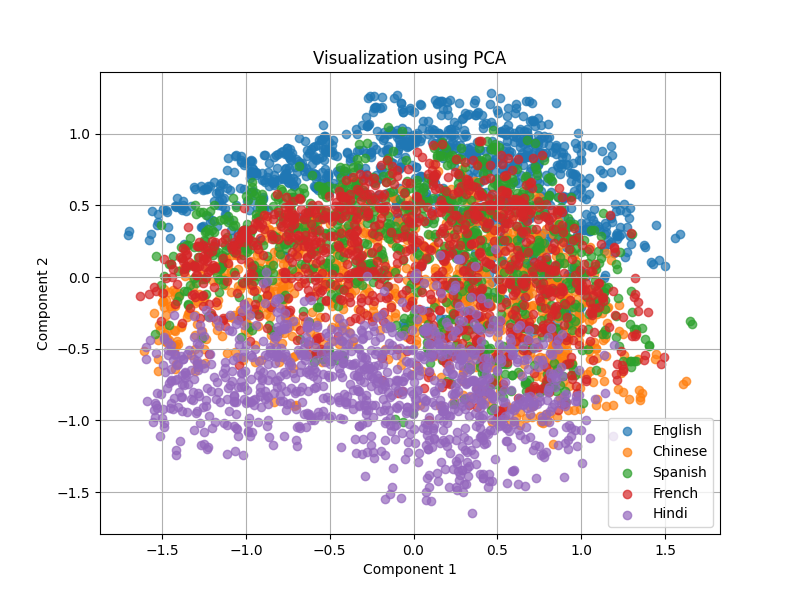}      \caption{PCA, Layer 7}        \end{subfigure}  \hfill  \begin{subfigure}{0.18\textwidth}      \includegraphics[width=\textwidth]{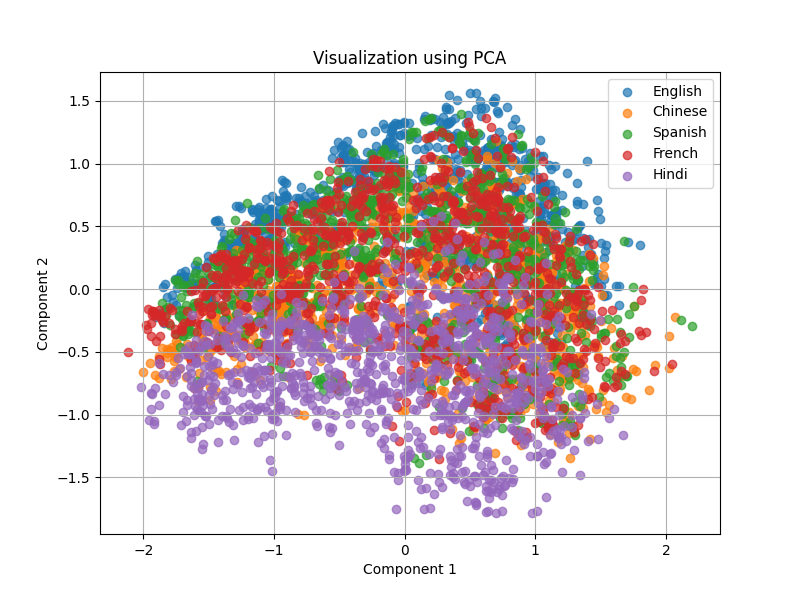}      \caption{PCA, Layer 8}        \end{subfigure}  \hfill  \begin{subfigure}{0.18\textwidth}      \includegraphics[width=\textwidth]{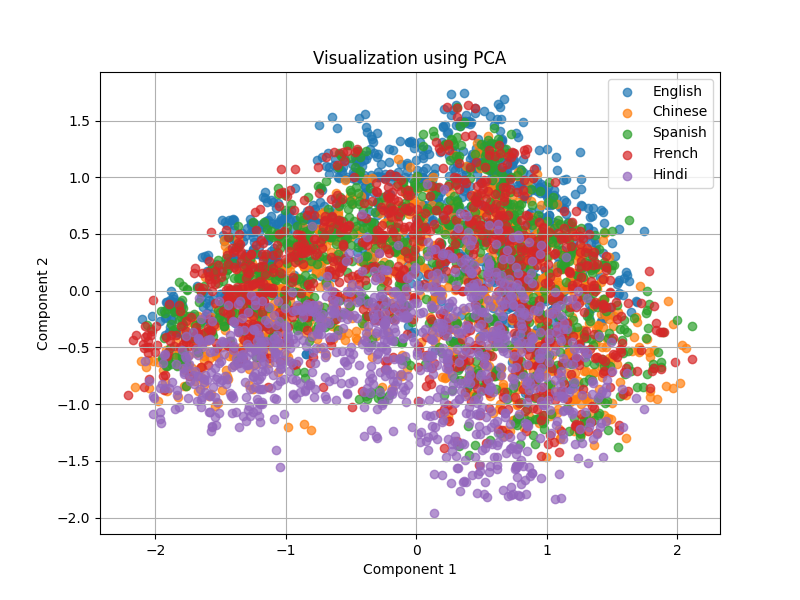}      \caption{PCA, Layer 9}        \end{subfigure}  \hfill  \begin{subfigure}{0.18\textwidth}      \includegraphics[width=\textwidth]{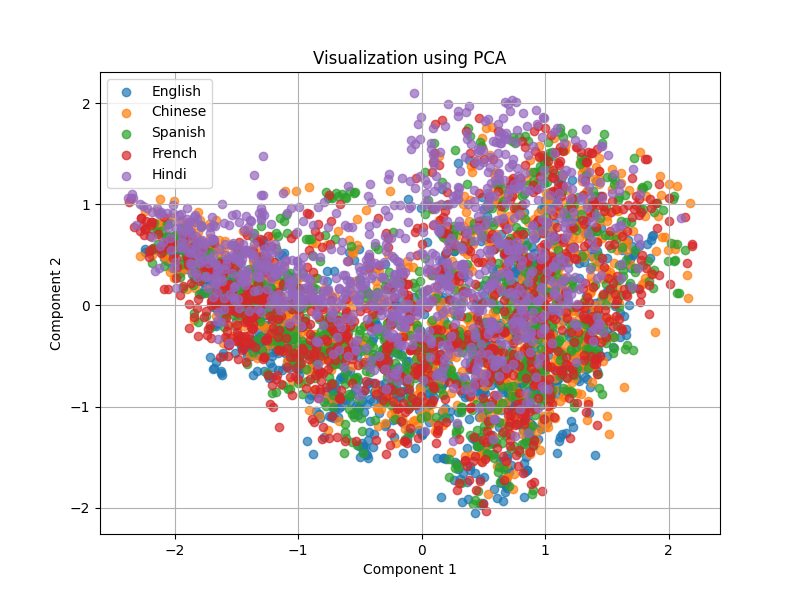}      \caption{PCA, Layer 10}        \end{subfigure}    \vspace{0.2in}    %
\begin{subfigure}{0.18\textwidth}      \includegraphics[width=\textwidth]{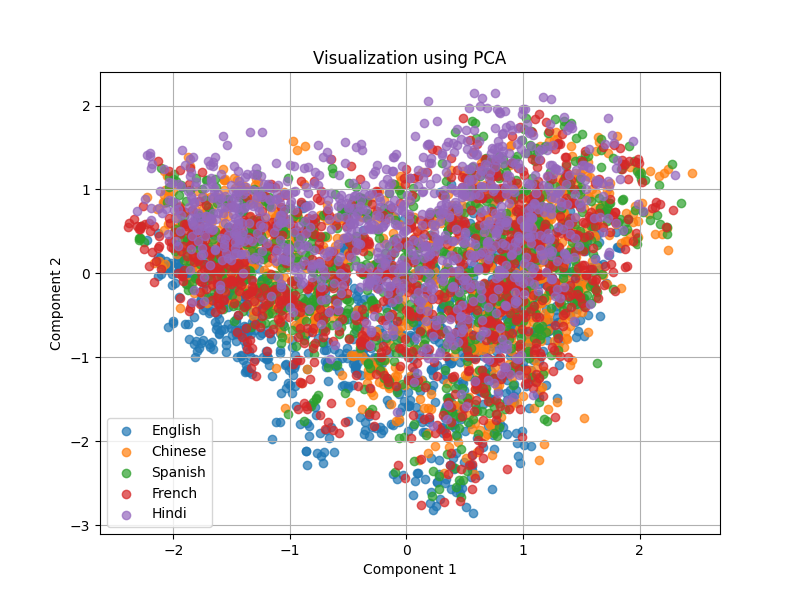}      \caption{PCA, Layer 11}        \end{subfigure}  \hfill  \begin{subfigure}{0.18\textwidth}      \includegraphics[width=\textwidth]{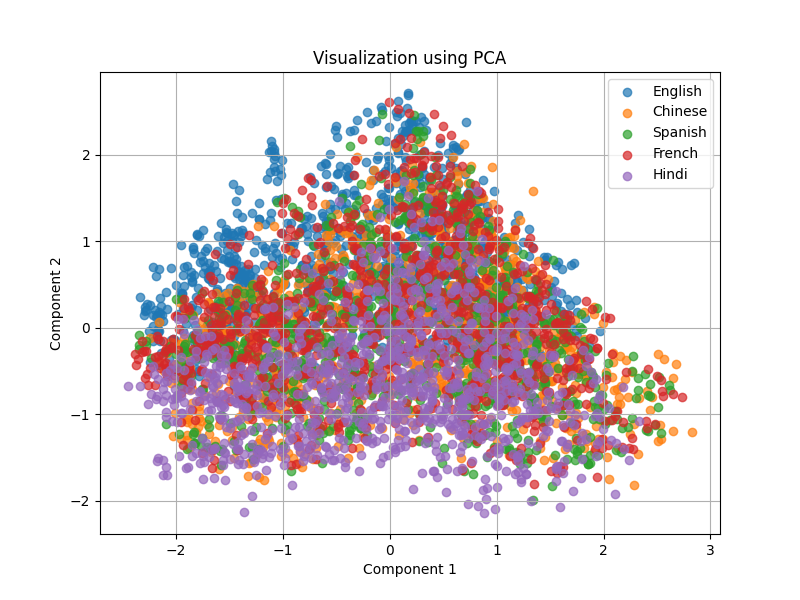}      \caption{PCA, Layer 12}        \end{subfigure}  \hfill  \begin{subfigure}{0.18\textwidth}      \includegraphics[width=\textwidth]{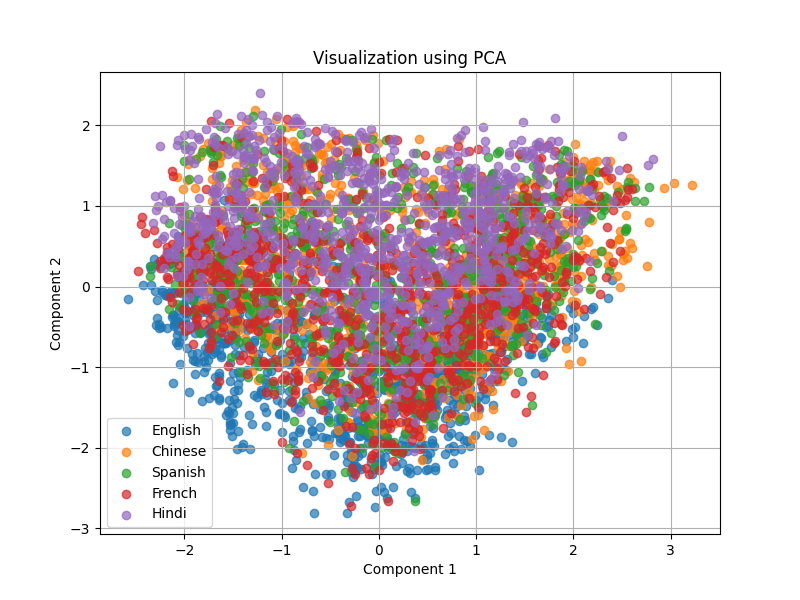}      \caption{PCA, Layer 13}        \end{subfigure}  \hfill  \begin{subfigure}{0.18\textwidth}      \includegraphics[width=\textwidth]{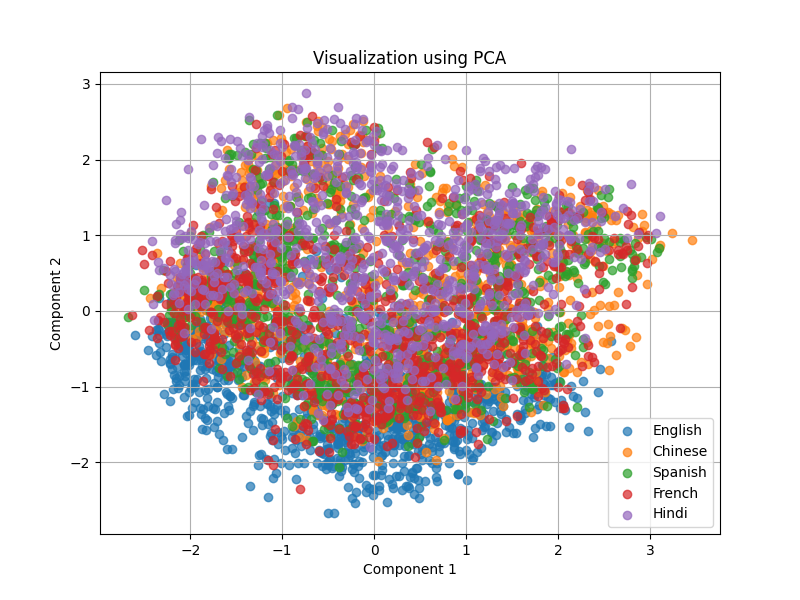}      \caption{PCA, Layer 14}        \end{subfigure}  \hfill  \begin{subfigure}{0.18\textwidth}      \includegraphics[width=\textwidth]{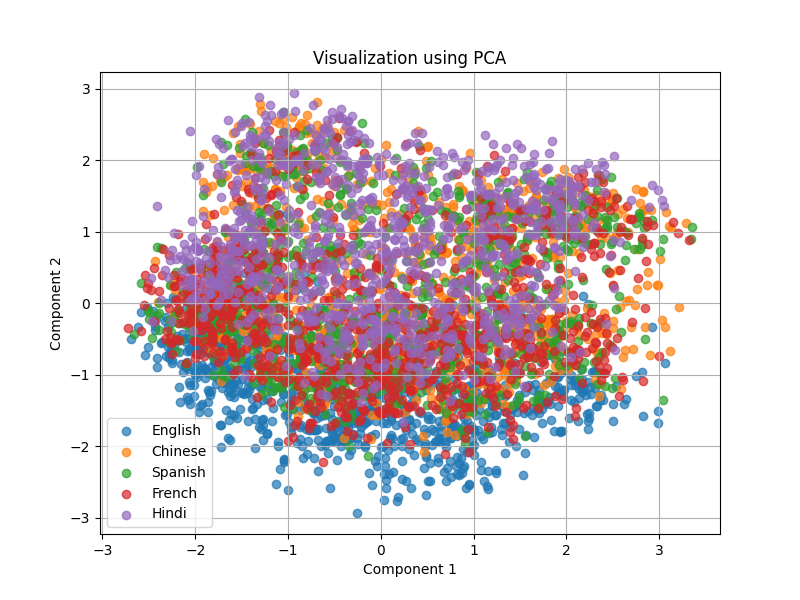}      \caption{PCA, Layer 15}        \end{subfigure}    \vspace{0.2in}    %
\begin{subfigure}{0.18\textwidth}      \includegraphics[width=\textwidth]{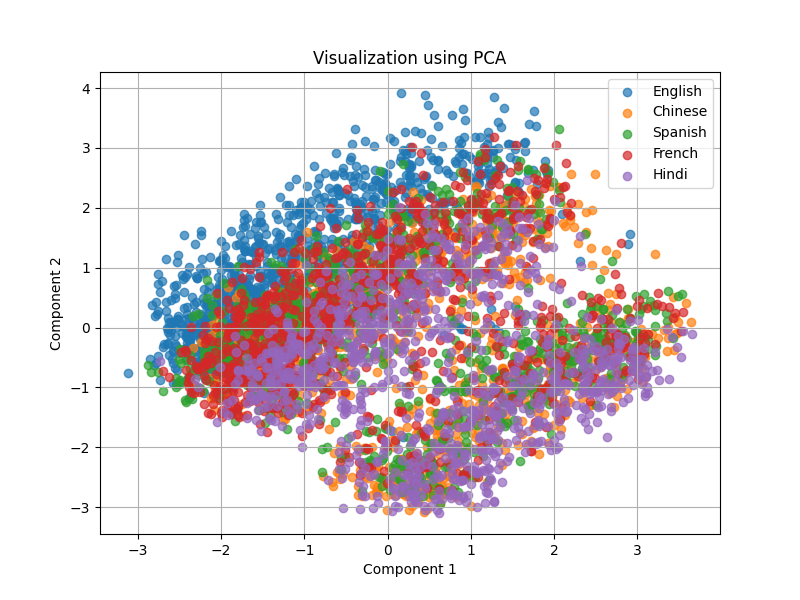}      \caption{PCA, Layer 16}        \end{subfigure}  \hfill  \begin{subfigure}{0.18\textwidth}      \includegraphics[width=\textwidth]{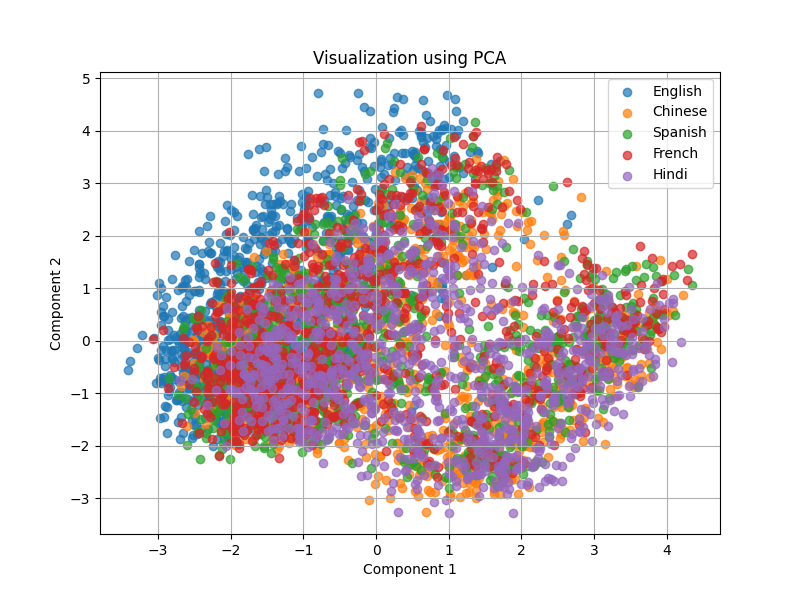}      \caption{PCA, Layer 17}        \end{subfigure}  \hfill  \begin{subfigure}{0.18\textwidth}      \includegraphics[width=\textwidth]{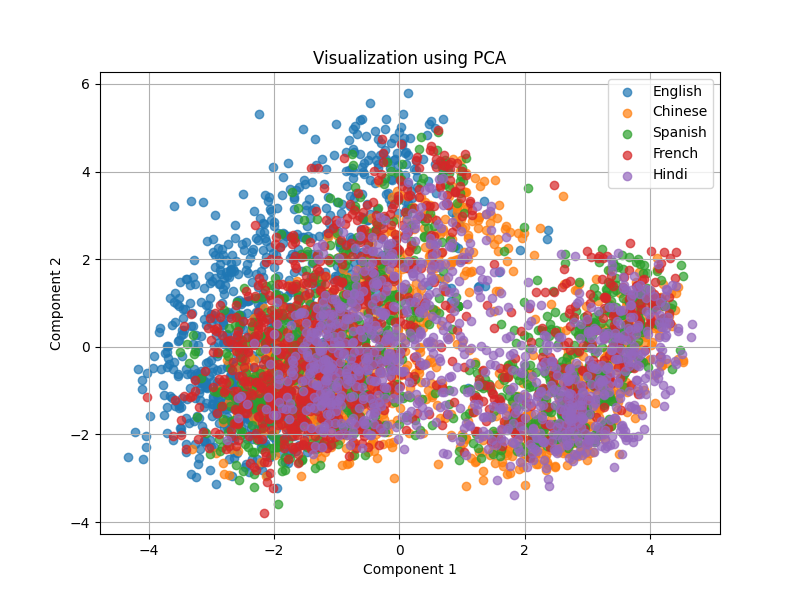}      \caption{PCA, Layer 18}        \end{subfigure}  \hfill  \begin{subfigure}{0.18\textwidth}      \includegraphics[width=\textwidth]{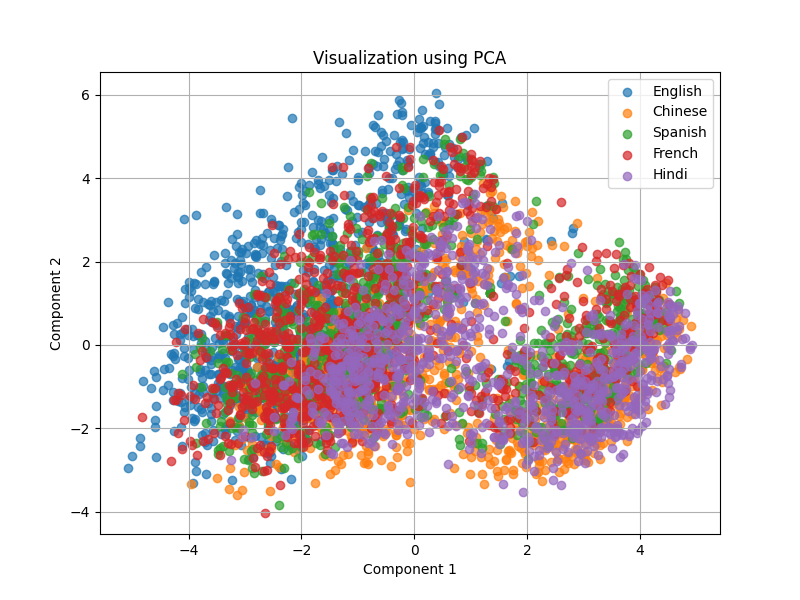}      \caption{PCA, Layer 19}        \end{subfigure}  \hfill  \begin{subfigure}{0.18\textwidth}      \includegraphics[width=\textwidth]{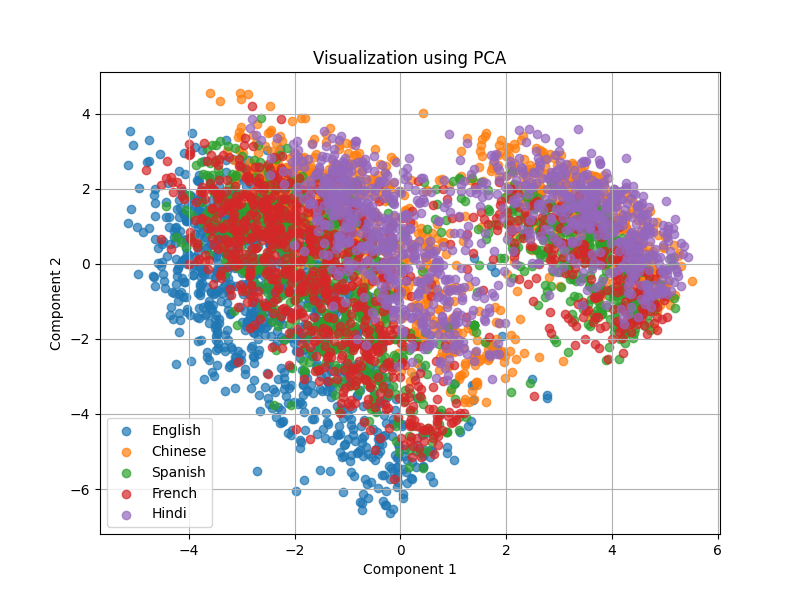}      \caption{PCA, Layer 20}        \end{subfigure}    \vspace{0.2in}    %
\begin{subfigure}{0.18\textwidth}      \includegraphics[width=\textwidth]{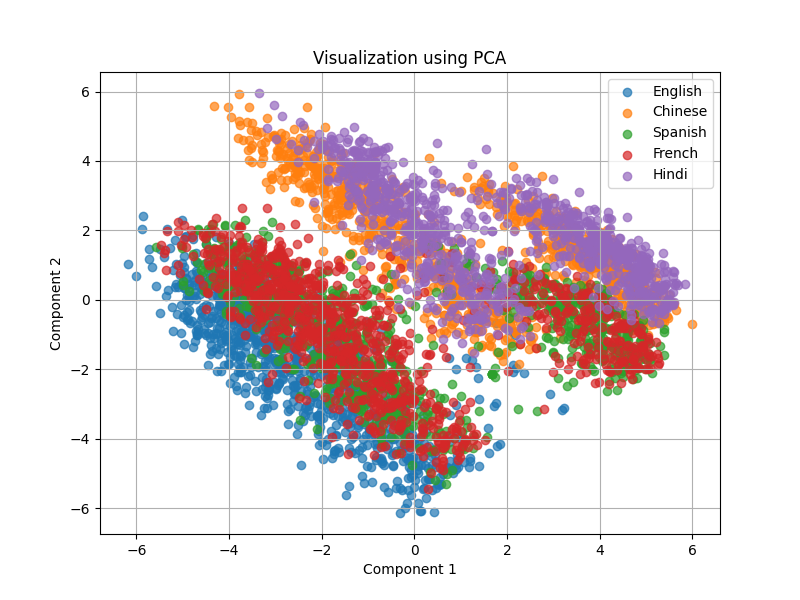}      \caption{PCA, Layer 21}        \end{subfigure}  \hfill  \begin{subfigure}{0.18\textwidth}      \includegraphics[width=\textwidth]{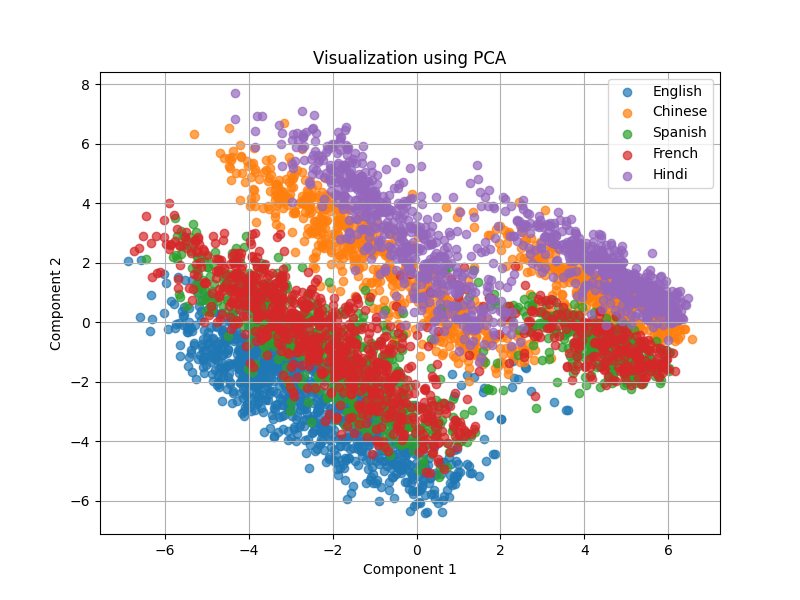}      \caption{PCA, Layer 22}        \end{subfigure}  \hfill  \begin{subfigure}{0.18\textwidth}      \includegraphics[width=\textwidth]{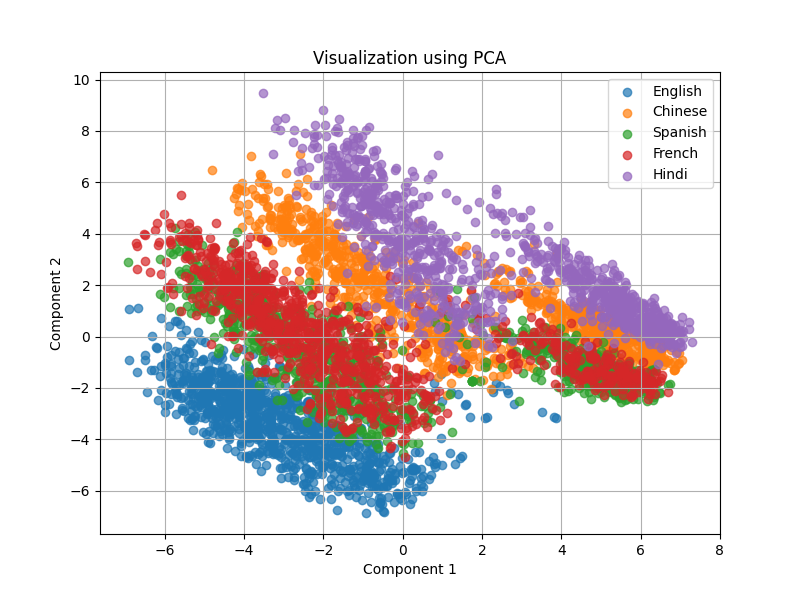}      \caption{PCA, Layer 23}        \end{subfigure}  \hfill  \begin{subfigure}{0.18\textwidth}      \includegraphics[width=\textwidth]{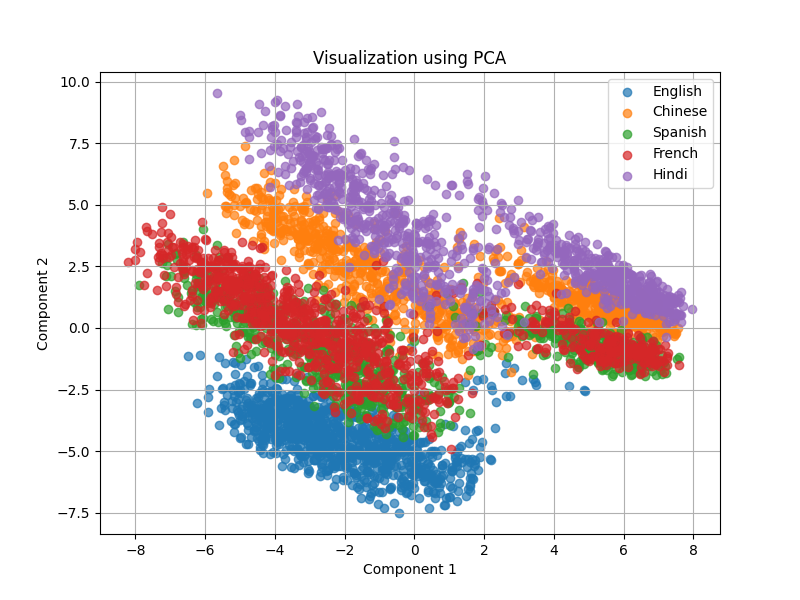}      \caption{PCA, Layer 24}        \end{subfigure}  \hfill  \begin{subfigure}{0.18\textwidth}      \includegraphics[width=\textwidth]{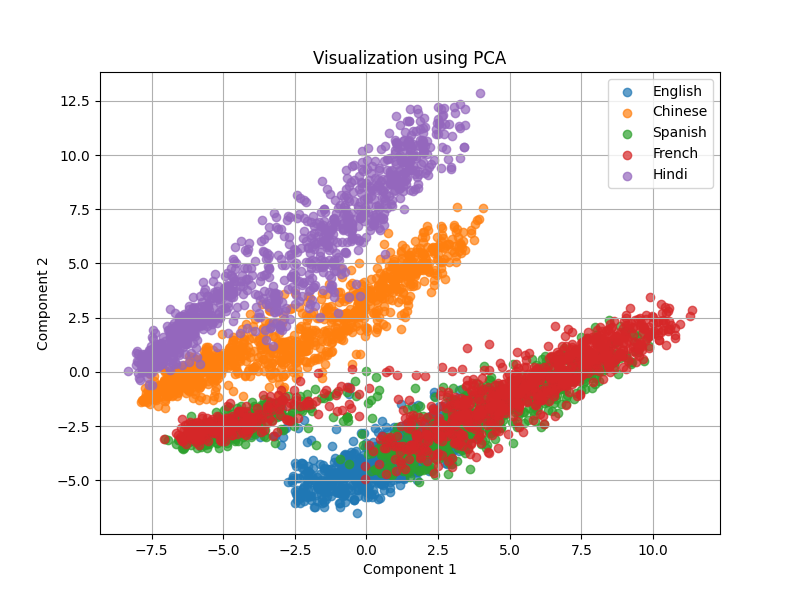}      \caption{PCA, Layer 25}        \end{subfigure}    \vspace{0.2in}    %
\begin{subfigure}{0.18\textwidth}      \includegraphics[width=\textwidth]{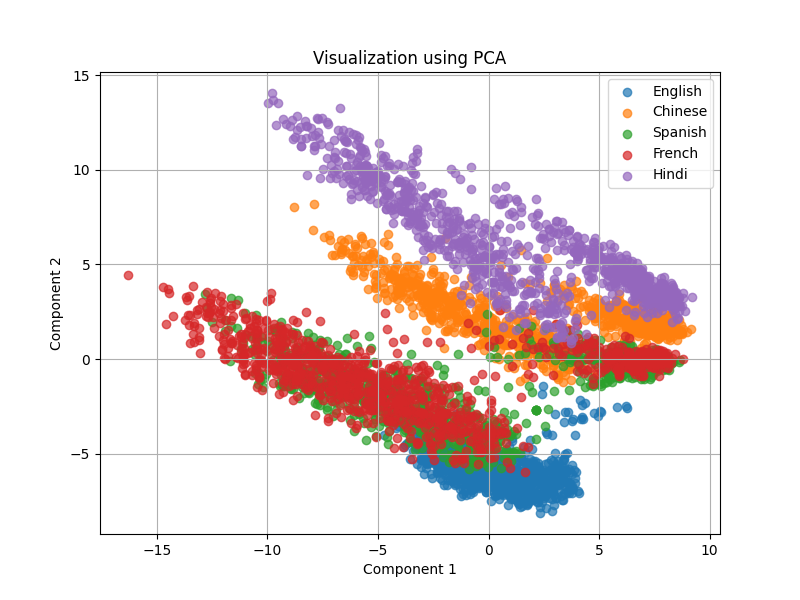}      \caption{PCA, Layer 26}        \end{subfigure}  \hfill  \begin{subfigure}{0.18\textwidth}      \includegraphics[width=\textwidth]{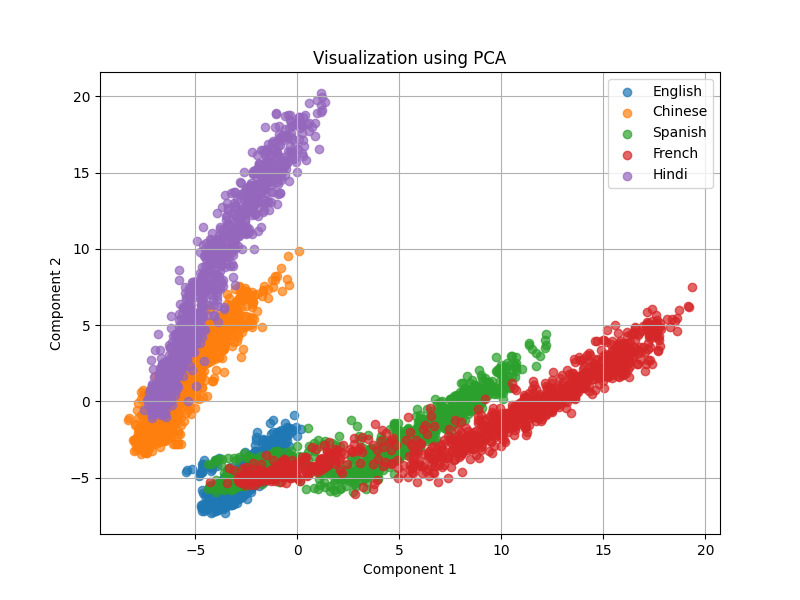}      \caption{PCA, Layer 27}        \end{subfigure}  \hfill  \begin{subfigure}{0.18\textwidth}      \includegraphics[width=\textwidth]{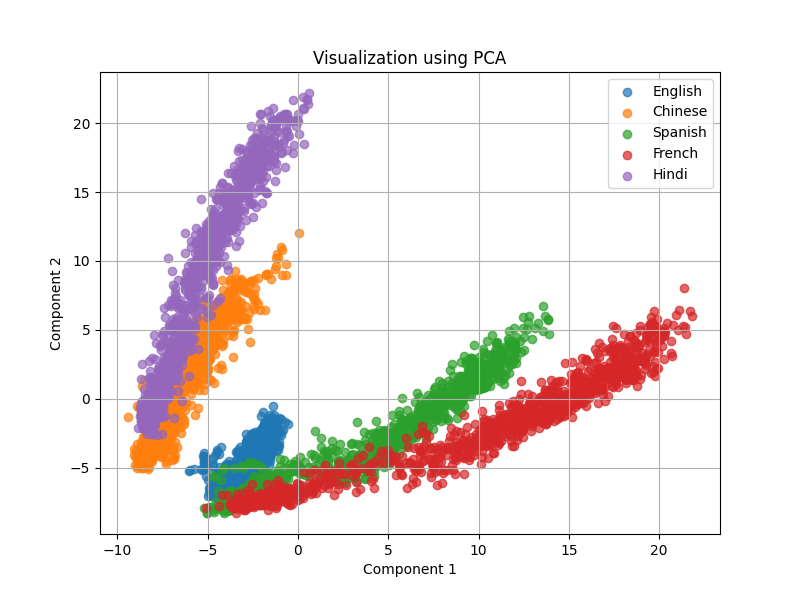}      \caption{PCA, Layer 28}        \end{subfigure}  \hfill  \begin{subfigure}{0.18\textwidth}      \includegraphics[width=\textwidth]{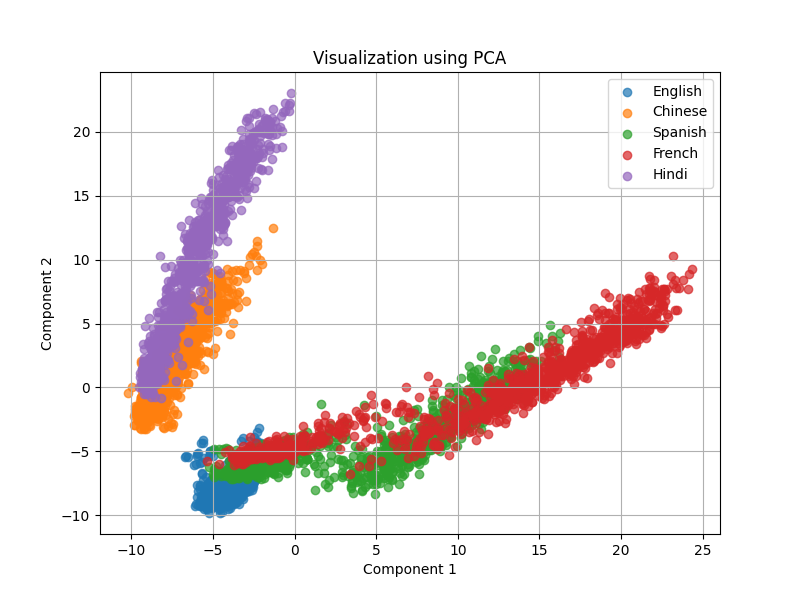}      \caption{PCA, Layer 29}        \end{subfigure}  \hfill  \begin{subfigure}{0.18\textwidth}      \includegraphics[width=\textwidth]{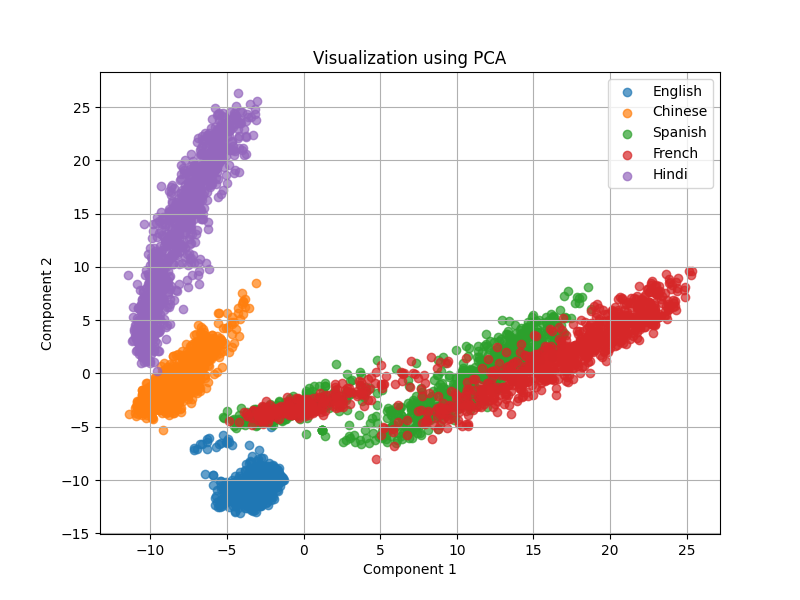}      \caption{PCA, Layer 30}        \end{subfigure}    \vspace{0.2in}    %
\begin{subfigure}{0.18\textwidth}      \includegraphics[width=\textwidth]{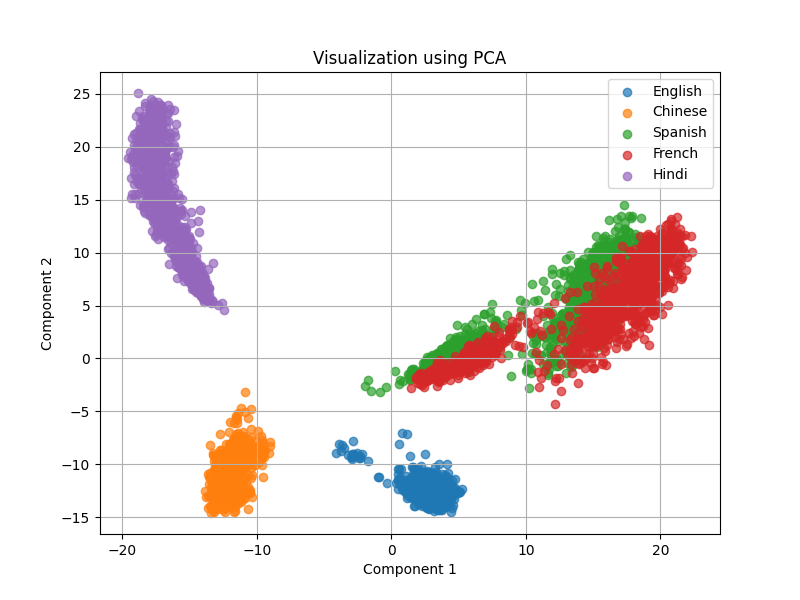}      \caption{PCA, Layer 31}        \end{subfigure}  \hfill  \begin{subfigure}{0.18\textwidth}      \includegraphics[width=\textwidth]{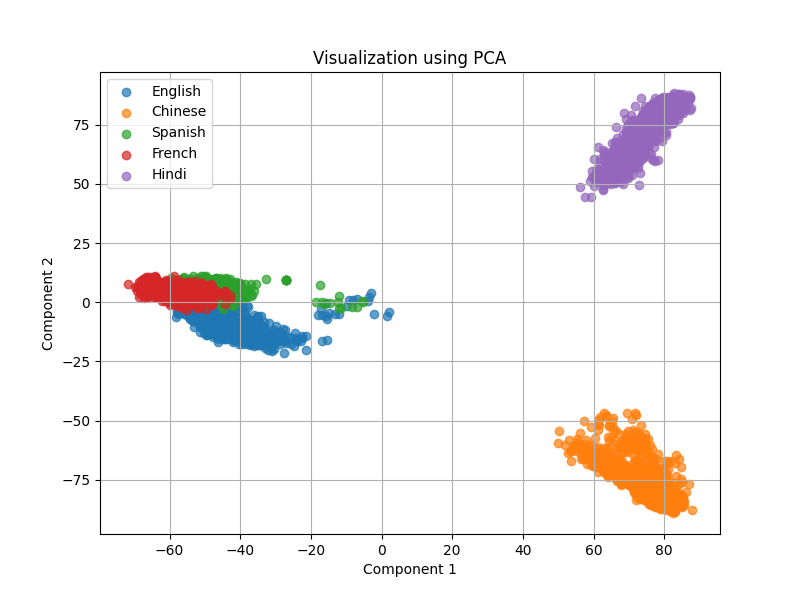}      \caption{PCA, Layer 32}        \end{subfigure}    \caption{PCA visualizations for layers 1-32 of Llama-3-8B-Instruct on the FOLIO dataset.}  
\end{figure*}

\begin{figure*}[htbp]
\centering
\begin{subfigure}{0.18\textwidth}
\includegraphics[width=\textwidth]{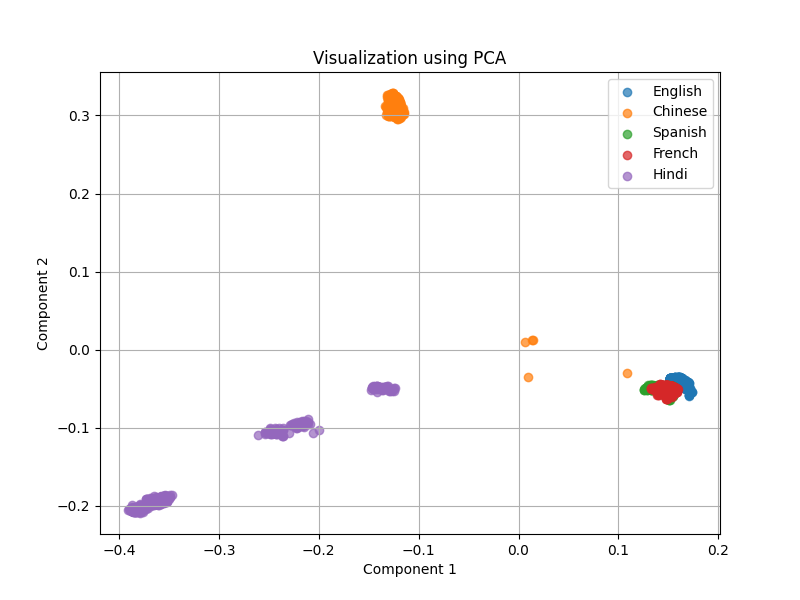}
\caption{PCA, Layer 1}
\end{subfigure}
\hfill
\begin{subfigure}{0.18\textwidth}
\includegraphics[width=\textwidth]{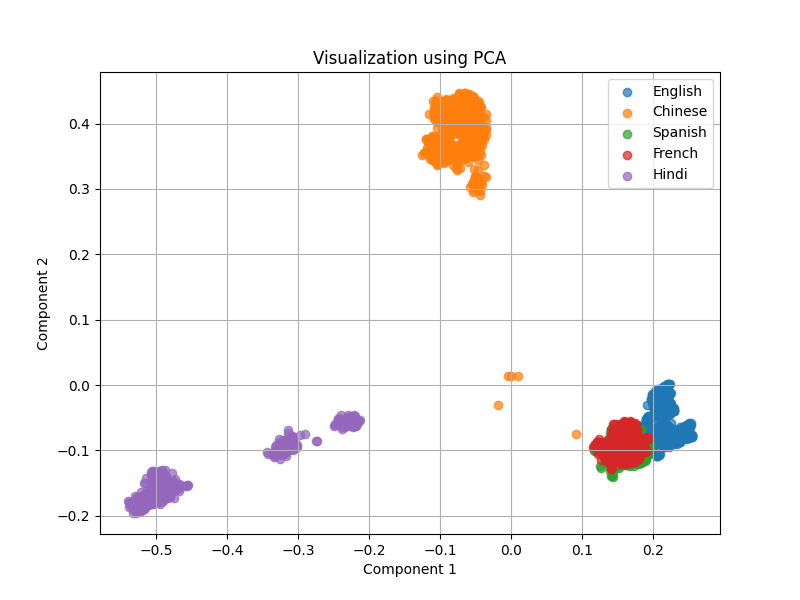}
\caption{PCA, Layer 2}

\end{subfigure}
\hfill
\begin{subfigure}{0.18\textwidth}
\includegraphics[width=\textwidth]{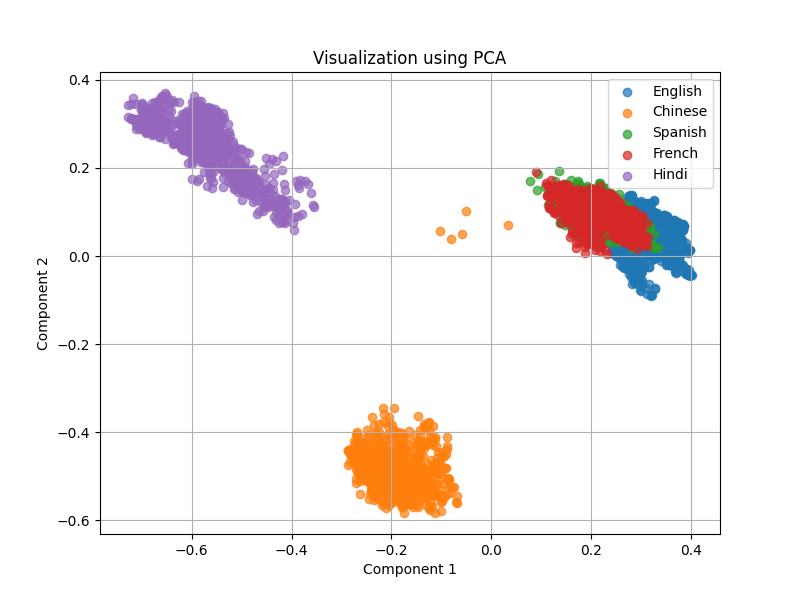}
\caption{PCA, Layer 3}

\end{subfigure}
\hfill
\begin{subfigure}{0.18\textwidth}
\includegraphics[width=\textwidth]{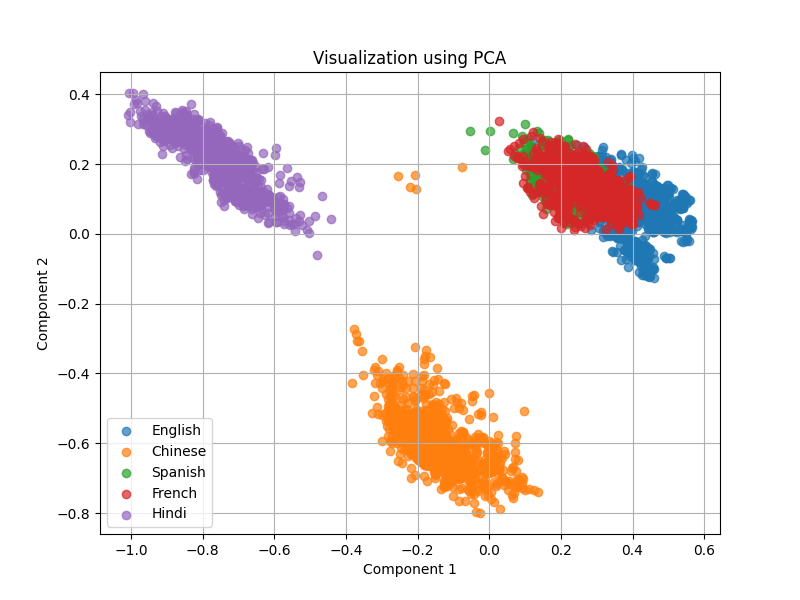}
\caption{PCA, Layer 4}

\end{subfigure}
\hfill
\begin{subfigure}{0.18\textwidth}
\includegraphics[width=\textwidth]{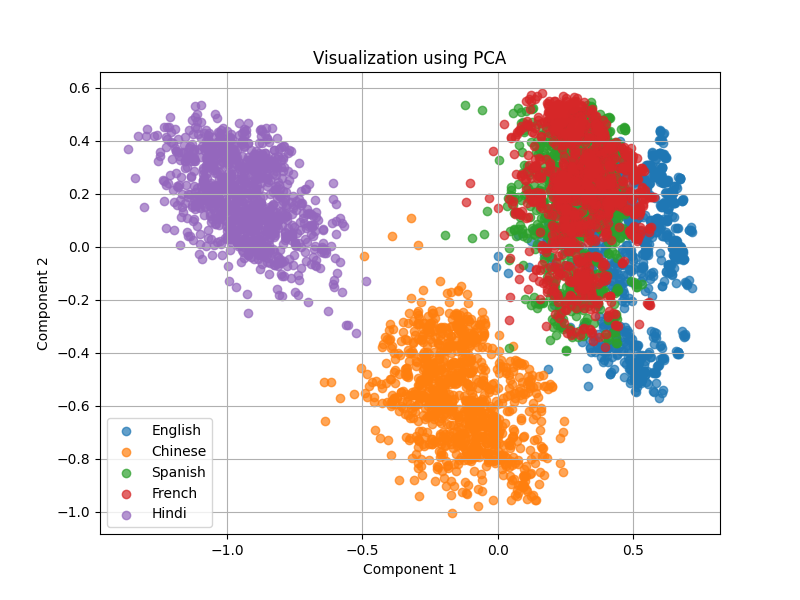}
\caption{PCA, Layer 5}

\end{subfigure}
\vspace{0.2in} %
\begin{subfigure}{0.18\textwidth}      \includegraphics[width=\textwidth]{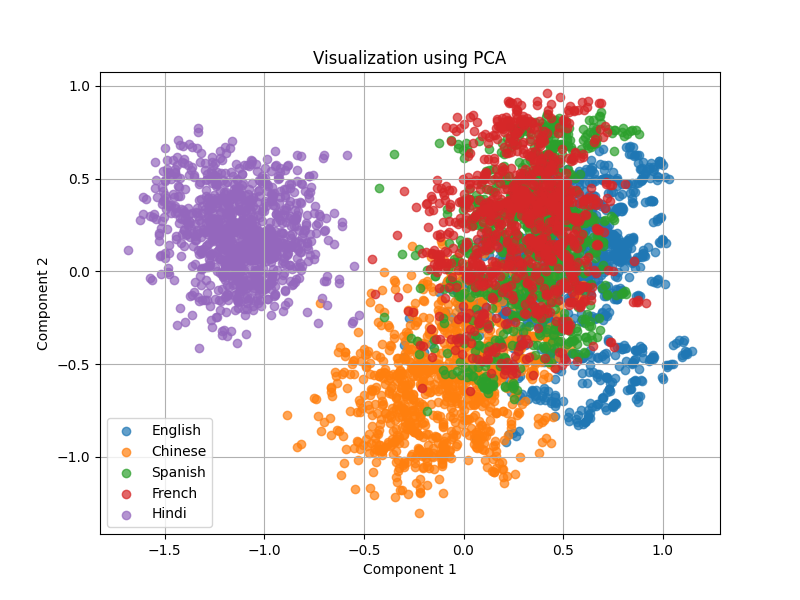}      \caption{PCA, Layer 6}        \end{subfigure}  \hfill  \begin{subfigure}{0.18\textwidth}      \includegraphics[width=\textwidth]{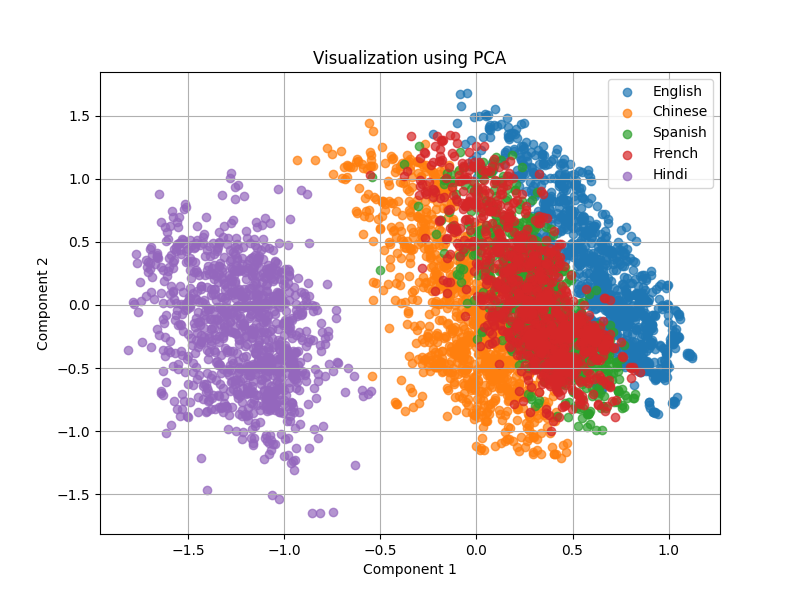}      \caption{PCA, Layer 7}        \end{subfigure}  \hfill  \begin{subfigure}{0.18\textwidth}      \includegraphics[width=\textwidth]{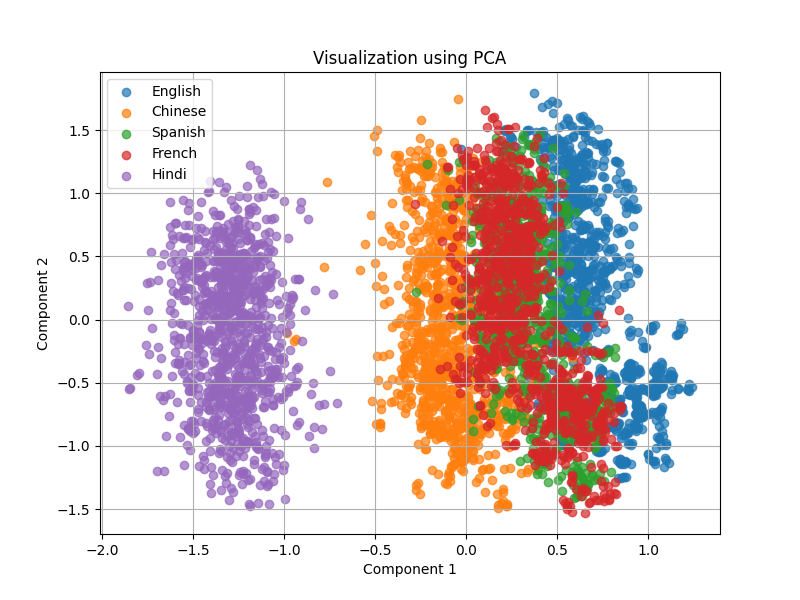}      \caption{PCA, Layer 8}        \end{subfigure}  \hfill  \begin{subfigure}{0.18\textwidth}      \includegraphics[width=\textwidth]{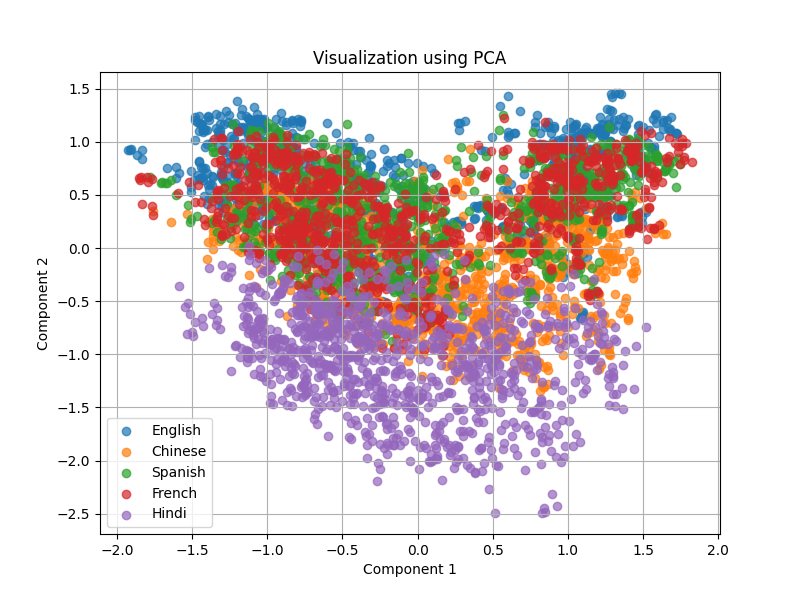}      \caption{PCA, Layer 9}        \end{subfigure}  \hfill  \begin{subfigure}{0.18\textwidth}      \includegraphics[width=\textwidth]{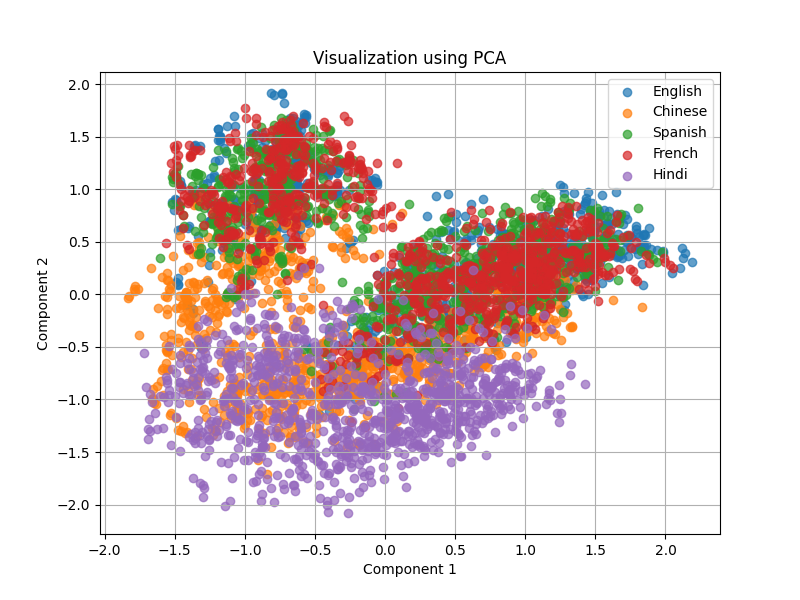}      \caption{PCA, Layer 10}        \end{subfigure}    \vspace{0.2in}    %
\begin{subfigure}{0.18\textwidth}      \includegraphics[width=\textwidth]{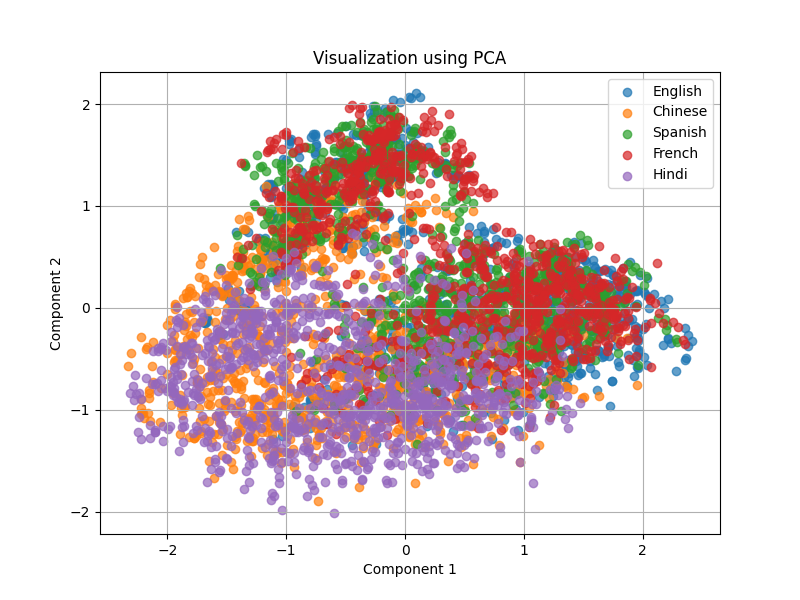}      \caption{PCA, Layer 11}        \end{subfigure}  \hfill  \begin{subfigure}{0.18\textwidth}      \includegraphics[width=\textwidth]{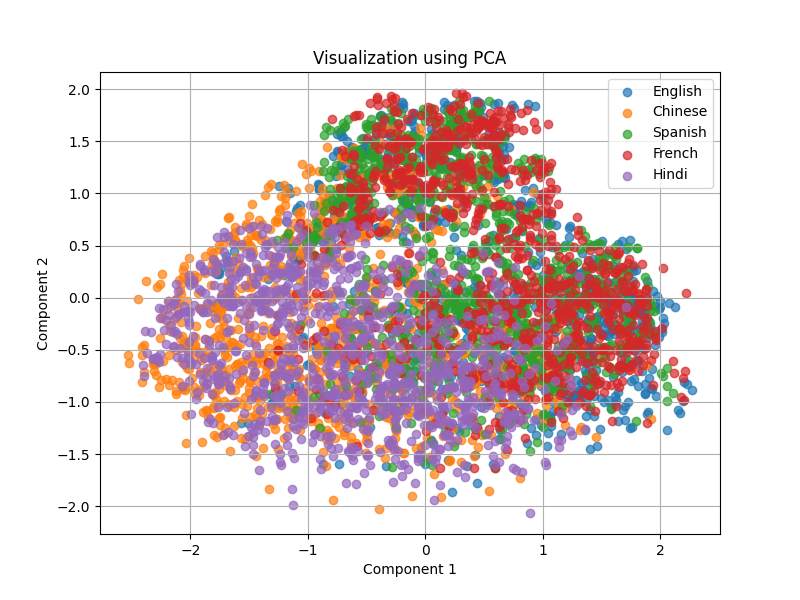}      \caption{PCA, Layer 12}        \end{subfigure}  \hfill  \begin{subfigure}{0.18\textwidth}      \includegraphics[width=\textwidth]{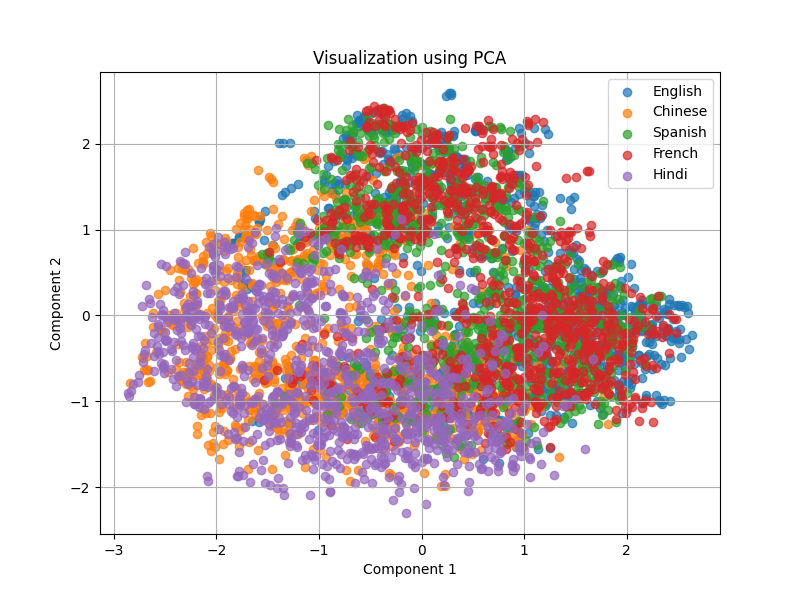}      \caption{PCA, Layer 13}        \end{subfigure}  \hfill  \begin{subfigure}{0.18\textwidth}      \includegraphics[width=\textwidth]{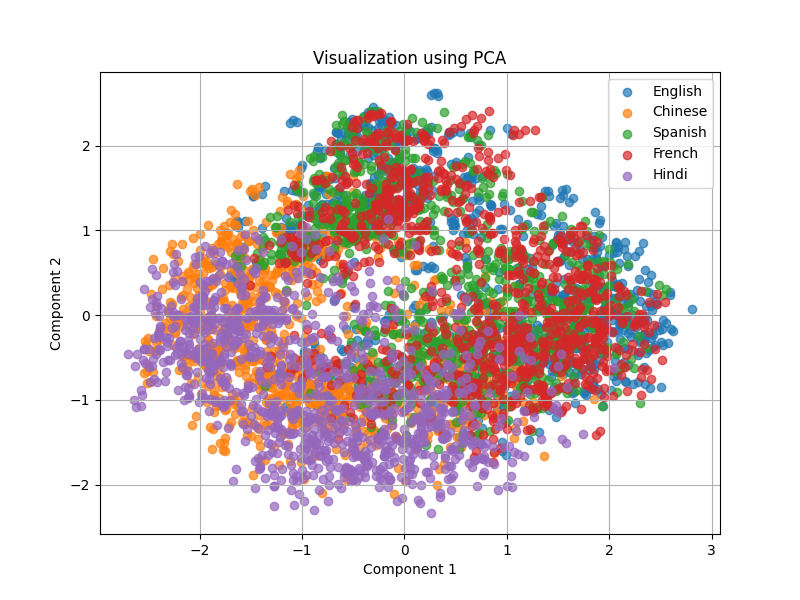}      \caption{PCA, Layer 14}        \end{subfigure}  \hfill  \begin{subfigure}{0.18\textwidth}      \includegraphics[width=\textwidth]{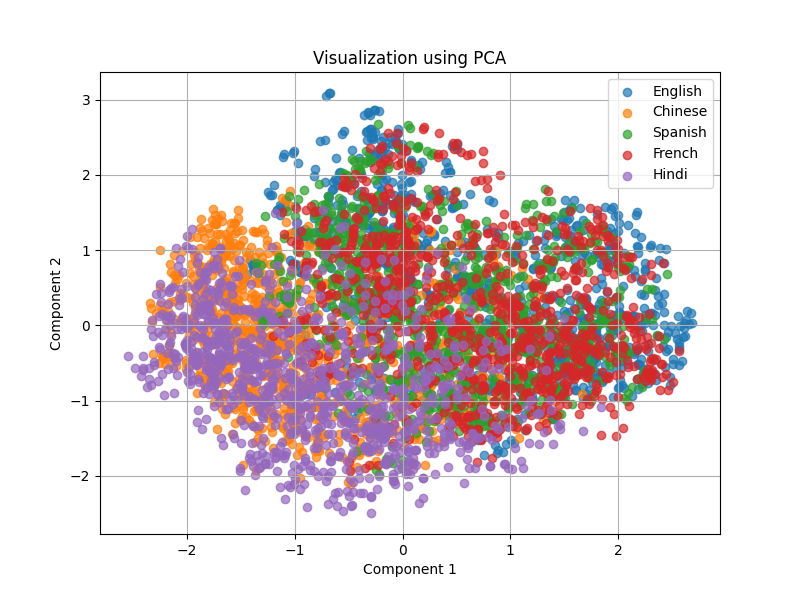}      \caption{PCA, Layer 15}        \end{subfigure}    \vspace{0.2in}    %
\begin{subfigure}{0.18\textwidth}      \includegraphics[width=\textwidth]{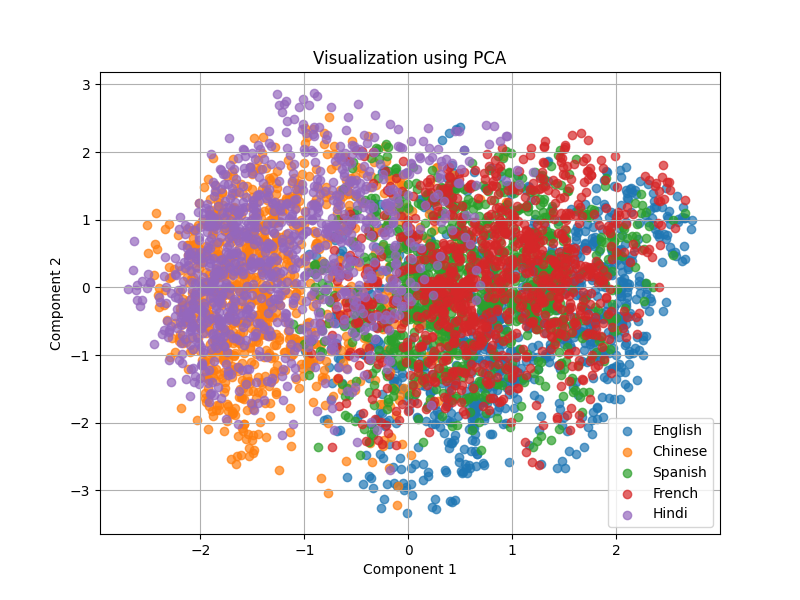}      \caption{PCA, Layer 16}        \end{subfigure}  \hfill  \begin{subfigure}{0.18\textwidth}      \includegraphics[width=\textwidth]{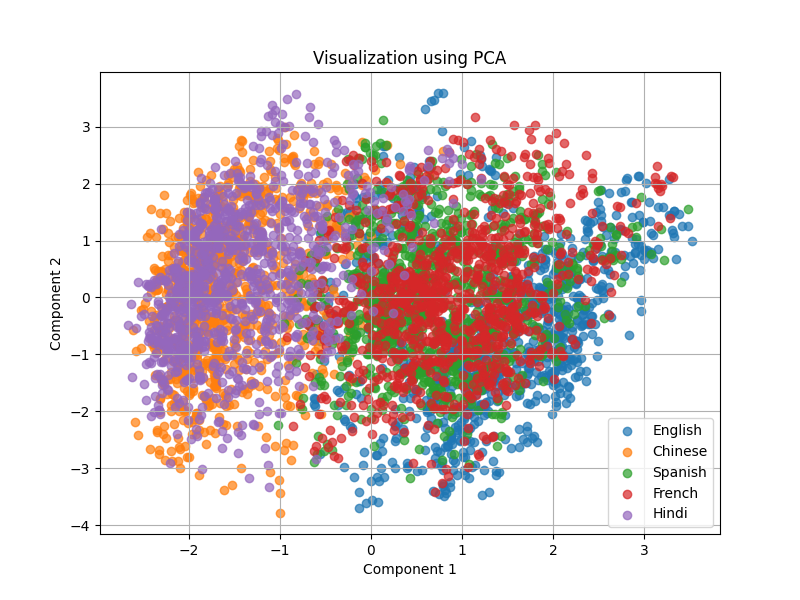}      \caption{PCA, Layer 17}        \end{subfigure}  \hfill  \begin{subfigure}{0.18\textwidth}      \includegraphics[width=\textwidth]{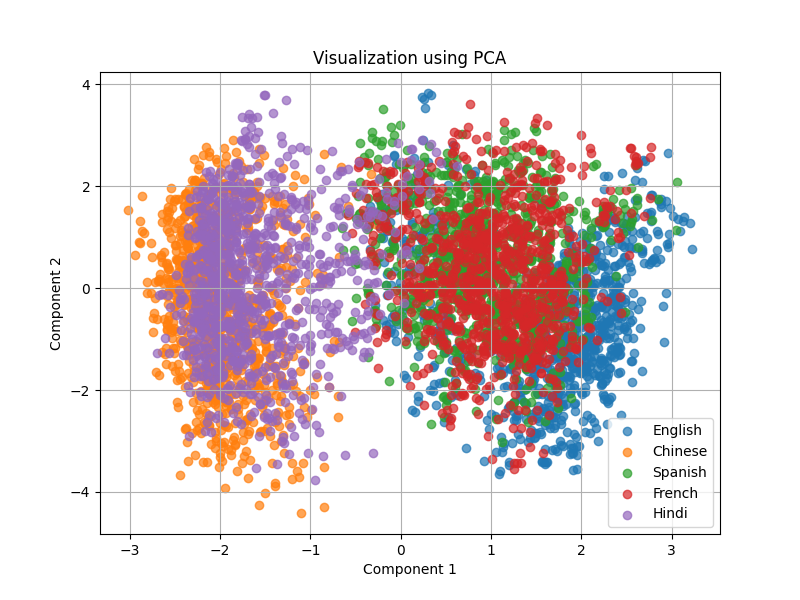}      \caption{PCA, Layer 18}        \end{subfigure}  \hfill  \begin{subfigure}{0.18\textwidth}      \includegraphics[width=\textwidth]{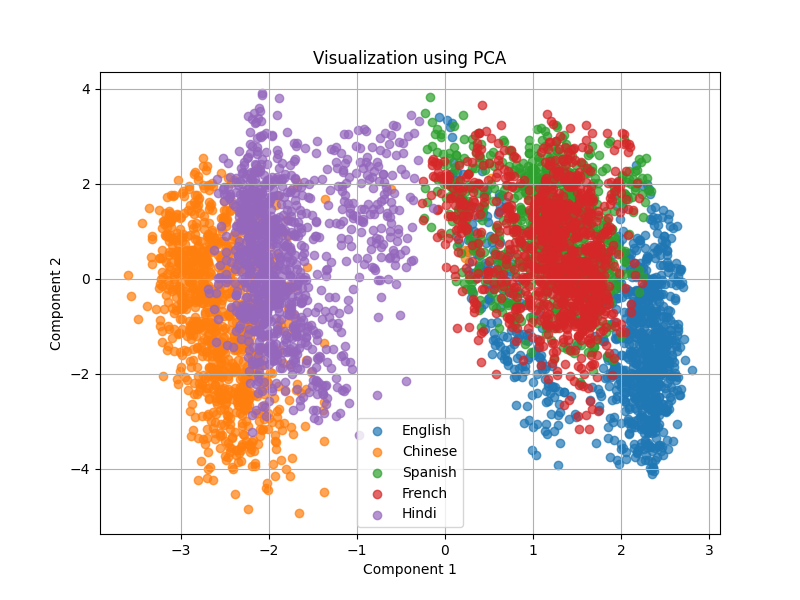}      \caption{PCA, Layer 19}        \end{subfigure}  \hfill  \begin{subfigure}{0.18\textwidth}      \includegraphics[width=\textwidth]{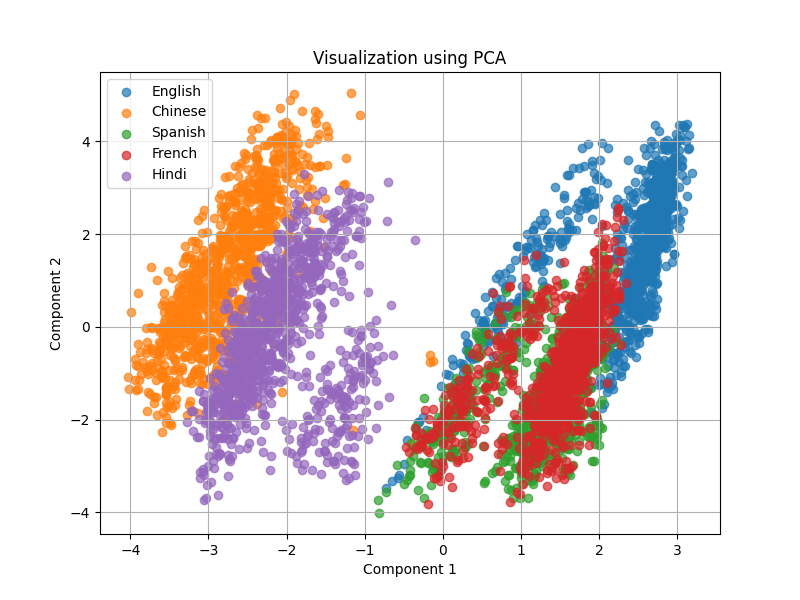}      \caption{PCA, Layer 20}        \end{subfigure}    \vspace{0.2in}    %
\begin{subfigure}{0.18\textwidth}      \includegraphics[width=\textwidth]{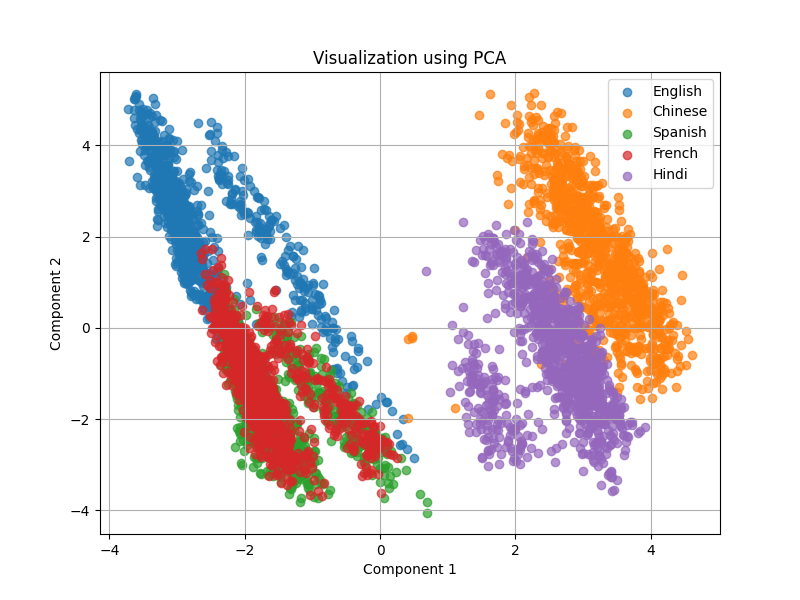}      \caption{PCA, Layer 21}        \end{subfigure}  \hfill  \begin{subfigure}{0.18\textwidth}      \includegraphics[width=\textwidth]{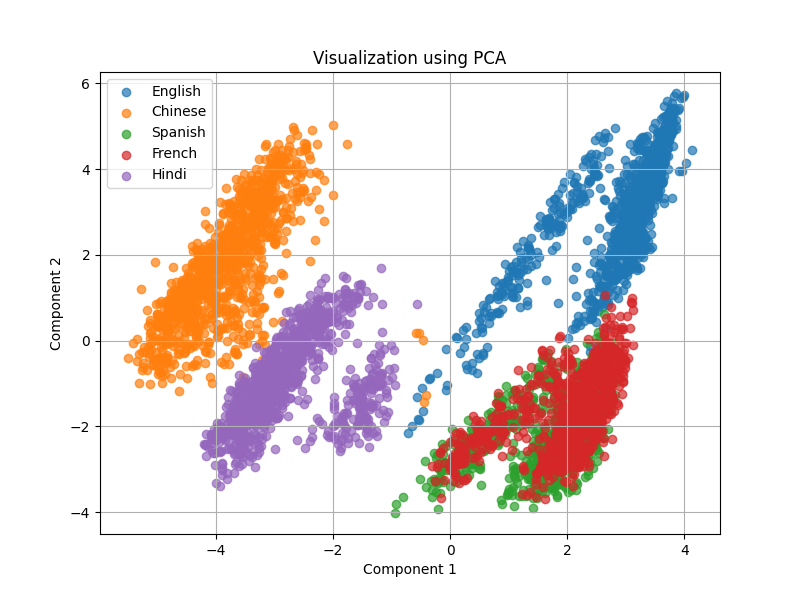}      \caption{PCA, Layer 22}        \end{subfigure}  \hfill  \begin{subfigure}{0.18\textwidth}      \includegraphics[width=\textwidth]{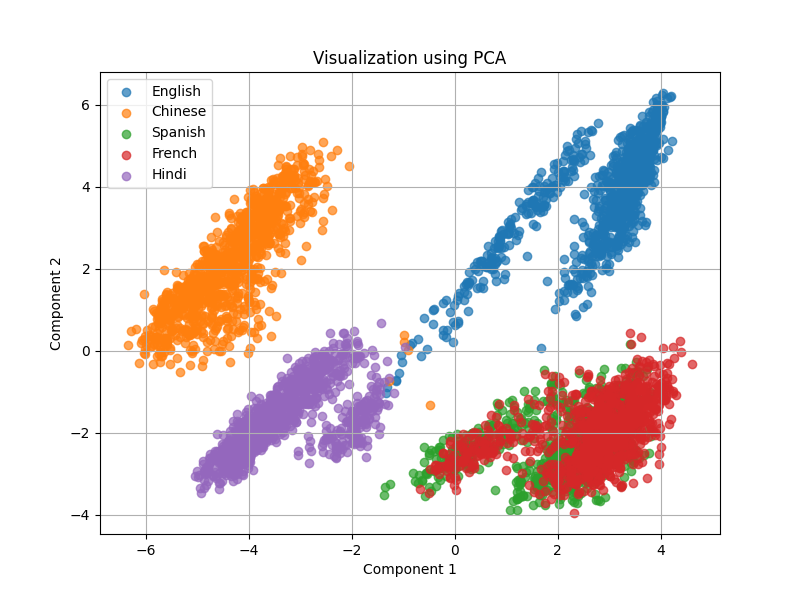}      \caption{PCA, Layer 23}        \end{subfigure}  \hfill  \begin{subfigure}{0.18\textwidth}      \includegraphics[width=\textwidth]{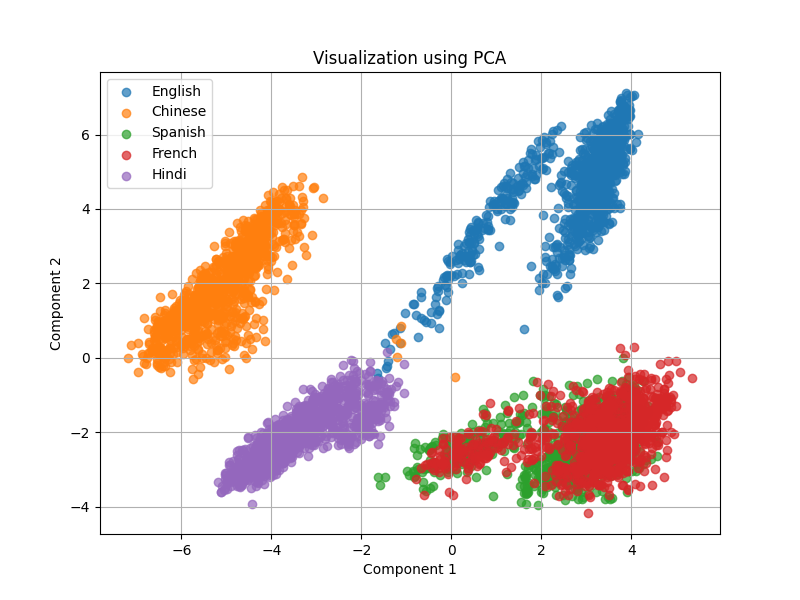}      \caption{PCA, Layer 24}        \end{subfigure}  \hfill  \begin{subfigure}{0.18\textwidth}      \includegraphics[width=\textwidth]{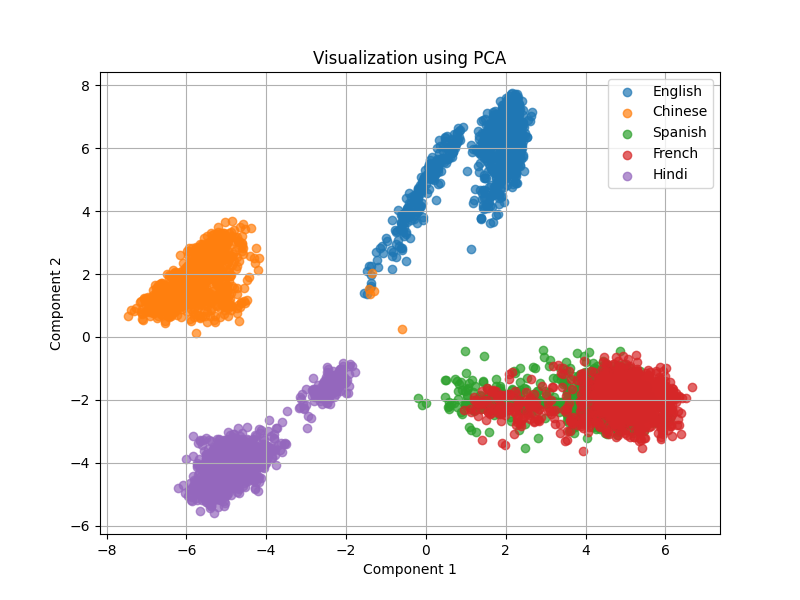}      \caption{PCA, Layer 25}        \end{subfigure}    \vspace{0.2in}    %
\begin{subfigure}{0.18\textwidth}      \includegraphics[width=\textwidth]{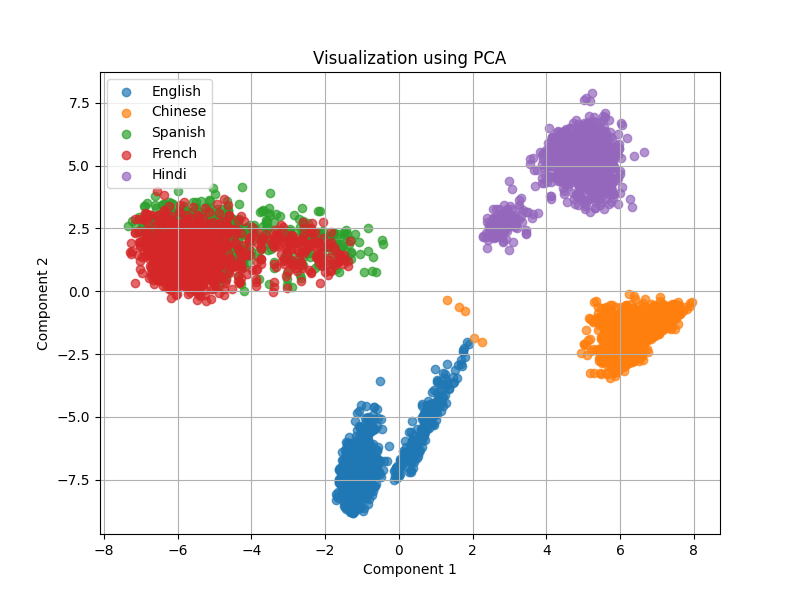}      \caption{PCA, Layer 26}        \end{subfigure}  \hfill  \begin{subfigure}{0.18\textwidth}      \includegraphics[width=\textwidth]{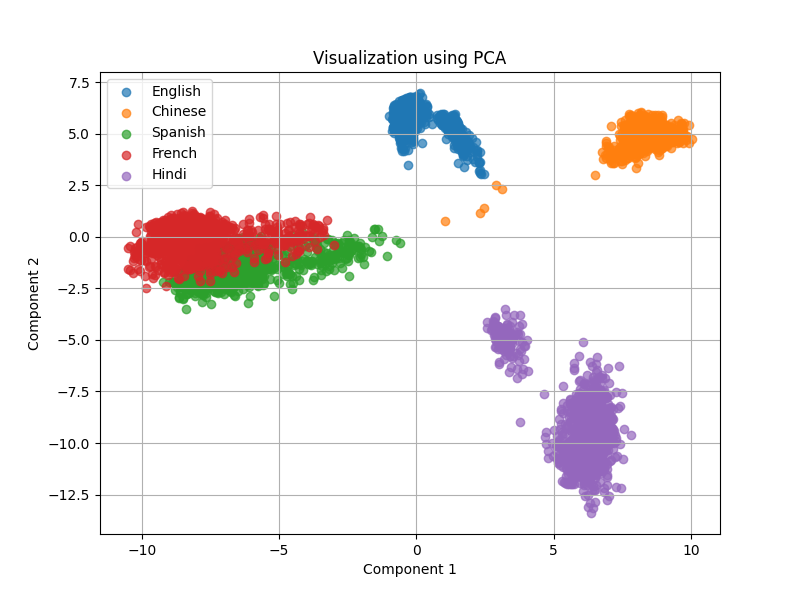}      \caption{PCA, Layer 27}        \end{subfigure}  \hfill  \begin{subfigure}{0.18\textwidth}      \includegraphics[width=\textwidth]{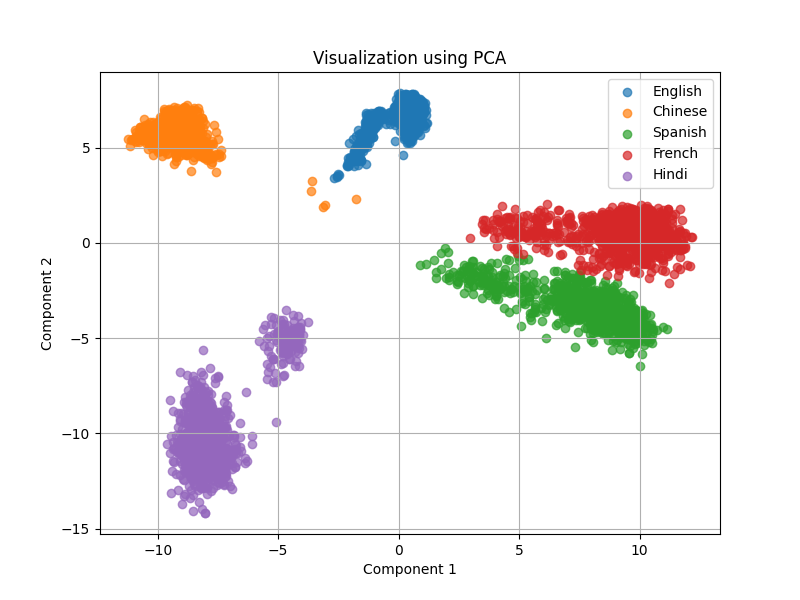}      \caption{PCA, Layer 28}        \end{subfigure}  \hfill  \begin{subfigure}{0.18\textwidth}      \includegraphics[width=\textwidth]{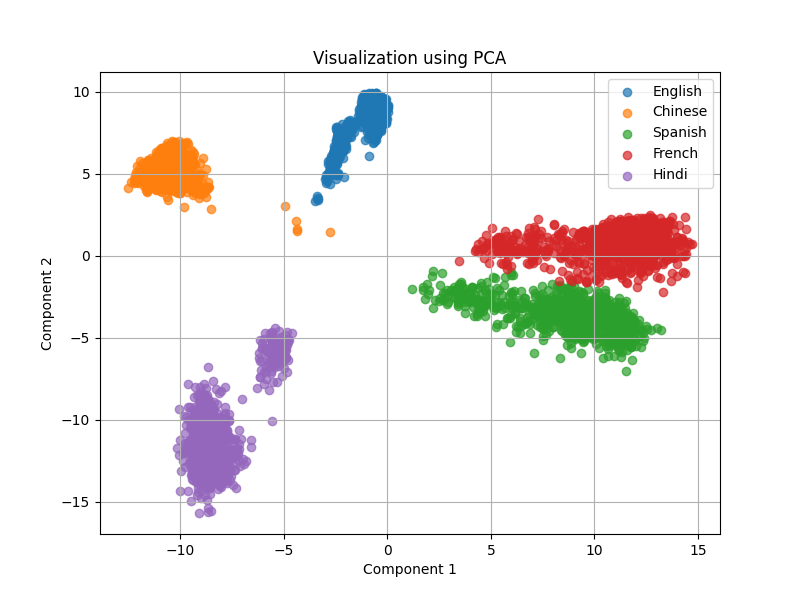}      \caption{PCA, Layer 29}        \end{subfigure}  \hfill  \begin{subfigure}{0.18\textwidth}      \includegraphics[width=\textwidth]{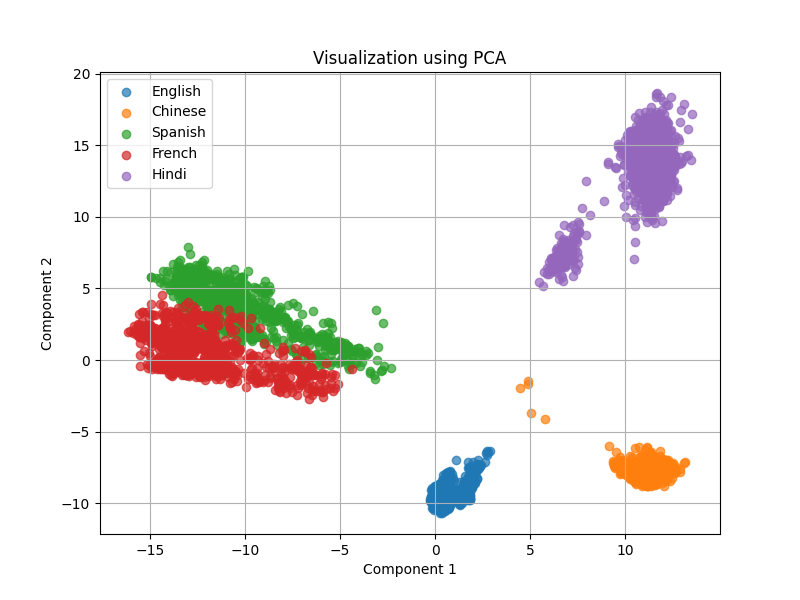}      \caption{PCA, Layer 30}        \end{subfigure}    \vspace{0.2in}    %
\begin{subfigure}{0.18\textwidth}      \includegraphics[width=\textwidth]{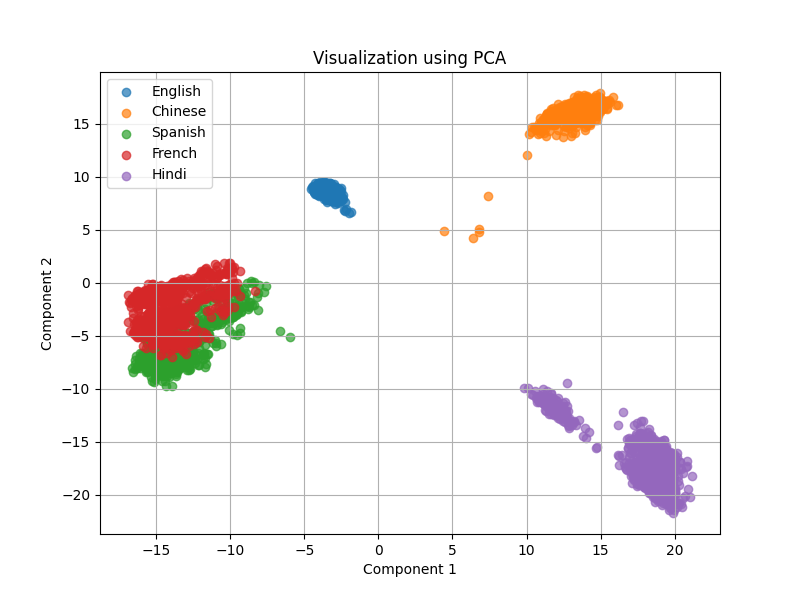}      \caption{PCA, Layer 31}        \end{subfigure}  \hfill  \begin{subfigure}{0.18\textwidth}      \includegraphics[width=\textwidth]{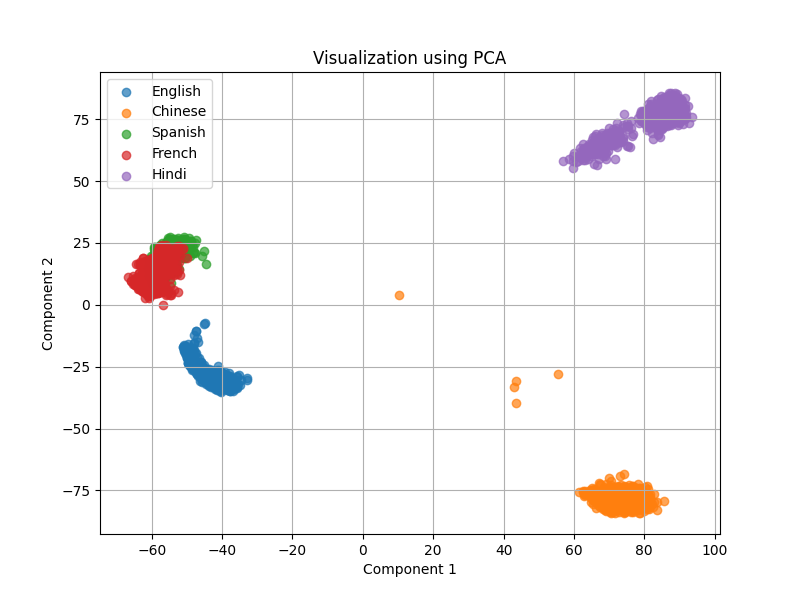}      \caption{PCA, Layer 32}        \end{subfigure}    \caption{PCA visualizations for layers 1-32 of Llama-3-8B-Instruct on the LogicalDeduction dataset.}  
\end{figure*}

\begin{figure*}[htbp]
\centering
\begin{subfigure}{0.18\textwidth}
\includegraphics[width=\textwidth]{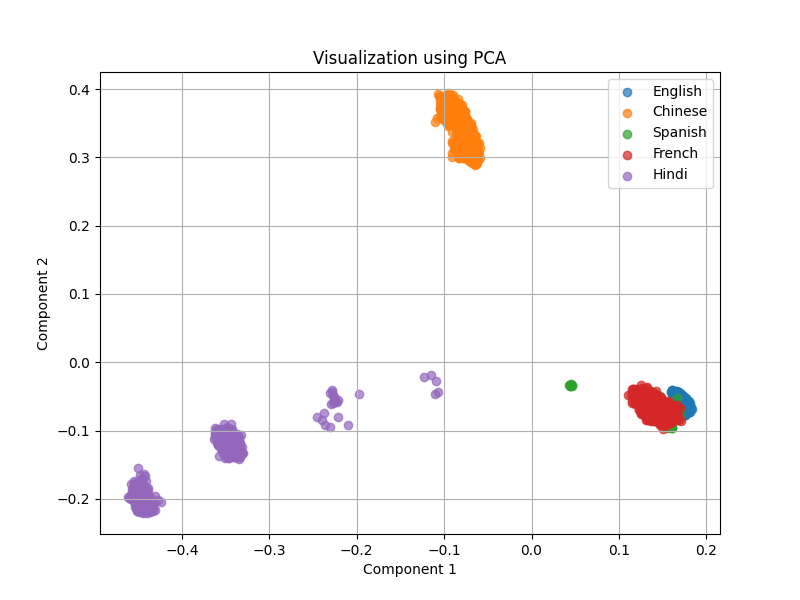}
\caption{PCA, Layer 1}
\end{subfigure}
\hfill
\begin{subfigure}{0.18\textwidth}
\includegraphics[width=\textwidth]{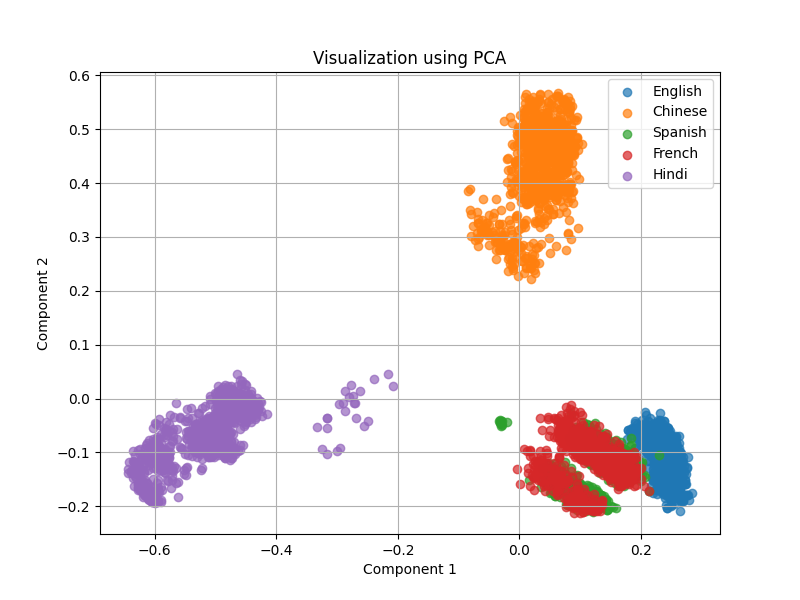}
\caption{PCA, Layer 2}

\end{subfigure}
\hfill
\begin{subfigure}{0.18\textwidth}
\includegraphics[width=\textwidth]{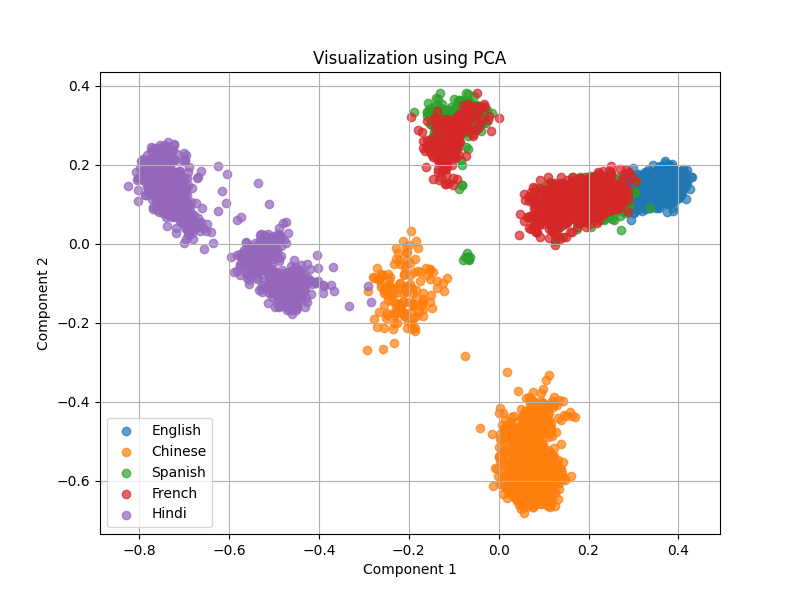}
\caption{PCA, Layer 3}

\end{subfigure}
\hfill
\begin{subfigure}{0.18\textwidth}
\includegraphics[width=\textwidth]{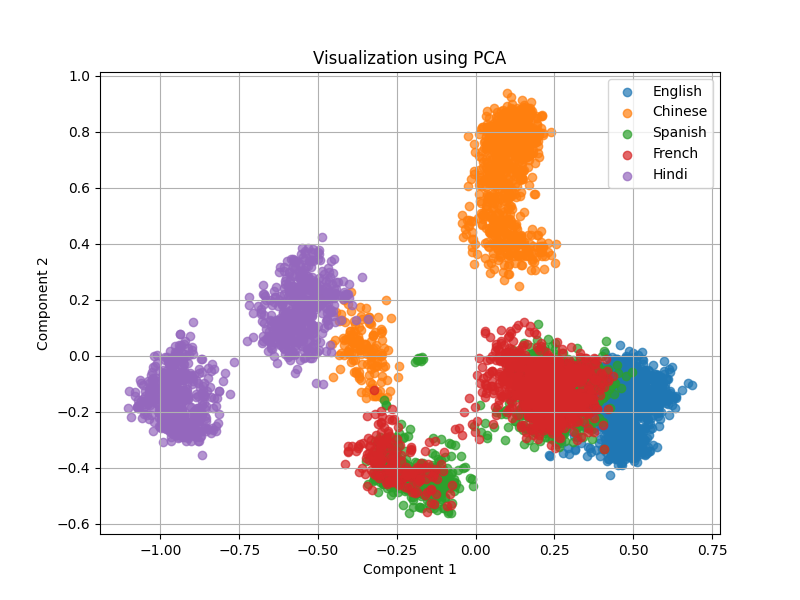}
\caption{PCA, Layer 4}

\end{subfigure}
\hfill
\begin{subfigure}{0.18\textwidth}
\includegraphics[width=\textwidth]{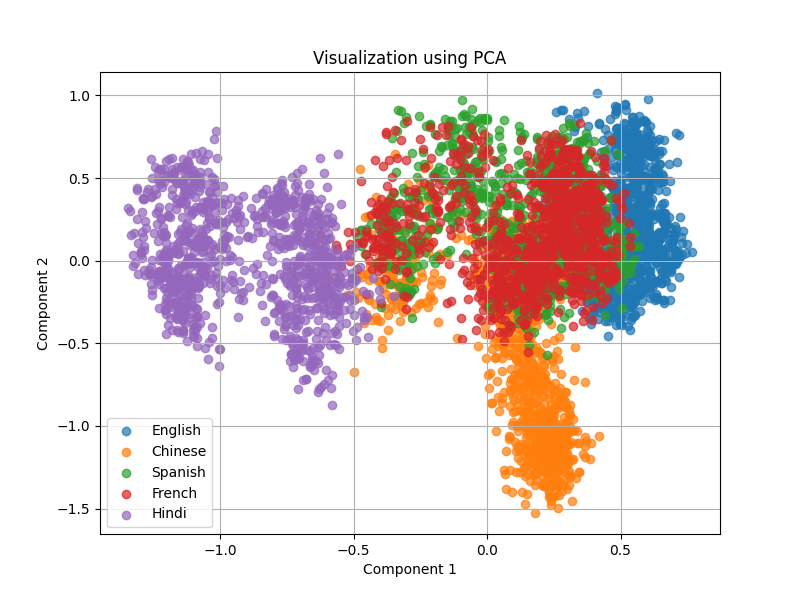}
\caption{PCA, Layer 5}

\end{subfigure}
\vspace{0.2in} %
\begin{subfigure}{0.18\textwidth}      \includegraphics[width=\textwidth]{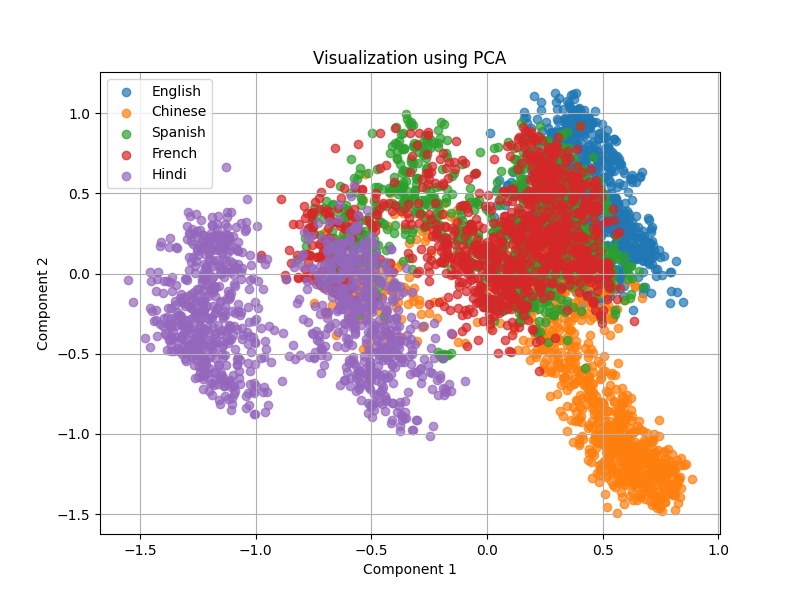}      \caption{PCA, Layer 6}        \end{subfigure}  \hfill  \begin{subfigure}{0.18\textwidth}      \includegraphics[width=\textwidth]{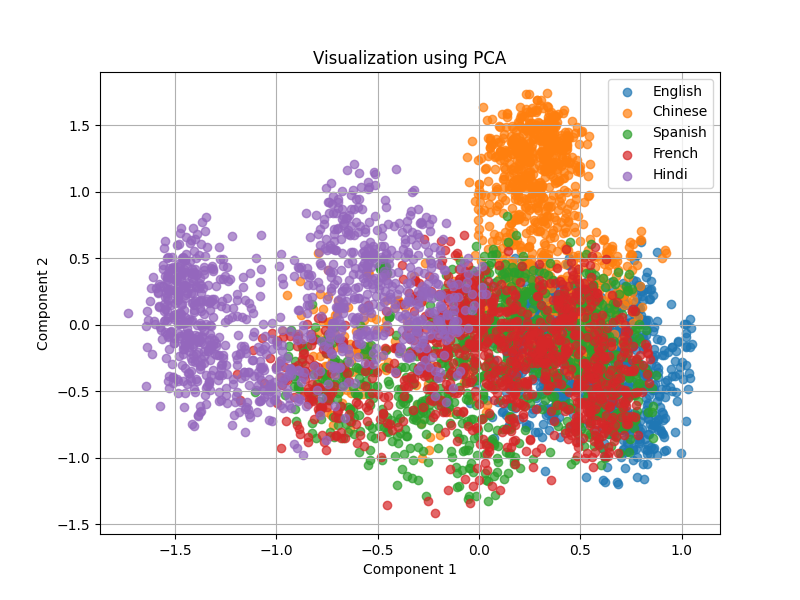}      \caption{PCA, Layer 7}        \end{subfigure}  \hfill  \begin{subfigure}{0.18\textwidth}      \includegraphics[width=\textwidth]{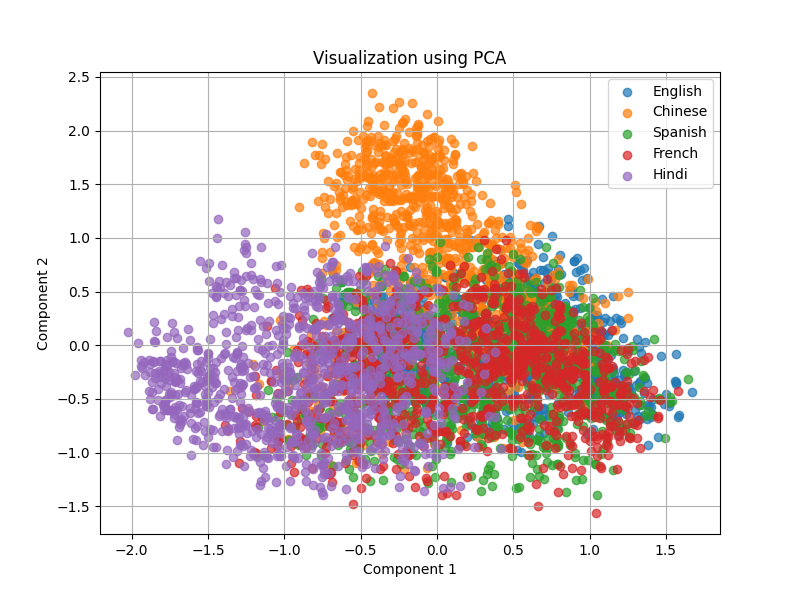}      \caption{PCA, Layer 8}        \end{subfigure}  \hfill  \begin{subfigure}{0.18\textwidth}      \includegraphics[width=\textwidth]{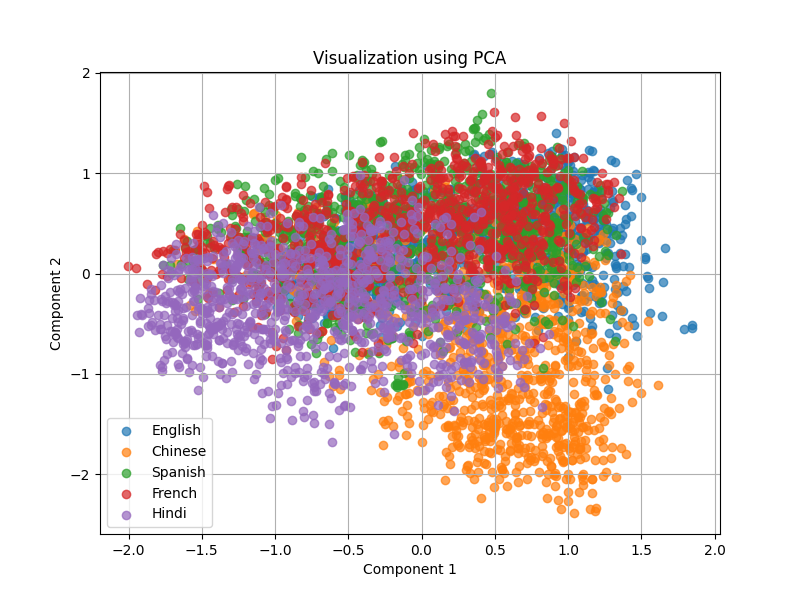}      \caption{PCA, Layer 9}        \end{subfigure}  \hfill  \begin{subfigure}{0.18\textwidth}      \includegraphics[width=\textwidth]{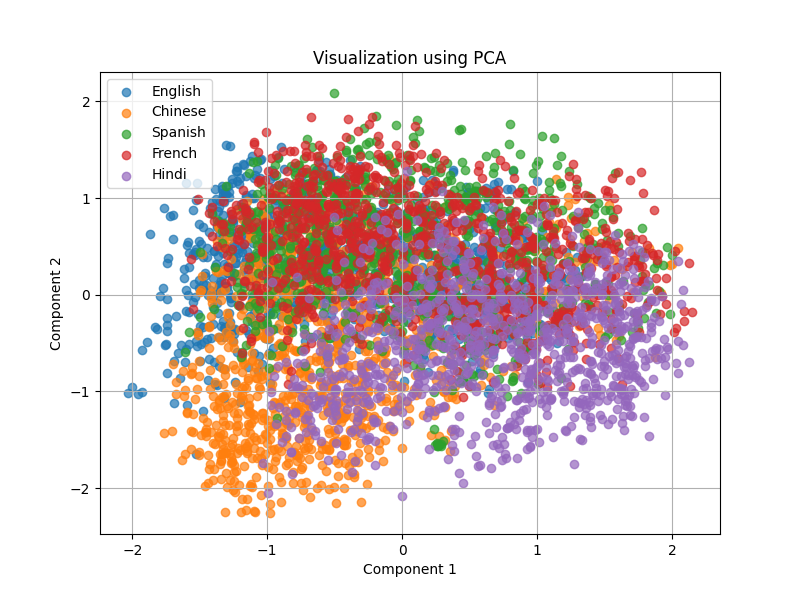}      \caption{PCA, Layer 10}        \end{subfigure}    \vspace{0.2in}    %
\begin{subfigure}{0.18\textwidth}      \includegraphics[width=\textwidth]{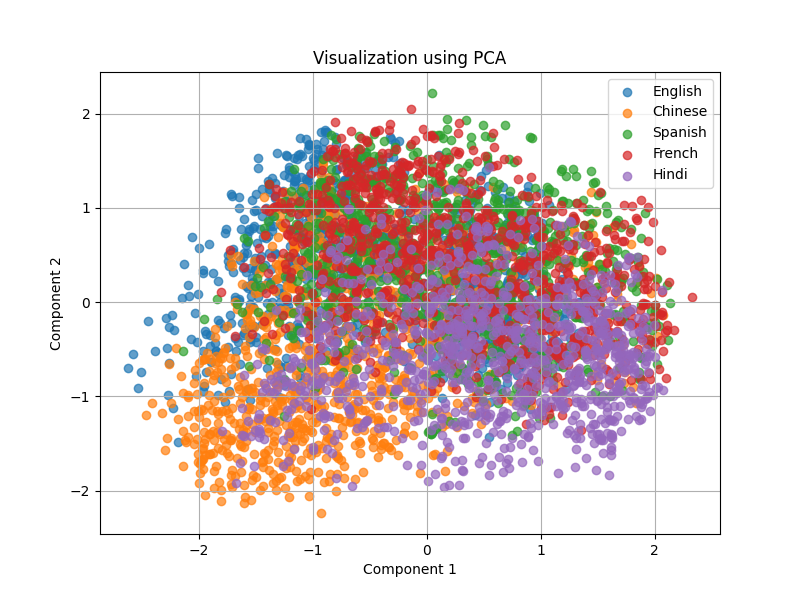}      \caption{PCA, Layer 11}        \end{subfigure}  \hfill  \begin{subfigure}{0.18\textwidth}      \includegraphics[width=\textwidth]{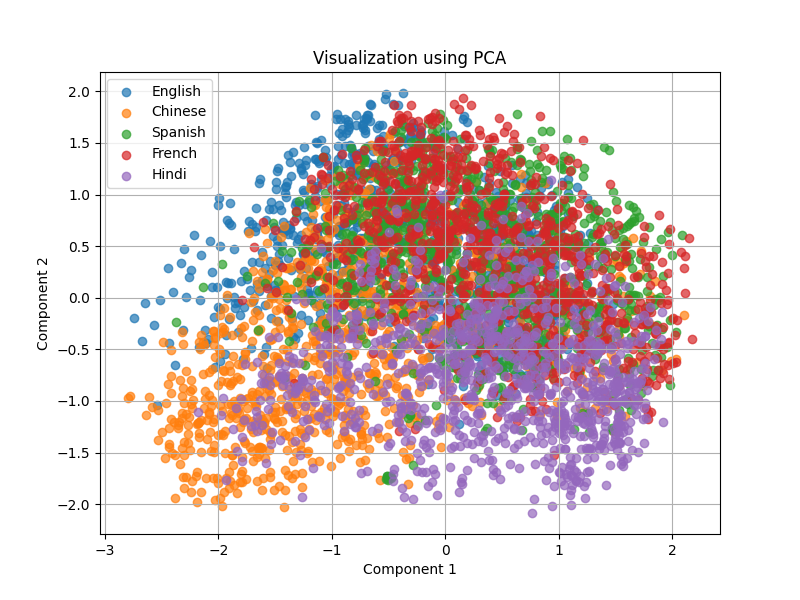}      \caption{PCA, Layer 12}        \end{subfigure}  \hfill  \begin{subfigure}{0.18\textwidth}      \includegraphics[width=\textwidth]{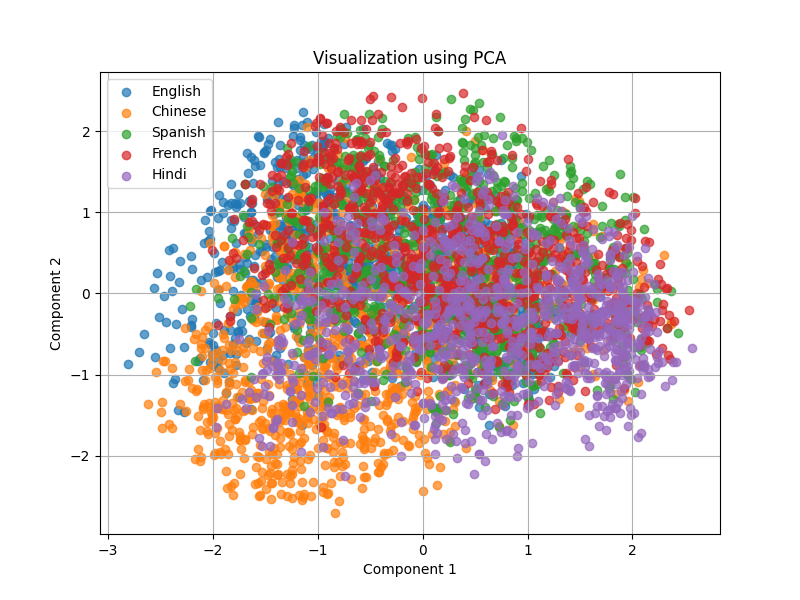}      \caption{PCA, Layer 13}        \end{subfigure}  \hfill  \begin{subfigure}{0.18\textwidth}      \includegraphics[width=\textwidth]{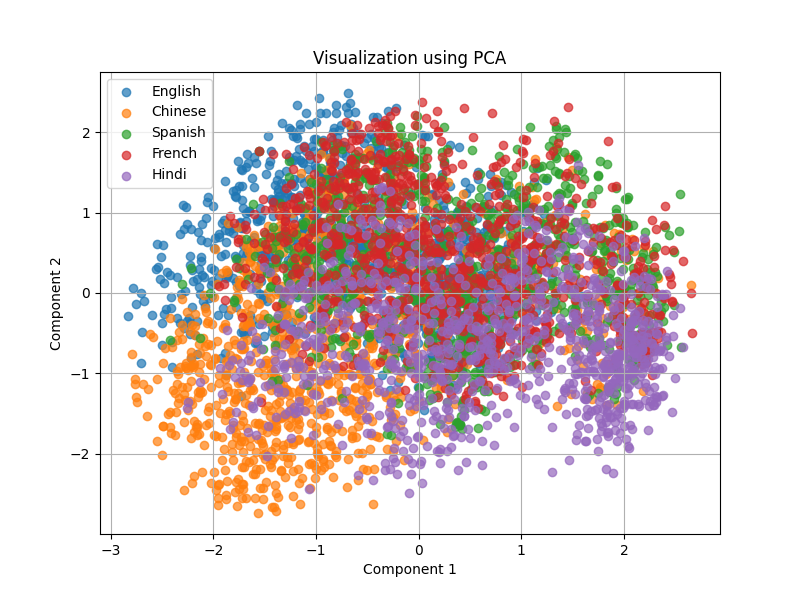}      \caption{PCA, Layer 14}        \end{subfigure}  \hfill  \begin{subfigure}{0.18\textwidth}      \includegraphics[width=\textwidth]{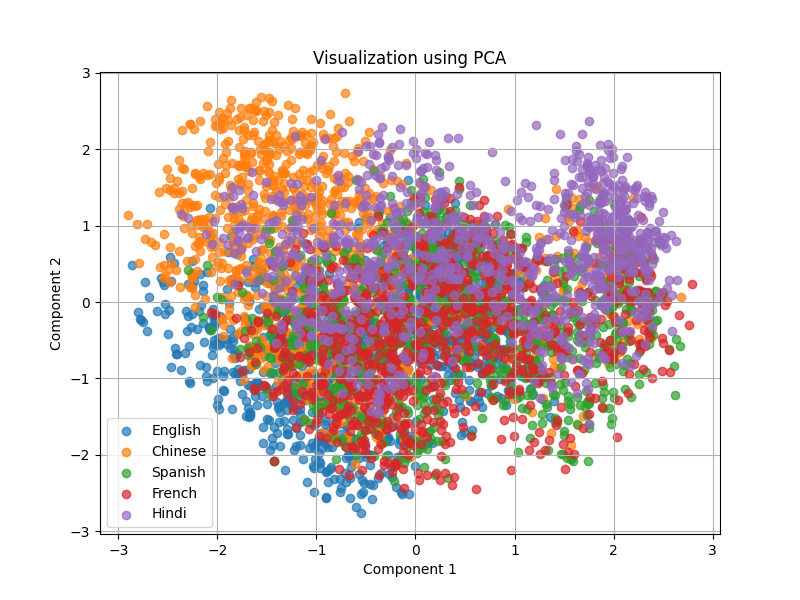}      \caption{PCA, Layer 15}        \end{subfigure}    \vspace{0.2in}    %
\begin{subfigure}{0.18\textwidth}      \includegraphics[width=\textwidth]{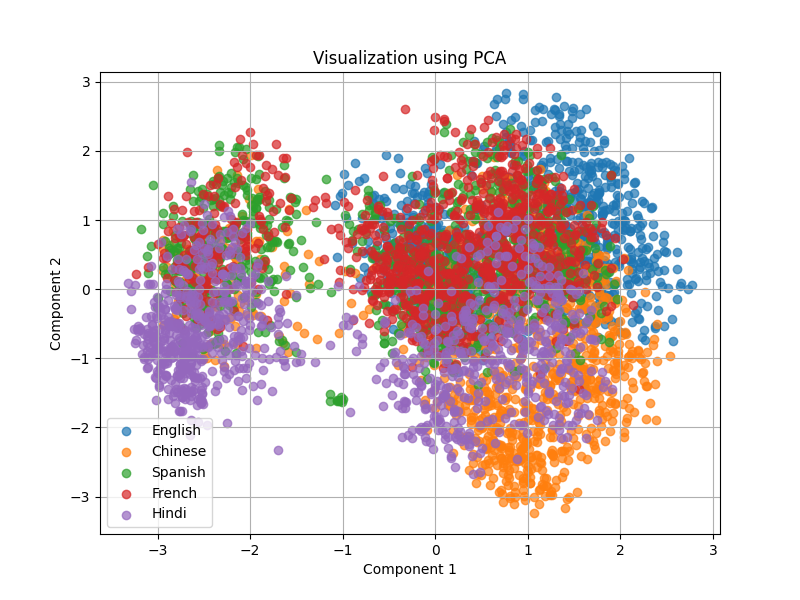}      \caption{PCA, Layer 16}        \end{subfigure}  \hfill  \begin{subfigure}{0.18\textwidth}      \includegraphics[width=\textwidth]{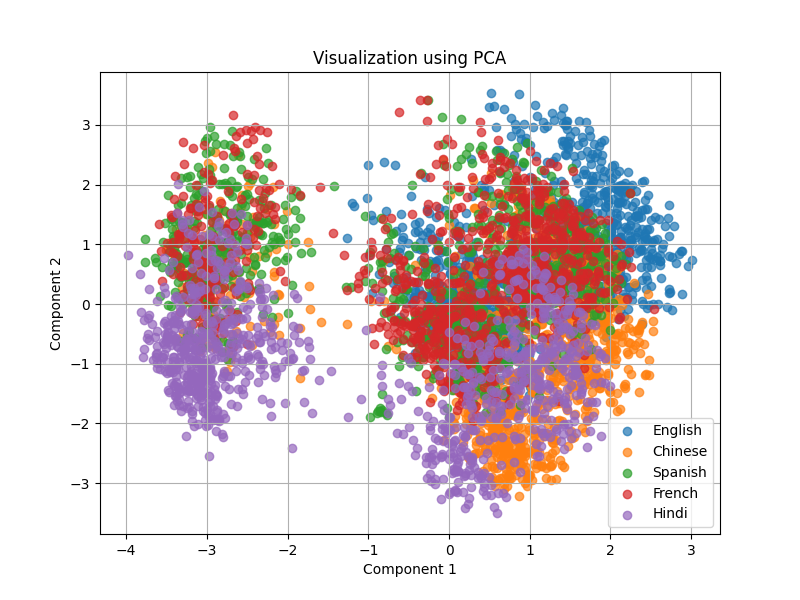}      \caption{PCA, Layer 17}        \end{subfigure}  \hfill  \begin{subfigure}{0.18\textwidth}      \includegraphics[width=\textwidth]{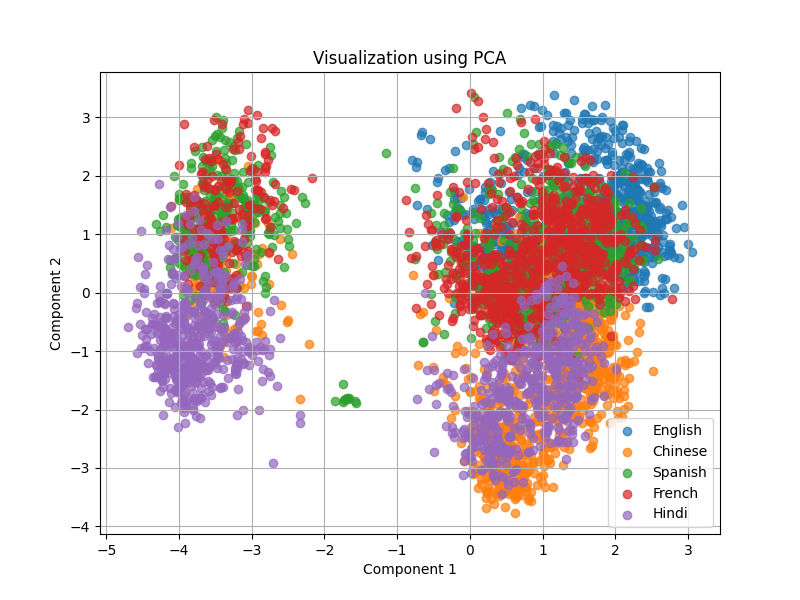}      \caption{PCA, Layer 18}        \end{subfigure}  \hfill  \begin{subfigure}{0.18\textwidth}      \includegraphics[width=\textwidth]{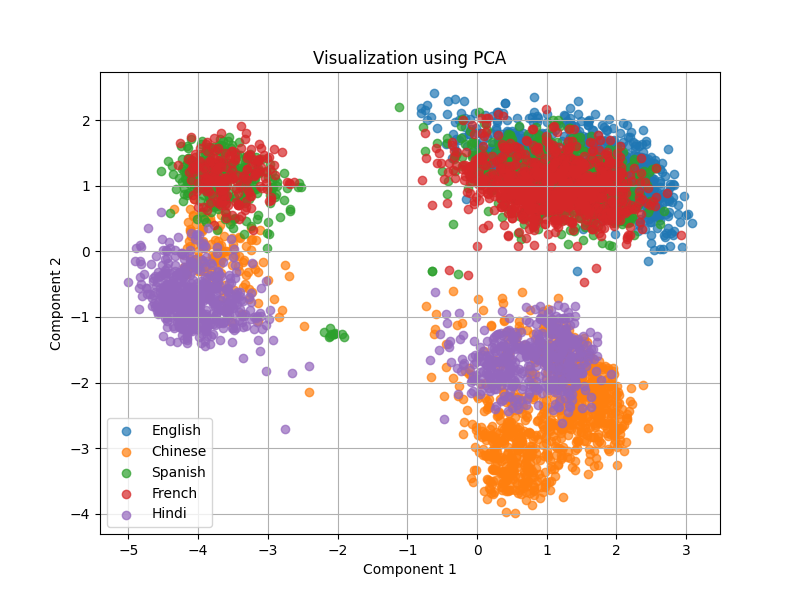}      \caption{PCA, Layer 19}        \end{subfigure}  \hfill  \begin{subfigure}{0.18\textwidth}      \includegraphics[width=\textwidth]{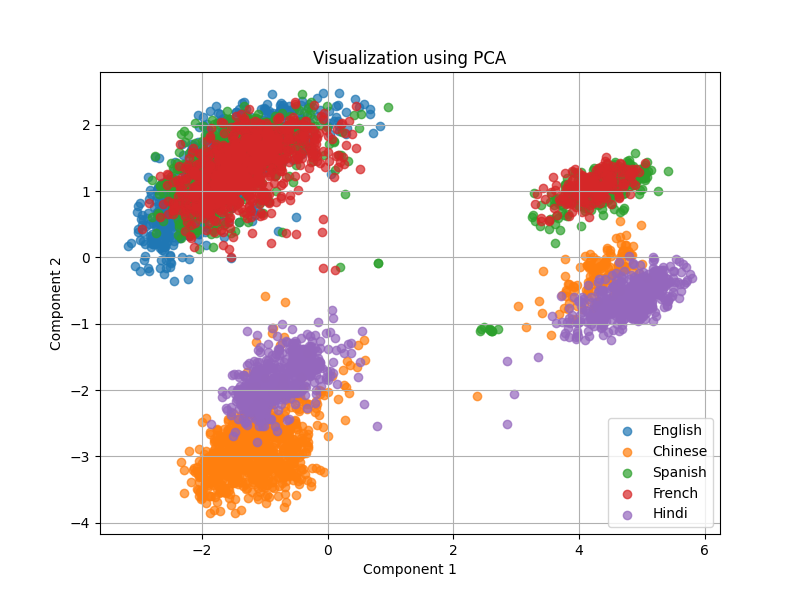}      \caption{PCA, Layer 20}        \end{subfigure}    \vspace{0.2in}    %
\begin{subfigure}{0.18\textwidth}      \includegraphics[width=\textwidth]{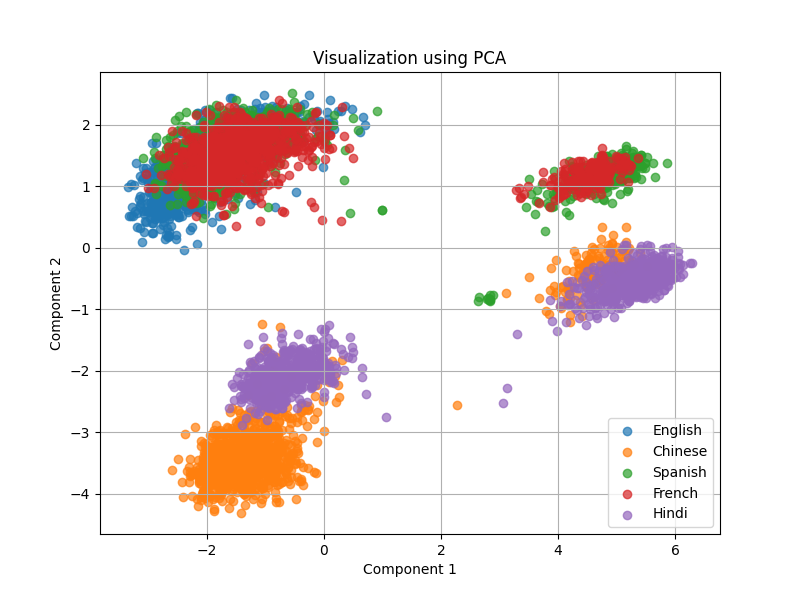}      \caption{PCA, Layer 21}        \end{subfigure}  \hfill  \begin{subfigure}{0.18\textwidth}      \includegraphics[width=\textwidth]{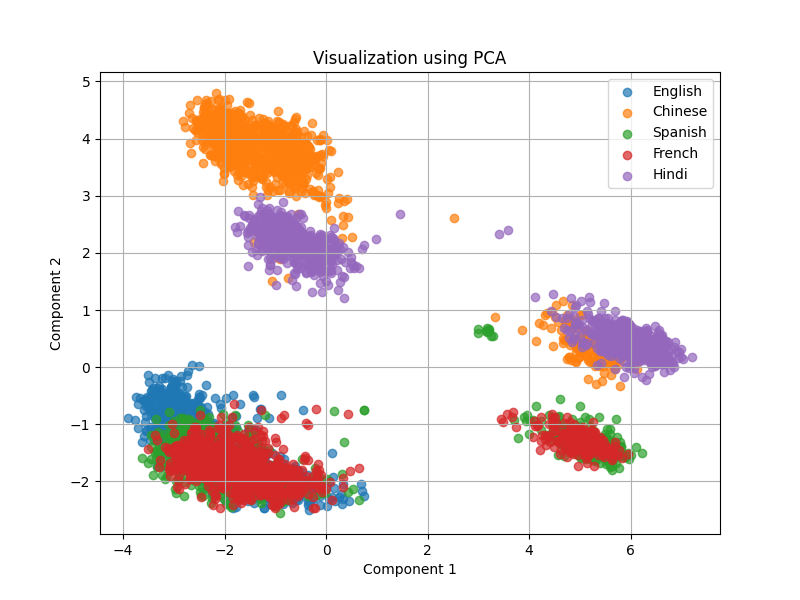}      \caption{PCA, Layer 22}        \end{subfigure}  \hfill  \begin{subfigure}{0.18\textwidth}      \includegraphics[width=\textwidth]{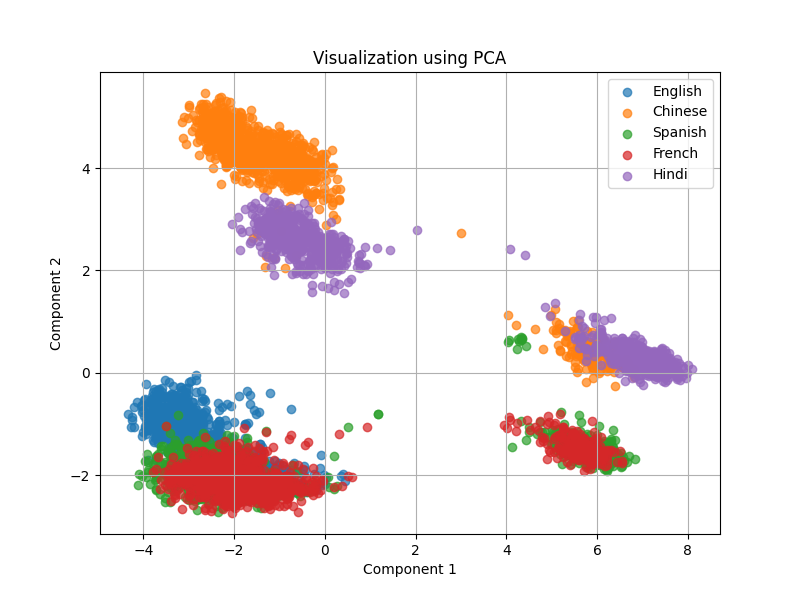}      \caption{PCA, Layer 23}        \end{subfigure}  \hfill  \begin{subfigure}{0.18\textwidth}      \includegraphics[width=\textwidth]{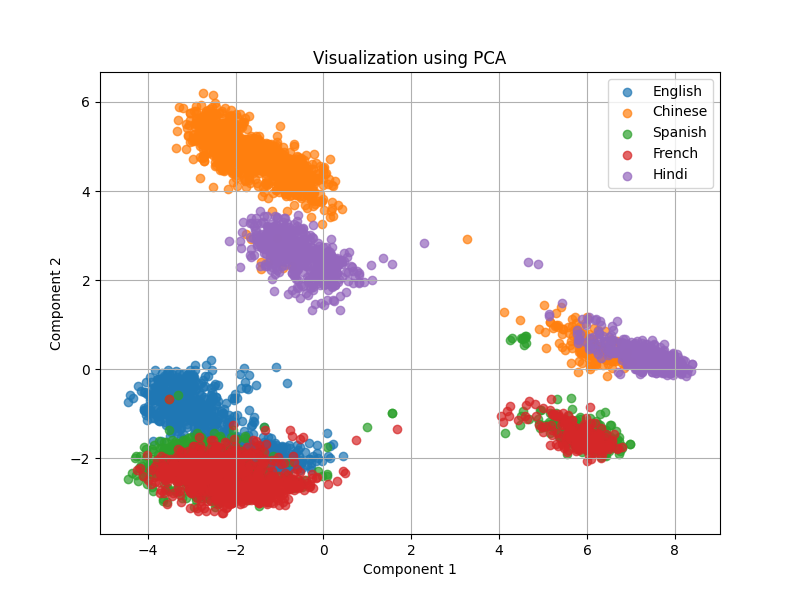}      \caption{PCA, Layer 24}        \end{subfigure}  \hfill  \begin{subfigure}{0.18\textwidth}      \includegraphics[width=\textwidth]{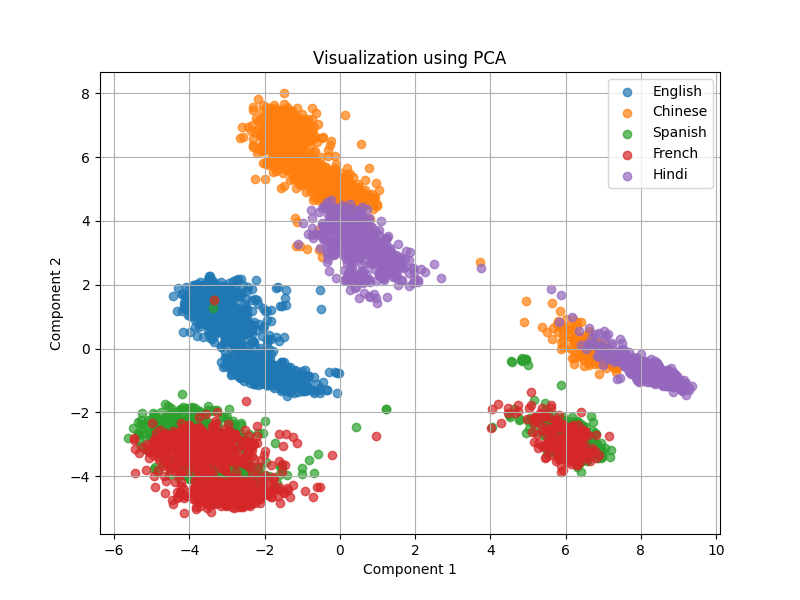}      \caption{PCA, Layer 25}        \end{subfigure}    \vspace{0.2in}    %
\begin{subfigure}{0.18\textwidth}      \includegraphics[width=\textwidth]{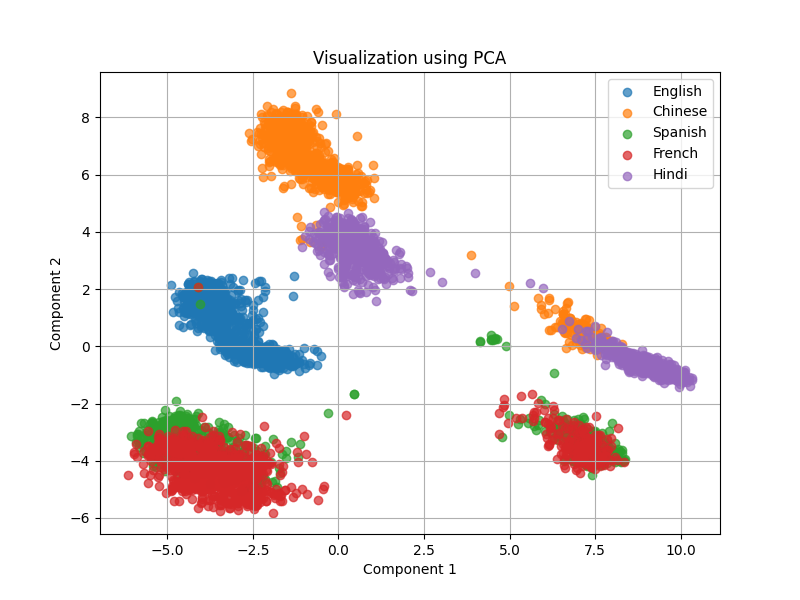}      \caption{PCA, Layer 26}        \end{subfigure}  \hfill  \begin{subfigure}{0.18\textwidth}      \includegraphics[width=\textwidth]{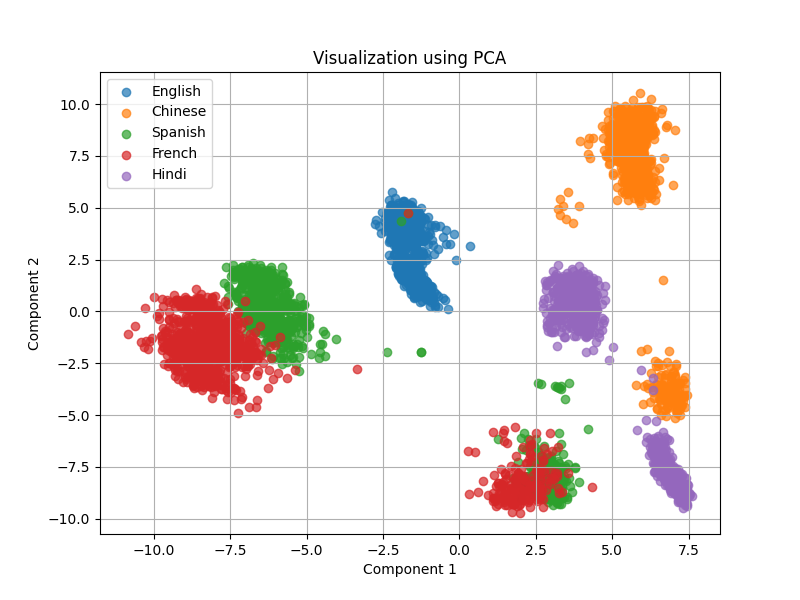}      \caption{PCA, Layer 27}        \end{subfigure}  \hfill  \begin{subfigure}{0.18\textwidth}      \includegraphics[width=\textwidth]{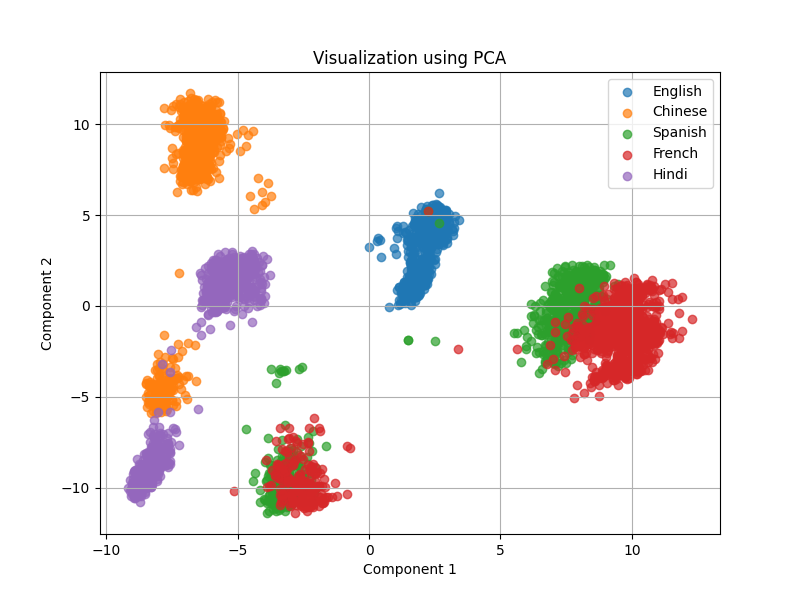}      \caption{PCA, Layer 28}        \end{subfigure}  \hfill  \begin{subfigure}{0.18\textwidth}      \includegraphics[width=\textwidth]{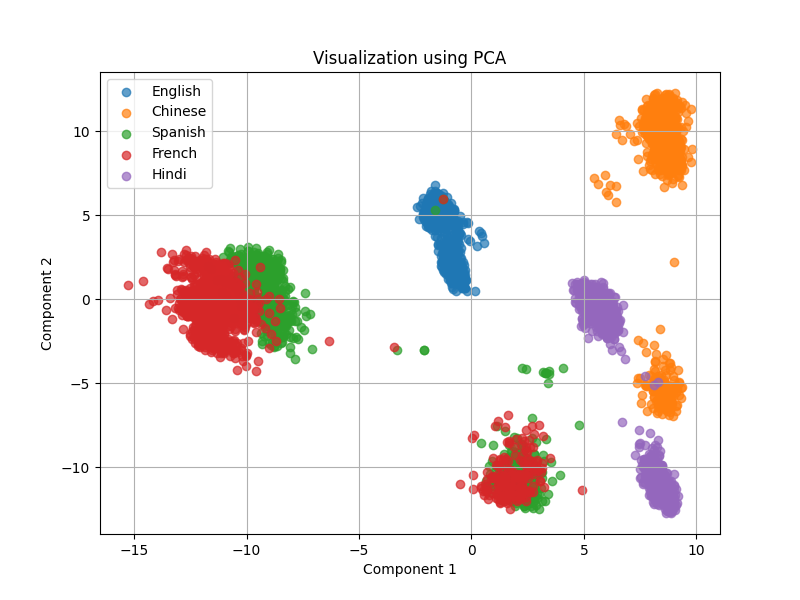}      \caption{PCA, Layer 29}        \end{subfigure}  \hfill  \begin{subfigure}{0.18\textwidth}      \includegraphics[width=\textwidth]{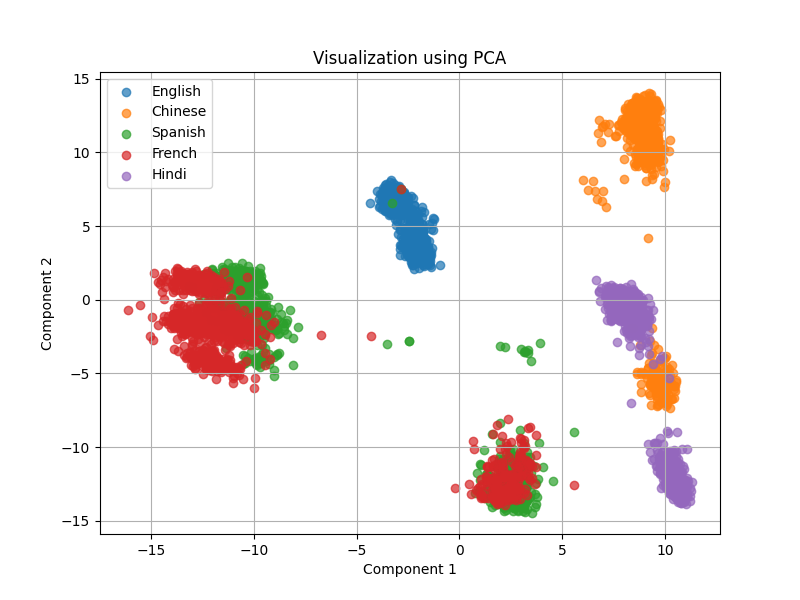}      \caption{PCA, Layer 30}        \end{subfigure}    \vspace{0.2in}    %
\begin{subfigure}{0.18\textwidth}      \includegraphics[width=\textwidth]{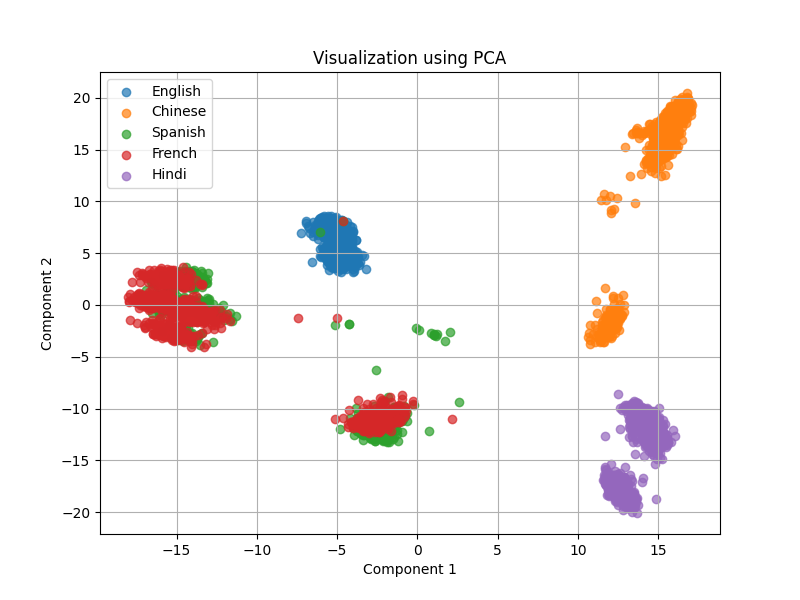}      \caption{PCA, Layer 31}        \end{subfigure}  \hfill  \begin{subfigure}{0.18\textwidth}      \includegraphics[width=\textwidth]{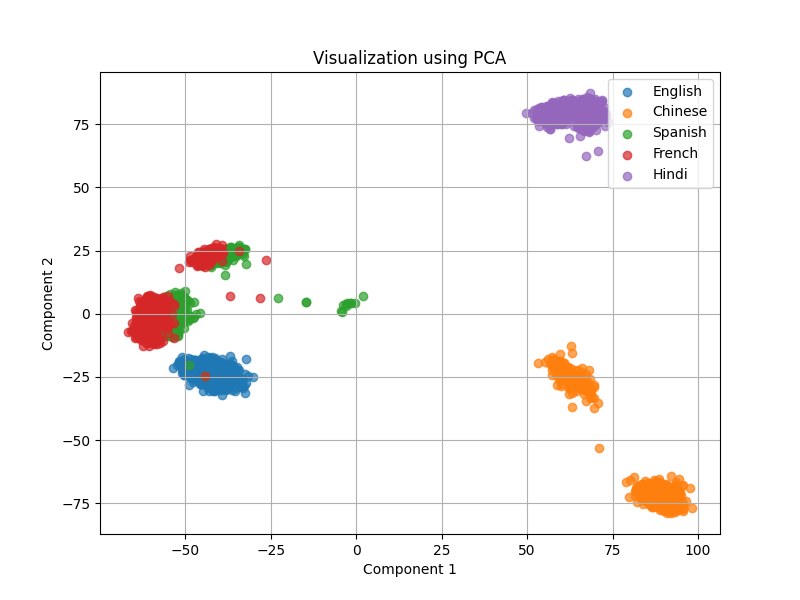}      \caption{PCA, Layer 32}        \end{subfigure}    \caption{PCA visualizations for layers 1-32 of Llama-3-8B-Instruct on the ProofWriter dataset.}  
\end{figure*}

\begin{figure*}[htbp]
\centering
\begin{subfigure}{0.18\textwidth}
\includegraphics[width=\textwidth]{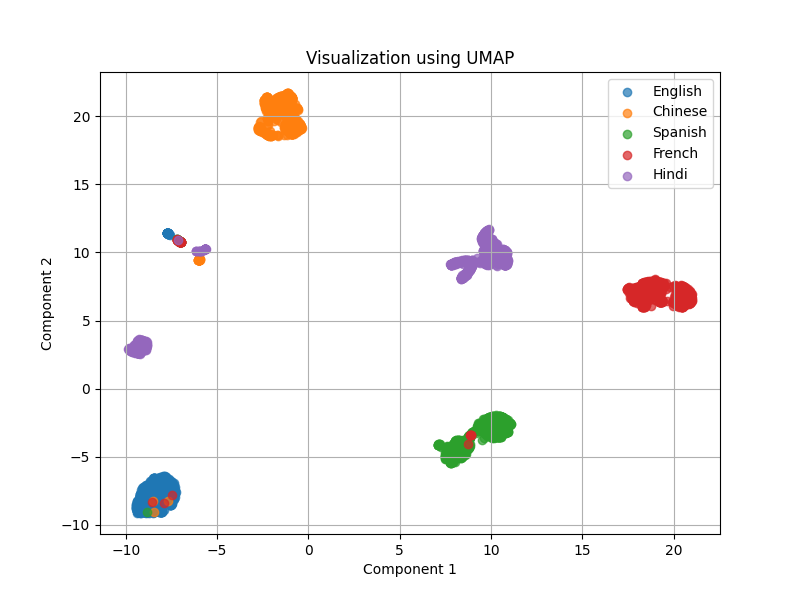}
\caption{UMAP, Layer 1}
\end{subfigure}
\hfill
\begin{subfigure}{0.18\textwidth}
\includegraphics[width=\textwidth]{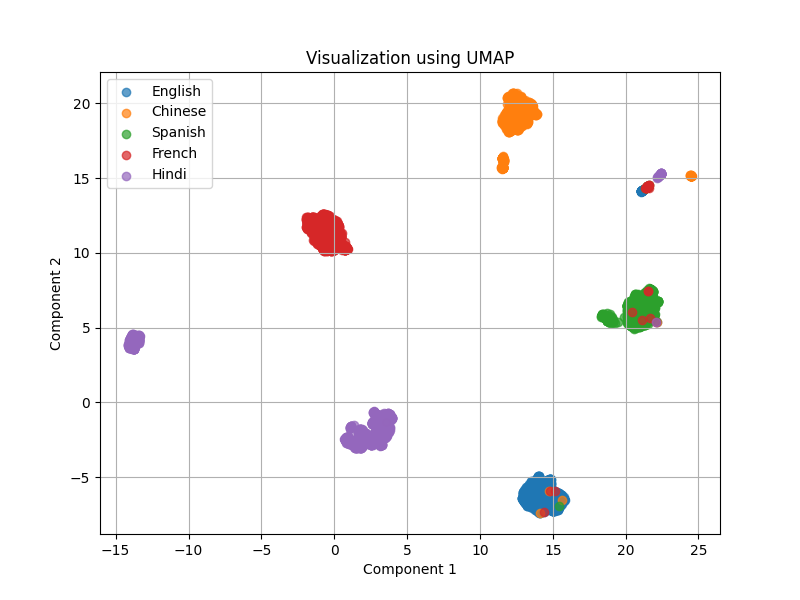}
\caption{UMAP, Layer 2}

\end{subfigure}
\hfill
\begin{subfigure}{0.18\textwidth}
\includegraphics[width=\textwidth]{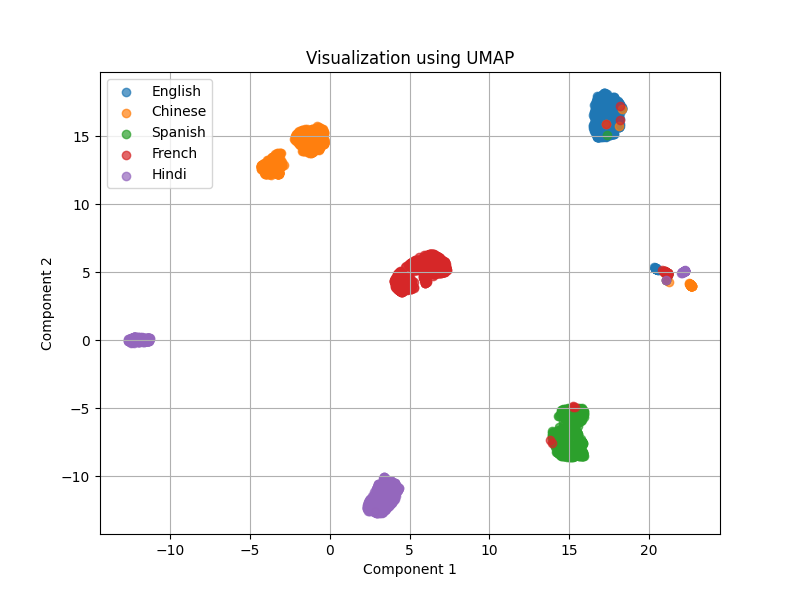}
\caption{UMAP, Layer 3}

\end{subfigure}
\hfill
\begin{subfigure}{0.18\textwidth}
\includegraphics[width=\textwidth]{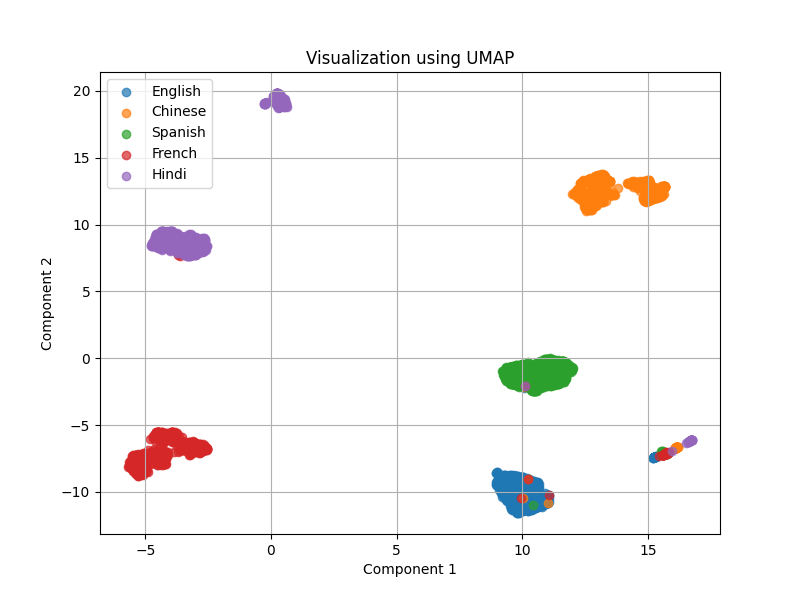}
\caption{UMAP, Layer 4}

\end{subfigure}
\hfill
\begin{subfigure}{0.18\textwidth}
\includegraphics[width=\textwidth]{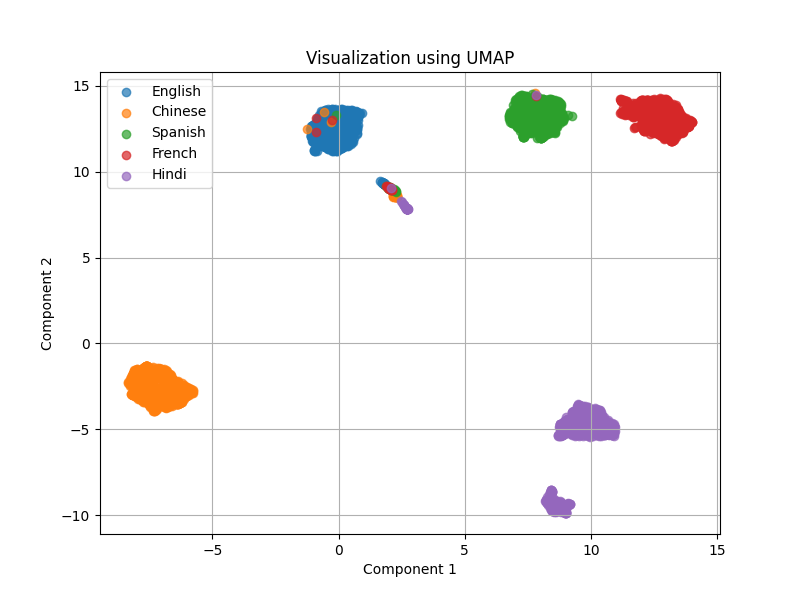}
\caption{UMAP, Layer 5}

\end{subfigure}
\vspace{0.2in} %
\begin{subfigure}{0.18\textwidth}      \includegraphics[width=\textwidth]{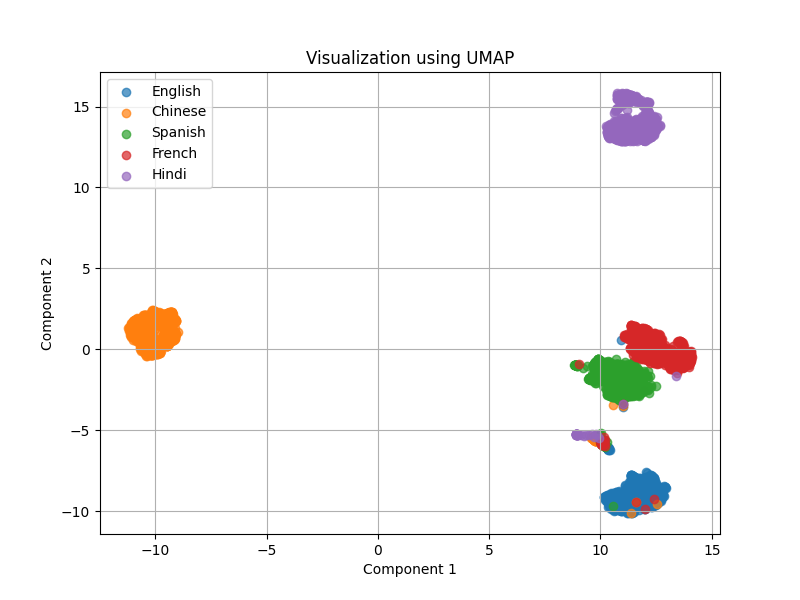}      \caption{UMAP, Layer 6}        \end{subfigure}  \hfill  \begin{subfigure}{0.18\textwidth}      \includegraphics[width=\textwidth]{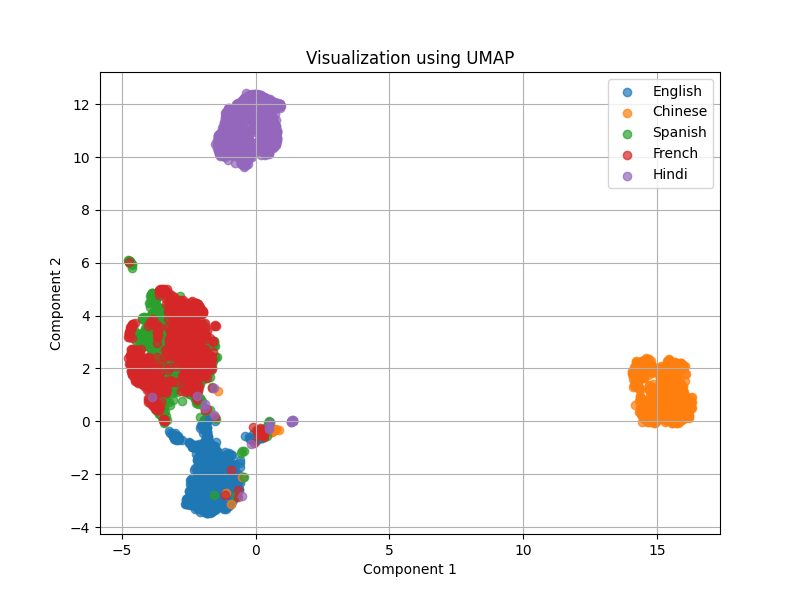}      \caption{UMAP, Layer 7}        \end{subfigure}  \hfill  \begin{subfigure}{0.18\textwidth}      \includegraphics[width=\textwidth]{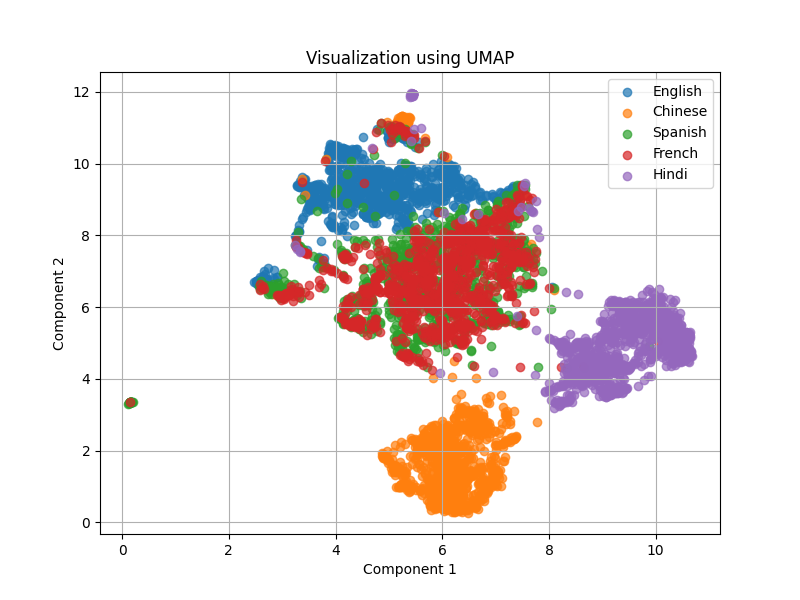}      \caption{UMAP, Layer 8}        \end{subfigure}  \hfill  \begin{subfigure}{0.18\textwidth}      \includegraphics[width=\textwidth]{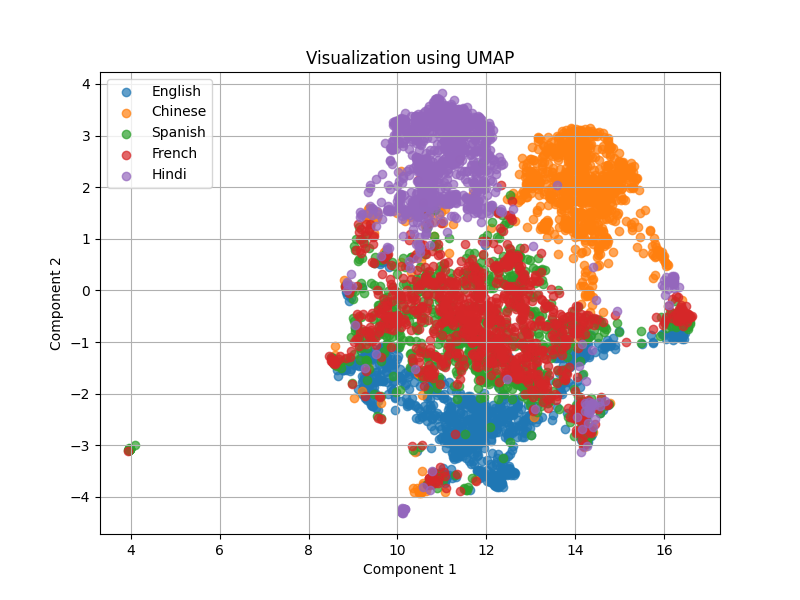}      \caption{UMAP, Layer 9}        \end{subfigure}  \hfill  \begin{subfigure}{0.18\textwidth}      \includegraphics[width=\textwidth]{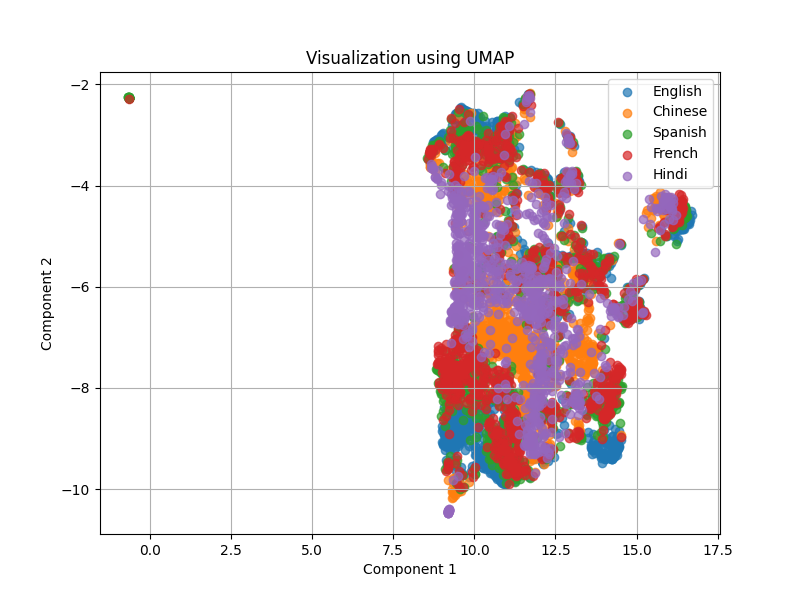}      \caption{UMAP, Layer 10}        \end{subfigure}    \vspace{0.2in}    %
\begin{subfigure}{0.18\textwidth}      \includegraphics[width=\textwidth]{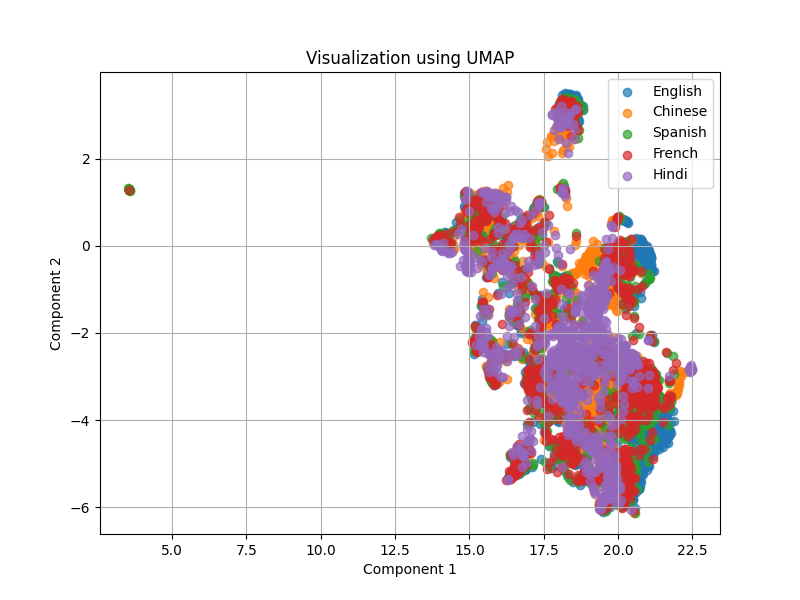}      \caption{UMAP, Layer 11}        \end{subfigure}  \hfill  \begin{subfigure}{0.18\textwidth}      \includegraphics[width=\textwidth]{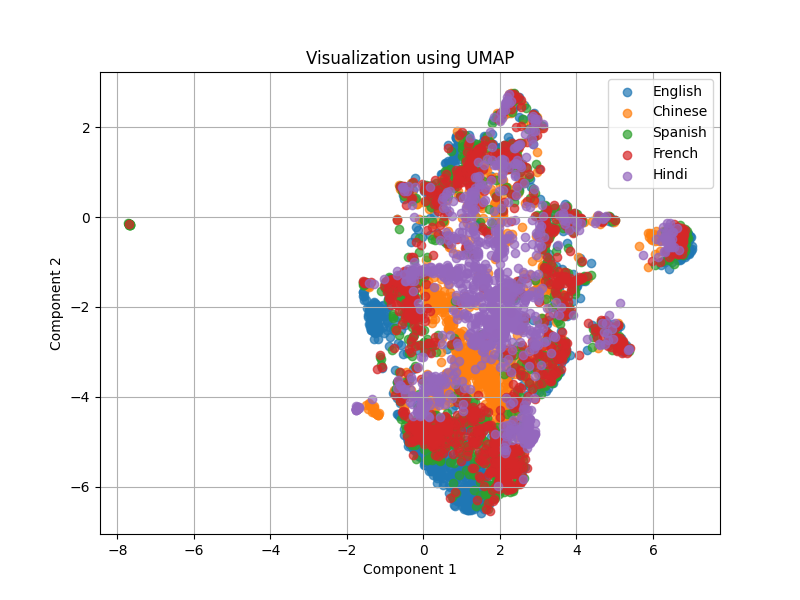}      \caption{UMAP, Layer 12}        \end{subfigure}  \hfill  \begin{subfigure}{0.18\textwidth}      \includegraphics[width=\textwidth]{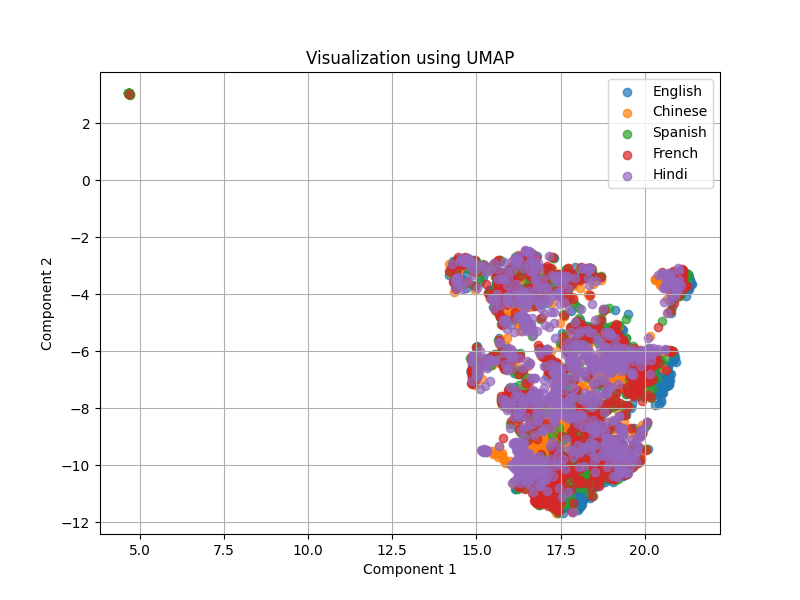}      \caption{UMAP, Layer 13}        \end{subfigure}  \hfill  \begin{subfigure}{0.18\textwidth}      \includegraphics[width=\textwidth]{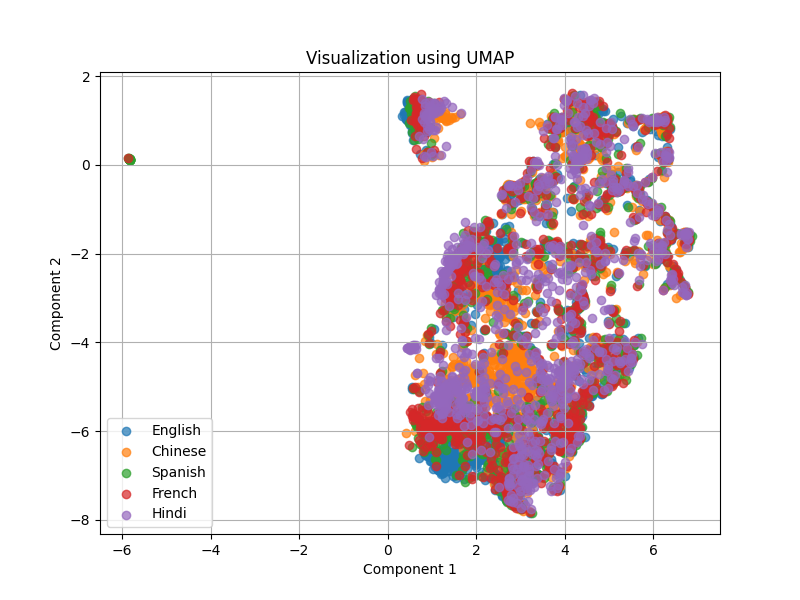}      \caption{UMAP, Layer 14}        \end{subfigure}  \hfill  \begin{subfigure}{0.18\textwidth}      \includegraphics[width=\textwidth]{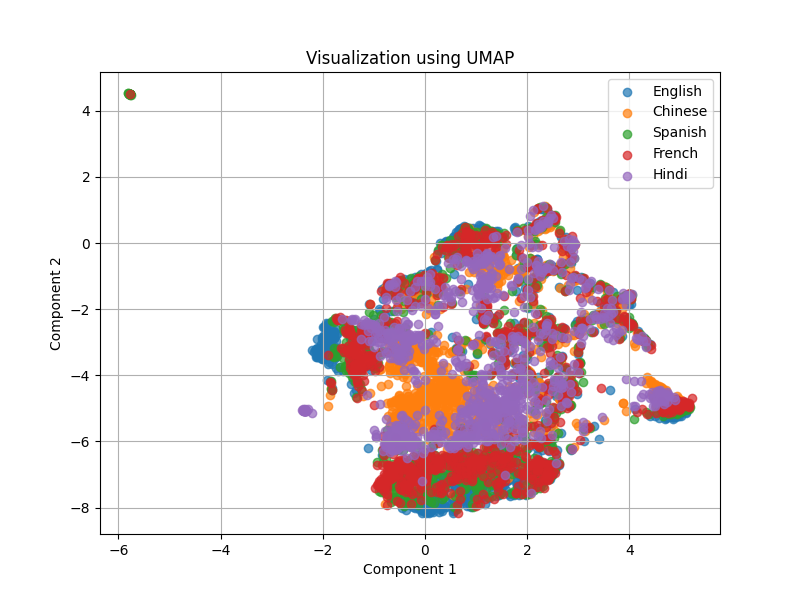}      \caption{UMAP, Layer 15}        \end{subfigure}    \vspace{0.2in}    %
\begin{subfigure}{0.18\textwidth}      \includegraphics[width=\textwidth]{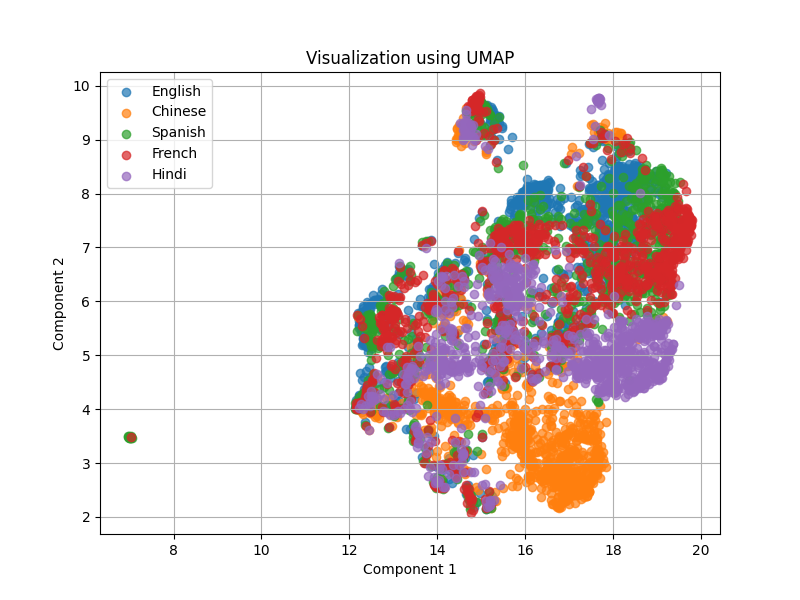}      \caption{UMAP, Layer 16}        \end{subfigure}  \hfill  \begin{subfigure}{0.18\textwidth}      \includegraphics[width=\textwidth]{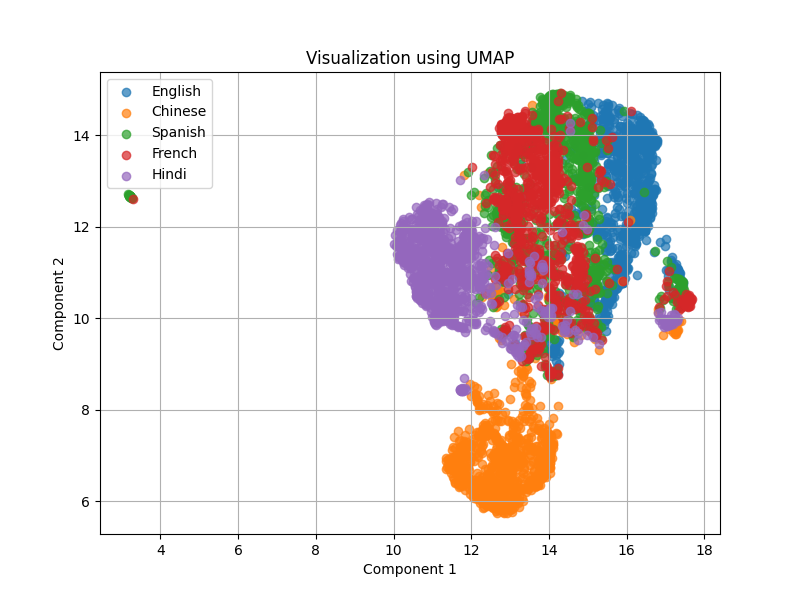}      \caption{UMAP, Layer 17}        \end{subfigure}  \hfill  \begin{subfigure}{0.18\textwidth}      \includegraphics[width=\textwidth]{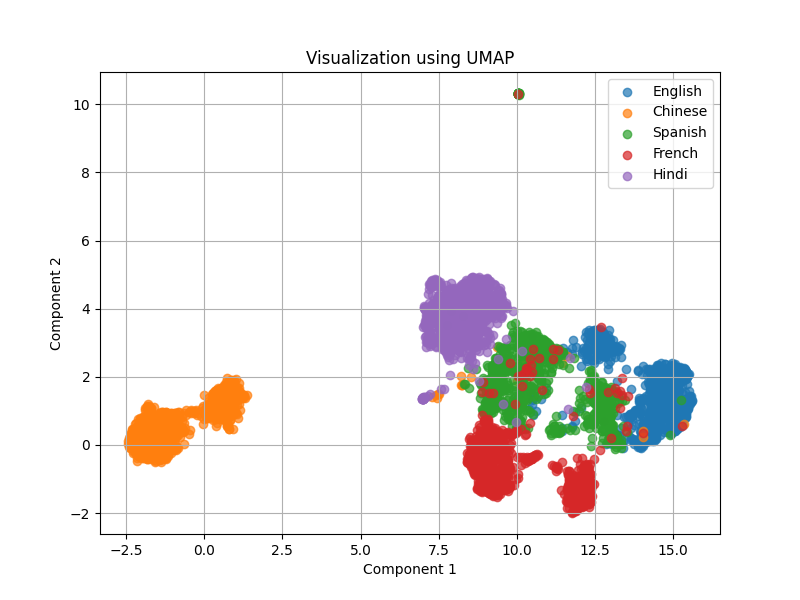}      \caption{UMAP, Layer 18}        \end{subfigure}  \hfill  \begin{subfigure}{0.18\textwidth}      \includegraphics[width=\textwidth]{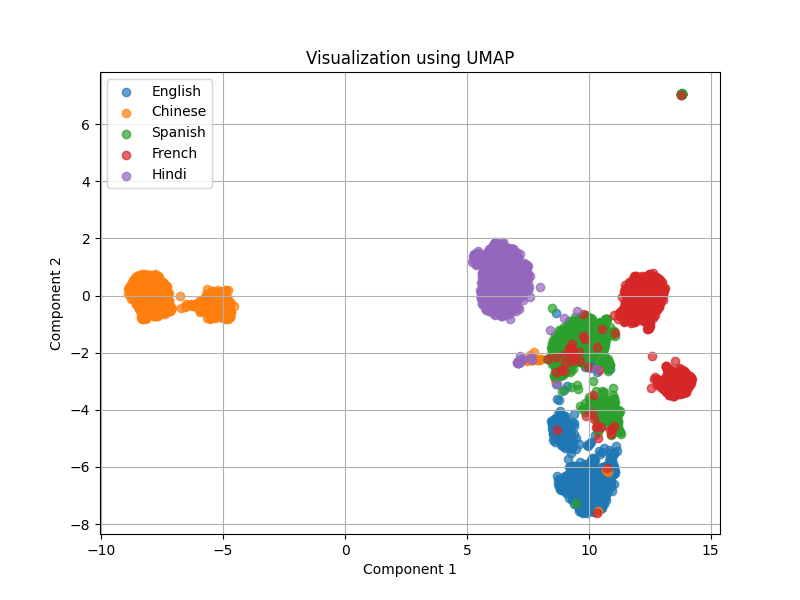}      \caption{UMAP, Layer 19}        \end{subfigure}  \hfill  \begin{subfigure}{0.18\textwidth}      \includegraphics[width=\textwidth]{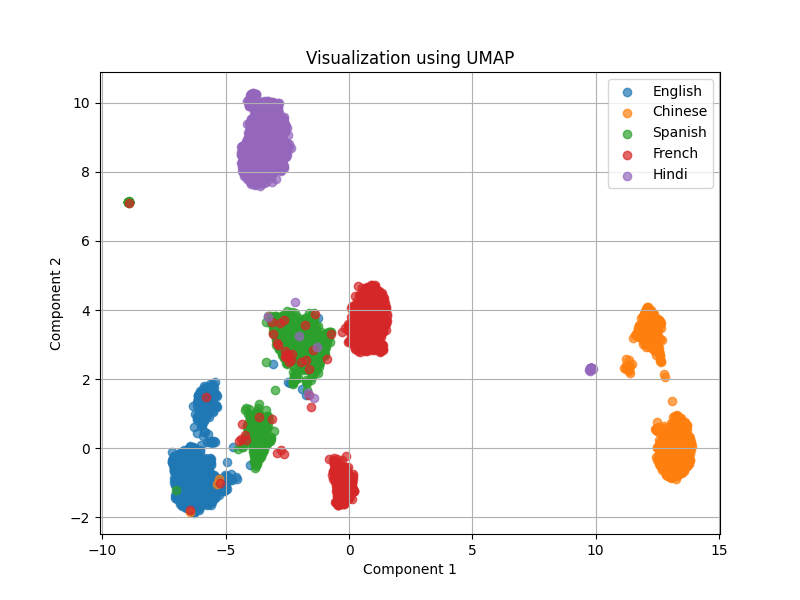}      \caption{UMAP, Layer 20}        \end{subfigure}    \vspace{0.2in}    %
\begin{subfigure}{0.18\textwidth}      \includegraphics[width=\textwidth]{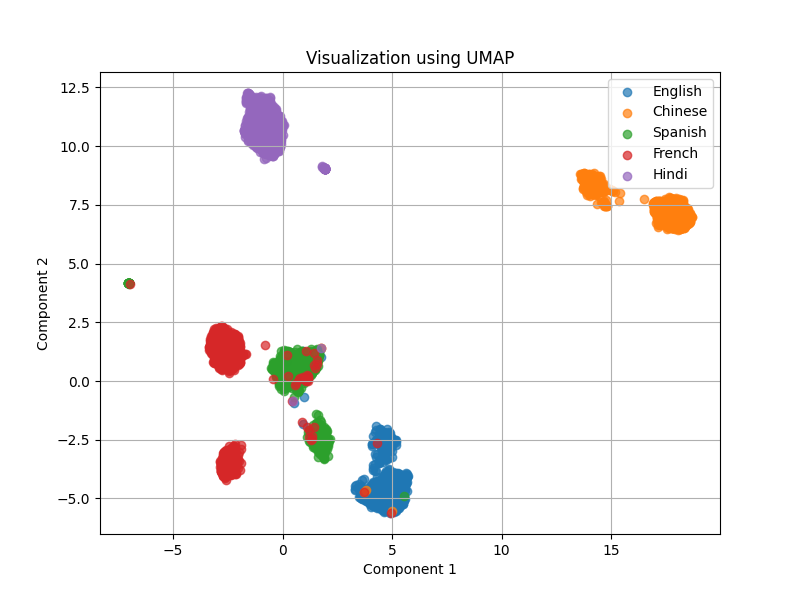}      \caption{UMAP, Layer 21}        \end{subfigure}  \hfill  \begin{subfigure}{0.18\textwidth}      \includegraphics[width=\textwidth]{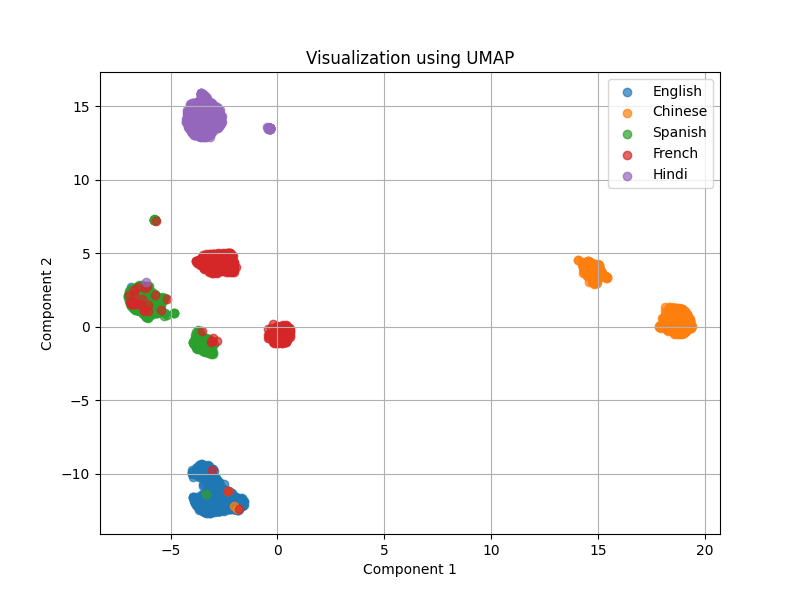}      \caption{UMAP, Layer 22}        \end{subfigure}  \hfill  \begin{subfigure}{0.18\textwidth}      \includegraphics[width=\textwidth]{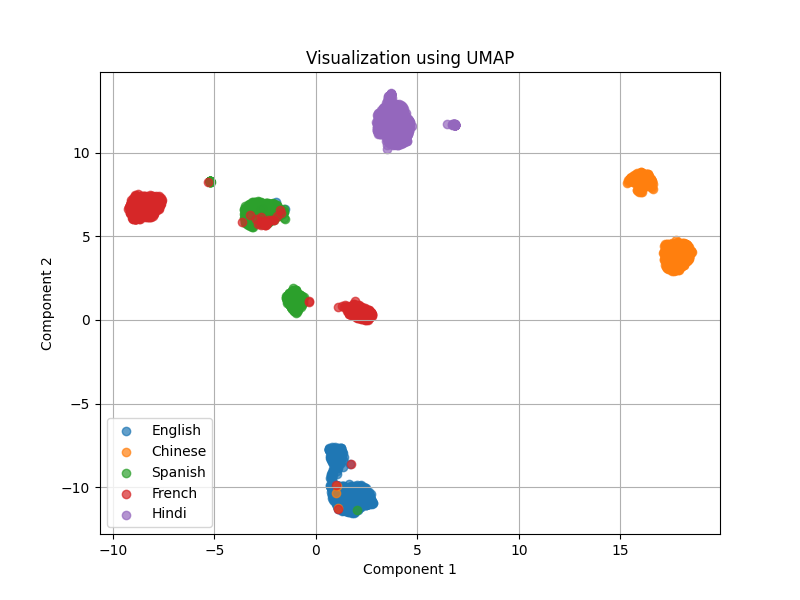}      \caption{UMAP, Layer 23}        \end{subfigure}  \hfill  \begin{subfigure}{0.18\textwidth}      \includegraphics[width=\textwidth]{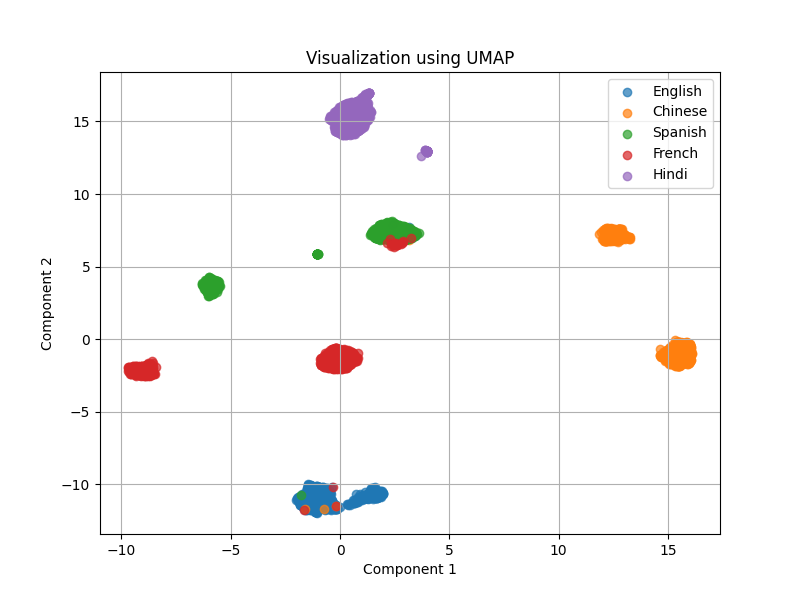}      \caption{UMAP, Layer 24}        \end{subfigure}  \hfill  \begin{subfigure}{0.18\textwidth}      \includegraphics[width=\textwidth]{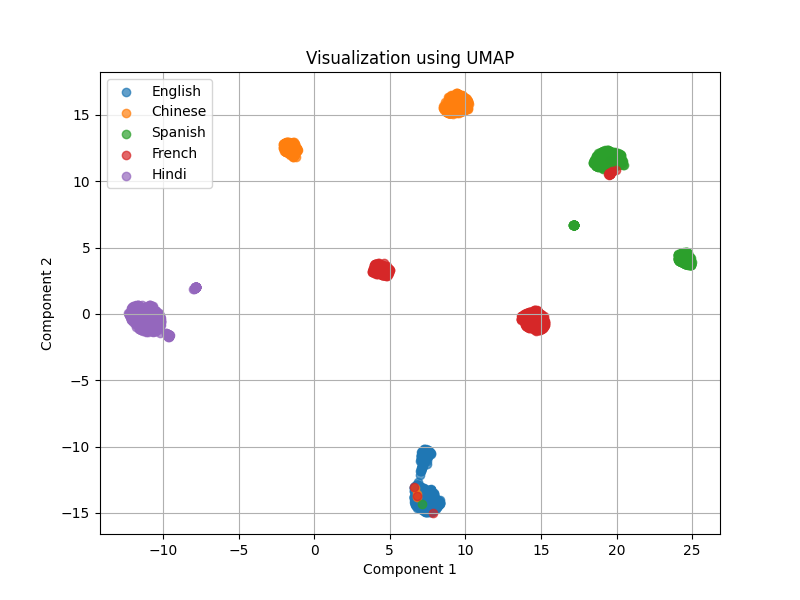}      \caption{UMAP, Layer 25}        \end{subfigure}    \vspace{0.2in}    %
\begin{subfigure}{0.18\textwidth}      \includegraphics[width=\textwidth]{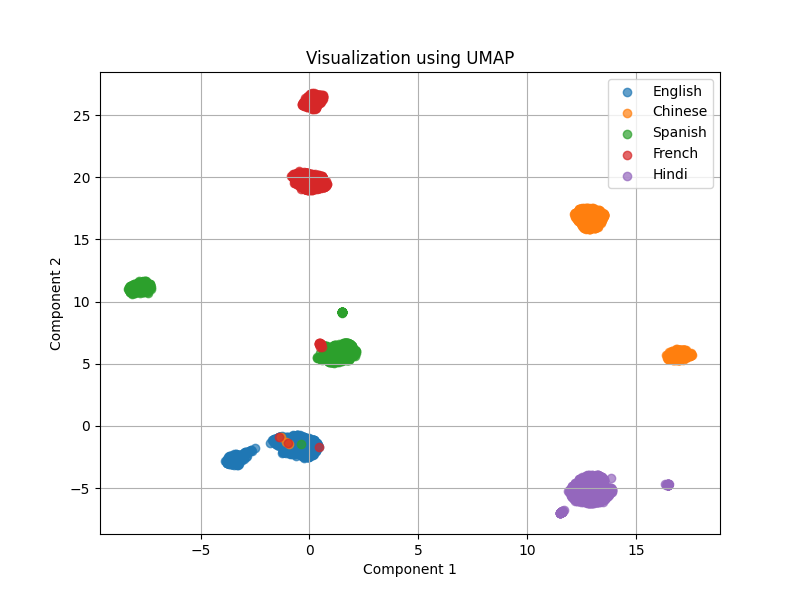}      \caption{UMAP, Layer 26}        \end{subfigure}  \hfill  \begin{subfigure}{0.18\textwidth}      \includegraphics[width=\textwidth]{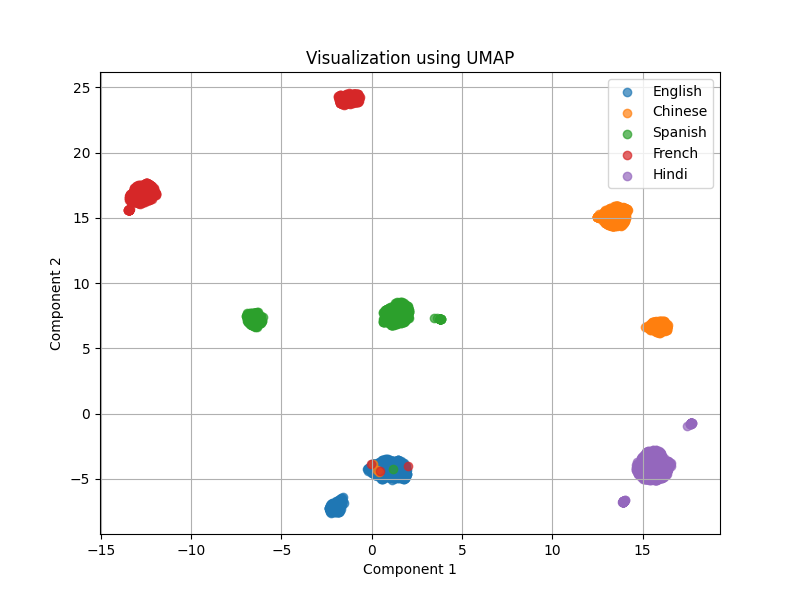}      \caption{UMAP, Layer 27}        \end{subfigure}  \hfill  \begin{subfigure}{0.18\textwidth}      \includegraphics[width=\textwidth]{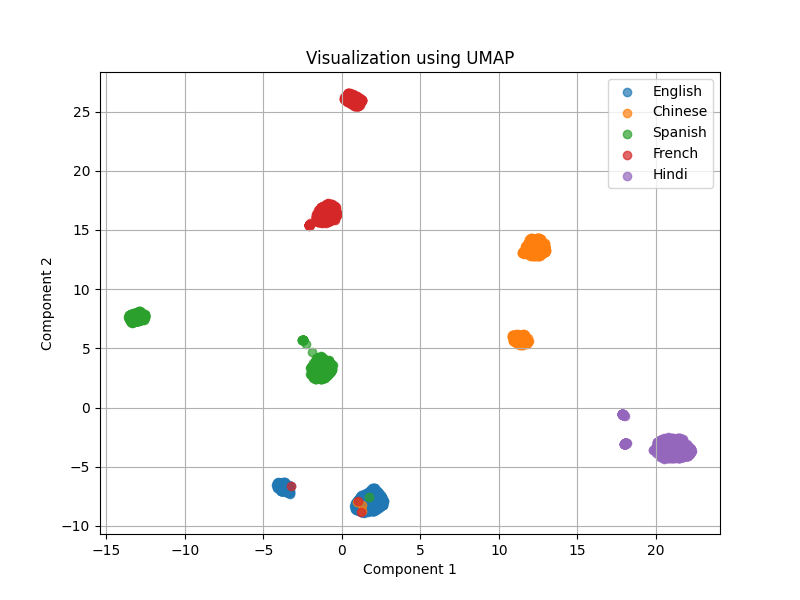}      \caption{UMAP, Layer 28}        \end{subfigure}  \hfill  \begin{subfigure}{0.18\textwidth}      \includegraphics[width=\textwidth]{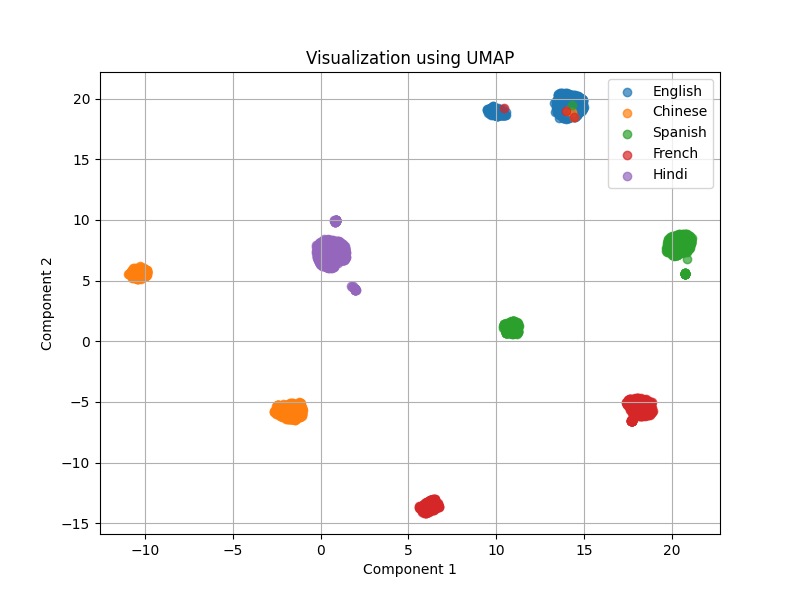}      \caption{UMAP, Layer 29}        \end{subfigure}  \hfill  \begin{subfigure}{0.18\textwidth}      \includegraphics[width=\textwidth]{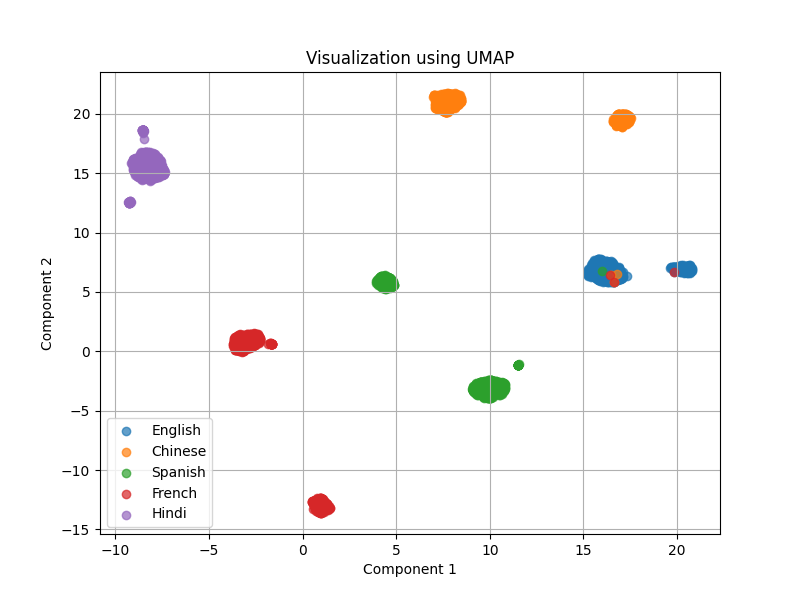}      \caption{UMAP, Layer 30}        \end{subfigure}    \vspace{0.2in}    %
\begin{subfigure}{0.18\textwidth}      \includegraphics[width=\textwidth]{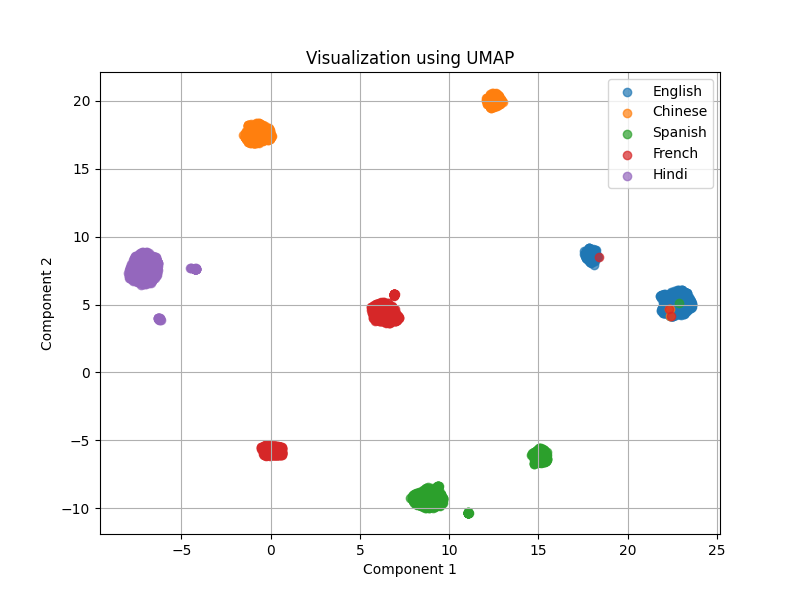}      \caption{UMAP, Layer 31}        \end{subfigure}  \hfill  \begin{subfigure}{0.18\textwidth}      \includegraphics[width=\textwidth]{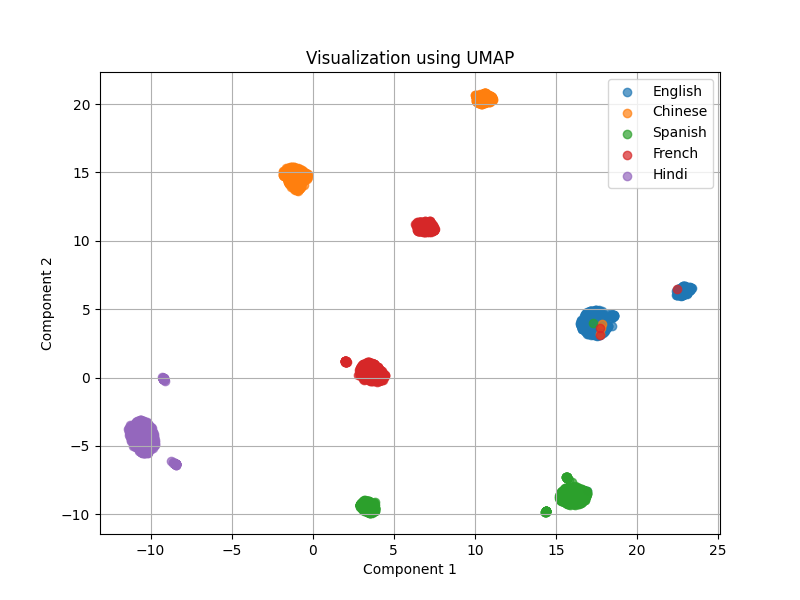}      \caption{UMAP, Layer 32}        \end{subfigure}    \caption{UMAP visualizations for layers 1-32 of Llama-3-8B-Instruct on the GSM8K dataset.}  
\end{figure*}

\begin{figure*}[htbp]
\centering
\begin{subfigure}{0.18\textwidth}
\includegraphics[width=\textwidth]{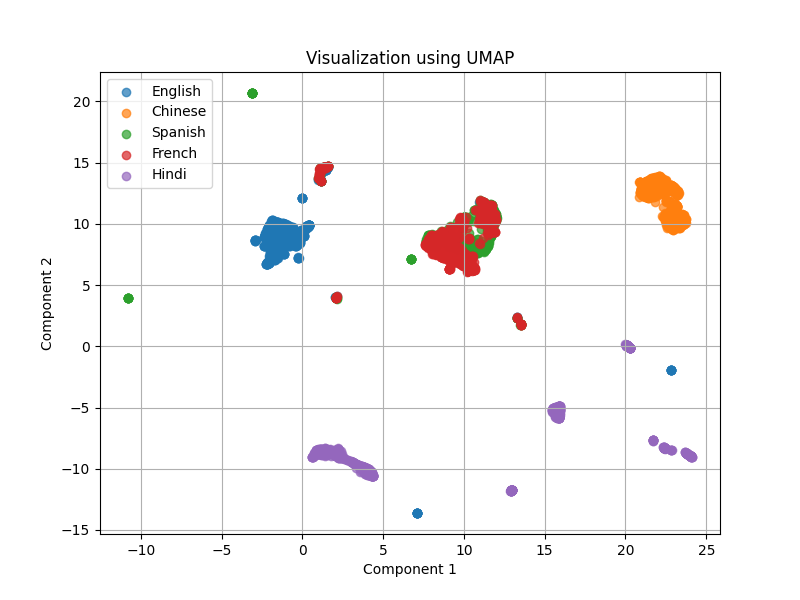}
\caption{UMAP, Layer 1}
\end{subfigure}
\hfill
\begin{subfigure}{0.18\textwidth}
\includegraphics[width=\textwidth]{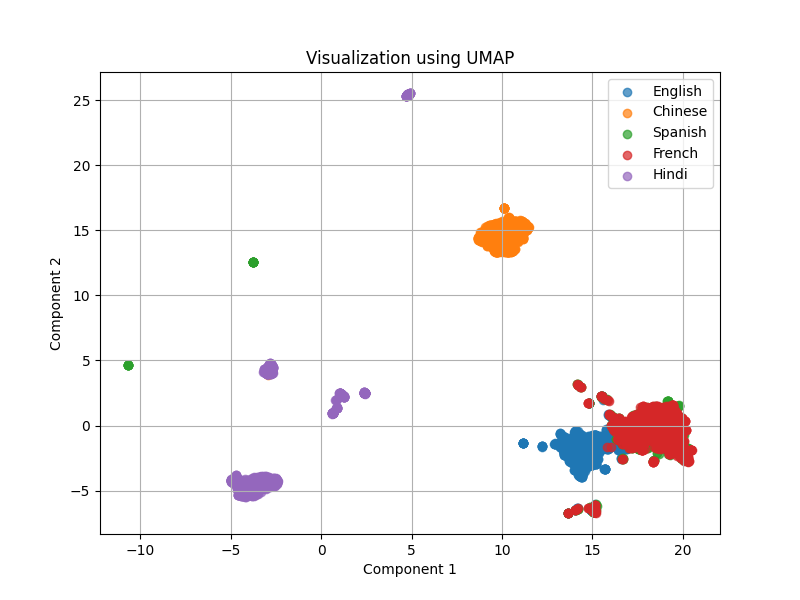}
\caption{UMAP, Layer 2}

\end{subfigure}
\hfill
\begin{subfigure}{0.18\textwidth}
\includegraphics[width=\textwidth]{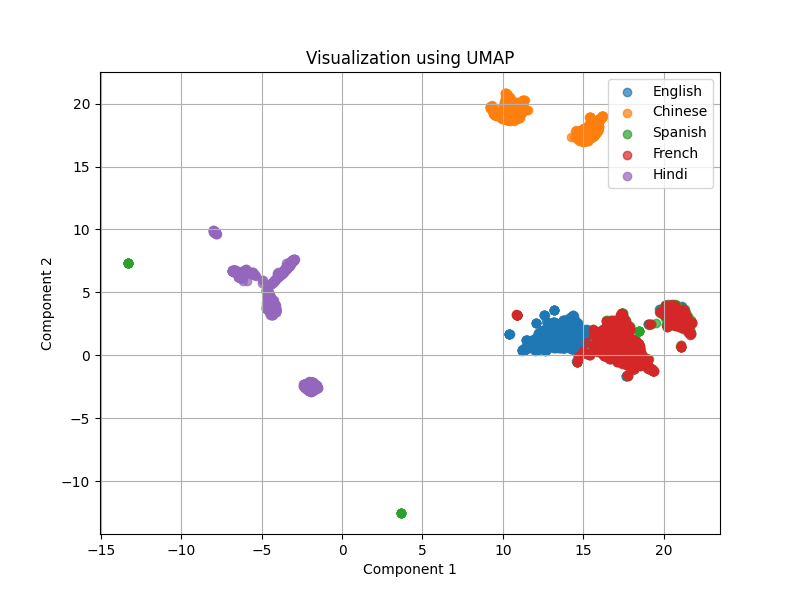}
\caption{UMAP, Layer 3}

\end{subfigure}
\hfill
\begin{subfigure}{0.18\textwidth}
\includegraphics[width=\textwidth]{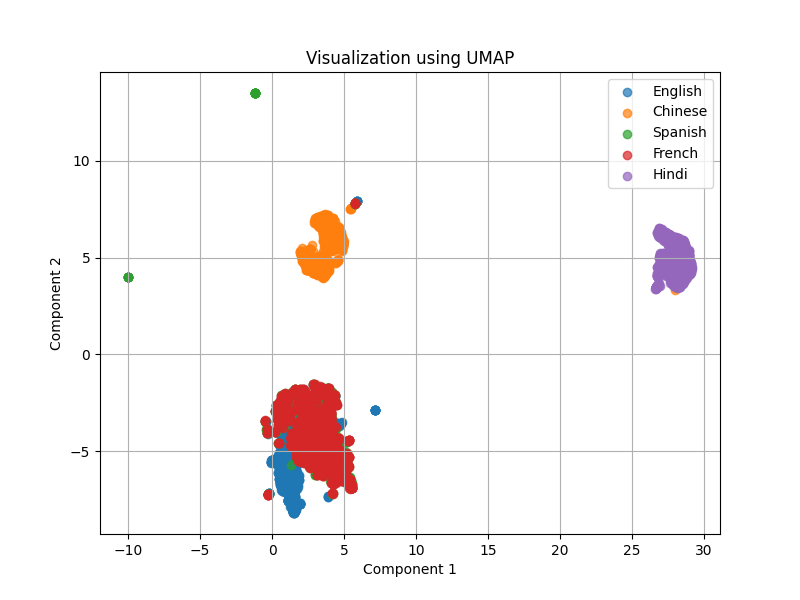}
\caption{UMAP, Layer 4}

\end{subfigure}
\hfill
\begin{subfigure}{0.18\textwidth}
\includegraphics[width=\textwidth]{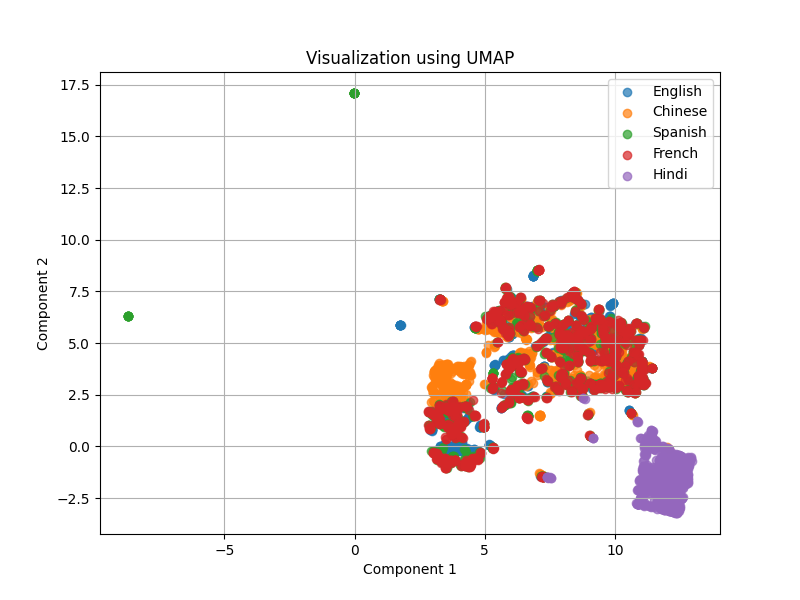}
\caption{UMAP, Layer 5}

\end{subfigure}
\vspace{0.2in} %
\begin{subfigure}{0.18\textwidth}      \includegraphics[width=\textwidth]{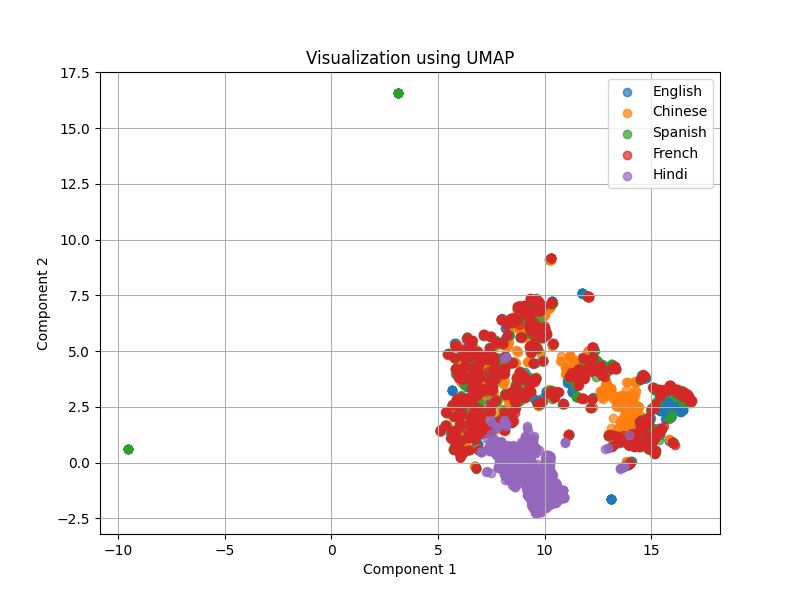}      \caption{UMAP, Layer 6}        \end{subfigure}  \hfill  \begin{subfigure}{0.18\textwidth}      \includegraphics[width=\textwidth]{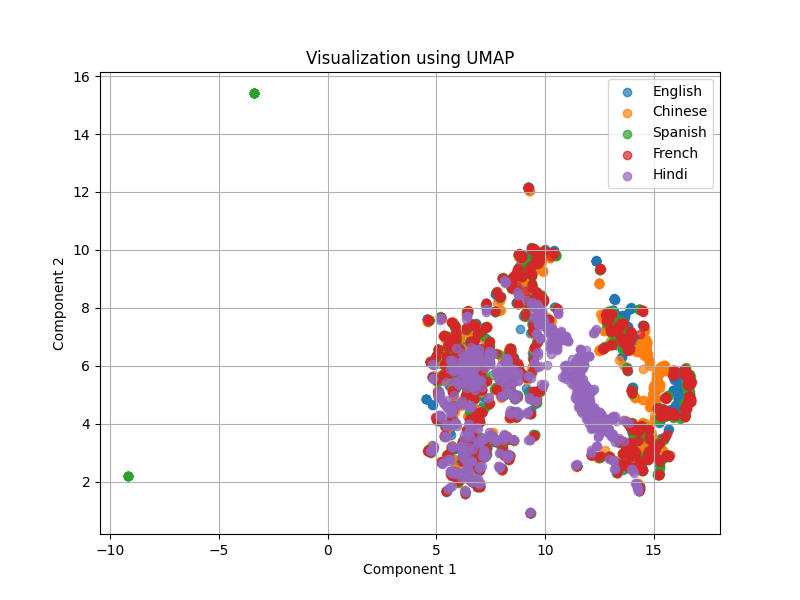}      \caption{UMAP, Layer 7}        \end{subfigure}  \hfill  \begin{subfigure}{0.18\textwidth}      \includegraphics[width=\textwidth]{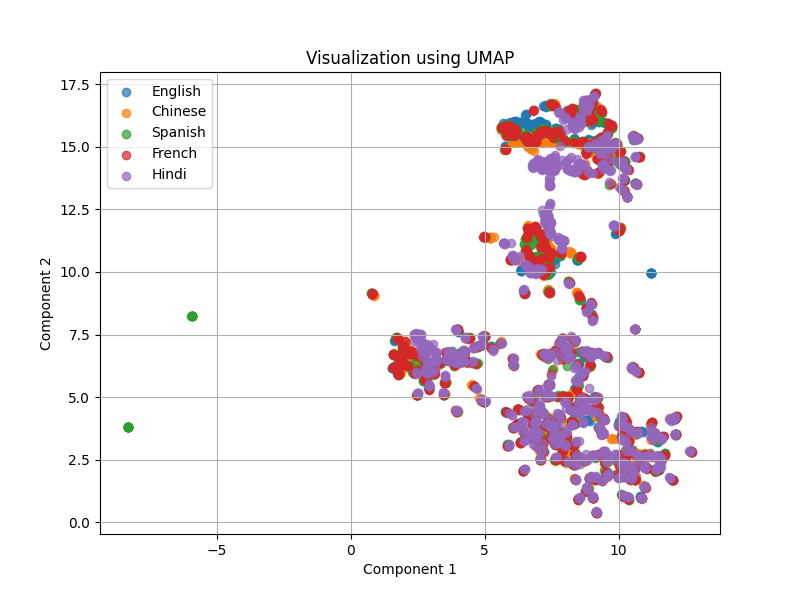}      \caption{UMAP, Layer 8}        \end{subfigure}  \hfill  \begin{subfigure}{0.18\textwidth}      \includegraphics[width=\textwidth]{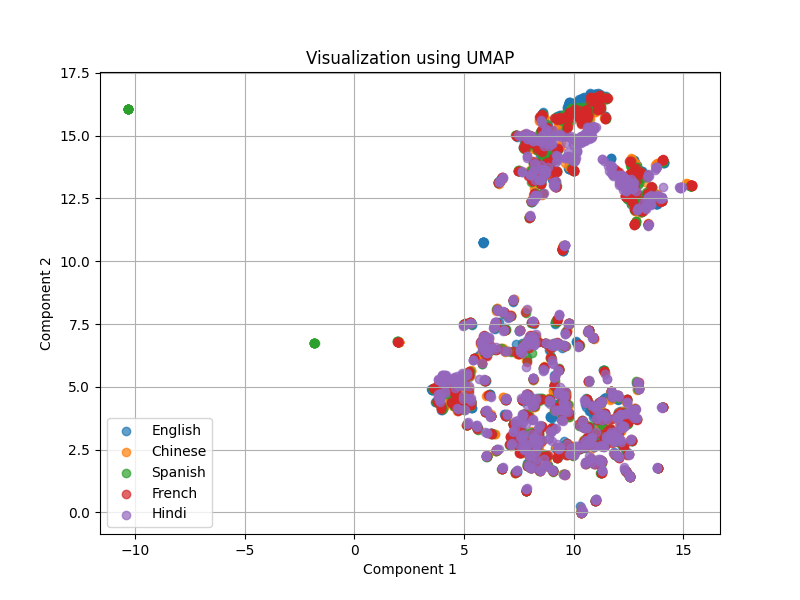}      \caption{UMAP, Layer 9}        \end{subfigure}  \hfill  \begin{subfigure}{0.18\textwidth}      \includegraphics[width=\textwidth]{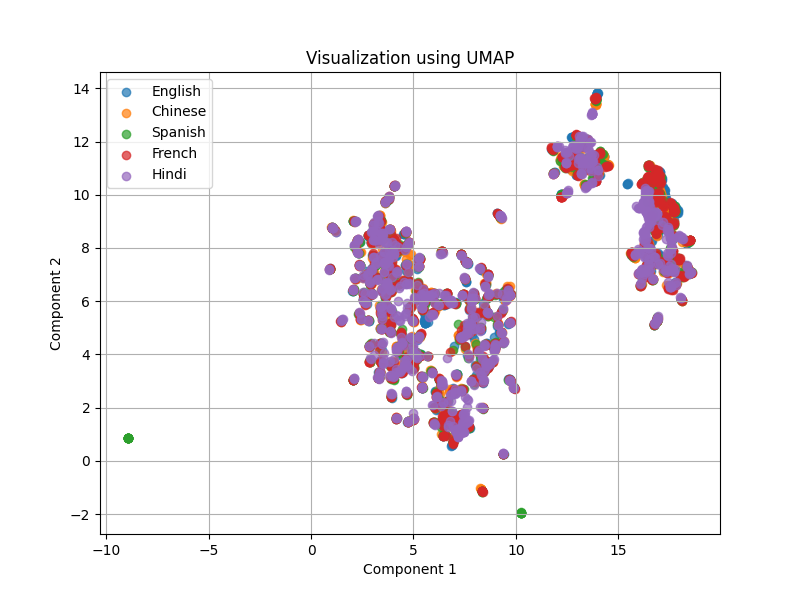}      \caption{UMAP, Layer 10}        \end{subfigure}    \vspace{0.2in}    %
\begin{subfigure}{0.18\textwidth}      \includegraphics[width=\textwidth]{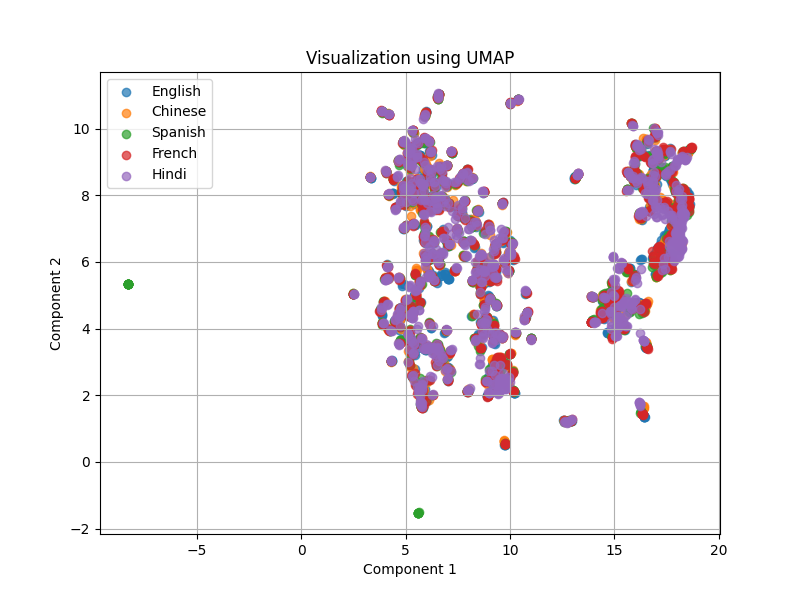}      \caption{UMAP, Layer 11}        \end{subfigure}  \hfill  \begin{subfigure}{0.18\textwidth}      \includegraphics[width=\textwidth]{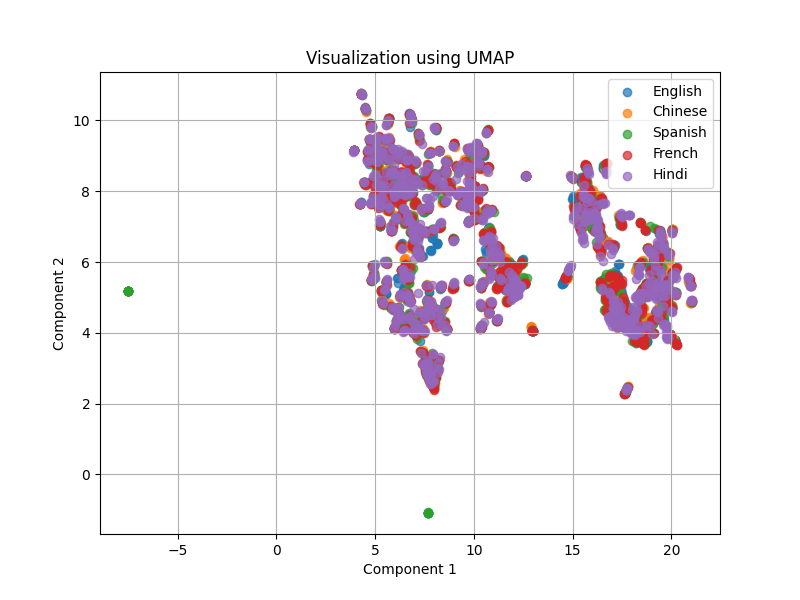}      \caption{UMAP, Layer 12}        \end{subfigure}  \hfill  \begin{subfigure}{0.18\textwidth}      \includegraphics[width=\textwidth]{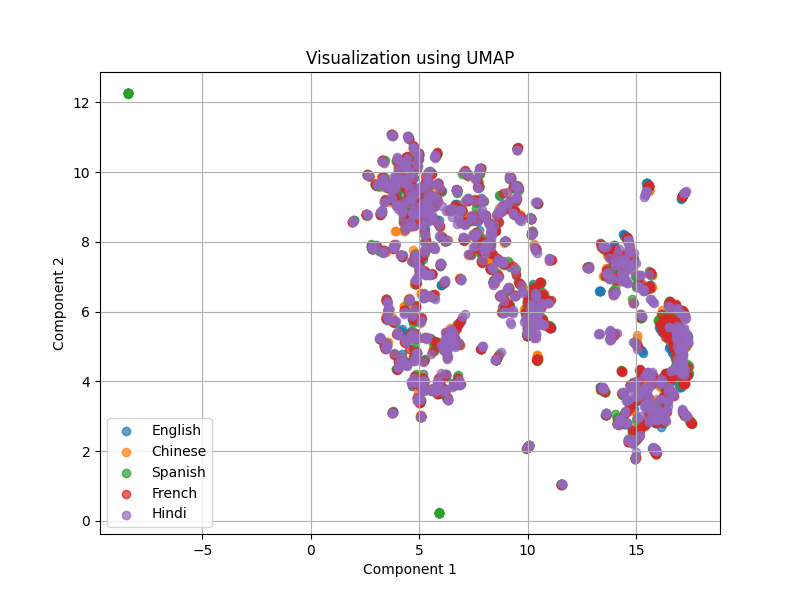}      \caption{UMAP, Layer 13}        \end{subfigure}  \hfill  \begin{subfigure}{0.18\textwidth}      \includegraphics[width=\textwidth]{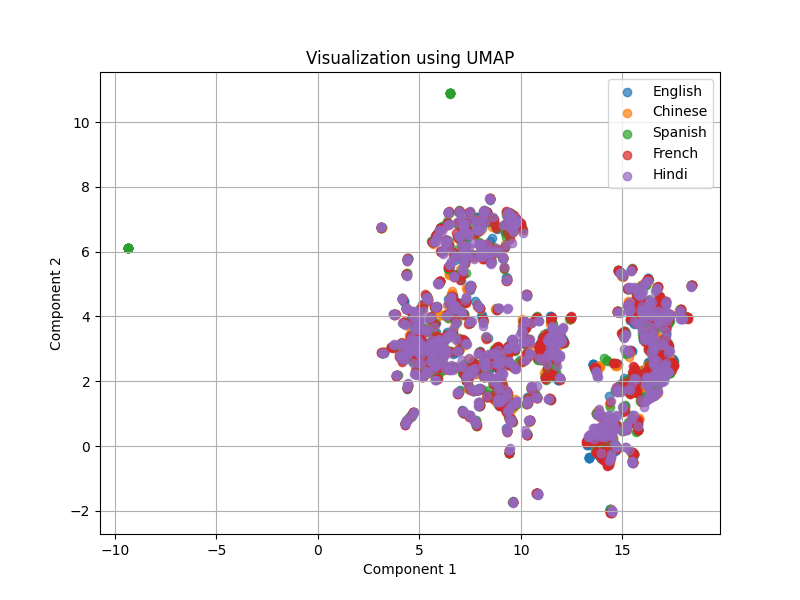}      \caption{UMAP, Layer 14}        \end{subfigure}  \hfill  \begin{subfigure}{0.18\textwidth}      \includegraphics[width=\textwidth]{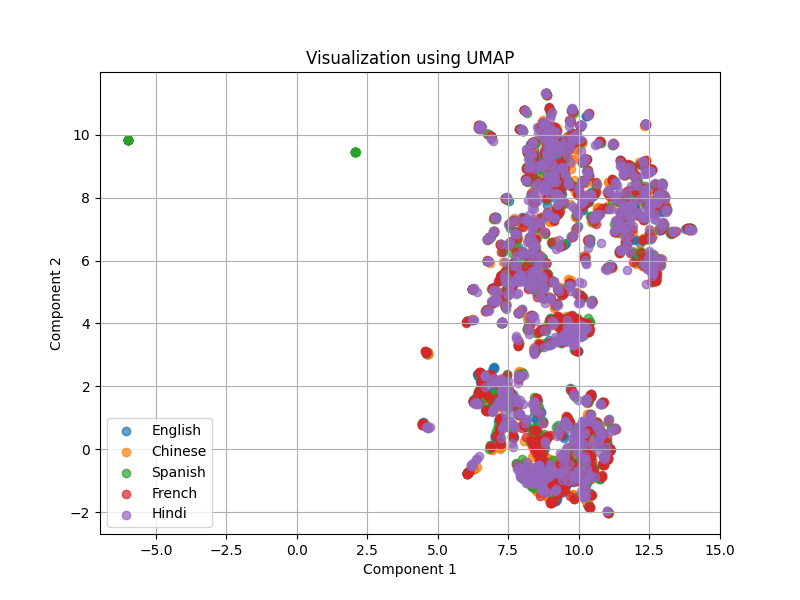}      \caption{UMAP, Layer 15}        \end{subfigure}    \vspace{0.2in}    %
\begin{subfigure}{0.18\textwidth}      \includegraphics[width=\textwidth]{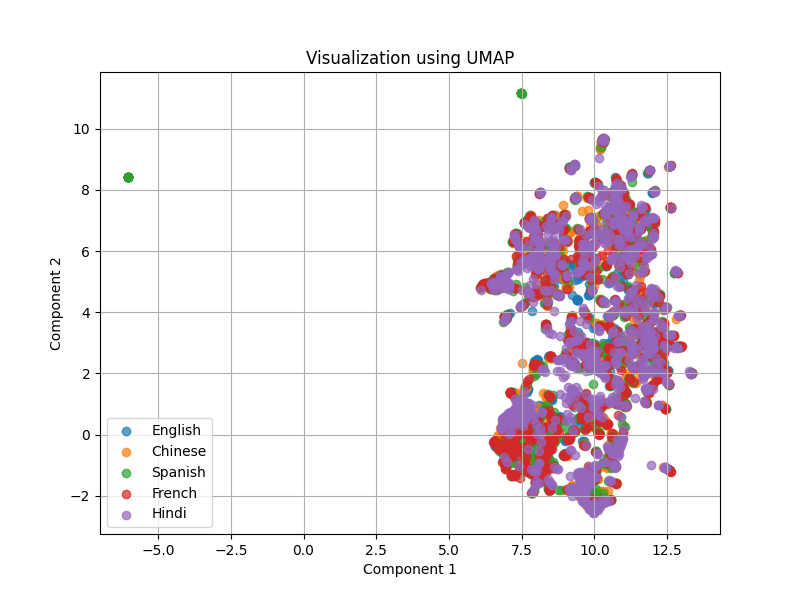}      \caption{UMAP, Layer 16}        \end{subfigure}  \hfill  \begin{subfigure}{0.18\textwidth}      \includegraphics[width=\textwidth]{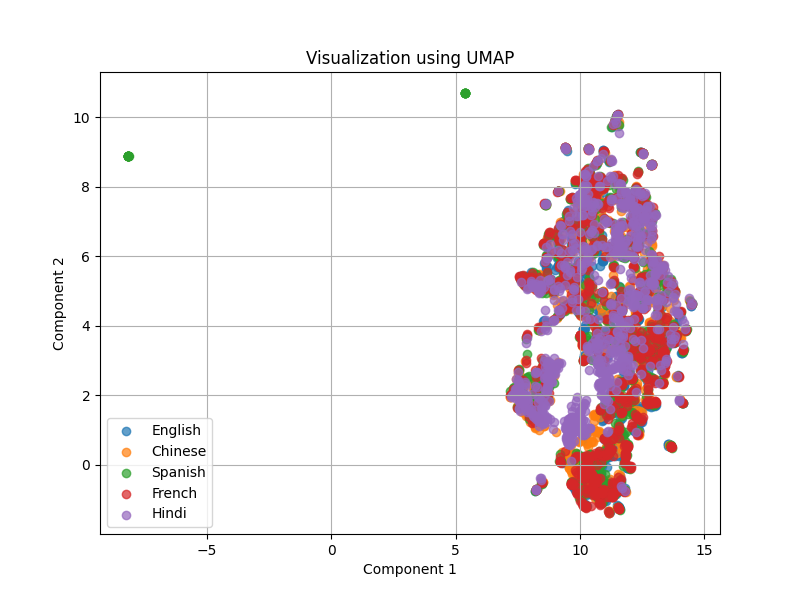}      \caption{UMAP, Layer 17}        \end{subfigure}  \hfill  \begin{subfigure}{0.18\textwidth}      \includegraphics[width=\textwidth]{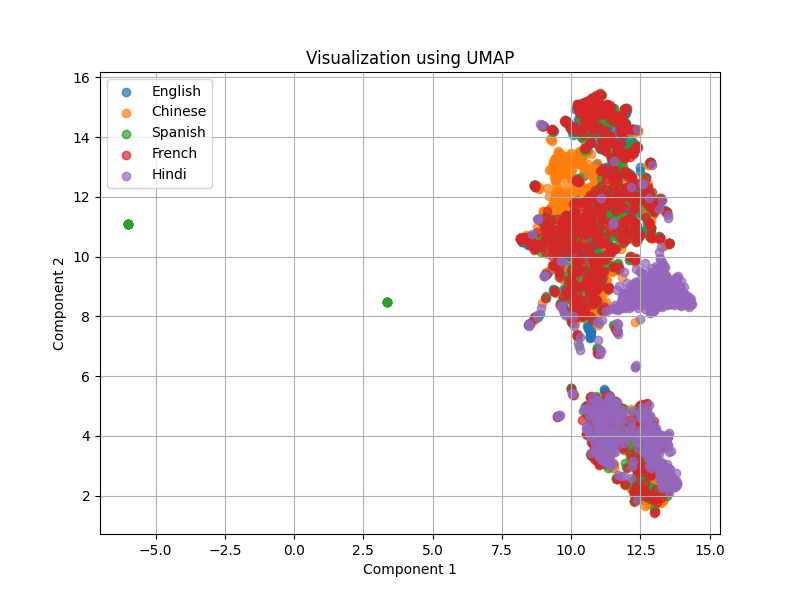}      \caption{UMAP, Layer 18}        \end{subfigure}  \hfill  \begin{subfigure}{0.18\textwidth}      \includegraphics[width=\textwidth]{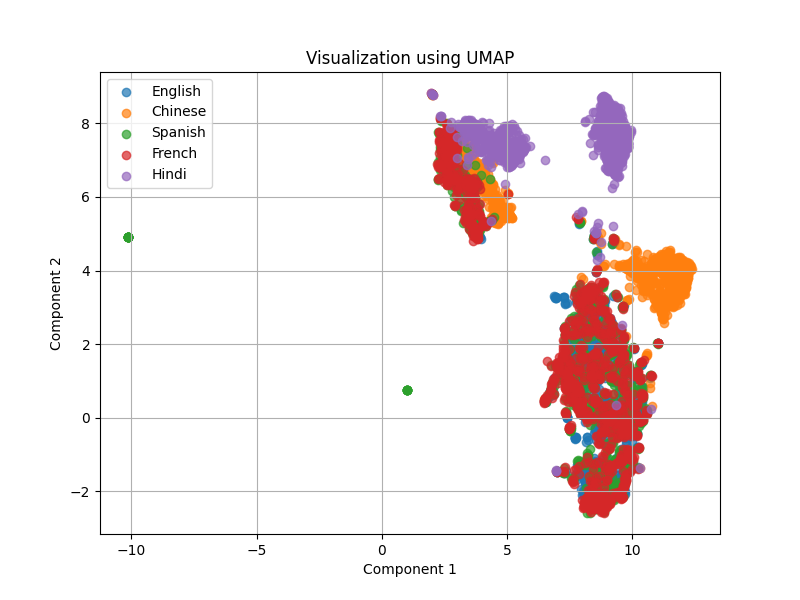}      \caption{UMAP, Layer 19}        \end{subfigure}  \hfill  \begin{subfigure}{0.18\textwidth}      \includegraphics[width=\textwidth]{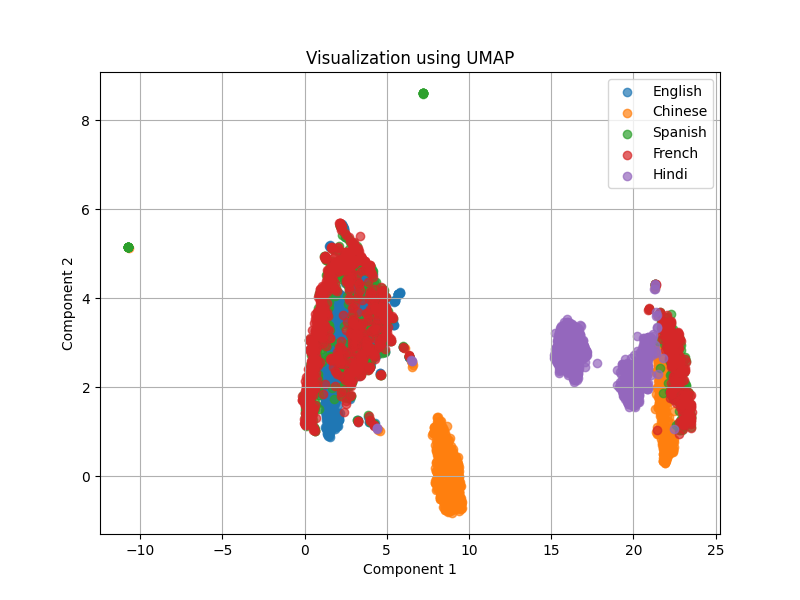}      \caption{UMAP, Layer 20}        \end{subfigure}    \vspace{0.2in}    %
\begin{subfigure}{0.18\textwidth}      \includegraphics[width=\textwidth]{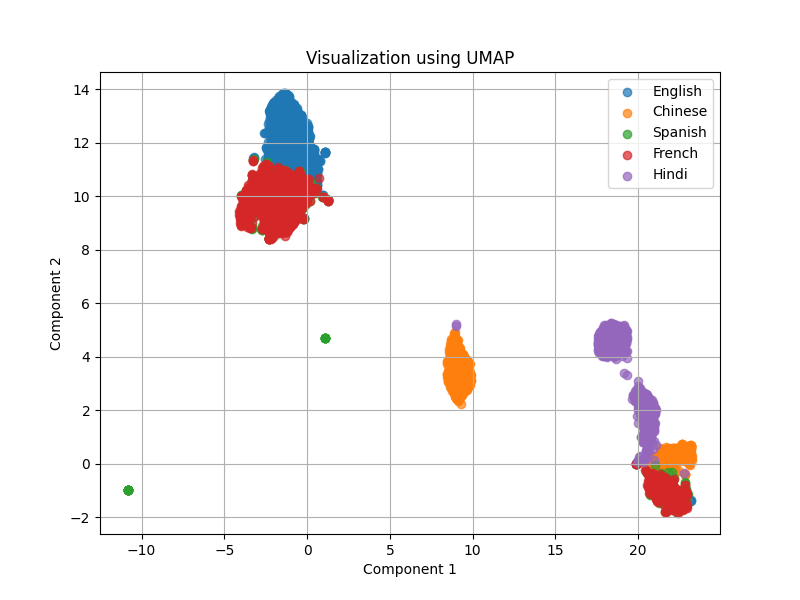}      \caption{UMAP, Layer 21}        \end{subfigure}  \hfill  \begin{subfigure}{0.18\textwidth}      \includegraphics[width=\textwidth]{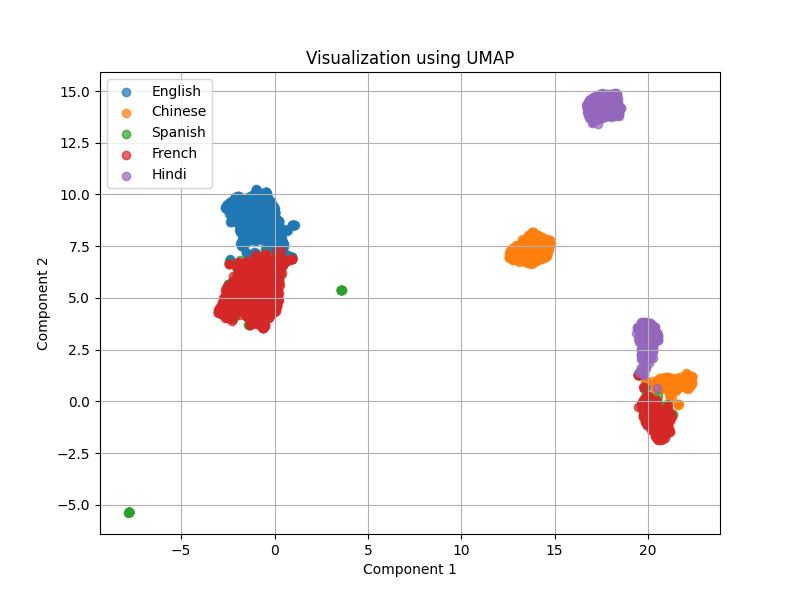}      \caption{UMAP, Layer 22}        \end{subfigure}  \hfill  \begin{subfigure}{0.18\textwidth}      \includegraphics[width=\textwidth]{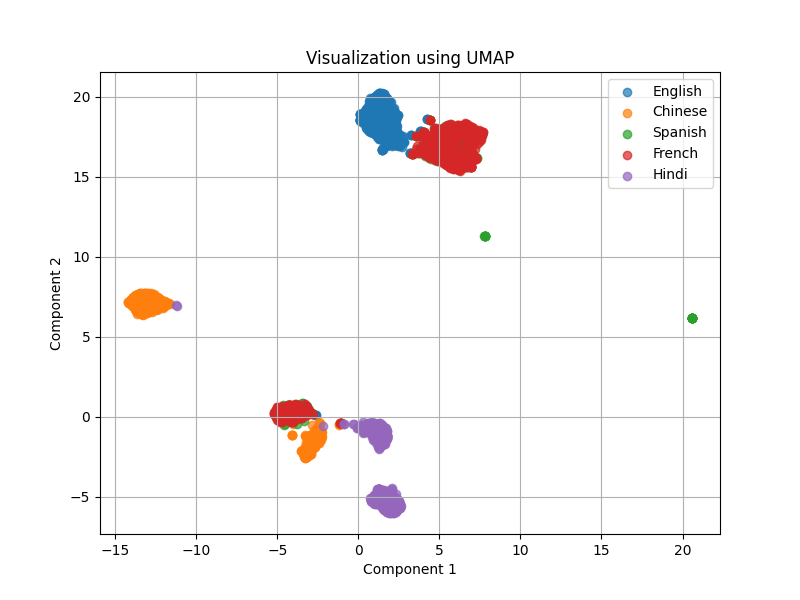}      \caption{UMAP, Layer 23}        \end{subfigure}  \hfill  \begin{subfigure}{0.18\textwidth}      \includegraphics[width=\textwidth]{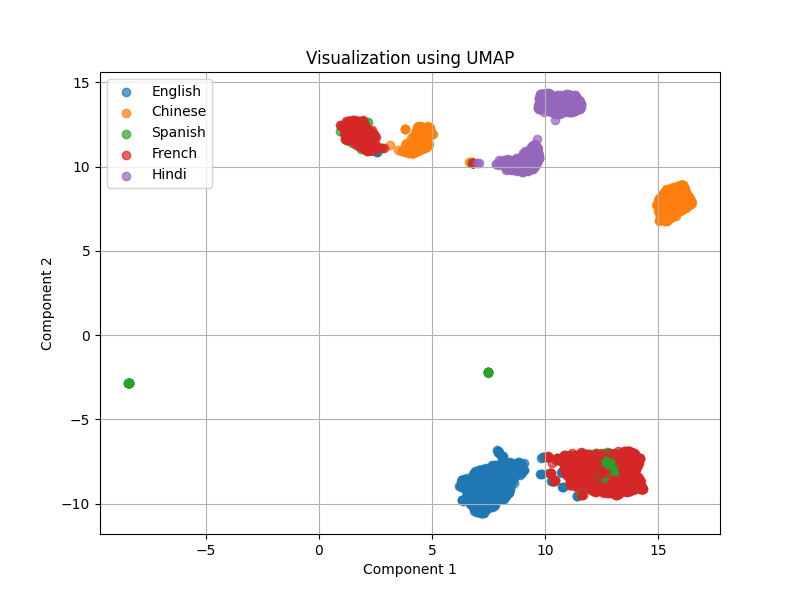}      \caption{UMAP, Layer 24}        \end{subfigure}  \hfill  \begin{subfigure}{0.18\textwidth}      \includegraphics[width=\textwidth]{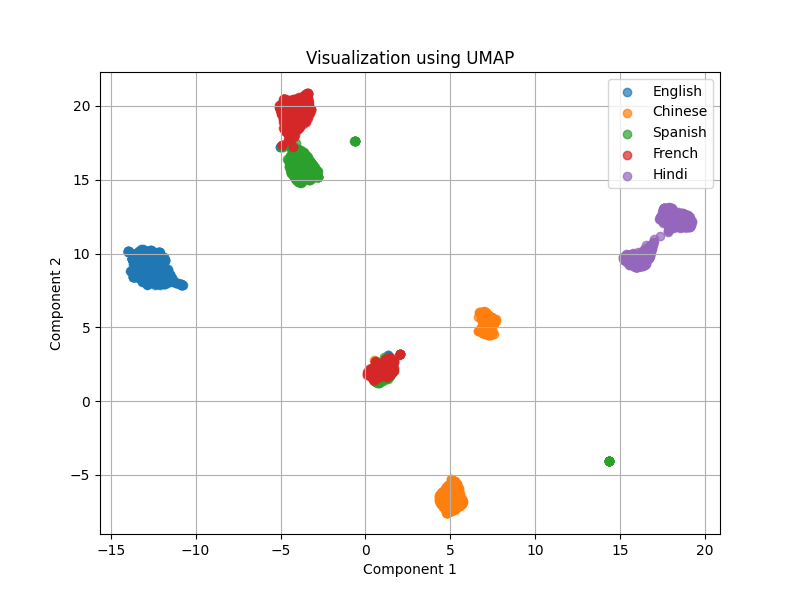}      \caption{UMAP, Layer 25}        \end{subfigure}    \vspace{0.2in}    %
\begin{subfigure}{0.18\textwidth}      \includegraphics[width=\textwidth]{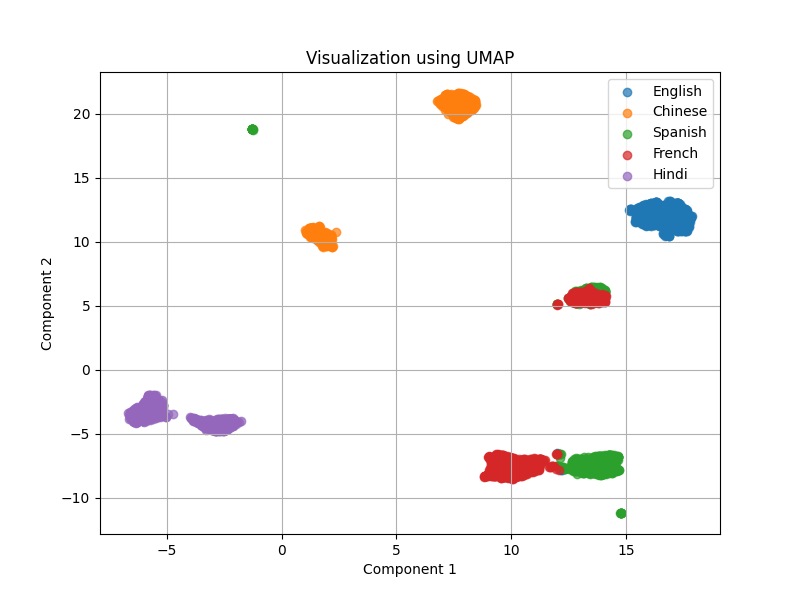}      \caption{UMAP, Layer 26}        \end{subfigure}  \hfill  \begin{subfigure}{0.18\textwidth}      \includegraphics[width=\textwidth]{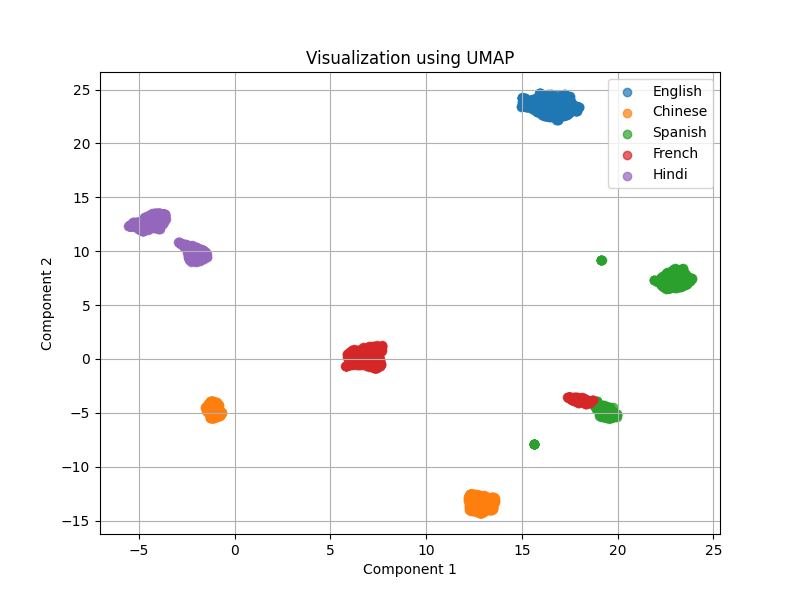}      \caption{UMAP, Layer 27}        \end{subfigure}  \hfill  \begin{subfigure}{0.18\textwidth}      \includegraphics[width=\textwidth]{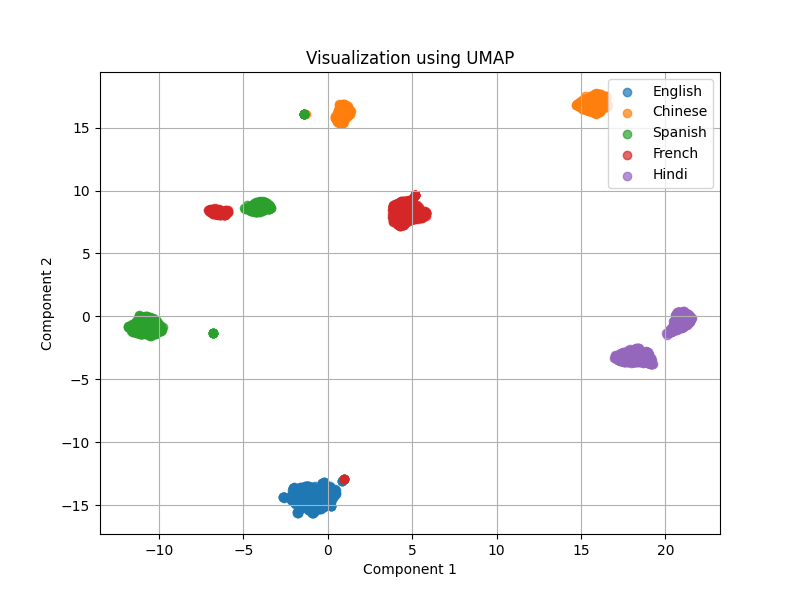}      \caption{UMAP, Layer 28}        \end{subfigure}  \hfill  \begin{subfigure}{0.18\textwidth}      \includegraphics[width=\textwidth]{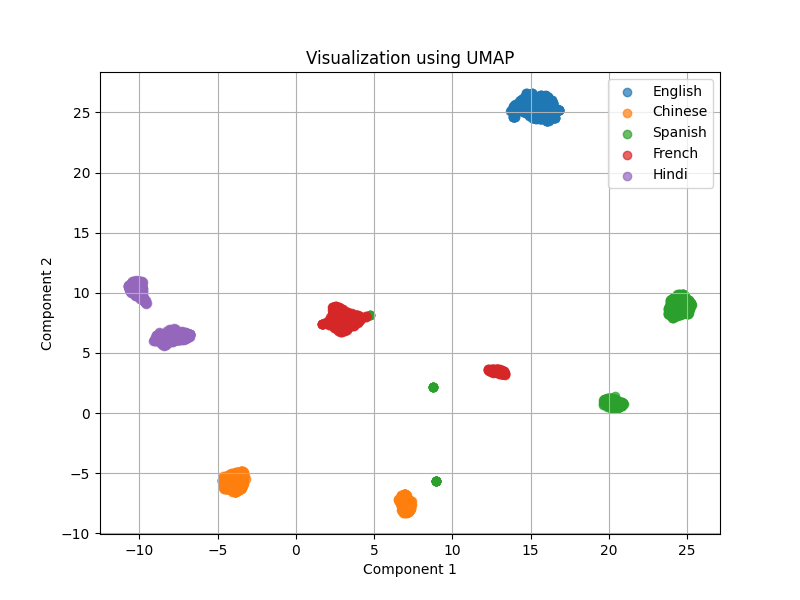}      \caption{UMAP, Layer 29}        \end{subfigure}  \hfill  \begin{subfigure}{0.18\textwidth}      \includegraphics[width=\textwidth]{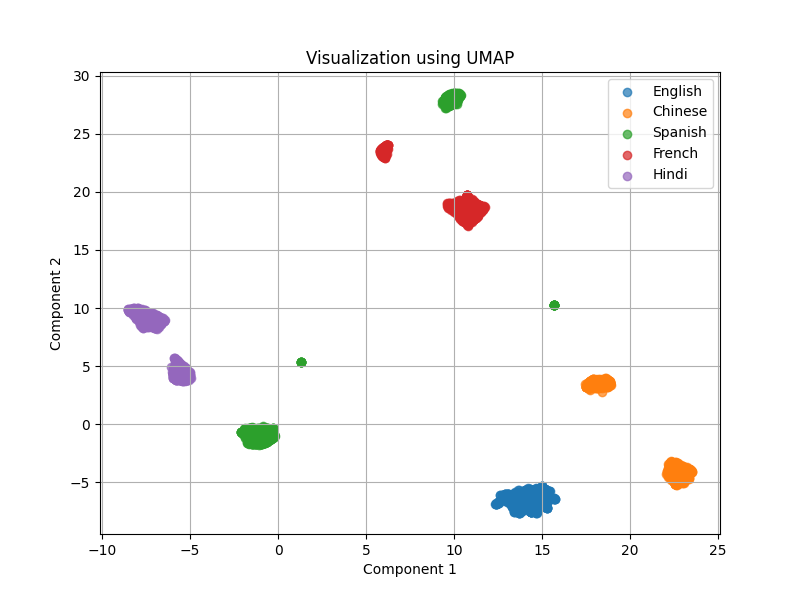}      \caption{UMAP, Layer 30}        \end{subfigure}    \vspace{0.2in}    %
\begin{subfigure}{0.18\textwidth}      \includegraphics[width=\textwidth]{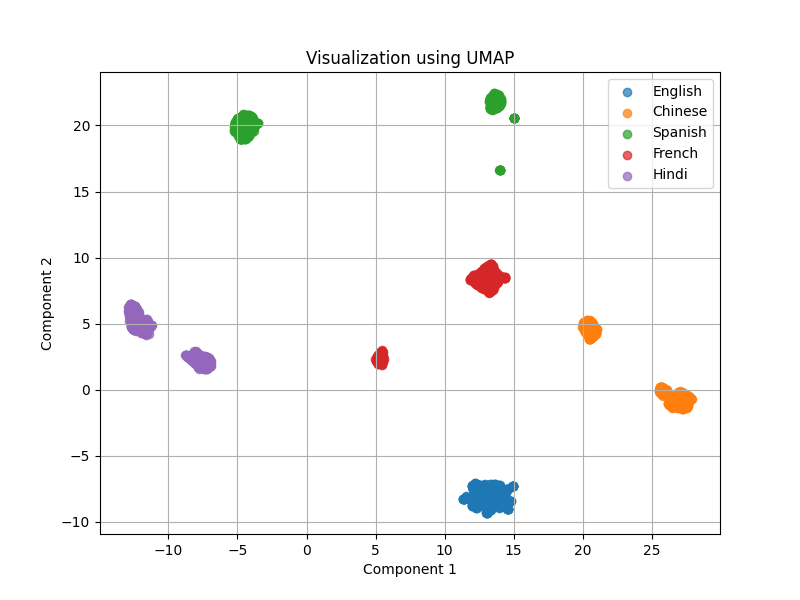}      \caption{UMAP, Layer 31}        \end{subfigure}  \hfill  \begin{subfigure}{0.18\textwidth}      \includegraphics[width=\textwidth]{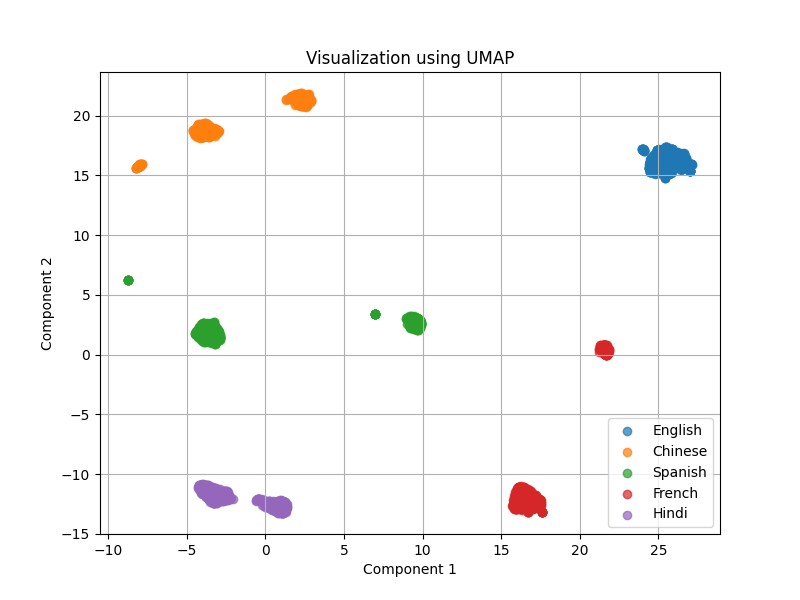}      \caption{UMAP, Layer 32}        \end{subfigure}    \caption{UMAP visualizations for layers 1-32 of Llama-3-8B-Instruct on the FOLIO dataset.}  
\end{figure*}

\begin{figure*}[htbp]
\centering
\begin{subfigure}{0.18\textwidth}
\includegraphics[width=\textwidth]{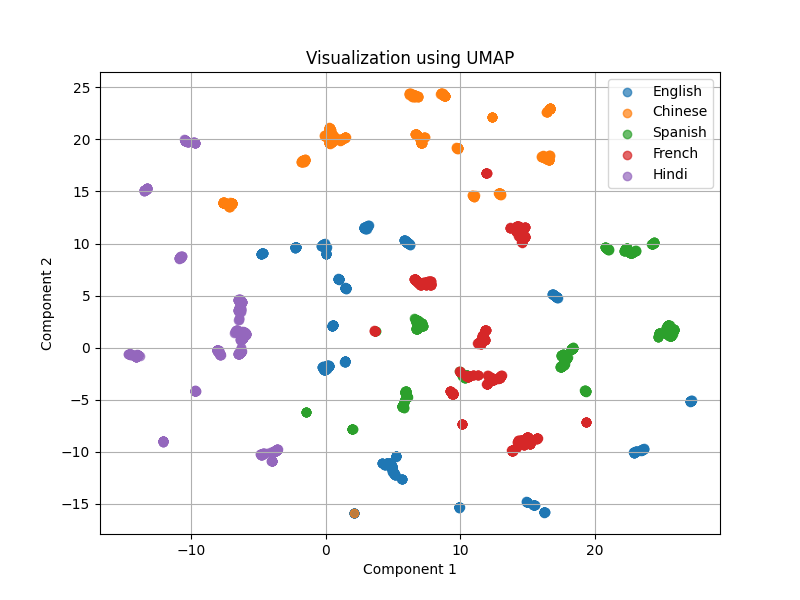}
\caption{UMAP, Layer 1}
\end{subfigure}
\hfill
\begin{subfigure}{0.18\textwidth}
\includegraphics[width=\textwidth]{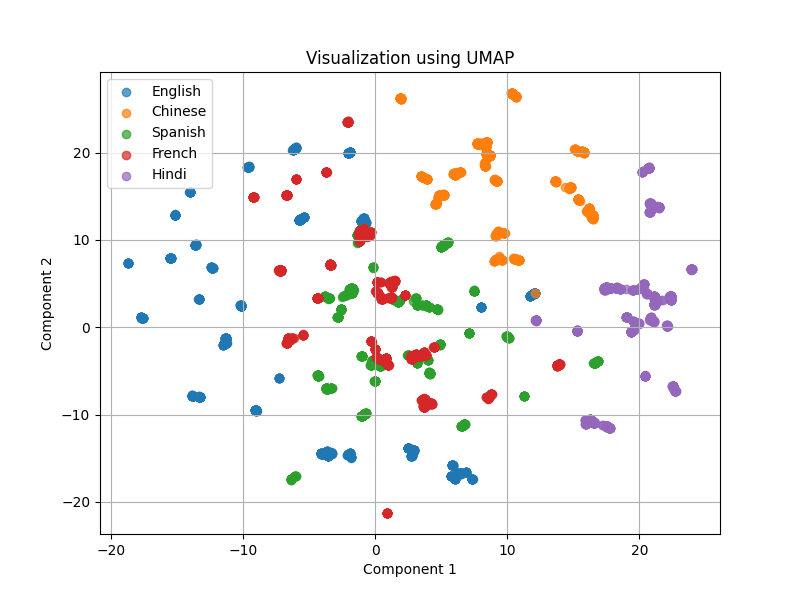}
\caption{UMAP, Layer 2}

\end{subfigure}
\hfill
\begin{subfigure}{0.18\textwidth}
\includegraphics[width=\textwidth]{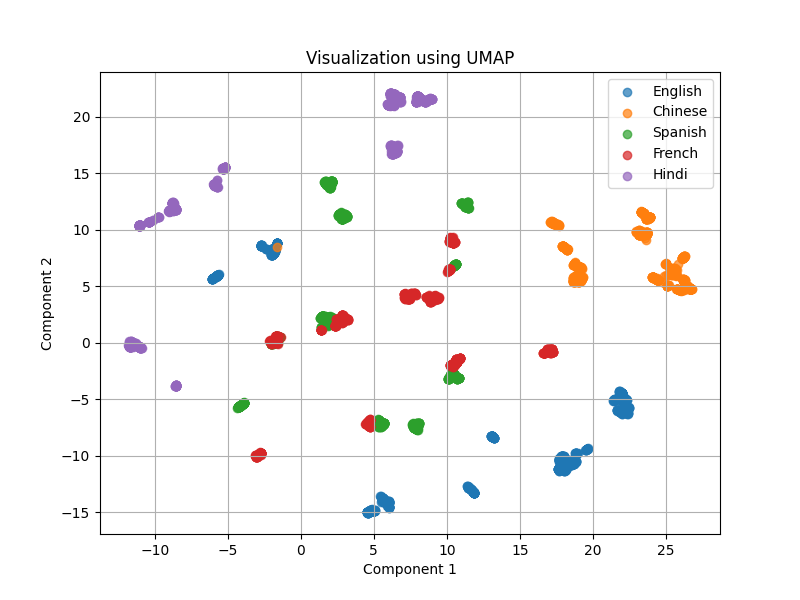}
\caption{UMAP, Layer 3}

\end{subfigure}
\hfill
\begin{subfigure}{0.18\textwidth}
\includegraphics[width=\textwidth]{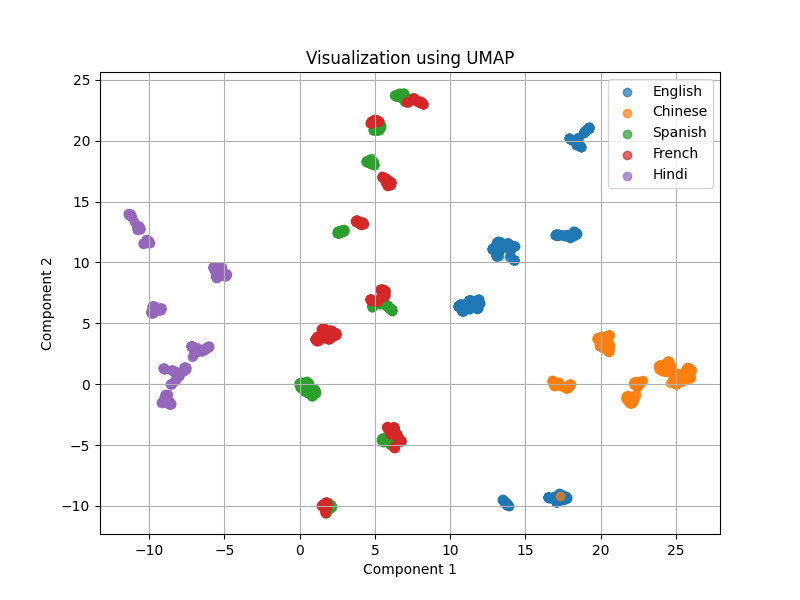}
\caption{UMAP, Layer 4}

\end{subfigure}
\hfill
\begin{subfigure}{0.18\textwidth}
\includegraphics[width=\textwidth]{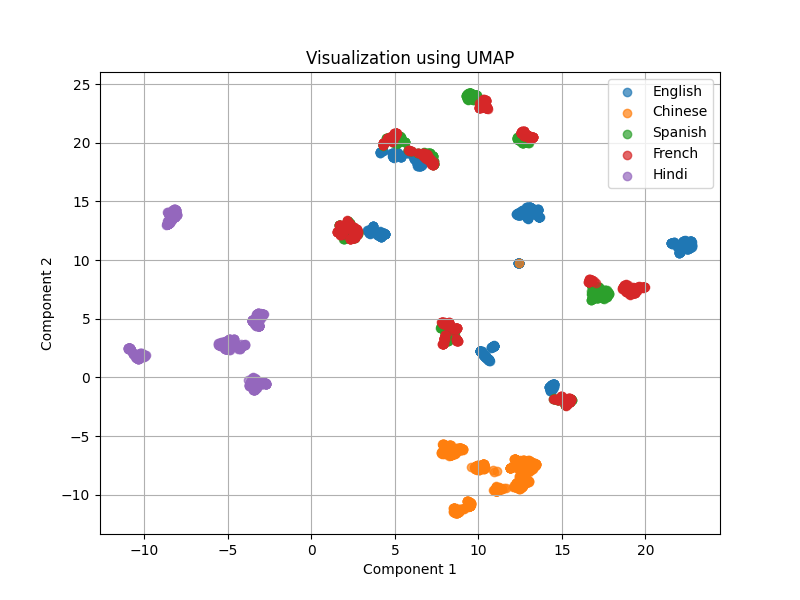}
\caption{UMAP, Layer 5}

\end{subfigure}
\vspace{0.2in} %
\begin{subfigure}{0.18\textwidth}      \includegraphics[width=\textwidth]{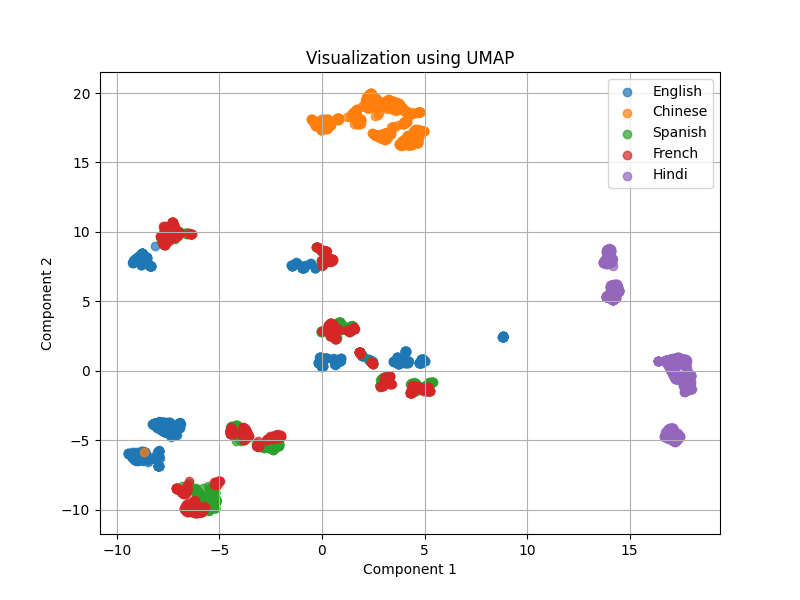}      \caption{UMAP, Layer 6}        \end{subfigure}  \hfill  \begin{subfigure}{0.18\textwidth}      \includegraphics[width=\textwidth]{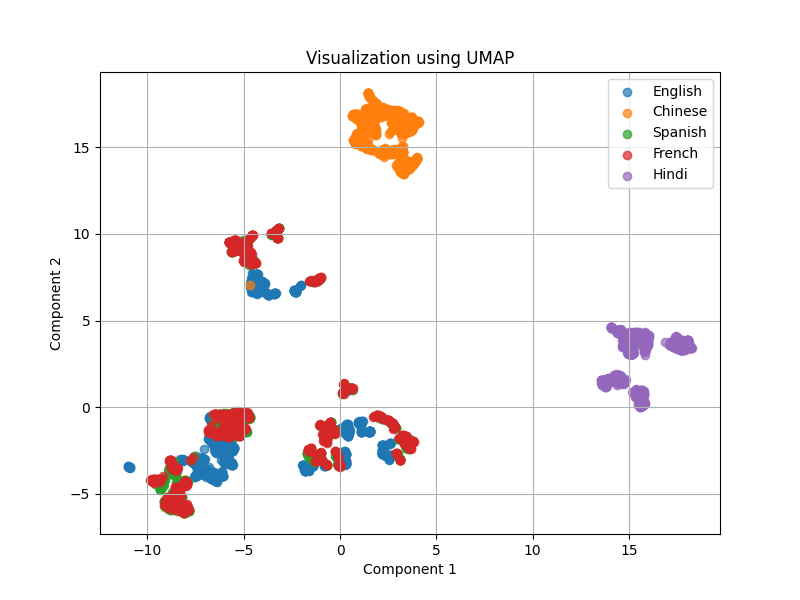}      \caption{UMAP, Layer 7}        \end{subfigure}  \hfill  \begin{subfigure}{0.18\textwidth}      \includegraphics[width=\textwidth]{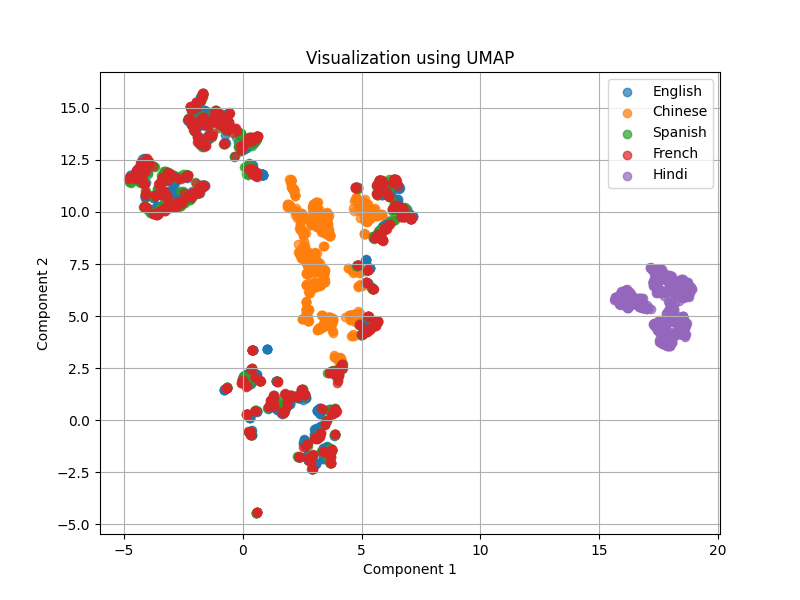}      \caption{UMAP, Layer 8}        \end{subfigure}  \hfill  \begin{subfigure}{0.18\textwidth}      \includegraphics[width=\textwidth]{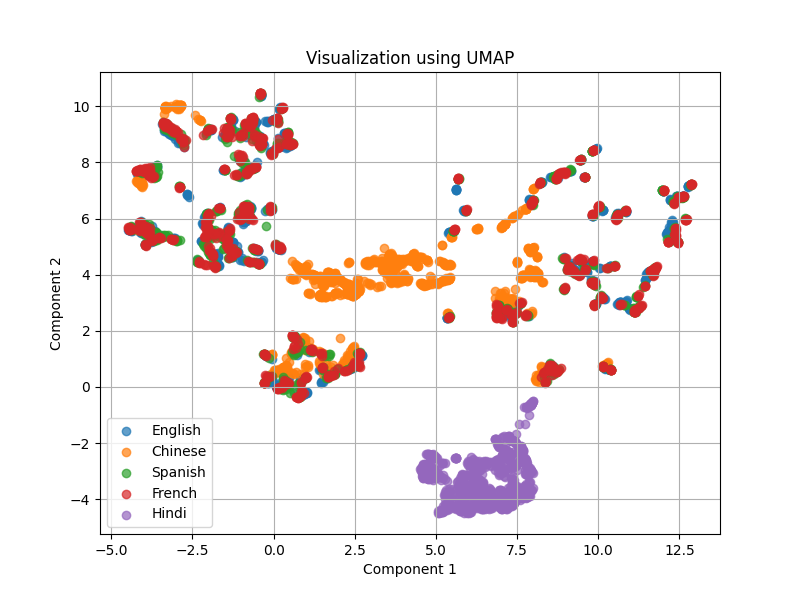}      \caption{UMAP, Layer 9}        \end{subfigure}  \hfill  \begin{subfigure}{0.18\textwidth}      \includegraphics[width=\textwidth]{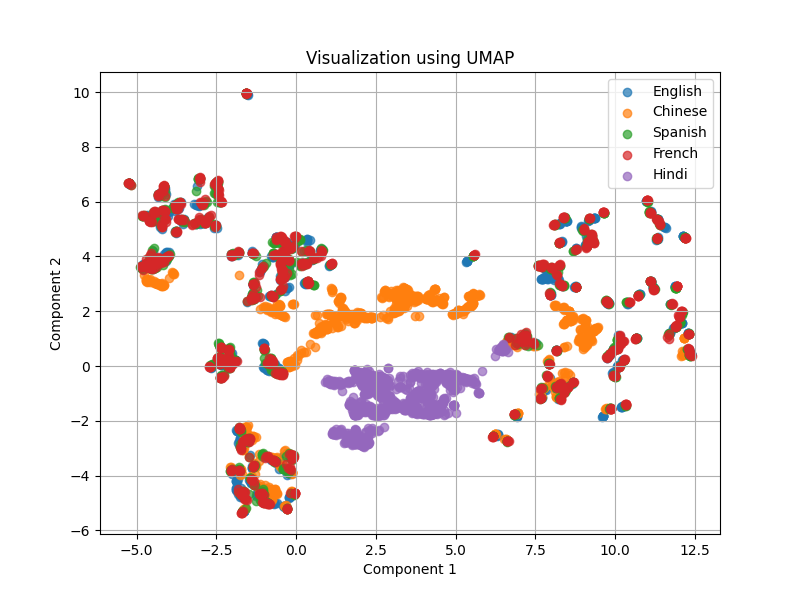}      \caption{UMAP, Layer 10}        \end{subfigure}    \vspace{0.2in}    %
\begin{subfigure}{0.18\textwidth}      \includegraphics[width=\textwidth]{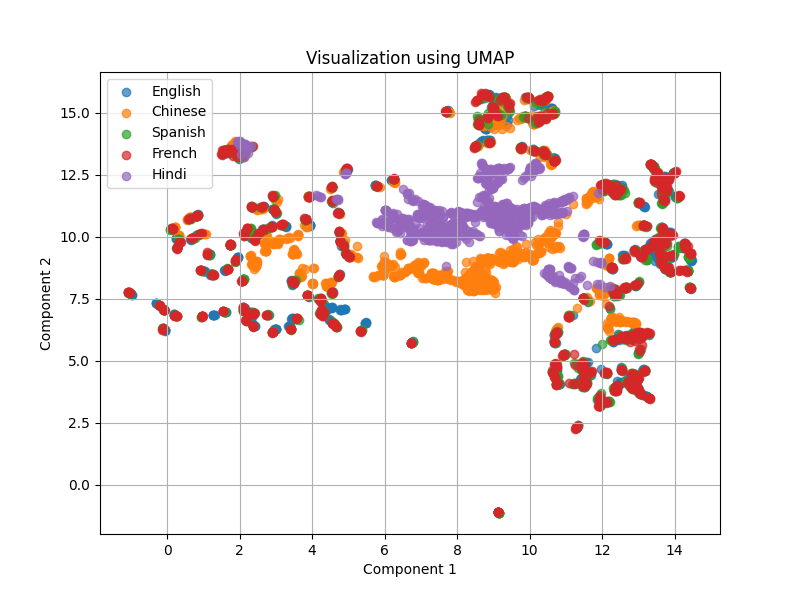}      \caption{UMAP, Layer 11}        \end{subfigure}  \hfill  \begin{subfigure}{0.18\textwidth}      \includegraphics[width=\textwidth]{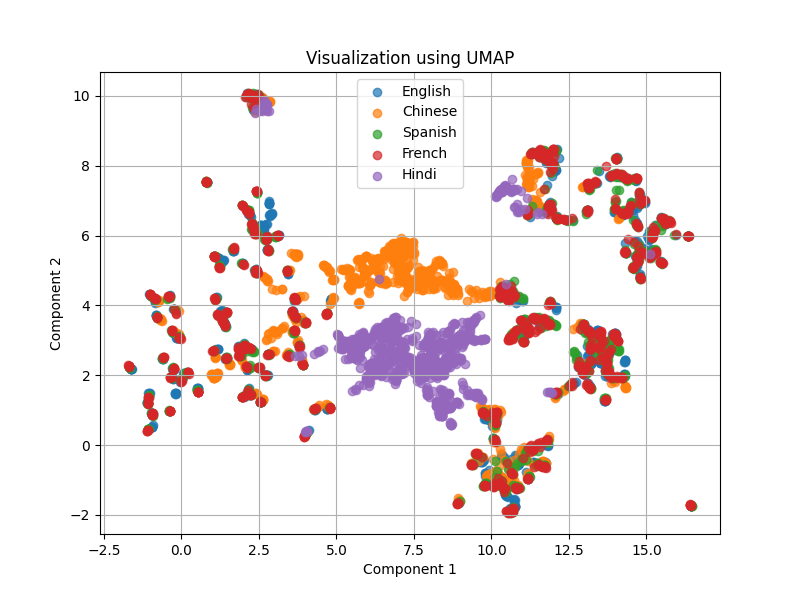}      \caption{UMAP, Layer 12}        \end{subfigure}  \hfill  \begin{subfigure}{0.18\textwidth}      \includegraphics[width=\textwidth]{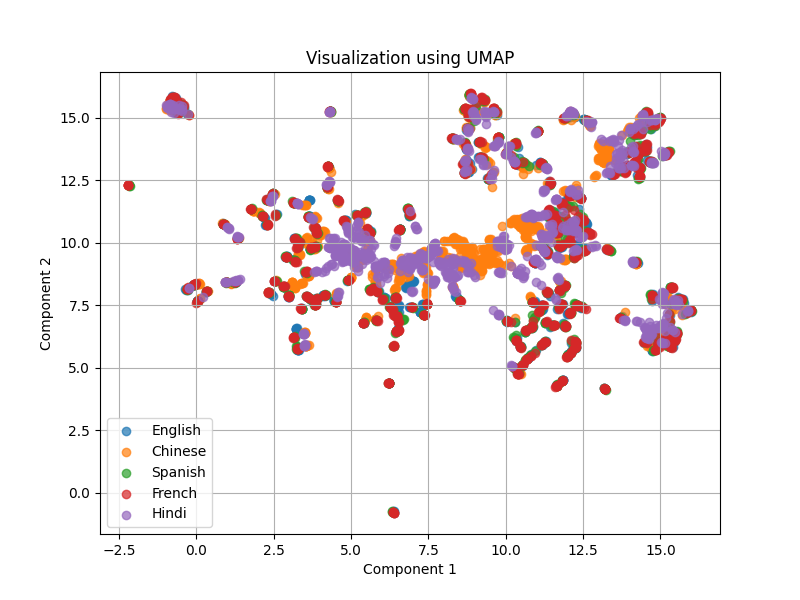}      \caption{UMAP, Layer 13}        \end{subfigure}  \hfill  \begin{subfigure}{0.18\textwidth}      \includegraphics[width=\textwidth]{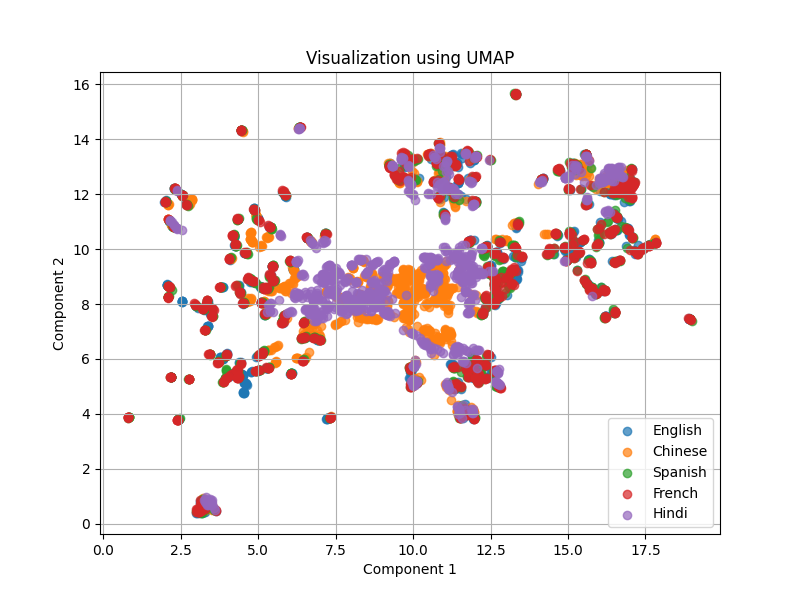}      \caption{UMAP, Layer 14}        \end{subfigure}  \hfill  \begin{subfigure}{0.18\textwidth}      \includegraphics[width=\textwidth]{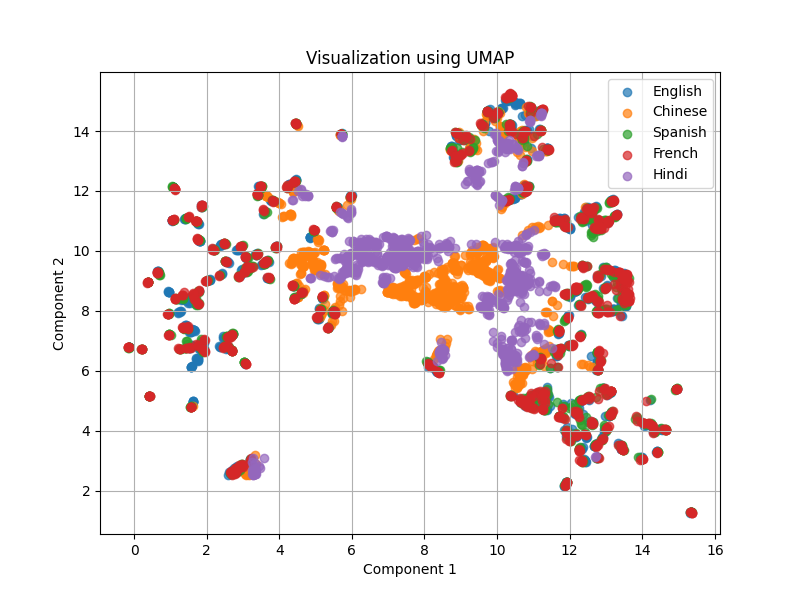}      \caption{UMAP, Layer 15}        \end{subfigure}    \vspace{0.2in}    %
\begin{subfigure}{0.18\textwidth}      \includegraphics[width=\textwidth]{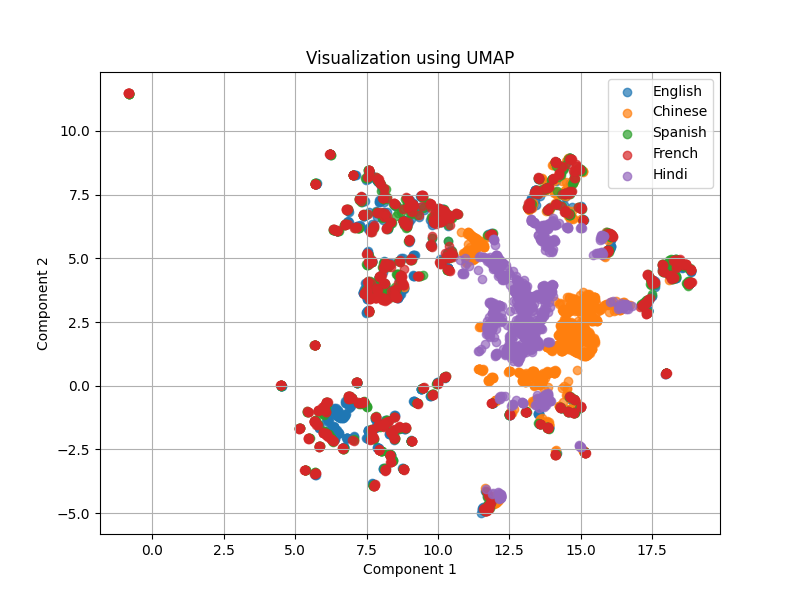}      \caption{UMAP, Layer 16}        \end{subfigure}  \hfill  \begin{subfigure}{0.18\textwidth}      \includegraphics[width=\textwidth]{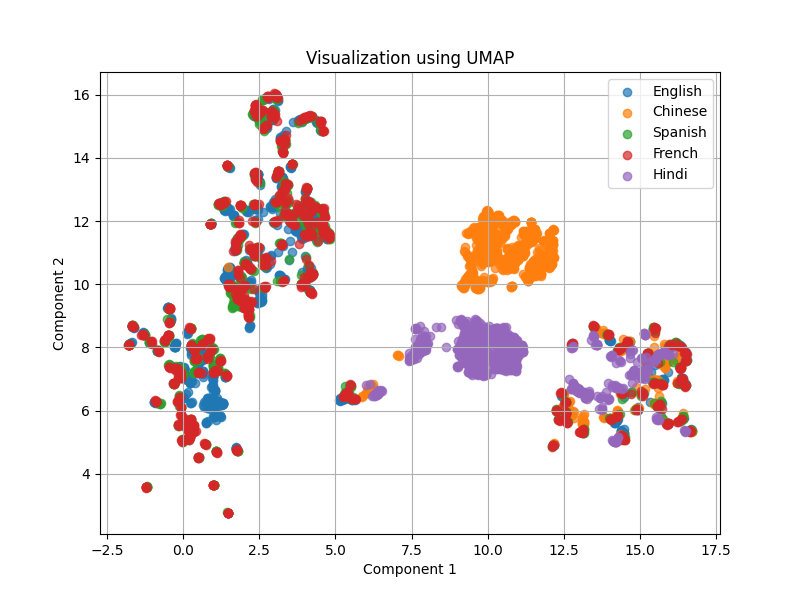}      \caption{UMAP, Layer 17}        \end{subfigure}  \hfill  \begin{subfigure}{0.18\textwidth}      \includegraphics[width=\textwidth]{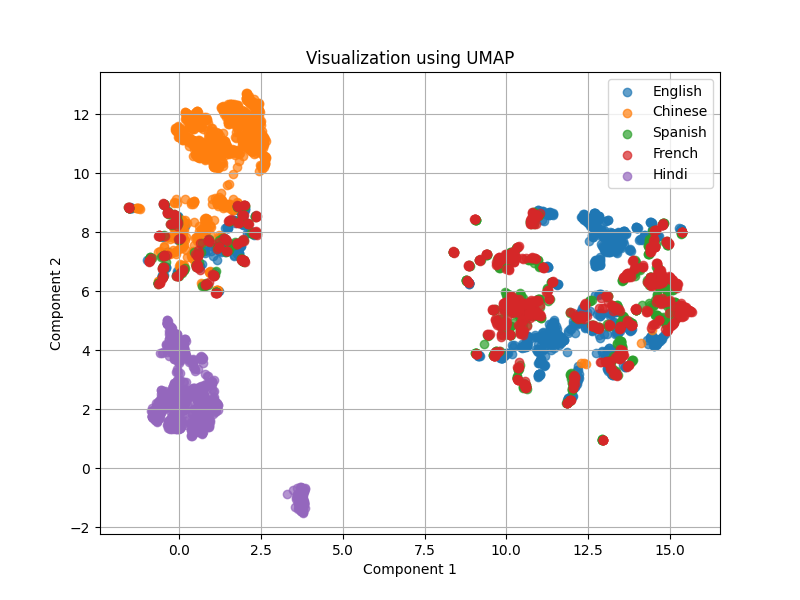}      \caption{UMAP, Layer 18}        \end{subfigure}  \hfill  \begin{subfigure}{0.18\textwidth}      \includegraphics[width=\textwidth]{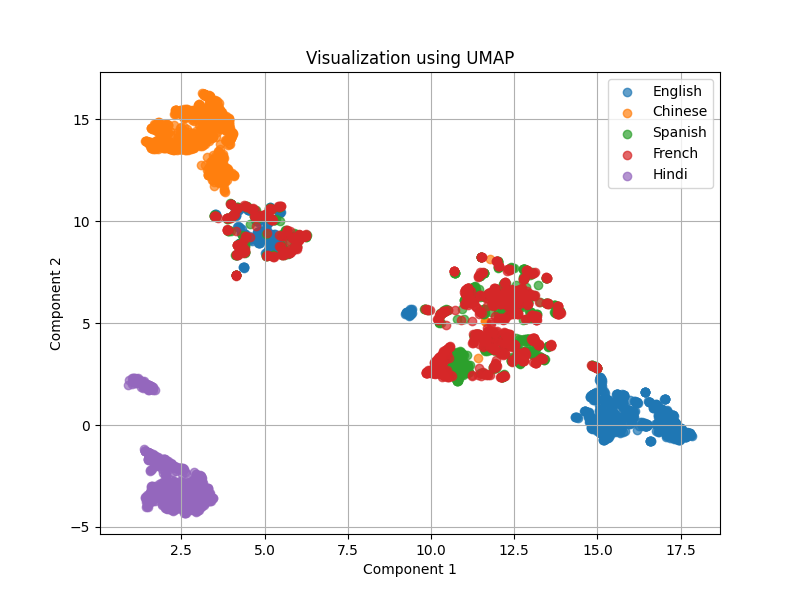}      \caption{UMAP, Layer 19}        \end{subfigure}  \hfill  \begin{subfigure}{0.18\textwidth}      \includegraphics[width=\textwidth]{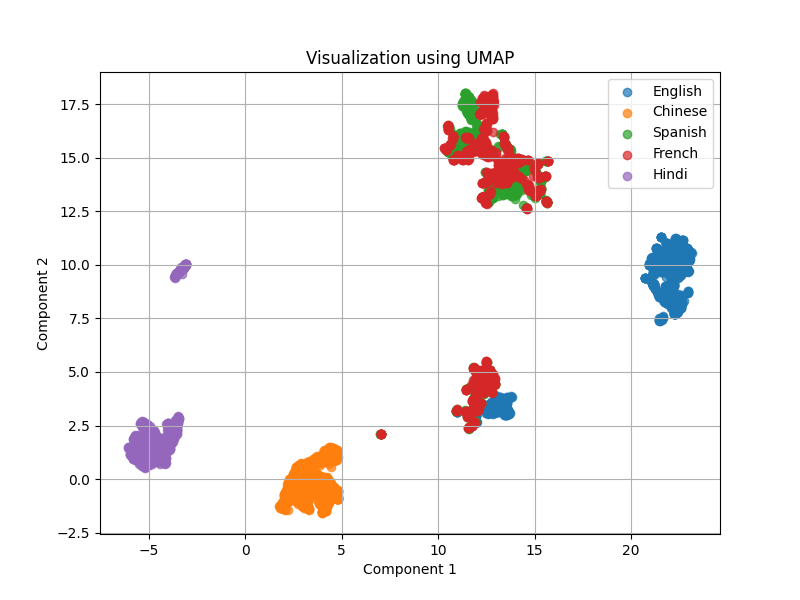}      \caption{UMAP, Layer 20}        \end{subfigure}    \vspace{0.2in}    %
\begin{subfigure}{0.18\textwidth}      \includegraphics[width=\textwidth]{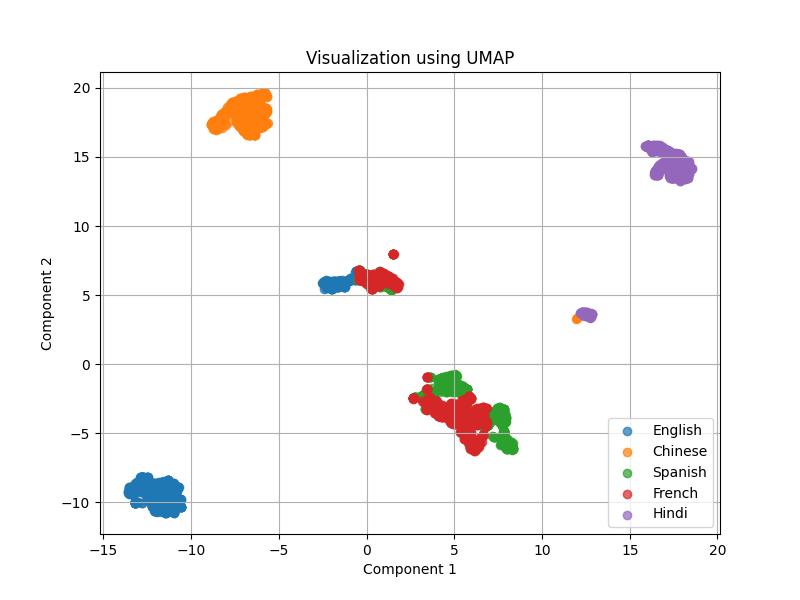}      \caption{UMAP, Layer 21}        \end{subfigure}  \hfill  \begin{subfigure}{0.18\textwidth}      \includegraphics[width=\textwidth]{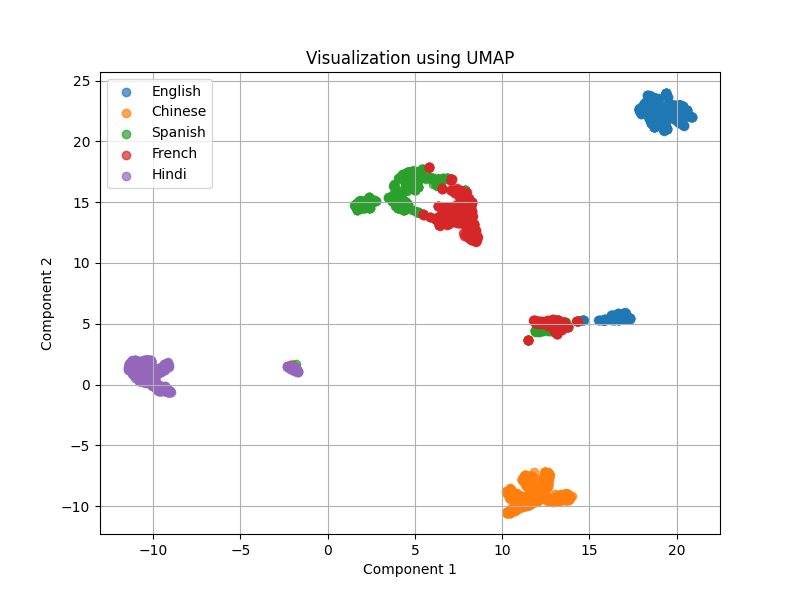}      \caption{UMAP, Layer 22}        \end{subfigure}  \hfill  \begin{subfigure}{0.18\textwidth}      \includegraphics[width=\textwidth]{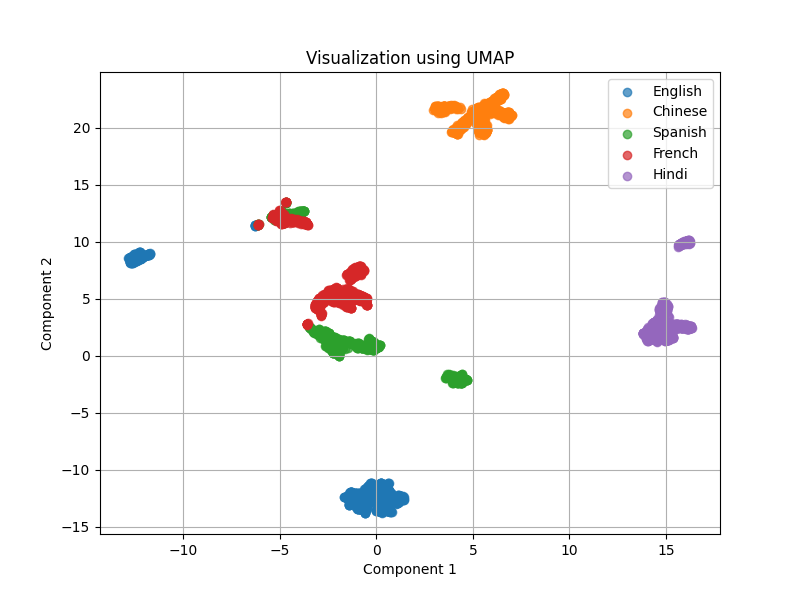}      \caption{UMAP, Layer 23}        \end{subfigure}  \hfill  \begin{subfigure}{0.18\textwidth}      \includegraphics[width=\textwidth]{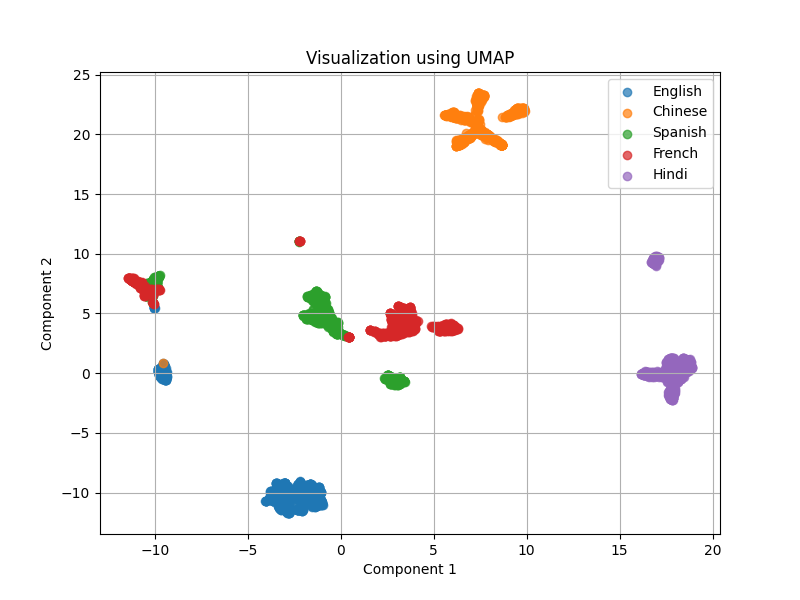}      \caption{UMAP, Layer 24}        \end{subfigure}  \hfill  \begin{subfigure}{0.18\textwidth}      \includegraphics[width=\textwidth]{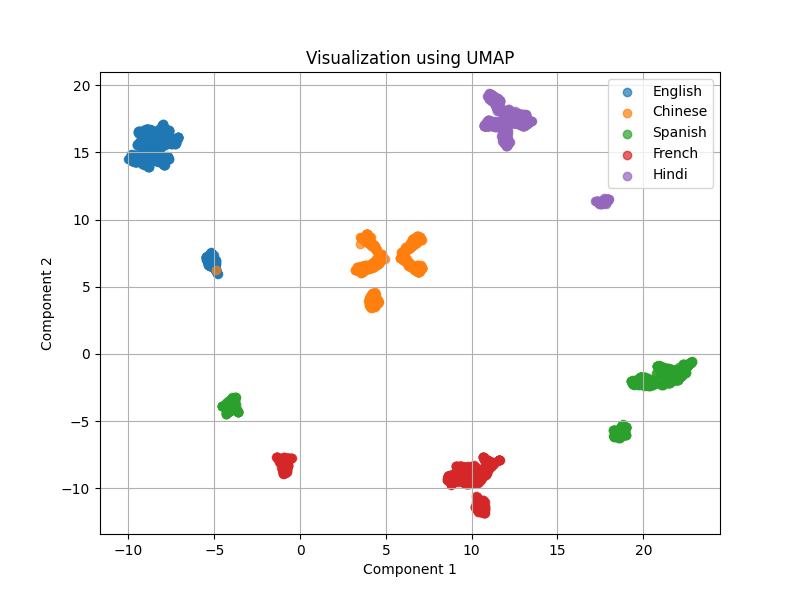}      \caption{UMAP, Layer 25}        \end{subfigure}    \vspace{0.2in}    %
\begin{subfigure}{0.18\textwidth}      \includegraphics[width=\textwidth]{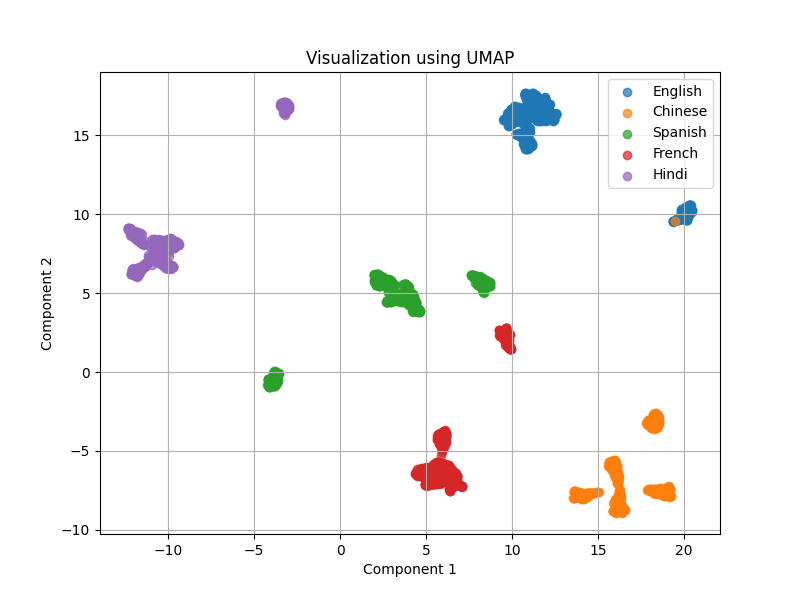}      \caption{UMAP, Layer 26}        \end{subfigure}  \hfill  \begin{subfigure}{0.18\textwidth}      \includegraphics[width=\textwidth]{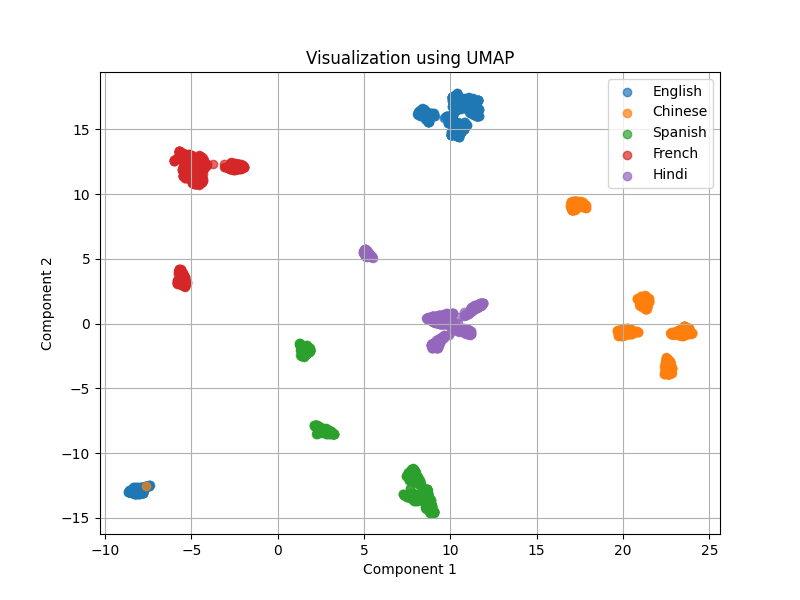}      \caption{UMAP, Layer 27}        \end{subfigure}  \hfill  \begin{subfigure}{0.18\textwidth}      \includegraphics[width=\textwidth]{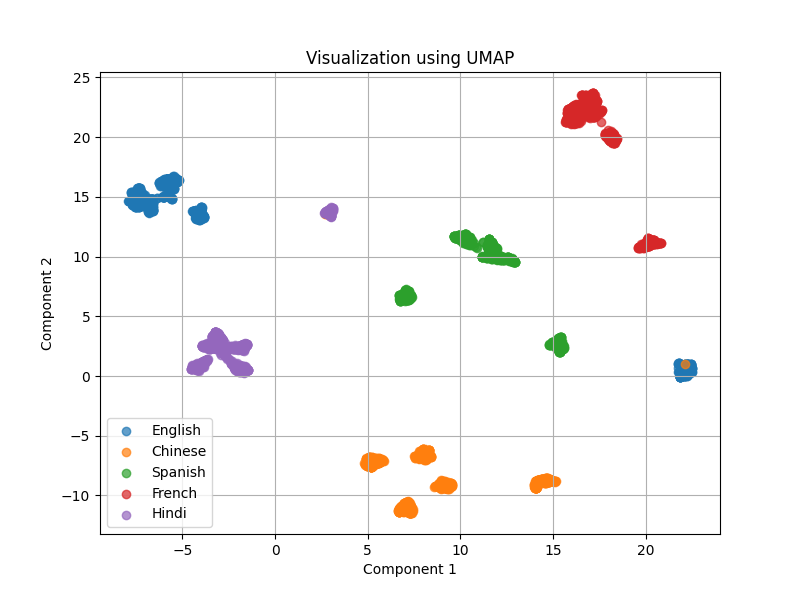}      \caption{UMAP, Layer 28}        \end{subfigure}  \hfill  \begin{subfigure}{0.18\textwidth}      \includegraphics[width=\textwidth]{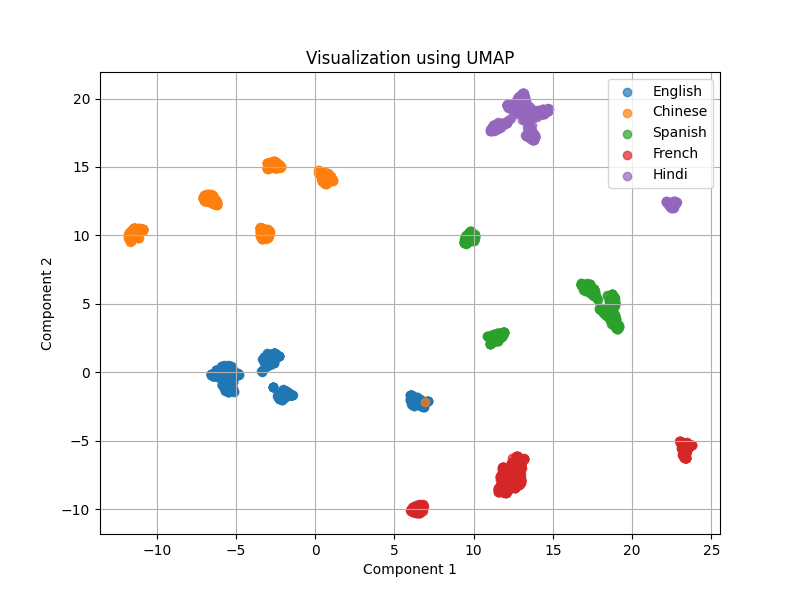}      \caption{UMAP, Layer 29}        \end{subfigure}  \hfill  \begin{subfigure}{0.18\textwidth}      \includegraphics[width=\textwidth]{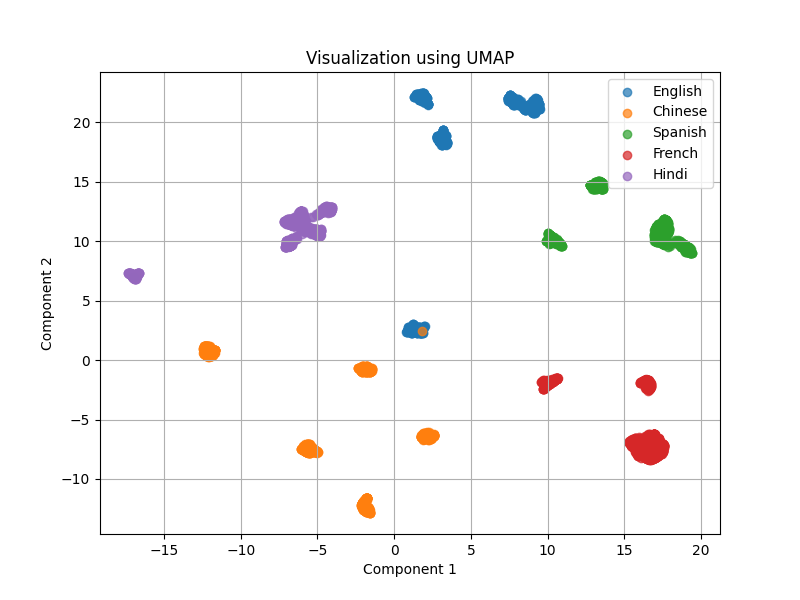}      \caption{UMAP, Layer 30}        \end{subfigure}    \vspace{0.2in}    %
\begin{subfigure}{0.18\textwidth}      \includegraphics[width=\textwidth]{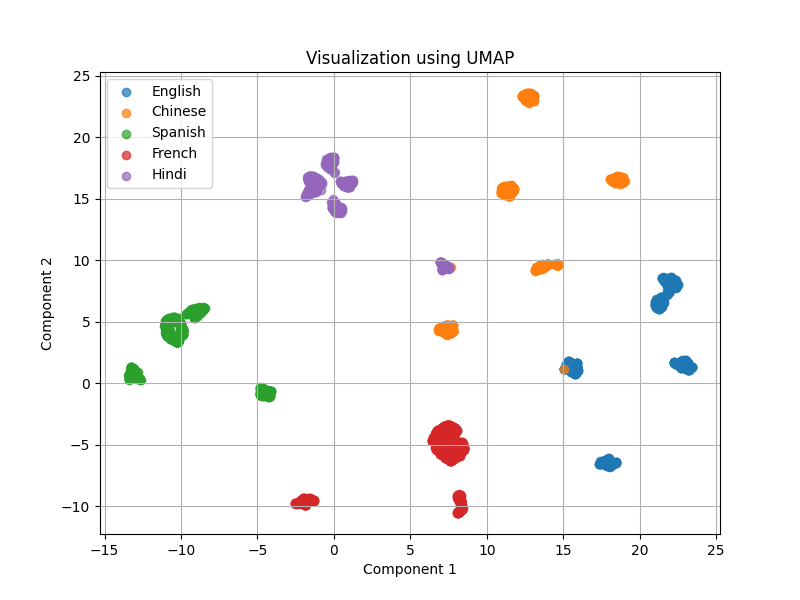}      \caption{UMAP, Layer 31}        \end{subfigure}  \hfill  \begin{subfigure}{0.18\textwidth}      \includegraphics[width=\textwidth]{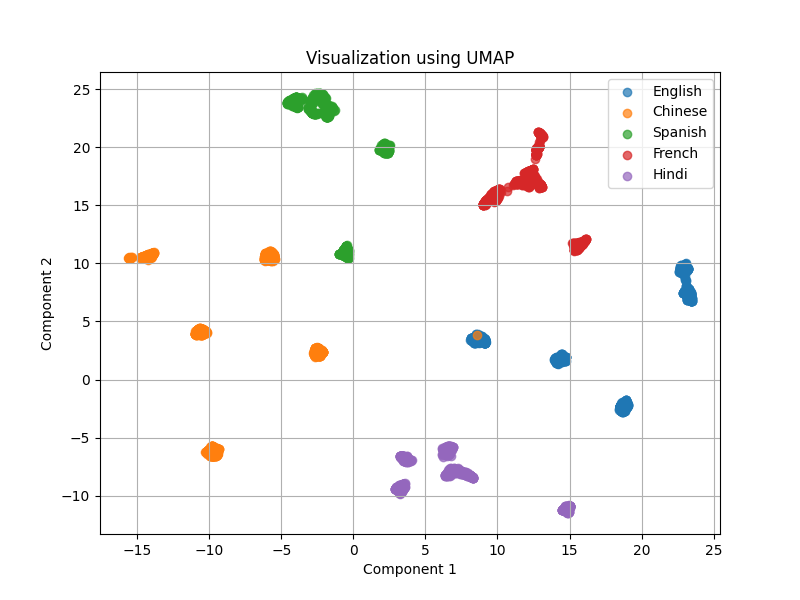}      \caption{UMAP, Layer 32}        \end{subfigure}    \caption{UMAP visualizations for layers 1-32 of Llama-3-8B-Instruct on the LogicalDeduction dataset.}  
\end{figure*}

\begin{figure*}[htbp]
\centering
\begin{subfigure}{0.18\textwidth}
\includegraphics[width=\textwidth]{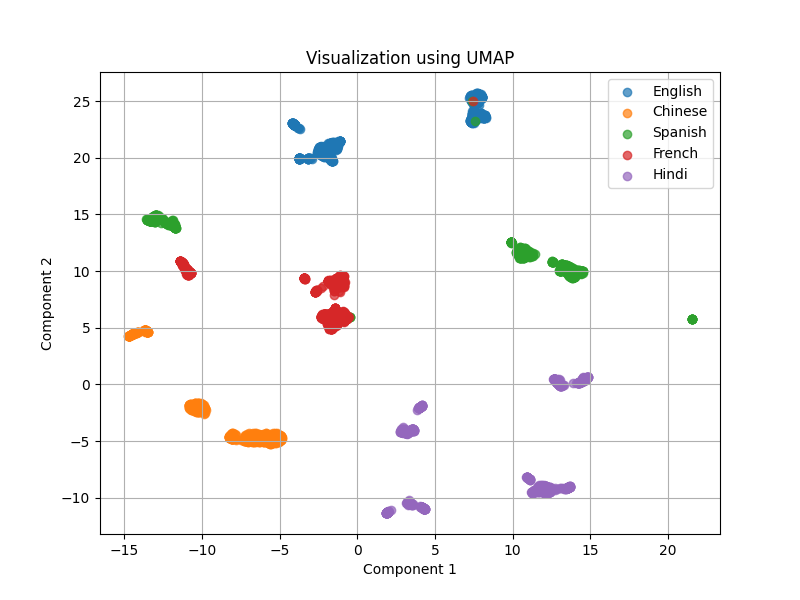}
\caption{UMAP, Layer 1}
\end{subfigure}
\hfill
\begin{subfigure}{0.18\textwidth}
\includegraphics[width=\textwidth]{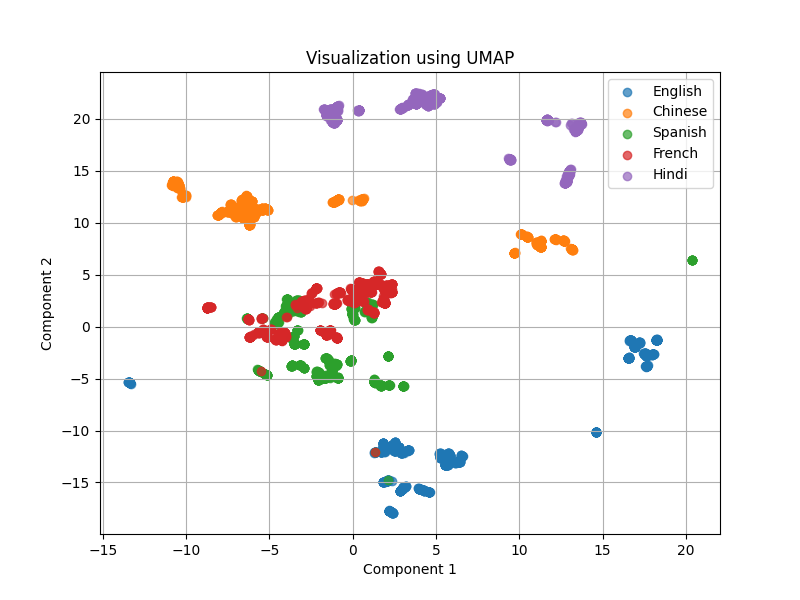}
\caption{UMAP, Layer 2}

\end{subfigure}
\hfill
\begin{subfigure}{0.18\textwidth}
\includegraphics[width=\textwidth]{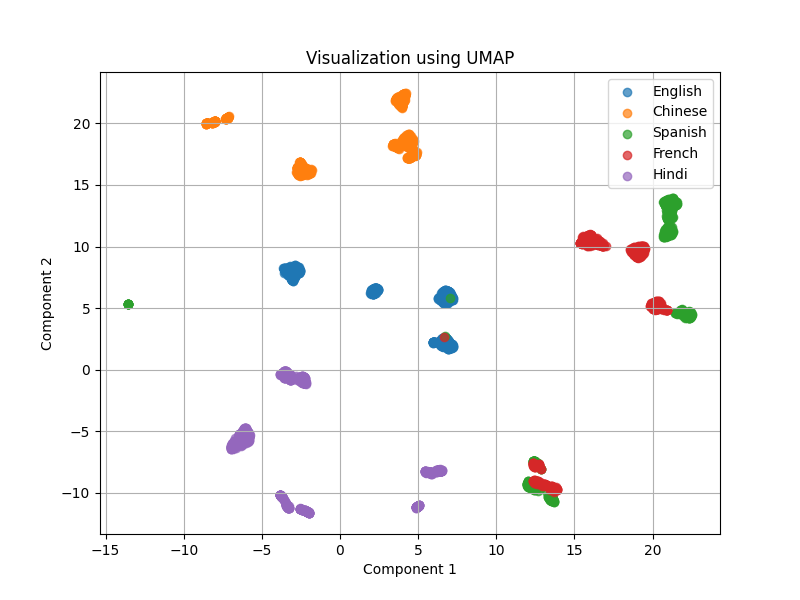}
\caption{UMAP, Layer 3}

\end{subfigure}
\hfill
\begin{subfigure}{0.18\textwidth}
\includegraphics[width=\textwidth]{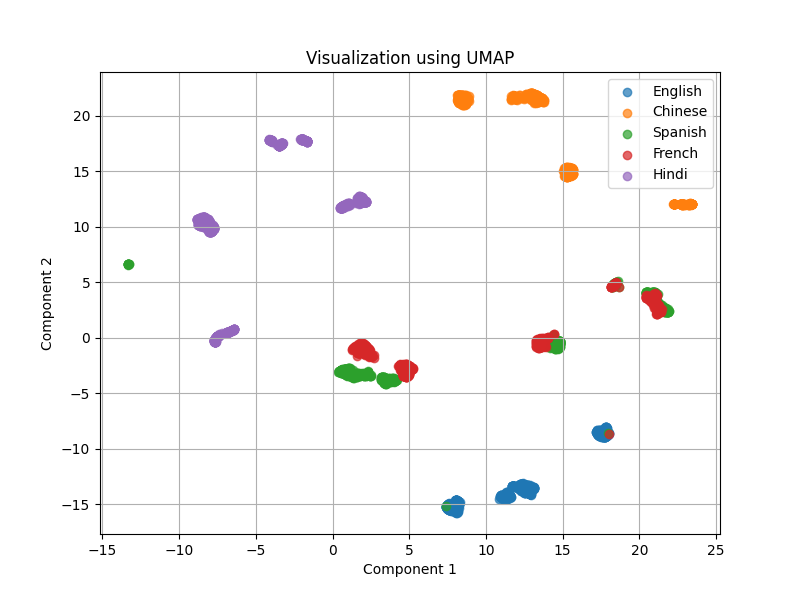}
\caption{UMAP, Layer 4}

\end{subfigure}
\hfill
\begin{subfigure}{0.18\textwidth}
\includegraphics[width=\textwidth]{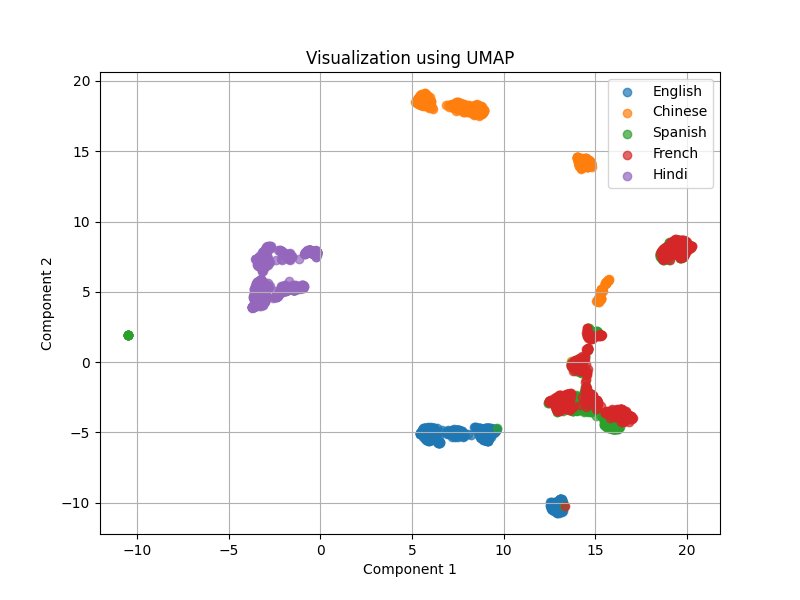}
\caption{UMAP, Layer 5}

\end{subfigure}
\vspace{0.2in} %
\begin{subfigure}{0.18\textwidth}      \includegraphics[width=\textwidth]{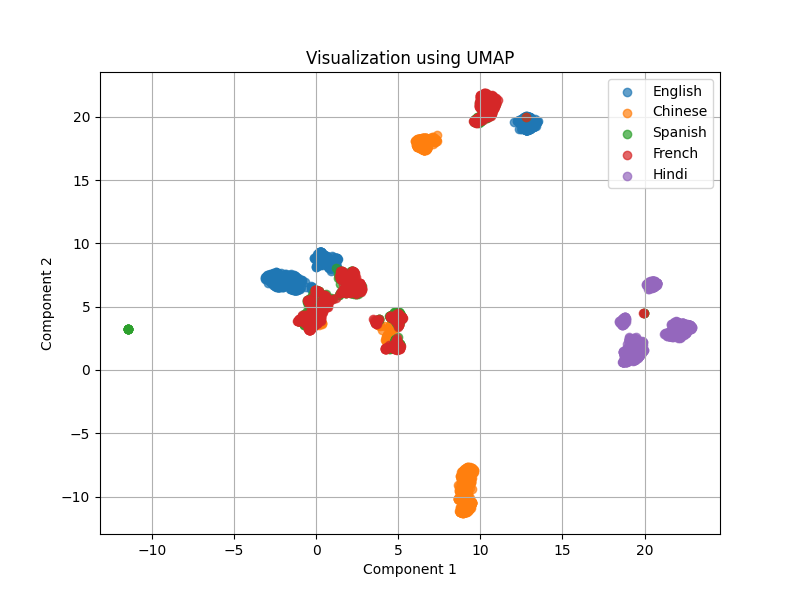}      \caption{UMAP, Layer 6}        \end{subfigure}  \hfill  \begin{subfigure}{0.18\textwidth}      \includegraphics[width=\textwidth]{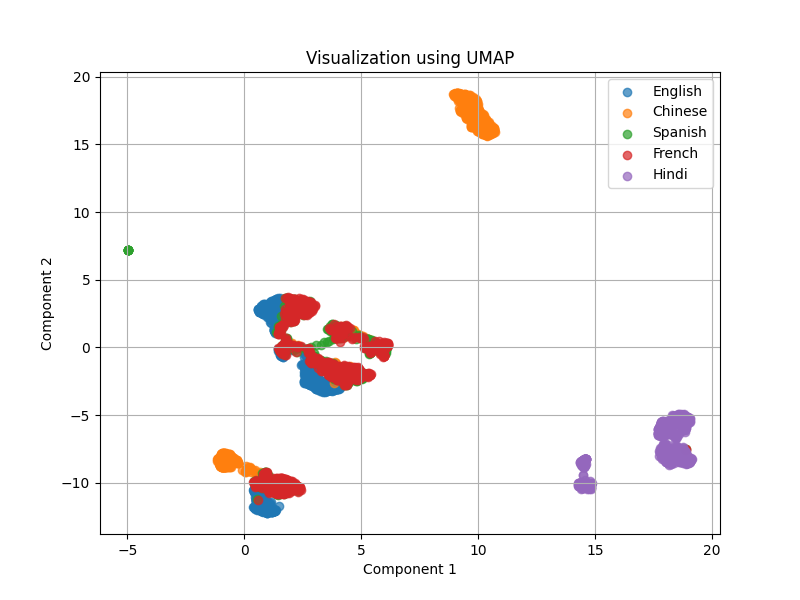}      \caption{UMAP, Layer 7}        \end{subfigure}  \hfill  \begin{subfigure}{0.18\textwidth}      \includegraphics[width=\textwidth]{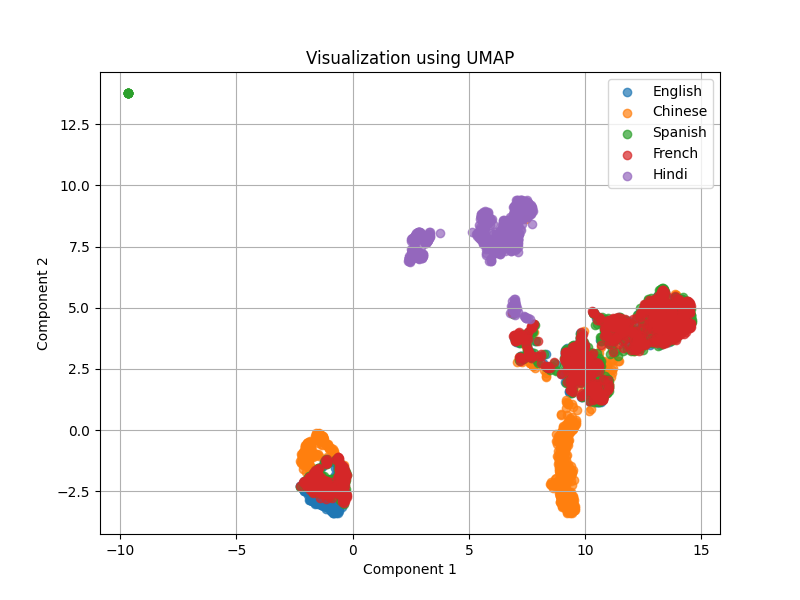}      \caption{UMAP, Layer 8}        \end{subfigure}  \hfill  \begin{subfigure}{0.18\textwidth}      \includegraphics[width=\textwidth]{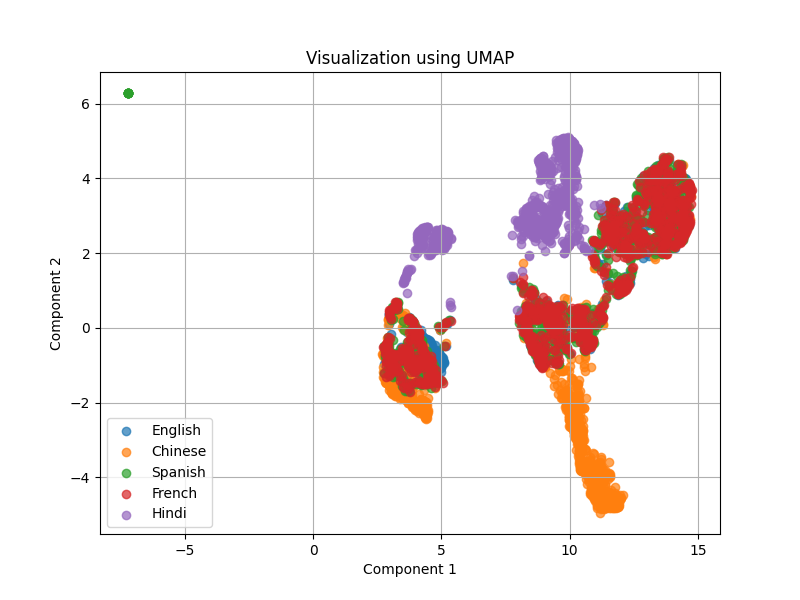}      \caption{UMAP, Layer 9}        \end{subfigure}  \hfill  \begin{subfigure}{0.18\textwidth}      \includegraphics[width=\textwidth]{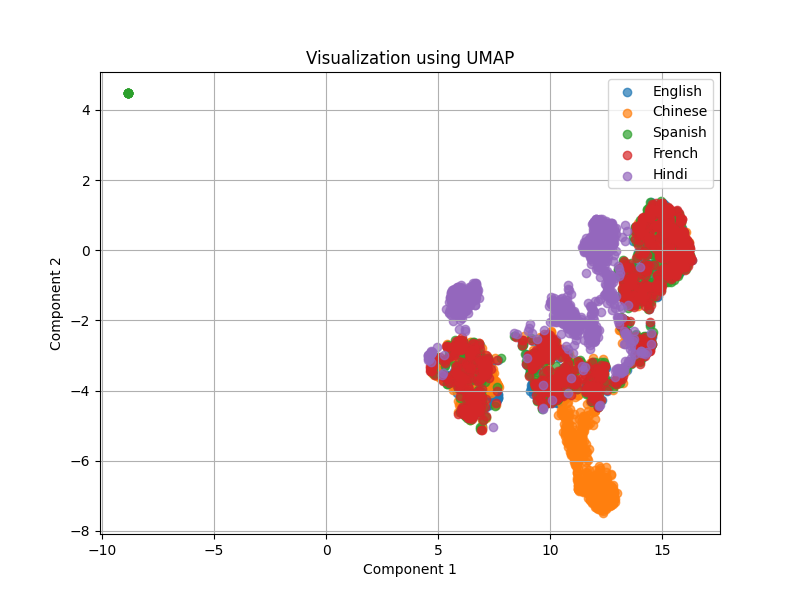}      \caption{UMAP, Layer 10}        \end{subfigure}    \vspace{0.2in}    %
\begin{subfigure}{0.18\textwidth}      \includegraphics[width=\textwidth]{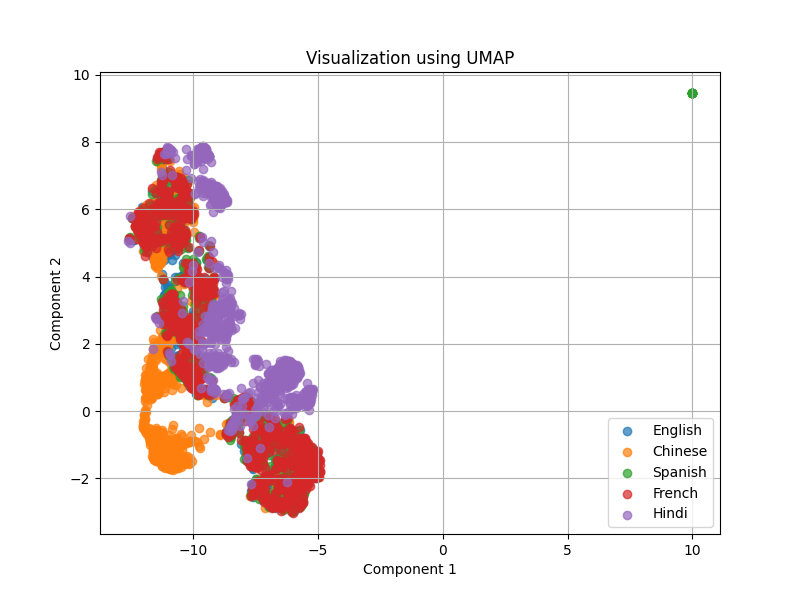}      \caption{UMAP, Layer 11}        \end{subfigure}  \hfill  \begin{subfigure}{0.18\textwidth}      \includegraphics[width=\textwidth]{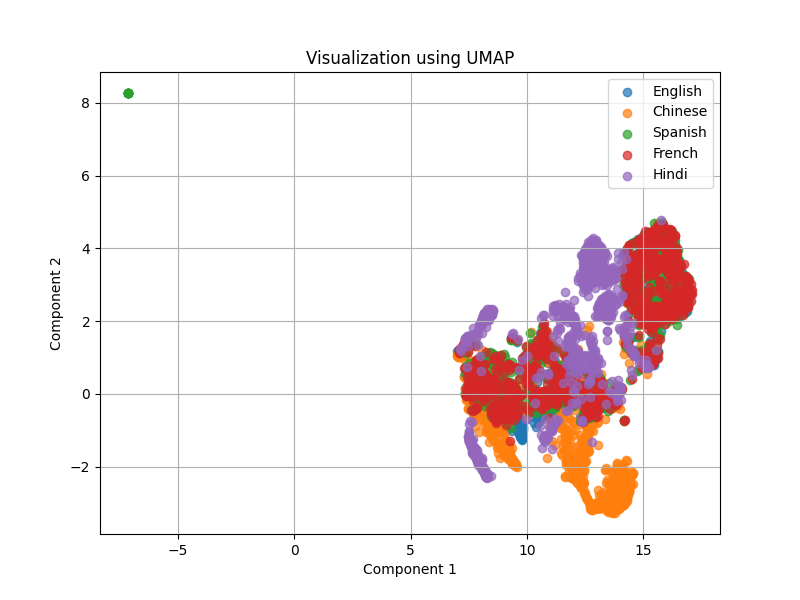}      \caption{UMAP, Layer 12}        \end{subfigure}  \hfill  \begin{subfigure}{0.18\textwidth}      \includegraphics[width=\textwidth]{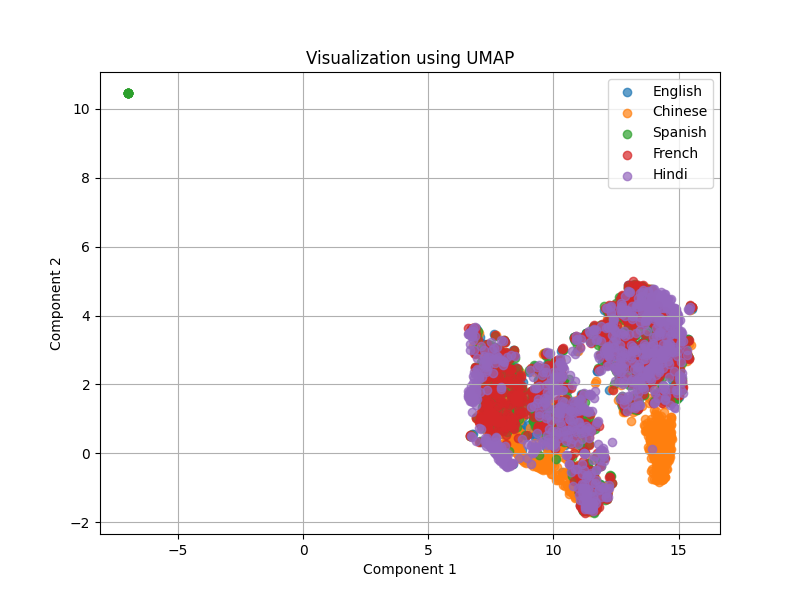}      \caption{UMAP, Layer 13}        \end{subfigure}  \hfill  \begin{subfigure}{0.18\textwidth}      \includegraphics[width=\textwidth]{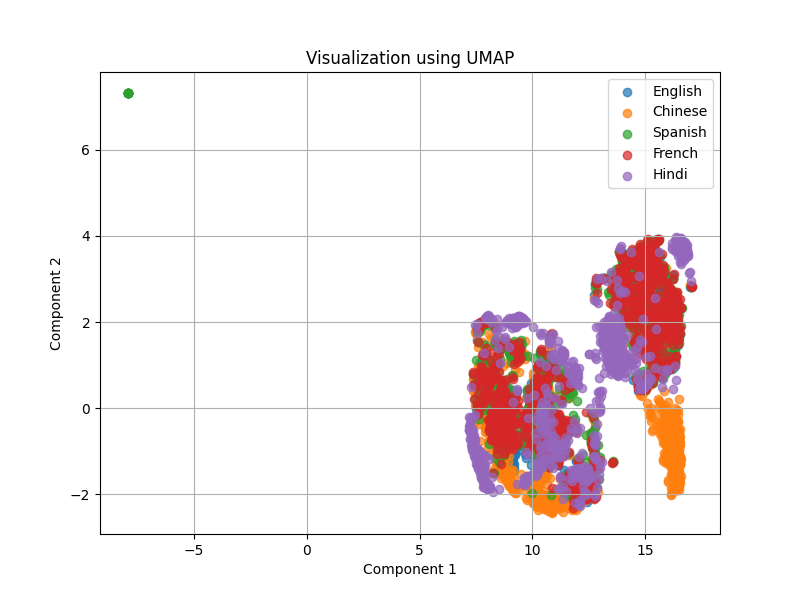}      \caption{UMAP, Layer 14}        \end{subfigure}  \hfill  \begin{subfigure}{0.18\textwidth}      \includegraphics[width=\textwidth]{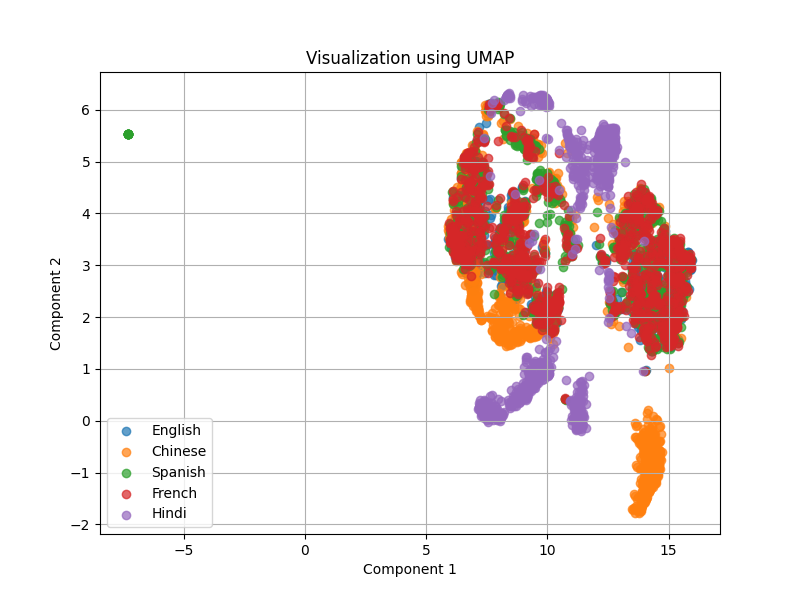}      \caption{UMAP, Layer 15}        \end{subfigure}    \vspace{0.2in}    %
\begin{subfigure}{0.18\textwidth}      \includegraphics[width=\textwidth]{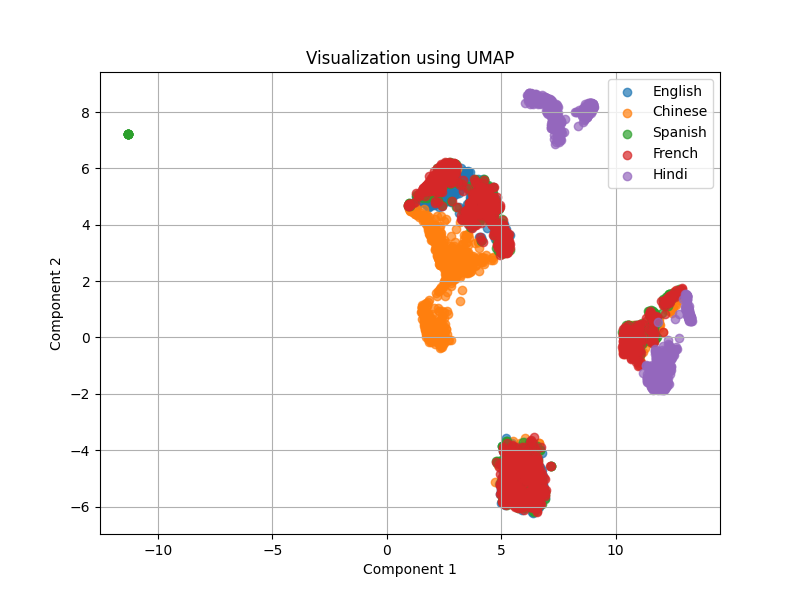}      \caption{UMAP, Layer 16}        \end{subfigure}  \hfill  \begin{subfigure}{0.18\textwidth}      \includegraphics[width=\textwidth]{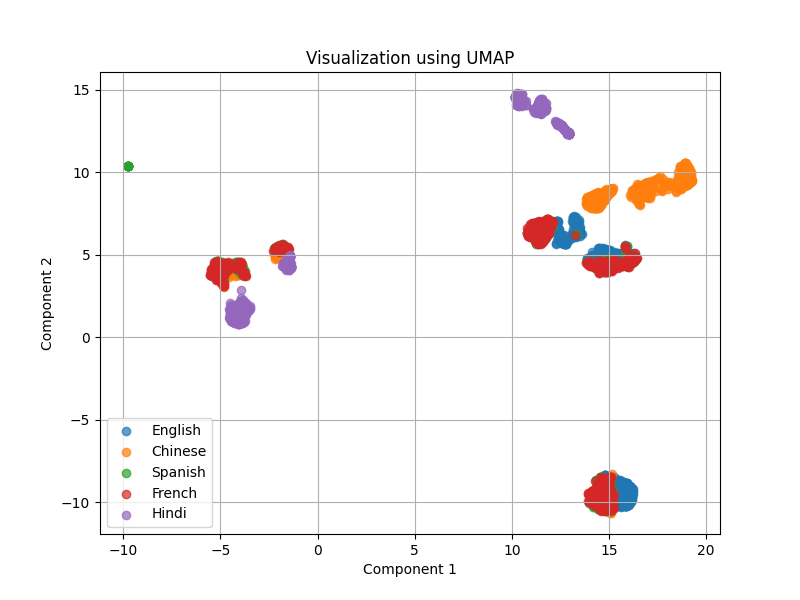}      \caption{UMAP, Layer 17}        \end{subfigure}  \hfill  \begin{subfigure}{0.18\textwidth}      \includegraphics[width=\textwidth]{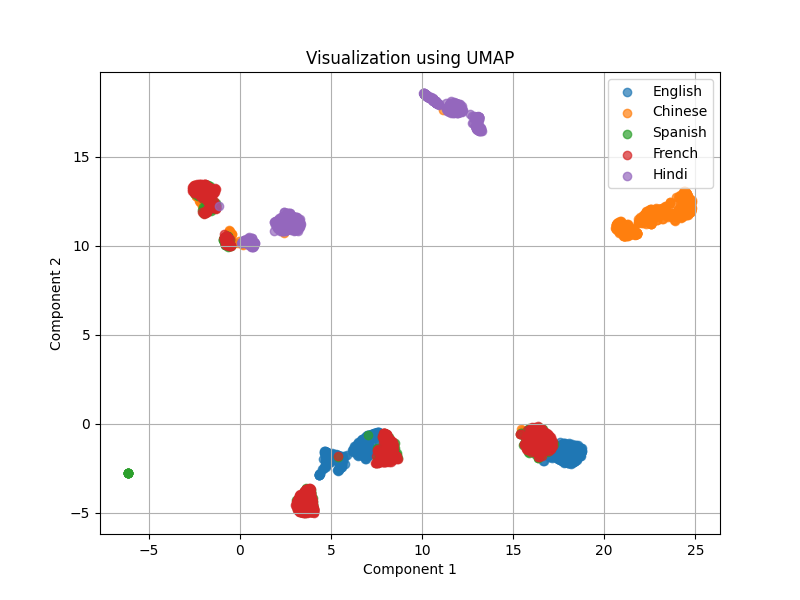}      \caption{UMAP, Layer 18}        \end{subfigure}  \hfill  \begin{subfigure}{0.18\textwidth}      \includegraphics[width=\textwidth]{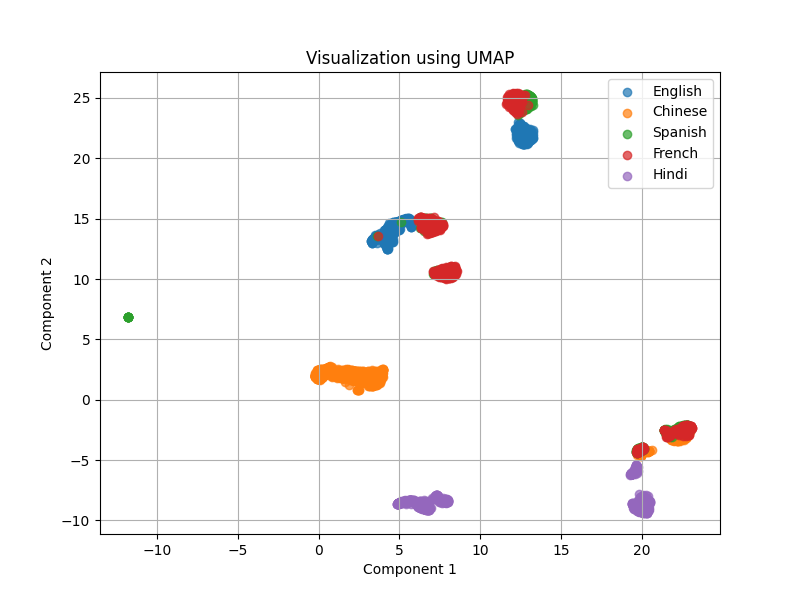}      \caption{UMAP, Layer 19}        \end{subfigure}  \hfill  \begin{subfigure}{0.18\textwidth}      \includegraphics[width=\textwidth]{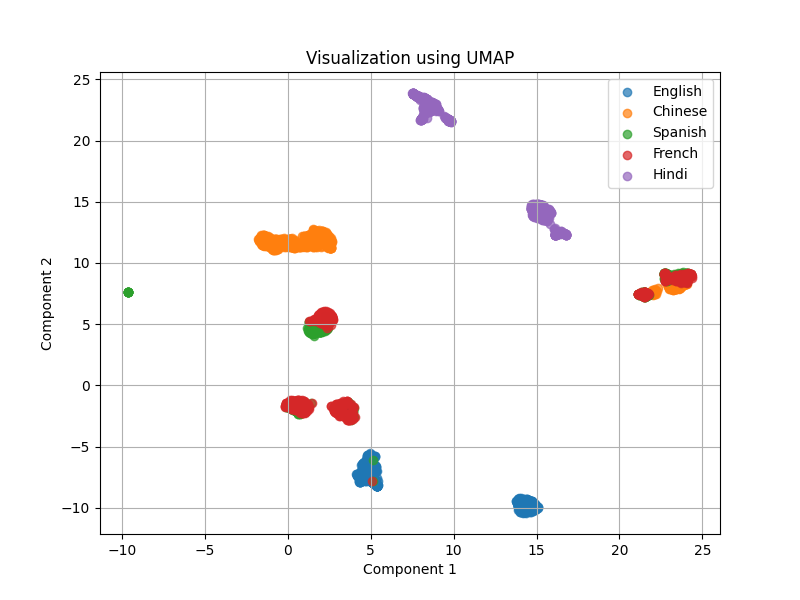}      \caption{UMAP, Layer 20}        \end{subfigure}    \vspace{0.2in}    %
\begin{subfigure}{0.18\textwidth}      \includegraphics[width=\textwidth]{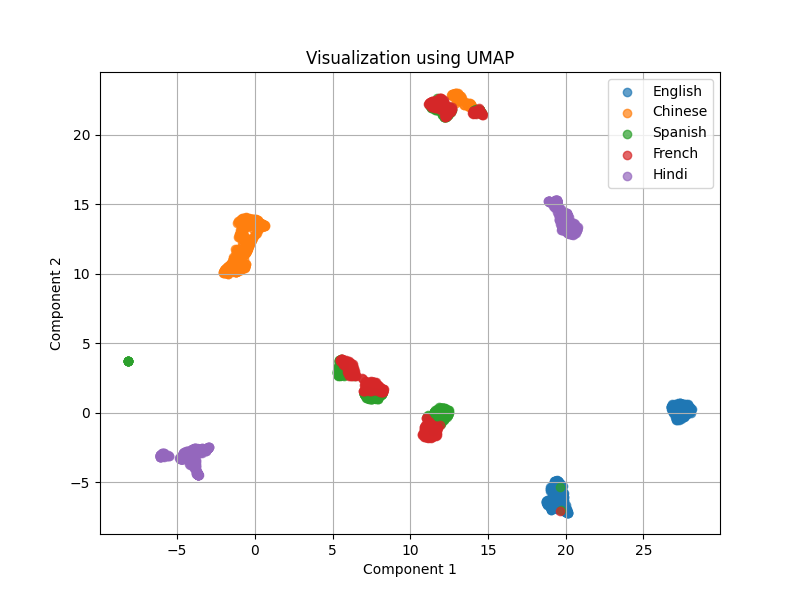}      \caption{UMAP, Layer 21}        \end{subfigure}  \hfill  \begin{subfigure}{0.18\textwidth}      \includegraphics[width=\textwidth]{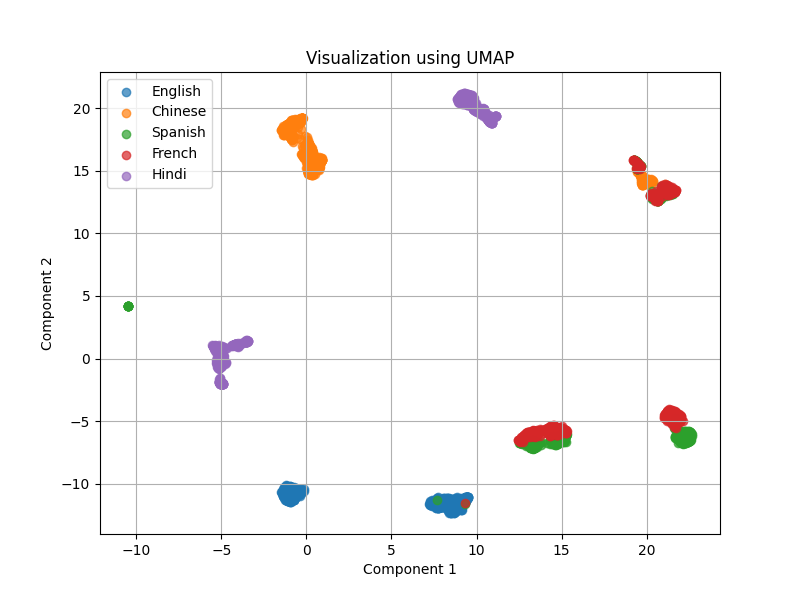}      \caption{UMAP, Layer 22}        \end{subfigure}  \hfill  \begin{subfigure}{0.18\textwidth}      \includegraphics[width=\textwidth]{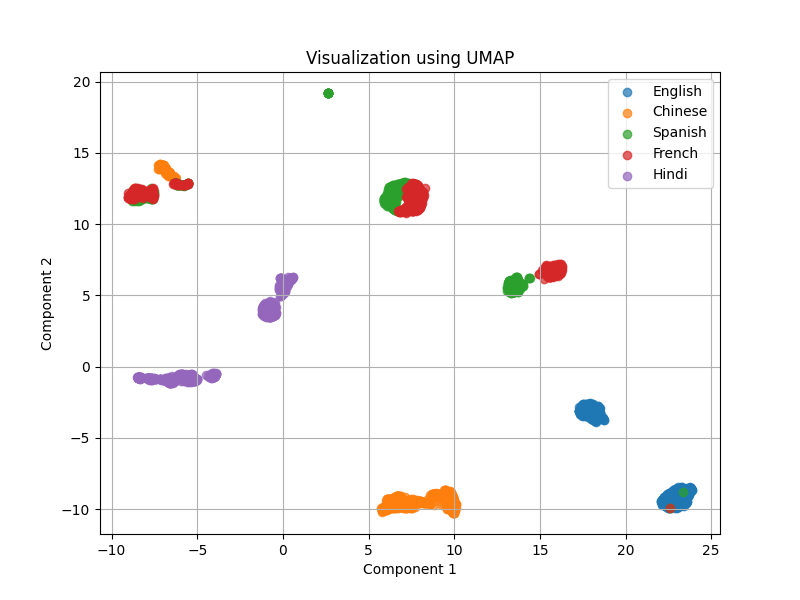}      \caption{UMAP, Layer 23}        \end{subfigure}  \hfill  \begin{subfigure}{0.18\textwidth}      \includegraphics[width=\textwidth]{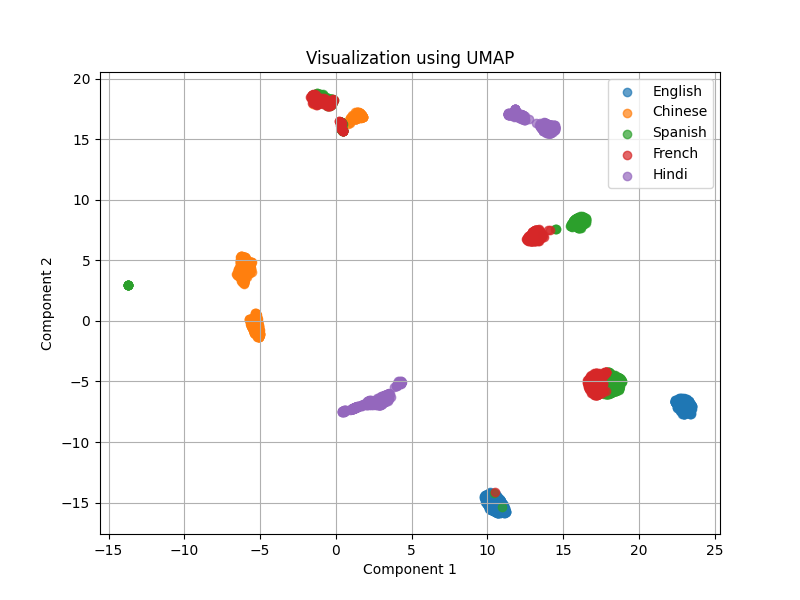}      \caption{UMAP, Layer 24}        \end{subfigure}  \hfill  \begin{subfigure}{0.18\textwidth}      \includegraphics[width=\textwidth]{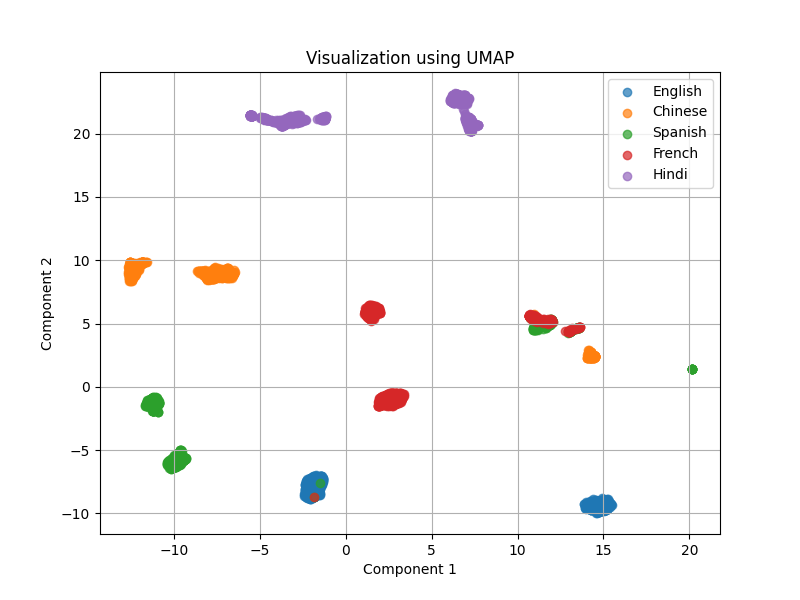}      \caption{UMAP, Layer 25}        \end{subfigure}    \vspace{0.2in}    %
\begin{subfigure}{0.18\textwidth}      \includegraphics[width=\textwidth]{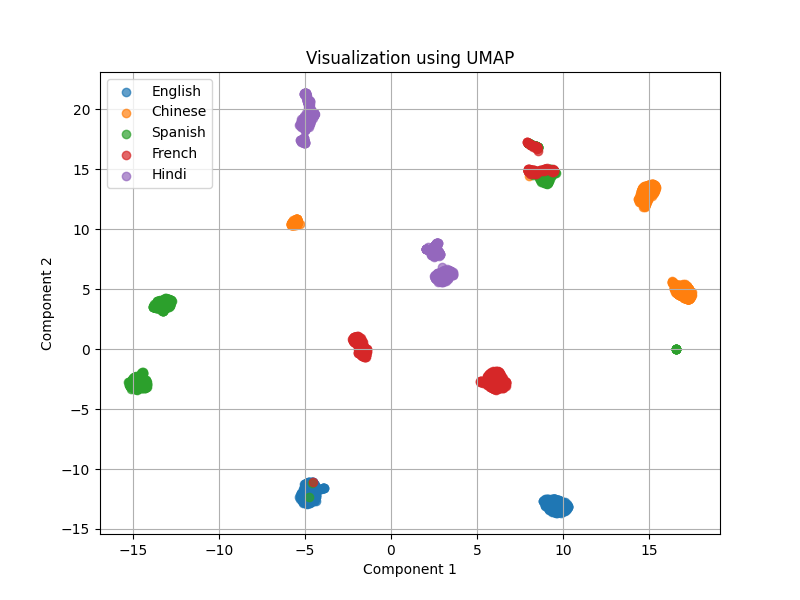}      \caption{UMAP, Layer 26}        \end{subfigure}  \hfill  \begin{subfigure}{0.18\textwidth}      \includegraphics[width=\textwidth]{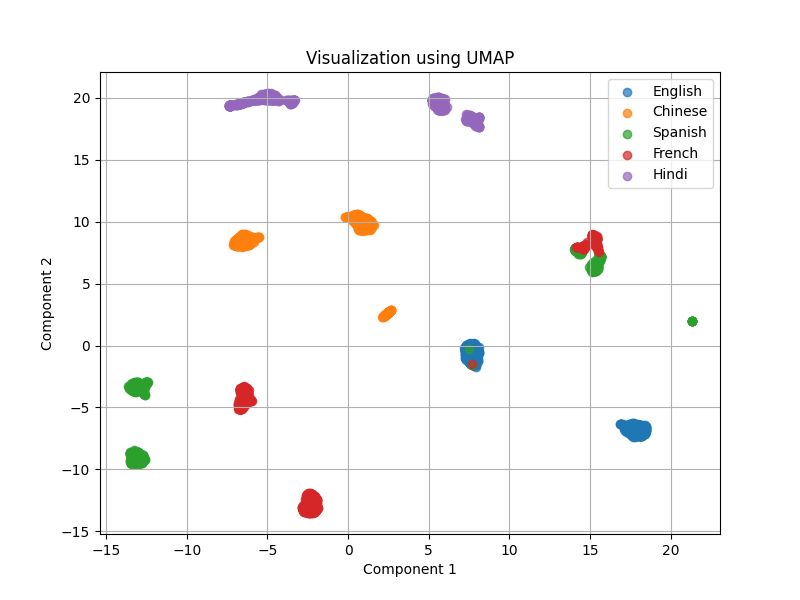}      \caption{UMAP, Layer 27}        \end{subfigure}  \hfill  \begin{subfigure}{0.18\textwidth}      \includegraphics[width=\textwidth]{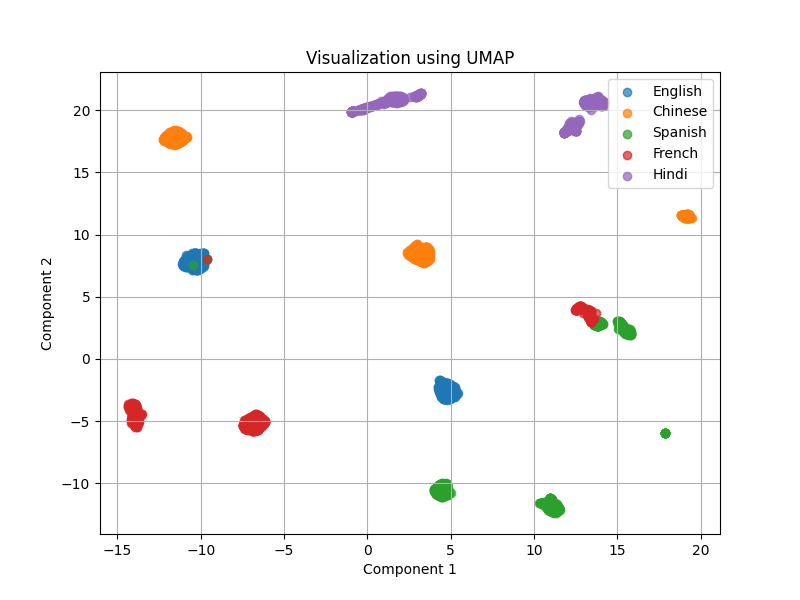}      \caption{UMAP, Layer 28}        \end{subfigure}  \hfill  \begin{subfigure}{0.18\textwidth}      \includegraphics[width=\textwidth]{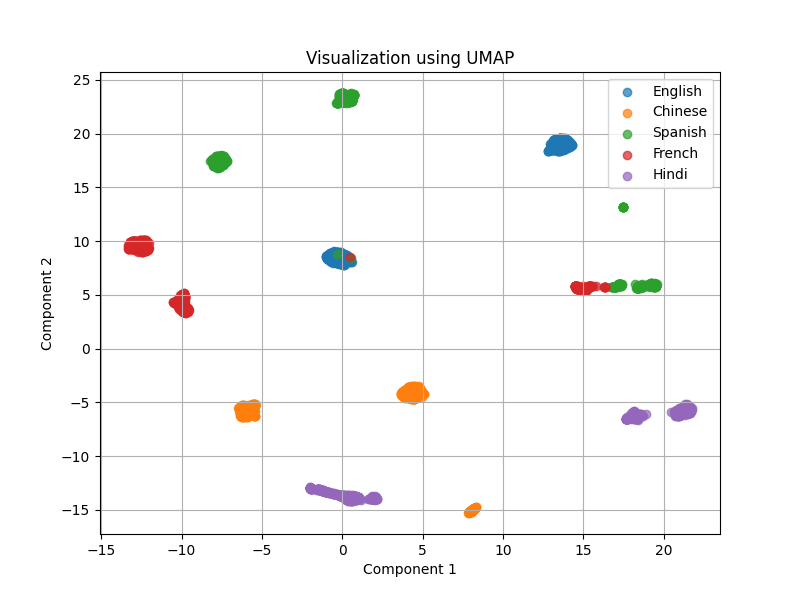}      \caption{UMAP, Layer 29}        \end{subfigure}  \hfill  \begin{subfigure}{0.18\textwidth}      \includegraphics[width=\textwidth]{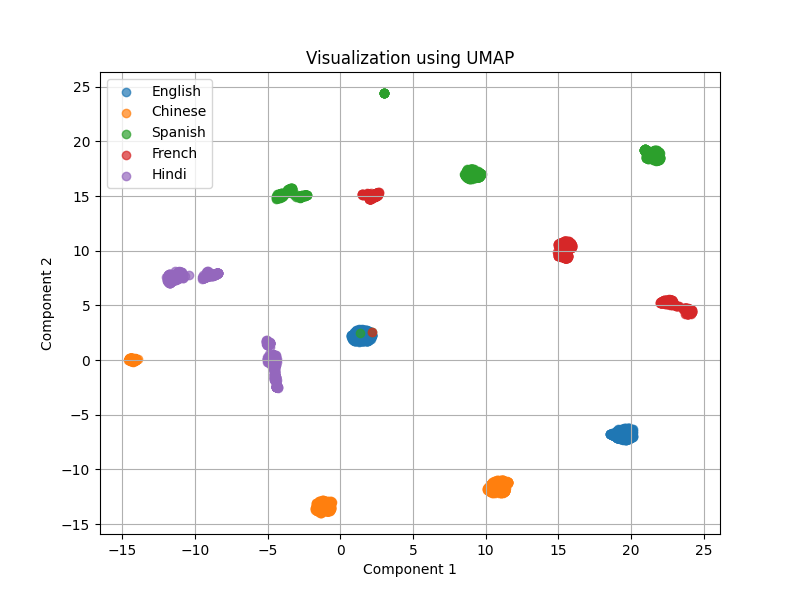}      \caption{UMAP, Layer 30}        \end{subfigure}    \vspace{0.2in}    %
\begin{subfigure}{0.18\textwidth}      \includegraphics[width=\textwidth]{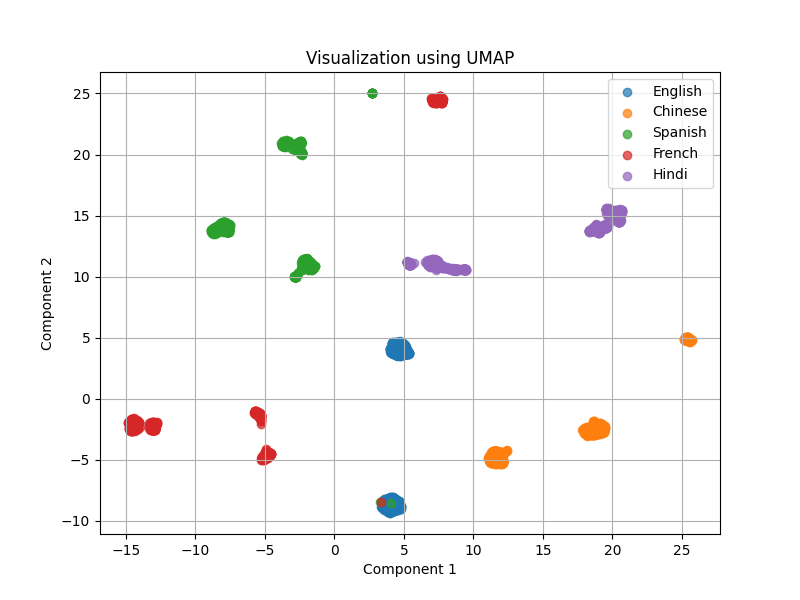}      \caption{UMAP, Layer 31}        \end{subfigure}  \hfill  \begin{subfigure}{0.18\textwidth}      \includegraphics[width=\textwidth]{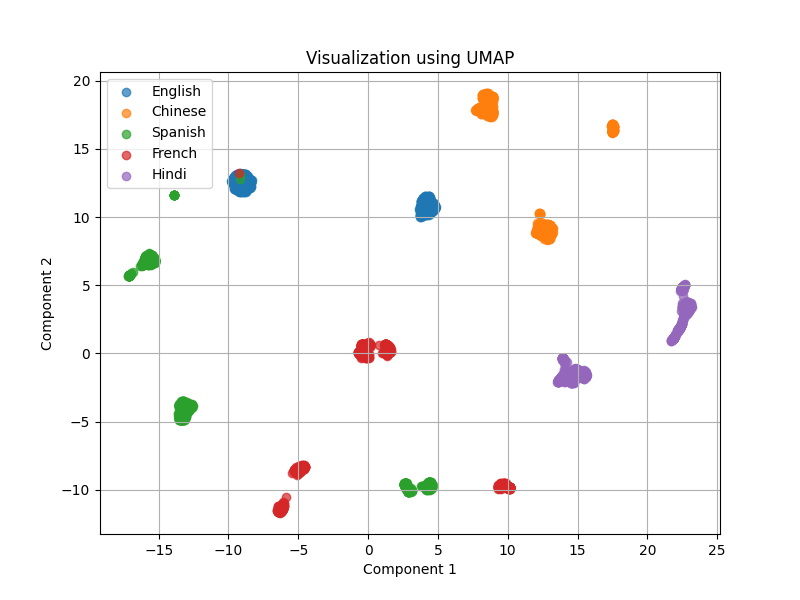}      \caption{UMAP, Layer 32}        \end{subfigure}    \caption{UMAP visualizations for layers 1-32 of Llama-3-8B-Instruct on the ProofWriter dataset.}  
\end{figure*}

\begin{figure*}[htbp]
\centering
\begin{subfigure}{0.18\textwidth}
\includegraphics[width=\textwidth]{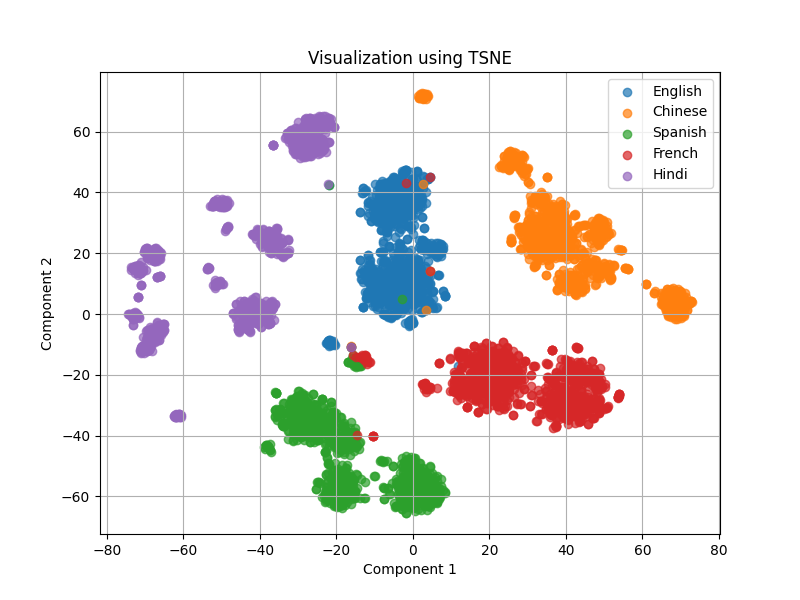}
\caption{T-SNE, Layer 1}
\end{subfigure}
\hfill
\begin{subfigure}{0.18\textwidth}
\includegraphics[width=\textwidth]{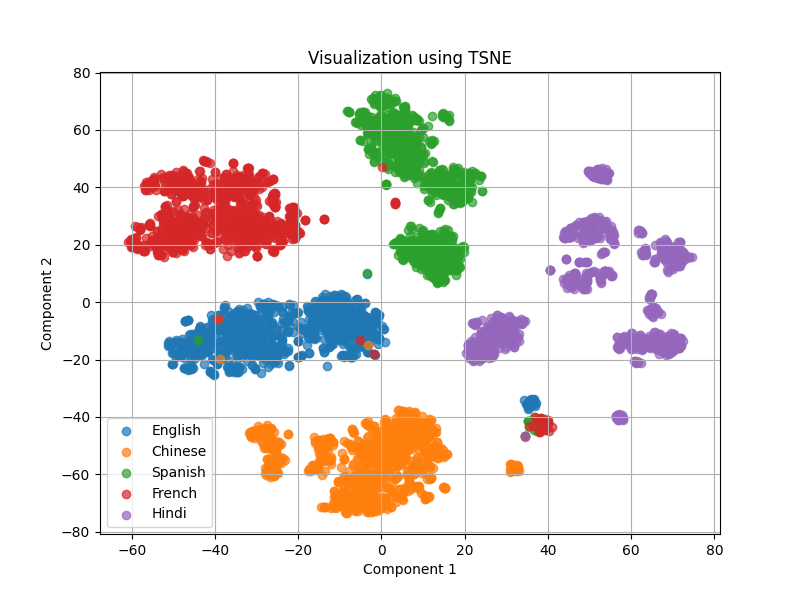}
\caption{T-SNE, Layer 2}

\end{subfigure}
\hfill
\begin{subfigure}{0.18\textwidth}
\includegraphics[width=\textwidth]{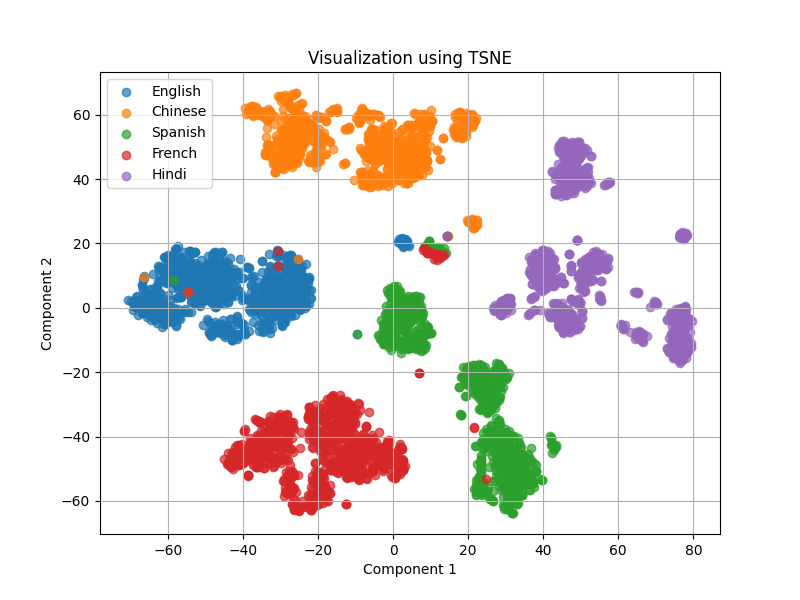}
\caption{T-SNE, Layer 3}

\end{subfigure}
\hfill
\begin{subfigure}{0.18\textwidth}
\includegraphics[width=\textwidth]{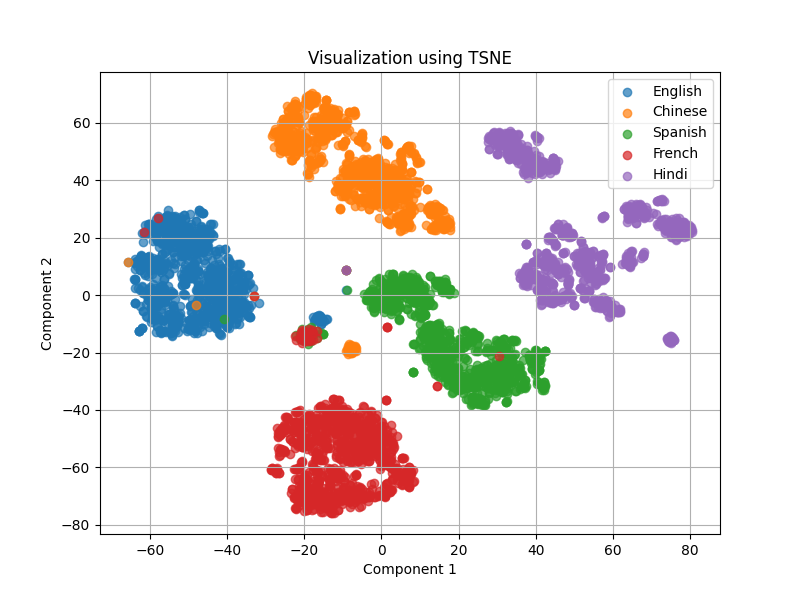}
\caption{T-SNE, Layer 4}

\end{subfigure}
\hfill
\begin{subfigure}{0.18\textwidth}
\includegraphics[width=\textwidth]{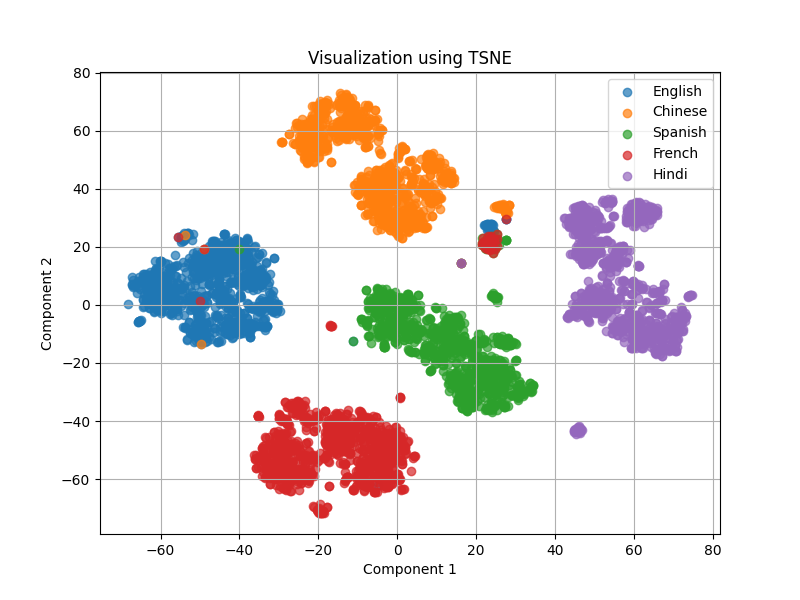}
\caption{T-SNE, Layer 5}

\end{subfigure}
\vspace{0.2in} %
\begin{subfigure}{0.18\textwidth}      \includegraphics[width=\textwidth]{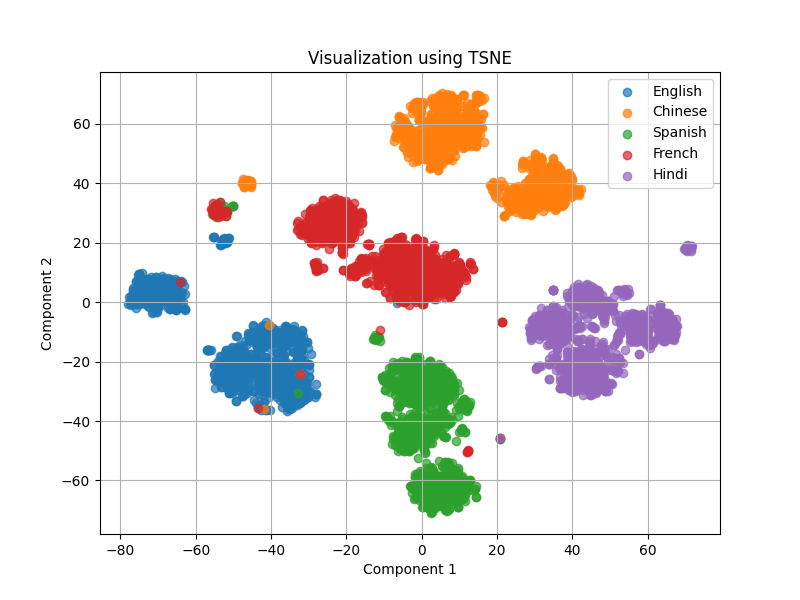}      \caption{T-SNE, Layer 6}        \end{subfigure}  \hfill  \begin{subfigure}{0.18\textwidth}      \includegraphics[width=\textwidth]{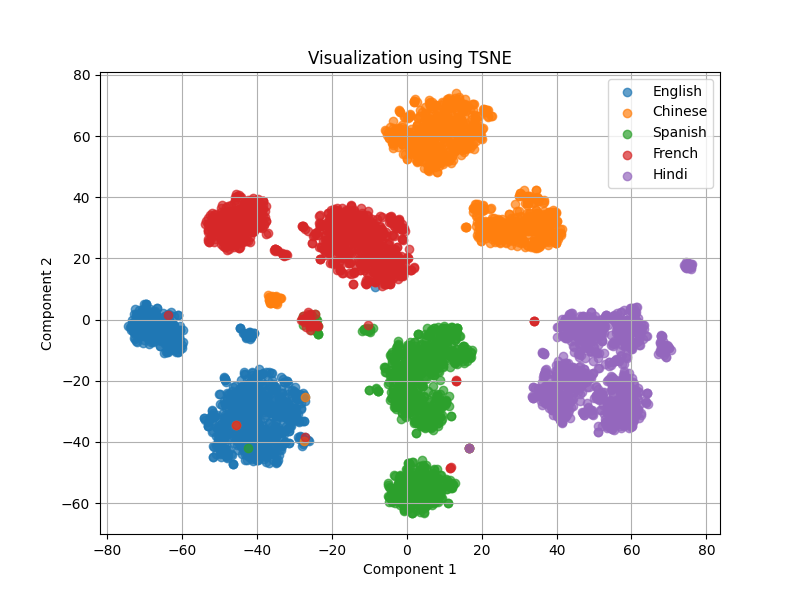}      \caption{T-SNE, Layer 7}        \end{subfigure}  \hfill  \begin{subfigure}{0.18\textwidth}      \includegraphics[width=\textwidth]{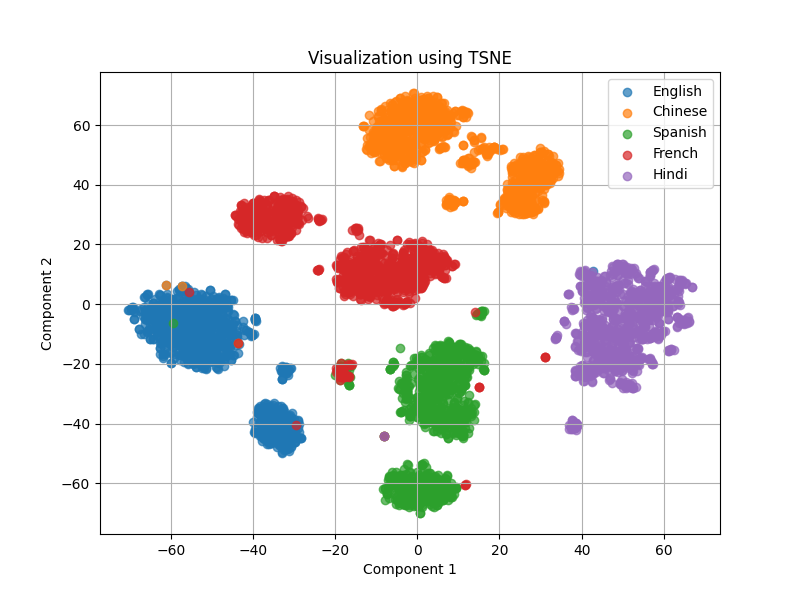}      \caption{T-SNE, Layer 8}        \end{subfigure}  \hfill  \begin{subfigure}{0.18\textwidth}      \includegraphics[width=\textwidth]{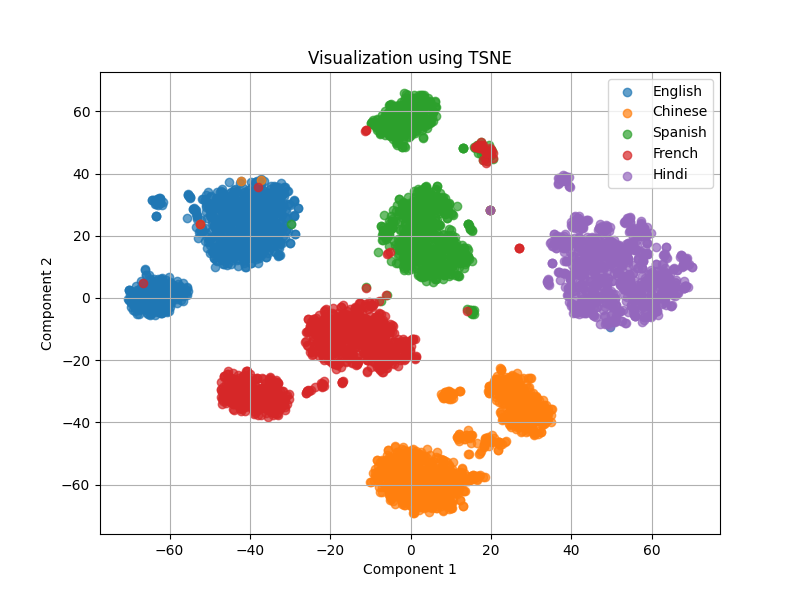}      \caption{T-SNE, Layer 9}        \end{subfigure}  \hfill  \begin{subfigure}{0.18\textwidth}      \includegraphics[width=\textwidth]{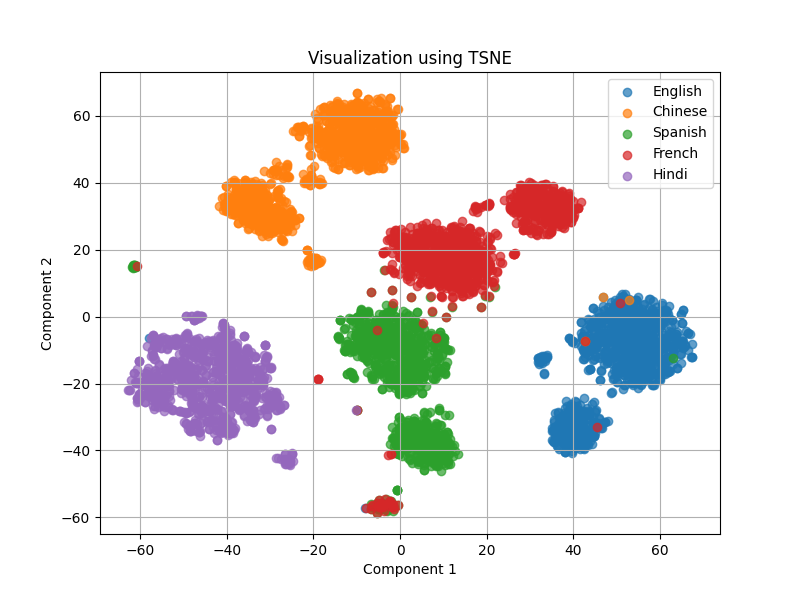}      \caption{T-SNE, Layer 10}        \end{subfigure}    \vspace{0.2in}    %
\begin{subfigure}{0.18\textwidth}      \includegraphics[width=\textwidth]{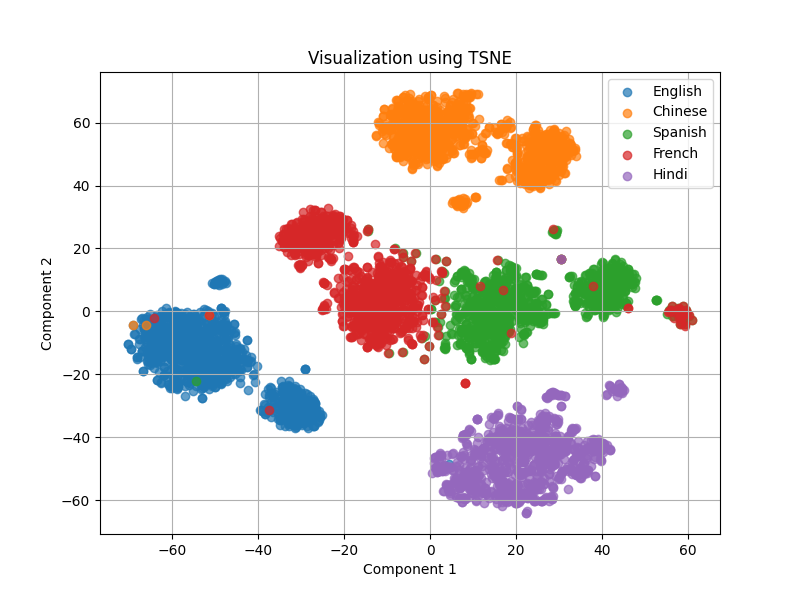}      \caption{T-SNE, Layer 11}        \end{subfigure}  \hfill  \begin{subfigure}{0.18\textwidth}      \includegraphics[width=\textwidth]{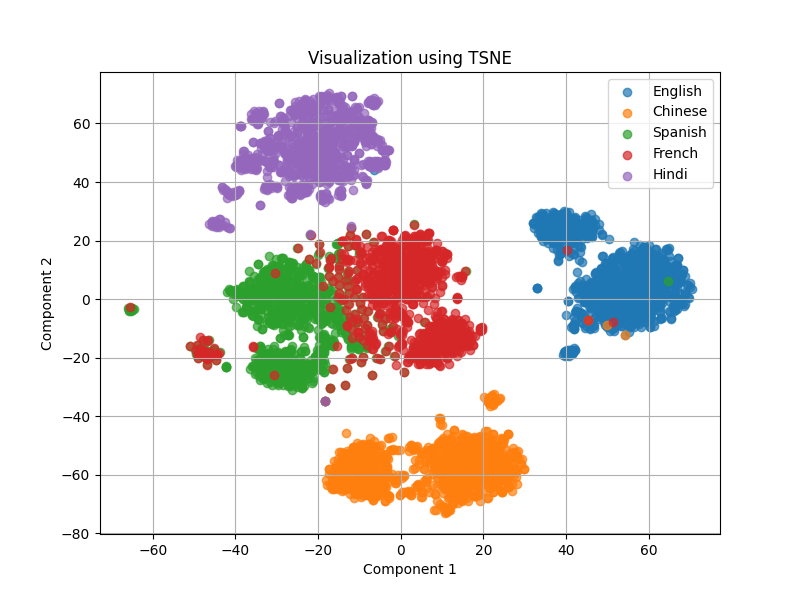}      \caption{T-SNE, Layer 12}        \end{subfigure}  \hfill  \begin{subfigure}{0.18\textwidth}      \includegraphics[width=\textwidth]{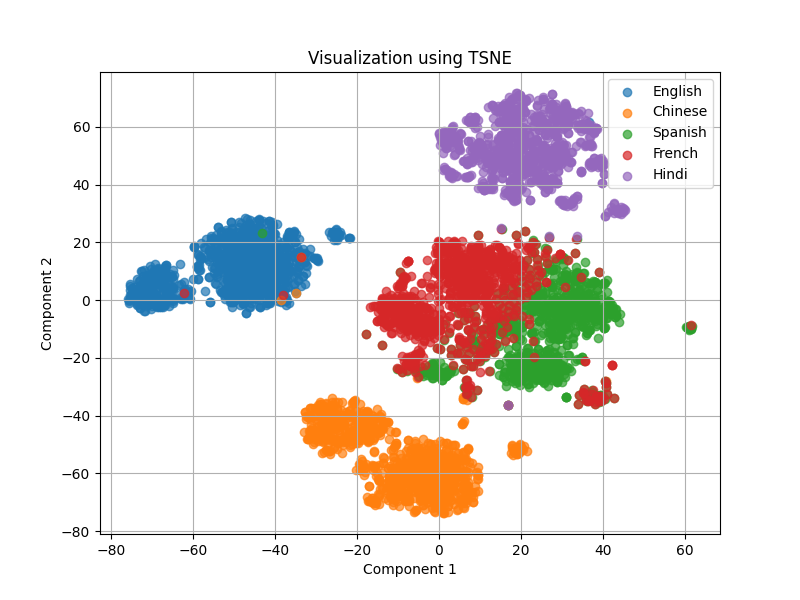}      \caption{T-SNE, Layer 13}        \end{subfigure}  \hfill  \begin{subfigure}{0.18\textwidth}      \includegraphics[width=\textwidth]{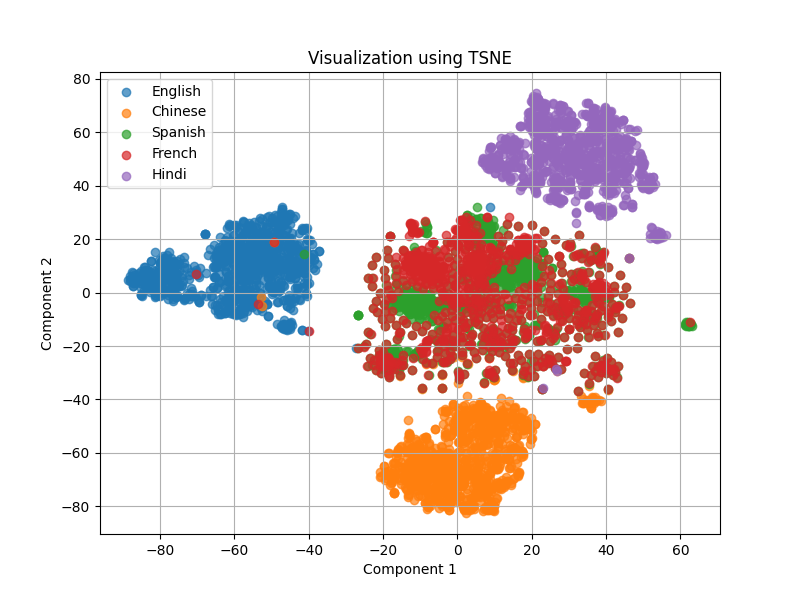}      \caption{T-SNE, Layer 14}        \end{subfigure}  \hfill  \begin{subfigure}{0.18\textwidth}      \includegraphics[width=\textwidth]{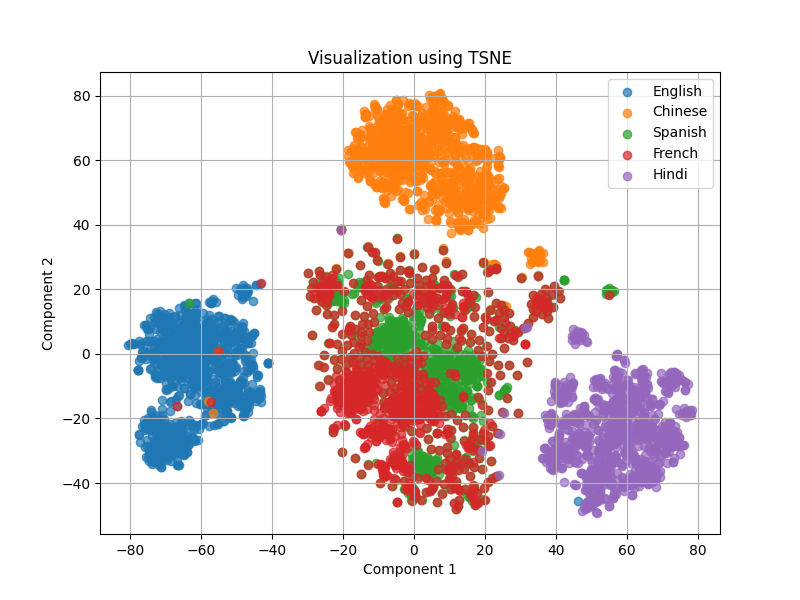}      \caption{T-SNE, Layer 15}        \end{subfigure}    \vspace{0.2in}    %
\begin{subfigure}{0.18\textwidth}      \includegraphics[width=\textwidth]{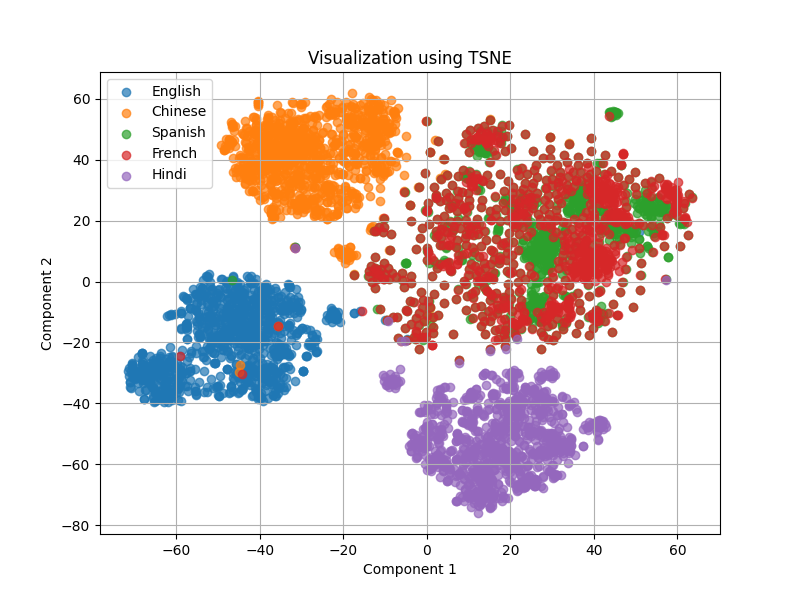}      \caption{T-SNE, Layer 16}        \end{subfigure}  \hfill  \begin{subfigure}{0.18\textwidth}      \includegraphics[width=\textwidth]{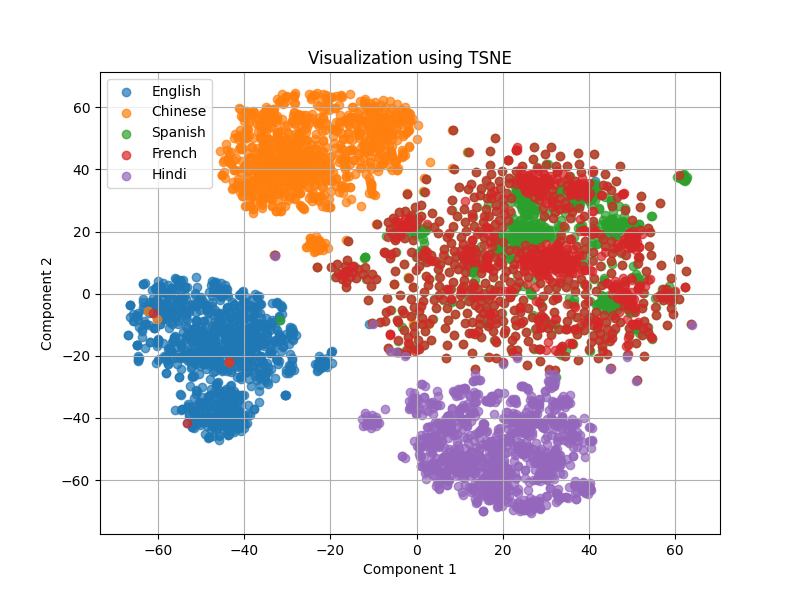}      \caption{T-SNE, Layer 17}        \end{subfigure}  \hfill  \begin{subfigure}{0.18\textwidth}      \includegraphics[width=\textwidth]{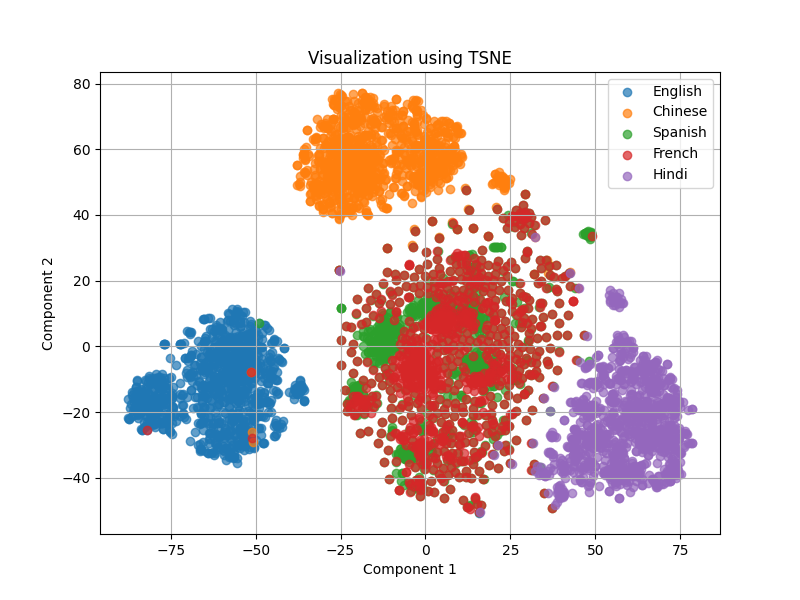}      \caption{T-SNE, Layer 18}        \end{subfigure}  \hfill  \begin{subfigure}{0.18\textwidth}      \includegraphics[width=\textwidth]{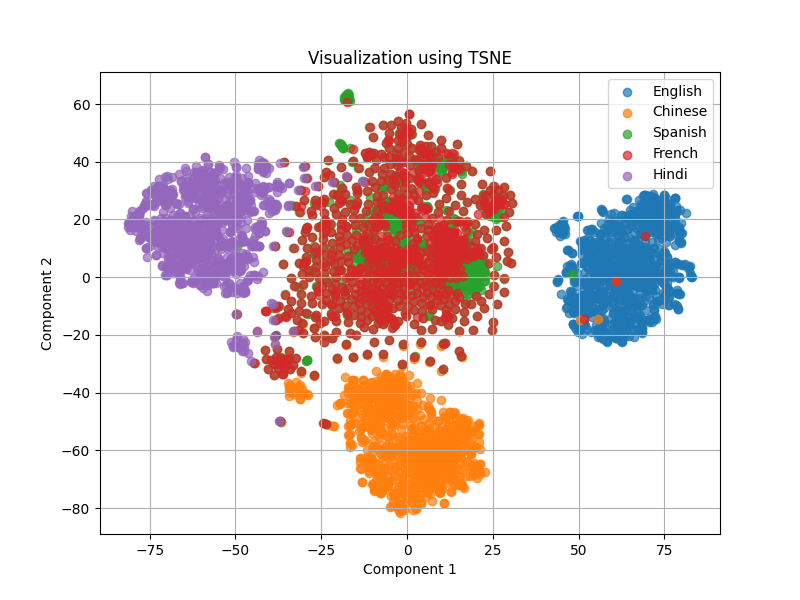}      \caption{T-SNE, Layer 19}        \end{subfigure}  \hfill  \begin{subfigure}{0.18\textwidth}      \includegraphics[width=\textwidth]{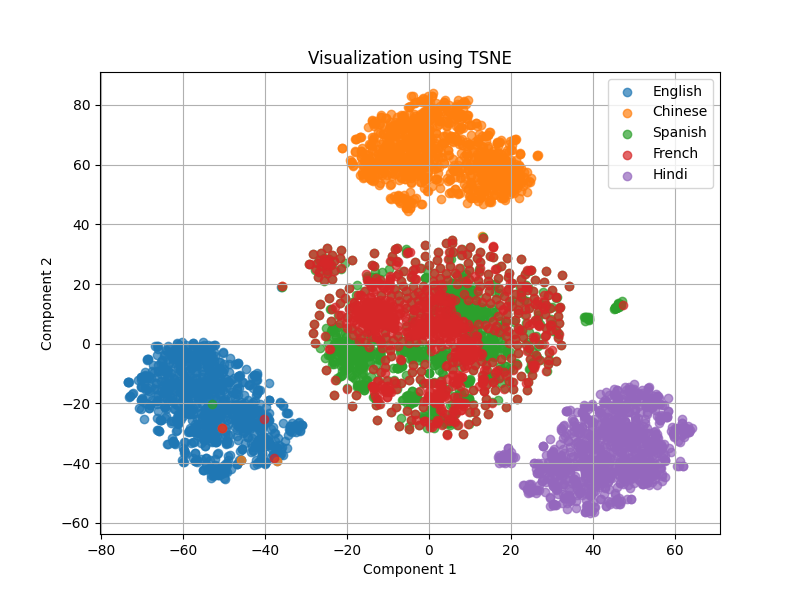}      \caption{T-SNE, Layer 20}        \end{subfigure}    \vspace{0.2in}    %
\begin{subfigure}{0.18\textwidth}      \includegraphics[width=\textwidth]{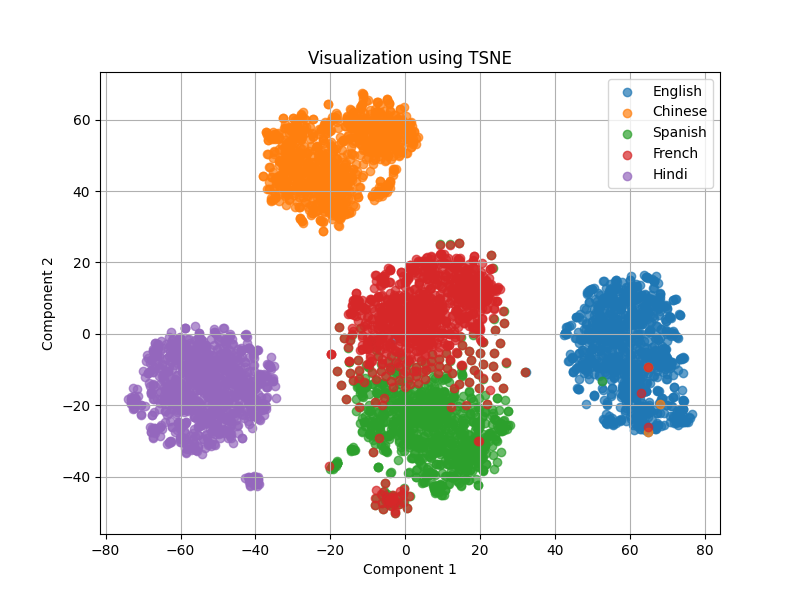}      \caption{T-SNE, Layer 21}        \end{subfigure}  \hfill  \begin{subfigure}{0.18\textwidth}      \includegraphics[width=\textwidth]{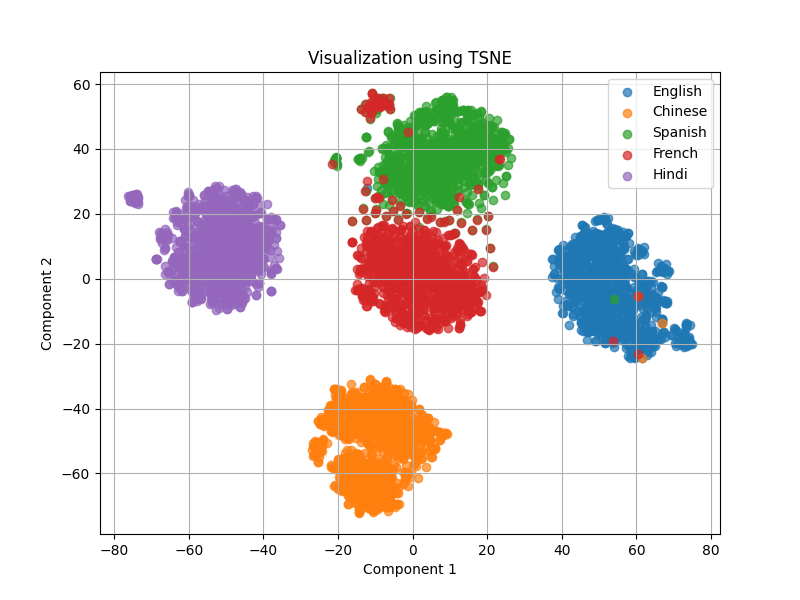}      \caption{T-SNE, Layer 22}        \end{subfigure}  \hfill  \begin{subfigure}{0.18\textwidth}      \includegraphics[width=\textwidth]{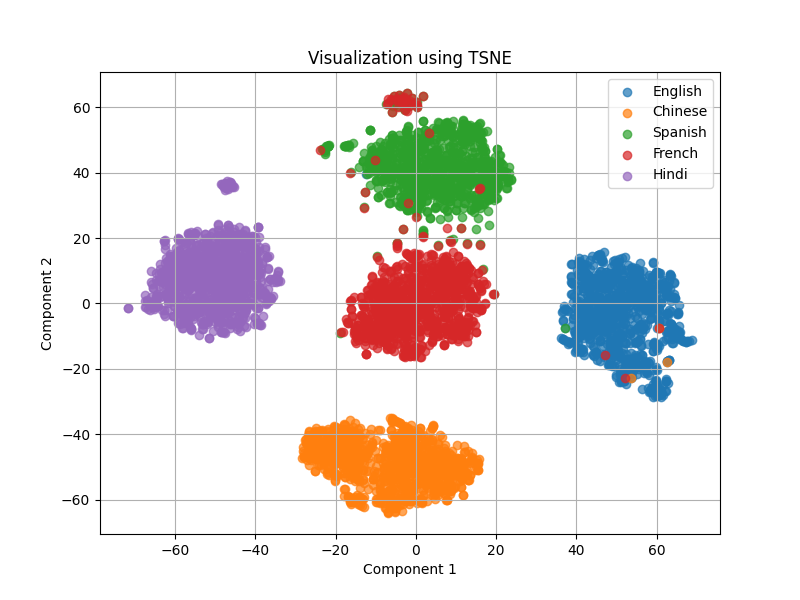}      \caption{T-SNE, Layer 23}        \end{subfigure}  \hfill  \begin{subfigure}{0.18\textwidth}      \includegraphics[width=\textwidth]{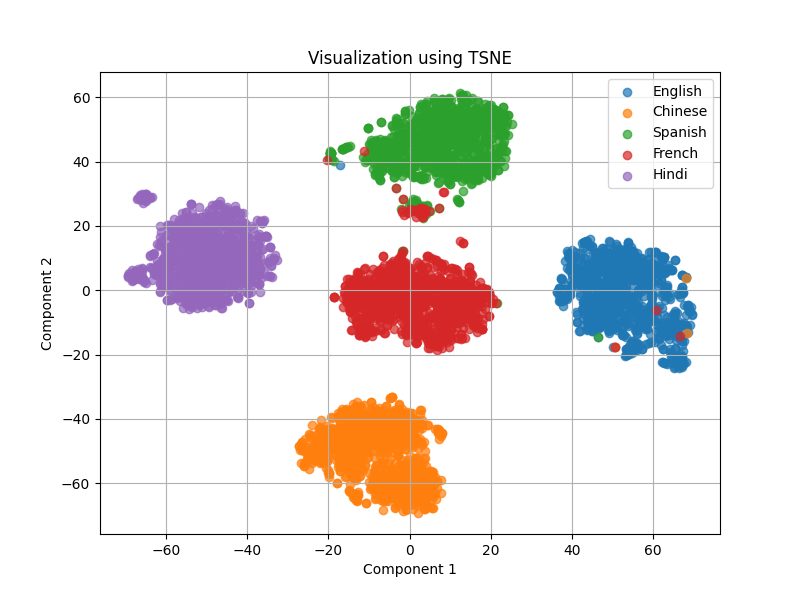}      \caption{T-SNE, Layer 24}        \end{subfigure}  \hfill  \begin{subfigure}{0.18\textwidth}      \includegraphics[width=\textwidth]{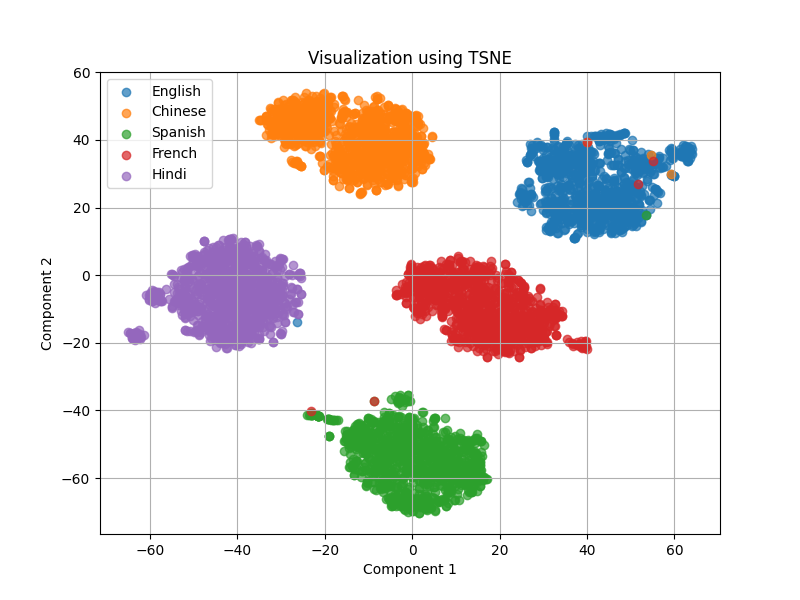}      \caption{T-SNE, Layer 25}        \end{subfigure}    \vspace{0.2in}    %
\begin{subfigure}{0.18\textwidth}      \includegraphics[width=\textwidth]{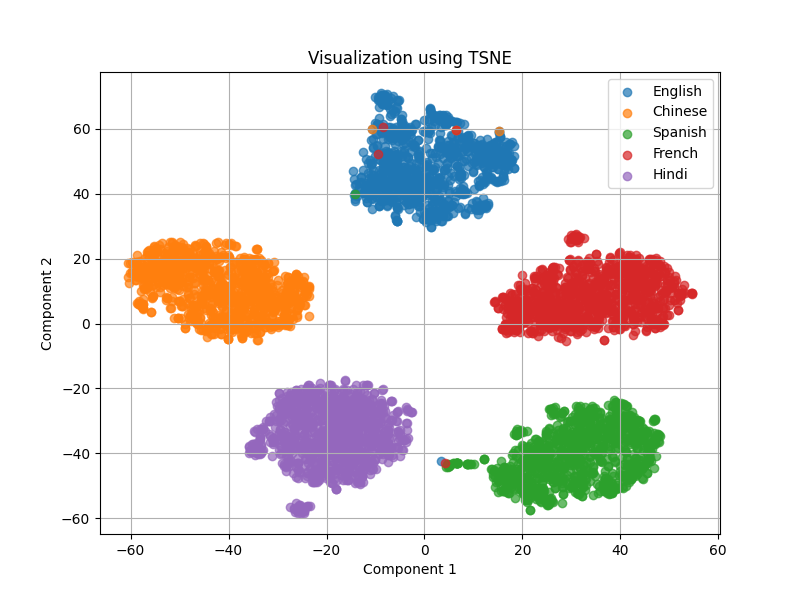}      \caption{T-SNE, Layer 26}        \end{subfigure}  \hfill  \begin{subfigure}{0.18\textwidth}      \includegraphics[width=\textwidth]{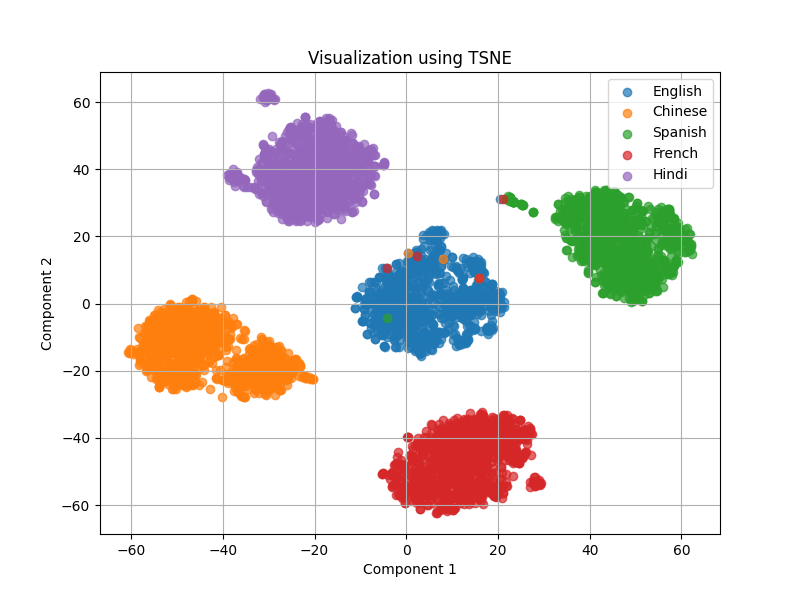}      \caption{T-SNE, Layer 27}        \end{subfigure}  \hfill  \begin{subfigure}{0.18\textwidth}      \includegraphics[width=\textwidth]{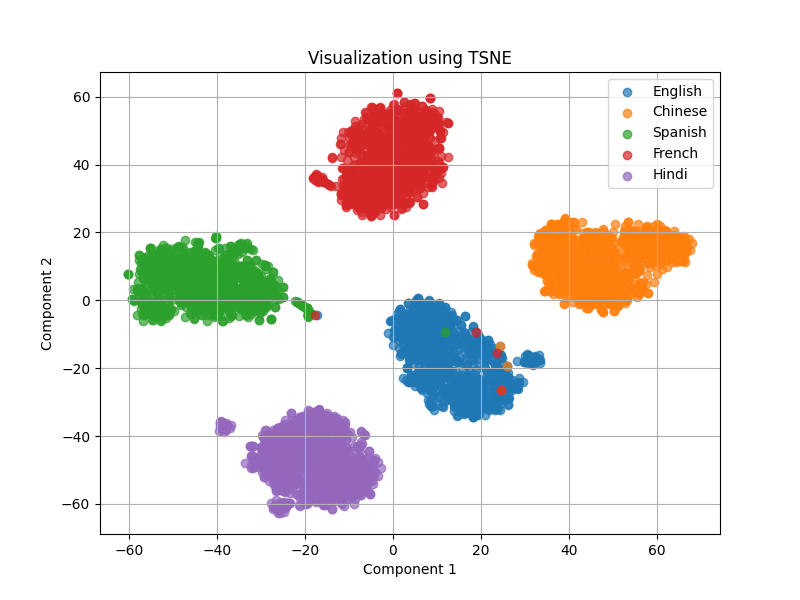}      \caption{T-SNE, Layer 28}        \end{subfigure}      \caption{T-SNE visualizations for layers 1-28 of Qwen2-7B-Instruct on the GSM8K dataset.}  
\end{figure*}

\begin{figure*}[htbp]
\centering
\begin{subfigure}{0.18\textwidth}
\includegraphics[width=\textwidth]{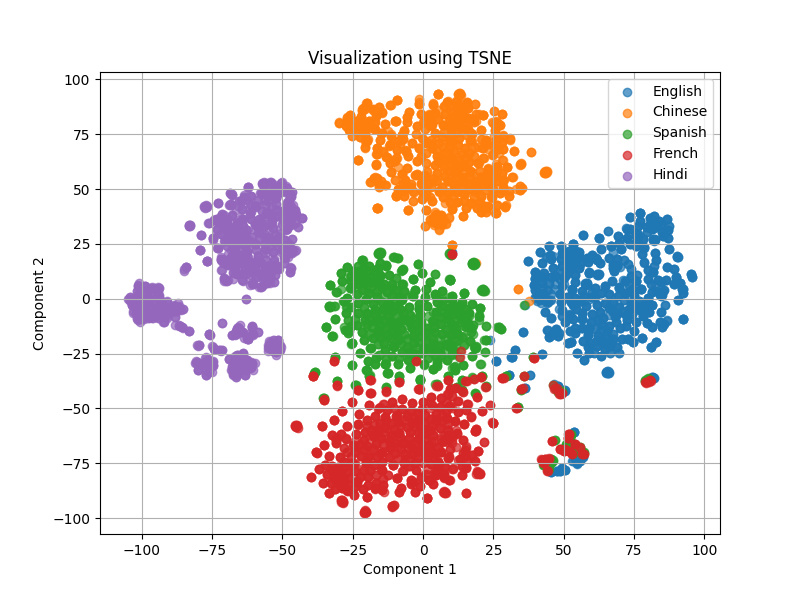}
\caption{T-SNE, Layer 1}
\end{subfigure}
\hfill
\begin{subfigure}{0.18\textwidth}
\includegraphics[width=\textwidth]{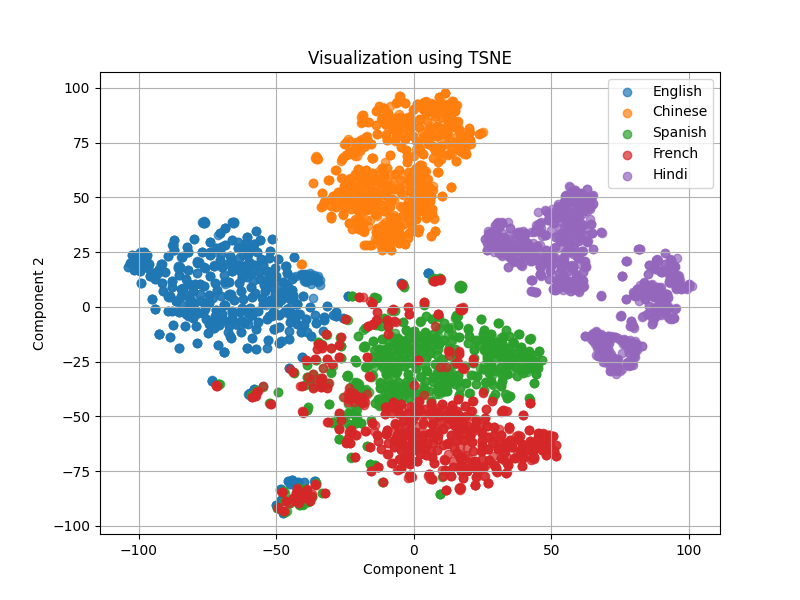}
\caption{T-SNE, Layer 2}

\end{subfigure}
\hfill
\begin{subfigure}{0.18\textwidth}
\includegraphics[width=\textwidth]{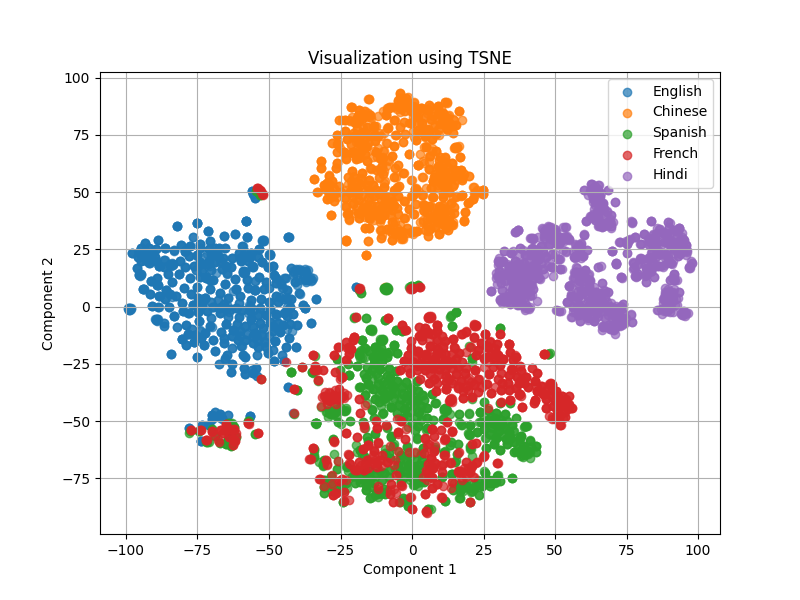}
\caption{T-SNE, Layer 3}

\end{subfigure}
\hfill
\begin{subfigure}{0.18\textwidth}
\includegraphics[width=\textwidth]{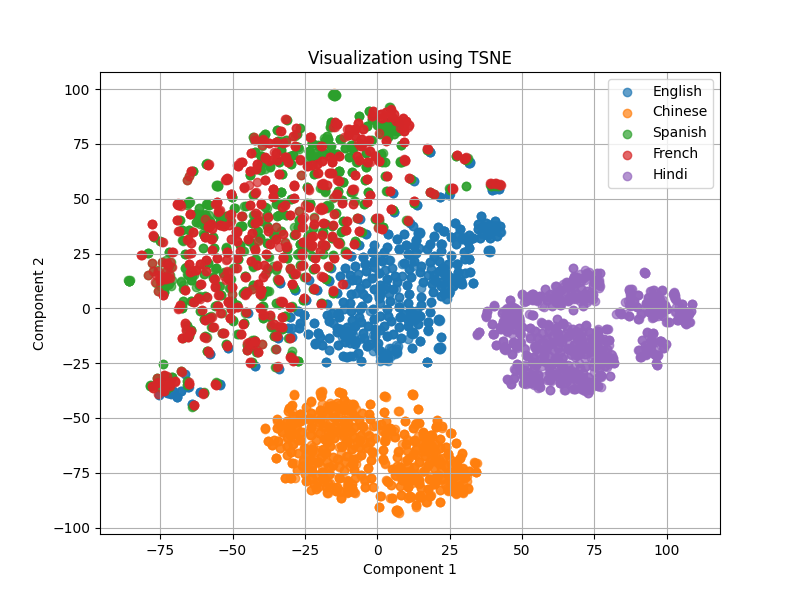}
\caption{T-SNE, Layer 4}

\end{subfigure}
\hfill
\begin{subfigure}{0.18\textwidth}
\includegraphics[width=\textwidth]{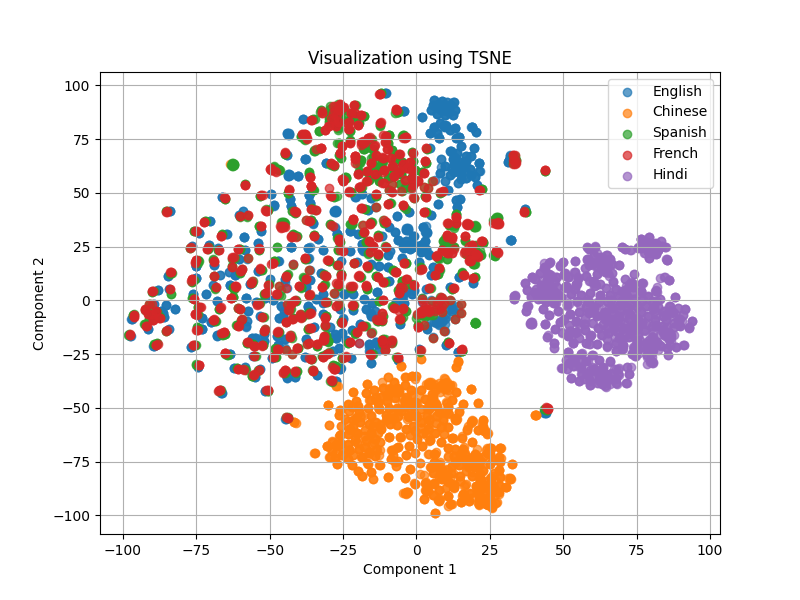}
\caption{T-SNE, Layer 5}

\end{subfigure}
\vspace{0.2in} %
\begin{subfigure}{0.18\textwidth}      \includegraphics[width=\textwidth]{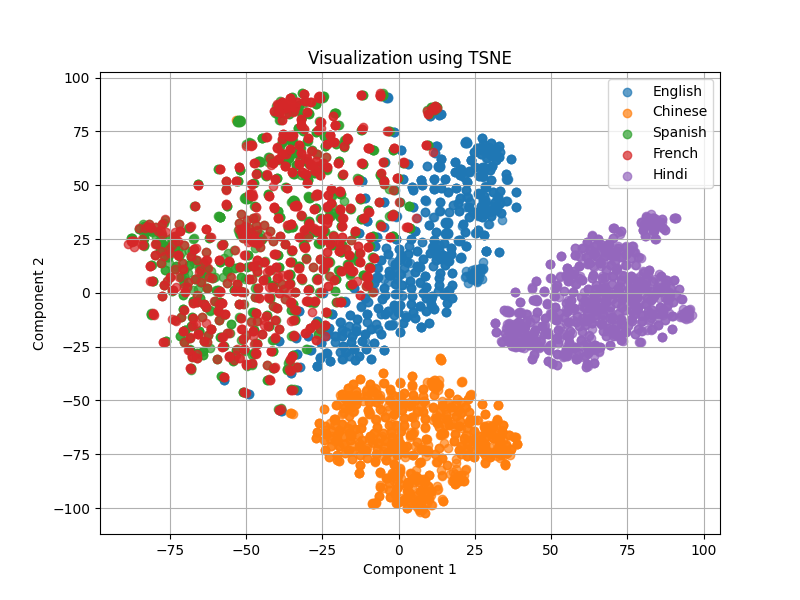}      \caption{T-SNE, Layer 6}        \end{subfigure}  \hfill  \begin{subfigure}{0.18\textwidth}      \includegraphics[width=\textwidth]{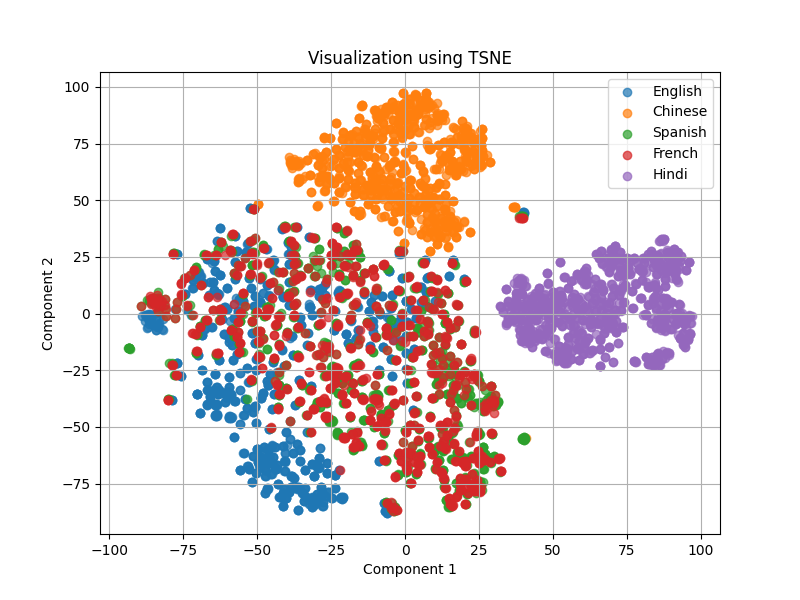}      \caption{T-SNE, Layer 7}        \end{subfigure}  \hfill  \begin{subfigure}{0.18\textwidth}      \includegraphics[width=\textwidth]{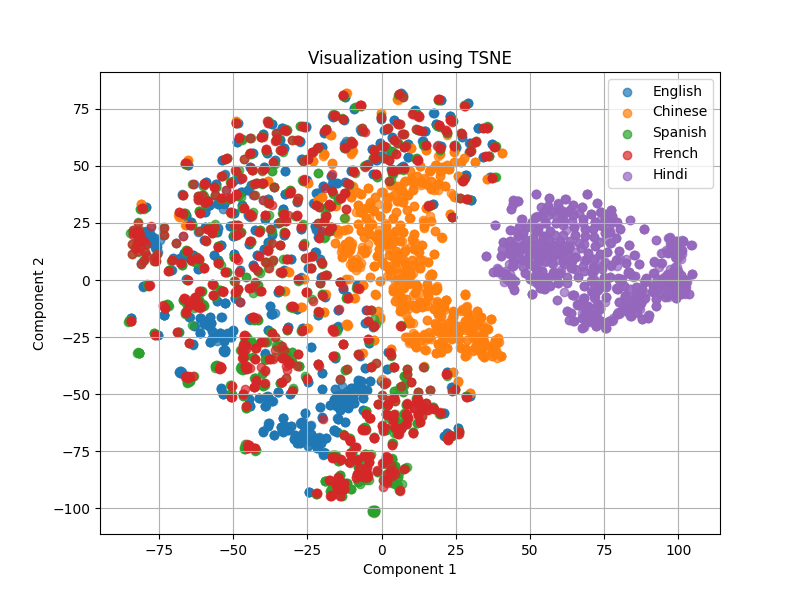}      \caption{T-SNE, Layer 8}        \end{subfigure}  \hfill  \begin{subfigure}{0.18\textwidth}      \includegraphics[width=\textwidth]{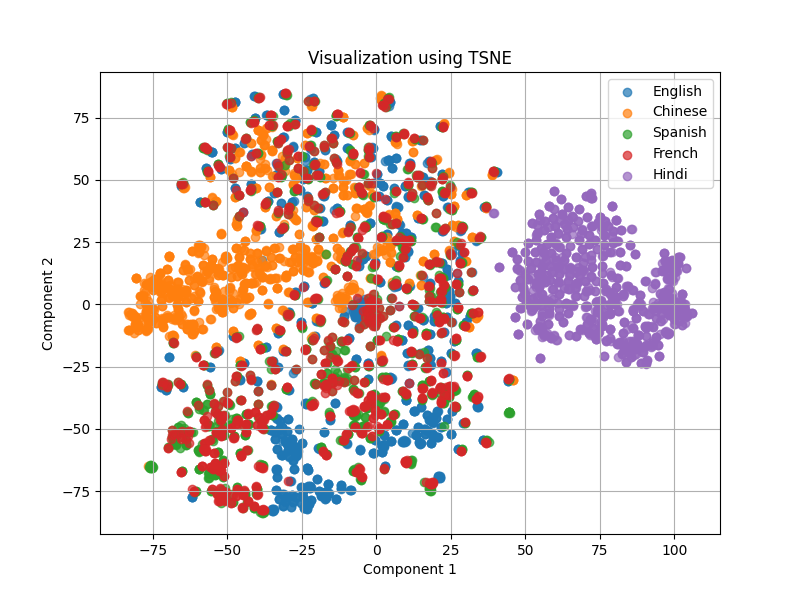}      \caption{T-SNE, Layer 9}        \end{subfigure}  \hfill  \begin{subfigure}{0.18\textwidth}      \includegraphics[width=\textwidth]{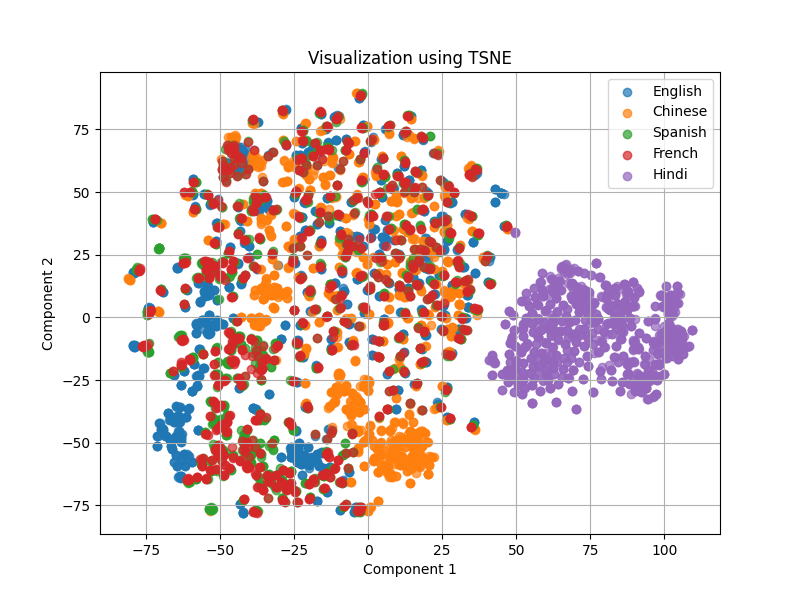}      \caption{T-SNE, Layer 10}        \end{subfigure}    \vspace{0.2in}    %
\begin{subfigure}{0.18\textwidth}      \includegraphics[width=\textwidth]{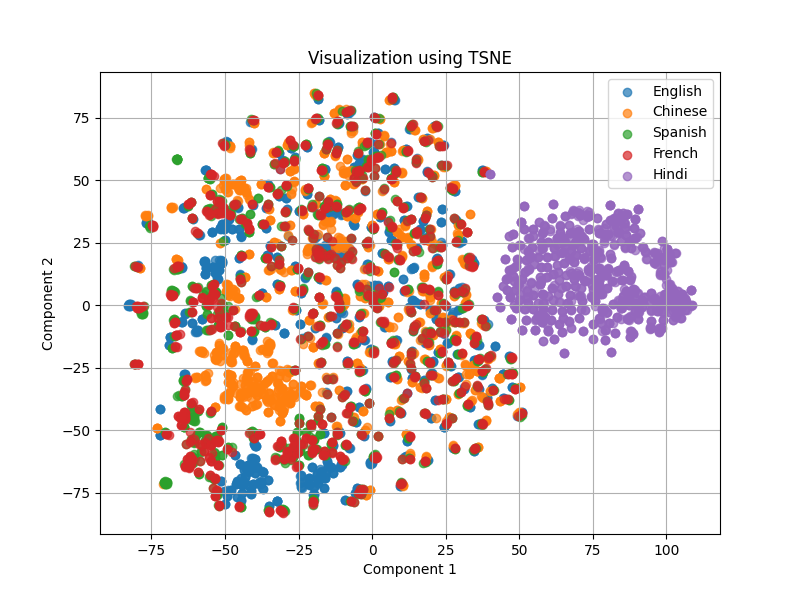}      \caption{T-SNE, Layer 11}        \end{subfigure}  \hfill  \begin{subfigure}{0.18\textwidth}      \includegraphics[width=\textwidth]{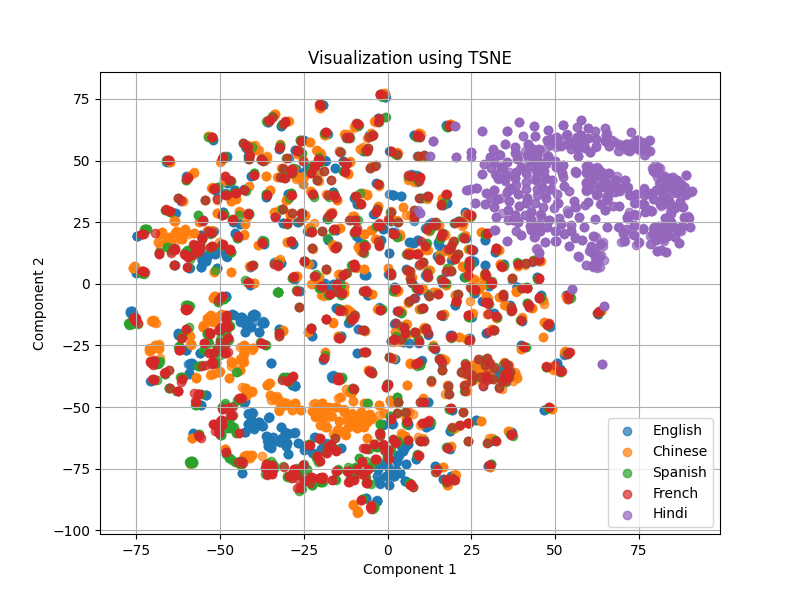}      \caption{T-SNE, Layer 12}        \end{subfigure}  \hfill  \begin{subfigure}{0.18\textwidth}      \includegraphics[width=\textwidth]{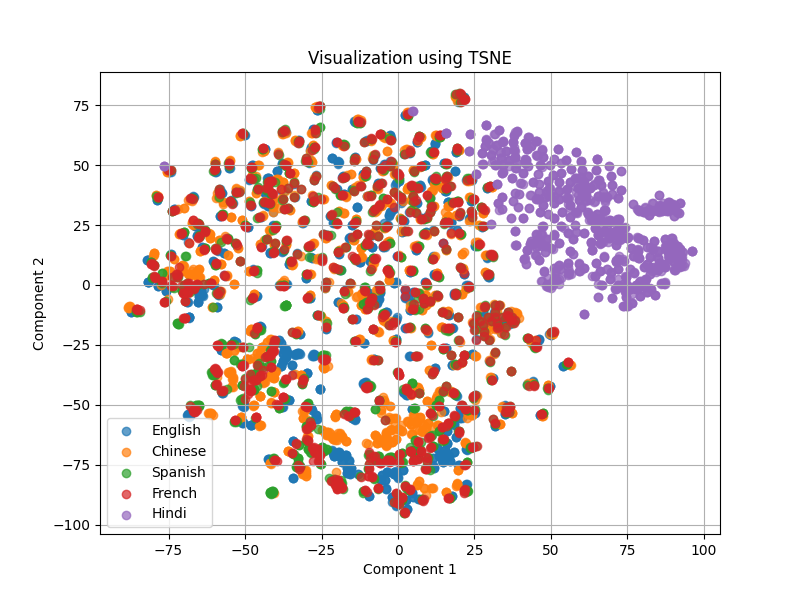}      \caption{T-SNE, Layer 13}        \end{subfigure}  \hfill  \begin{subfigure}{0.18\textwidth}      \includegraphics[width=\textwidth]{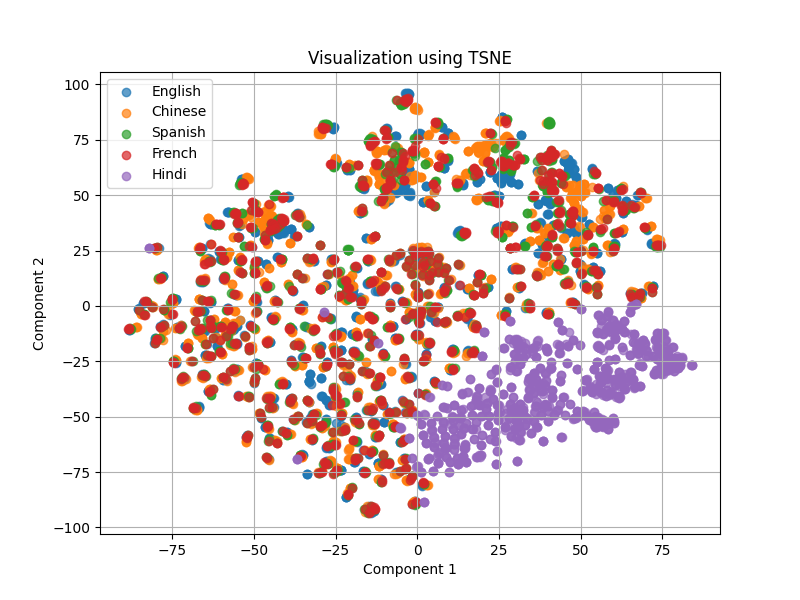}      \caption{T-SNE, Layer 14}        \end{subfigure}  \hfill  \begin{subfigure}{0.18\textwidth}      \includegraphics[width=\textwidth]{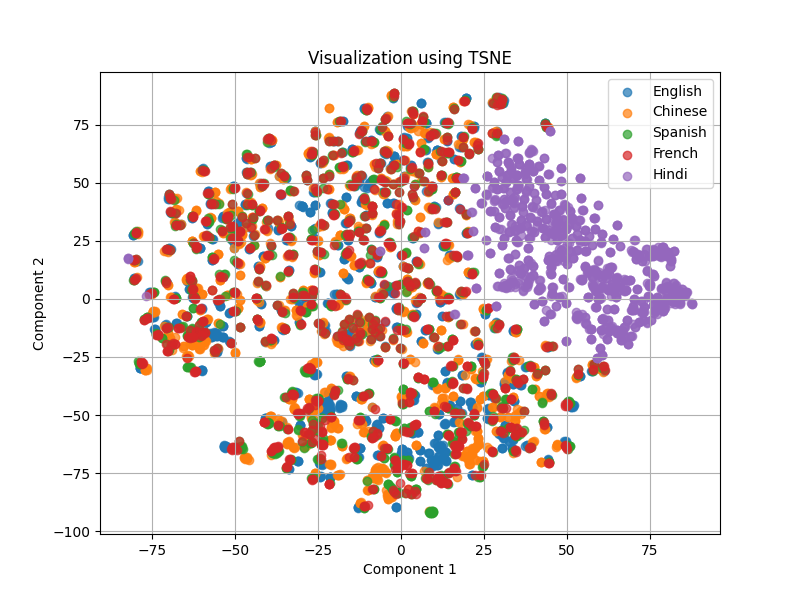}      \caption{T-SNE, Layer 15}        \end{subfigure}    \vspace{0.2in}    %
\begin{subfigure}{0.18\textwidth}      \includegraphics[width=\textwidth]{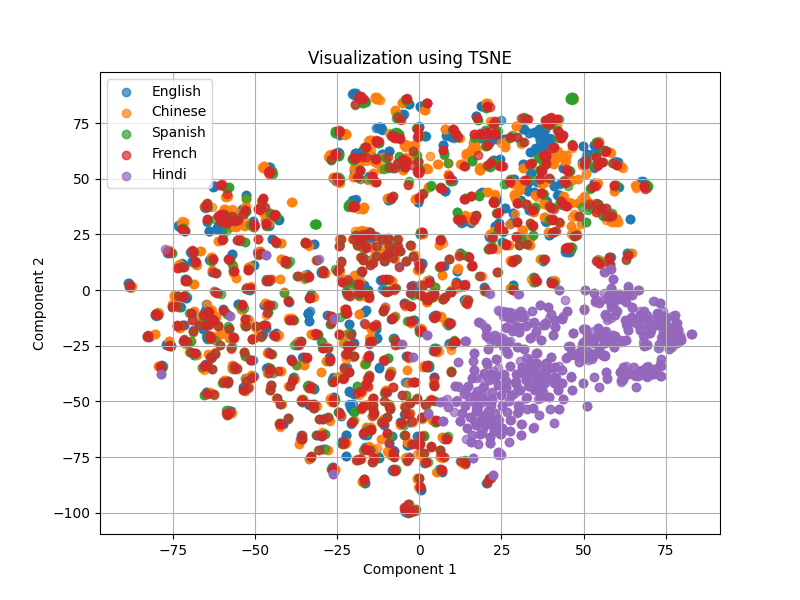}      \caption{T-SNE, Layer 16}        \end{subfigure}  \hfill  \begin{subfigure}{0.18\textwidth}      \includegraphics[width=\textwidth]{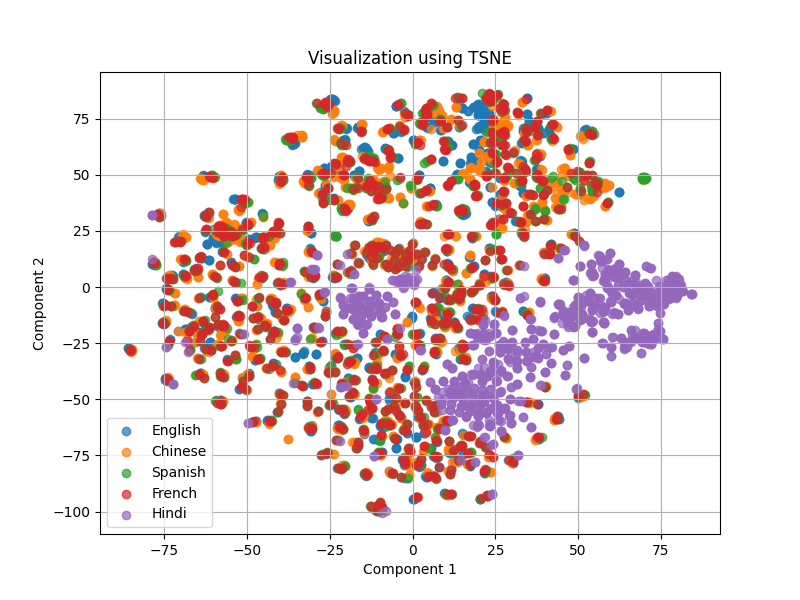}      \caption{T-SNE, Layer 17}        \end{subfigure}  \hfill  \begin{subfigure}{0.18\textwidth}      \includegraphics[width=\textwidth]{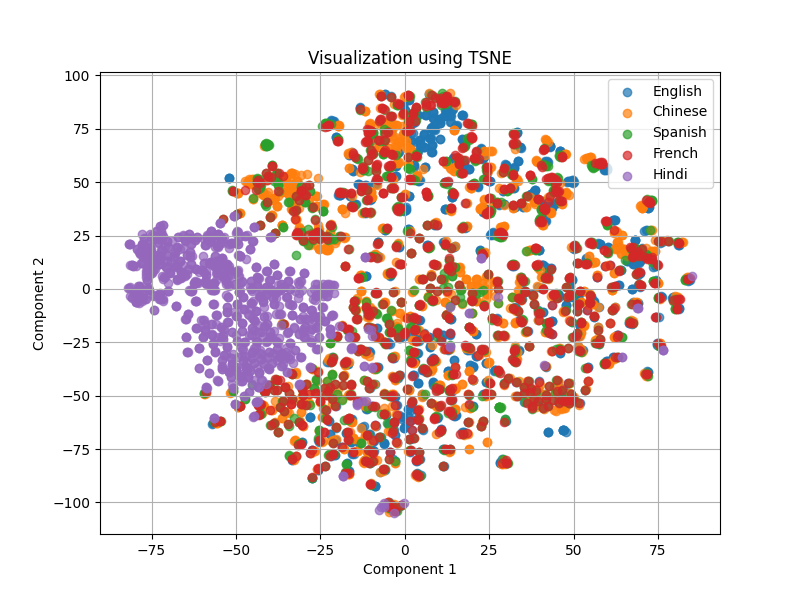}      \caption{T-SNE, Layer 18}        \end{subfigure}  \hfill  \begin{subfigure}{0.18\textwidth}      \includegraphics[width=\textwidth]{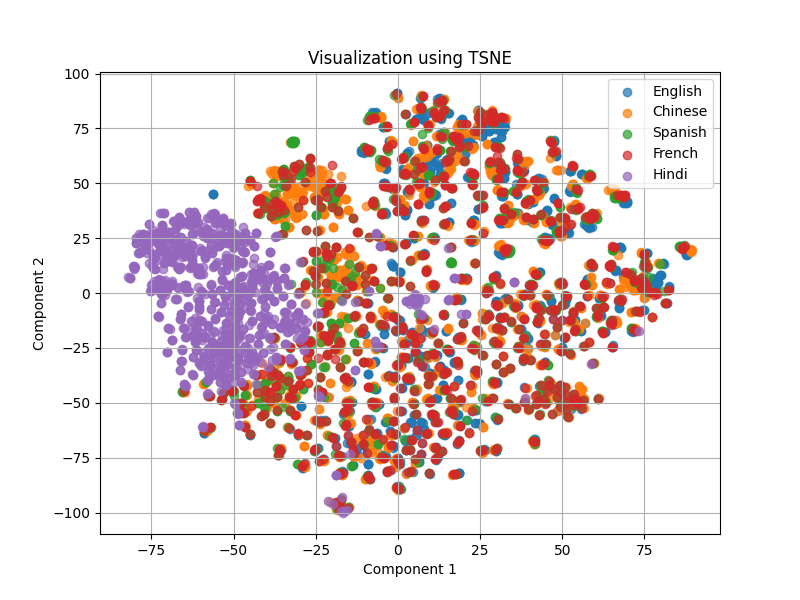}      \caption{T-SNE, Layer 19}        \end{subfigure}  \hfill  \begin{subfigure}{0.18\textwidth}      \includegraphics[width=\textwidth]{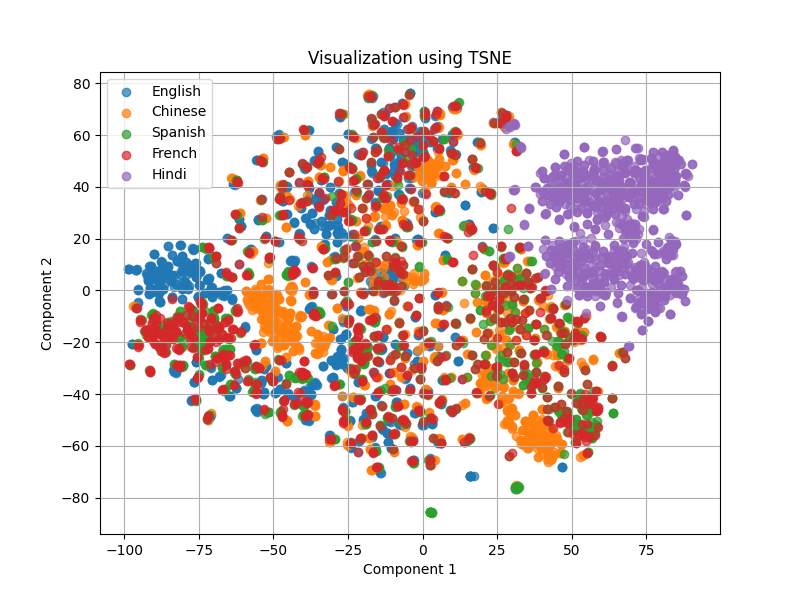}      \caption{T-SNE, Layer 20}        \end{subfigure}    \vspace{0.2in}    %
\begin{subfigure}{0.18\textwidth}      \includegraphics[width=\textwidth]{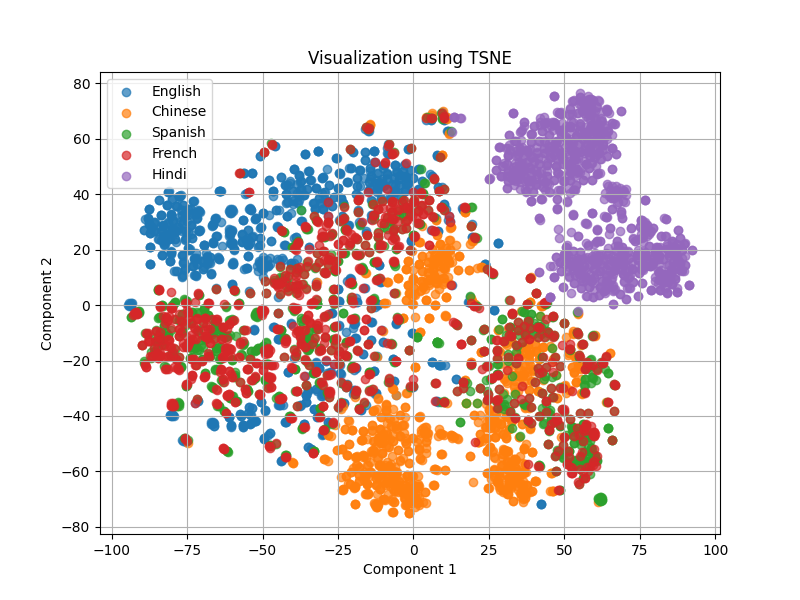}      \caption{T-SNE, Layer 21}        \end{subfigure}  \hfill  \begin{subfigure}{0.18\textwidth}      \includegraphics[width=\textwidth]{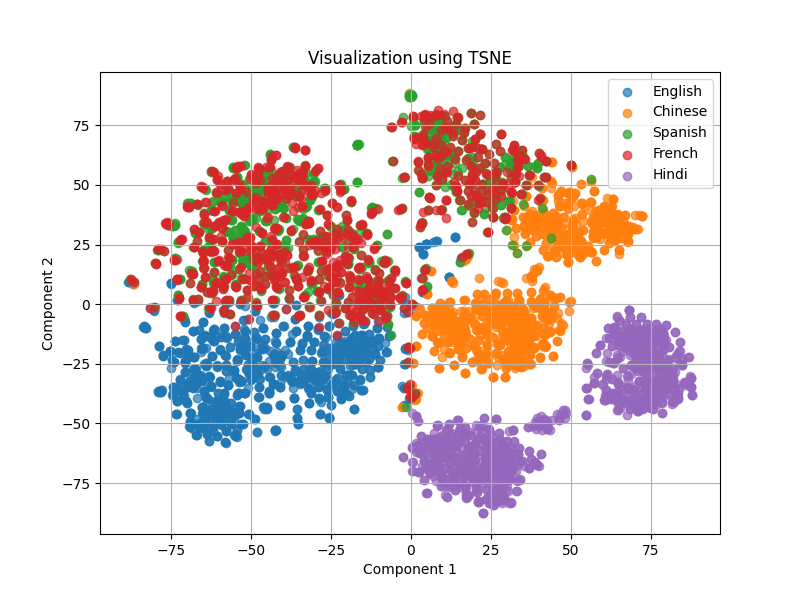}      \caption{T-SNE, Layer 22}        \end{subfigure}  \hfill  \begin{subfigure}{0.18\textwidth}      \includegraphics[width=\textwidth]{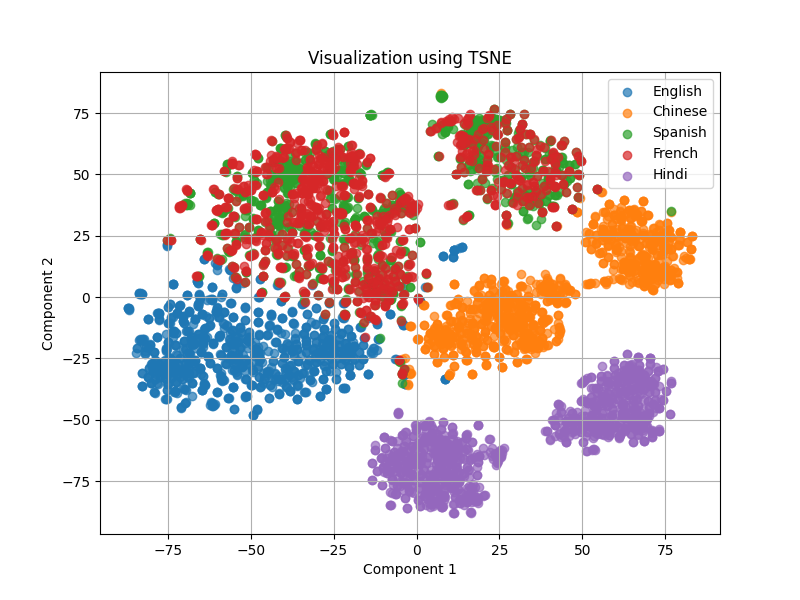}      \caption{T-SNE, Layer 23}        \end{subfigure}  \hfill  \begin{subfigure}{0.18\textwidth}      \includegraphics[width=\textwidth]{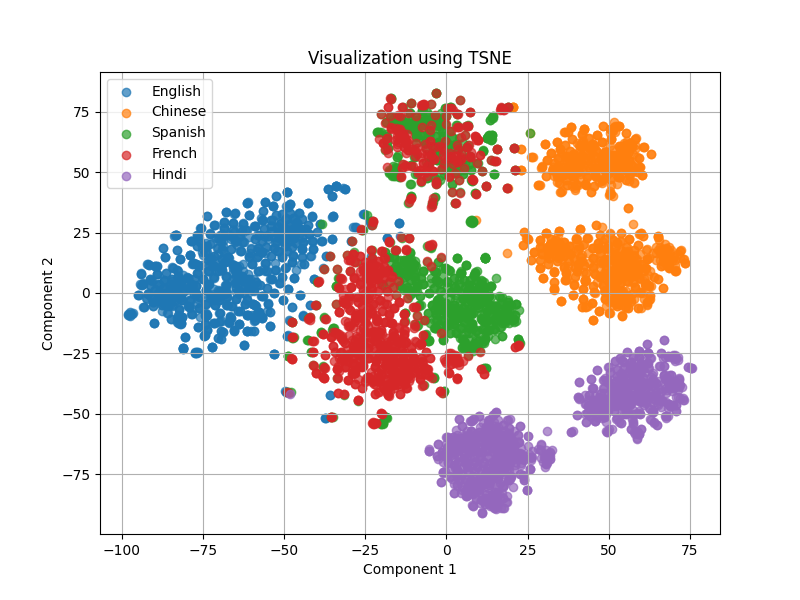}      \caption{T-SNE, Layer 24}        \end{subfigure}  \hfill  \begin{subfigure}{0.18\textwidth}      \includegraphics[width=\textwidth]{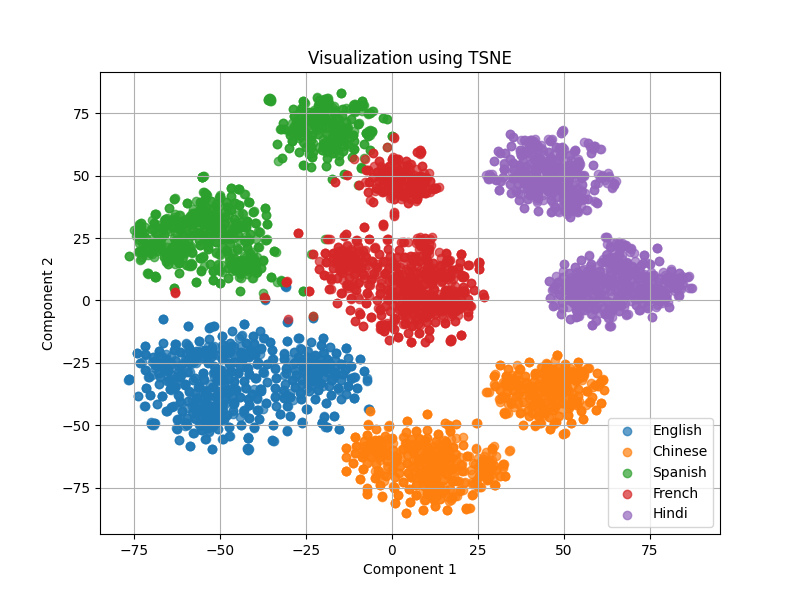}      \caption{T-SNE, Layer 25}        \end{subfigure}    \vspace{0.2in}    %
\begin{subfigure}{0.18\textwidth}      \includegraphics[width=\textwidth]{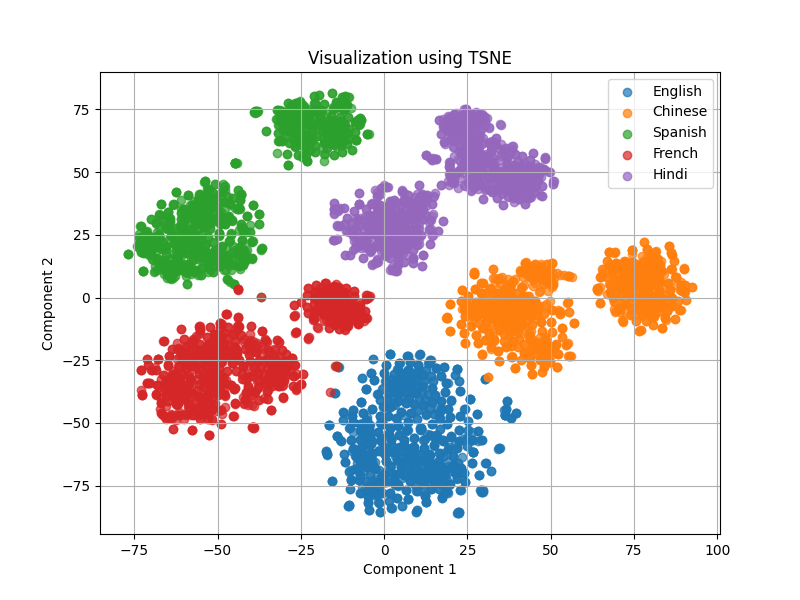}      \caption{T-SNE, Layer 26}        \end{subfigure}  \hfill  \begin{subfigure}{0.18\textwidth}      \includegraphics[width=\textwidth]{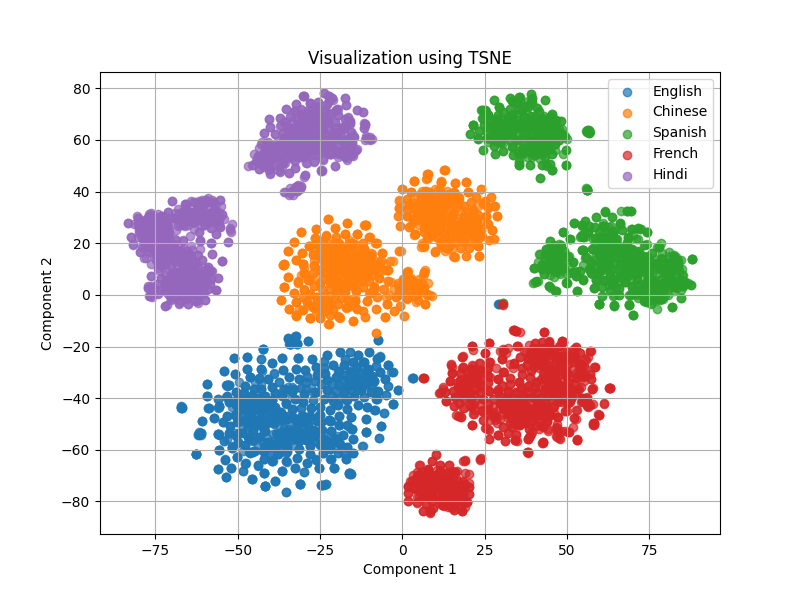}      \caption{T-SNE, Layer 27}        \end{subfigure}  \hfill  \begin{subfigure}{0.18\textwidth}      \includegraphics[width=\textwidth]{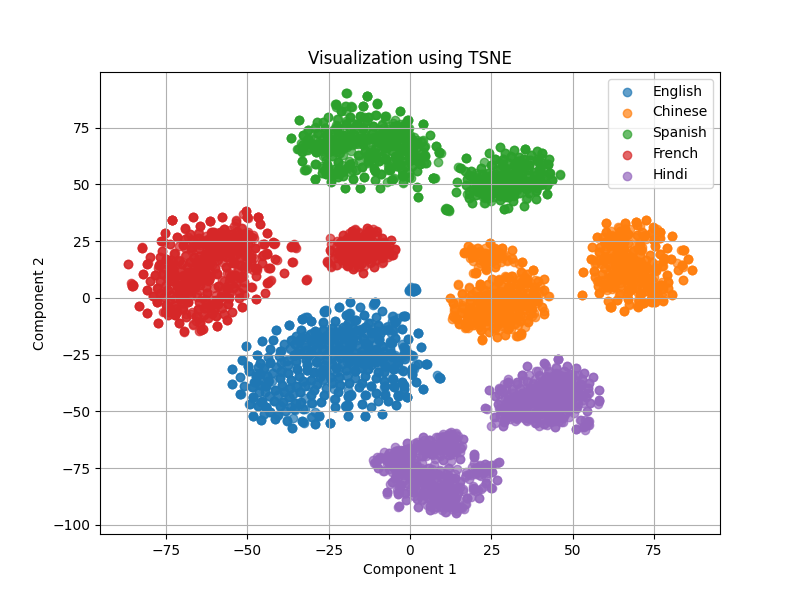}      \caption{T-SNE, Layer 28}        \end{subfigure}      \caption{T-SNE visualizations for layers 1-28 of Qwen2-7B-Instruct on the FOLIO dataset.}  
\end{figure*}

\begin{figure*}[htbp]
\centering
\begin{subfigure}{0.18\textwidth}
\includegraphics[width=\textwidth]{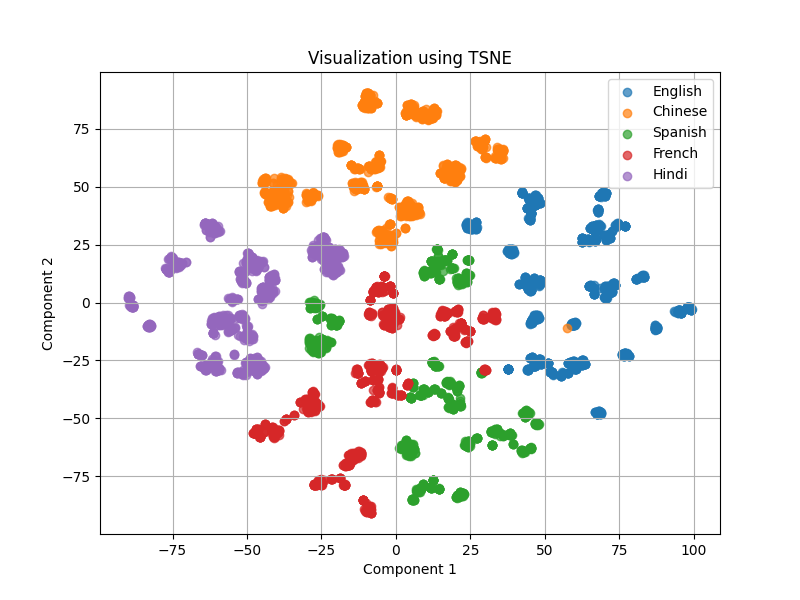}
\caption{T-SNE, Layer 1}
\end{subfigure}
\hfill
\begin{subfigure}{0.18\textwidth}
\includegraphics[width=\textwidth]{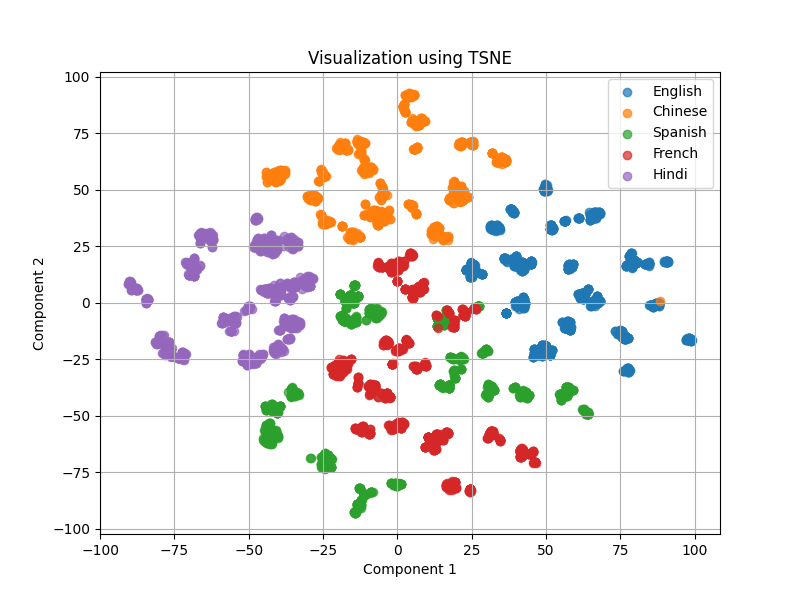}
\caption{T-SNE, Layer 2}

\end{subfigure}
\hfill
\begin{subfigure}{0.18\textwidth}
\includegraphics[width=\textwidth]{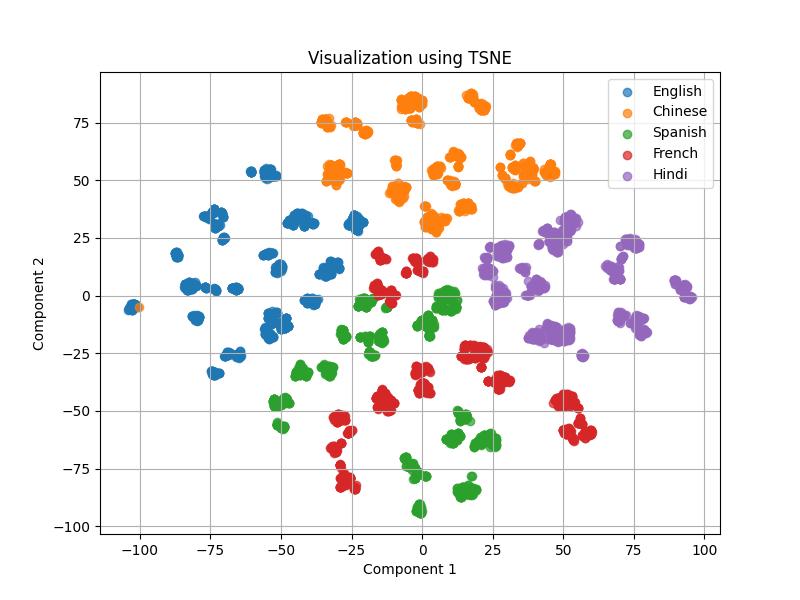}
\caption{T-SNE, Layer 3}

\end{subfigure}
\hfill
\begin{subfigure}{0.18\textwidth}
\includegraphics[width=\textwidth]{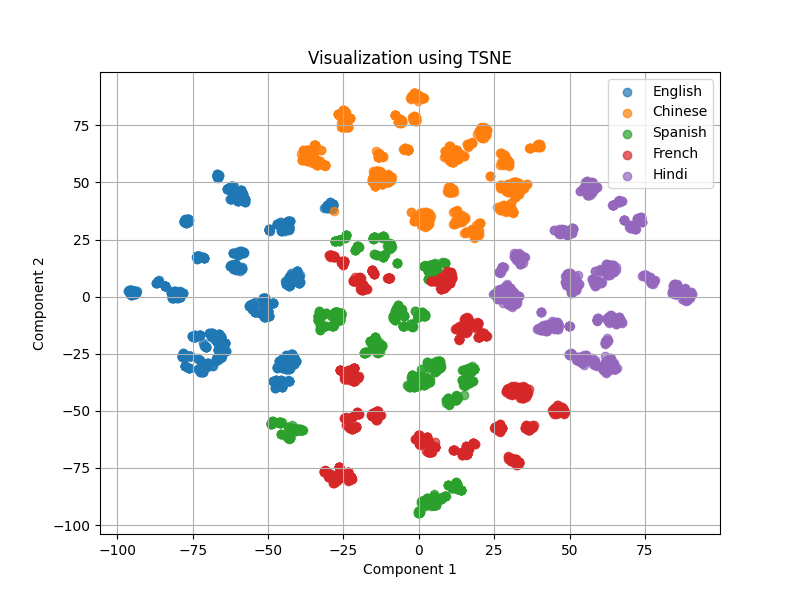}
\caption{T-SNE, Layer 4}

\end{subfigure}
\hfill
\begin{subfigure}{0.18\textwidth}
\includegraphics[width=\textwidth]{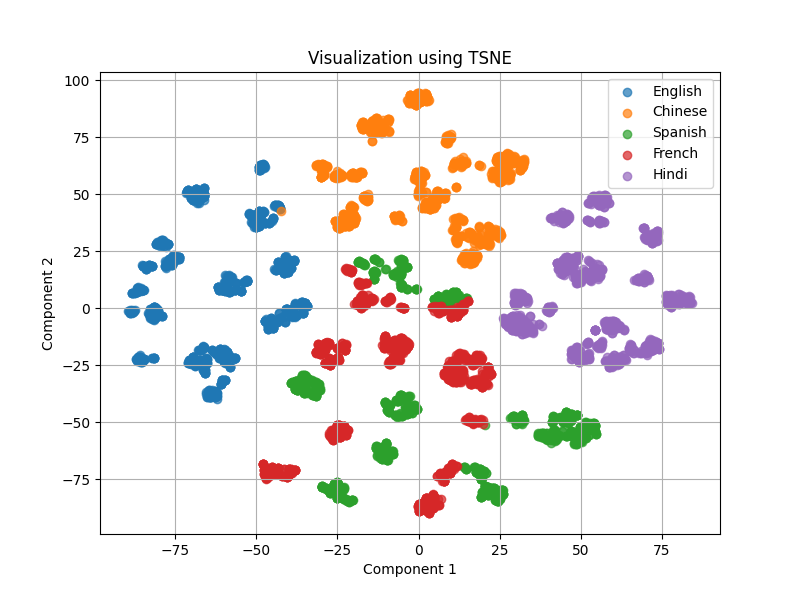}
\caption{T-SNE, Layer 5}

\end{subfigure}
\vspace{0.2in} %
\begin{subfigure}{0.18\textwidth}      \includegraphics[width=\textwidth]{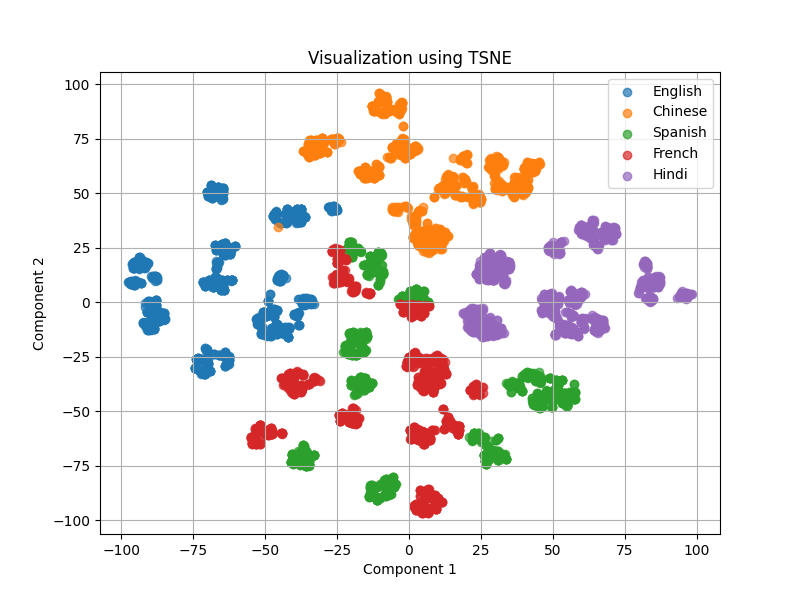}      \caption{T-SNE, Layer 6}        \end{subfigure}  \hfill  \begin{subfigure}{0.18\textwidth}      \includegraphics[width=\textwidth]{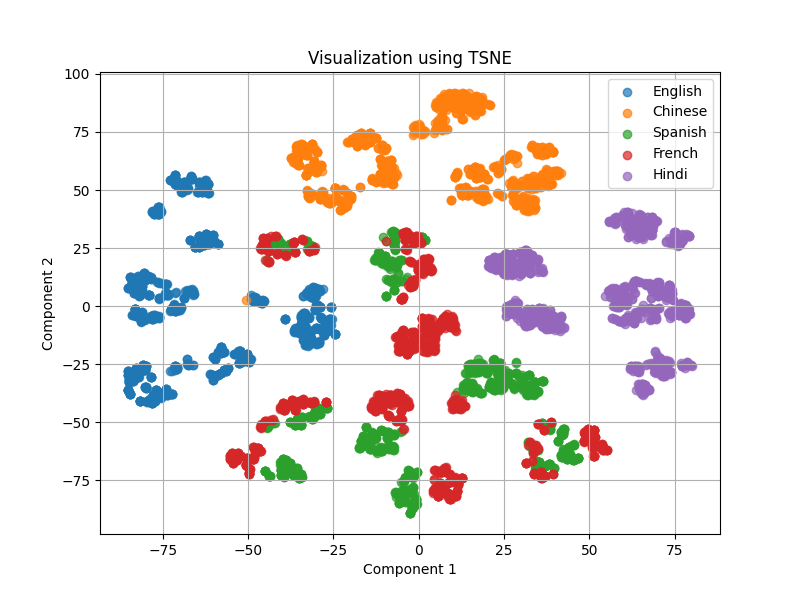}      \caption{T-SNE, Layer 7}        \end{subfigure}  \hfill  \begin{subfigure}{0.18\textwidth}      \includegraphics[width=\textwidth]{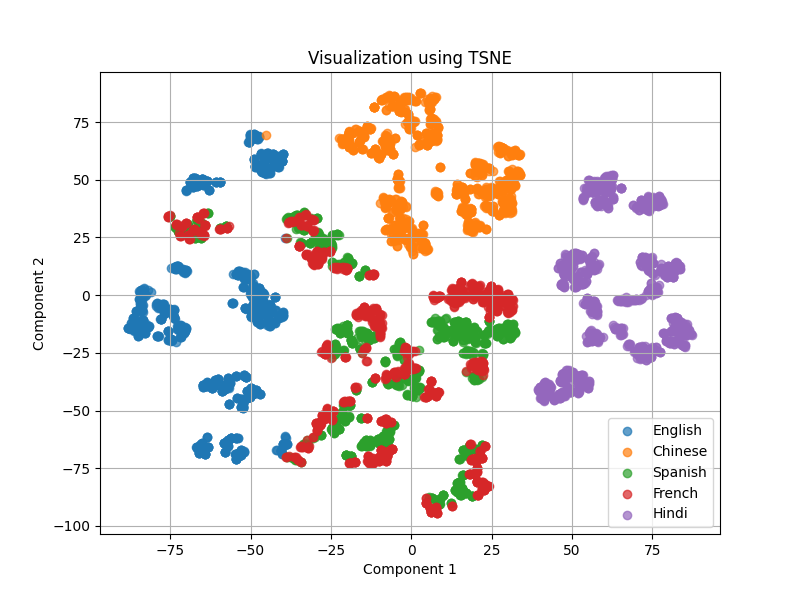}      \caption{T-SNE, Layer 8}        \end{subfigure}  \hfill  \begin{subfigure}{0.18\textwidth}      \includegraphics[width=\textwidth]{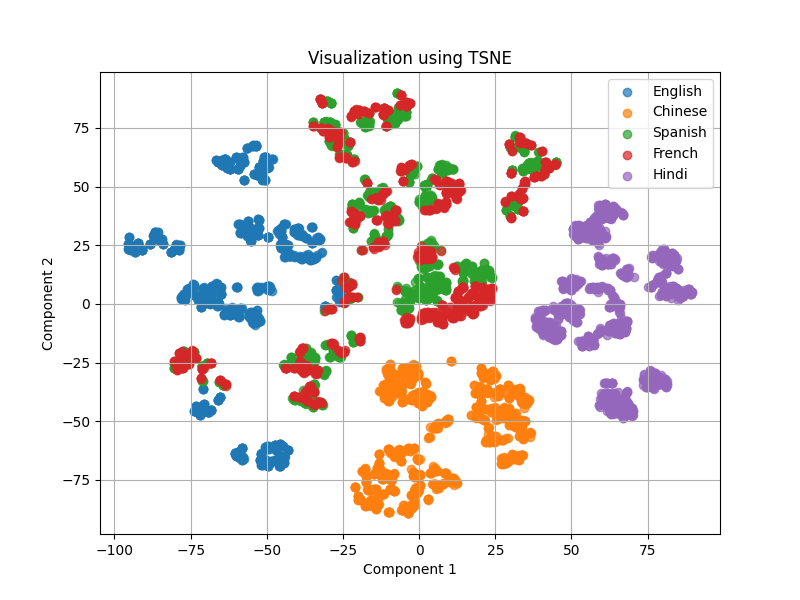}      \caption{T-SNE, Layer 9}        \end{subfigure}  \hfill  \begin{subfigure}{0.18\textwidth}      \includegraphics[width=\textwidth]{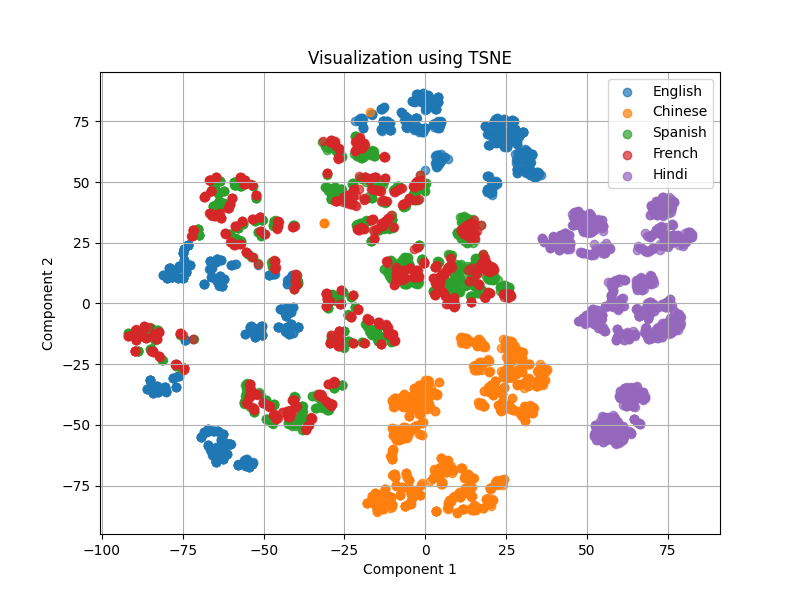}      \caption{T-SNE, Layer 10}        \end{subfigure}    \vspace{0.2in}    %
\begin{subfigure}{0.18\textwidth}      \includegraphics[width=\textwidth]{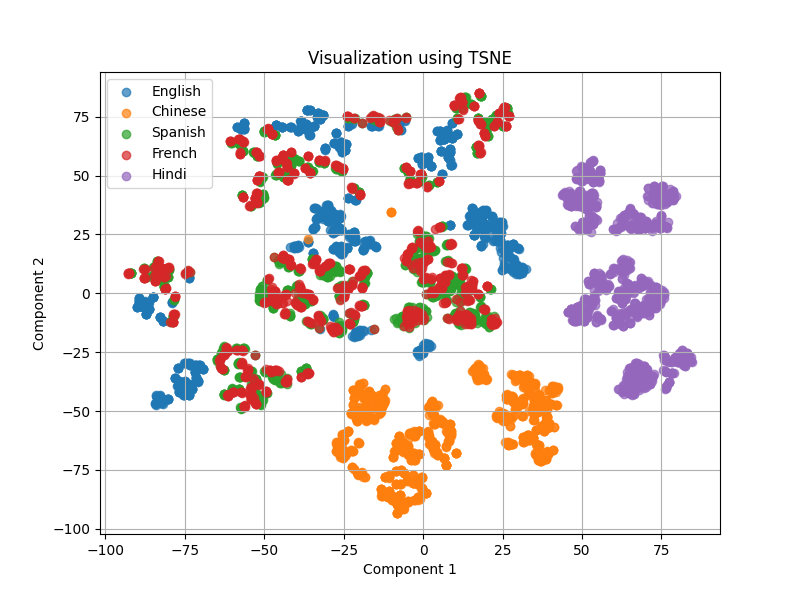}      \caption{T-SNE, Layer 11}        \end{subfigure}  \hfill  \begin{subfigure}{0.18\textwidth}      \includegraphics[width=\textwidth]{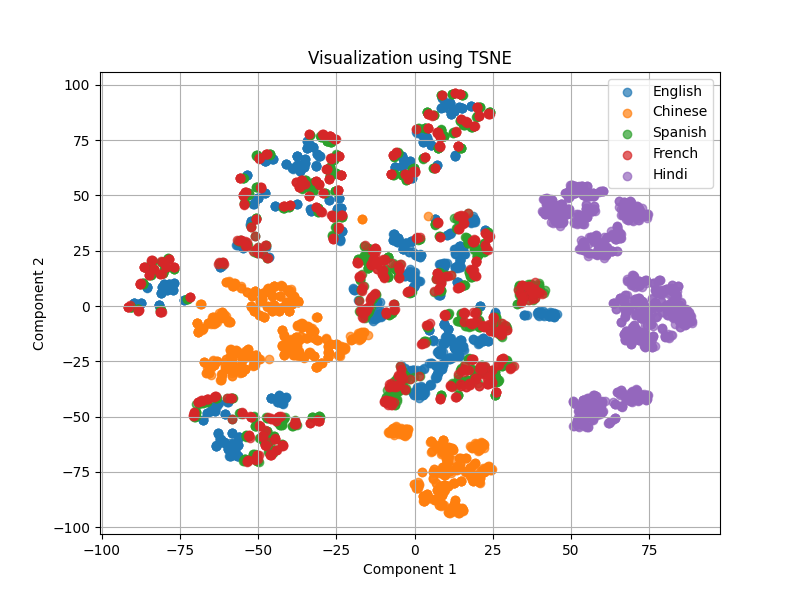}      \caption{T-SNE, Layer 12}        \end{subfigure}  \hfill  \begin{subfigure}{0.18\textwidth}      \includegraphics[width=\textwidth]{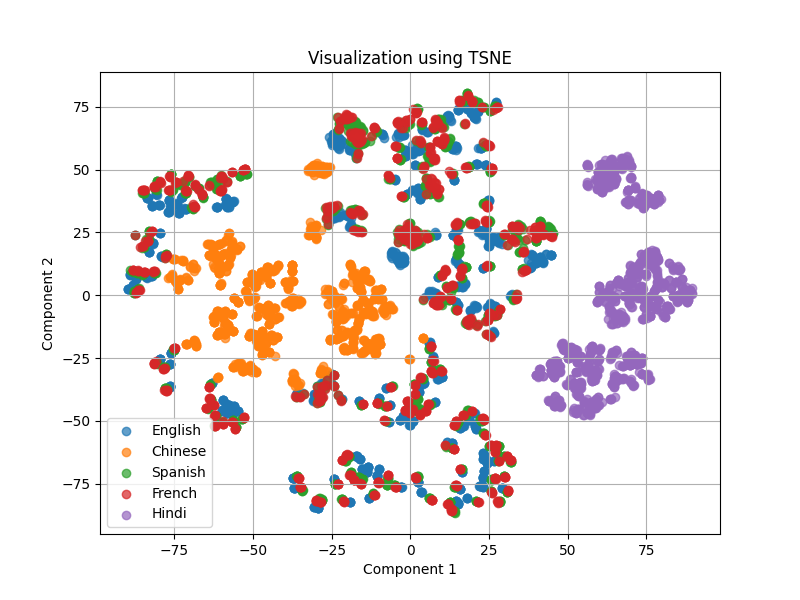}      \caption{T-SNE, Layer 13}        \end{subfigure}  \hfill  \begin{subfigure}{0.18\textwidth}      \includegraphics[width=\textwidth]{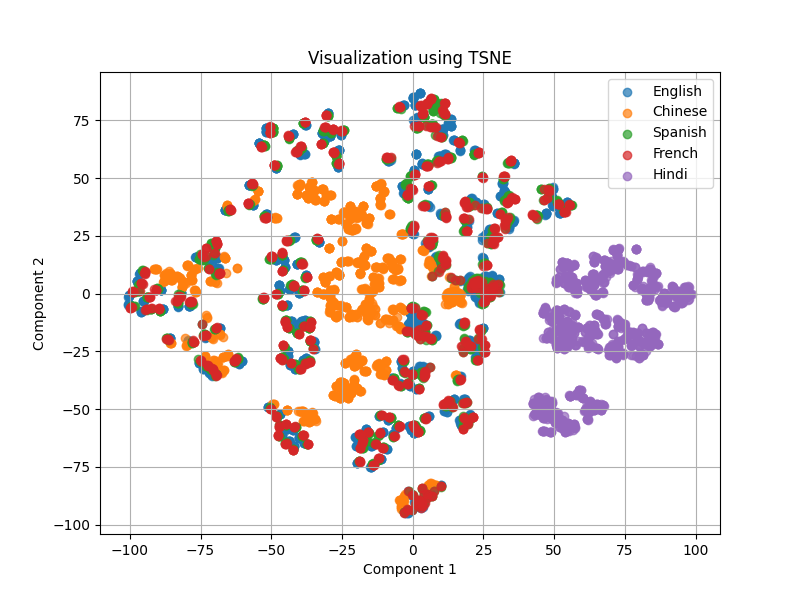}      \caption{T-SNE, Layer 14}        \end{subfigure}  \hfill  \begin{subfigure}{0.18\textwidth}      \includegraphics[width=\textwidth]{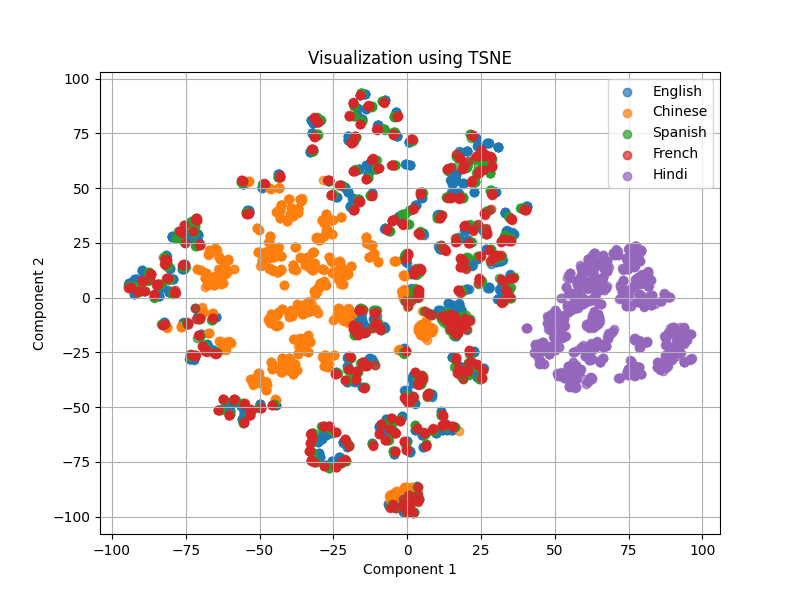}      \caption{T-SNE, Layer 15}        \end{subfigure}    \vspace{0.2in}    %
\begin{subfigure}{0.18\textwidth}      \includegraphics[width=\textwidth]{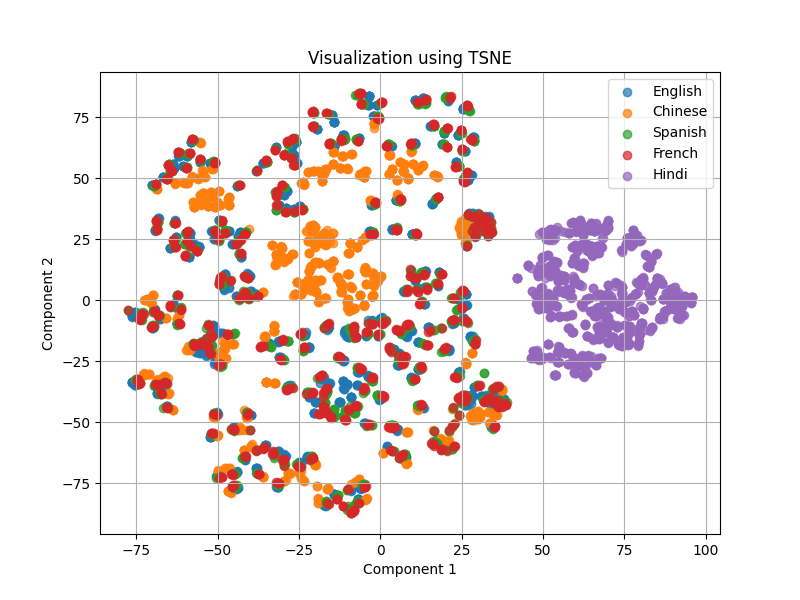}      \caption{T-SNE, Layer 16}        \end{subfigure}  \hfill  \begin{subfigure}{0.18\textwidth}      \includegraphics[width=\textwidth]{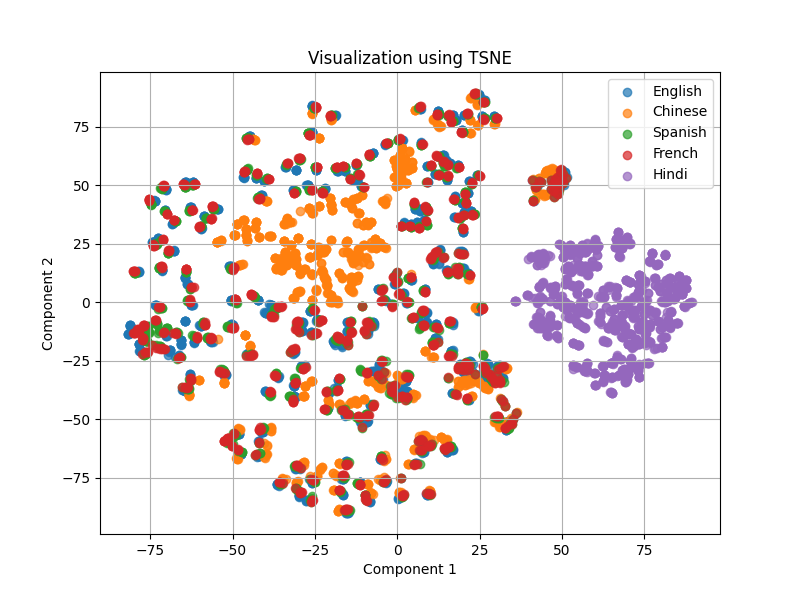}      \caption{T-SNE, Layer 17}        \end{subfigure}  \hfill  \begin{subfigure}{0.18\textwidth}      \includegraphics[width=\textwidth]{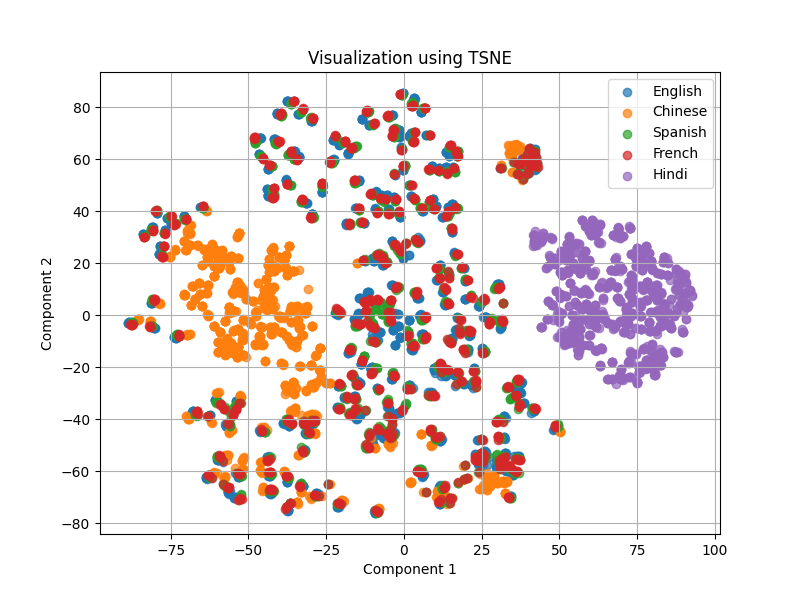}      \caption{T-SNE, Layer 18}        \end{subfigure}  \hfill  \begin{subfigure}{0.18\textwidth}      \includegraphics[width=\textwidth]{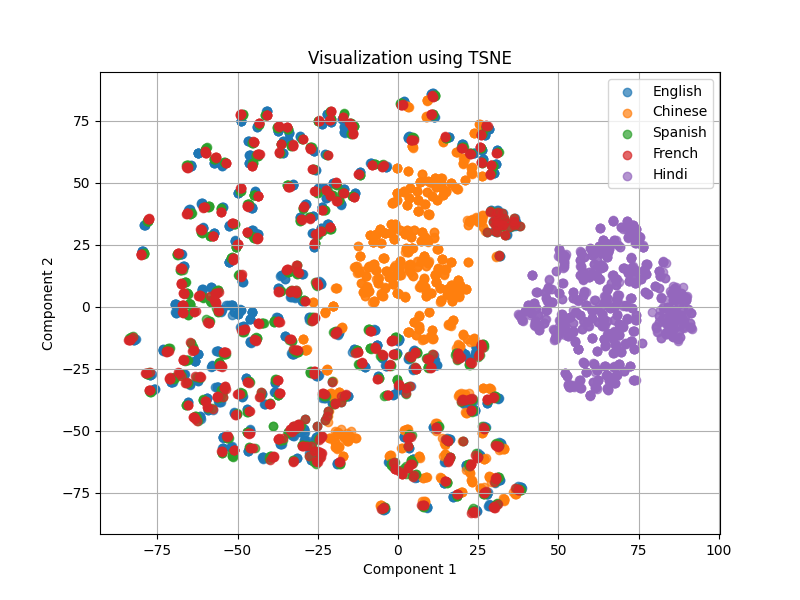}      \caption{T-SNE, Layer 19}        \end{subfigure}  \hfill  \begin{subfigure}{0.18\textwidth}      \includegraphics[width=\textwidth]{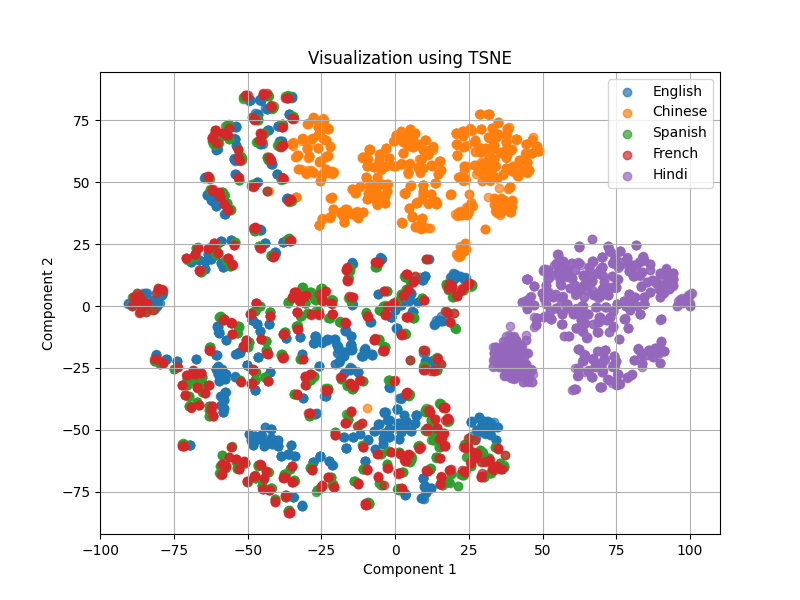}      \caption{T-SNE, Layer 20}        \end{subfigure}    \vspace{0.2in}    %
\begin{subfigure}{0.18\textwidth}      \includegraphics[width=\textwidth]{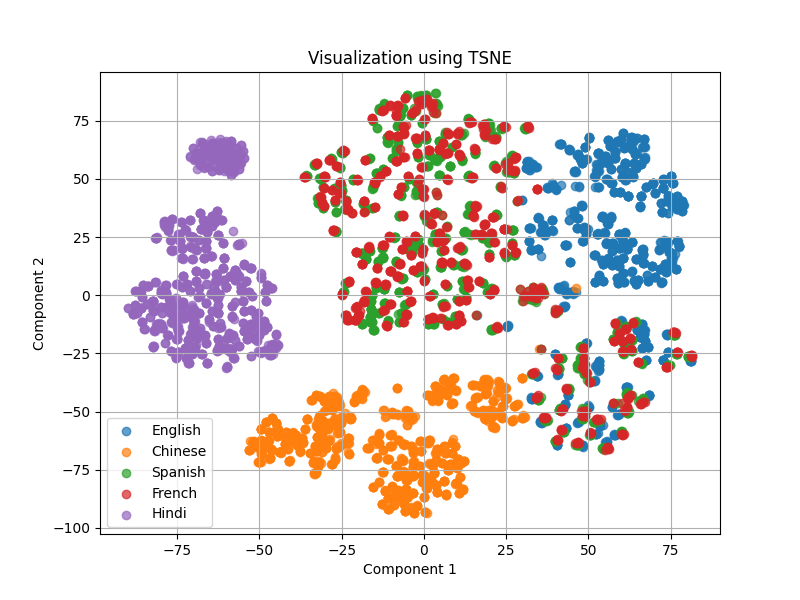}      \caption{T-SNE, Layer 21}        \end{subfigure}  \hfill  \begin{subfigure}{0.18\textwidth}      \includegraphics[width=\textwidth]{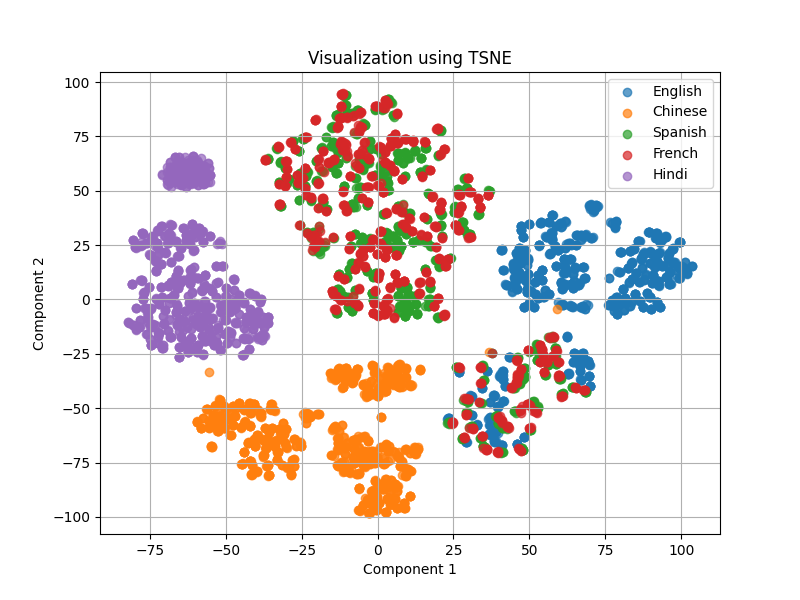}      \caption{T-SNE, Layer 22}        \end{subfigure}  \hfill  \begin{subfigure}{0.18\textwidth}      \includegraphics[width=\textwidth]{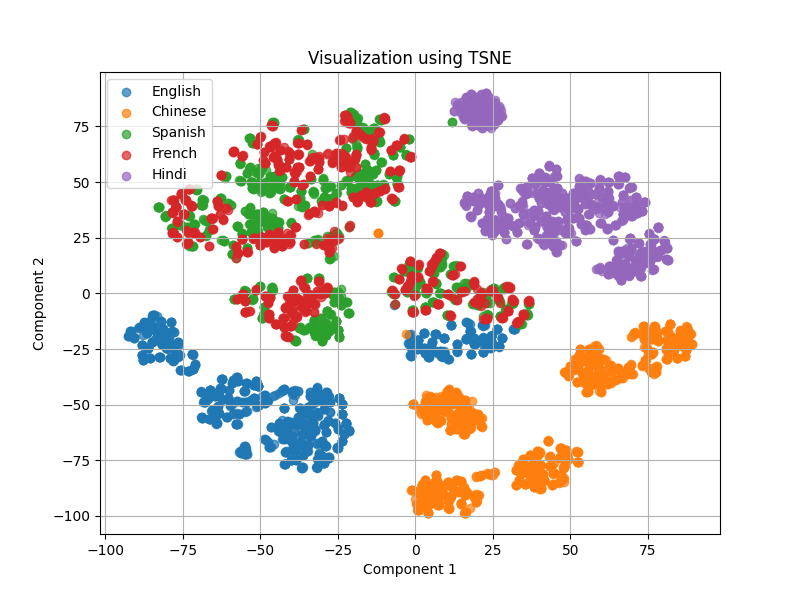}      \caption{T-SNE, Layer 23}        \end{subfigure}  \hfill  \begin{subfigure}{0.18\textwidth}      \includegraphics[width=\textwidth]{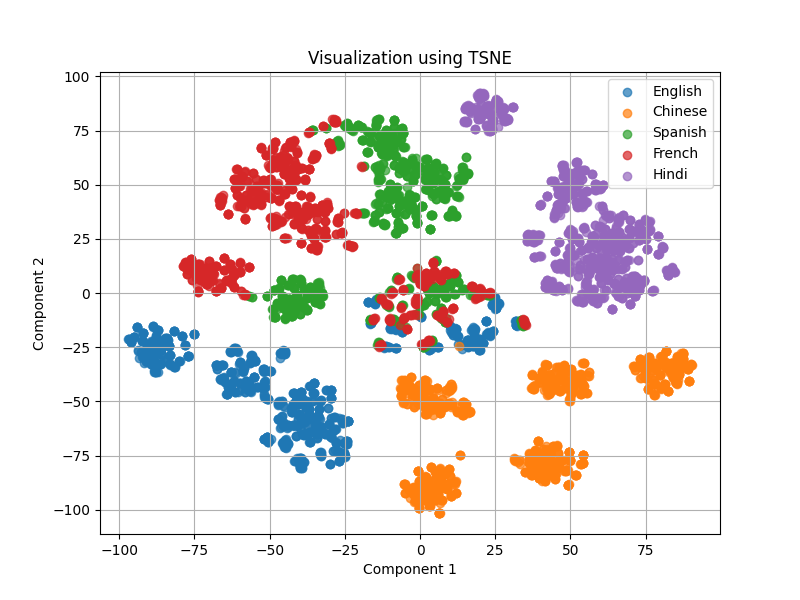}      \caption{T-SNE, Layer 24}        \end{subfigure}  \hfill  \begin{subfigure}{0.18\textwidth}      \includegraphics[width=\textwidth]{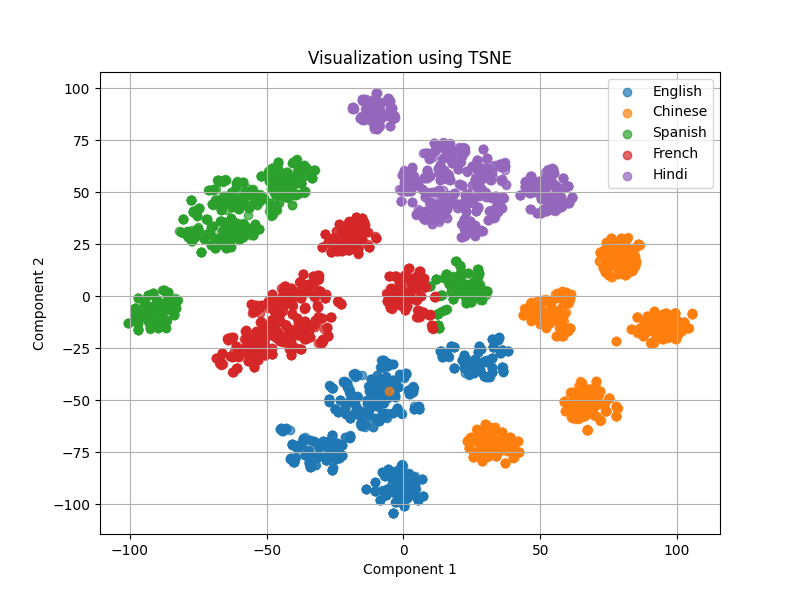}      \caption{T-SNE, Layer 25}        \end{subfigure}    \vspace{0.2in}    %
\begin{subfigure}{0.18\textwidth}      \includegraphics[width=\textwidth]{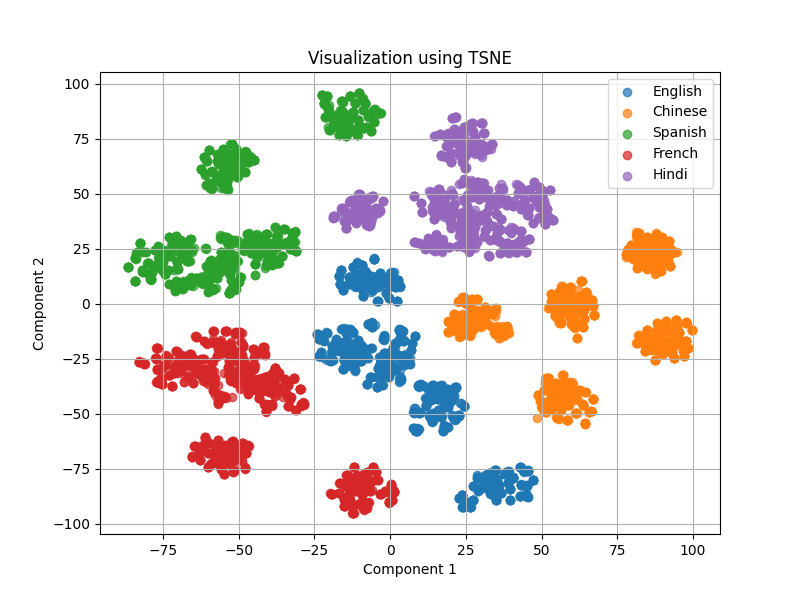}      \caption{T-SNE, Layer 26}        \end{subfigure}  \hfill  \begin{subfigure}{0.18\textwidth}      \includegraphics[width=\textwidth]{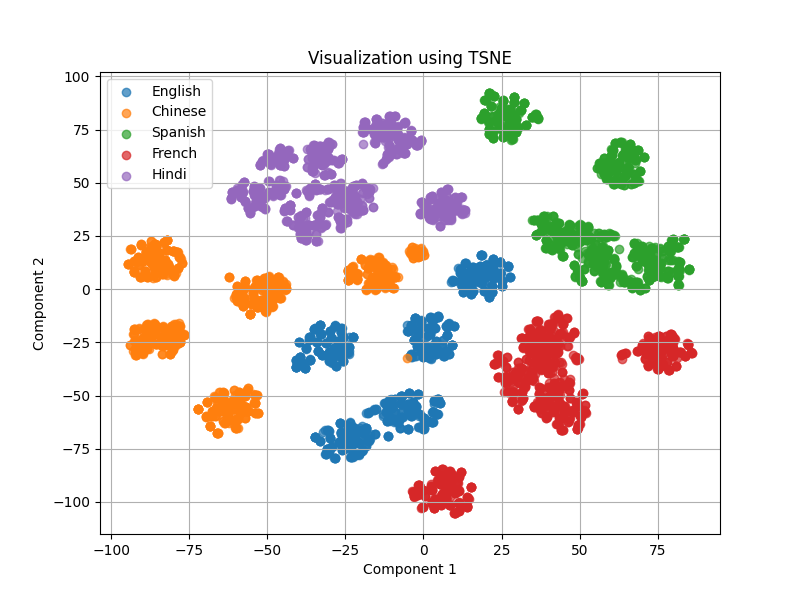}      \caption{T-SNE, Layer 27}        \end{subfigure}  \hfill  \begin{subfigure}{0.18\textwidth}      \includegraphics[width=\textwidth]{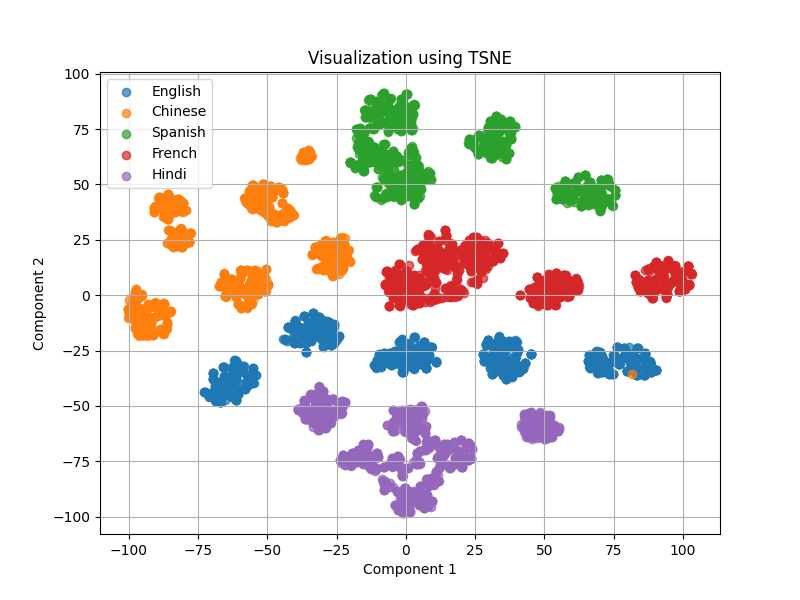}      \caption{T-SNE, Layer 28}        \end{subfigure}      \caption{T-SNE visualizations for layers 1-28 of Qwen2-7B-Instruct on the LogicalDeduction dataset.}  
\end{figure*}

\begin{figure*}[htbp]
\centering
\begin{subfigure}{0.18\textwidth}
\includegraphics[width=\textwidth]{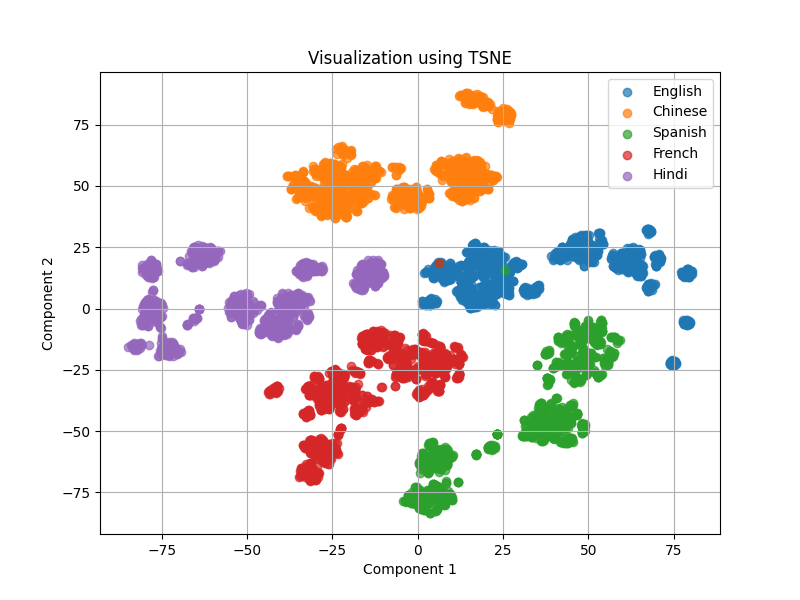}
\caption{T-SNE, Layer 1}
\end{subfigure}
\hfill
\begin{subfigure}{0.18\textwidth}
\includegraphics[width=\textwidth]{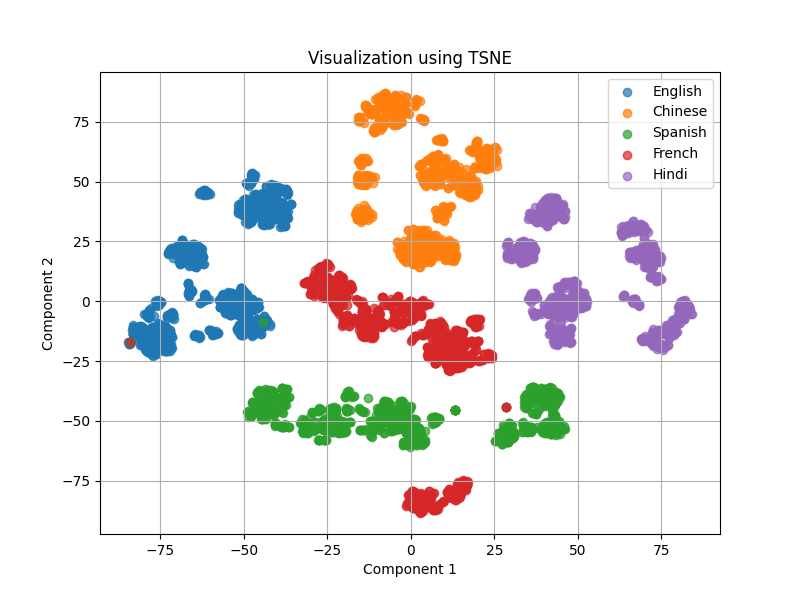}
\caption{T-SNE, Layer 2}

\end{subfigure}
\hfill
\begin{subfigure}{0.18\textwidth}
\includegraphics[width=\textwidth]{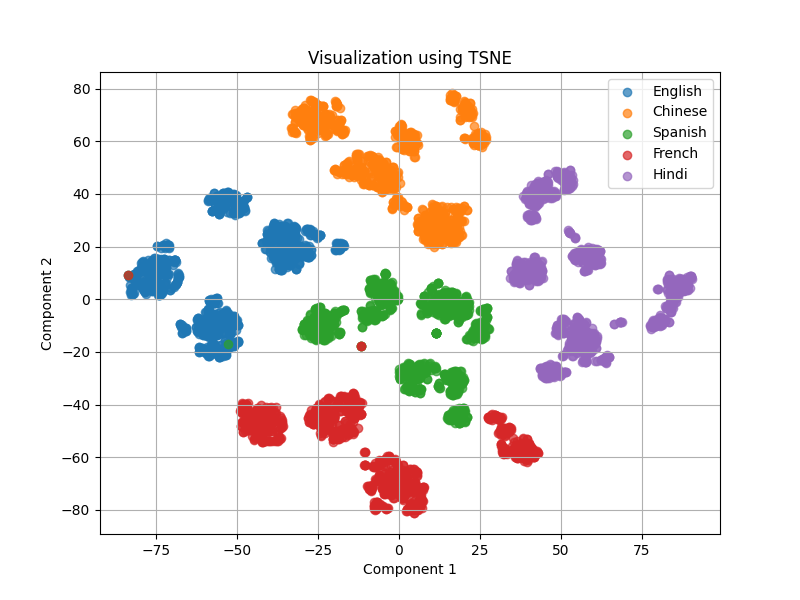}
\caption{T-SNE, Layer 3}

\end{subfigure}
\hfill
\begin{subfigure}{0.18\textwidth}
\includegraphics[width=\textwidth]{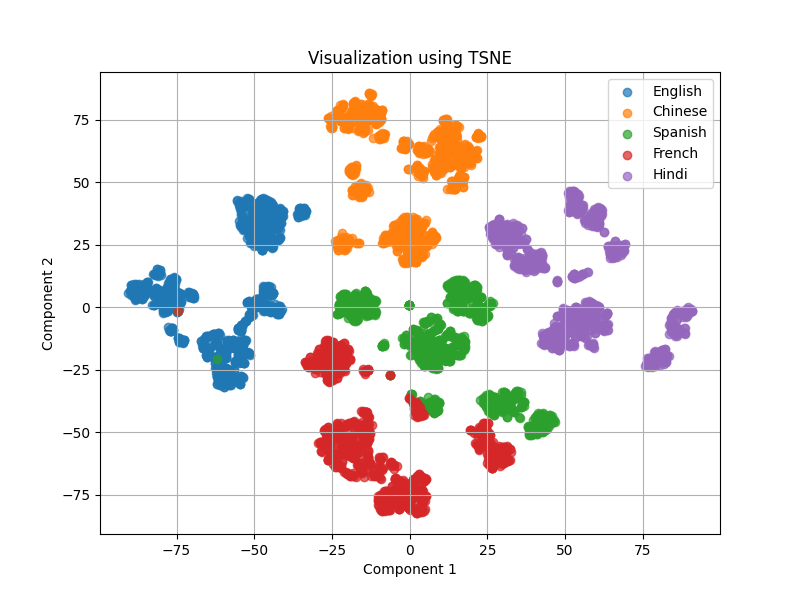}
\caption{T-SNE, Layer 4}

\end{subfigure}
\hfill
\begin{subfigure}{0.18\textwidth}
\includegraphics[width=\textwidth]{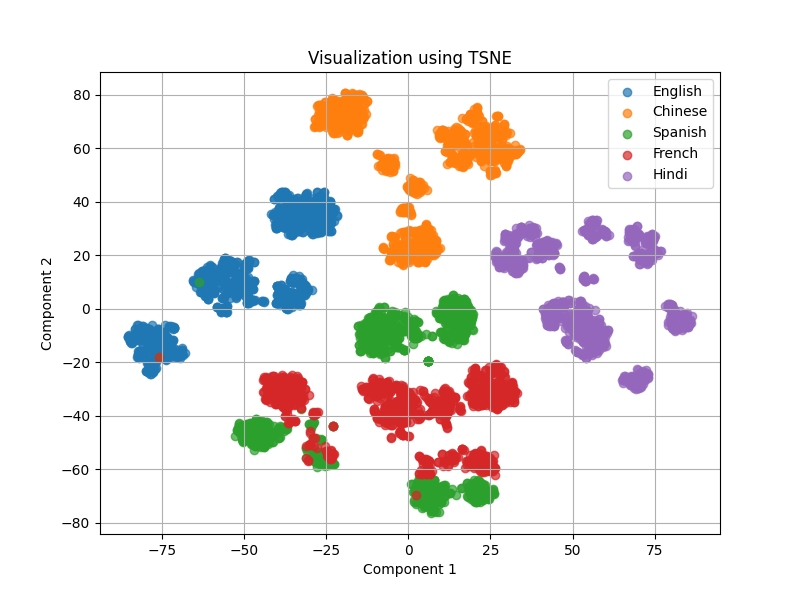}
\caption{T-SNE, Layer 5}

\end{subfigure}
\vspace{0.2in} %
\begin{subfigure}{0.18\textwidth}      \includegraphics[width=\textwidth]{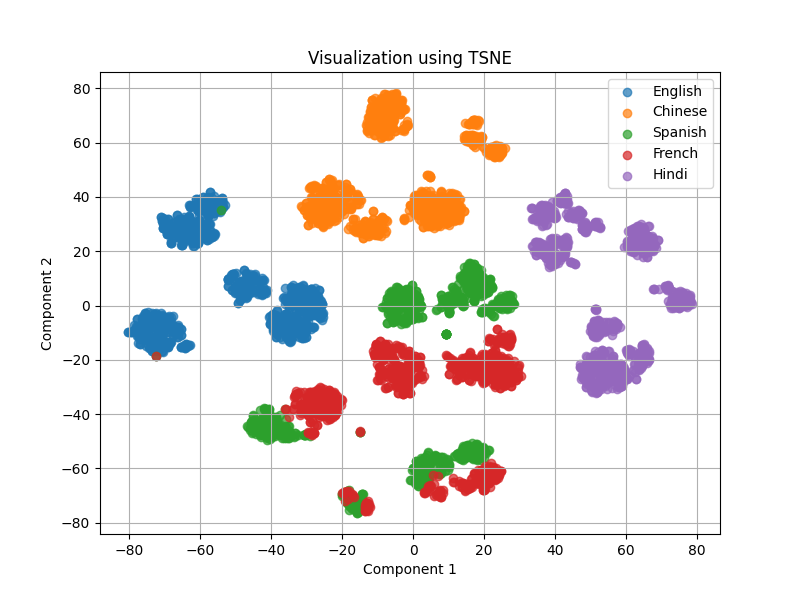}      \caption{T-SNE, Layer 6}        \end{subfigure}  \hfill  \begin{subfigure}{0.18\textwidth}      \includegraphics[width=\textwidth]{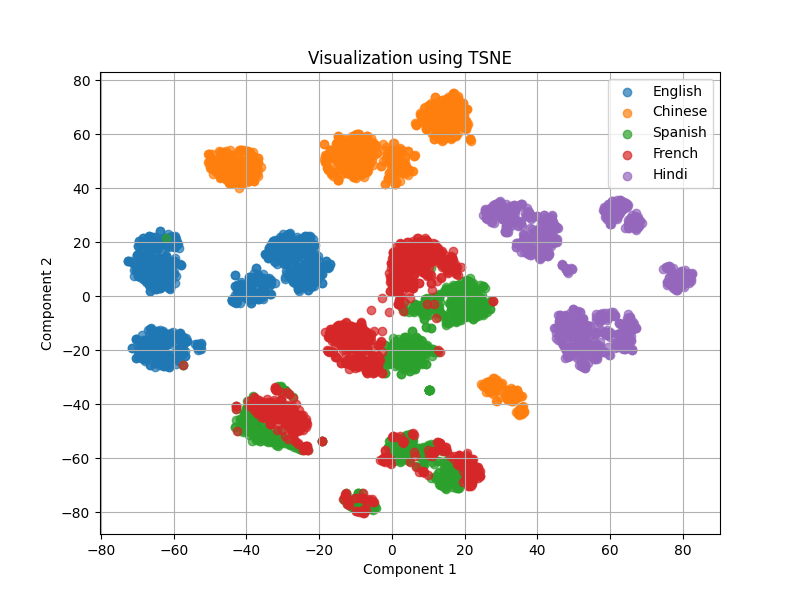}      \caption{T-SNE, Layer 7}        \end{subfigure}  \hfill  \begin{subfigure}{0.18\textwidth}      \includegraphics[width=\textwidth]{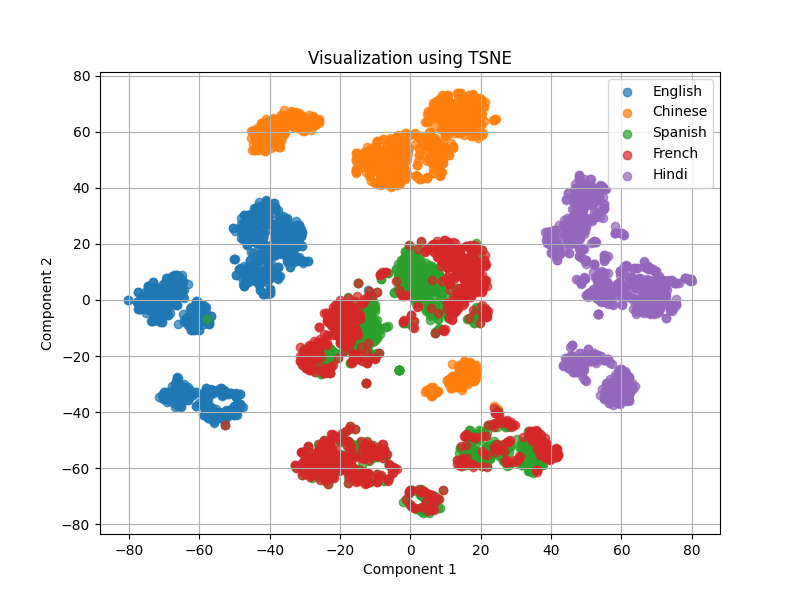}      \caption{T-SNE, Layer 8}        \end{subfigure}  \hfill  \begin{subfigure}{0.18\textwidth}      \includegraphics[width=\textwidth]{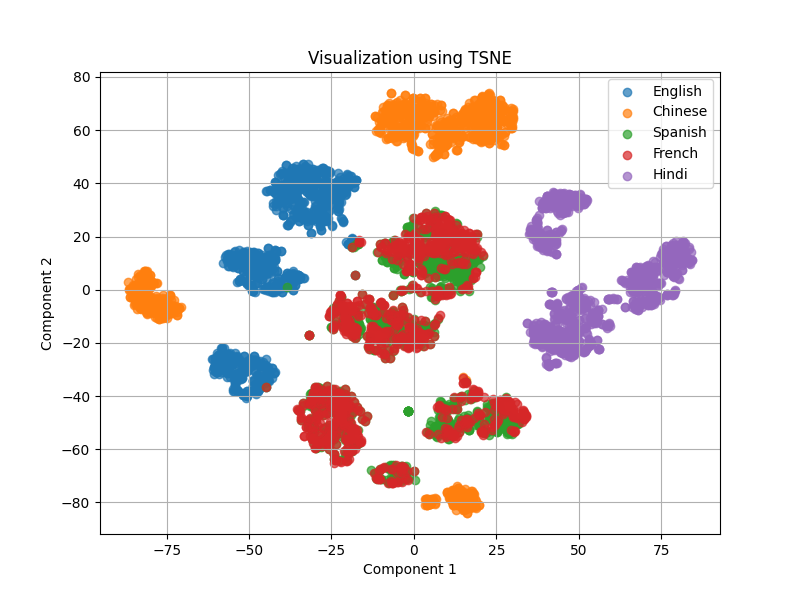}      \caption{T-SNE, Layer 9}        \end{subfigure}  \hfill  \begin{subfigure}{0.18\textwidth}      \includegraphics[width=\textwidth]{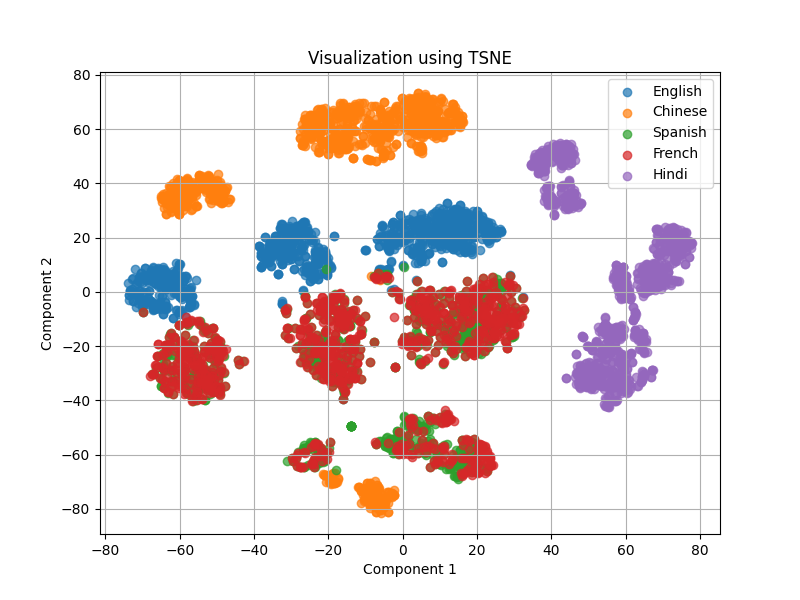}      \caption{T-SNE, Layer 10}        \end{subfigure}    \vspace{0.2in}    %
\begin{subfigure}{0.18\textwidth}      \includegraphics[width=\textwidth]{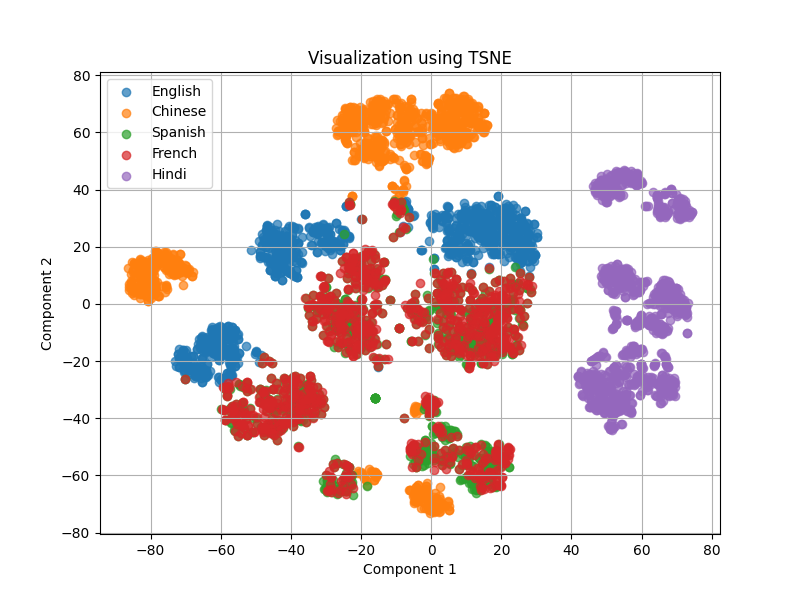}      \caption{T-SNE, Layer 11}        \end{subfigure}  \hfill  \begin{subfigure}{0.18\textwidth}      \includegraphics[width=\textwidth]{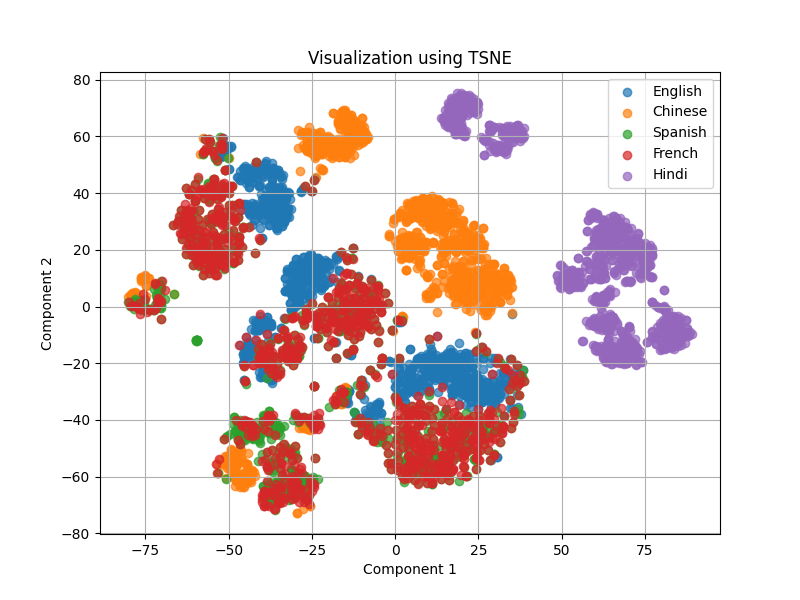}      \caption{T-SNE, Layer 12}        \end{subfigure}  \hfill  \begin{subfigure}{0.18\textwidth}      \includegraphics[width=\textwidth]{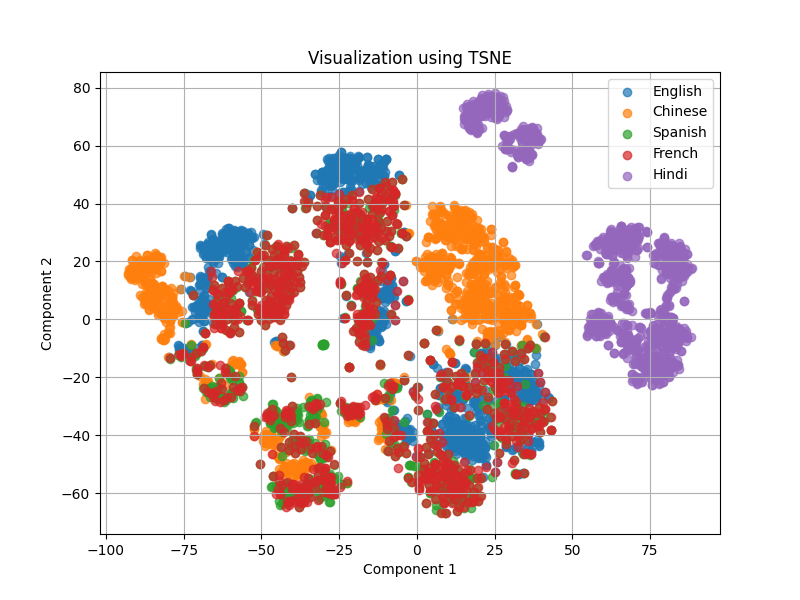}      \caption{T-SNE, Layer 13}        \end{subfigure}  \hfill  \begin{subfigure}{0.18\textwidth}      \includegraphics[width=\textwidth]{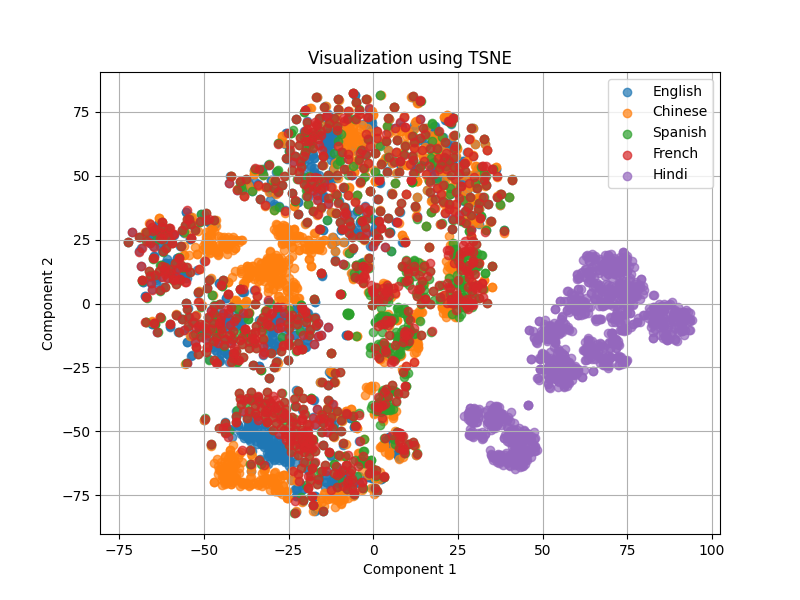}      \caption{T-SNE, Layer 14}        \end{subfigure}  \hfill  \begin{subfigure}{0.18\textwidth}      \includegraphics[width=\textwidth]{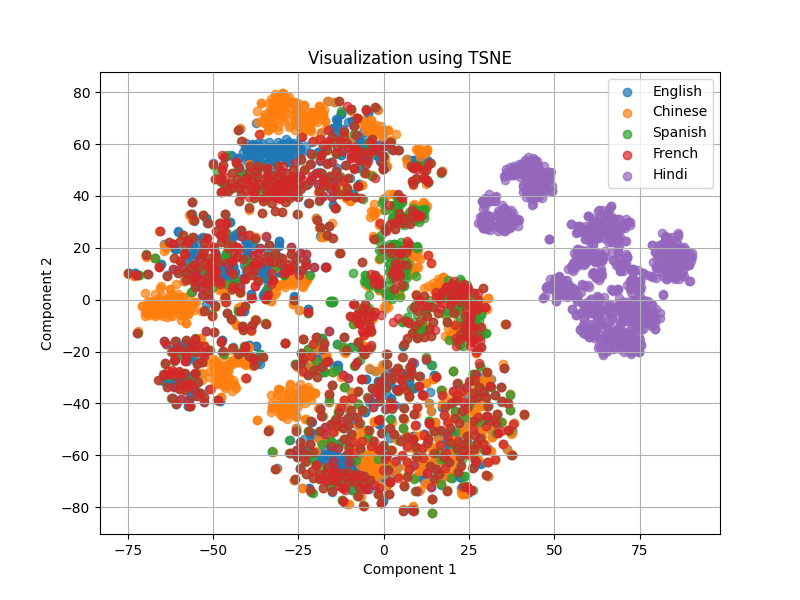}      \caption{T-SNE, Layer 15}        \end{subfigure}    \vspace{0.2in}    %
\begin{subfigure}{0.18\textwidth}      \includegraphics[width=\textwidth]{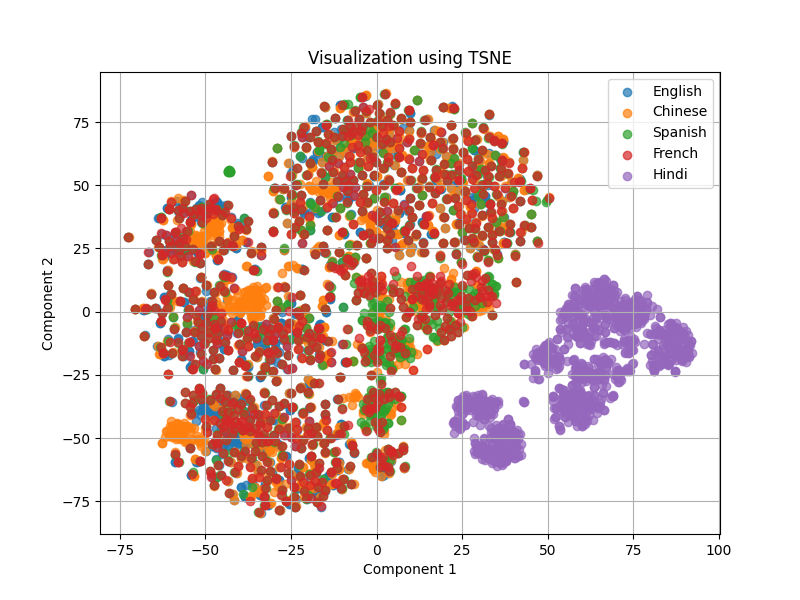}      \caption{T-SNE, Layer 16}        \end{subfigure}  \hfill  \begin{subfigure}{0.18\textwidth}      \includegraphics[width=\textwidth]{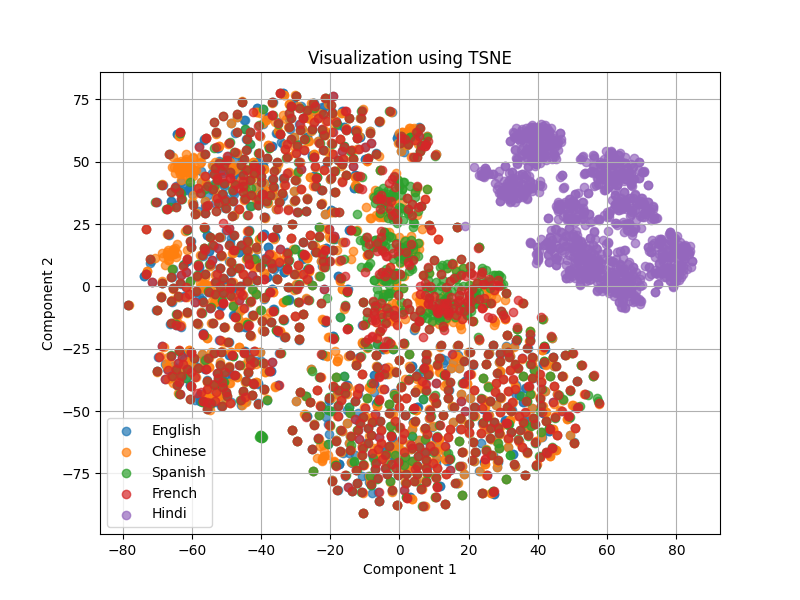}      \caption{T-SNE, Layer 17}        \end{subfigure}  \hfill  \begin{subfigure}{0.18\textwidth}      \includegraphics[width=\textwidth]{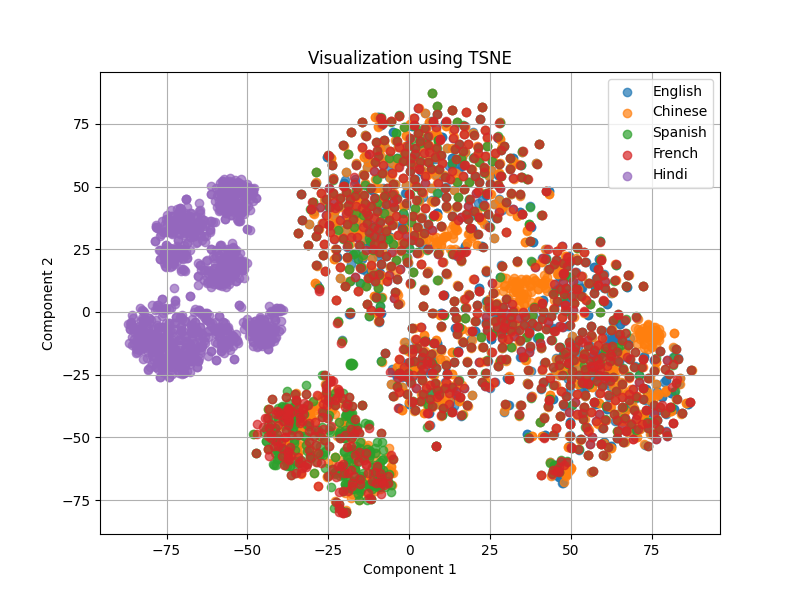}      \caption{T-SNE, Layer 18}        \end{subfigure}  \hfill  \begin{subfigure}{0.18\textwidth}      \includegraphics[width=\textwidth]{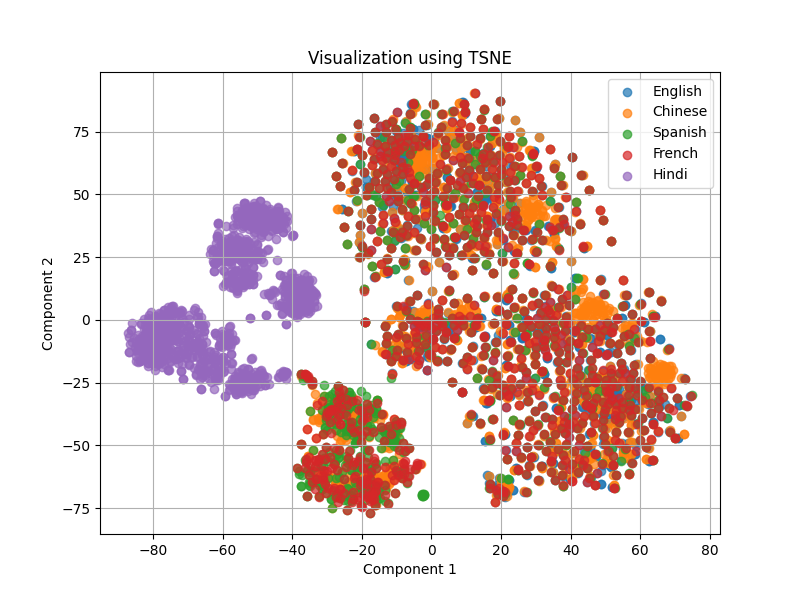}      \caption{T-SNE, Layer 19}        \end{subfigure}  \hfill  \begin{subfigure}{0.18\textwidth}      \includegraphics[width=\textwidth]{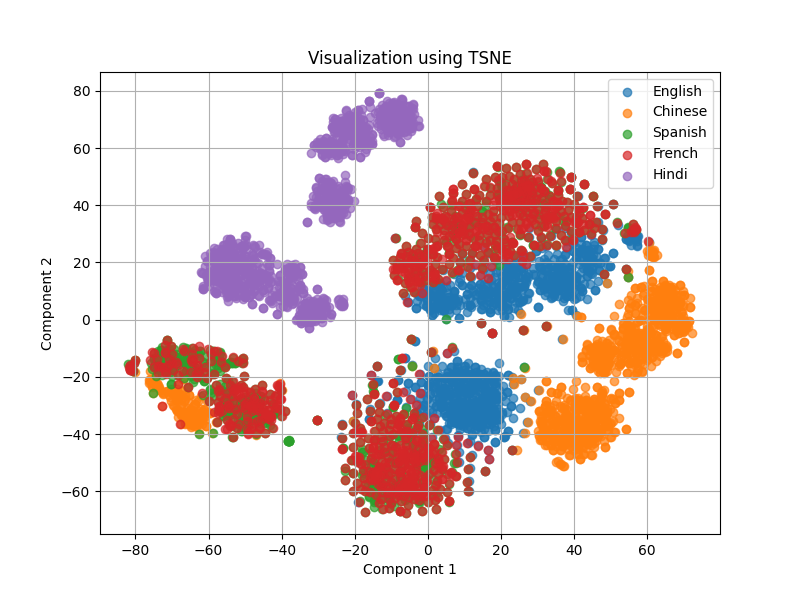}      \caption{T-SNE, Layer 20}        \end{subfigure}    \vspace{0.2in}    %
\begin{subfigure}{0.18\textwidth}      \includegraphics[width=\textwidth]{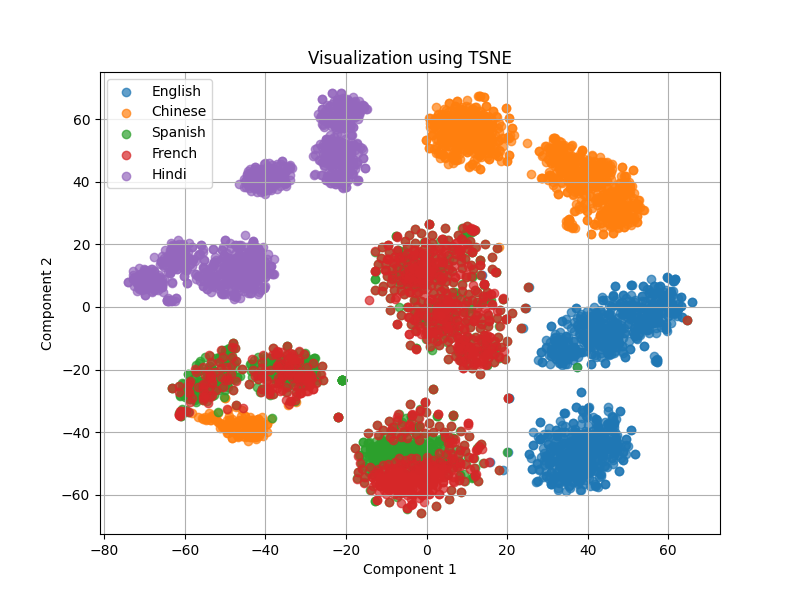}      \caption{T-SNE, Layer 21}        \end{subfigure}  \hfill  \begin{subfigure}{0.18\textwidth}      \includegraphics[width=\textwidth]{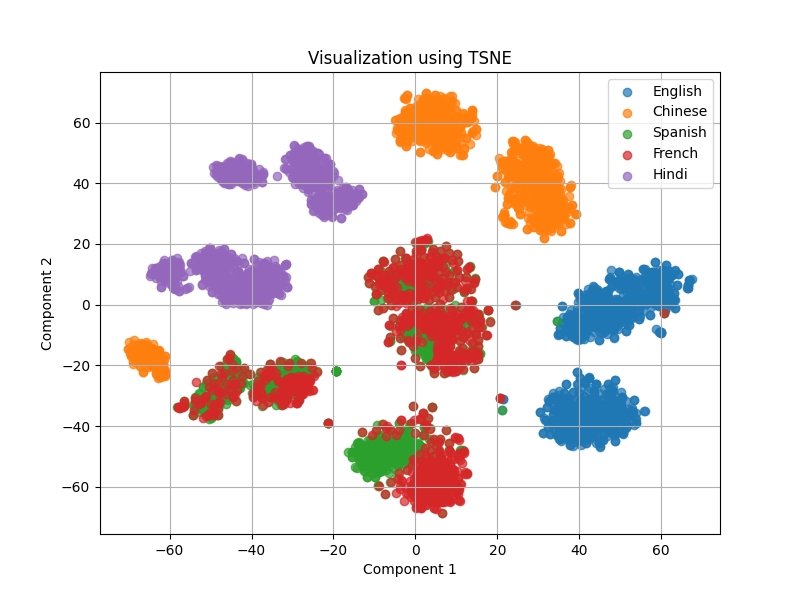}      \caption{T-SNE, Layer 22}        \end{subfigure}  \hfill  \begin{subfigure}{0.18\textwidth}      \includegraphics[width=\textwidth]{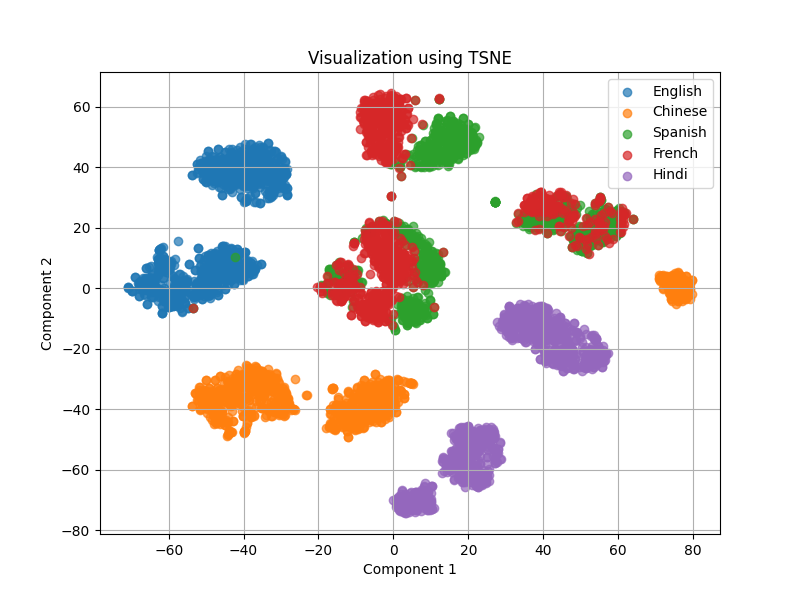}      \caption{T-SNE, Layer 23}        \end{subfigure}  \hfill  \begin{subfigure}{0.18\textwidth}      \includegraphics[width=\textwidth]{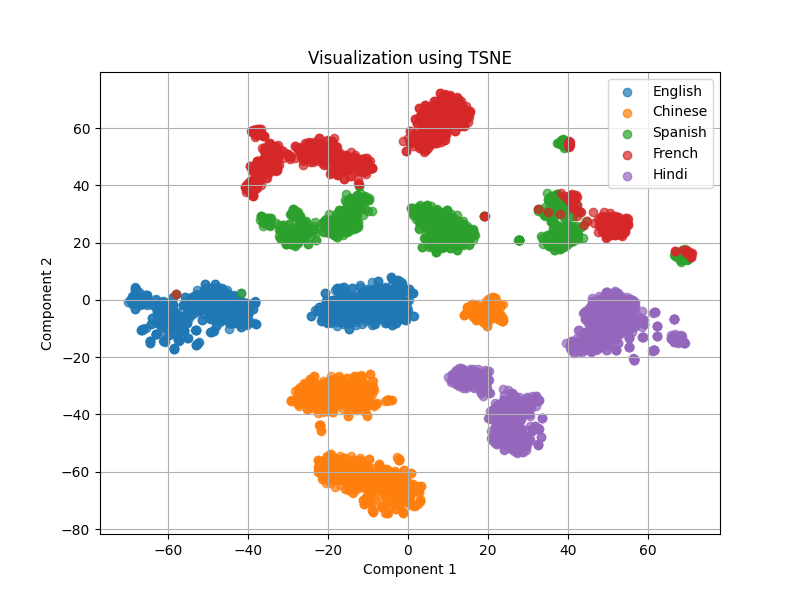}      \caption{T-SNE, Layer 24}        \end{subfigure}  \hfill  \begin{subfigure}{0.18\textwidth}      \includegraphics[width=\textwidth]{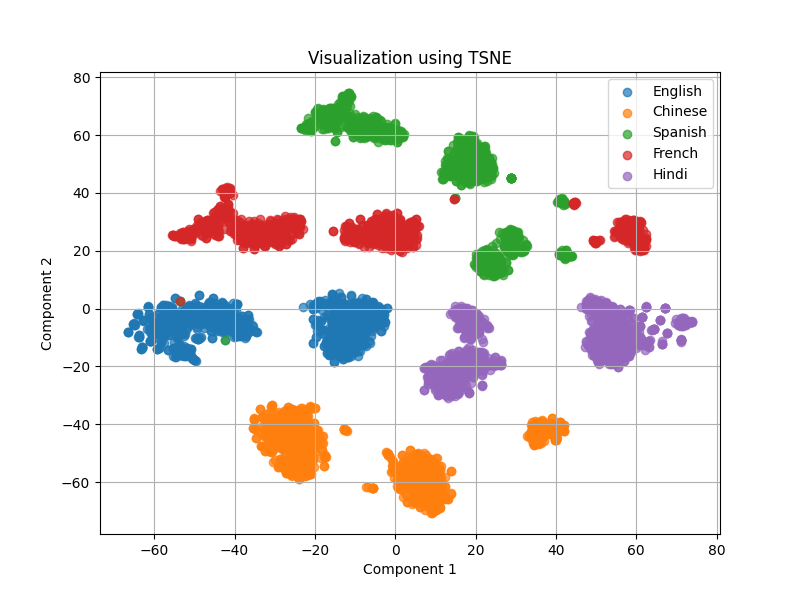}      \caption{T-SNE, Layer 25}        \end{subfigure}    \vspace{0.2in}    %
\begin{subfigure}{0.18\textwidth}      \includegraphics[width=\textwidth]{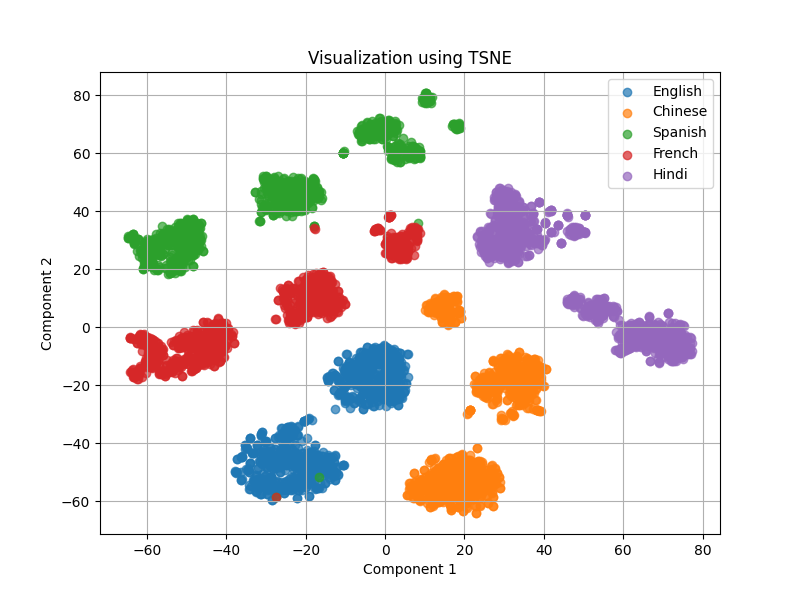}      \caption{T-SNE, Layer 26}        \end{subfigure}  \hfill  \begin{subfigure}{0.18\textwidth}      \includegraphics[width=\textwidth]{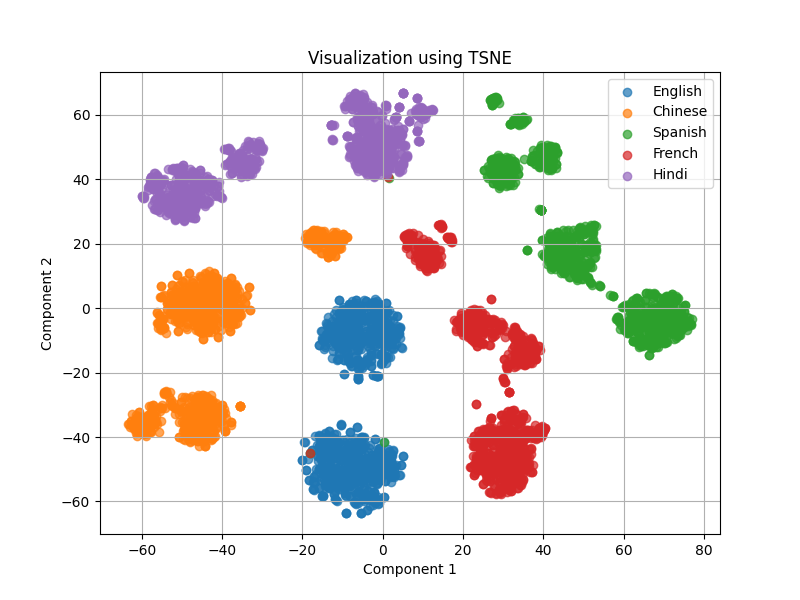}      \caption{T-SNE, Layer 27}        \end{subfigure}  \hfill  \begin{subfigure}{0.18\textwidth}      \includegraphics[width=\textwidth]{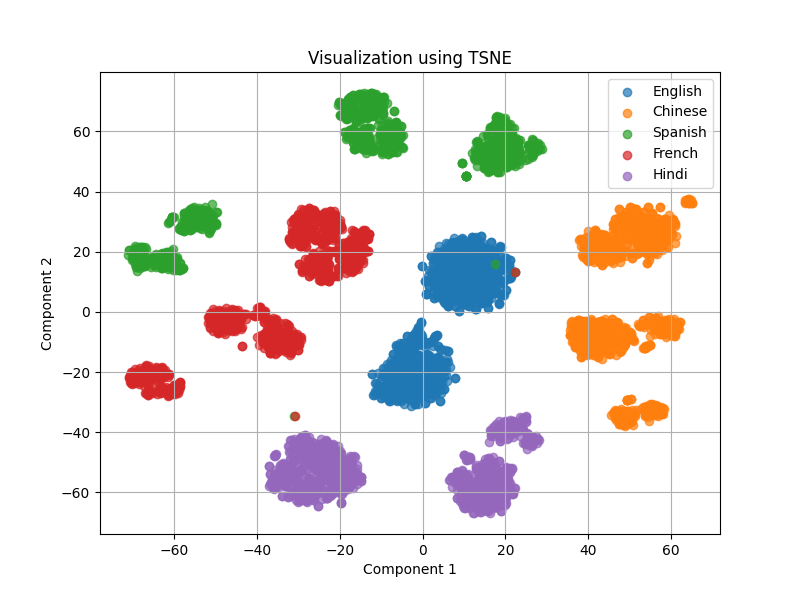}      \caption{T-SNE, Layer 28}        \end{subfigure}      \caption{T-SNE visualizations for layers 1-28 of Qwen2-7B-Instruct on the ProofWriter dataset.}  
\end{figure*}

\begin{figure*}[htbp]
\centering
\begin{subfigure}{0.18\textwidth}
\includegraphics[width=\textwidth]{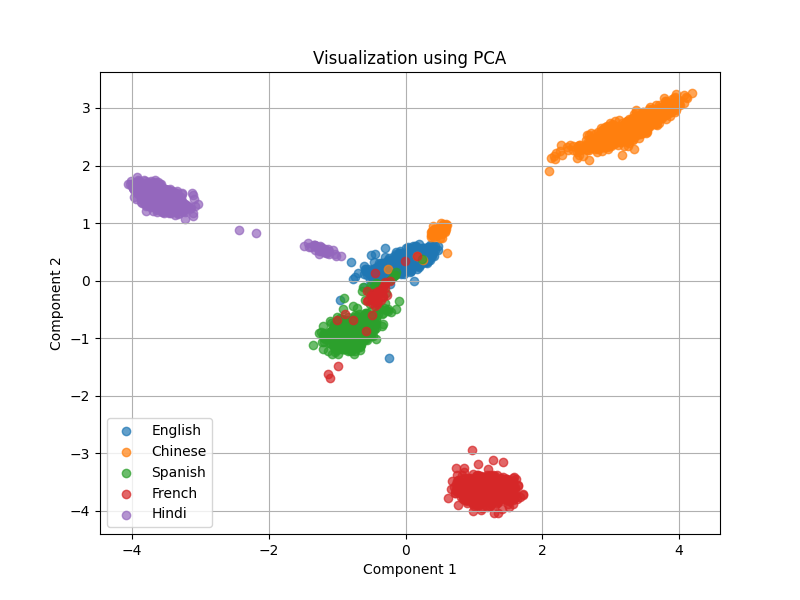}
\caption{PCA, Layer 1}
\end{subfigure}
\hfill
\begin{subfigure}{0.18\textwidth}
\includegraphics[width=\textwidth]{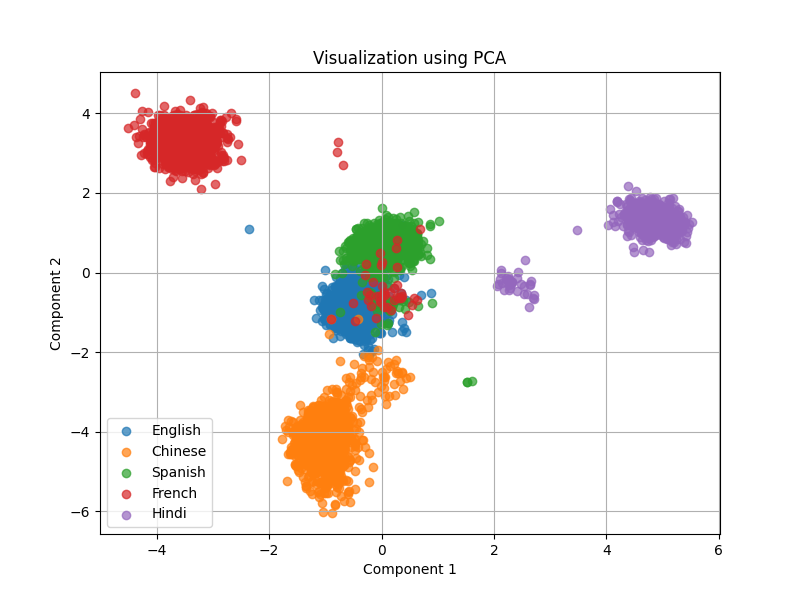}
\caption{PCA, Layer 2}

\end{subfigure}
\hfill
\begin{subfigure}{0.18\textwidth}
\includegraphics[width=\textwidth]{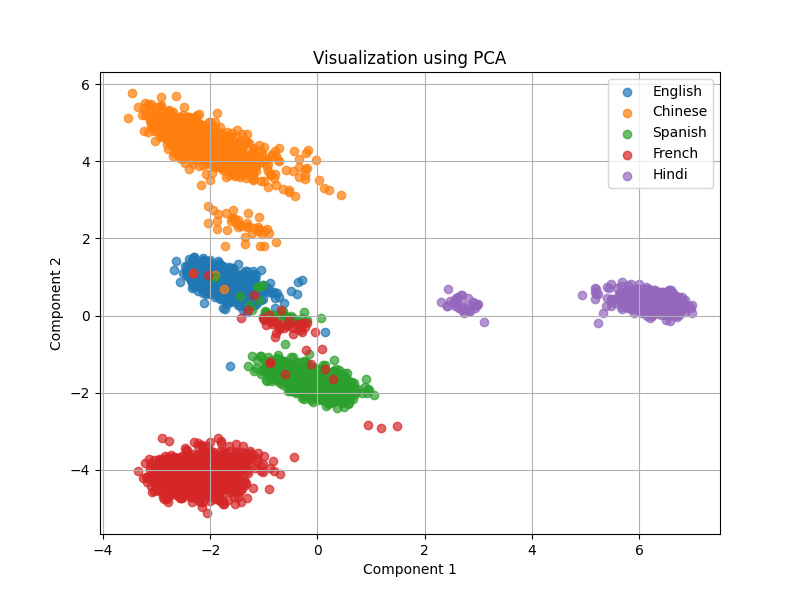}
\caption{PCA, Layer 3}

\end{subfigure}
\hfill
\begin{subfigure}{0.18\textwidth}
\includegraphics[width=\textwidth]{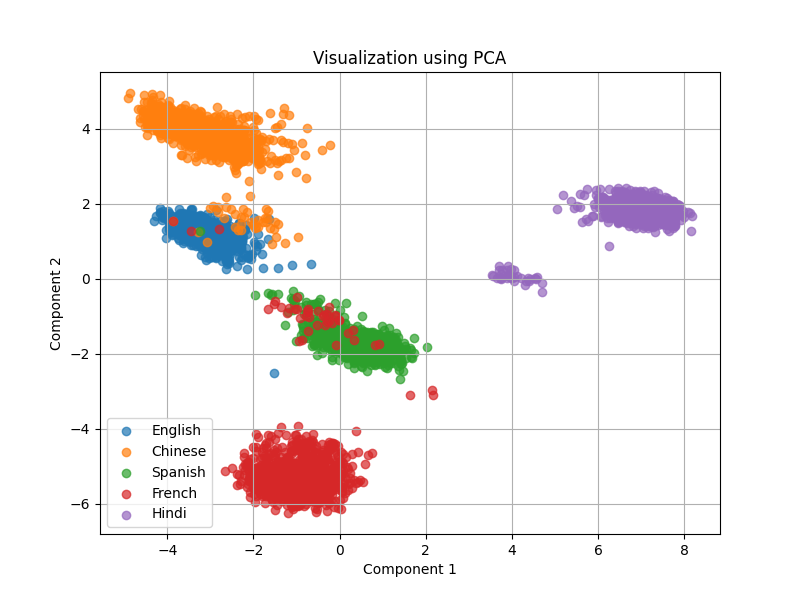}
\caption{PCA, Layer 4}

\end{subfigure}
\hfill
\begin{subfigure}{0.18\textwidth}
\includegraphics[width=\textwidth]{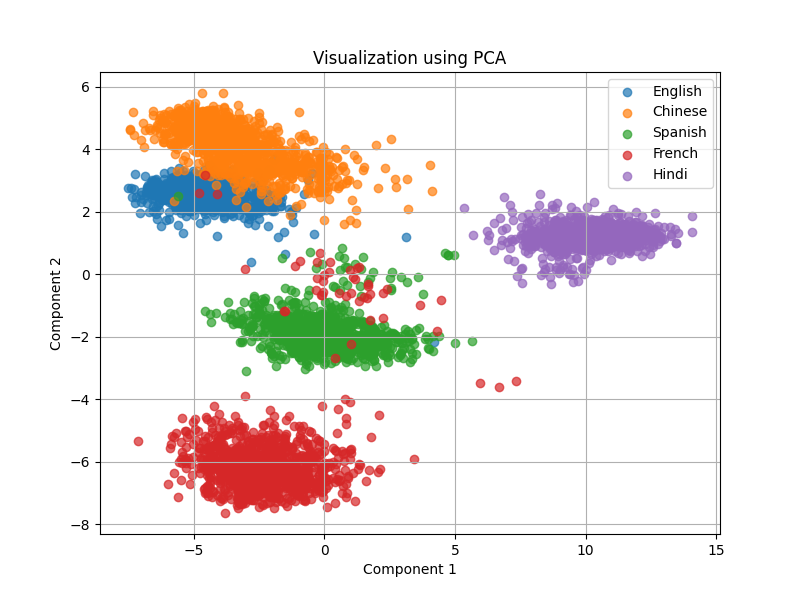}
\caption{PCA, Layer 5}

\end{subfigure}
\vspace{0.2in} %
\begin{subfigure}{0.18\textwidth}      \includegraphics[width=\textwidth]{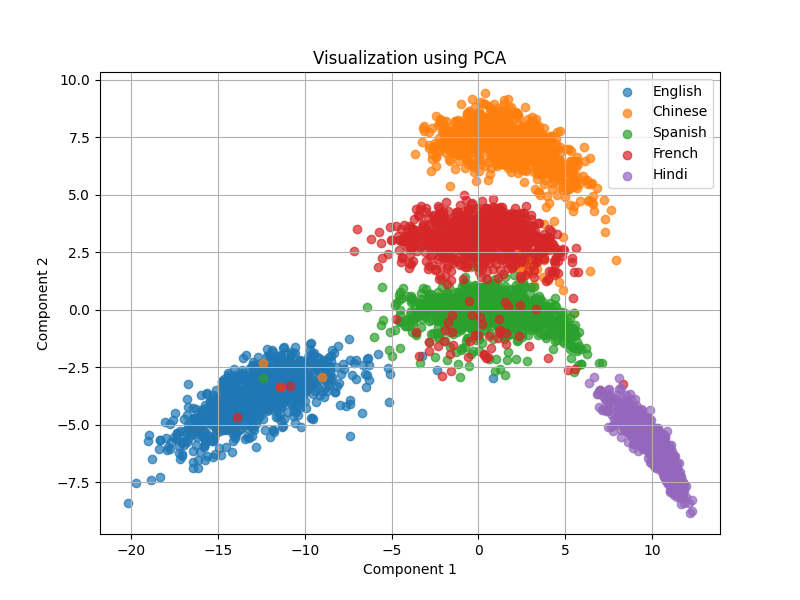}      \caption{PCA, Layer 6}        \end{subfigure}  \hfill  \begin{subfigure}{0.18\textwidth}      \includegraphics[width=\textwidth]{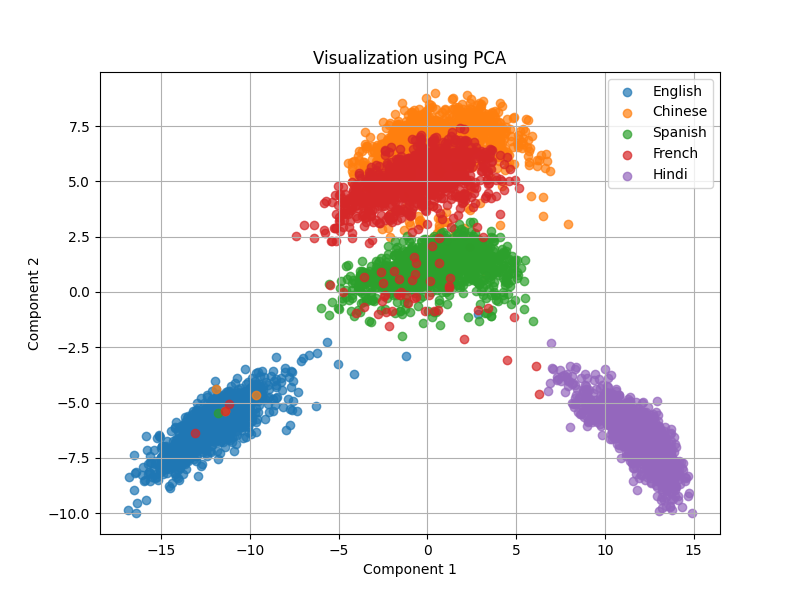}      \caption{PCA, Layer 7}        \end{subfigure}  \hfill  \begin{subfigure}{0.18\textwidth}      \includegraphics[width=\textwidth]{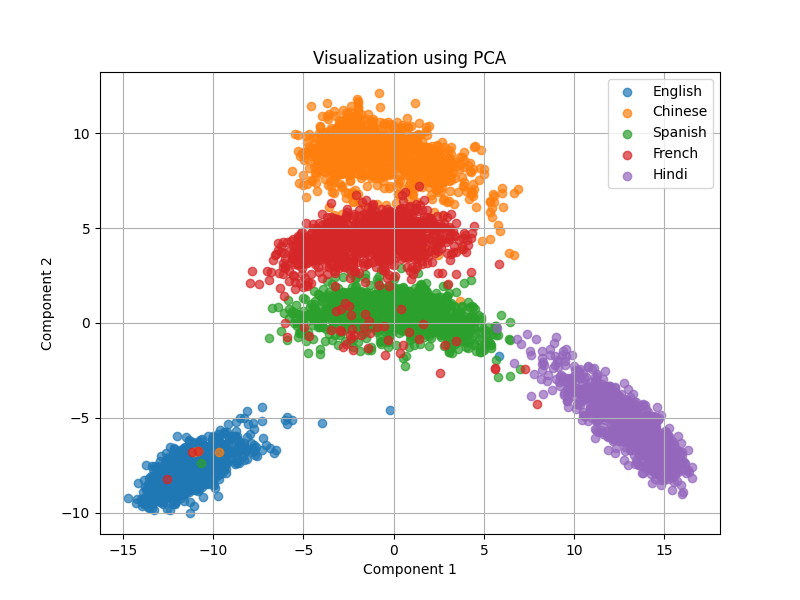}      \caption{PCA, Layer 8}        \end{subfigure}  \hfill  \begin{subfigure}{0.18\textwidth}      \includegraphics[width=\textwidth]{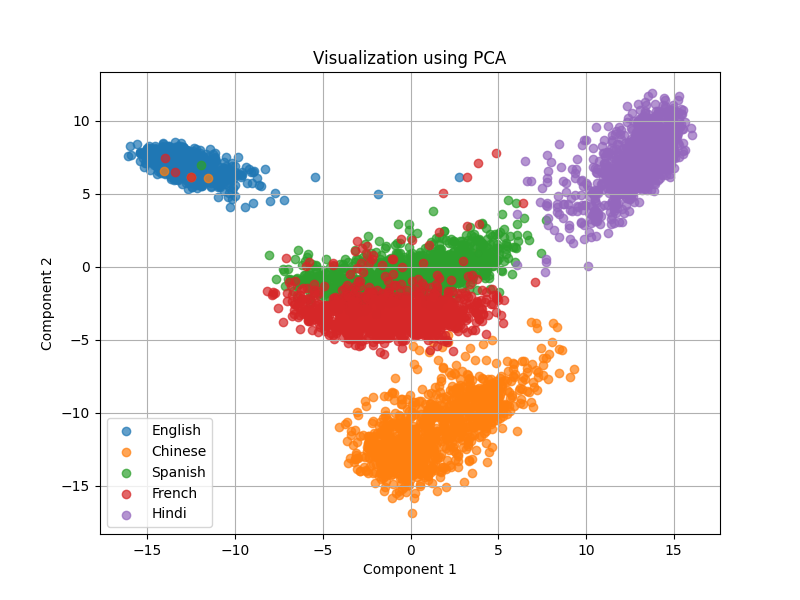}      \caption{PCA, Layer 9}        \end{subfigure}  \hfill  \begin{subfigure}{0.18\textwidth}      \includegraphics[width=\textwidth]{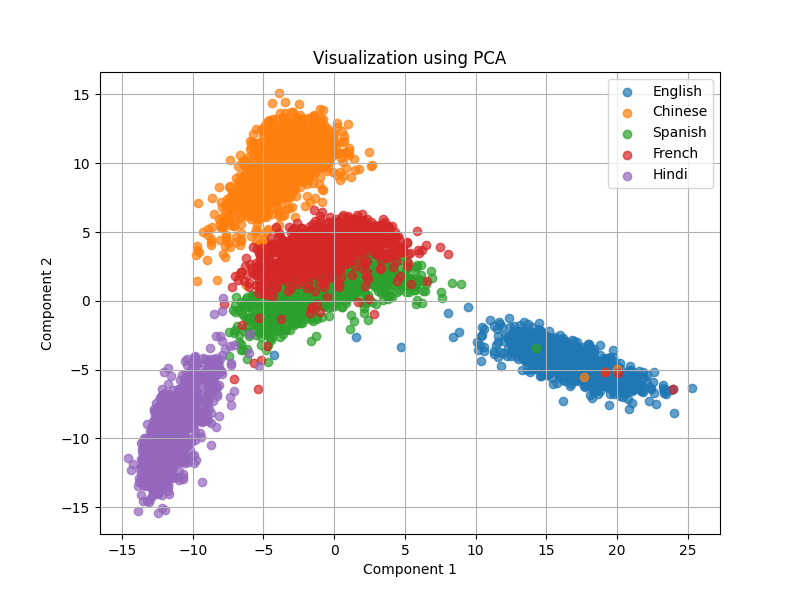}      \caption{PCA, Layer 10}        \end{subfigure}    \vspace{0.2in}    %
\begin{subfigure}{0.18\textwidth}      \includegraphics[width=\textwidth]{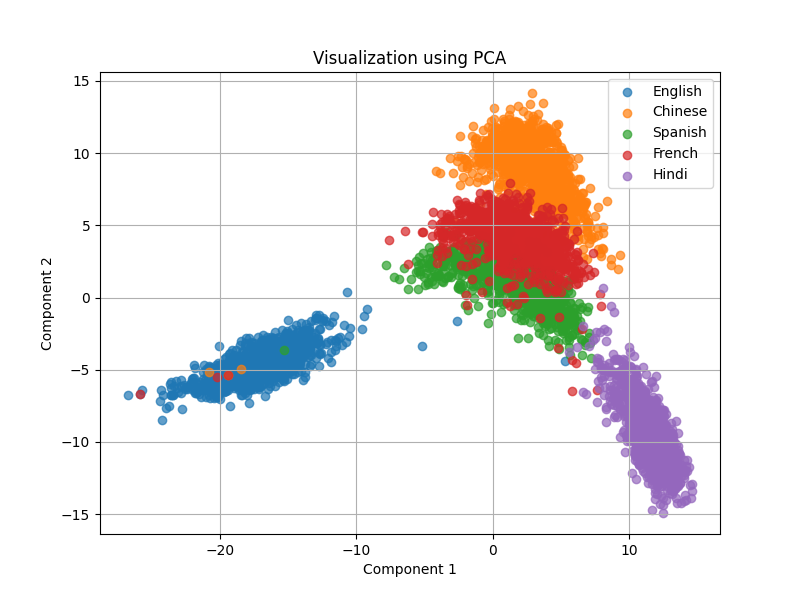}      \caption{PCA, Layer 11}        \end{subfigure}  \hfill  \begin{subfigure}{0.18\textwidth}      \includegraphics[width=\textwidth]{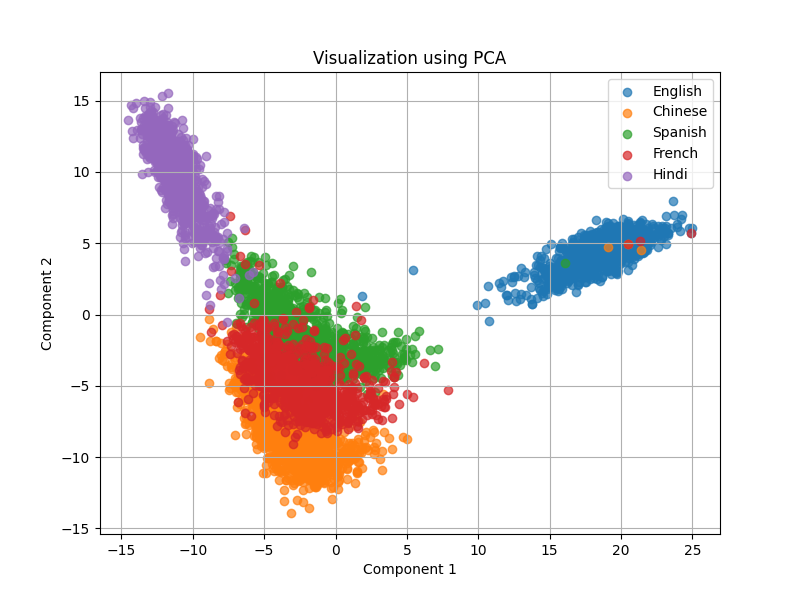}      \caption{PCA, Layer 12}        \end{subfigure}  \hfill  \begin{subfigure}{0.18\textwidth}      \includegraphics[width=\textwidth]{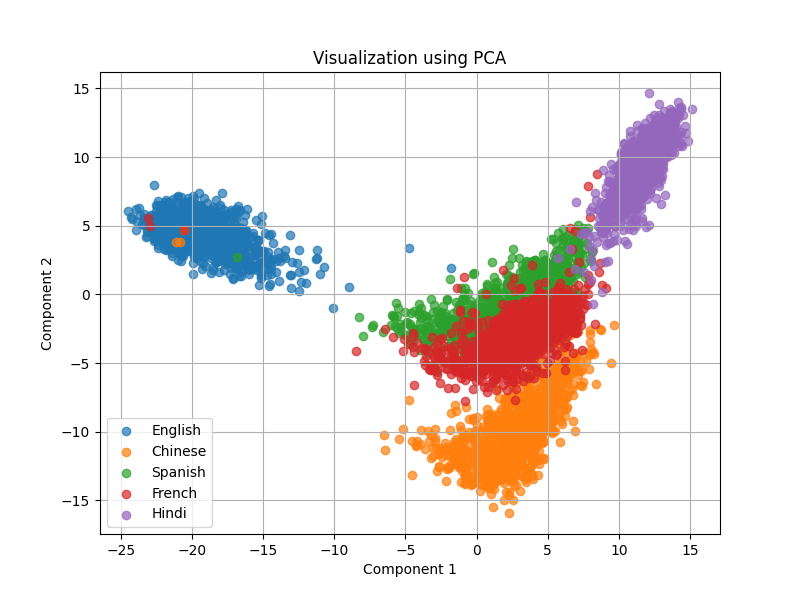}      \caption{PCA, Layer 13}        \end{subfigure}  \hfill  \begin{subfigure}{0.18\textwidth}      \includegraphics[width=\textwidth]{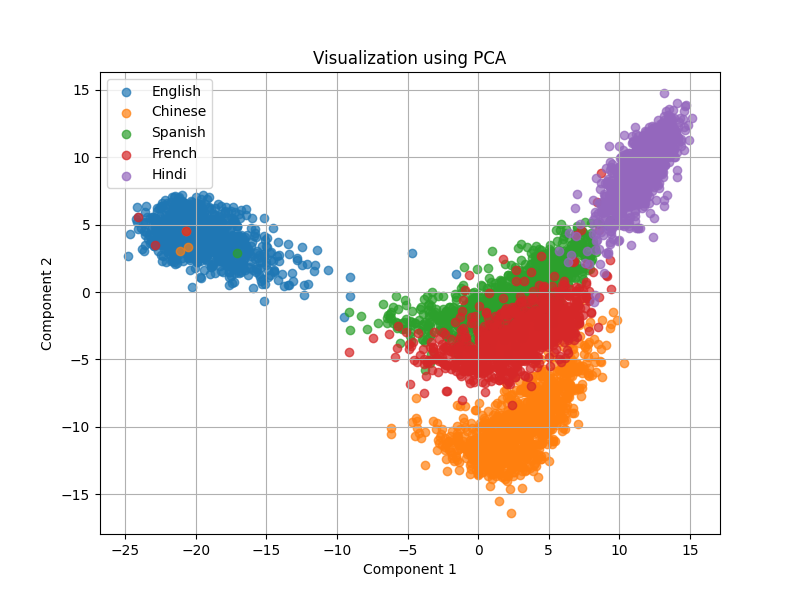}      \caption{PCA, Layer 14}        \end{subfigure}  \hfill  \begin{subfigure}{0.18\textwidth}      \includegraphics[width=\textwidth]{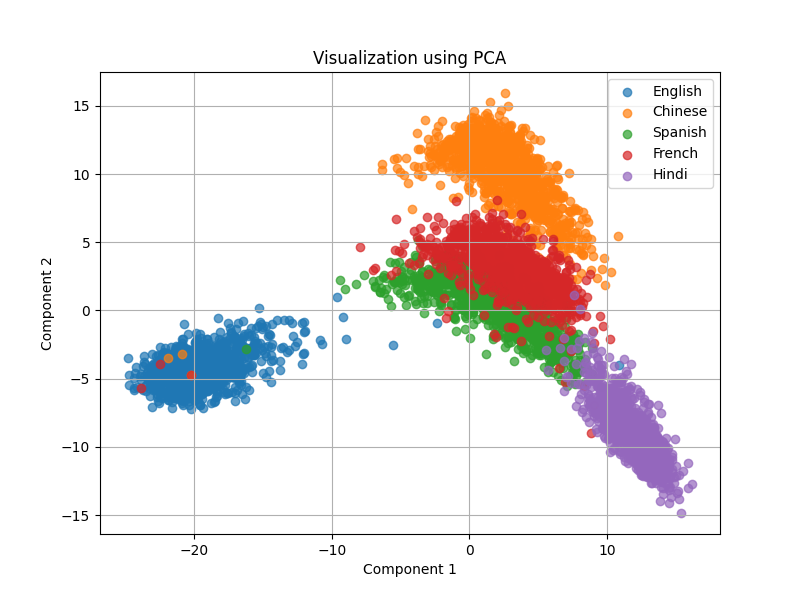}      \caption{PCA, Layer 15}        \end{subfigure}    \vspace{0.2in}    %
\begin{subfigure}{0.18\textwidth}      \includegraphics[width=\textwidth]{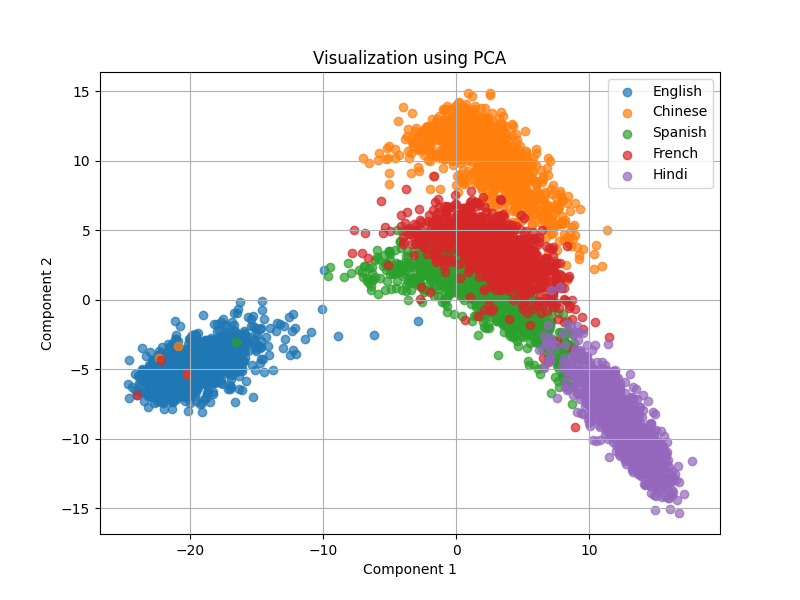}      \caption{PCA, Layer 16}        \end{subfigure}  \hfill  \begin{subfigure}{0.18\textwidth}      \includegraphics[width=\textwidth]{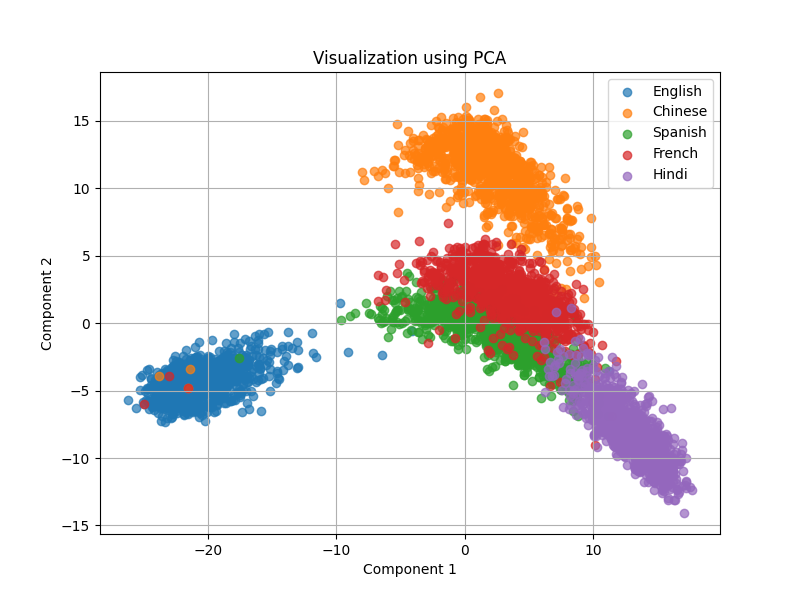}      \caption{PCA, Layer 17}        \end{subfigure}  \hfill  \begin{subfigure}{0.18\textwidth}      \includegraphics[width=\textwidth]{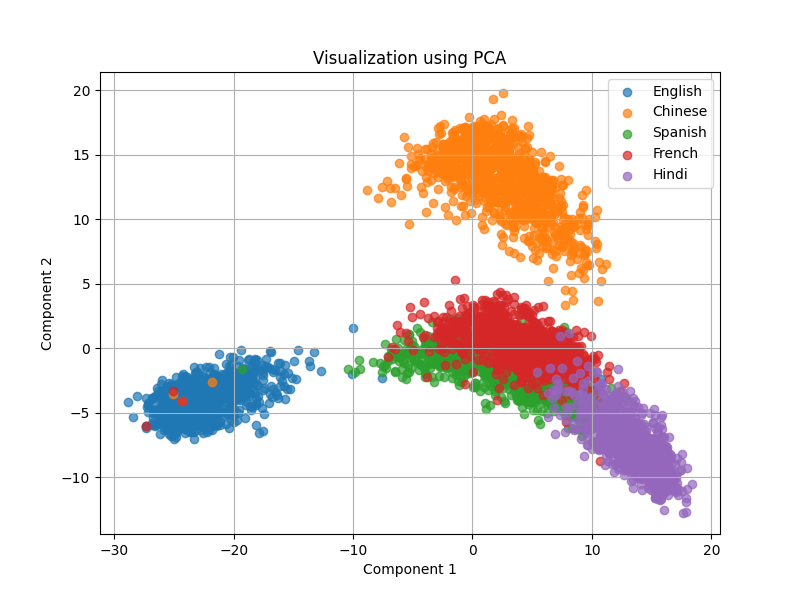}      \caption{PCA, Layer 18}        \end{subfigure}  \hfill  \begin{subfigure}{0.18\textwidth}      \includegraphics[width=\textwidth]{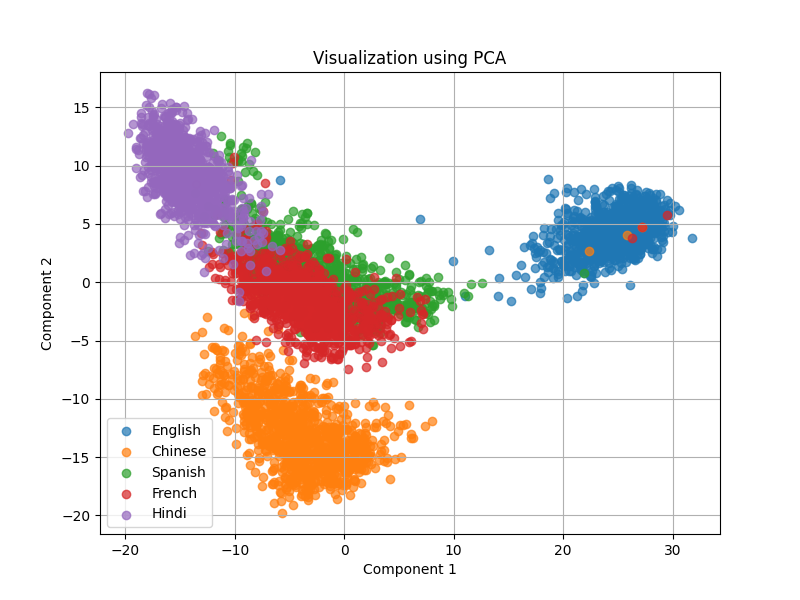}      \caption{PCA, Layer 19}        \end{subfigure}  \hfill  \begin{subfigure}{0.18\textwidth}      \includegraphics[width=\textwidth]{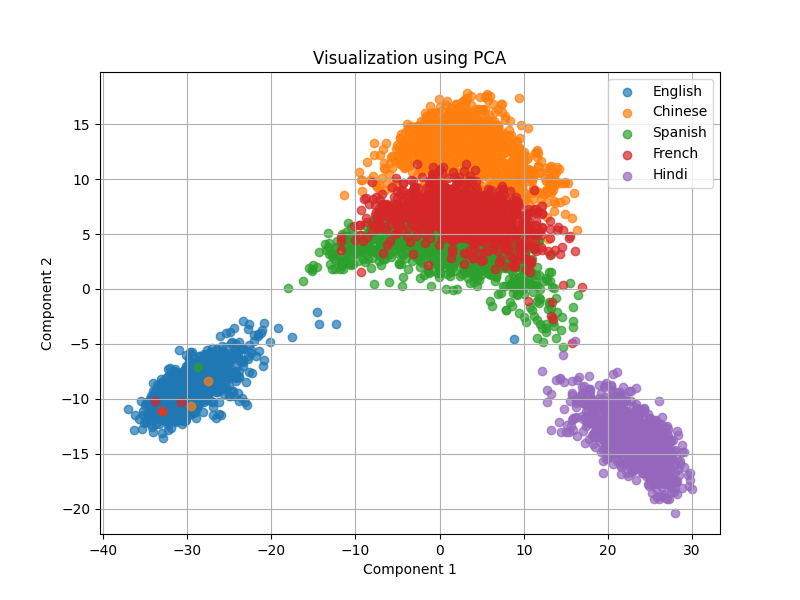}      \caption{PCA, Layer 20}        \end{subfigure}    \vspace{0.2in}    %
\begin{subfigure}{0.18\textwidth}      \includegraphics[width=\textwidth]{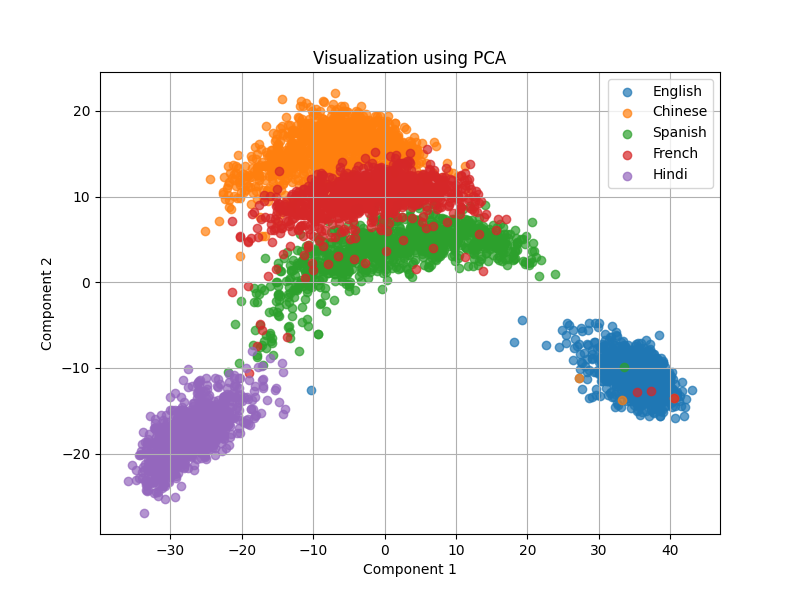}      \caption{PCA, Layer 21}        \end{subfigure}  \hfill  \begin{subfigure}{0.18\textwidth}      \includegraphics[width=\textwidth]{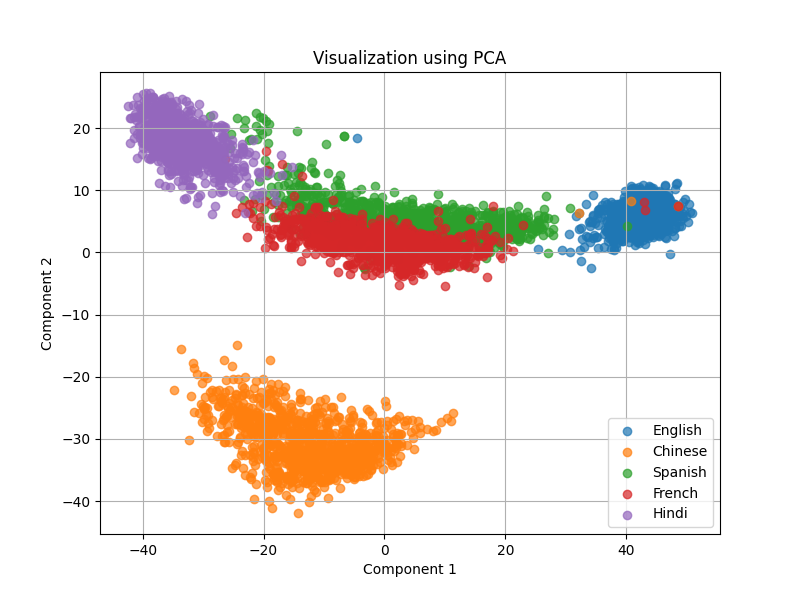}      \caption{PCA, Layer 22}        \end{subfigure}  \hfill  \begin{subfigure}{0.18\textwidth}      \includegraphics[width=\textwidth]{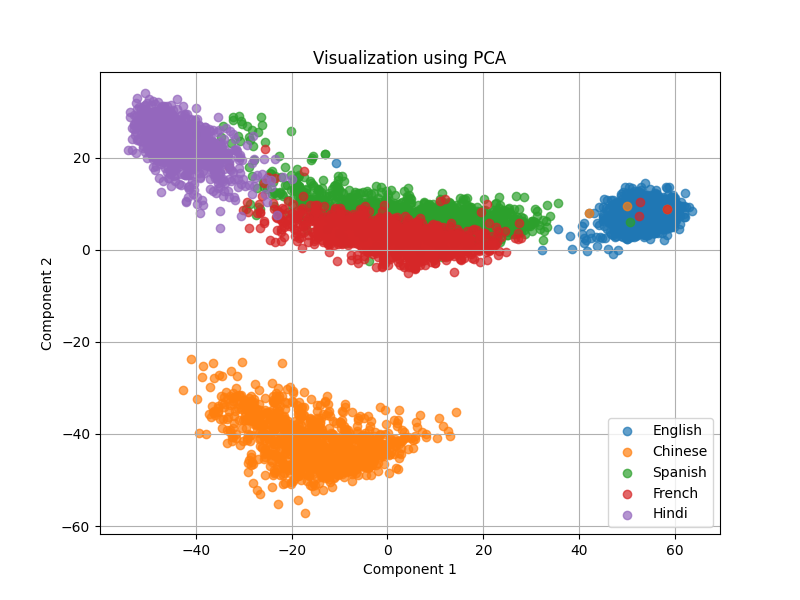}      \caption{PCA, Layer 23}        \end{subfigure}  \hfill  \begin{subfigure}{0.18\textwidth}      \includegraphics[width=\textwidth]{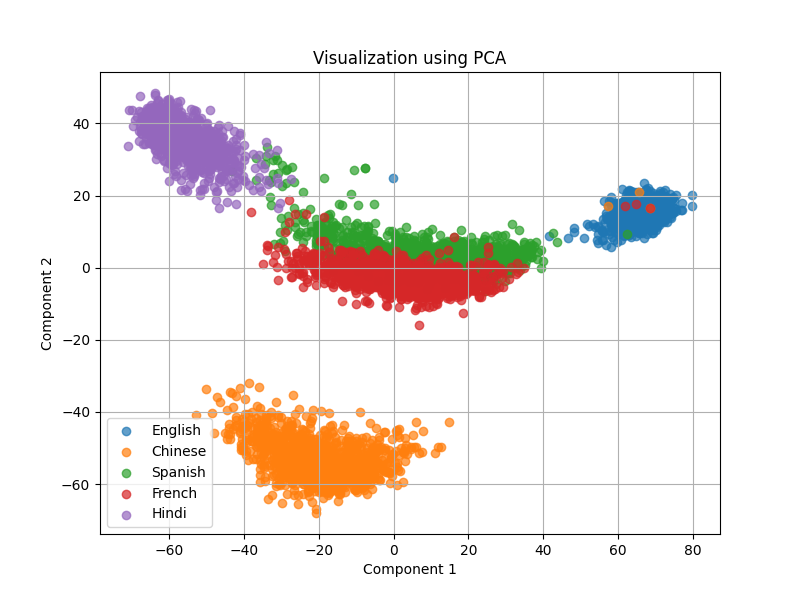}      \caption{PCA, Layer 24}        \end{subfigure}  \hfill  \begin{subfigure}{0.18\textwidth}      \includegraphics[width=\textwidth]{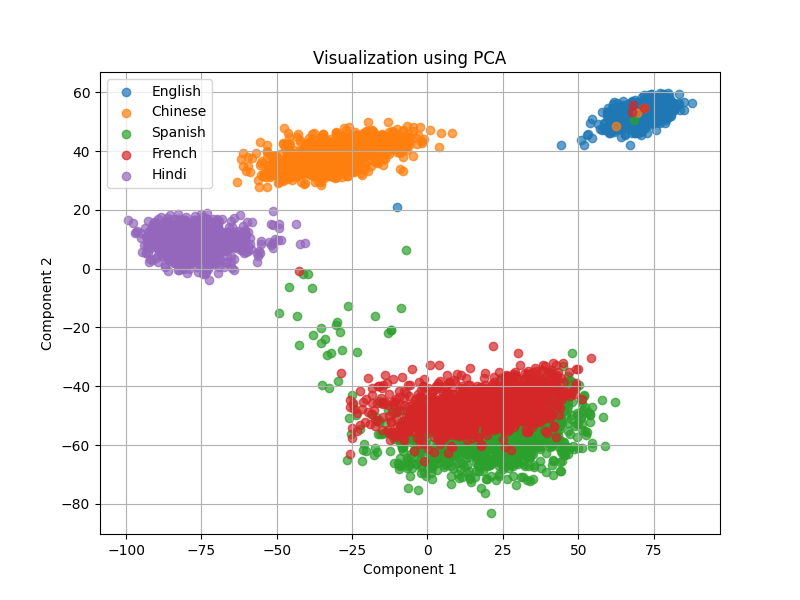}      \caption{PCA, Layer 25}        \end{subfigure}    \vspace{0.2in}    %
\begin{subfigure}{0.18\textwidth}      \includegraphics[width=\textwidth]{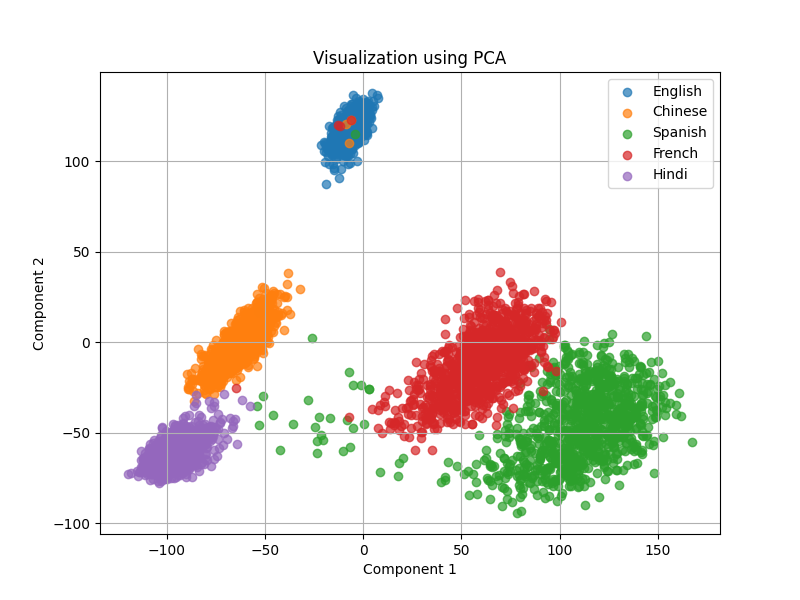}      \caption{PCA, Layer 26}        \end{subfigure}  \hfill  \begin{subfigure}{0.18\textwidth}      \includegraphics[width=\textwidth]{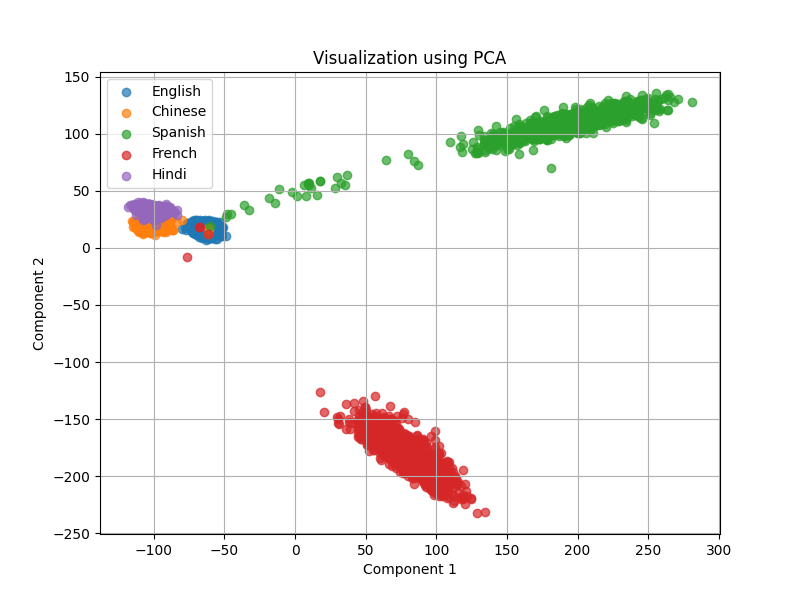}      \caption{PCA, Layer 27}        \end{subfigure}  \hfill  \begin{subfigure}{0.18\textwidth}      \includegraphics[width=\textwidth]{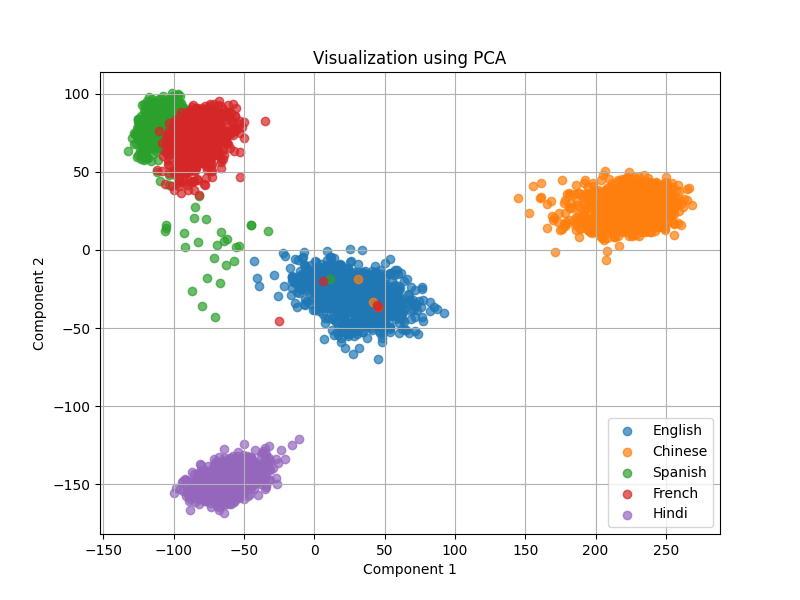}      \caption{PCA, Layer 28}        \end{subfigure}      \caption{PCA visualizations for layers 1-28 of Qwen2-7B-Instruct on the GSM8K dataset.}  
\end{figure*}

\begin{figure*}[htbp]
\centering
\begin{subfigure}{0.18\textwidth}
\includegraphics[width=\textwidth]{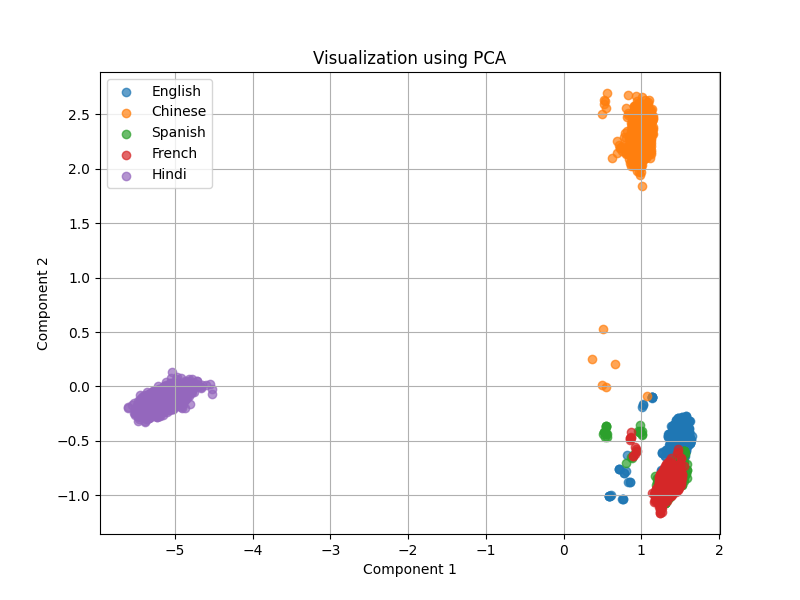}
\caption{PCA, Layer 1}
\end{subfigure}
\hfill
\begin{subfigure}{0.18\textwidth}
\includegraphics[width=\textwidth]{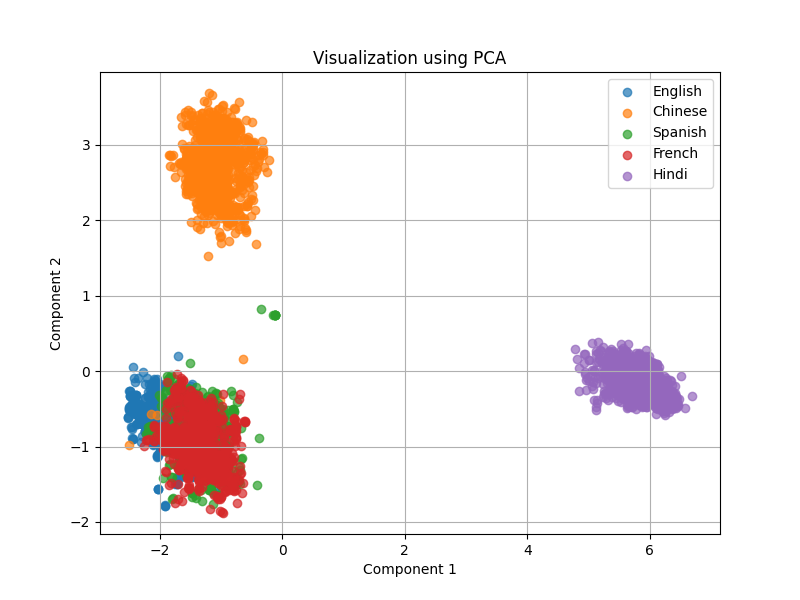}
\caption{PCA, Layer 2}

\end{subfigure}
\hfill
\begin{subfigure}{0.18\textwidth}
\includegraphics[width=\textwidth]{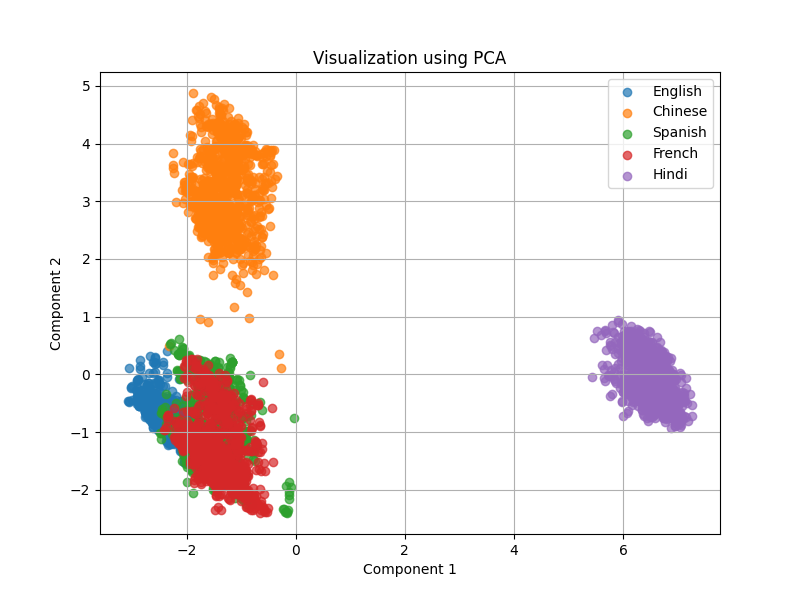}
\caption{PCA, Layer 3}

\end{subfigure}
\hfill
\begin{subfigure}{0.18\textwidth}
\includegraphics[width=\textwidth]{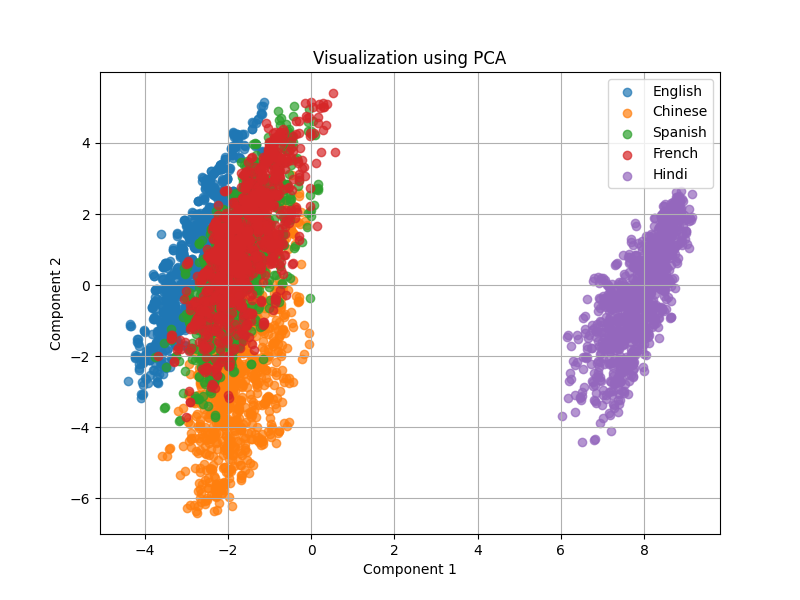}
\caption{PCA, Layer 4}

\end{subfigure}
\hfill
\begin{subfigure}{0.18\textwidth}
\includegraphics[width=\textwidth]{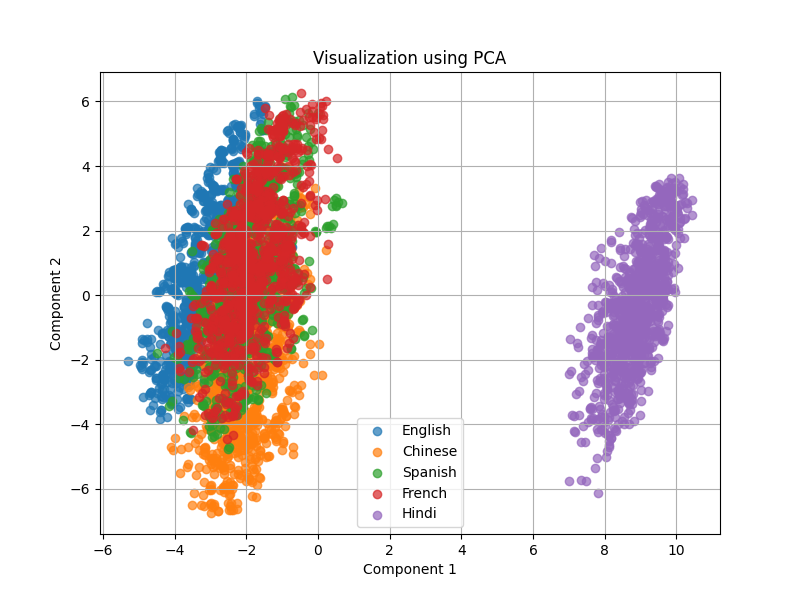}
\caption{PCA, Layer 5}

\end{subfigure}
\vspace{0.2in} %
\begin{subfigure}{0.18\textwidth}      \includegraphics[width=\textwidth]{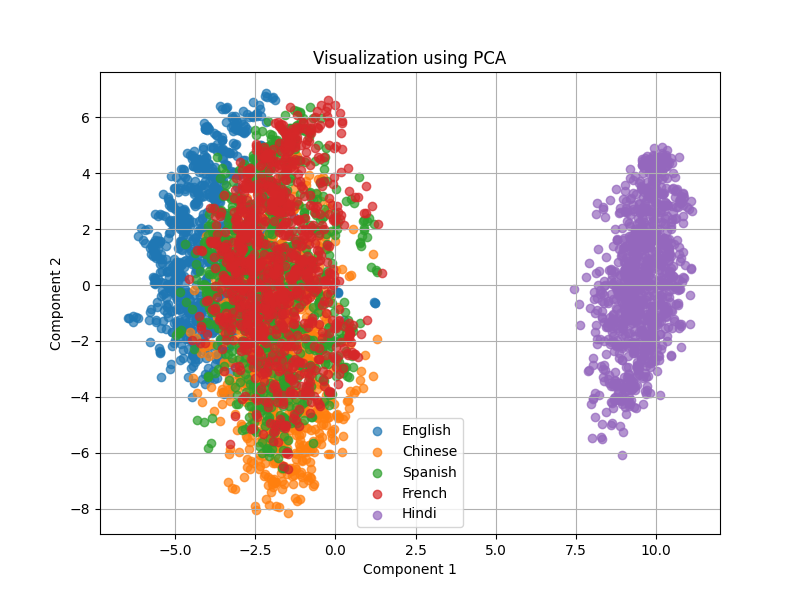}      \caption{PCA, Layer 6}        \end{subfigure}  \hfill  \begin{subfigure}{0.18\textwidth}      \includegraphics[width=\textwidth]{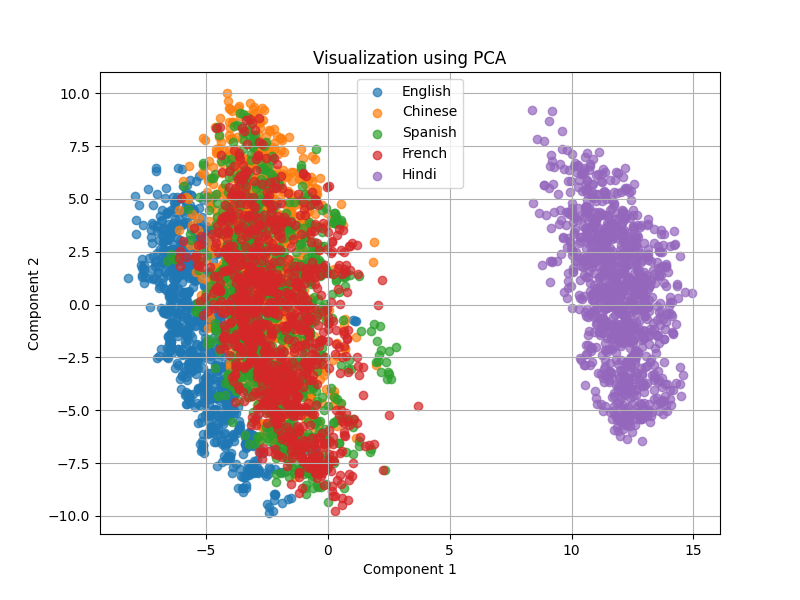}      \caption{PCA, Layer 7}        \end{subfigure}  \hfill  \begin{subfigure}{0.18\textwidth}      \includegraphics[width=\textwidth]{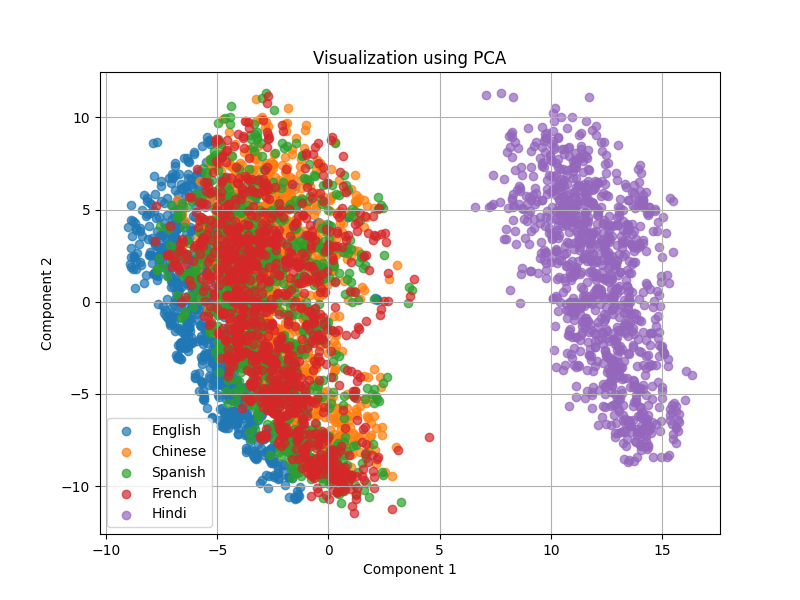}      \caption{PCA, Layer 8}        \end{subfigure}  \hfill  \begin{subfigure}{0.18\textwidth}      \includegraphics[width=\textwidth]{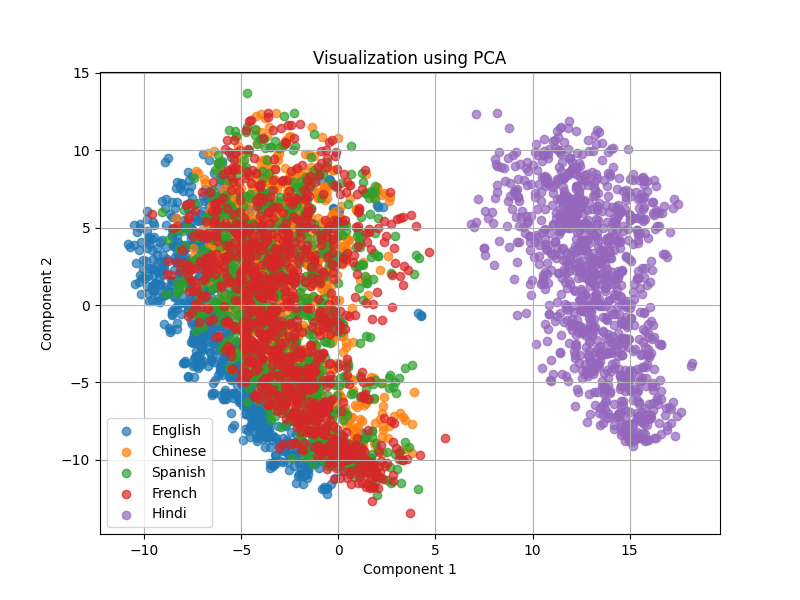}      \caption{PCA, Layer 9}        \end{subfigure}  \hfill  \begin{subfigure}{0.18\textwidth}      \includegraphics[width=\textwidth]{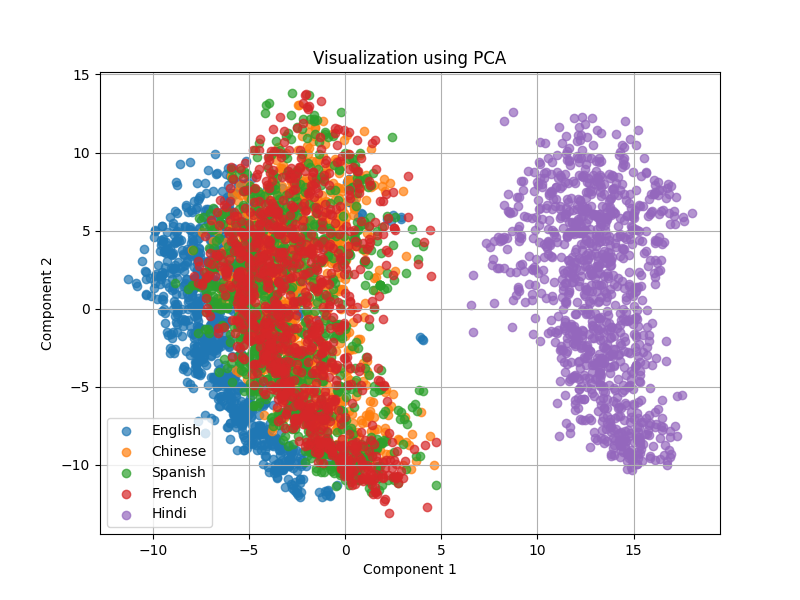}      \caption{PCA, Layer 10}        \end{subfigure}    \vspace{0.2in}    %
\begin{subfigure}{0.18\textwidth}      \includegraphics[width=\textwidth]{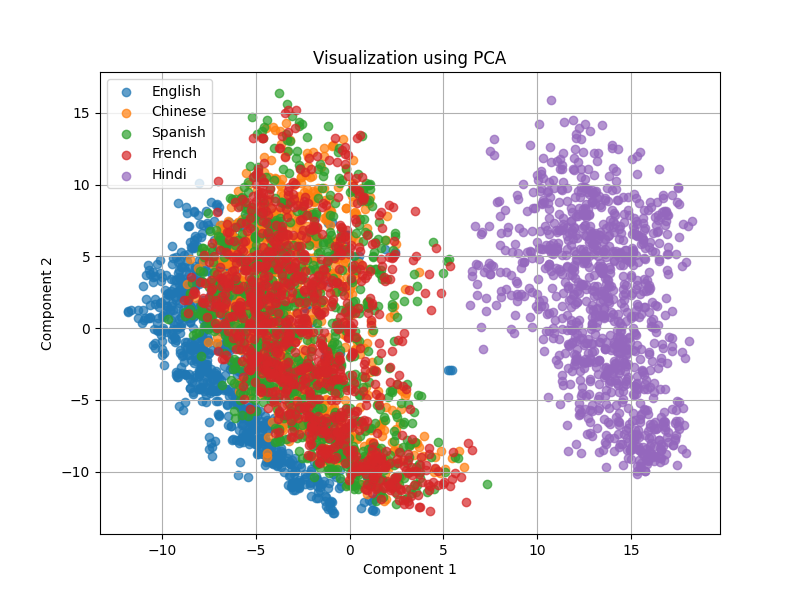}      \caption{PCA, Layer 11}        \end{subfigure}  \hfill  \begin{subfigure}{0.18\textwidth}      \includegraphics[width=\textwidth]{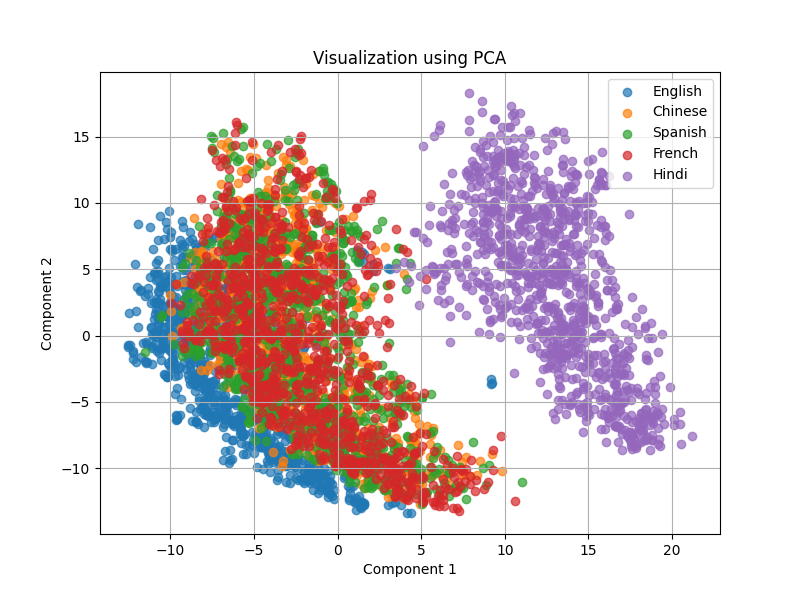}      \caption{PCA, Layer 12}        \end{subfigure}  \hfill  \begin{subfigure}{0.18\textwidth}      \includegraphics[width=\textwidth]{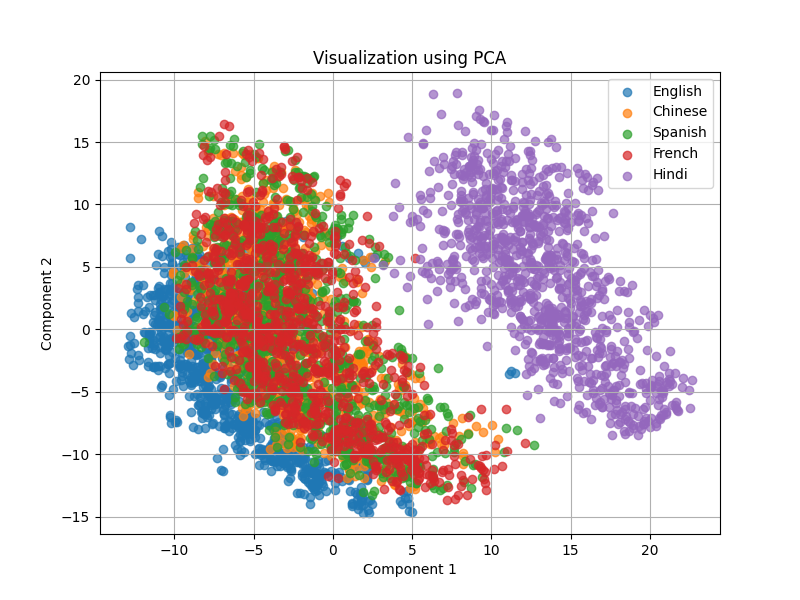}      \caption{PCA, Layer 13}        \end{subfigure}  \hfill  \begin{subfigure}{0.18\textwidth}      \includegraphics[width=\textwidth]{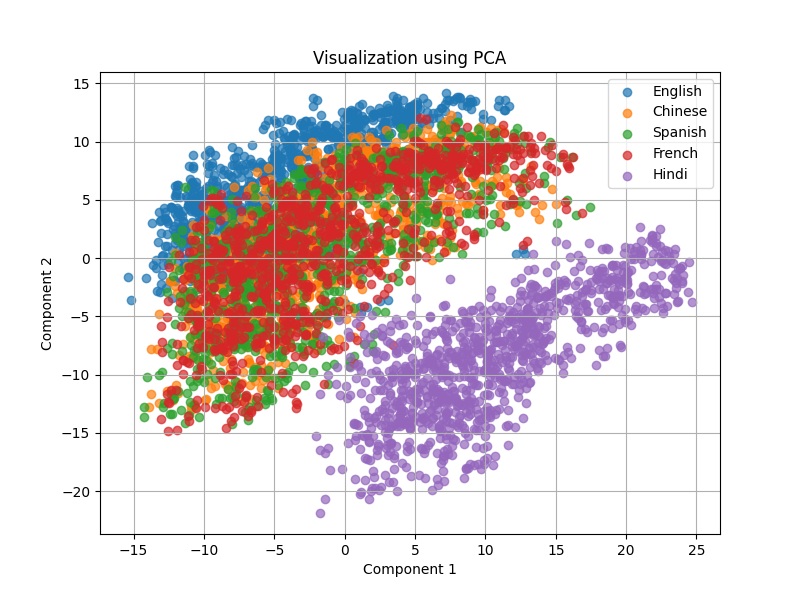}      \caption{PCA, Layer 14}        \end{subfigure}  \hfill  \begin{subfigure}{0.18\textwidth}      \includegraphics[width=\textwidth]{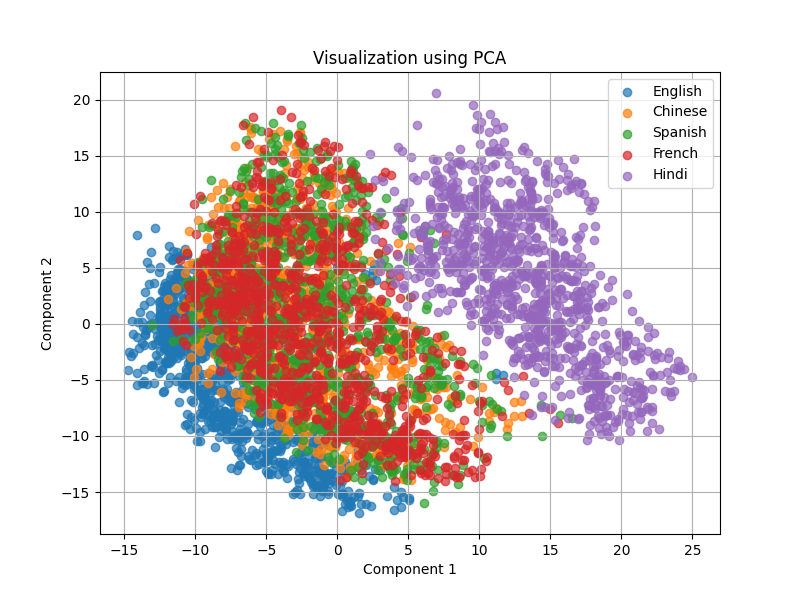}      \caption{PCA, Layer 15}        \end{subfigure}    \vspace{0.2in}    %
\begin{subfigure}{0.18\textwidth}      \includegraphics[width=\textwidth]{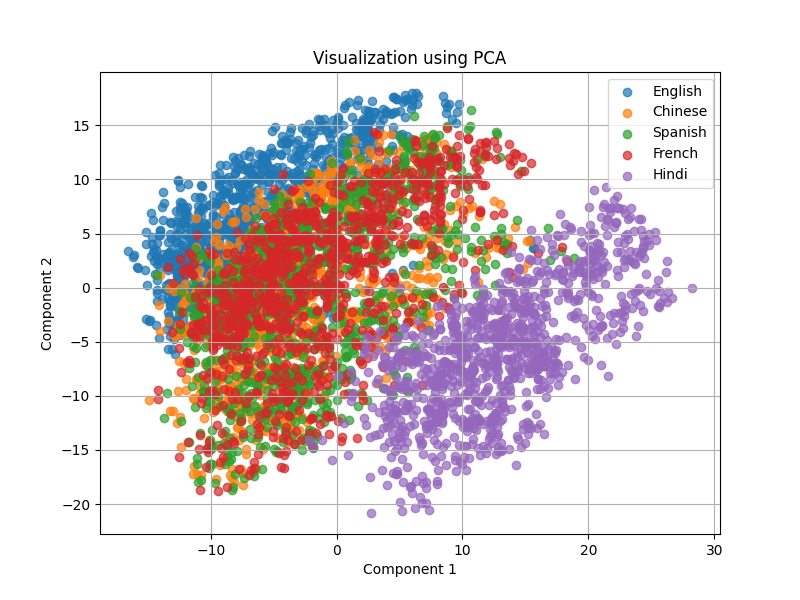}      \caption{PCA, Layer 16}        \end{subfigure}  \hfill  \begin{subfigure}{0.18\textwidth}      \includegraphics[width=\textwidth]{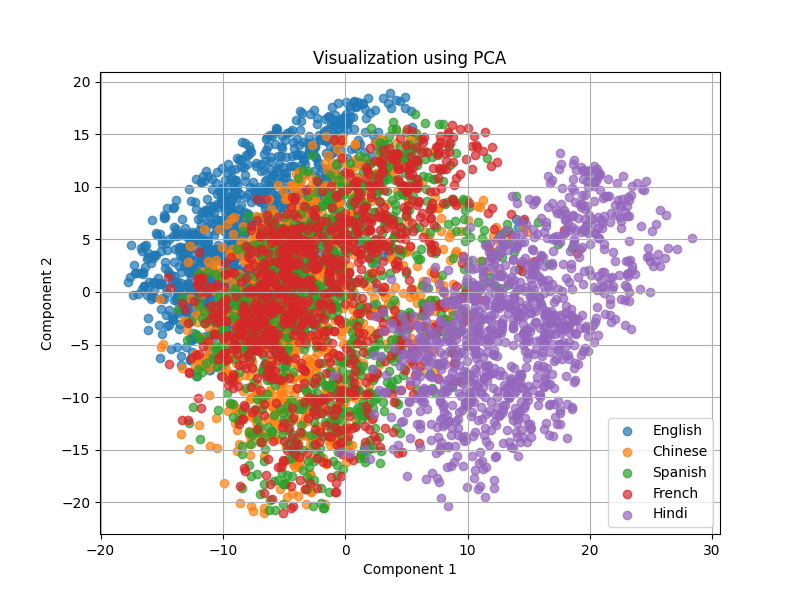}      \caption{PCA, Layer 17}        \end{subfigure}  \hfill  \begin{subfigure}{0.18\textwidth}      \includegraphics[width=\textwidth]{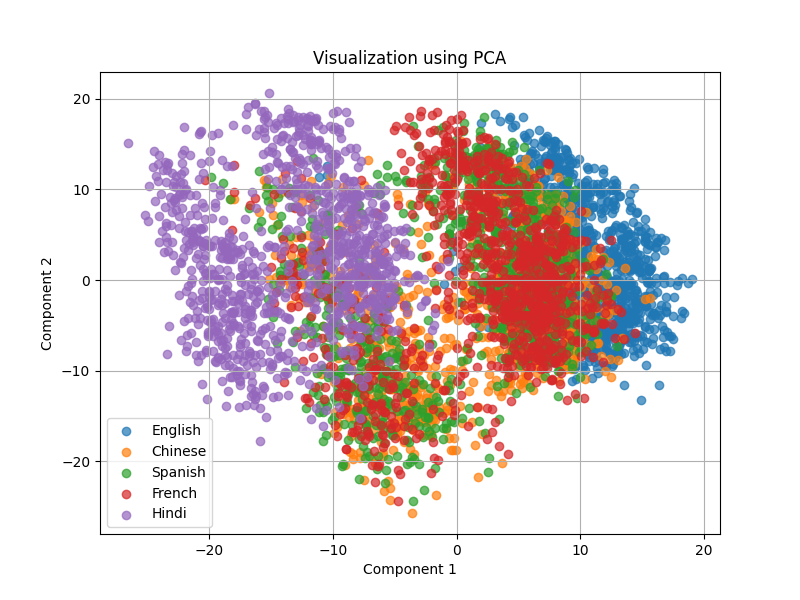}      \caption{PCA, Layer 18}        \end{subfigure}  \hfill  \begin{subfigure}{0.18\textwidth}      \includegraphics[width=\textwidth]{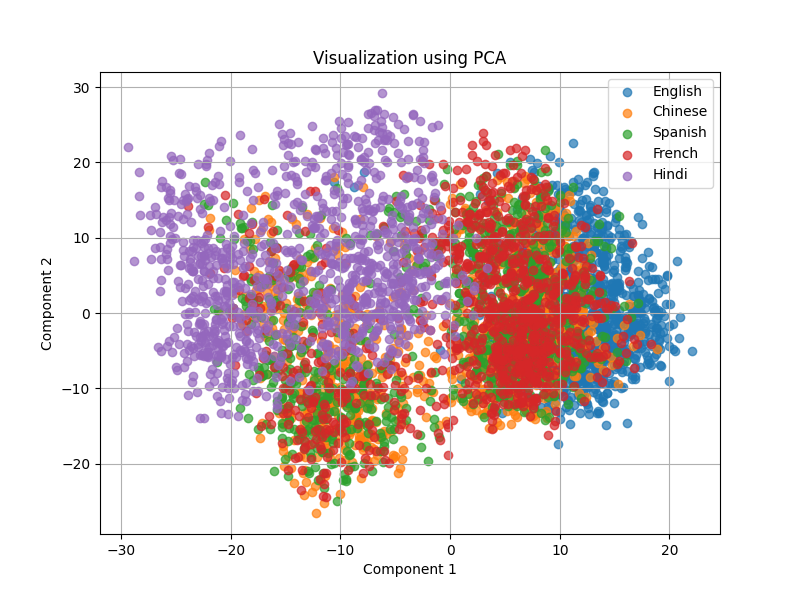}      \caption{PCA, Layer 19}        \end{subfigure}  \hfill  \begin{subfigure}{0.18\textwidth}      \includegraphics[width=\textwidth]{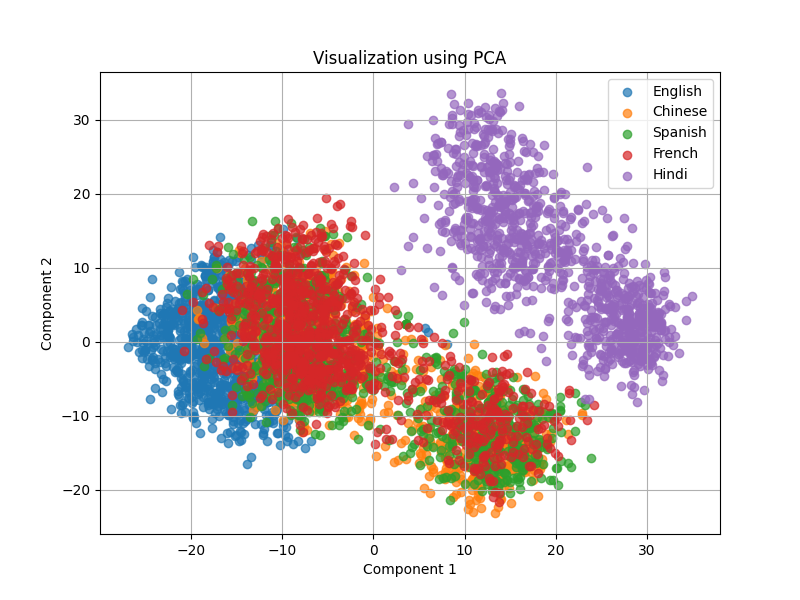}      \caption{PCA, Layer 20}        \end{subfigure}    \vspace{0.2in}    %
\begin{subfigure}{0.18\textwidth}      \includegraphics[width=\textwidth]{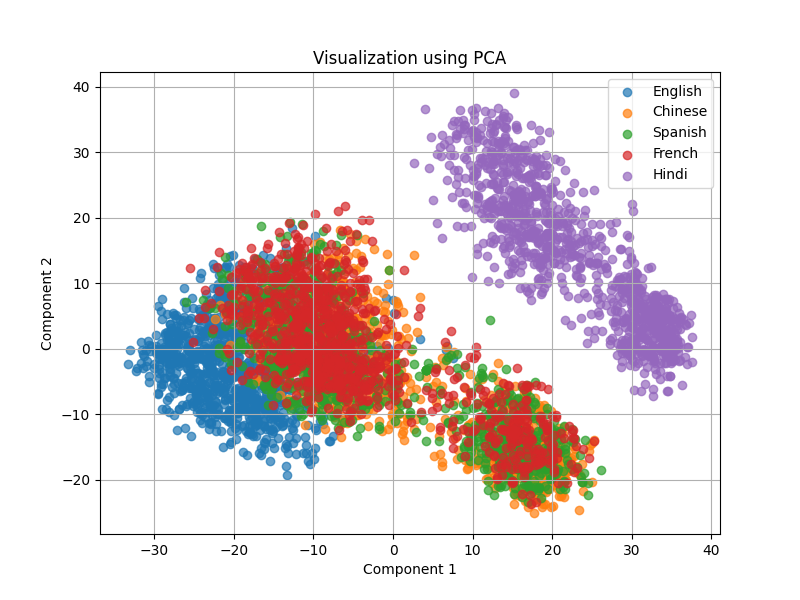}      \caption{PCA, Layer 21}        \end{subfigure}  \hfill  \begin{subfigure}{0.18\textwidth}      \includegraphics[width=\textwidth]{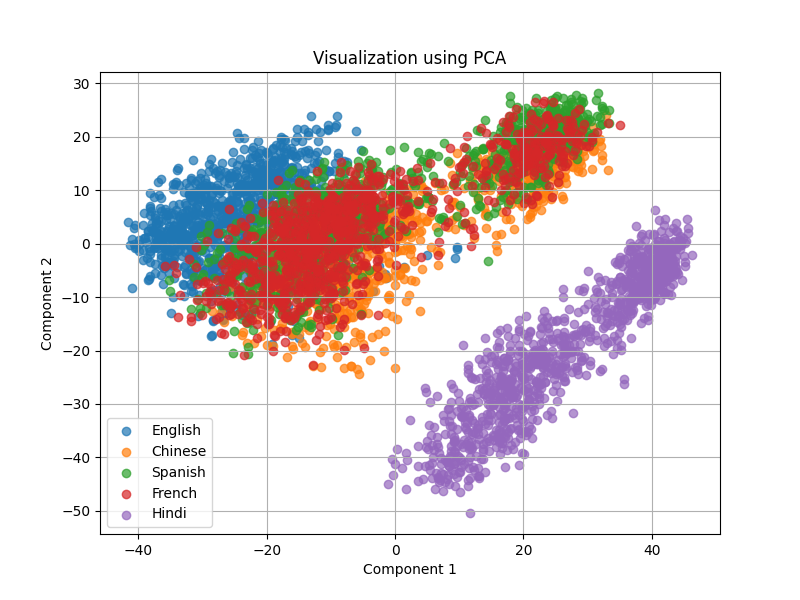}      \caption{PCA, Layer 22}        \end{subfigure}  \hfill  \begin{subfigure}{0.18\textwidth}      \includegraphics[width=\textwidth]{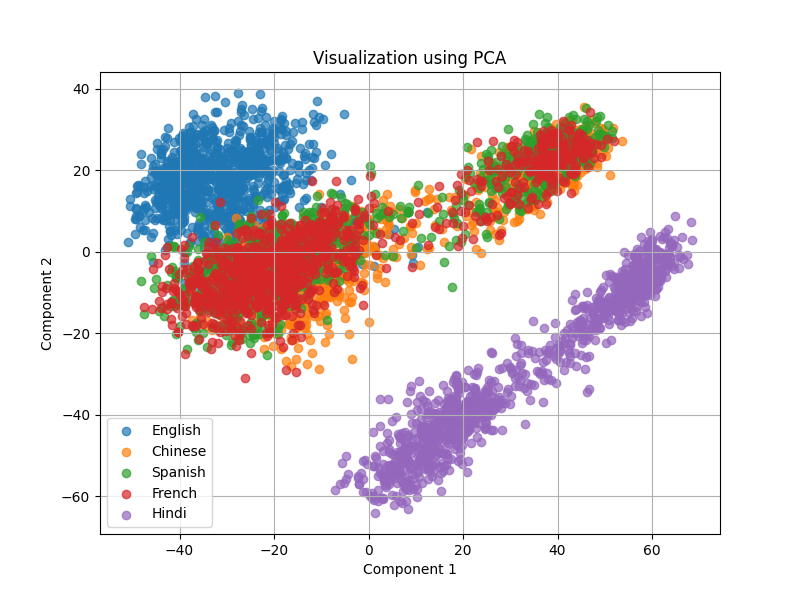}      \caption{PCA, Layer 23}        \end{subfigure}  \hfill  \begin{subfigure}{0.18\textwidth}      \includegraphics[width=\textwidth]{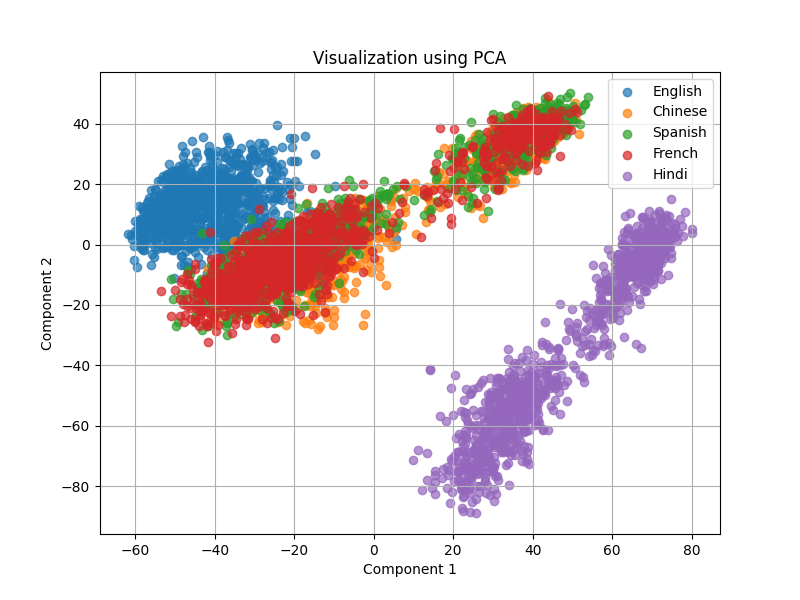}      \caption{PCA, Layer 24}        \end{subfigure}  \hfill  \begin{subfigure}{0.18\textwidth}      \includegraphics[width=\textwidth]{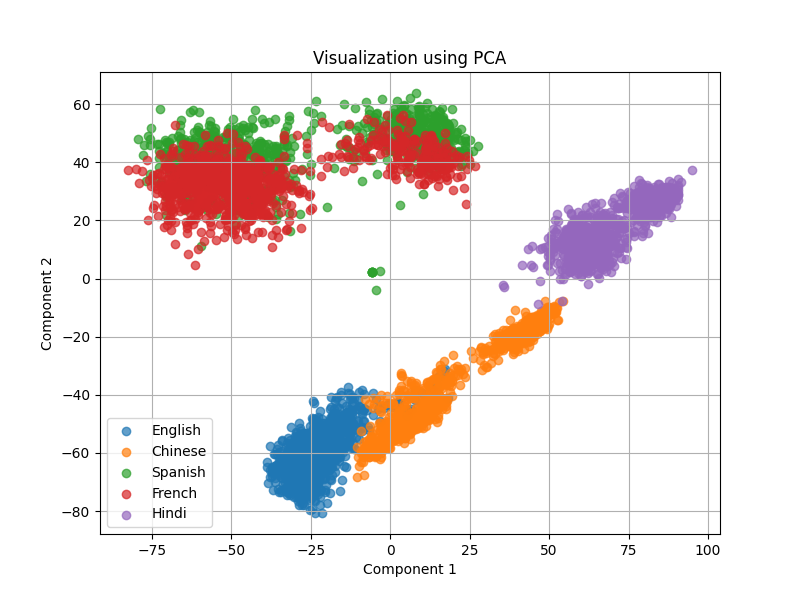}      \caption{PCA, Layer 25}        \end{subfigure}    \vspace{0.2in}    %
\begin{subfigure}{0.18\textwidth}      \includegraphics[width=\textwidth]{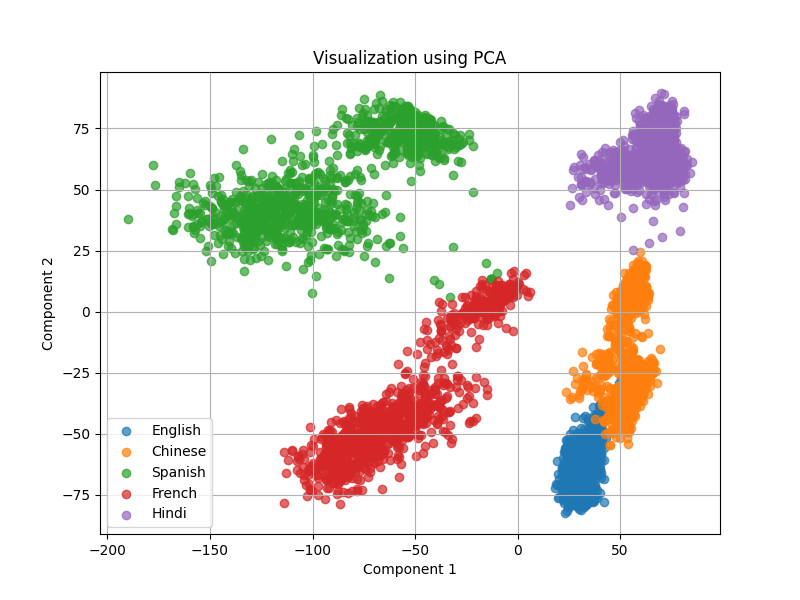}      \caption{PCA, Layer 26}        \end{subfigure}  \hfill  \begin{subfigure}{0.18\textwidth}      \includegraphics[width=\textwidth]{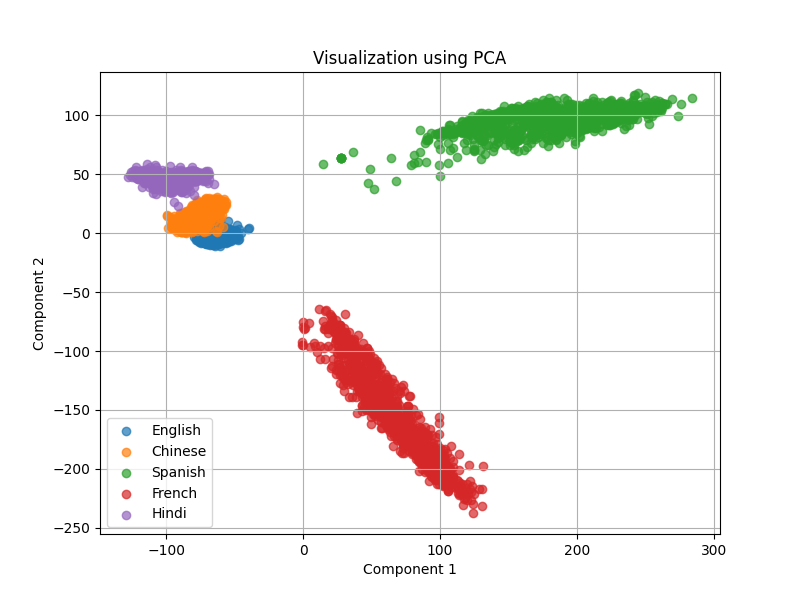}      \caption{PCA, Layer 27}        \end{subfigure}  \hfill  \begin{subfigure}{0.18\textwidth}      \includegraphics[width=\textwidth]{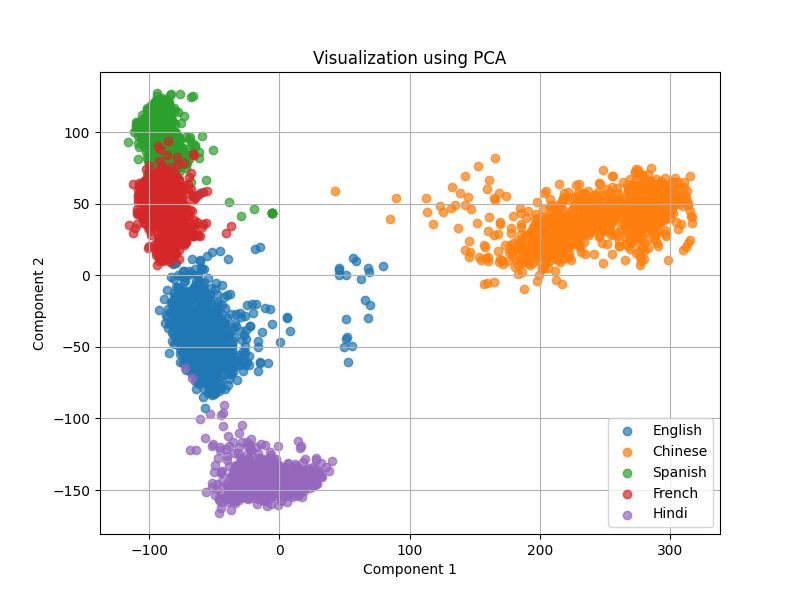}      \caption{PCA, Layer 28}        \end{subfigure}      \caption{PCA visualizations for layers 1-28 of Qwen2-7B-Instruct on the FOLIO dataset.}  
\end{figure*}

\begin{figure*}[htbp]
\centering
\begin{subfigure}{0.18\textwidth}
\includegraphics[width=\textwidth]{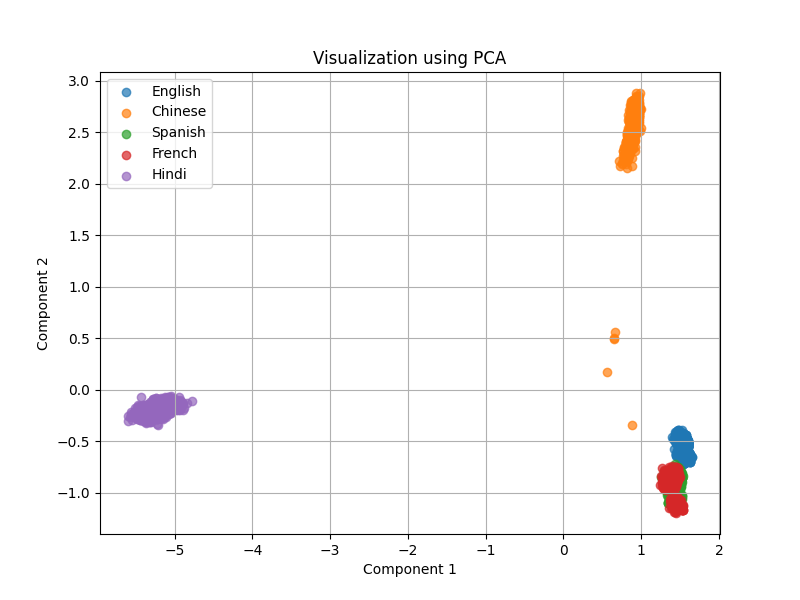}
\caption{PCA, Layer 1}
\end{subfigure}
\hfill
\begin{subfigure}{0.18\textwidth}
\includegraphics[width=\textwidth]{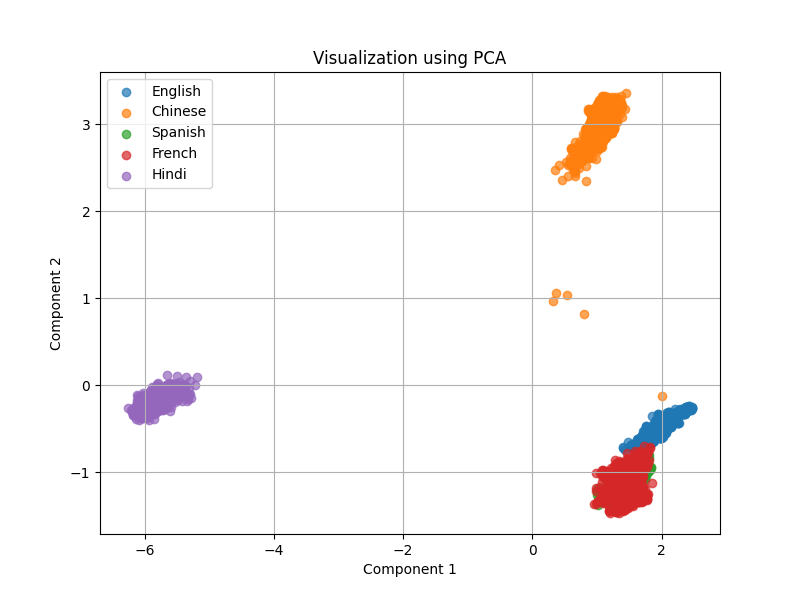}
\caption{PCA, Layer 2}

\end{subfigure}
\hfill
\begin{subfigure}{0.18\textwidth}
\includegraphics[width=\textwidth]{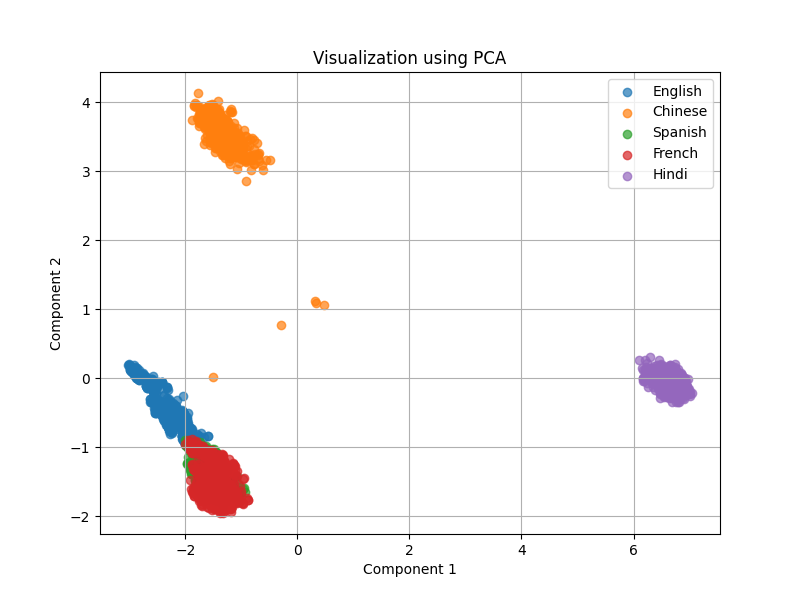}
\caption{PCA, Layer 3}

\end{subfigure}
\hfill
\begin{subfigure}{0.18\textwidth}
\includegraphics[width=\textwidth]{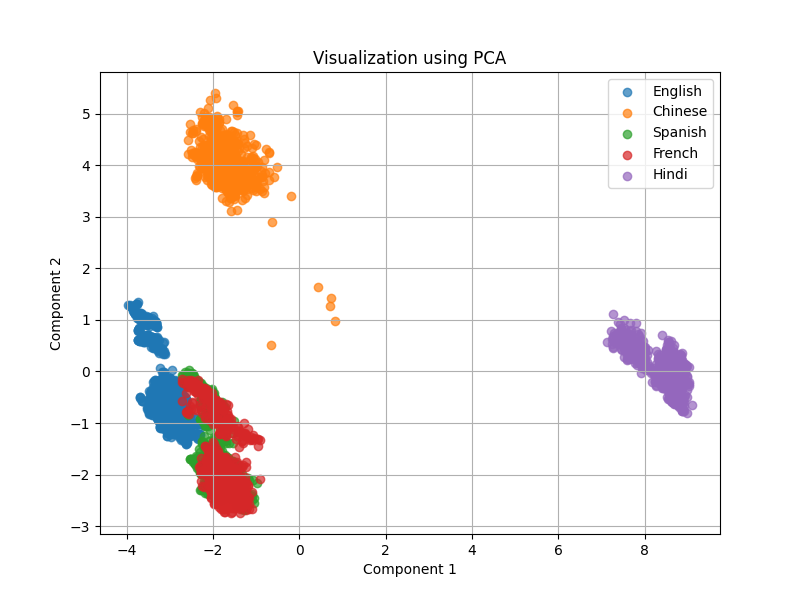}
\caption{PCA, Layer 4}

\end{subfigure}
\hfill
\begin{subfigure}{0.18\textwidth}
\includegraphics[width=\textwidth]{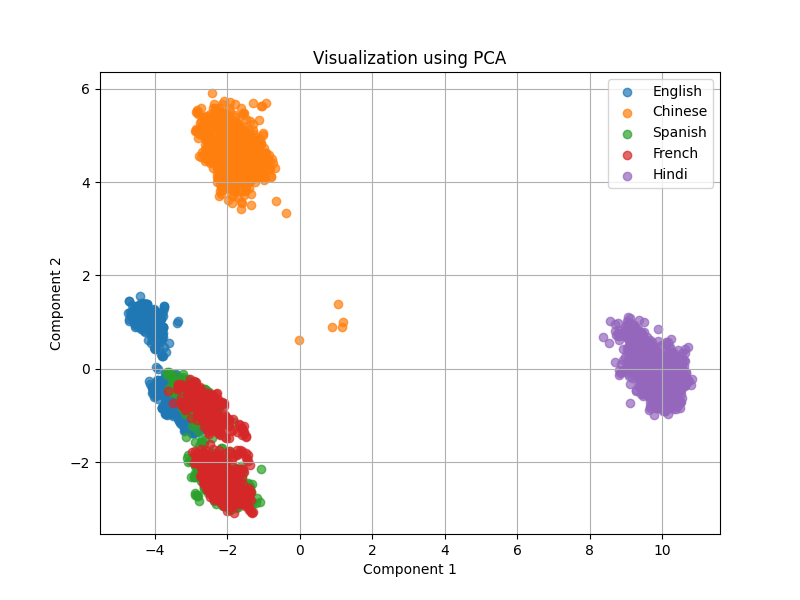}
\caption{PCA, Layer 5}

\end{subfigure}
\vspace{0.2in} %
\begin{subfigure}{0.18\textwidth}      \includegraphics[width=\textwidth]{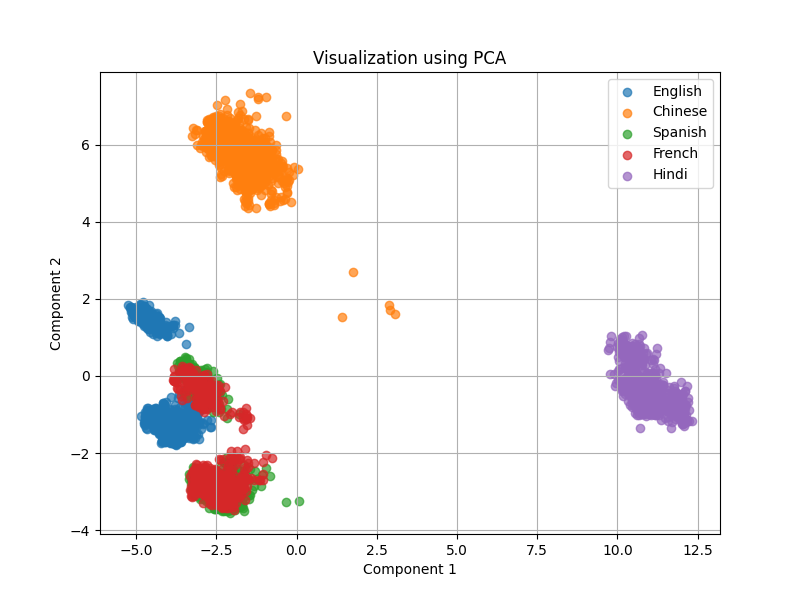}      \caption{PCA, Layer 6}        \end{subfigure}  \hfill  \begin{subfigure}{0.18\textwidth}      \includegraphics[width=\textwidth]{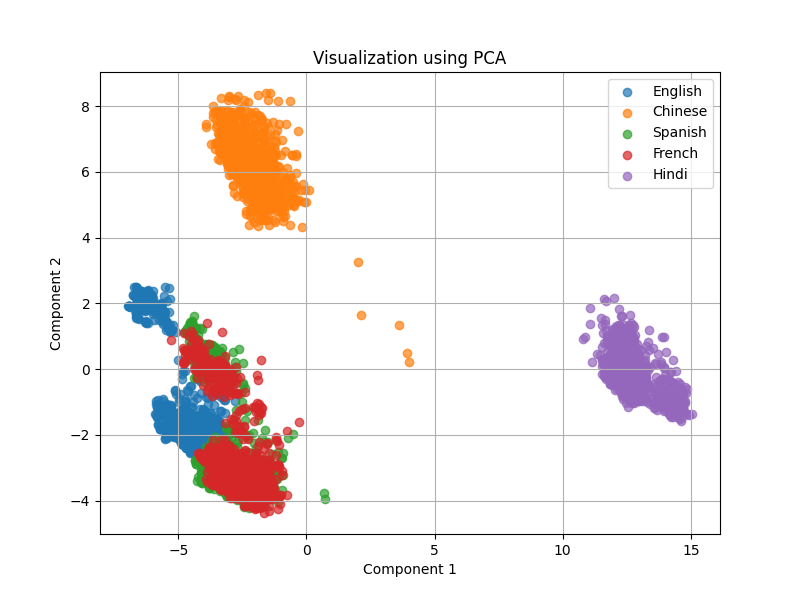}      \caption{PCA, Layer 7}        \end{subfigure}  \hfill  \begin{subfigure}{0.18\textwidth}      \includegraphics[width=\textwidth]{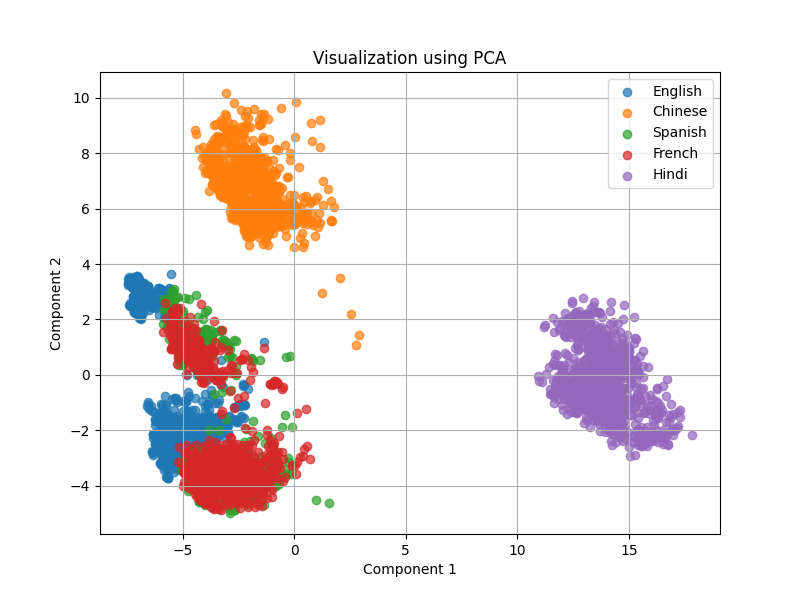}      \caption{PCA, Layer 8}        \end{subfigure}  \hfill  \begin{subfigure}{0.18\textwidth}      \includegraphics[width=\textwidth]{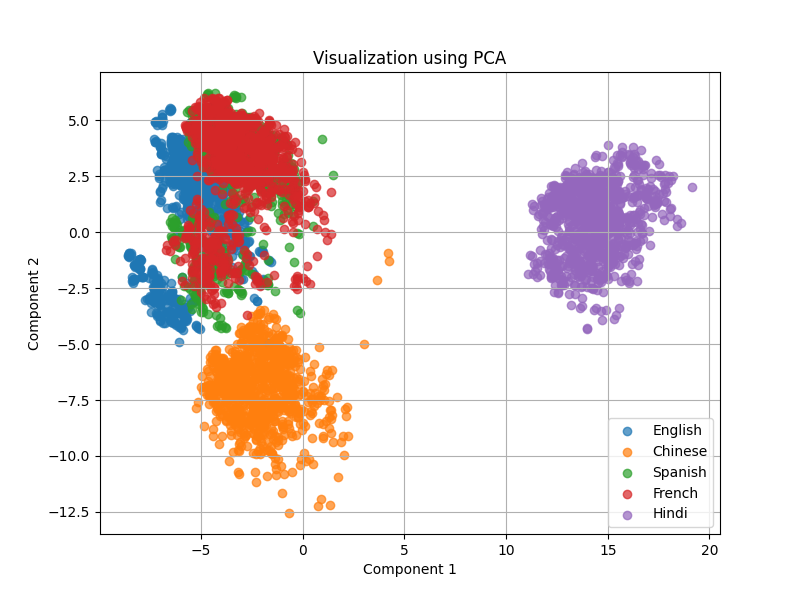}      \caption{PCA, Layer 9}        \end{subfigure}  \hfill  \begin{subfigure}{0.18\textwidth}      \includegraphics[width=\textwidth]{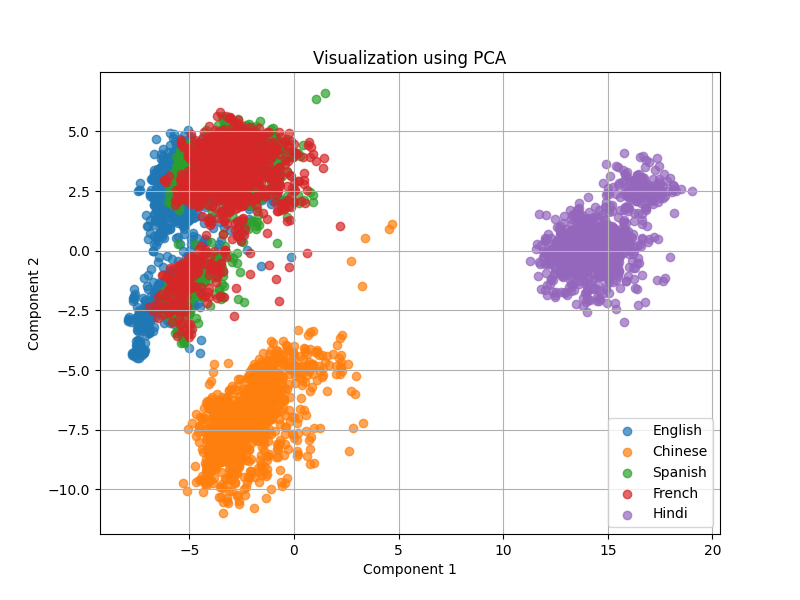}      \caption{PCA, Layer 10}        \end{subfigure}    \vspace{0.2in}    %
\begin{subfigure}{0.18\textwidth}      \includegraphics[width=\textwidth]{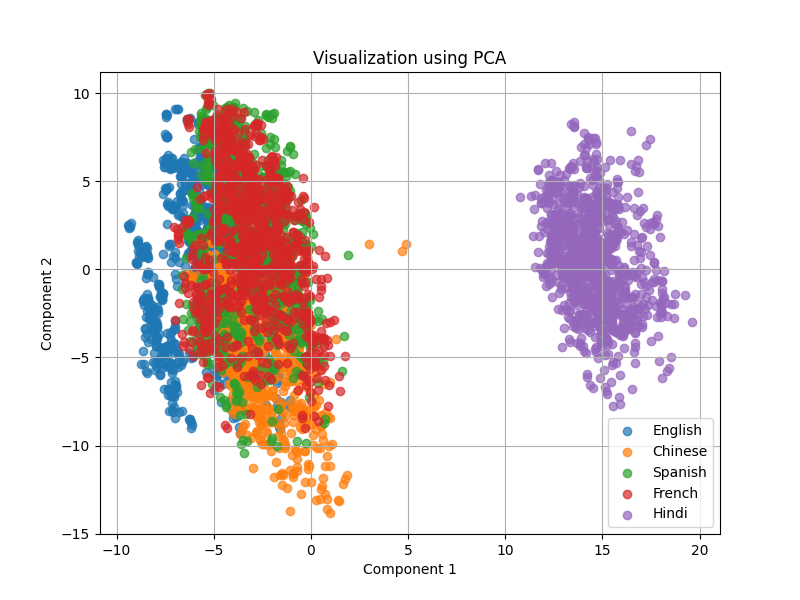}      \caption{PCA, Layer 11}        \end{subfigure}  \hfill  \begin{subfigure}{0.18\textwidth}      \includegraphics[width=\textwidth]{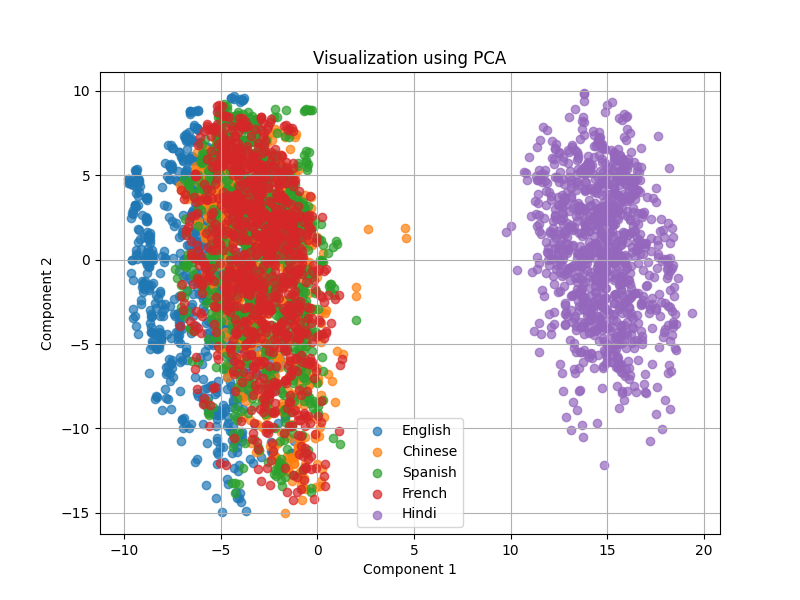}      \caption{PCA, Layer 12}        \end{subfigure}  \hfill  \begin{subfigure}{0.18\textwidth}      \includegraphics[width=\textwidth]{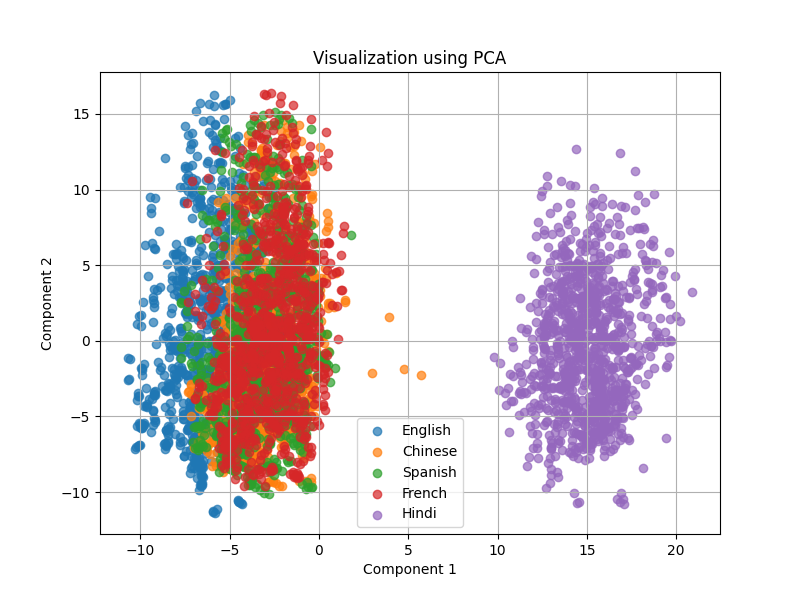}      \caption{PCA, Layer 13}        \end{subfigure}  \hfill  \begin{subfigure}{0.18\textwidth}      \includegraphics[width=\textwidth]{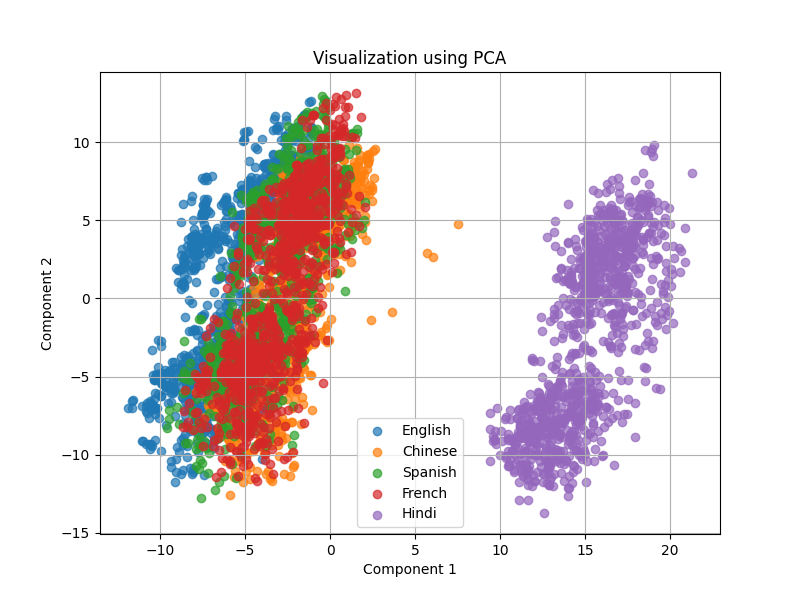}      \caption{PCA, Layer 14}        \end{subfigure}  \hfill  \begin{subfigure}{0.18\textwidth}      \includegraphics[width=\textwidth]{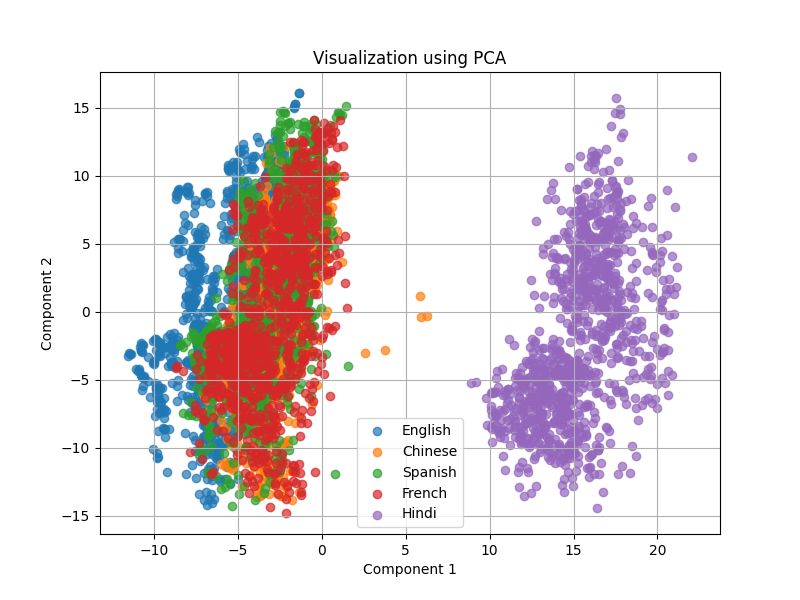}      \caption{PCA, Layer 15}        \end{subfigure}    \vspace{0.2in}    %
\begin{subfigure}{0.18\textwidth}      \includegraphics[width=\textwidth]{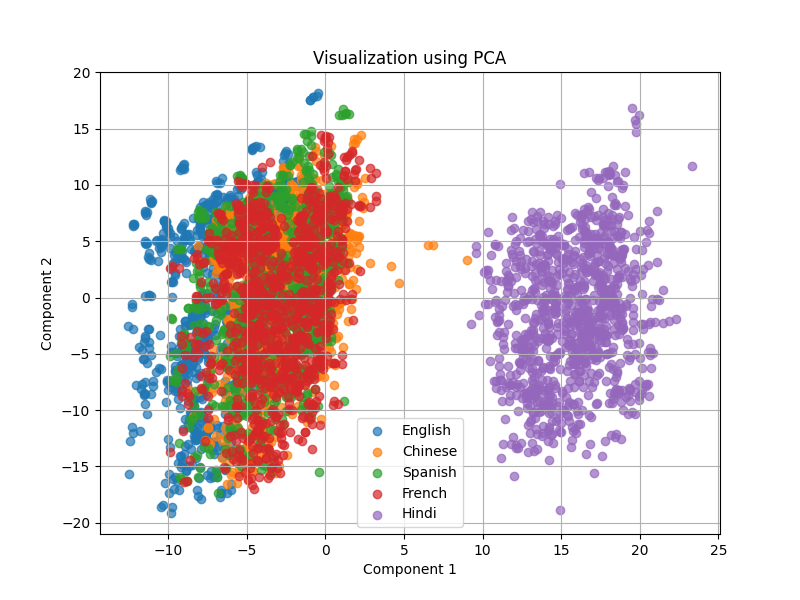}      \caption{PCA, Layer 16}        \end{subfigure}  \hfill  \begin{subfigure}{0.18\textwidth}      \includegraphics[width=\textwidth]{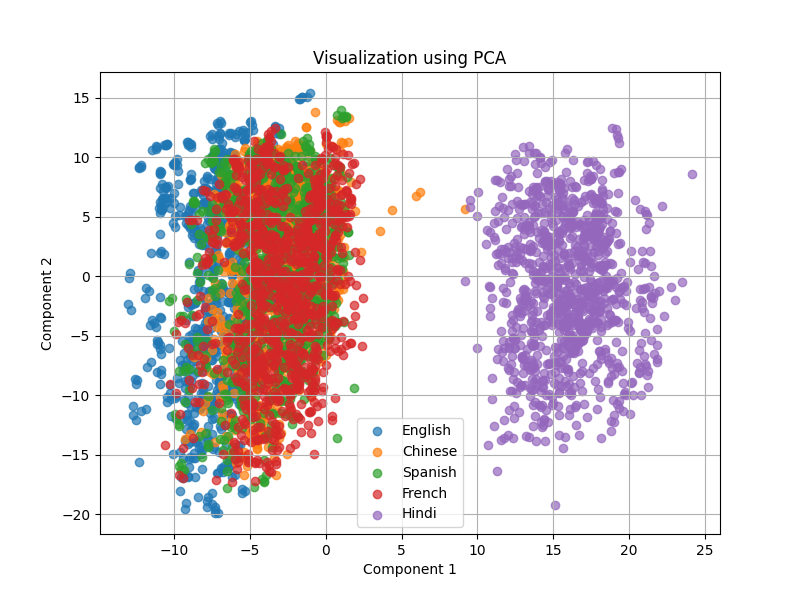}      \caption{PCA, Layer 17}        \end{subfigure}  \hfill  \begin{subfigure}{0.18\textwidth}      \includegraphics[width=\textwidth]{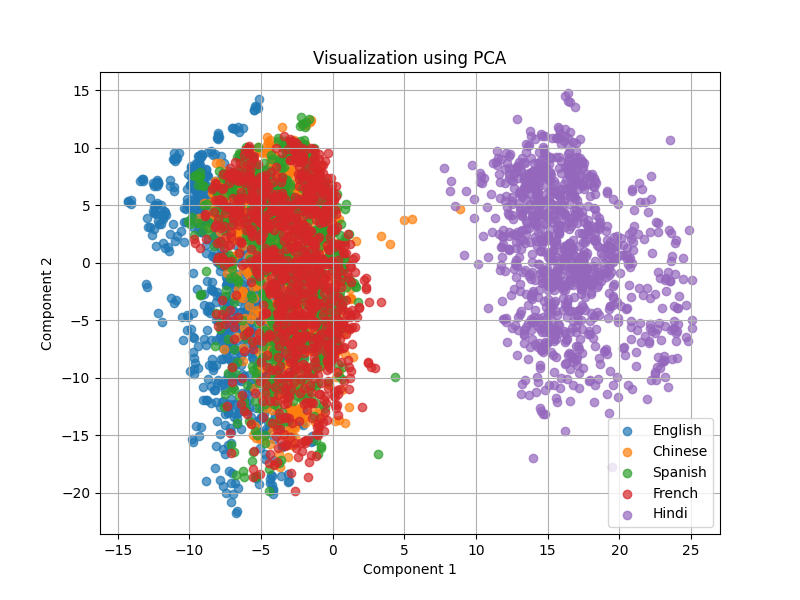}      \caption{PCA, Layer 18}        \end{subfigure}  \hfill  \begin{subfigure}{0.18\textwidth}      \includegraphics[width=\textwidth]{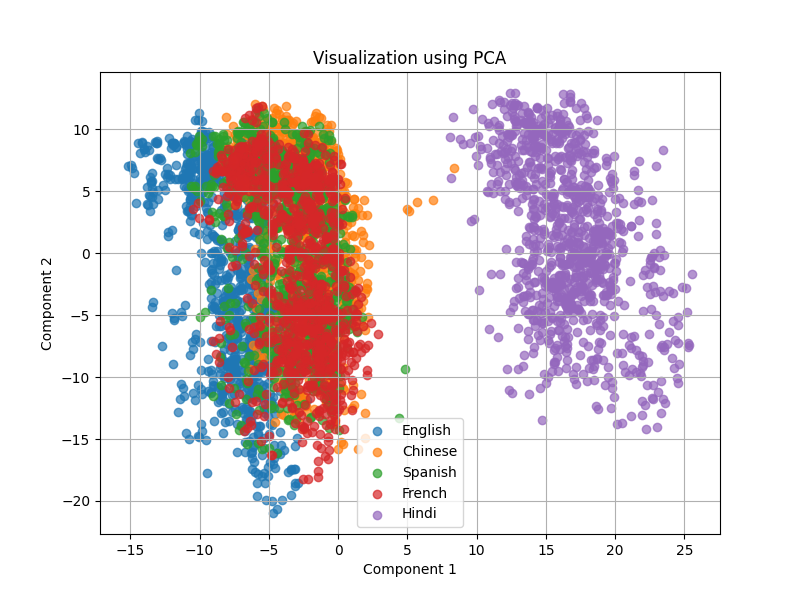}      \caption{PCA, Layer 19}        \end{subfigure}  \hfill  \begin{subfigure}{0.18\textwidth}      \includegraphics[width=\textwidth]{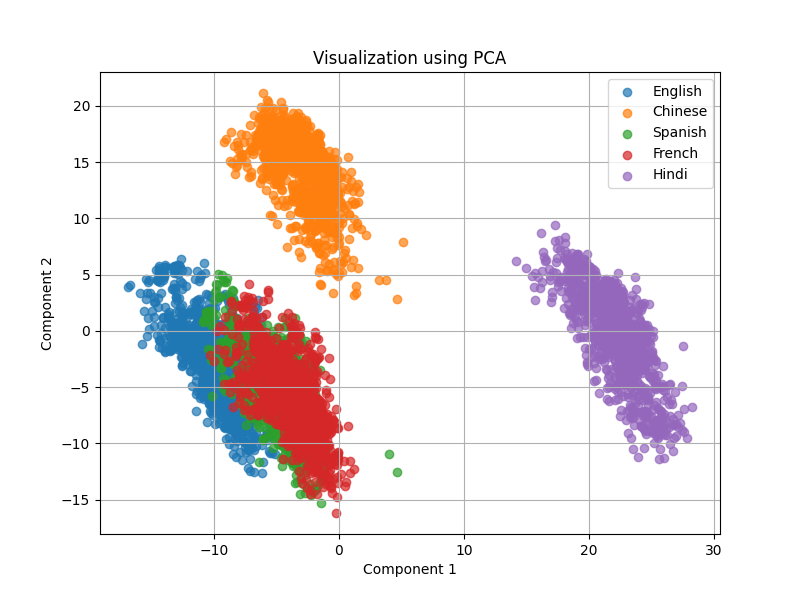}      \caption{PCA, Layer 20}        \end{subfigure}    \vspace{0.2in}    %
\begin{subfigure}{0.18\textwidth}      \includegraphics[width=\textwidth]{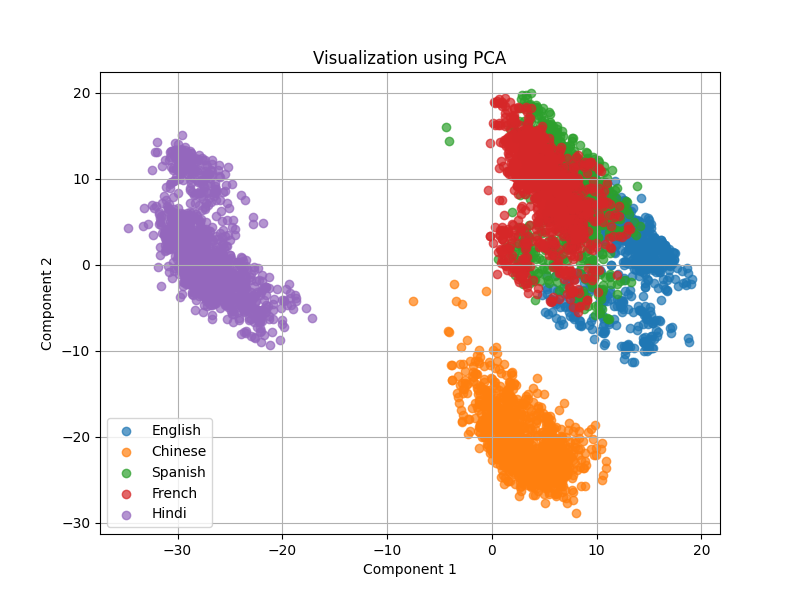}      \caption{PCA, Layer 21}        \end{subfigure}  \hfill  \begin{subfigure}{0.18\textwidth}      \includegraphics[width=\textwidth]{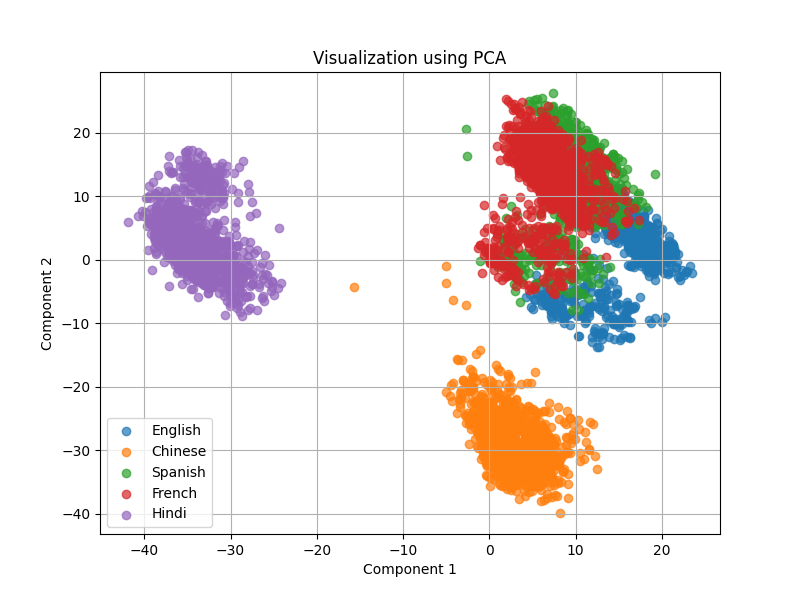}      \caption{PCA, Layer 22}        \end{subfigure}  \hfill  \begin{subfigure}{0.18\textwidth}      \includegraphics[width=\textwidth]{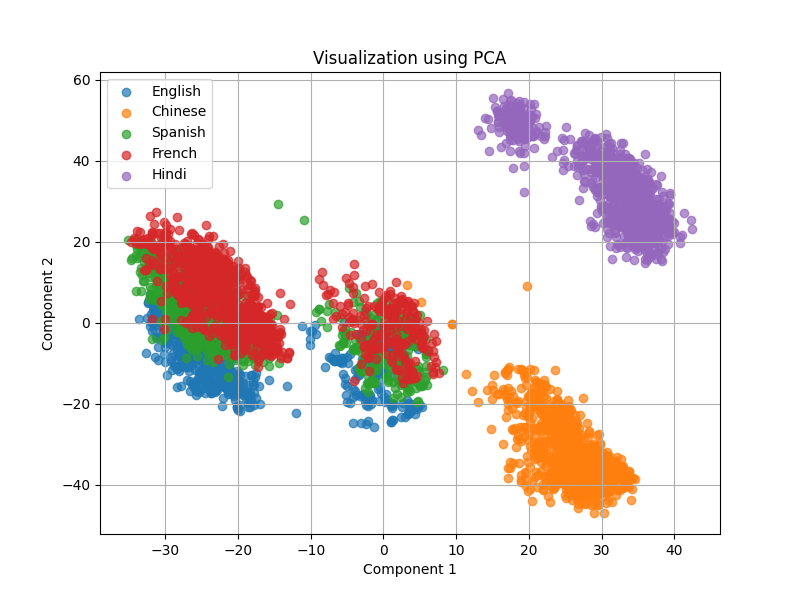}      \caption{PCA, Layer 23}        \end{subfigure}  \hfill  \begin{subfigure}{0.18\textwidth}      \includegraphics[width=\textwidth]{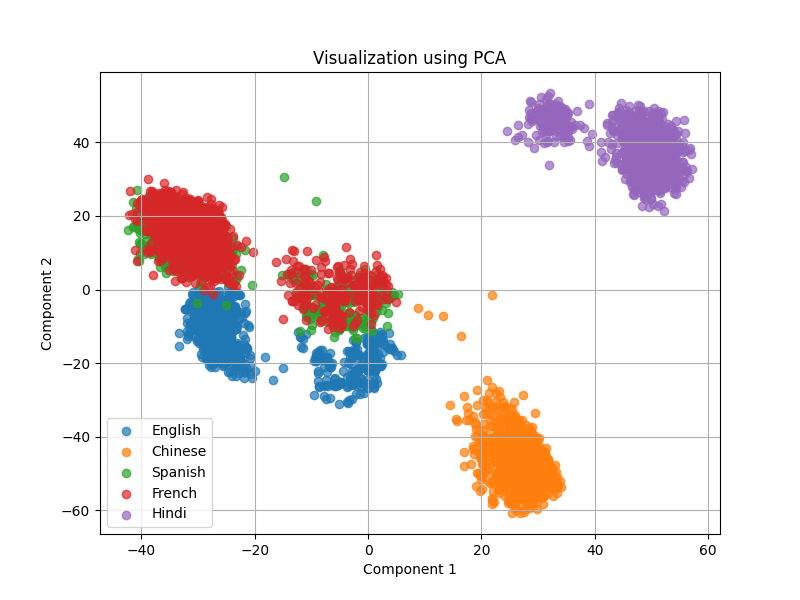}      \caption{PCA, Layer 24}        \end{subfigure}  \hfill  \begin{subfigure}{0.18\textwidth}      \includegraphics[width=\textwidth]{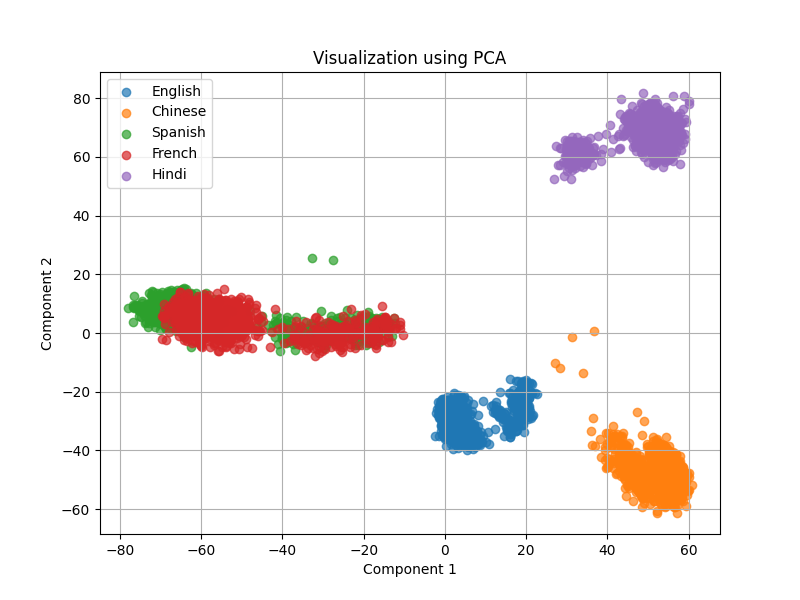}      \caption{PCA, Layer 25}        \end{subfigure}    \vspace{0.2in}    %
\begin{subfigure}{0.18\textwidth}      \includegraphics[width=\textwidth]{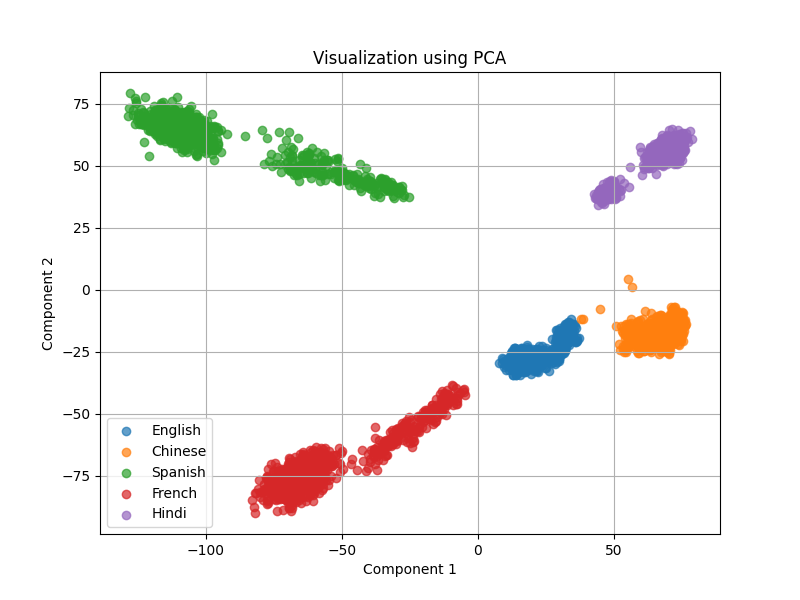}      \caption{PCA, Layer 26}        \end{subfigure}  \hfill  \begin{subfigure}{0.18\textwidth}      \includegraphics[width=\textwidth]{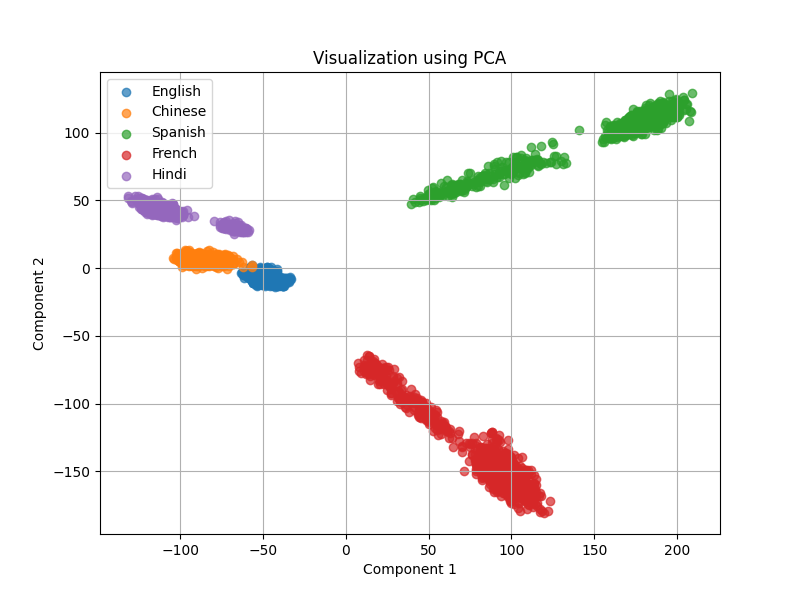}      \caption{PCA, Layer 27}        \end{subfigure}  \hfill  \begin{subfigure}{0.18\textwidth}      \includegraphics[width=\textwidth]{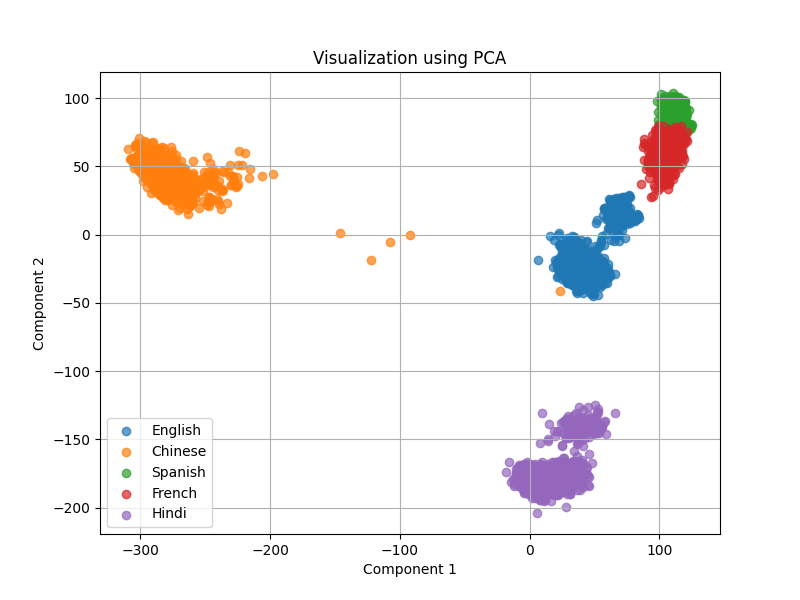}      \caption{PCA, Layer 28}        \end{subfigure}      \caption{PCA visualizations for layers 1-28 of Qwen2-7B-Instruct on the LogicalDeduction dataset.}  
\end{figure*}

\begin{figure*}[htbp]
\centering
\begin{subfigure}{0.18\textwidth}
\includegraphics[width=\textwidth]{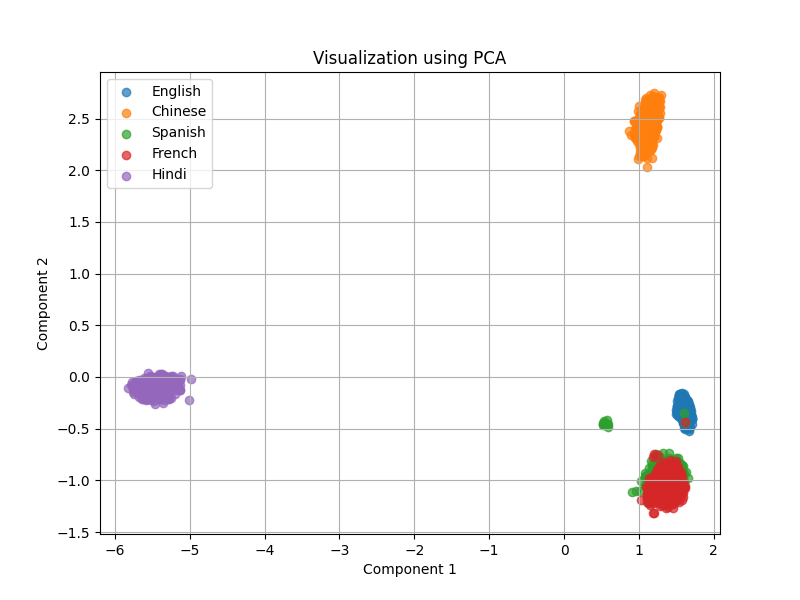}
\caption{PCA, Layer 1}
\end{subfigure}
\hfill
\begin{subfigure}{0.18\textwidth}
\includegraphics[width=\textwidth]{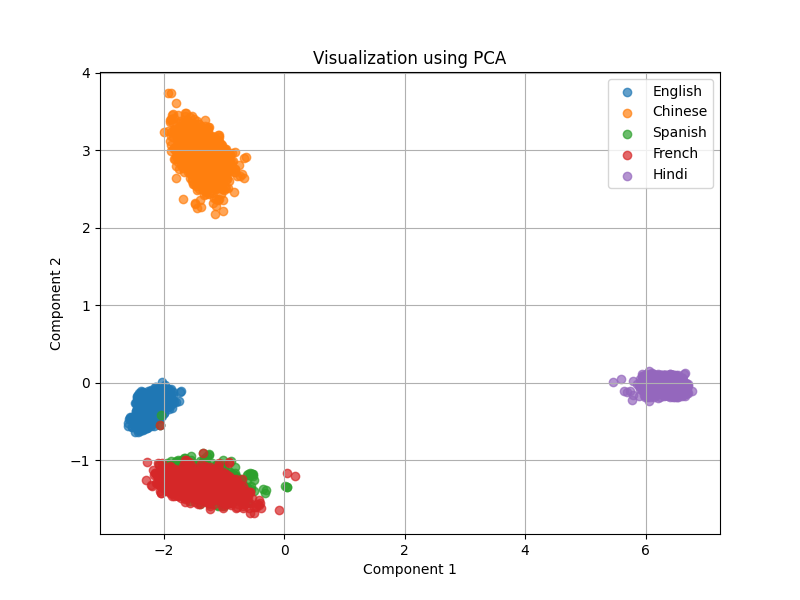}
\caption{PCA, Layer 2}

\end{subfigure}
\hfill
\begin{subfigure}{0.18\textwidth}
\includegraphics[width=\textwidth]{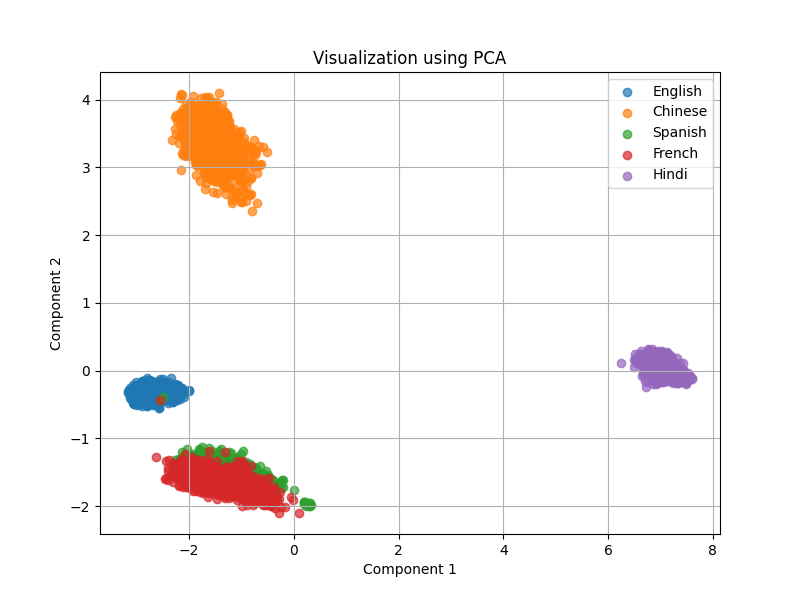}
\caption{PCA, Layer 3}

\end{subfigure}
\hfill
\begin{subfigure}{0.18\textwidth}
\includegraphics[width=\textwidth]{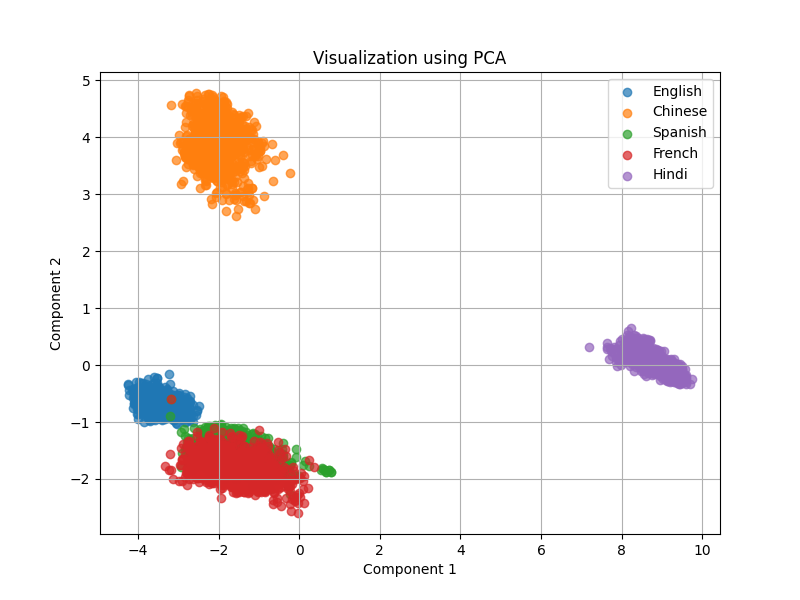}
\caption{PCA, Layer 4}

\end{subfigure}
\hfill
\begin{subfigure}{0.18\textwidth}
\includegraphics[width=\textwidth]{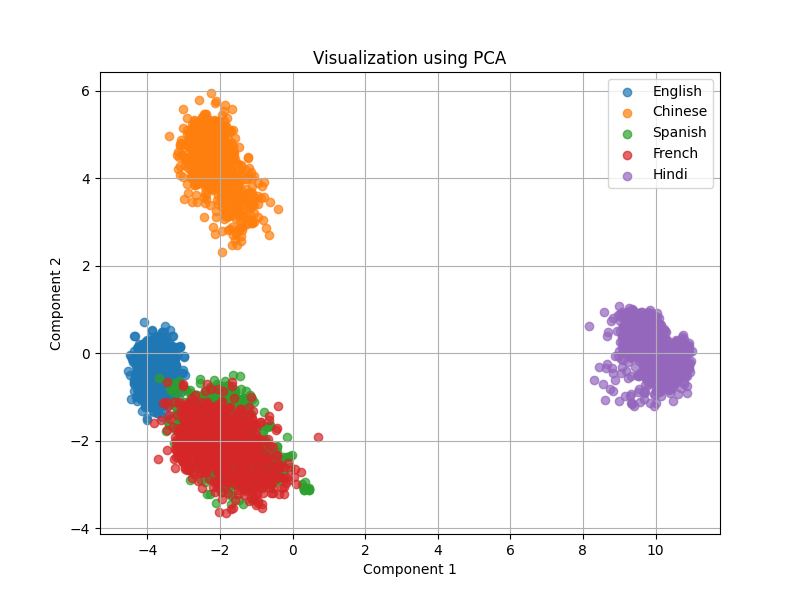}
\caption{PCA, Layer 5}

\end{subfigure}
\vspace{0.2in} %
\begin{subfigure}{0.18\textwidth}      \includegraphics[width=\textwidth]{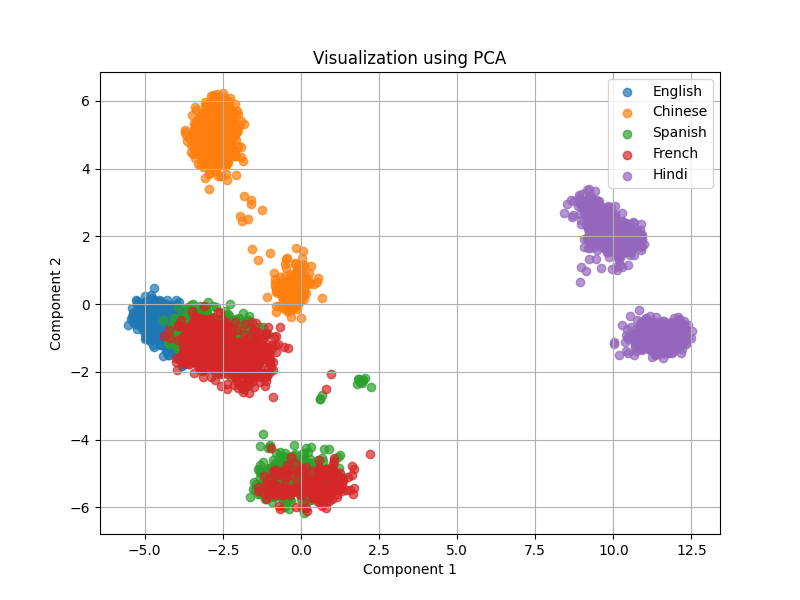}      \caption{PCA, Layer 6}        \end{subfigure}  \hfill  \begin{subfigure}{0.18\textwidth}      \includegraphics[width=\textwidth]{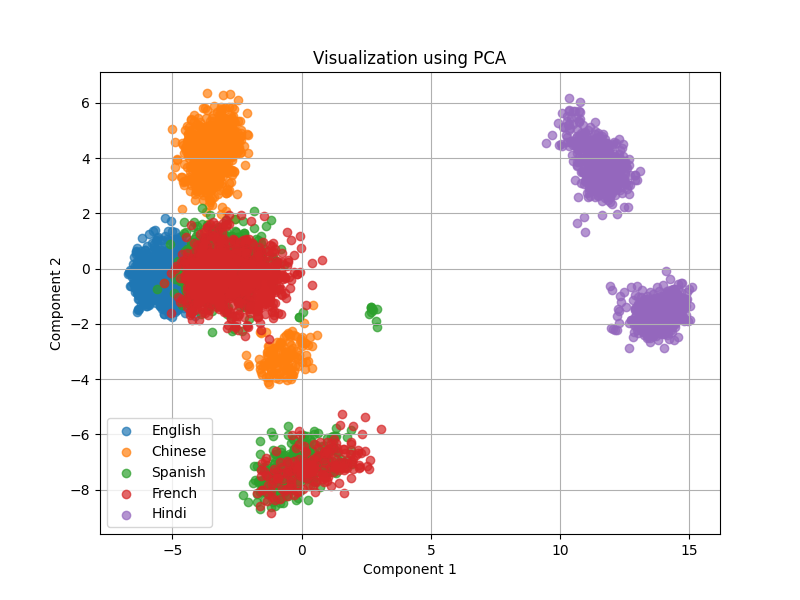}      \caption{PCA, Layer 7}        \end{subfigure}  \hfill  \begin{subfigure}{0.18\textwidth}      \includegraphics[width=\textwidth]{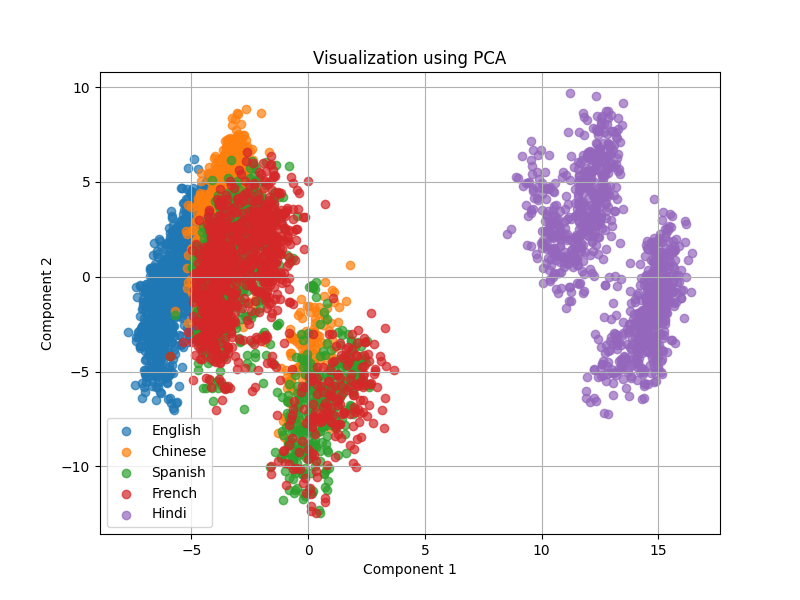}      \caption{PCA, Layer 8}        \end{subfigure}  \hfill  \begin{subfigure}{0.18\textwidth}      \includegraphics[width=\textwidth]{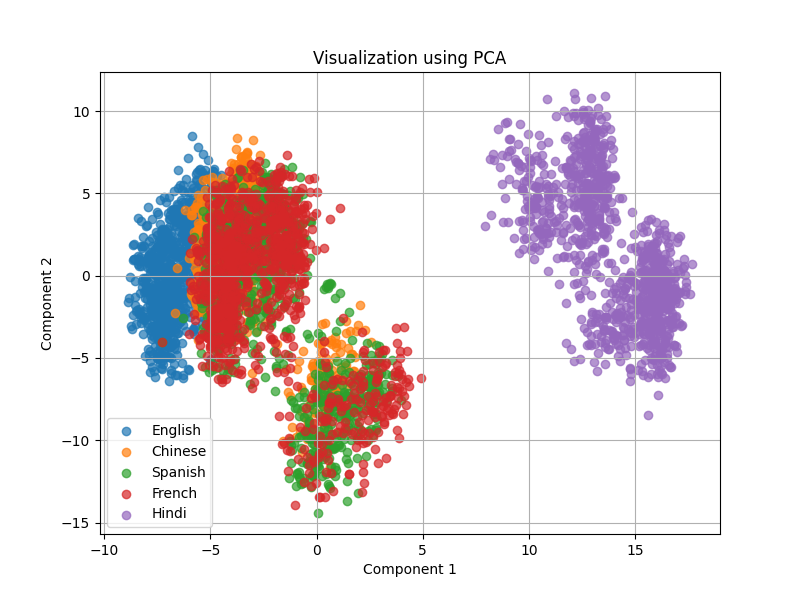}      \caption{PCA, Layer 9}        \end{subfigure}  \hfill  \begin{subfigure}{0.18\textwidth}      \includegraphics[width=\textwidth]{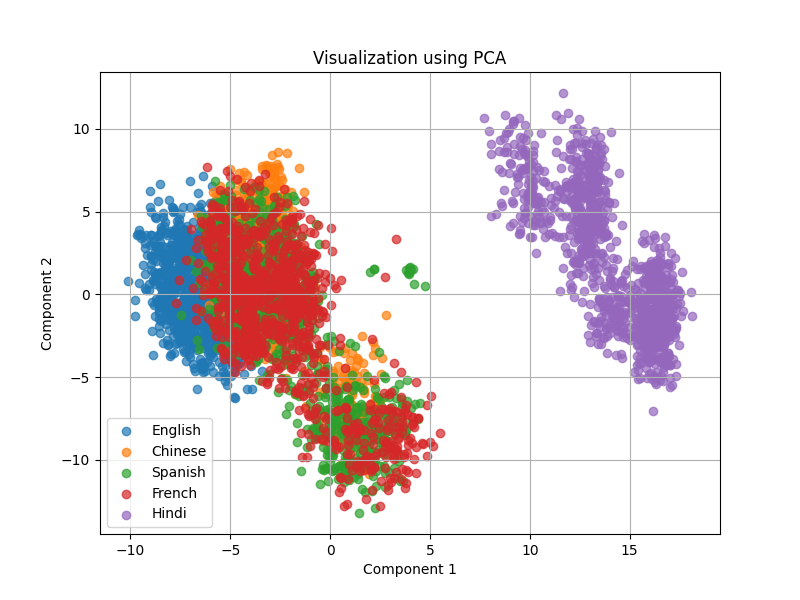}      \caption{PCA, Layer 10}        \end{subfigure}    \vspace{0.2in}    %
\begin{subfigure}{0.18\textwidth}      \includegraphics[width=\textwidth]{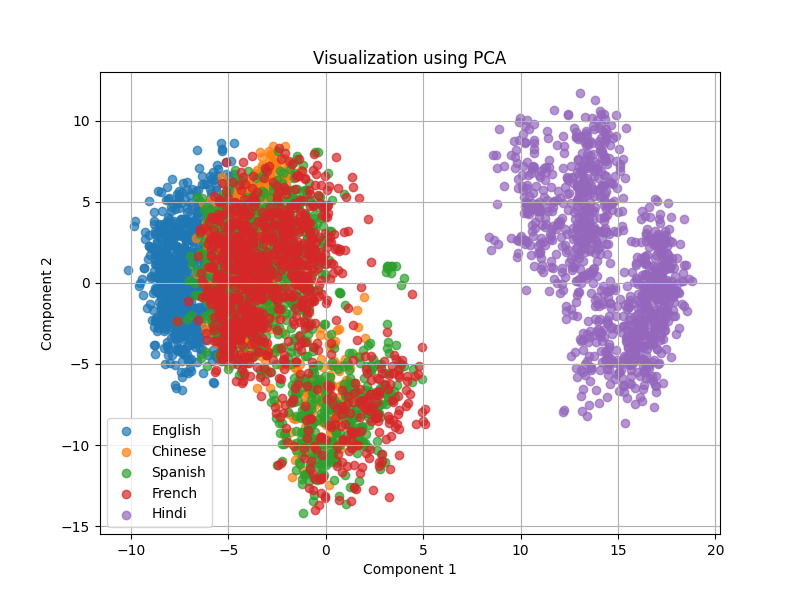}      \caption{PCA, Layer 11}        \end{subfigure}  \hfill  \begin{subfigure}{0.18\textwidth}      \includegraphics[width=\textwidth]{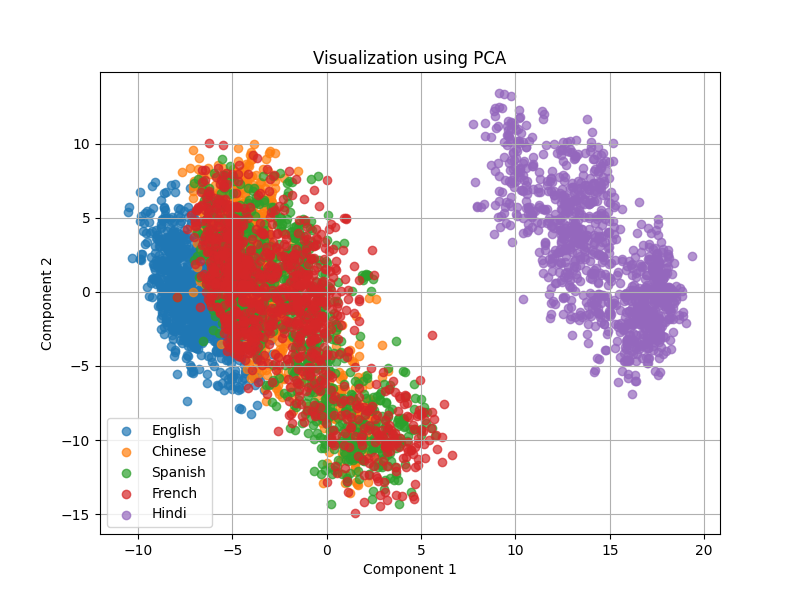}      \caption{PCA, Layer 12}        \end{subfigure}  \hfill  \begin{subfigure}{0.18\textwidth}      \includegraphics[width=\textwidth]{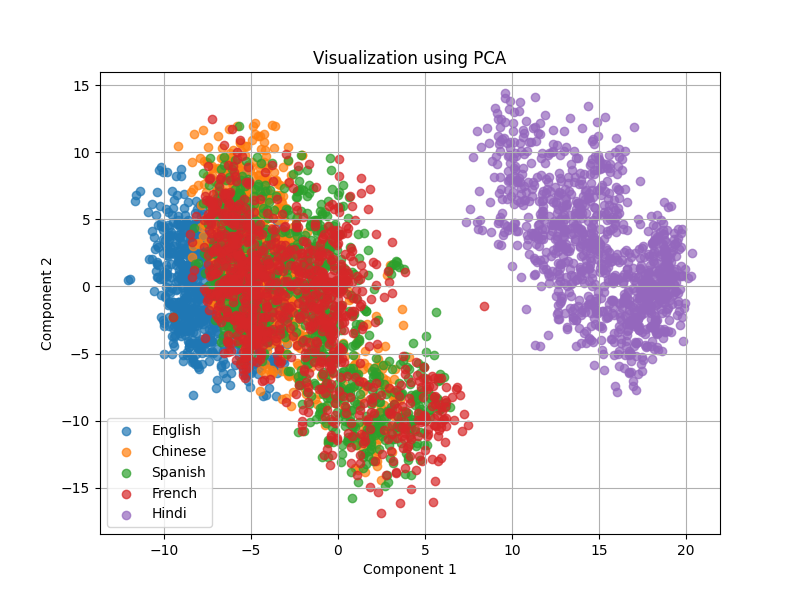}      \caption{PCA, Layer 13}        \end{subfigure}  \hfill  \begin{subfigure}{0.18\textwidth}      \includegraphics[width=\textwidth]{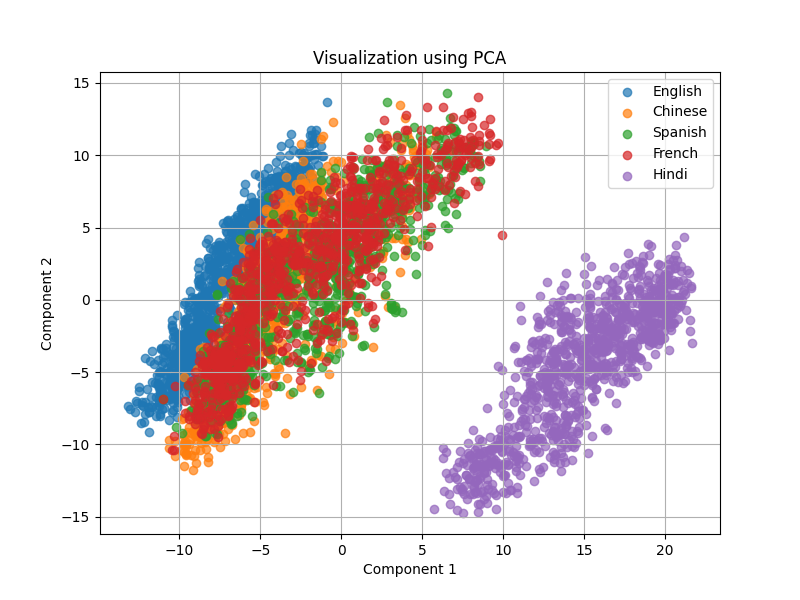}      \caption{PCA, Layer 14}        \end{subfigure}  \hfill  \begin{subfigure}{0.18\textwidth}      \includegraphics[width=\textwidth]{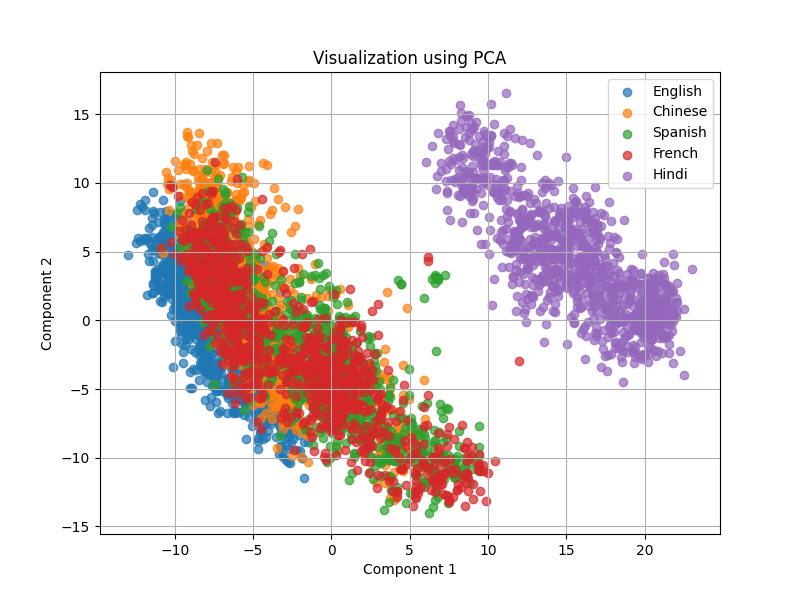}      \caption{PCA, Layer 15}        \end{subfigure}    \vspace{0.2in}    %
\begin{subfigure}{0.18\textwidth}      \includegraphics[width=\textwidth]{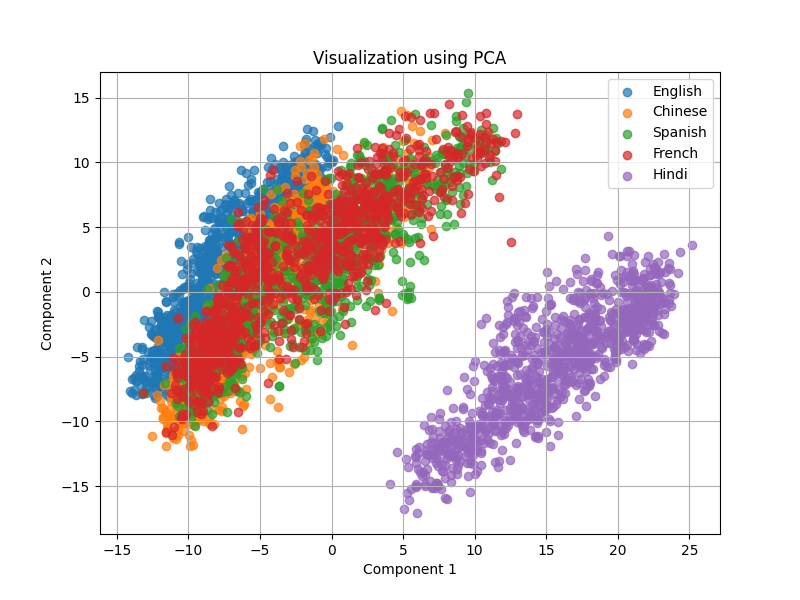}      \caption{PCA, Layer 16}        \end{subfigure}  \hfill  \begin{subfigure}{0.18\textwidth}      \includegraphics[width=\textwidth]{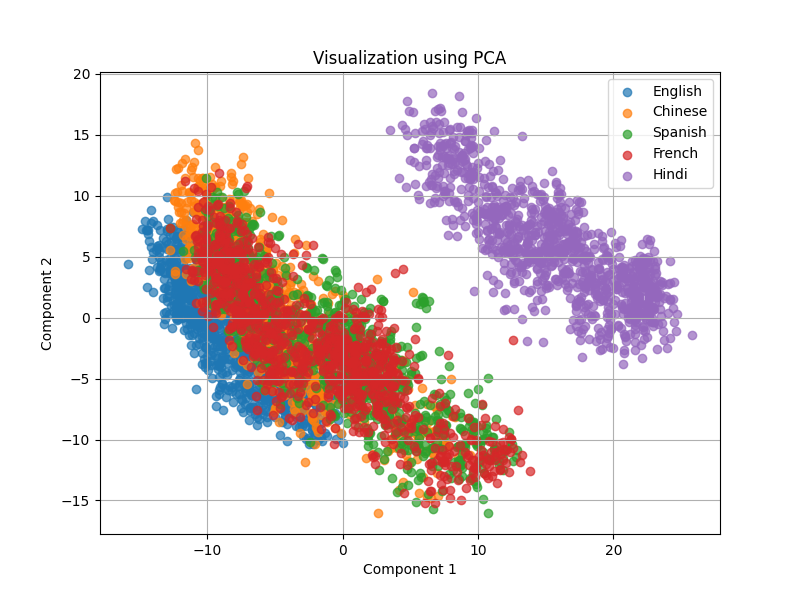}      \caption{PCA, Layer 17}        \end{subfigure}  \hfill  \begin{subfigure}{0.18\textwidth}      \includegraphics[width=\textwidth]{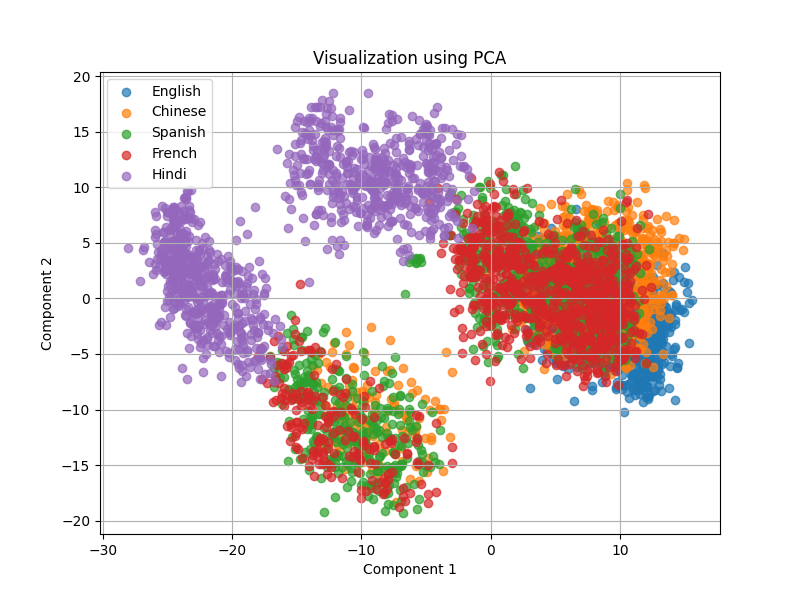}      \caption{PCA, Layer 18}        \end{subfigure}  \hfill  \begin{subfigure}{0.18\textwidth}      \includegraphics[width=\textwidth]{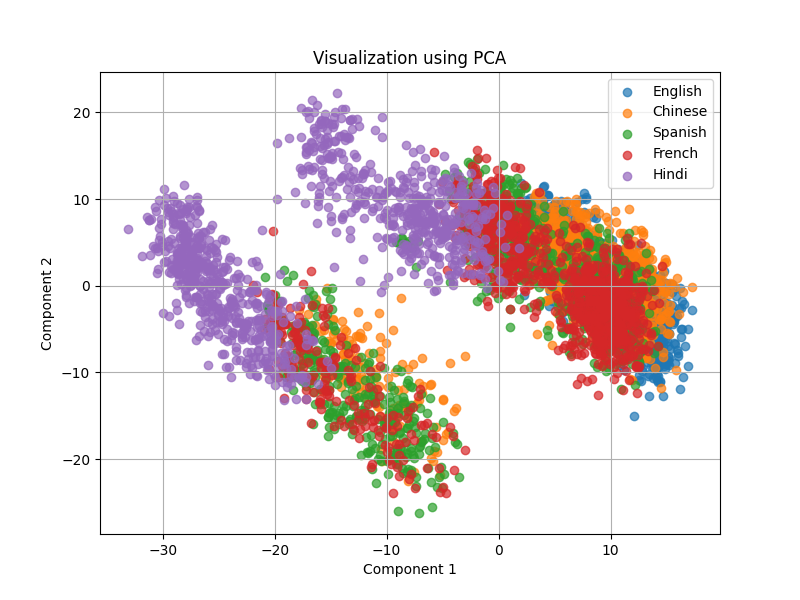}      \caption{PCA, Layer 19}        \end{subfigure}  \hfill  \begin{subfigure}{0.18\textwidth}      \includegraphics[width=\textwidth]{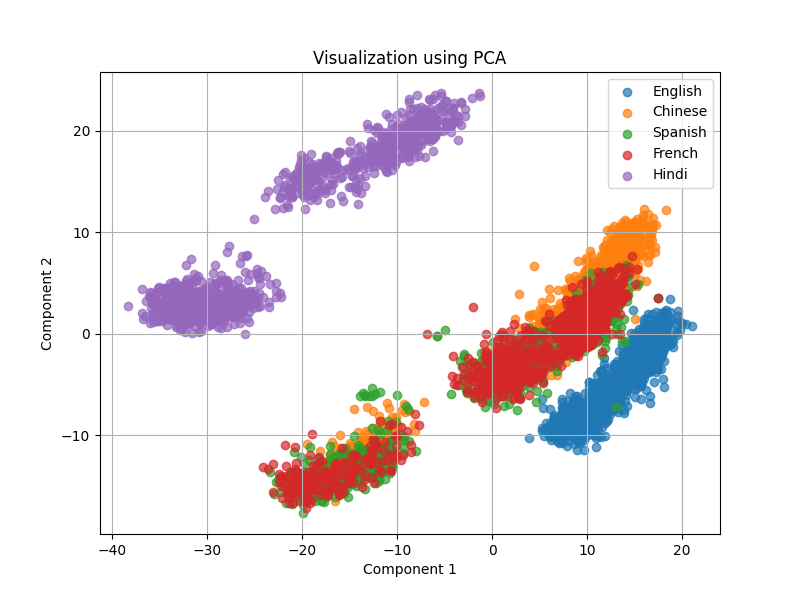}      \caption{PCA, Layer 20}        \end{subfigure}    \vspace{0.2in}    %
\begin{subfigure}{0.18\textwidth}      \includegraphics[width=\textwidth]{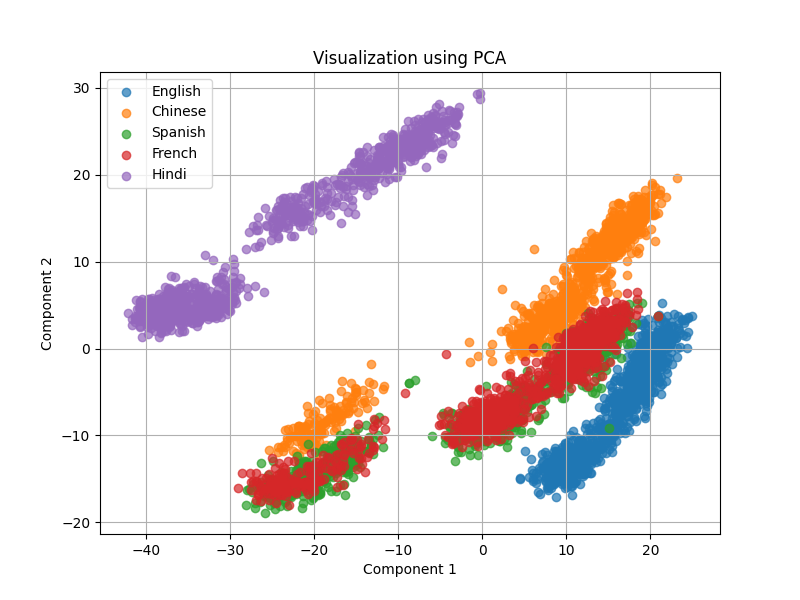}      \caption{PCA, Layer 21}        \end{subfigure}  \hfill  \begin{subfigure}{0.18\textwidth}      \includegraphics[width=\textwidth]{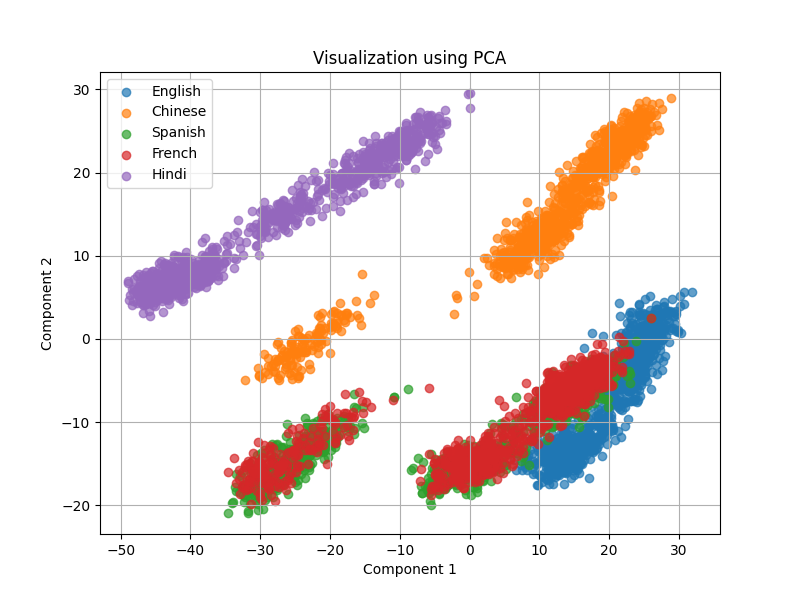}      \caption{PCA, Layer 22}        \end{subfigure}  \hfill  \begin{subfigure}{0.18\textwidth}      \includegraphics[width=\textwidth]{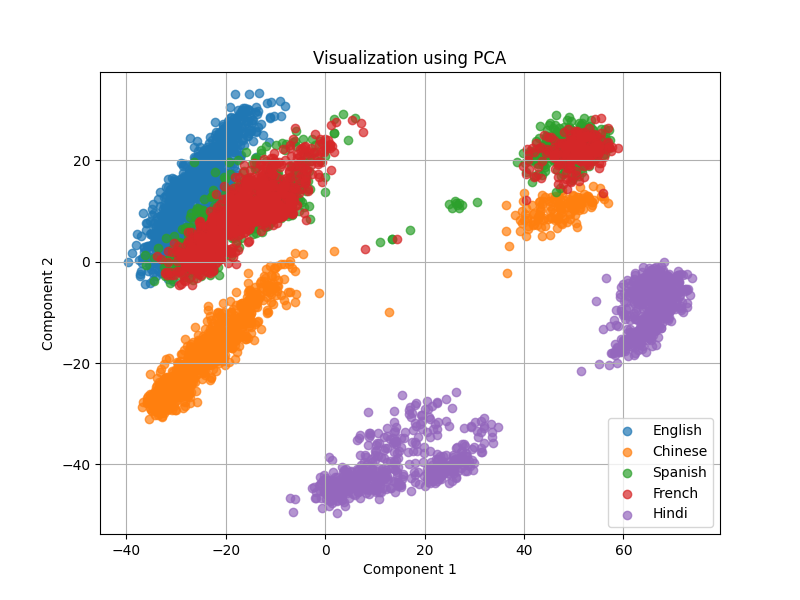}      \caption{PCA, Layer 23}        \end{subfigure}  \hfill  \begin{subfigure}{0.18\textwidth}      \includegraphics[width=\textwidth]{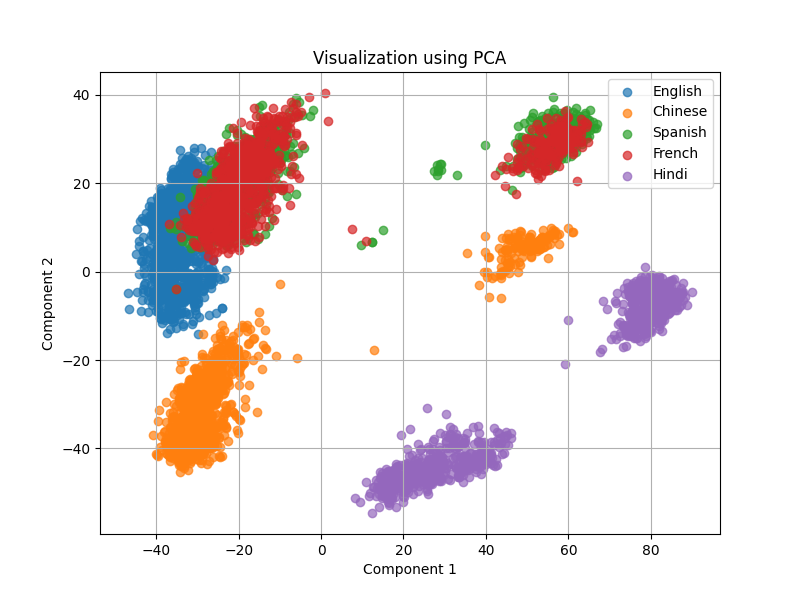}      \caption{PCA, Layer 24}        \end{subfigure}  \hfill  \begin{subfigure}{0.18\textwidth}      \includegraphics[width=\textwidth]{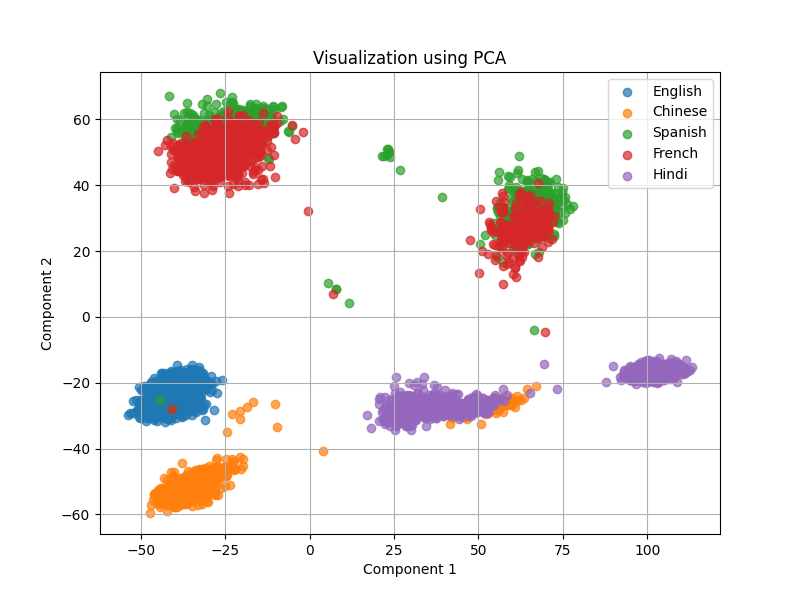}      \caption{PCA, Layer 25}        \end{subfigure}    \vspace{0.2in}    %
\begin{subfigure}{0.18\textwidth}      \includegraphics[width=\textwidth]{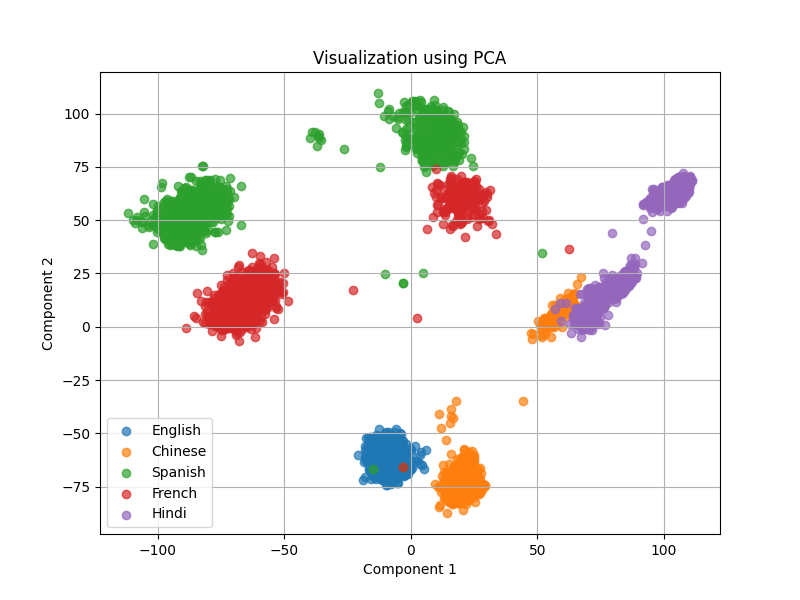}      \caption{PCA, Layer 26}        \end{subfigure}  \hfill  \begin{subfigure}{0.18\textwidth}      \includegraphics[width=\textwidth]{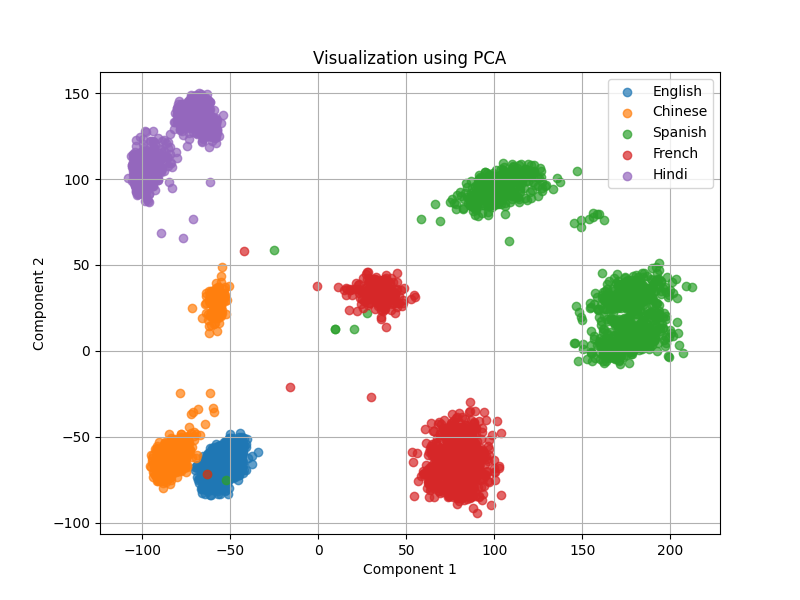}      \caption{PCA, Layer 27}        \end{subfigure}  \hfill  \begin{subfigure}{0.18\textwidth}      \includegraphics[width=\textwidth]{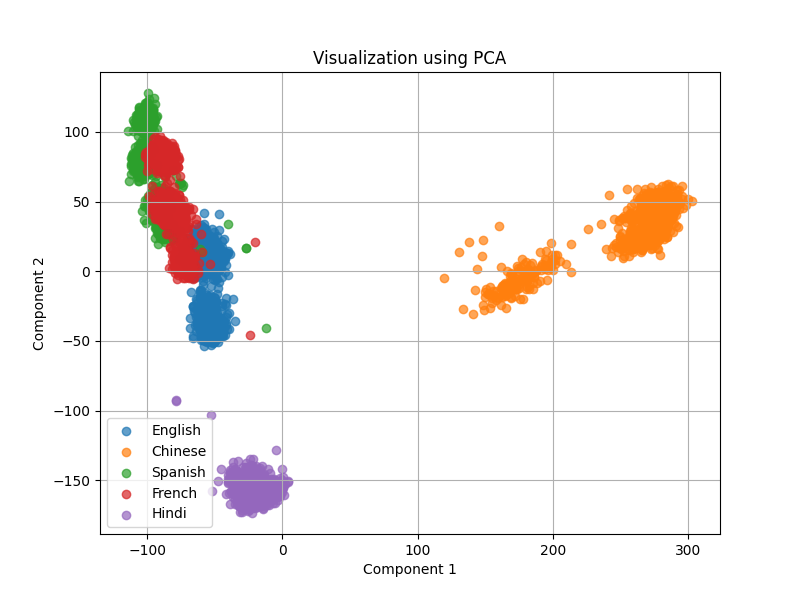}      \caption{PCA, Layer 28}        \end{subfigure}      \caption{PCA visualizations for layers 1-28 of Qwen2-7B-Instruct on the ProofWriter dataset.}  
\end{figure*}

\begin{figure*}[htbp]
\centering
\begin{subfigure}{0.18\textwidth}
\includegraphics[width=\textwidth]{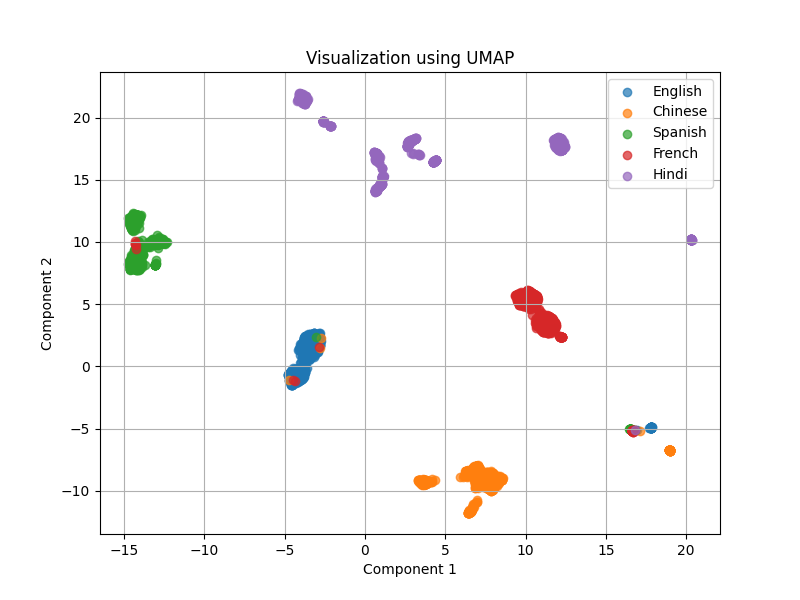}
\caption{UMAP, Layer 1}
\end{subfigure}
\hfill
\begin{subfigure}{0.18\textwidth}
\includegraphics[width=\textwidth]{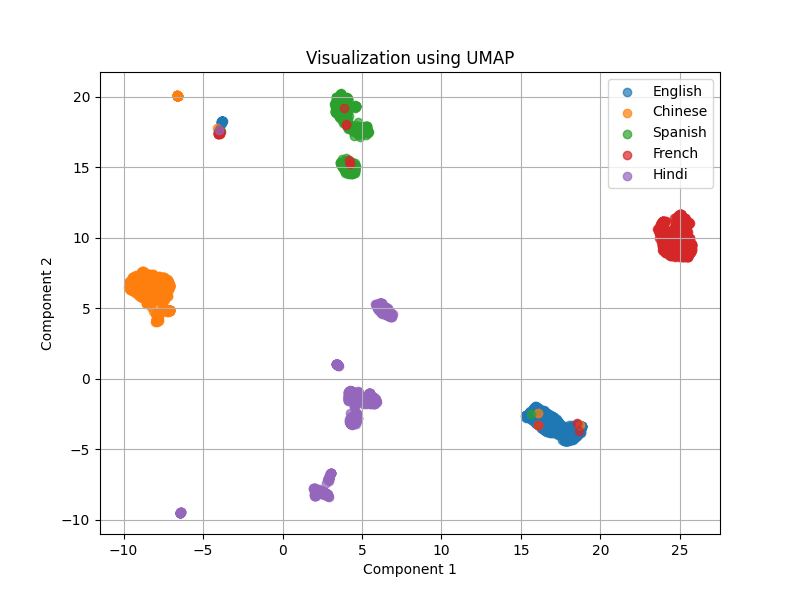}
\caption{UMAP, Layer 2}

\end{subfigure}
\hfill
\begin{subfigure}{0.18\textwidth}
\includegraphics[width=\textwidth]{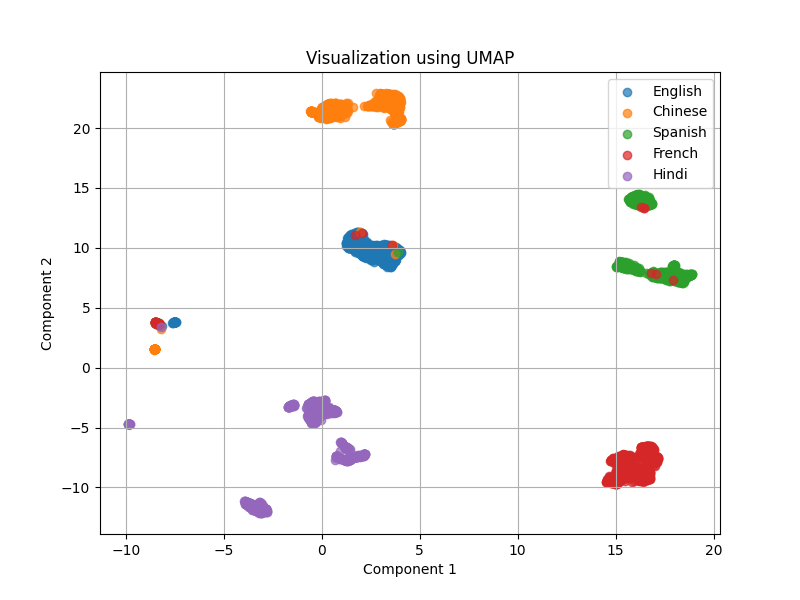}
\caption{UMAP, Layer 3}

\end{subfigure}
\hfill
\begin{subfigure}{0.18\textwidth}
\includegraphics[width=\textwidth]{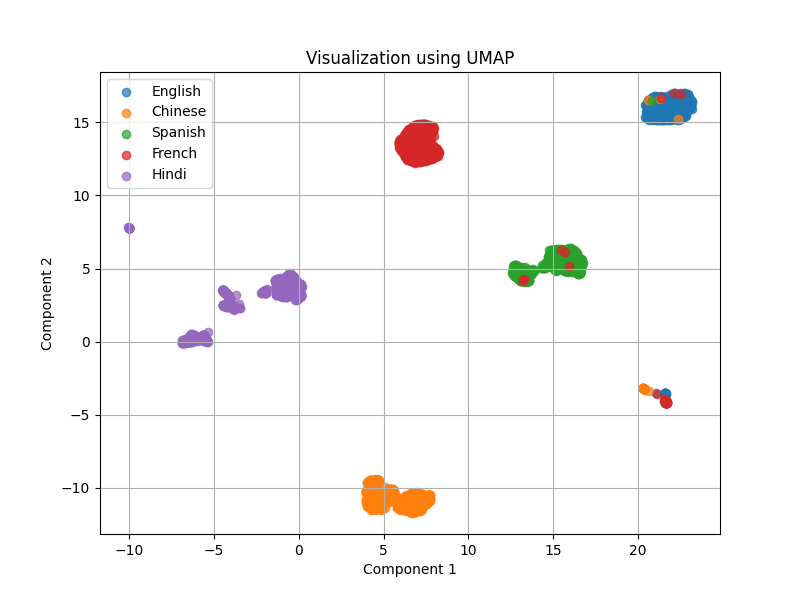}
\caption{UMAP, Layer 4}

\end{subfigure}
\hfill
\begin{subfigure}{0.18\textwidth}
\includegraphics[width=\textwidth]{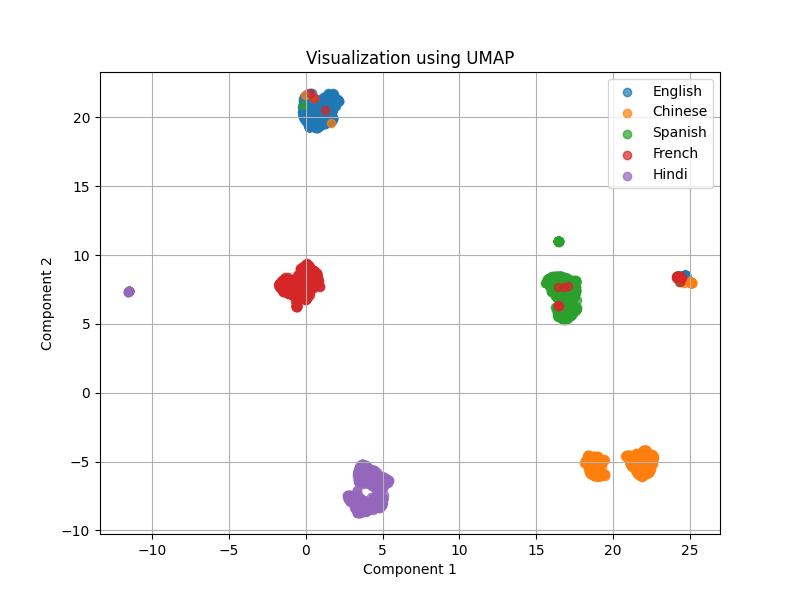}
\caption{UMAP, Layer 5}

\end{subfigure}
\vspace{0.2in} %
\begin{subfigure}{0.18\textwidth}      \includegraphics[width=\textwidth]{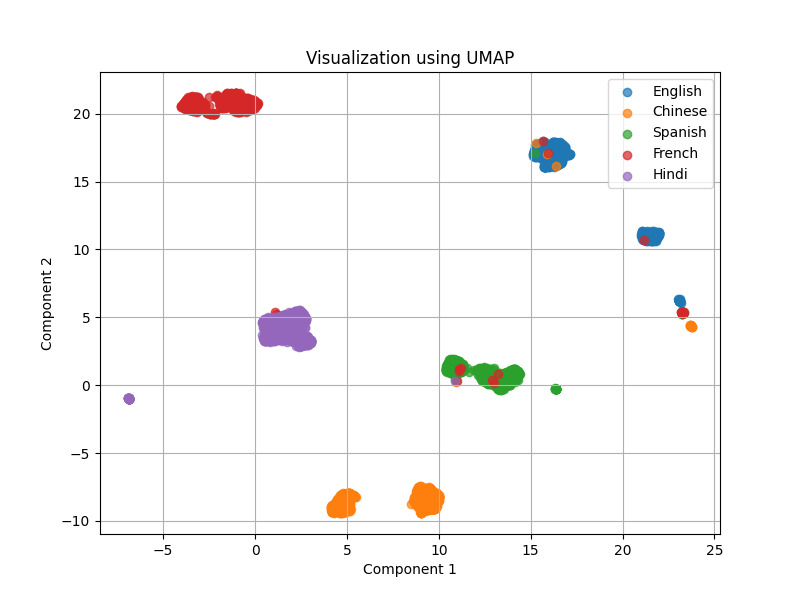}      \caption{UMAP, Layer 6}        \end{subfigure}  \hfill  \begin{subfigure}{0.18\textwidth}      \includegraphics[width=\textwidth]{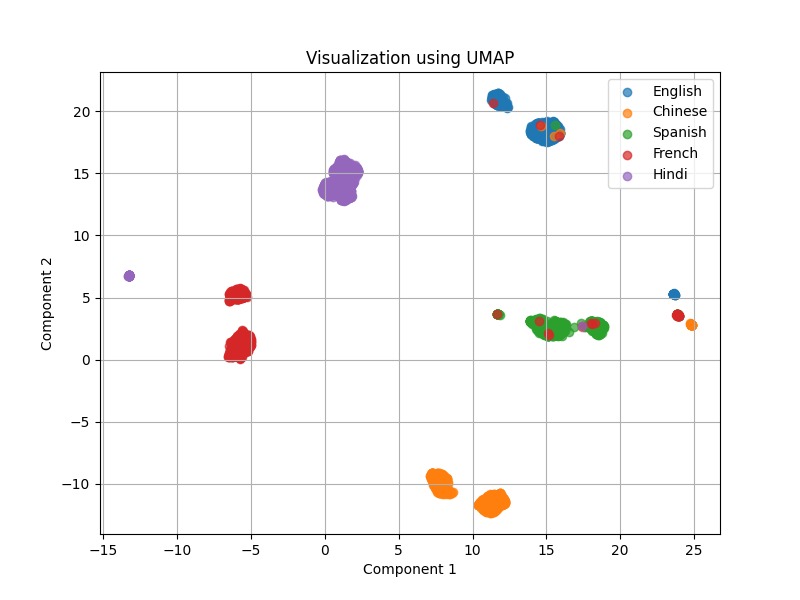}      \caption{UMAP, Layer 7}        \end{subfigure}  \hfill  \begin{subfigure}{0.18\textwidth}      \includegraphics[width=\textwidth]{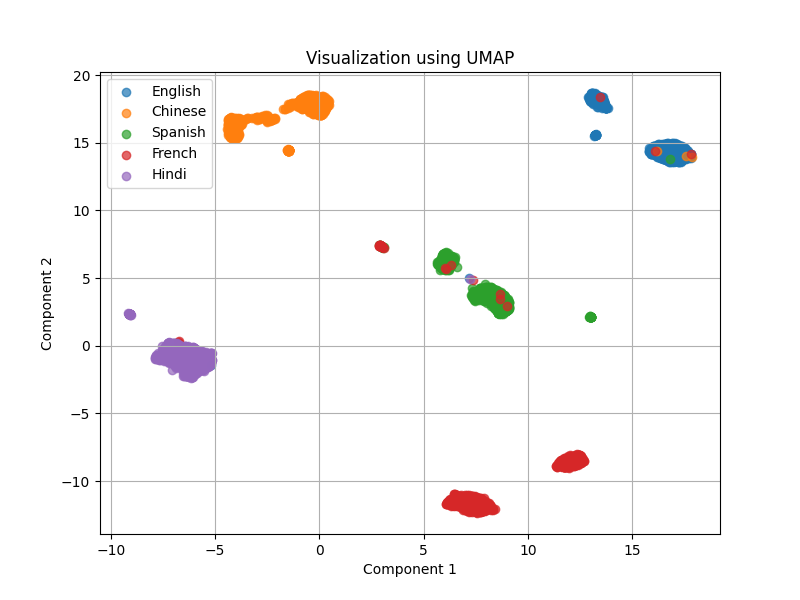}      \caption{UMAP, Layer 8}        \end{subfigure}  \hfill  \begin{subfigure}{0.18\textwidth}      \includegraphics[width=\textwidth]{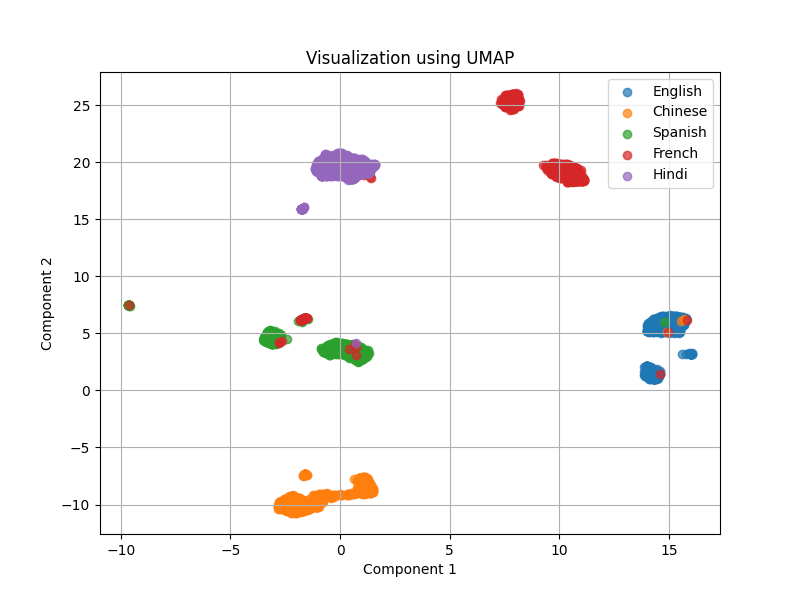}      \caption{UMAP, Layer 9}        \end{subfigure}  \hfill  \begin{subfigure}{0.18\textwidth}      \includegraphics[width=\textwidth]{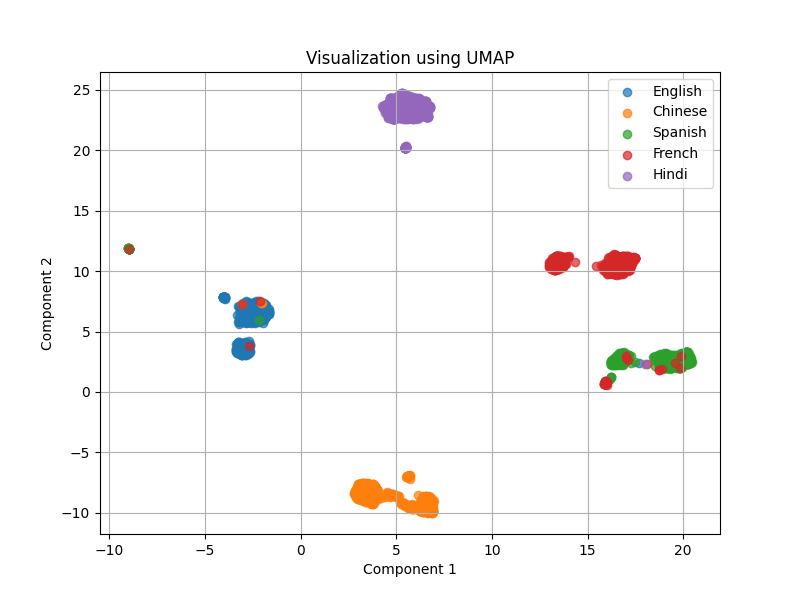}      \caption{UMAP, Layer 10}        \end{subfigure}    \vspace{0.2in}    %
\begin{subfigure}{0.18\textwidth}      \includegraphics[width=\textwidth]{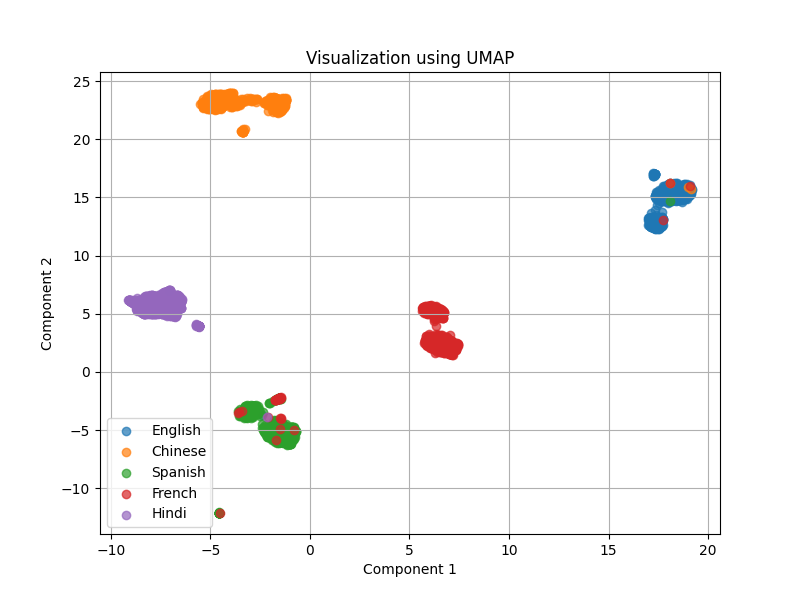}      \caption{UMAP, Layer 11}        \end{subfigure}  \hfill  \begin{subfigure}{0.18\textwidth}      \includegraphics[width=\textwidth]{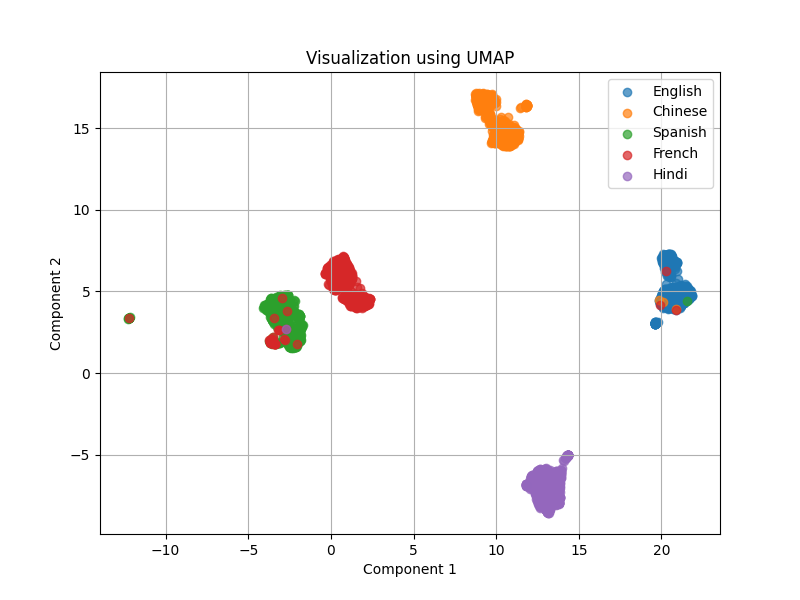}      \caption{UMAP, Layer 12}        \end{subfigure}  \hfill  \begin{subfigure}{0.18\textwidth}      \includegraphics[width=\textwidth]{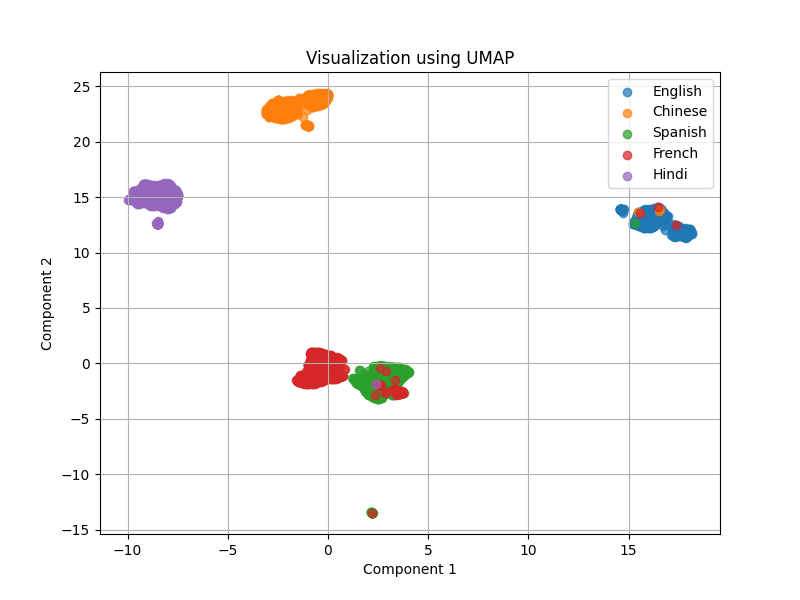}      \caption{UMAP, Layer 13}        \end{subfigure}  \hfill  \begin{subfigure}{0.18\textwidth}      \includegraphics[width=\textwidth]{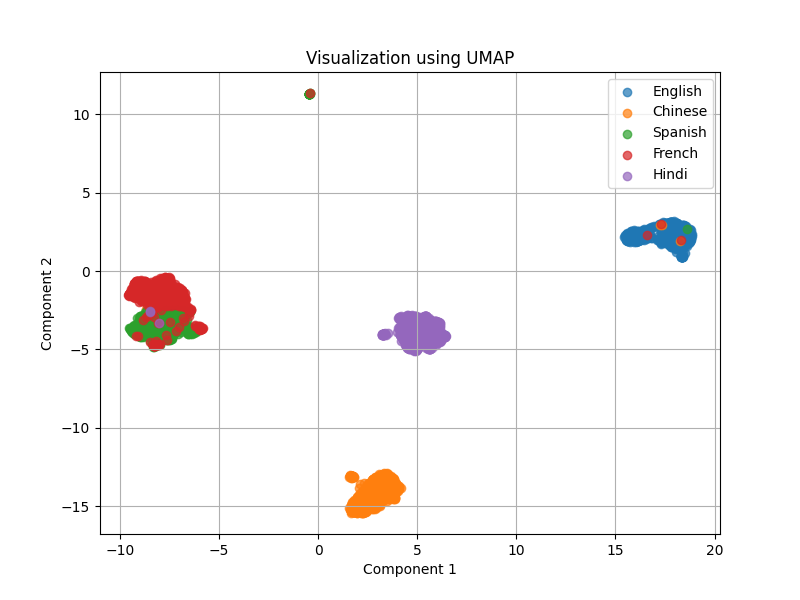}      \caption{UMAP, Layer 14}        \end{subfigure}  \hfill  \begin{subfigure}{0.18\textwidth}      \includegraphics[width=\textwidth]{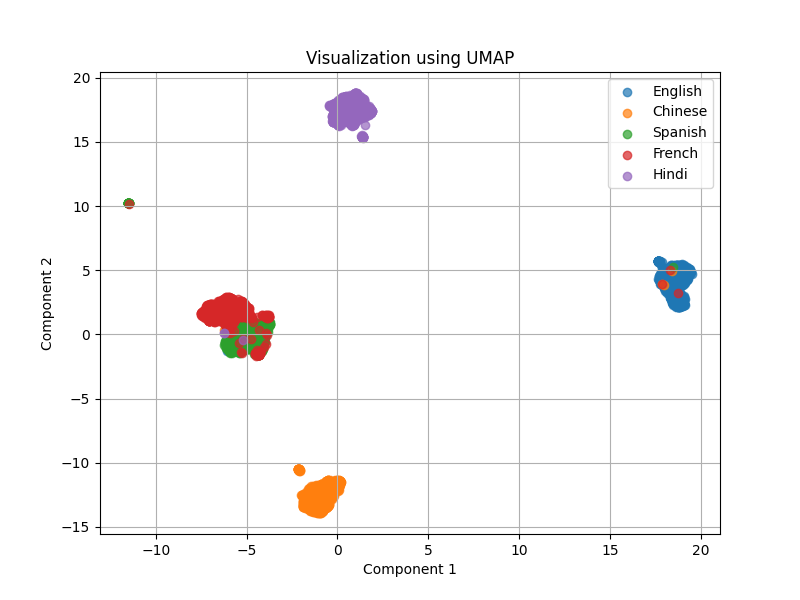}      \caption{UMAP, Layer 15}        \end{subfigure}    \vspace{0.2in}    %
\begin{subfigure}{0.18\textwidth}      \includegraphics[width=\textwidth]{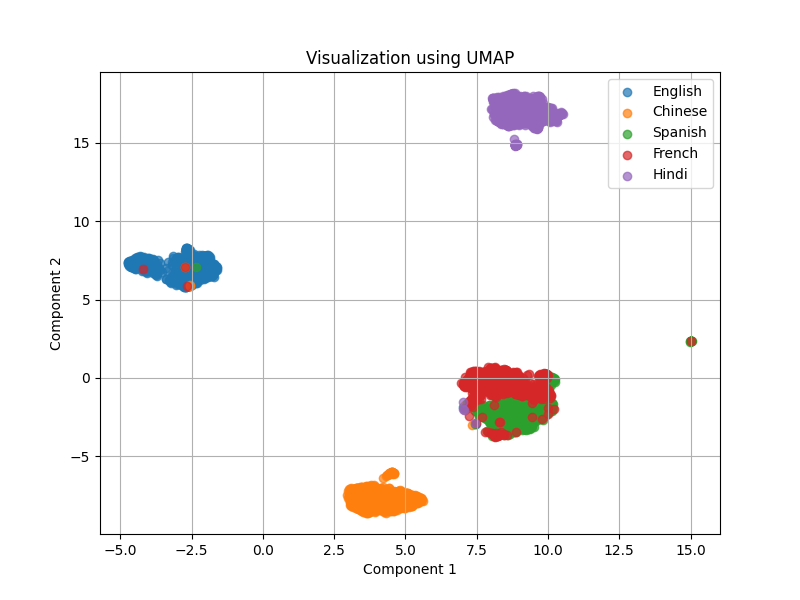}      \caption{UMAP, Layer 16}        \end{subfigure}  \hfill  \begin{subfigure}{0.18\textwidth}      \includegraphics[width=\textwidth]{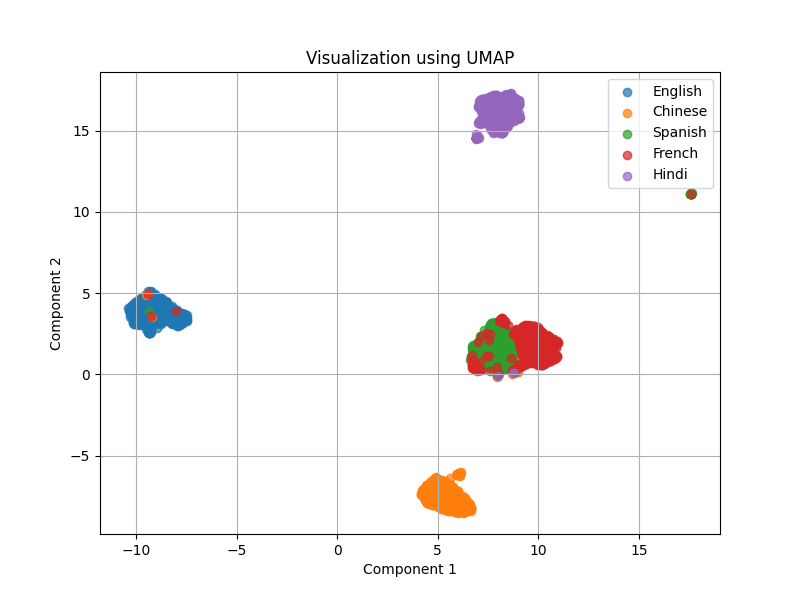}      \caption{UMAP, Layer 17}        \end{subfigure}  \hfill  \begin{subfigure}{0.18\textwidth}      \includegraphics[width=\textwidth]{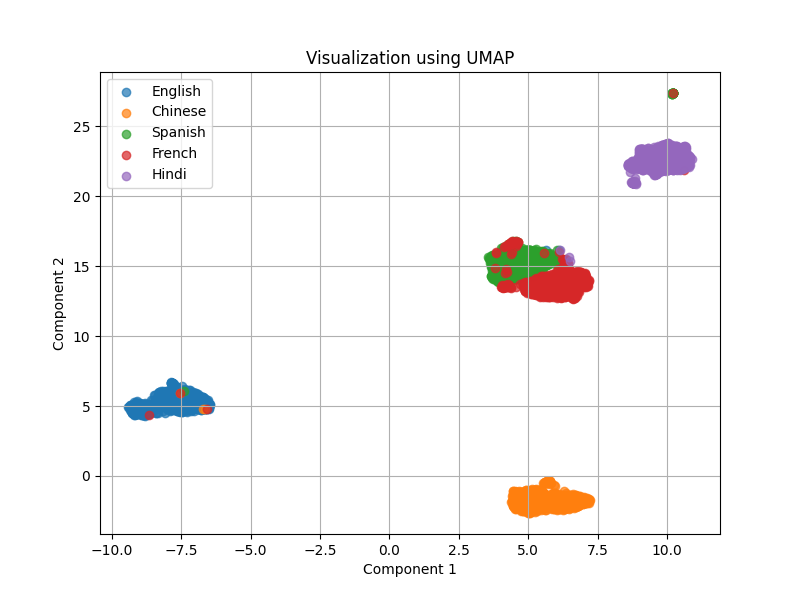}      \caption{UMAP, Layer 18}        \end{subfigure}  \hfill  \begin{subfigure}{0.18\textwidth}      \includegraphics[width=\textwidth]{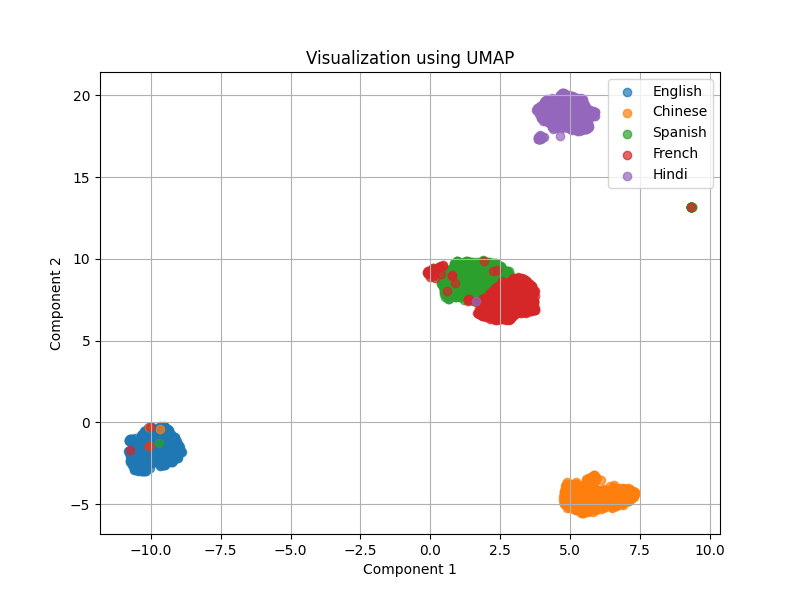}      \caption{UMAP, Layer 19}        \end{subfigure}  \hfill  \begin{subfigure}{0.18\textwidth}      \includegraphics[width=\textwidth]{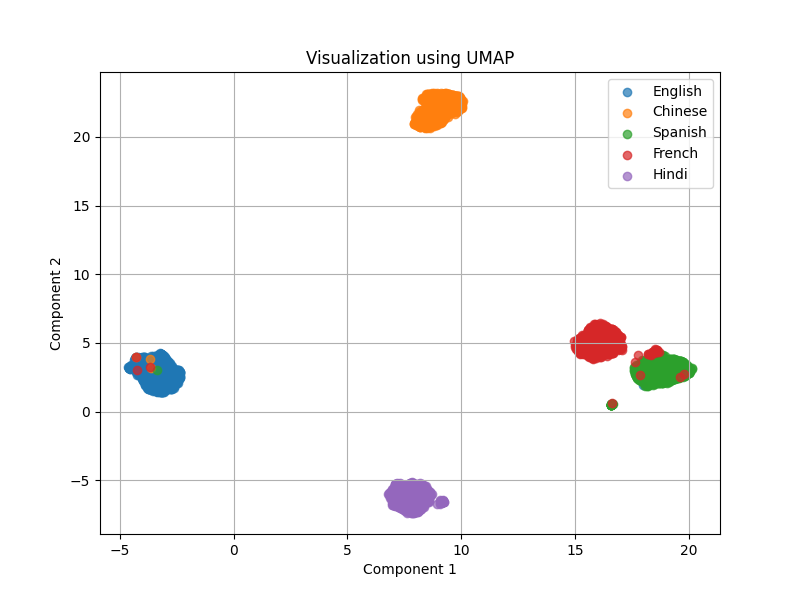}      \caption{UMAP, Layer 20}        \end{subfigure}    \vspace{0.2in}    %
\begin{subfigure}{0.18\textwidth}      \includegraphics[width=\textwidth]{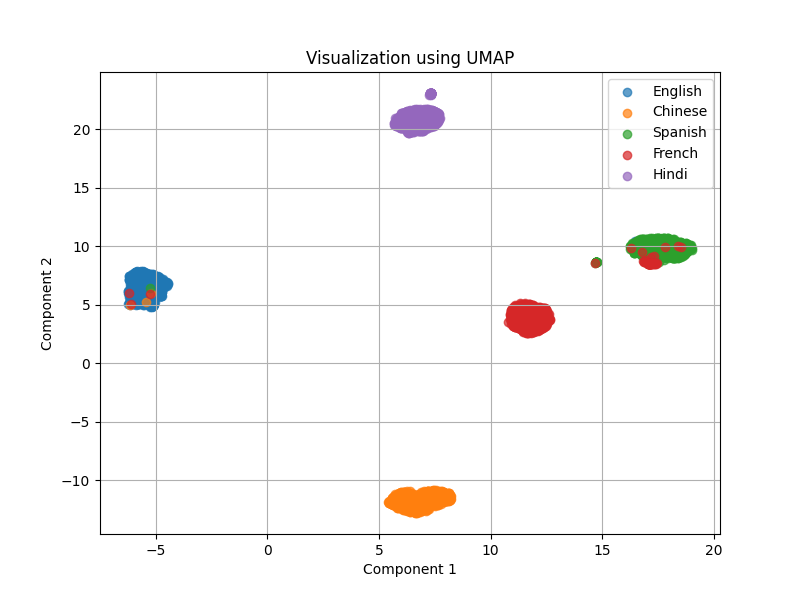}      \caption{UMAP, Layer 21}        \end{subfigure}  \hfill  \begin{subfigure}{0.18\textwidth}      \includegraphics[width=\textwidth]{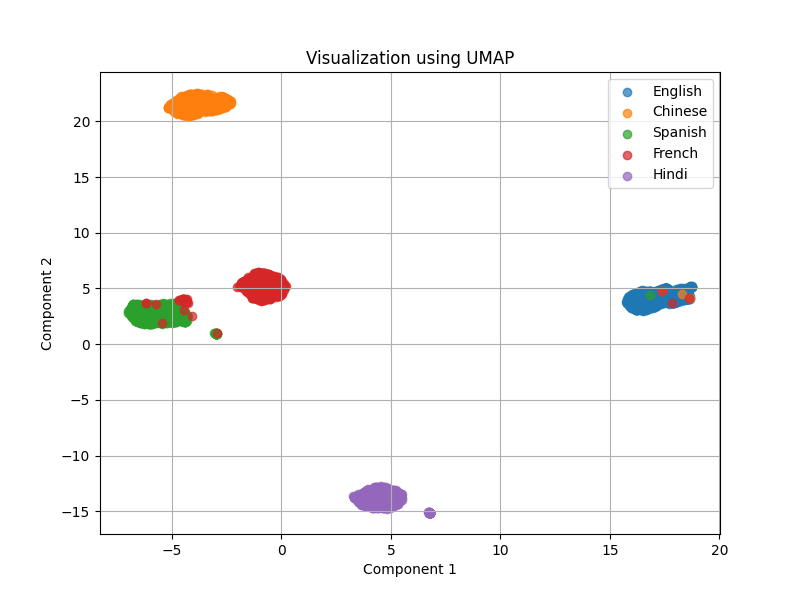}      \caption{UMAP, Layer 22}        \end{subfigure}  \hfill  \begin{subfigure}{0.18\textwidth}      \includegraphics[width=\textwidth]{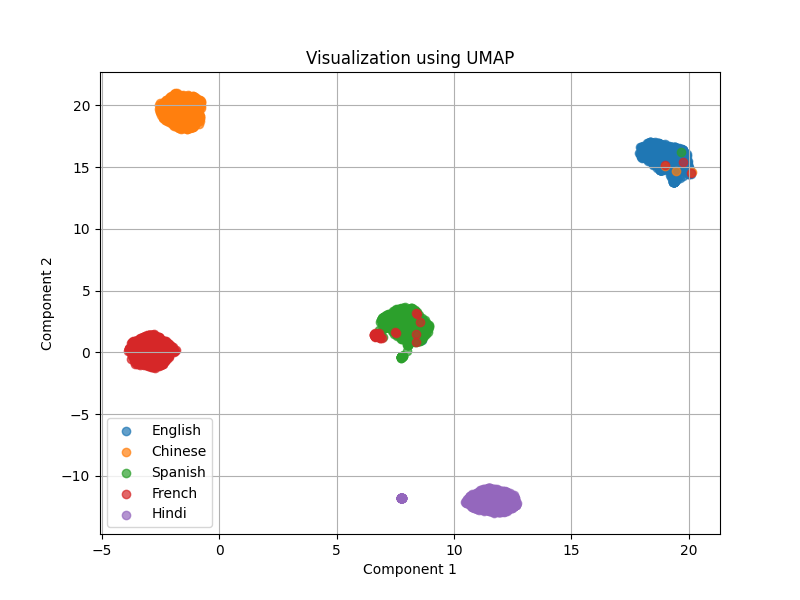}      \caption{UMAP, Layer 23}        \end{subfigure}  \hfill  \begin{subfigure}{0.18\textwidth}      \includegraphics[width=\textwidth]{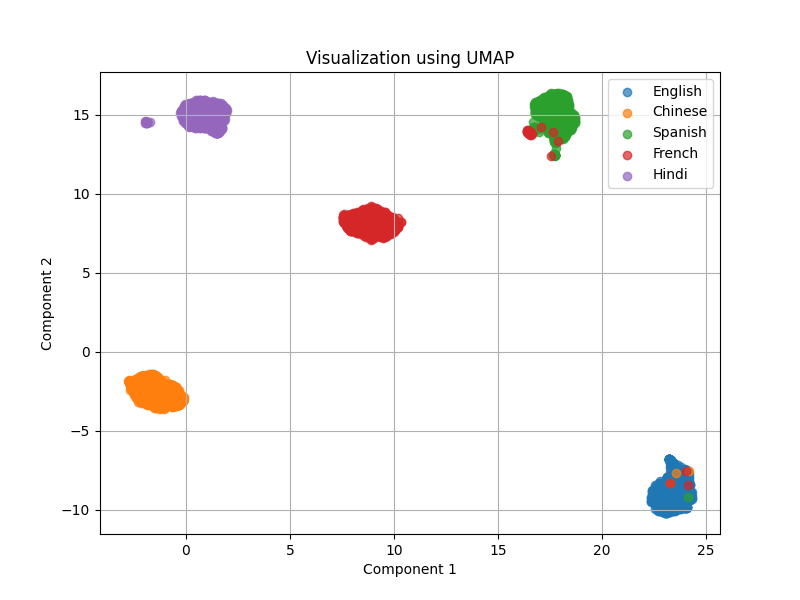}      \caption{UMAP, Layer 24}        \end{subfigure}  \hfill  \begin{subfigure}{0.18\textwidth}      \includegraphics[width=\textwidth]{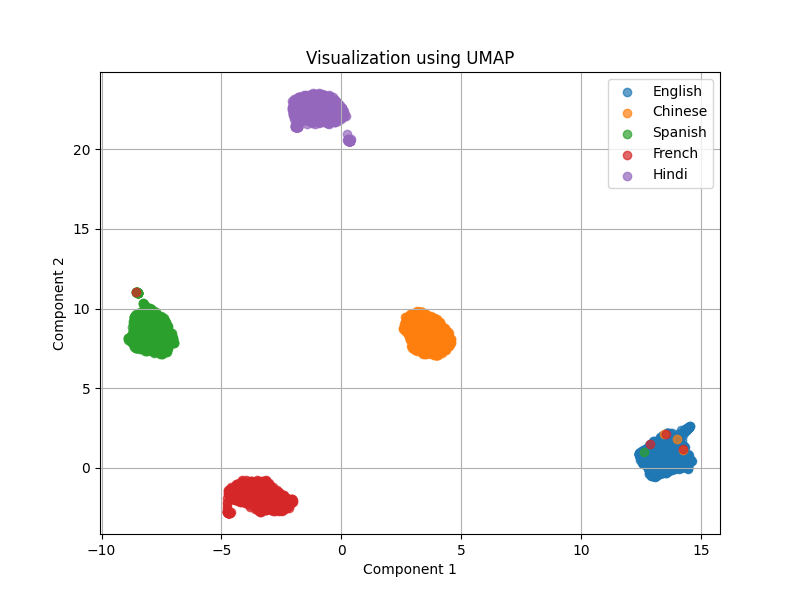}      \caption{UMAP, Layer 25}        \end{subfigure}    \vspace{0.2in}    %
\begin{subfigure}{0.18\textwidth}      \includegraphics[width=\textwidth]{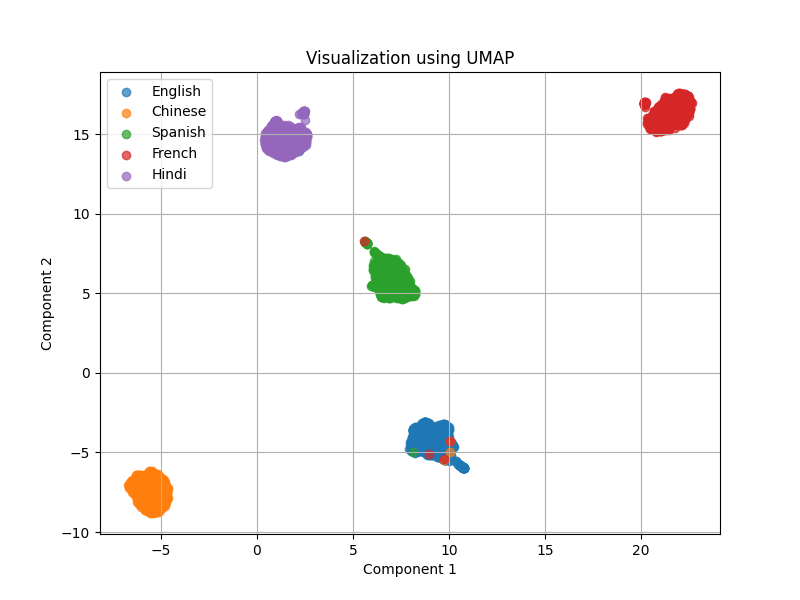}      \caption{UMAP, Layer 26}        \end{subfigure}  \hfill  \begin{subfigure}{0.18\textwidth}      \includegraphics[width=\textwidth]{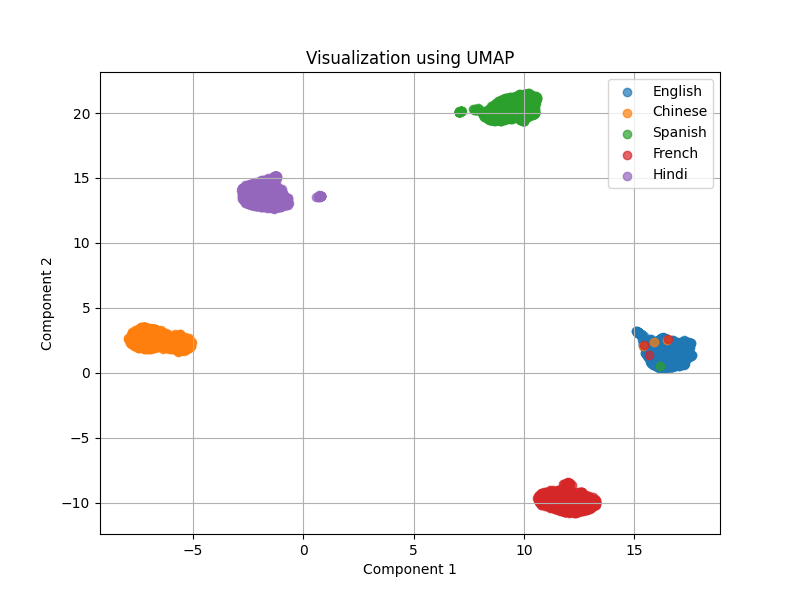}      \caption{UMAP, Layer 27}        \end{subfigure}  \hfill  \begin{subfigure}{0.18\textwidth}      \includegraphics[width=\textwidth]{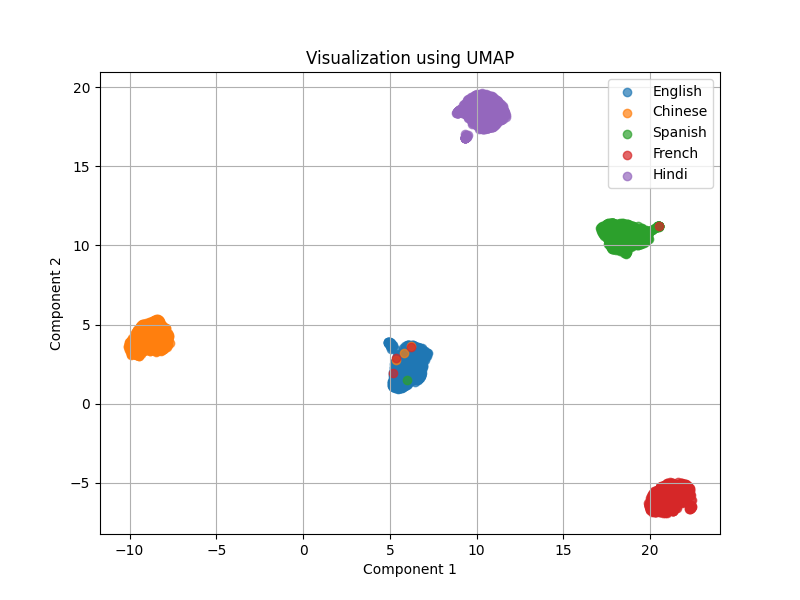}      \caption{UMAP, Layer 28}        \end{subfigure}      \caption{UMAP visualizations for layers 1-28 of Qwen2-7B-Instruct on the GSM8K dataset.}  
\end{figure*}

\begin{figure*}[htbp]
\centering
\begin{subfigure}{0.18\textwidth}
\includegraphics[width=\textwidth]{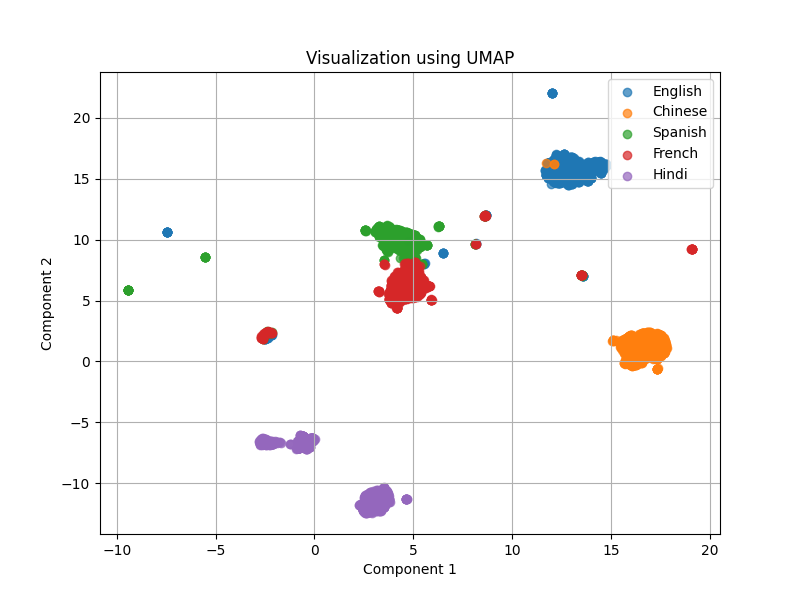}
\caption{UMAP, Layer 1}
\end{subfigure}
\hfill
\begin{subfigure}{0.18\textwidth}
\includegraphics[width=\textwidth]{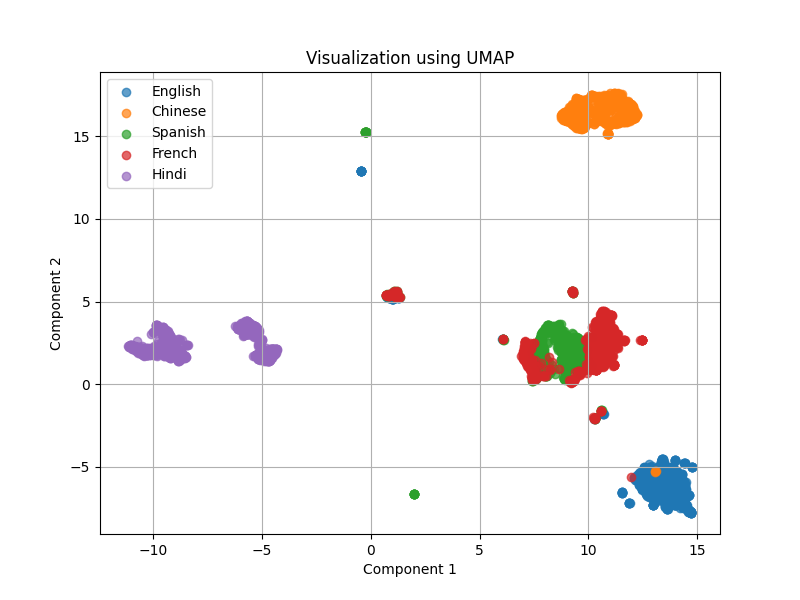}
\caption{UMAP, Layer 2}

\end{subfigure}
\hfill
\begin{subfigure}{0.18\textwidth}
\includegraphics[width=\textwidth]{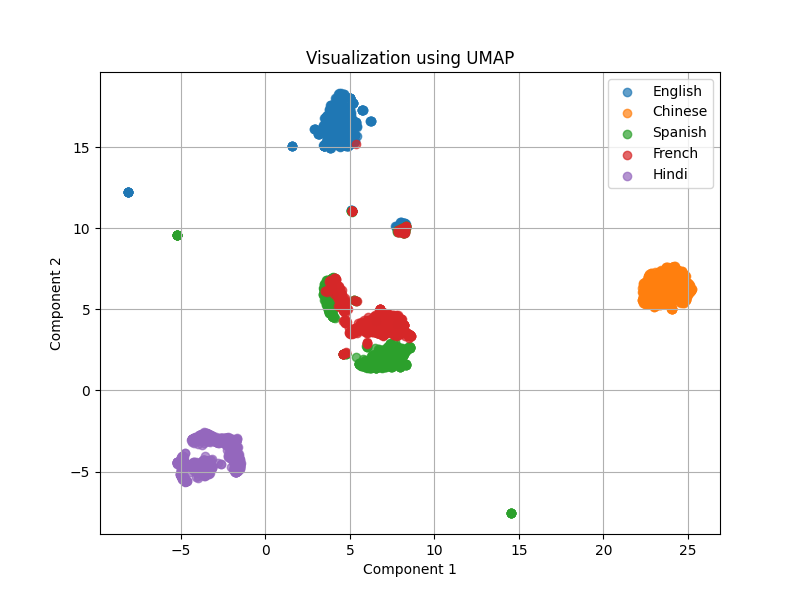}
\caption{UMAP, Layer 3}

\end{subfigure}
\hfill
\begin{subfigure}{0.18\textwidth}
\includegraphics[width=\textwidth]{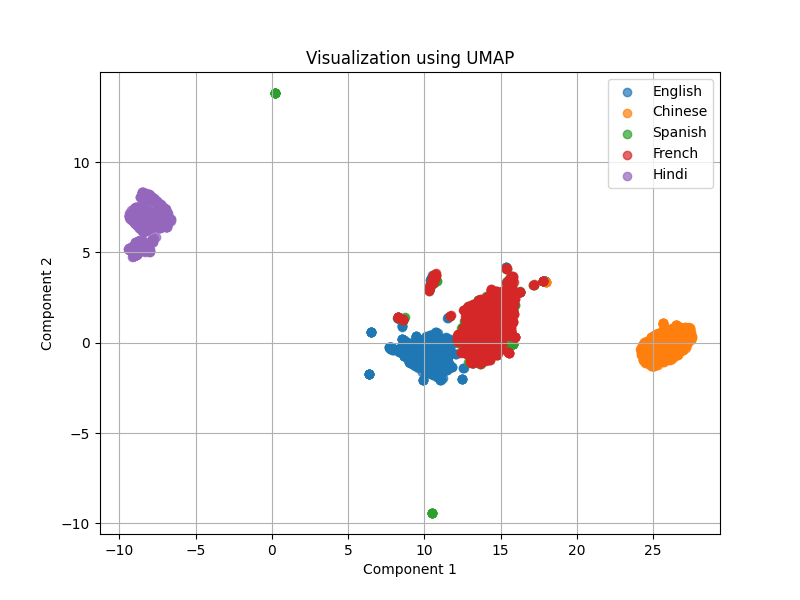}
\caption{UMAP, Layer 4}

\end{subfigure}
\hfill
\begin{subfigure}{0.18\textwidth}
\includegraphics[width=\textwidth]{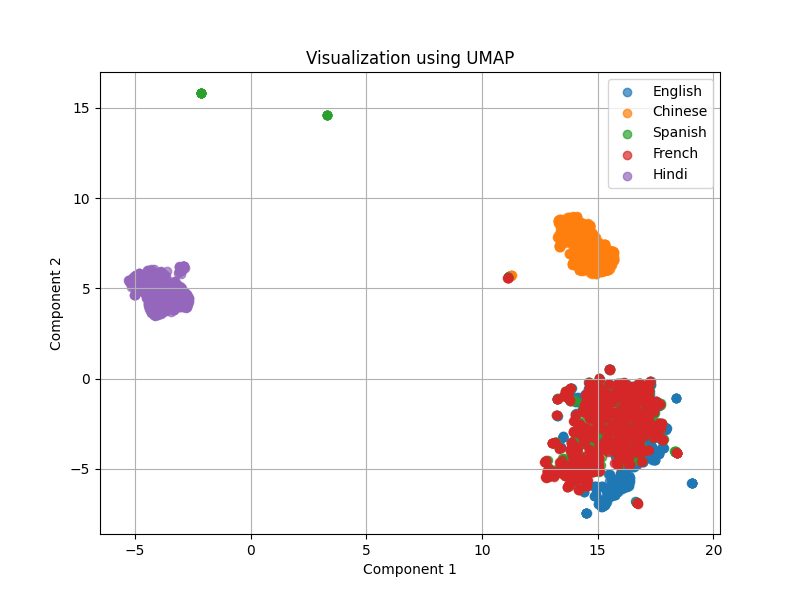}
\caption{UMAP, Layer 5}

\end{subfigure}
\vspace{0.2in} %
\begin{subfigure}{0.18\textwidth}      \includegraphics[width=\textwidth]{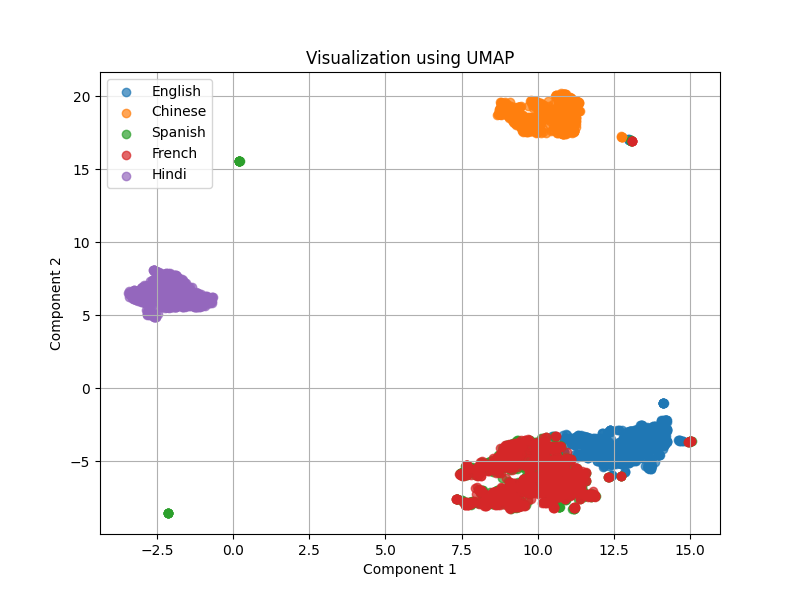}      \caption{UMAP, Layer 6}        \end{subfigure}  \hfill  \begin{subfigure}{0.18\textwidth}      \includegraphics[width=\textwidth]{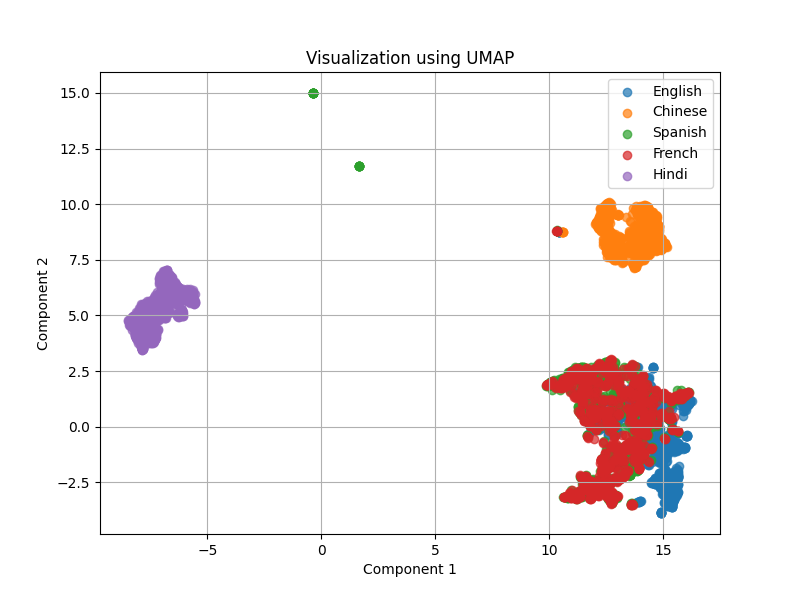}      \caption{UMAP, Layer 7}        \end{subfigure}  \hfill  \begin{subfigure}{0.18\textwidth}      \includegraphics[width=\textwidth]{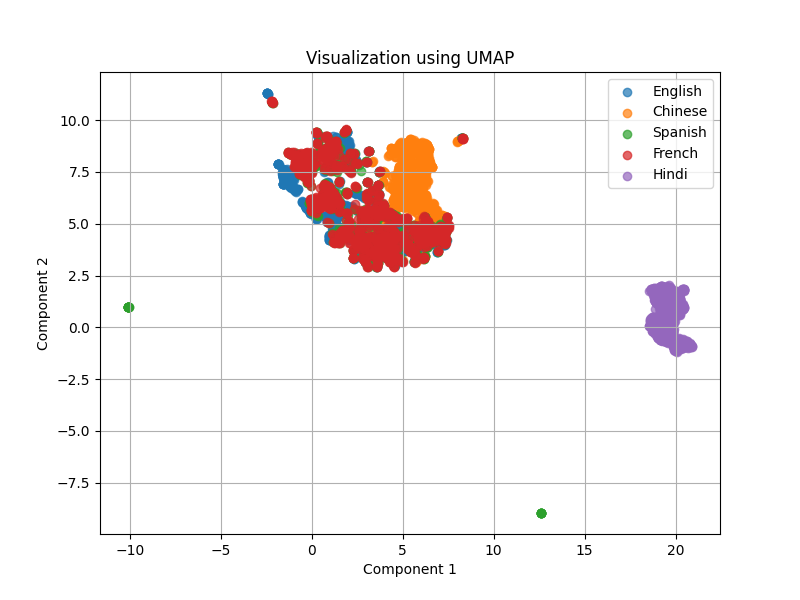}      \caption{UMAP, Layer 8}        \end{subfigure}  \hfill  \begin{subfigure}{0.18\textwidth}      \includegraphics[width=\textwidth]{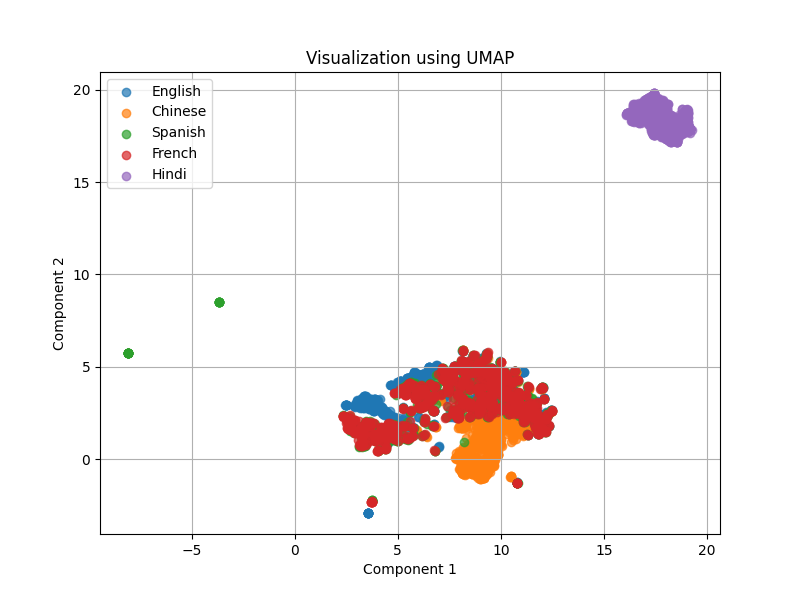}      \caption{UMAP, Layer 9}        \end{subfigure}  \hfill  \begin{subfigure}{0.18\textwidth}      \includegraphics[width=\textwidth]{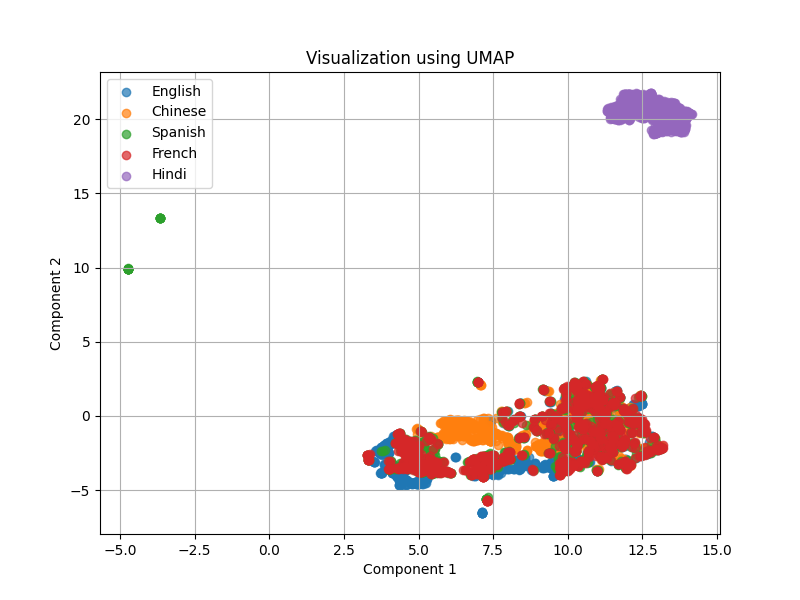}      \caption{UMAP, Layer 10}        \end{subfigure}    \vspace{0.2in}    %
\begin{subfigure}{0.18\textwidth}      \includegraphics[width=\textwidth]{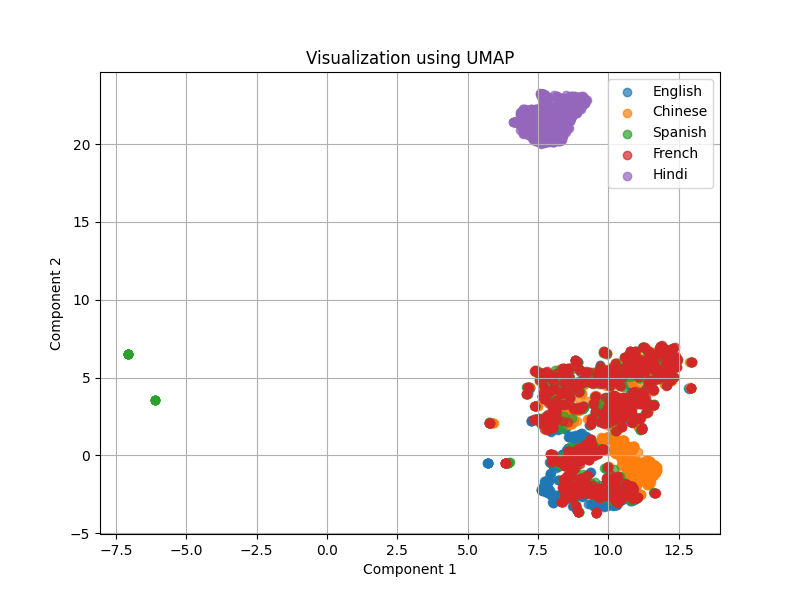}      \caption{UMAP, Layer 11}        \end{subfigure}  \hfill  \begin{subfigure}{0.18\textwidth}      \includegraphics[width=\textwidth]{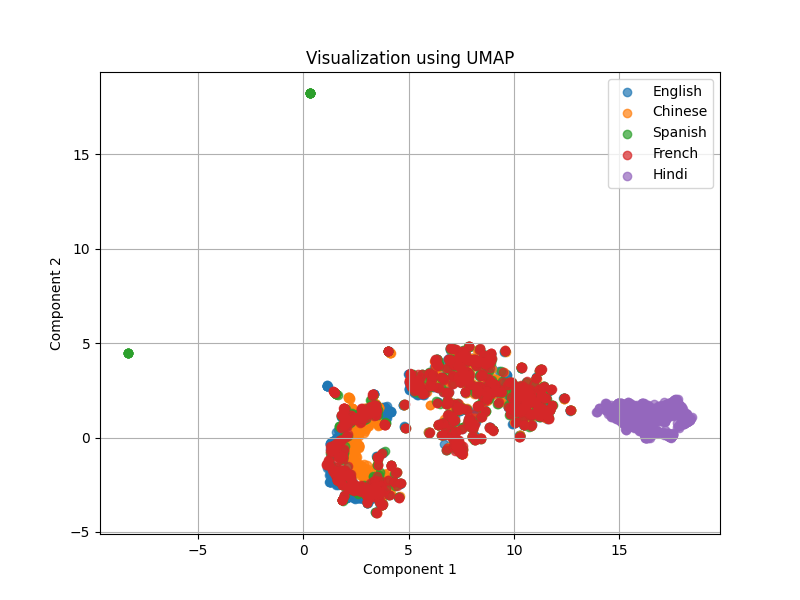}      \caption{UMAP, Layer 12}        \end{subfigure}  \hfill  \begin{subfigure}{0.18\textwidth}      \includegraphics[width=\textwidth]{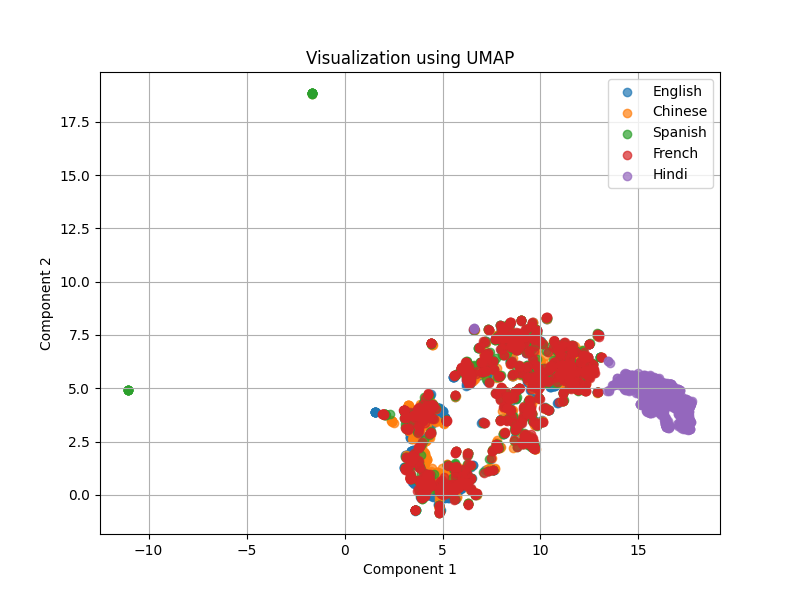}      \caption{UMAP, Layer 13}        \end{subfigure}  \hfill  \begin{subfigure}{0.18\textwidth}      \includegraphics[width=\textwidth]{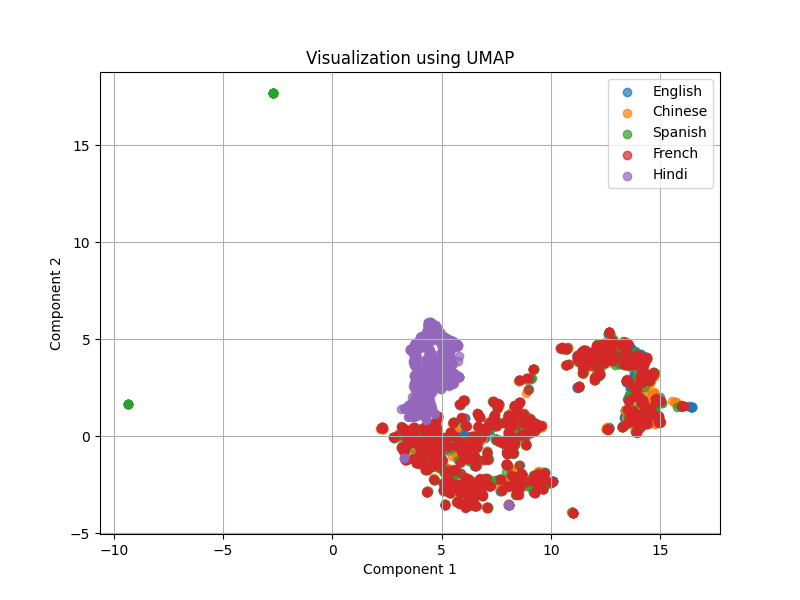}      \caption{UMAP, Layer 14}        \end{subfigure}  \hfill  \begin{subfigure}{0.18\textwidth}      \includegraphics[width=\textwidth]{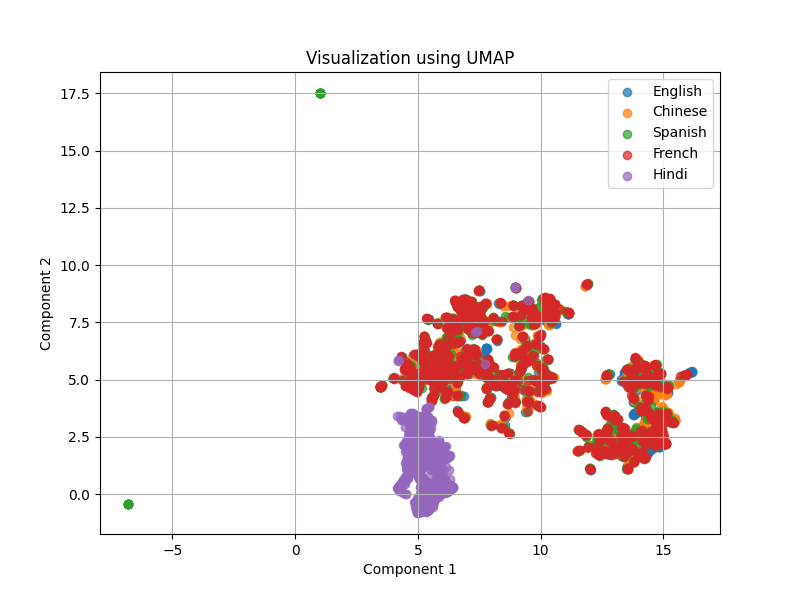}      \caption{UMAP, Layer 15}        \end{subfigure}    \vspace{0.2in}    %
\begin{subfigure}{0.18\textwidth}      \includegraphics[width=\textwidth]{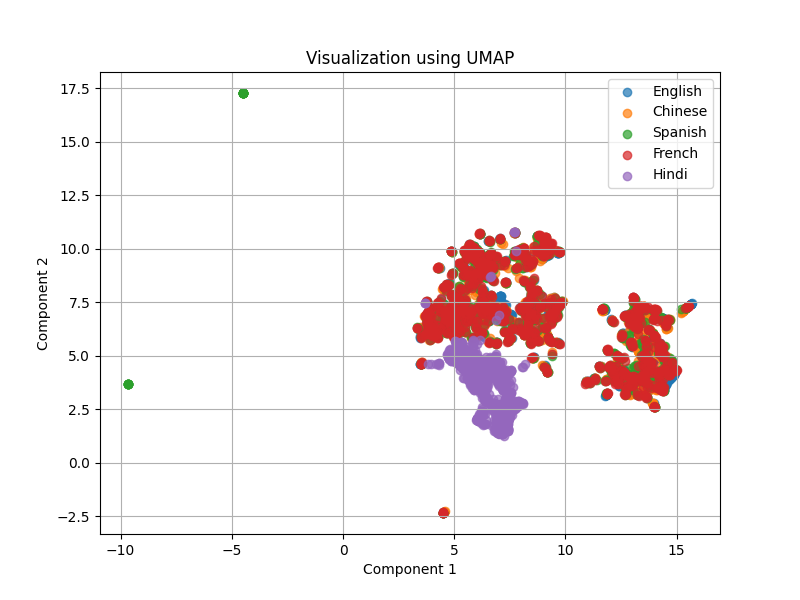}      \caption{UMAP, Layer 16}        \end{subfigure}  \hfill  \begin{subfigure}{0.18\textwidth}      \includegraphics[width=\textwidth]{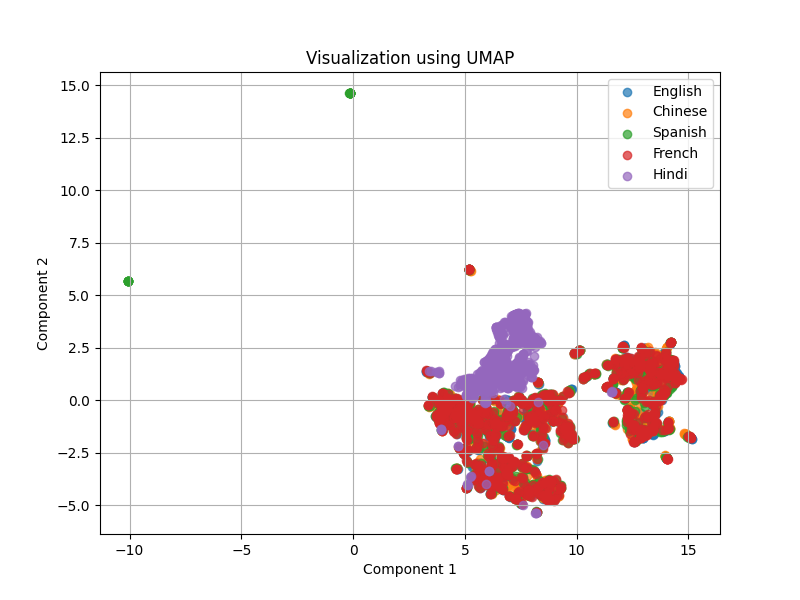}      \caption{UMAP, Layer 17}        \end{subfigure}  \hfill  \begin{subfigure}{0.18\textwidth}      \includegraphics[width=\textwidth]{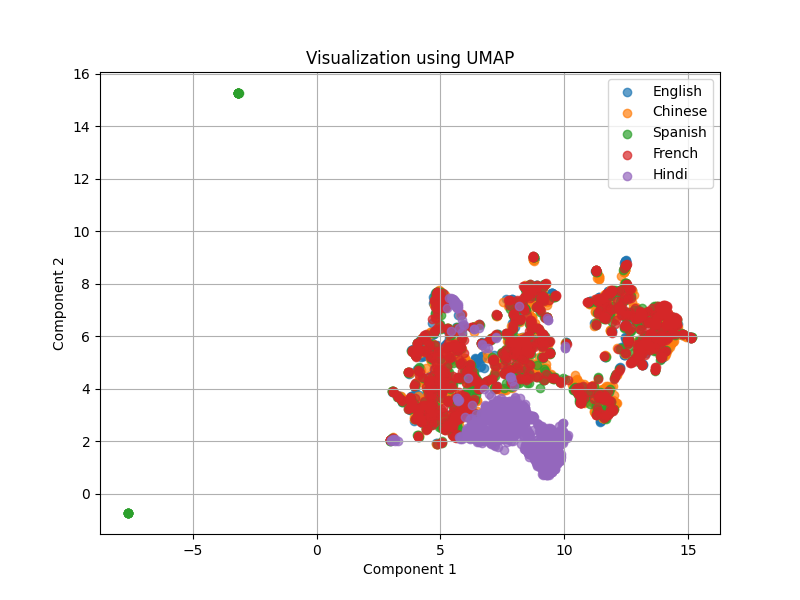}      \caption{UMAP, Layer 18}        \end{subfigure}  \hfill  \begin{subfigure}{0.18\textwidth}      \includegraphics[width=\textwidth]{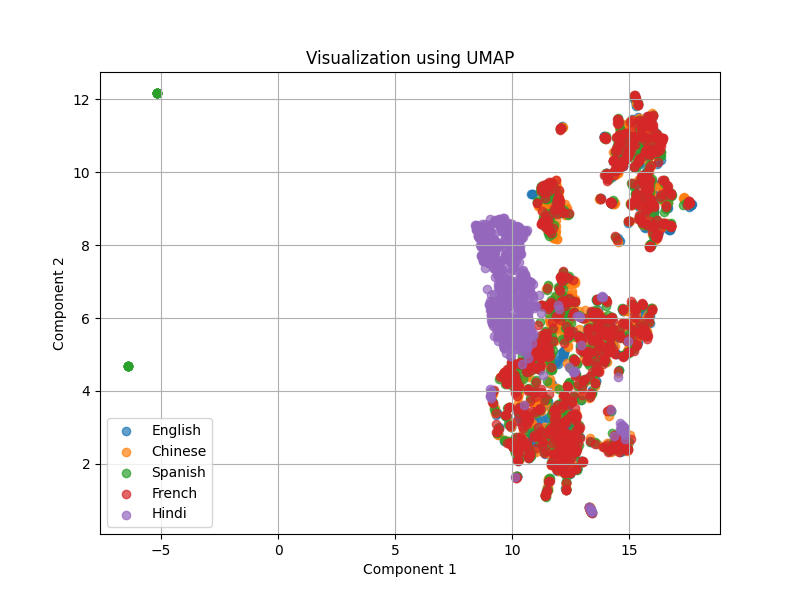}      \caption{UMAP, Layer 19}        \end{subfigure}  \hfill  \begin{subfigure}{0.18\textwidth}      \includegraphics[width=\textwidth]{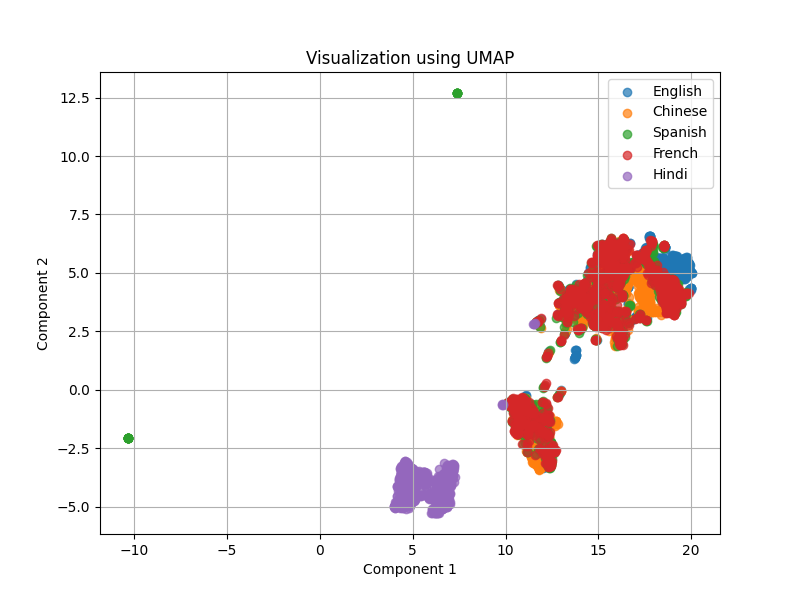}      \caption{UMAP, Layer 20}        \end{subfigure}    \vspace{0.2in}    %
\begin{subfigure}{0.18\textwidth}      \includegraphics[width=\textwidth]{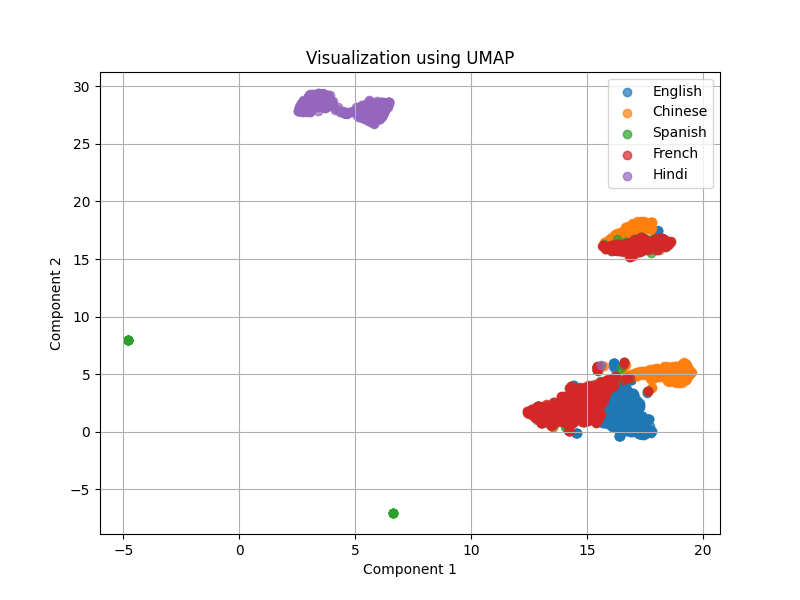}      \caption{UMAP, Layer 21}        \end{subfigure}  \hfill  \begin{subfigure}{0.18\textwidth}      \includegraphics[width=\textwidth]{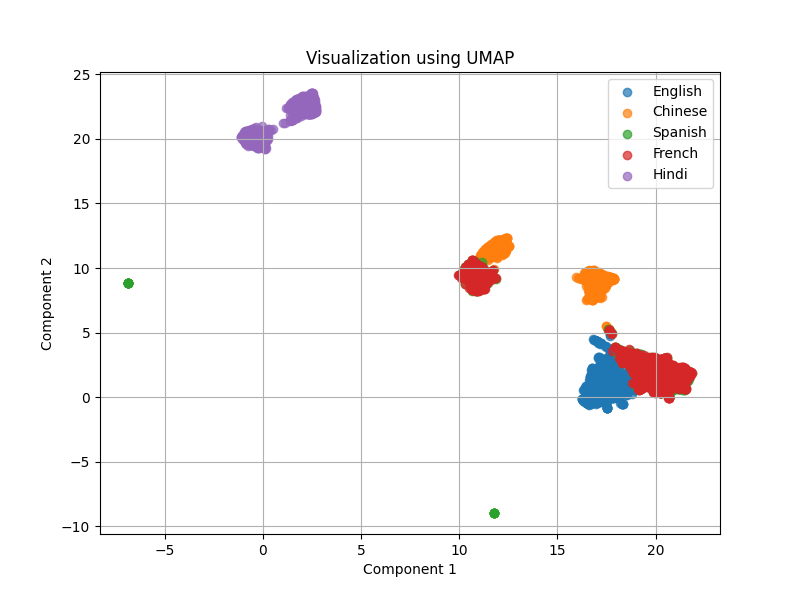}      \caption{UMAP, Layer 22}        \end{subfigure}  \hfill  \begin{subfigure}{0.18\textwidth}      \includegraphics[width=\textwidth]{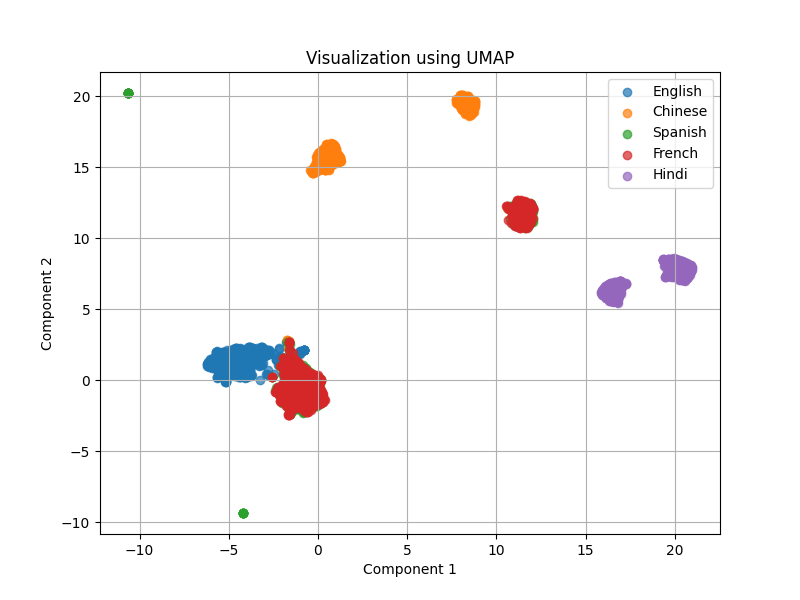}      \caption{UMAP, Layer 23}        \end{subfigure}  \hfill  \begin{subfigure}{0.18\textwidth}      \includegraphics[width=\textwidth]{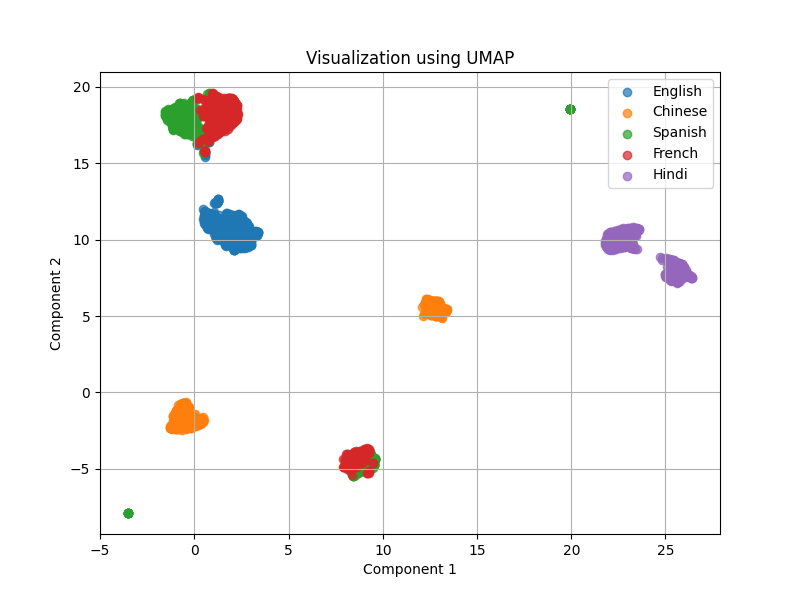}      \caption{UMAP, Layer 24}        \end{subfigure}  \hfill  \begin{subfigure}{0.18\textwidth}      \includegraphics[width=\textwidth]{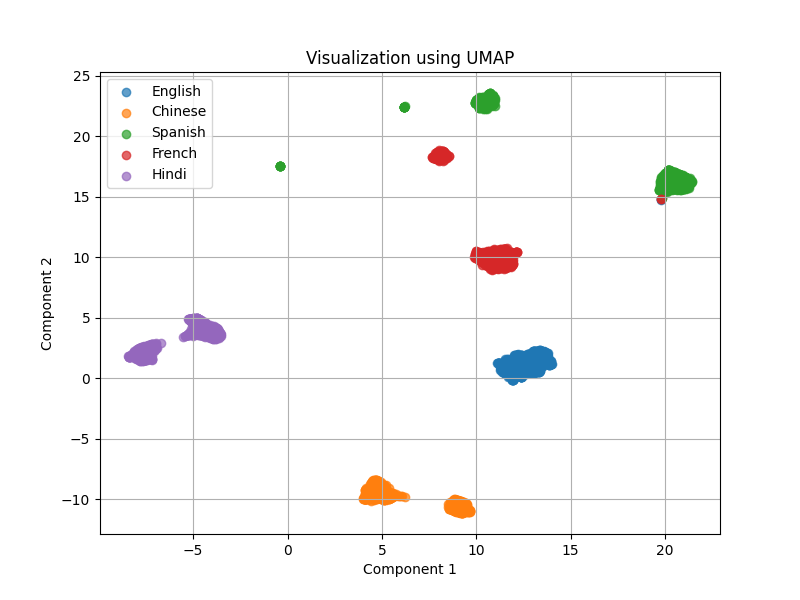}      \caption{UMAP, Layer 25}        \end{subfigure}    \vspace{0.2in}    %
\begin{subfigure}{0.18\textwidth}      \includegraphics[width=\textwidth]{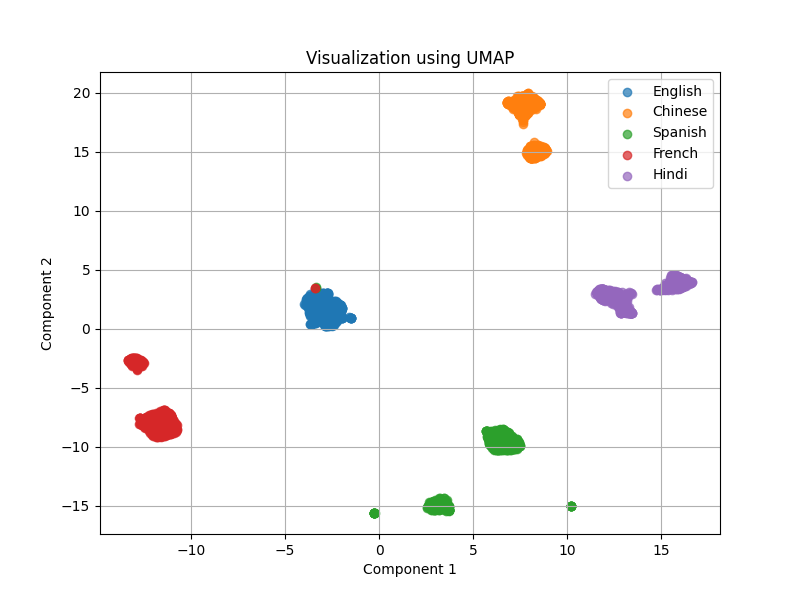}      \caption{UMAP, Layer 26}        \end{subfigure}  \hfill  \begin{subfigure}{0.18\textwidth}      \includegraphics[width=\textwidth]{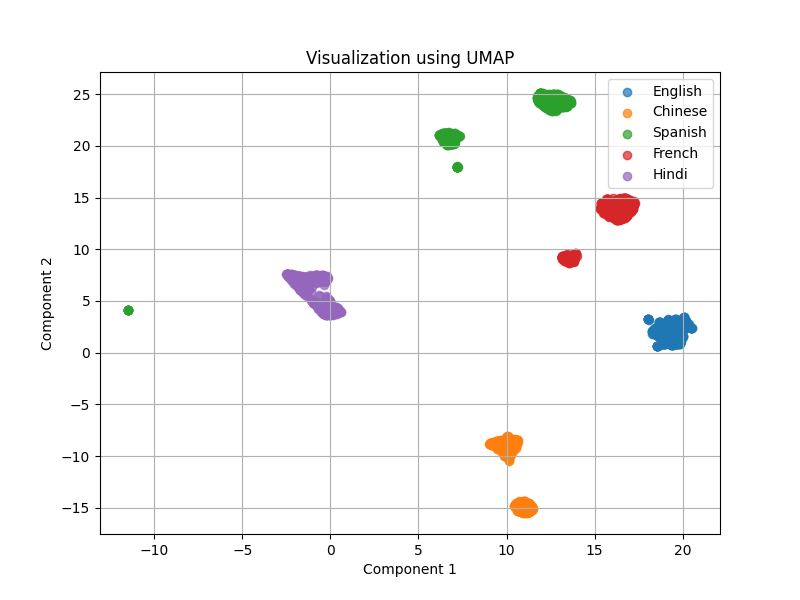}      \caption{UMAP, Layer 27}        \end{subfigure}  \hfill  \begin{subfigure}{0.18\textwidth}      \includegraphics[width=\textwidth]{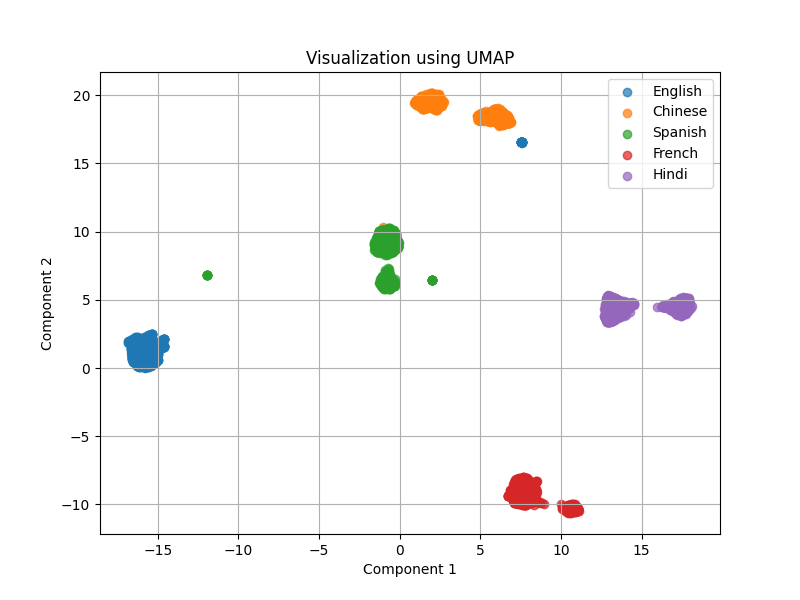}      \caption{UMAP, Layer 28}        \end{subfigure}      \caption{UMAP visualizations for layers 1-28 of Qwen2-7B-Instruct on the FOLIO dataset.}  
\end{figure*}

\begin{figure*}[htbp]
\centering
\begin{subfigure}{0.18\textwidth}
\includegraphics[width=\textwidth]{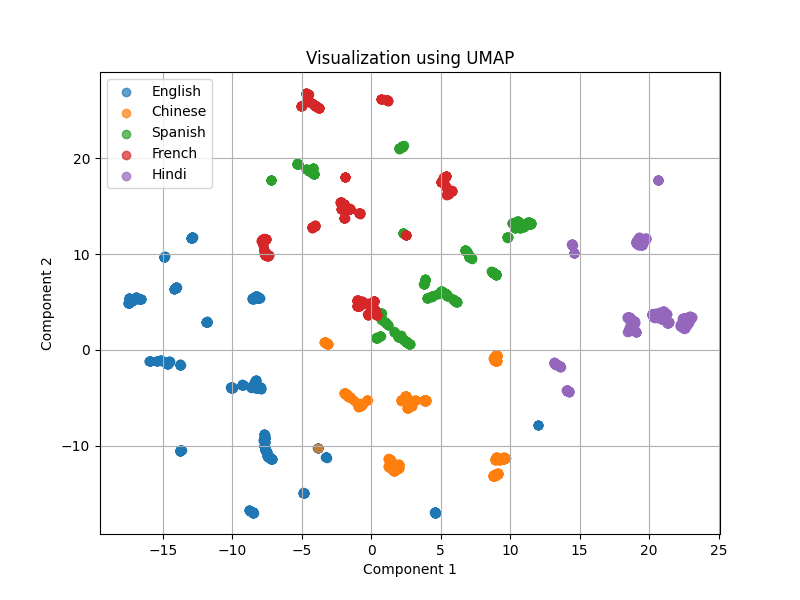}
\caption{UMAP, Layer 1}
\end{subfigure}
\hfill
\begin{subfigure}{0.18\textwidth}
\includegraphics[width=\textwidth]{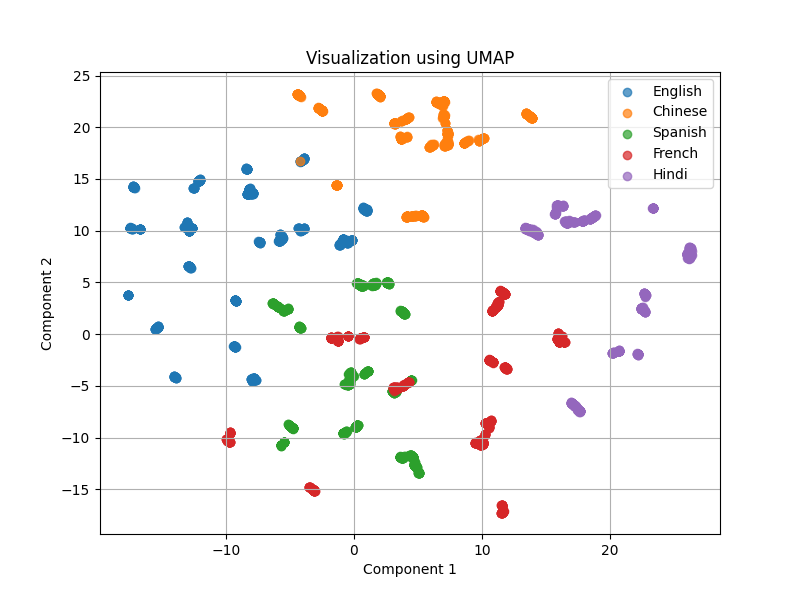}
\caption{UMAP, Layer 2}

\end{subfigure}
\hfill
\begin{subfigure}{0.18\textwidth}
\includegraphics[width=\textwidth]{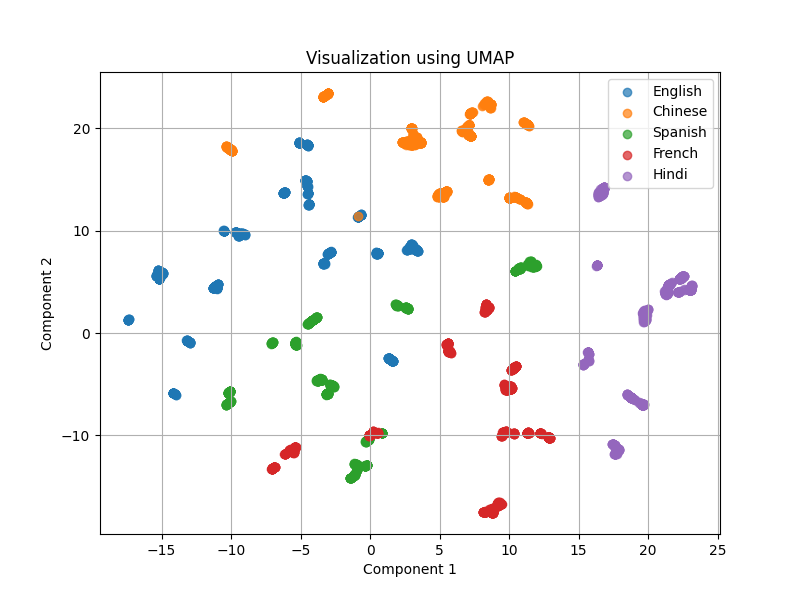}
\caption{UMAP, Layer 3}

\end{subfigure}
\hfill
\begin{subfigure}{0.18\textwidth}
\includegraphics[width=\textwidth]{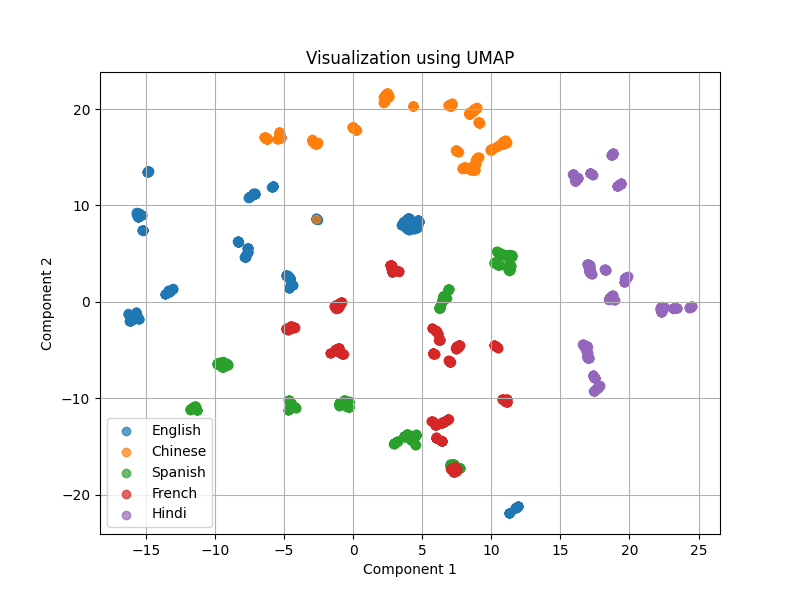}
\caption{UMAP, Layer 4}

\end{subfigure}
\hfill
\begin{subfigure}{0.18\textwidth}
\includegraphics[width=\textwidth]{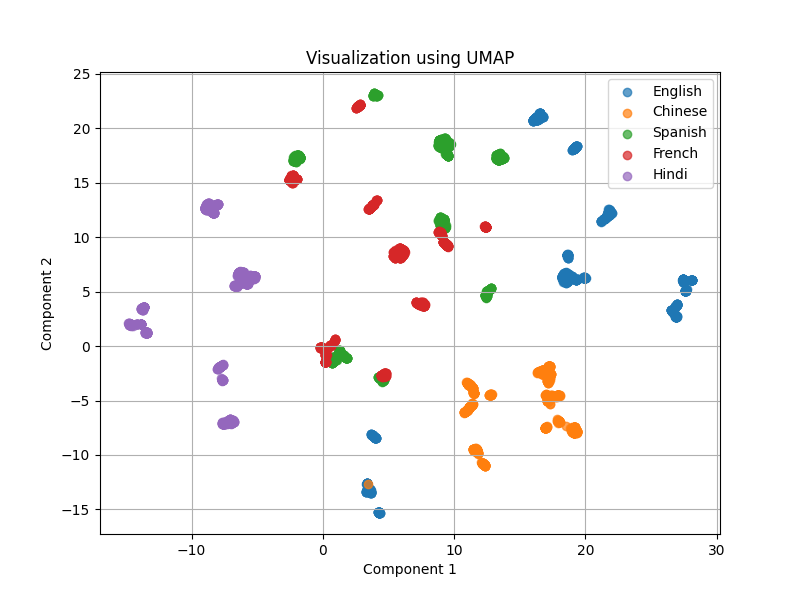}
\caption{UMAP, Layer 5}

\end{subfigure}
\vspace{0.2in} %
\begin{subfigure}{0.18\textwidth}      \includegraphics[width=\textwidth]{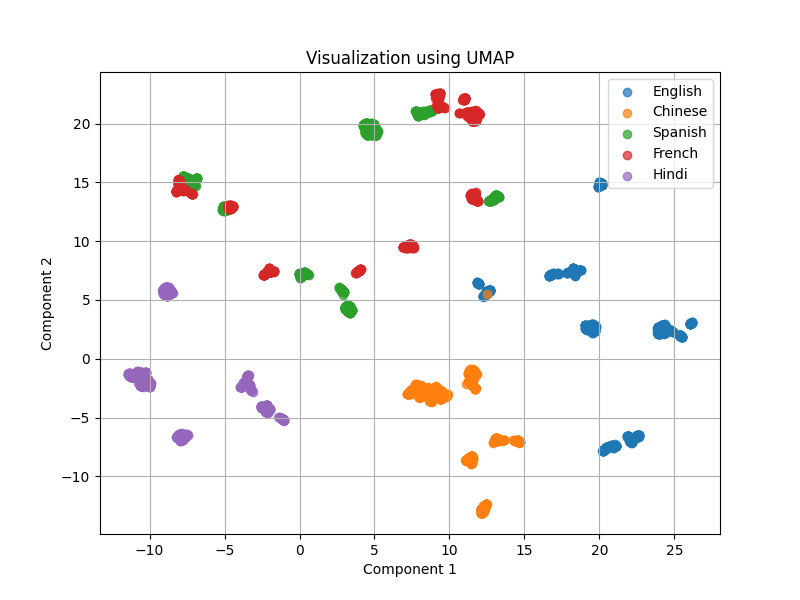}      \caption{UMAP, Layer 6}        \end{subfigure}  \hfill  \begin{subfigure}{0.18\textwidth}      \includegraphics[width=\textwidth]{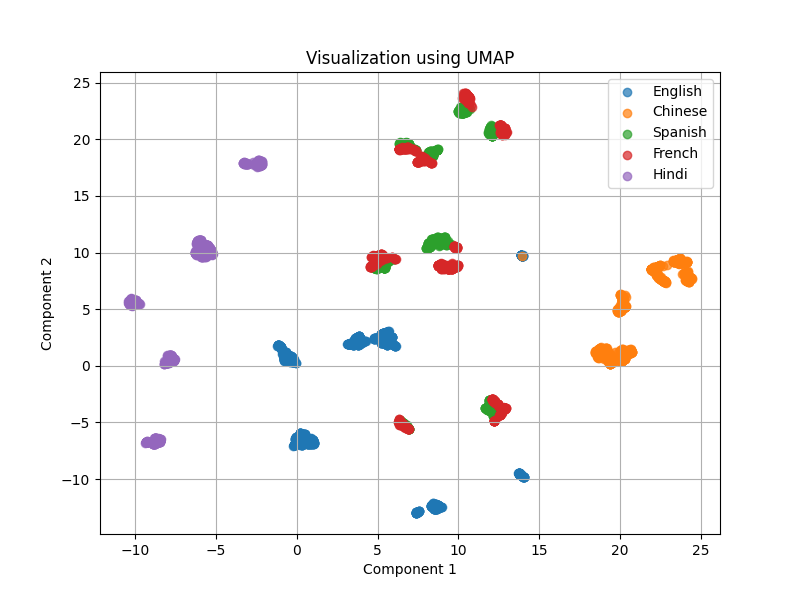}      \caption{UMAP, Layer 7}        \end{subfigure}  \hfill  \begin{subfigure}{0.18\textwidth}      \includegraphics[width=\textwidth]{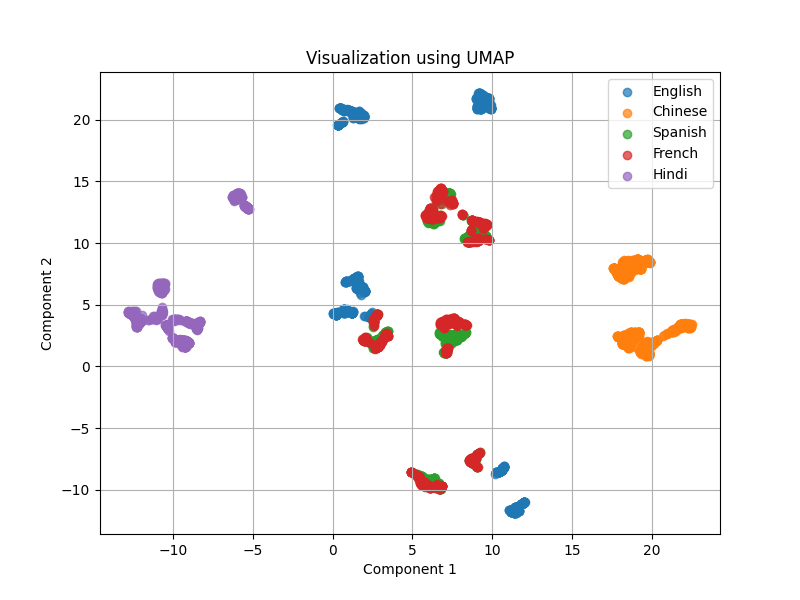}      \caption{UMAP, Layer 8}        \end{subfigure}  \hfill  \begin{subfigure}{0.18\textwidth}      \includegraphics[width=\textwidth]{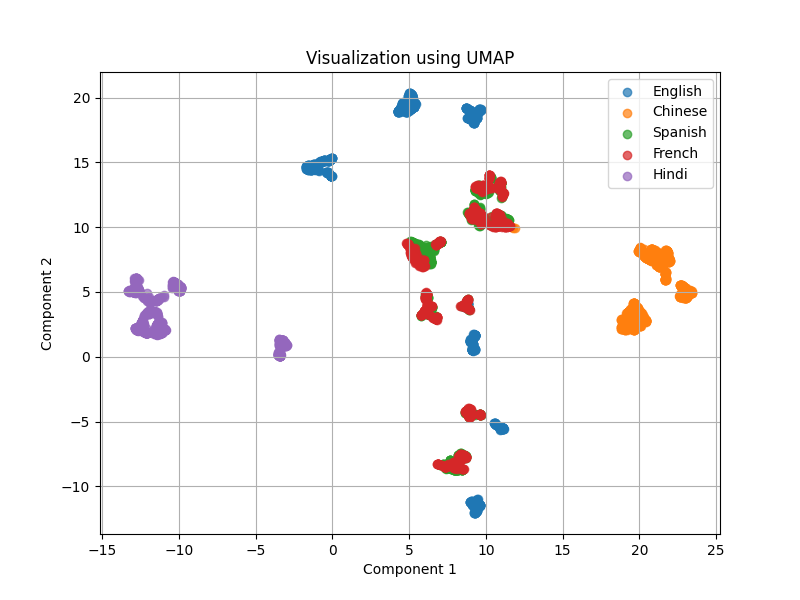}      \caption{UMAP, Layer 9}        \end{subfigure}  \hfill  \begin{subfigure}{0.18\textwidth}      \includegraphics[width=\textwidth]{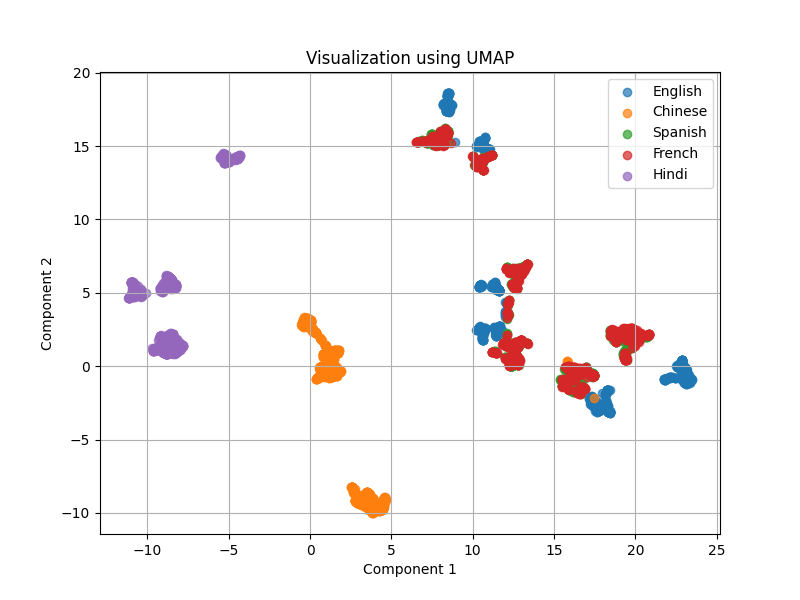}      \caption{UMAP, Layer 10}        \end{subfigure}    \vspace{0.2in}    %
\begin{subfigure}{0.18\textwidth}      \includegraphics[width=\textwidth]{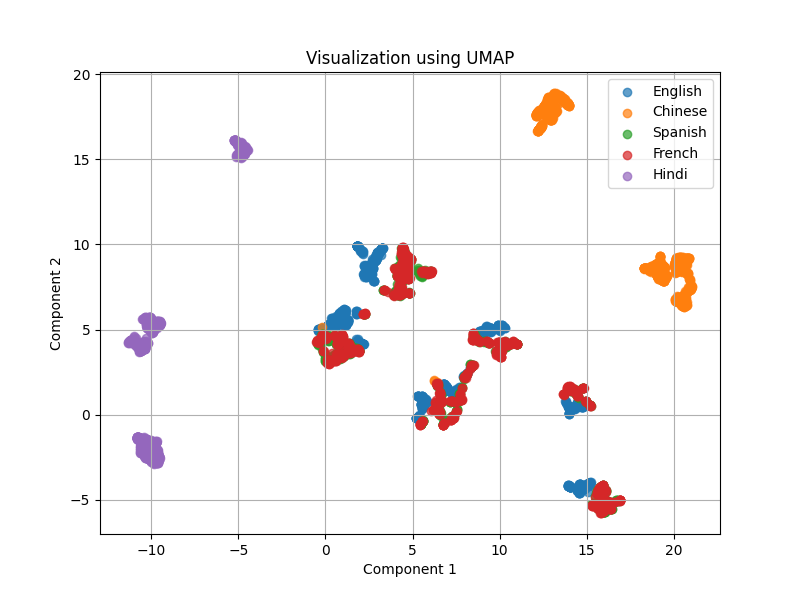}      \caption{UMAP, Layer 11}        \end{subfigure}  \hfill  \begin{subfigure}{0.18\textwidth}      \includegraphics[width=\textwidth]{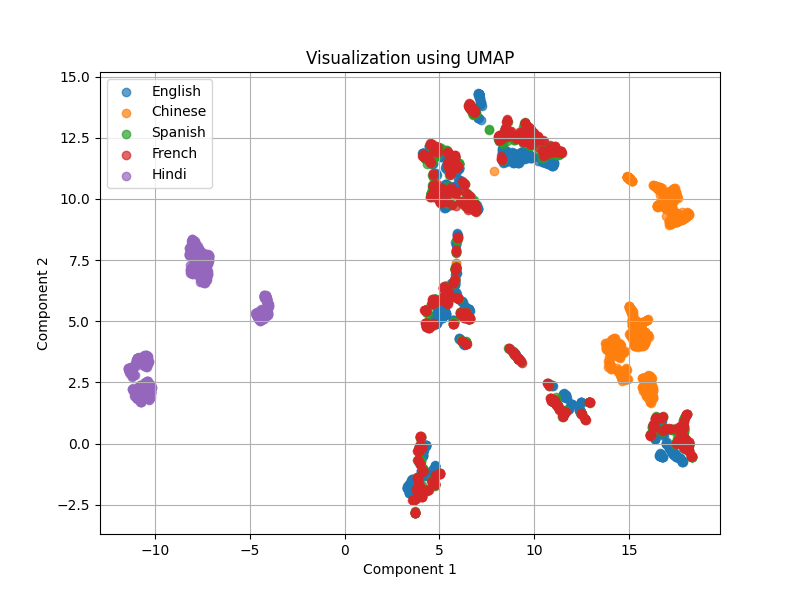}      \caption{UMAP, Layer 12}        \end{subfigure}  \hfill  \begin{subfigure}{0.18\textwidth}      \includegraphics[width=\textwidth]{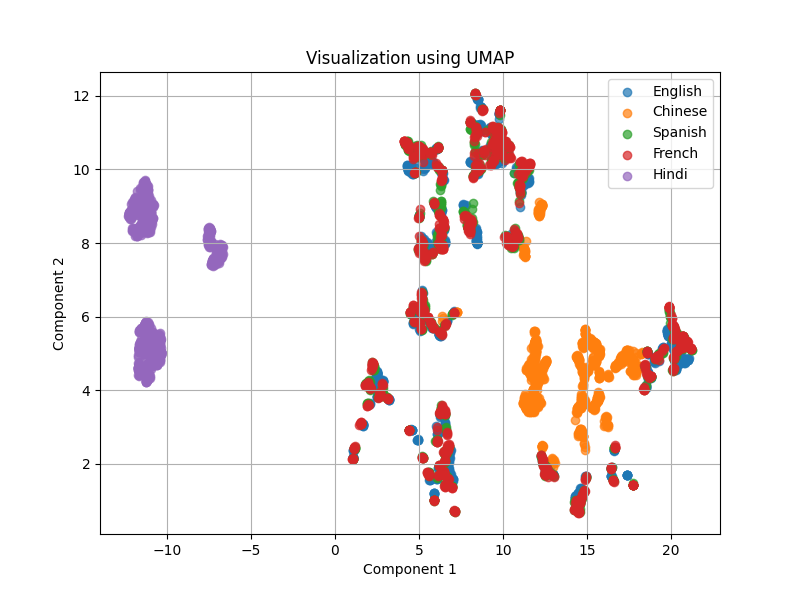}      \caption{UMAP, Layer 13}        \end{subfigure}  \hfill  \begin{subfigure}{0.18\textwidth}      \includegraphics[width=\textwidth]{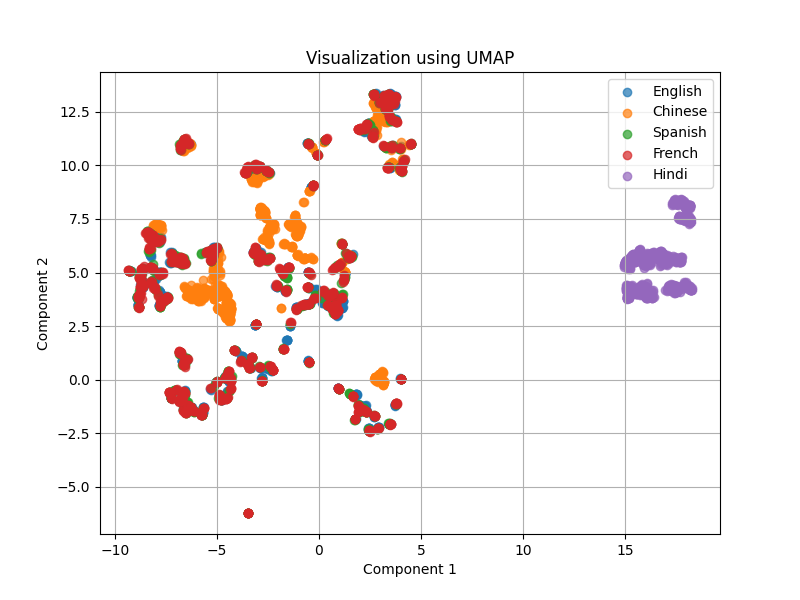}      \caption{UMAP, Layer 14}        \end{subfigure}  \hfill  \begin{subfigure}{0.18\textwidth}      \includegraphics[width=\textwidth]{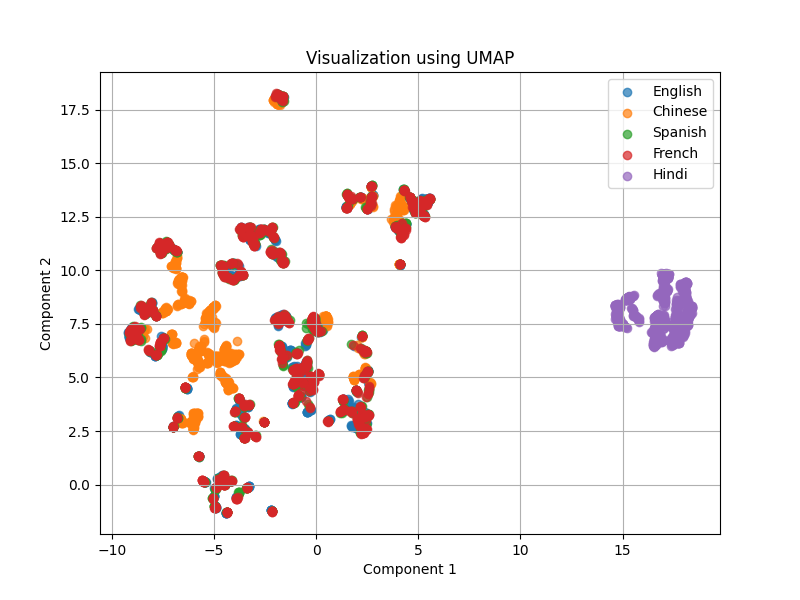}      \caption{UMAP, Layer 15}        \end{subfigure}    \vspace{0.2in}    %
\begin{subfigure}{0.18\textwidth}      \includegraphics[width=\textwidth]{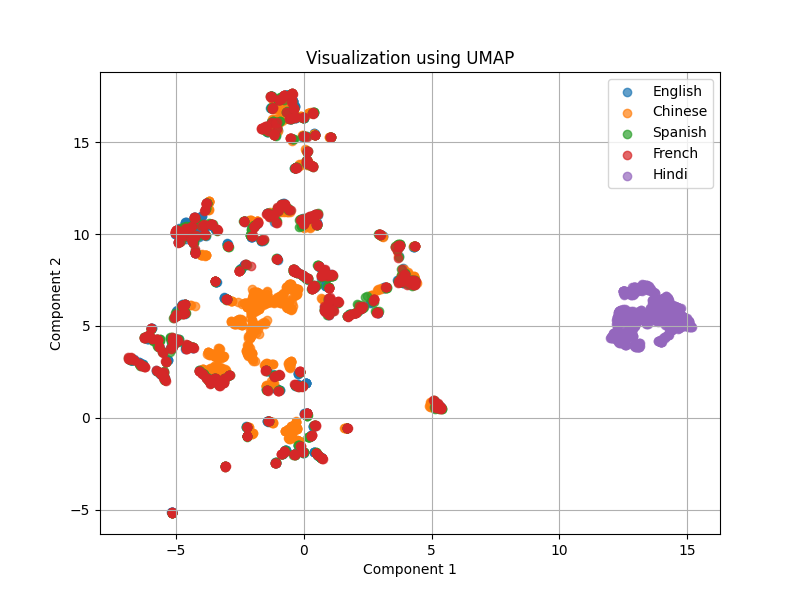}      \caption{UMAP, Layer 16}        \end{subfigure}  \hfill  \begin{subfigure}{0.18\textwidth}      \includegraphics[width=\textwidth]{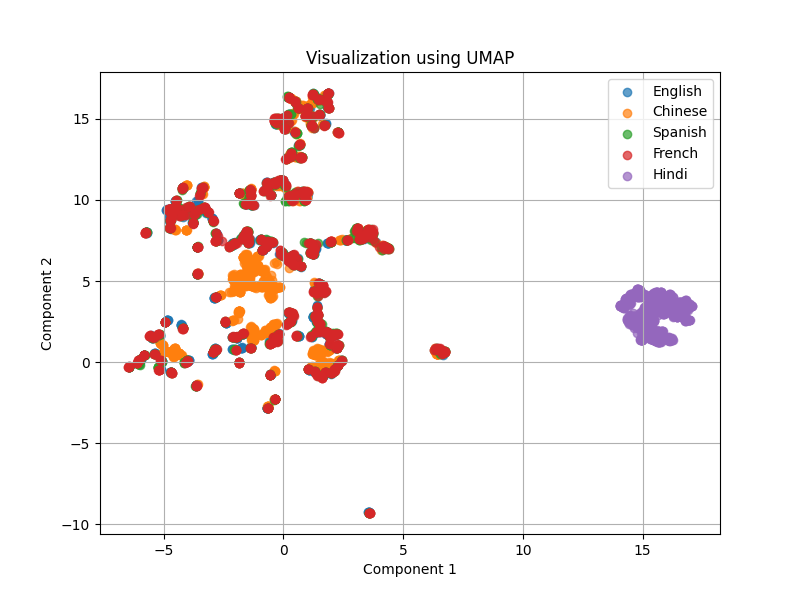}      \caption{UMAP, Layer 17}        \end{subfigure}  \hfill  \begin{subfigure}{0.18\textwidth}      \includegraphics[width=\textwidth]{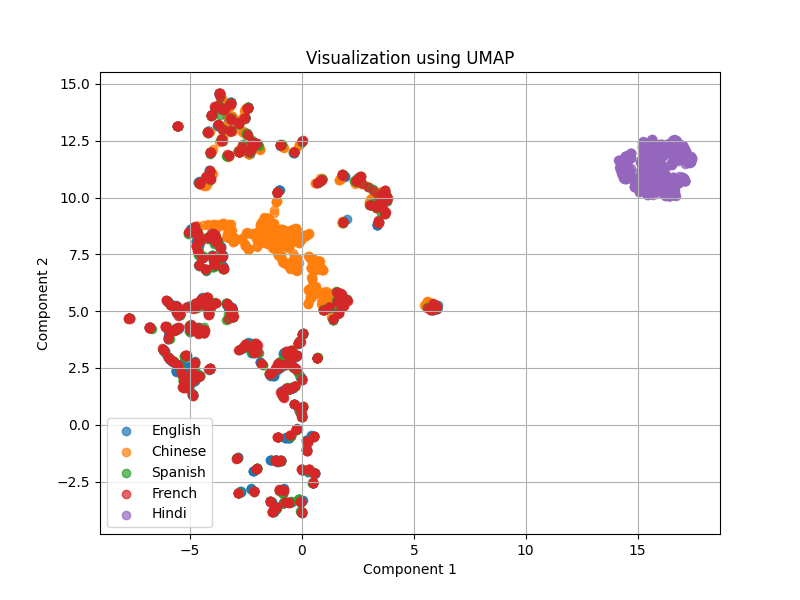}      \caption{UMAP, Layer 18}        \end{subfigure}  \hfill  \begin{subfigure}{0.18\textwidth}      \includegraphics[width=\textwidth]{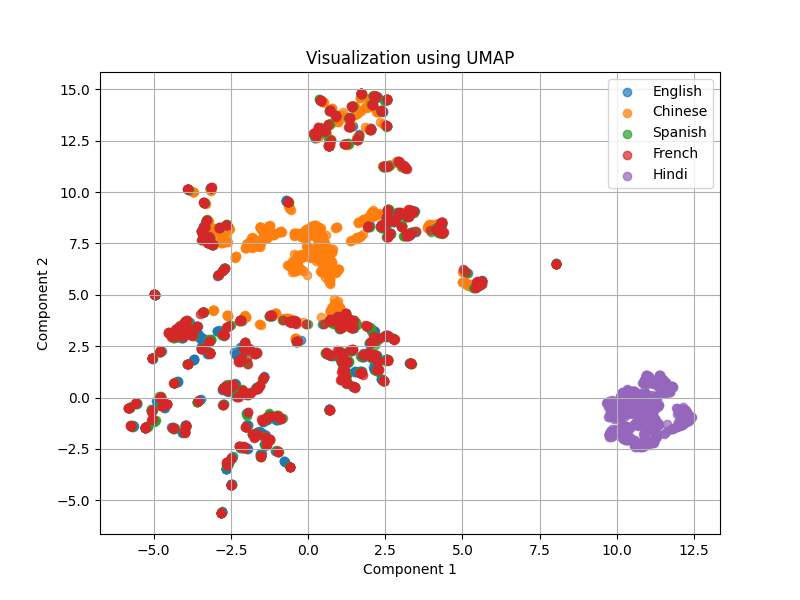}      \caption{UMAP, Layer 19}        \end{subfigure}  \hfill  \begin{subfigure}{0.18\textwidth}      \includegraphics[width=\textwidth]{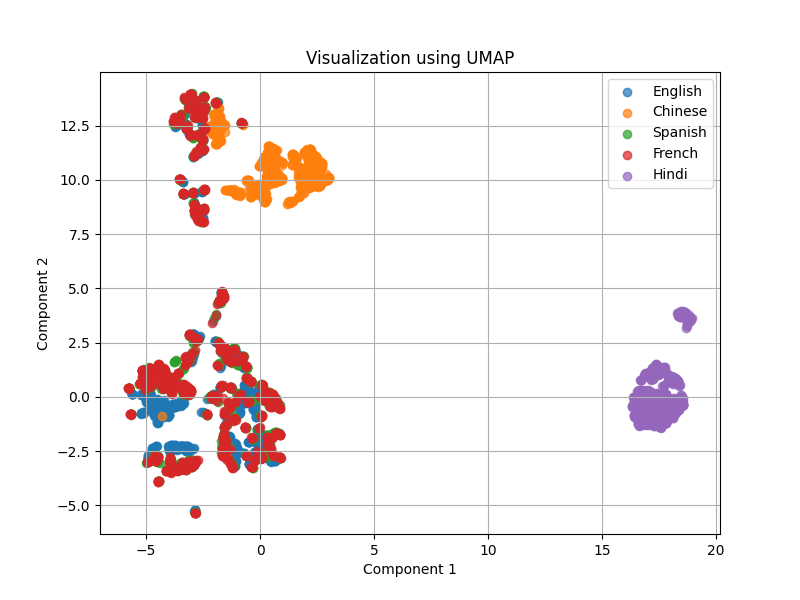}      \caption{UMAP, Layer 20}        \end{subfigure}    \vspace{0.2in}    %
\begin{subfigure}{0.18\textwidth}      \includegraphics[width=\textwidth]{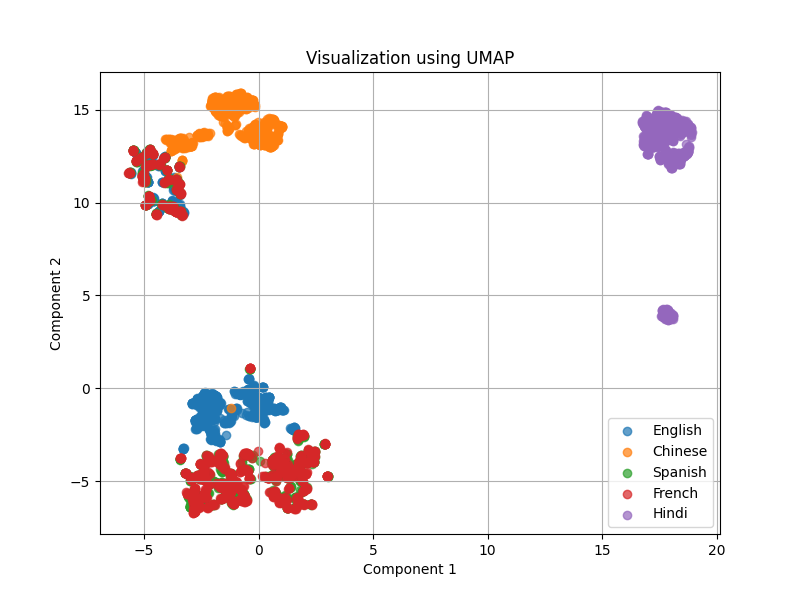}      \caption{UMAP, Layer 21}        \end{subfigure}  \hfill  \begin{subfigure}{0.18\textwidth}      \includegraphics[width=\textwidth]{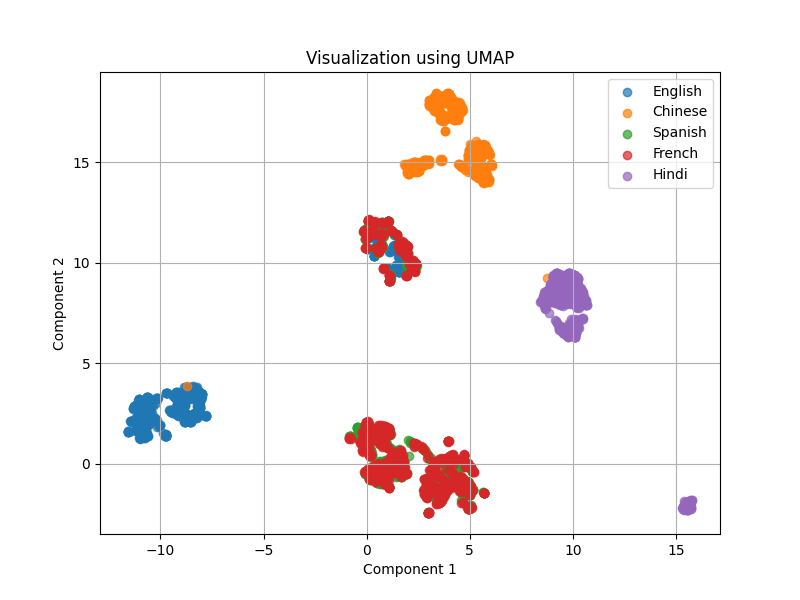}      \caption{UMAP, Layer 22}        \end{subfigure}  \hfill  \begin{subfigure}{0.18\textwidth}      \includegraphics[width=\textwidth]{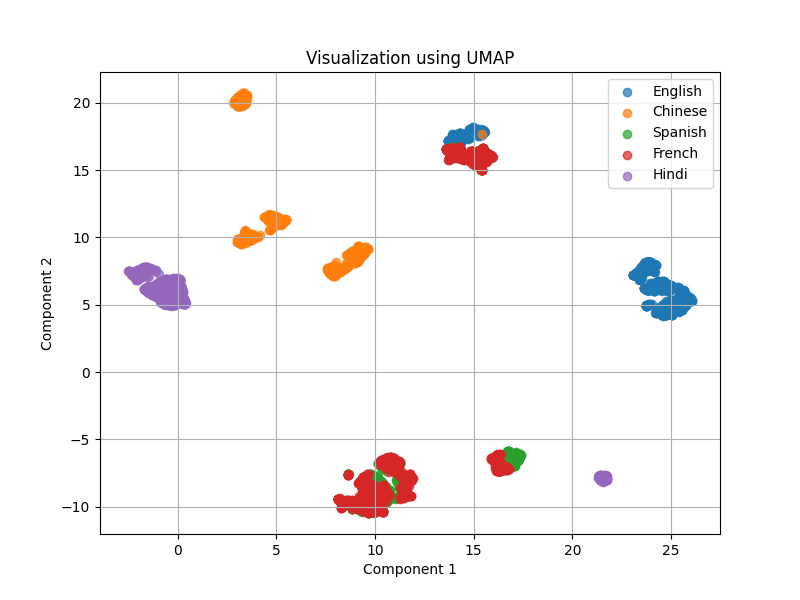}      \caption{UMAP, Layer 23}        \end{subfigure}  \hfill  \begin{subfigure}{0.18\textwidth}      \includegraphics[width=\textwidth]{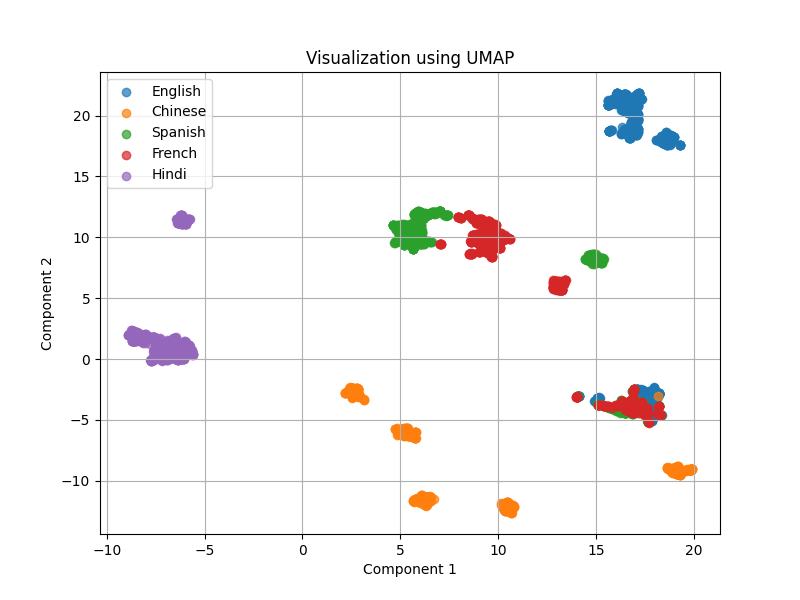}      \caption{UMAP, Layer 24}        \end{subfigure}  \hfill  \begin{subfigure}{0.18\textwidth}      \includegraphics[width=\textwidth]{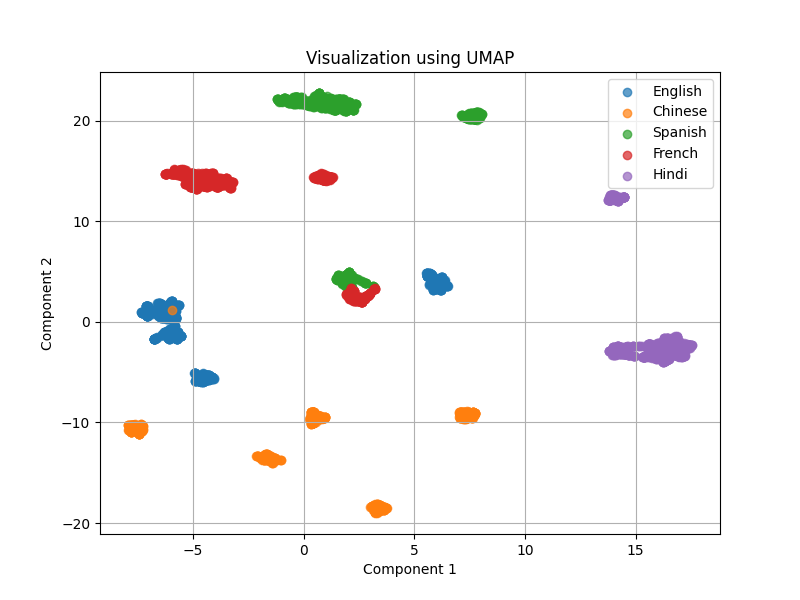}      \caption{UMAP, Layer 25}        \end{subfigure}    \vspace{0.2in}    %
\begin{subfigure}{0.18\textwidth}      \includegraphics[width=\textwidth]{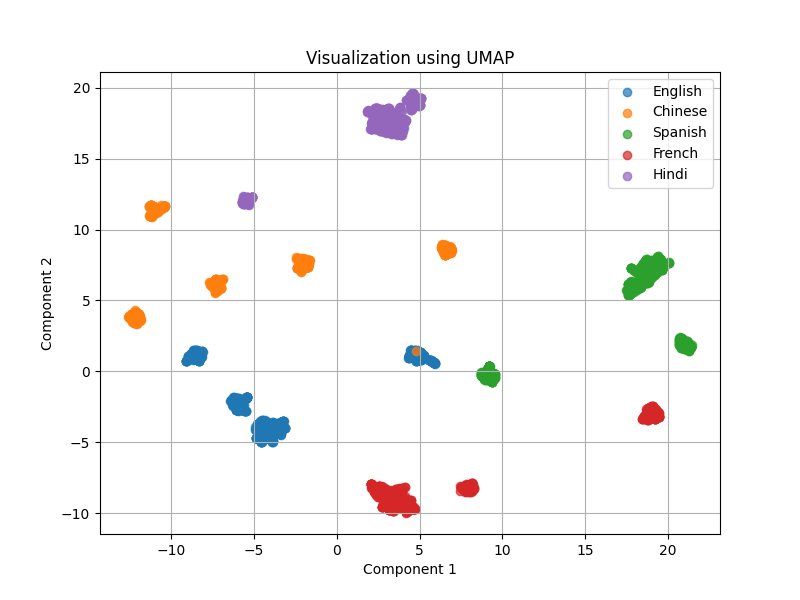}      \caption{UMAP, Layer 26}        \end{subfigure}  \hfill  \begin{subfigure}{0.18\textwidth}      \includegraphics[width=\textwidth]{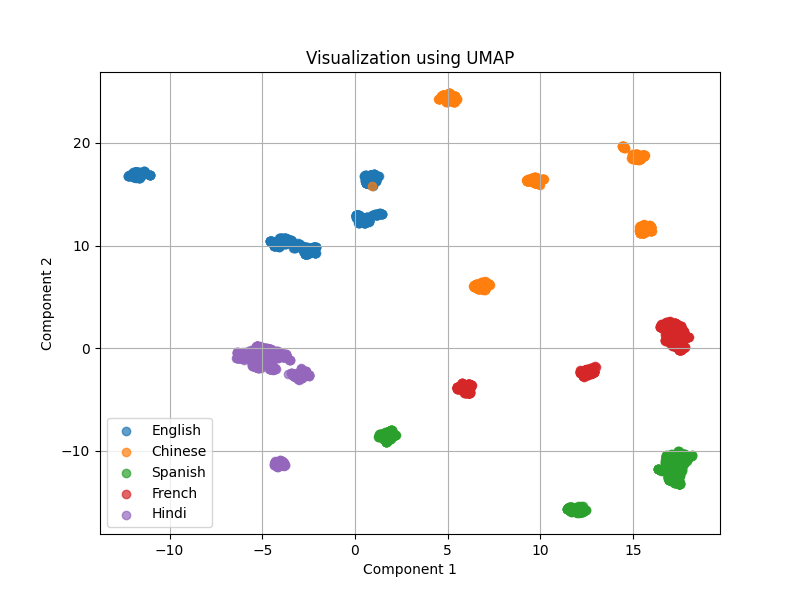}      \caption{UMAP, Layer 27}        \end{subfigure}  \hfill  \begin{subfigure}{0.18\textwidth}      \includegraphics[width=\textwidth]{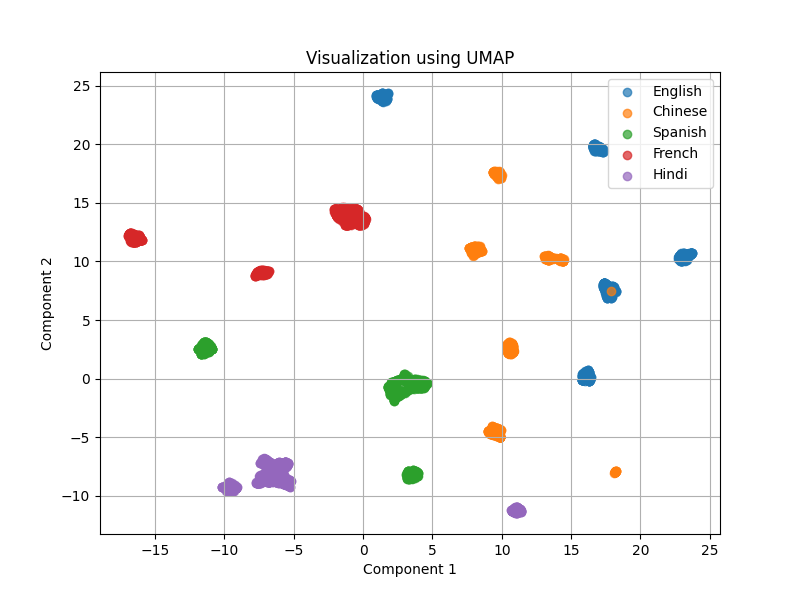}      \caption{UMAP, Layer 28}        \end{subfigure}      \caption{UMAP visualizations for layers 1-28 of Qwen2-7B-Instruct on the LogicalDeduction dataset.}  
\end{figure*}

\begin{figure*}[htbp]
\centering
\begin{subfigure}{0.18\textwidth}
\includegraphics[width=\textwidth]{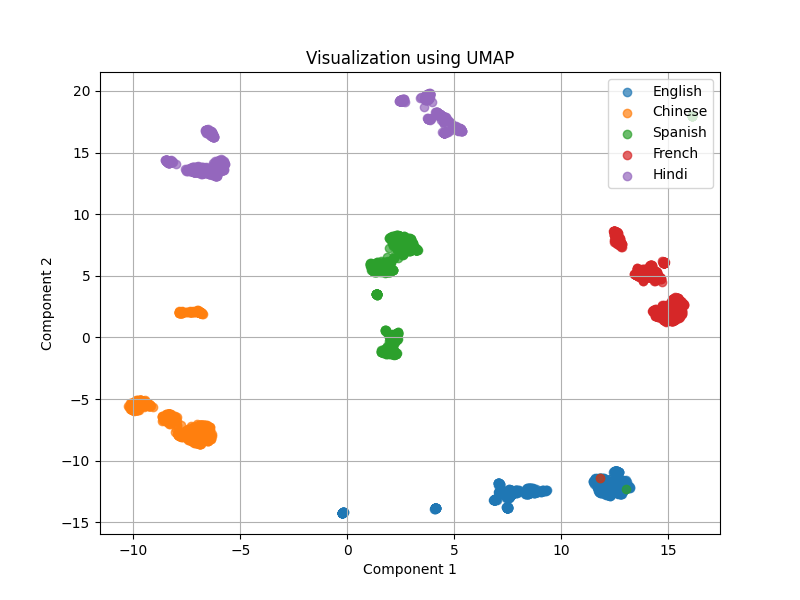}
\caption{UMAP, Layer 1}
\end{subfigure}
\hfill
\begin{subfigure}{0.18\textwidth}
\includegraphics[width=\textwidth]{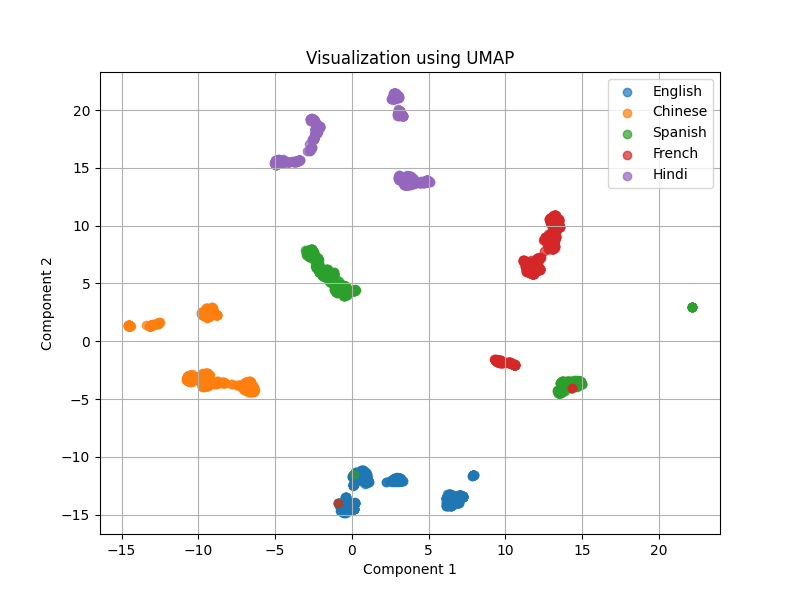}
\caption{UMAP, Layer 2}

\end{subfigure}
\hfill
\begin{subfigure}{0.18\textwidth}
\includegraphics[width=\textwidth]{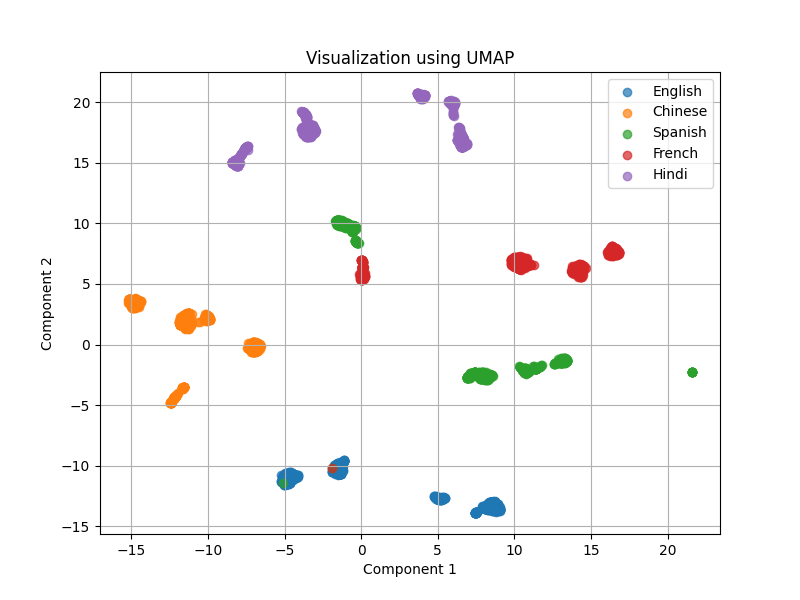}
\caption{UMAP, Layer 3}

\end{subfigure}
\hfill
\begin{subfigure}{0.18\textwidth}
\includegraphics[width=\textwidth]{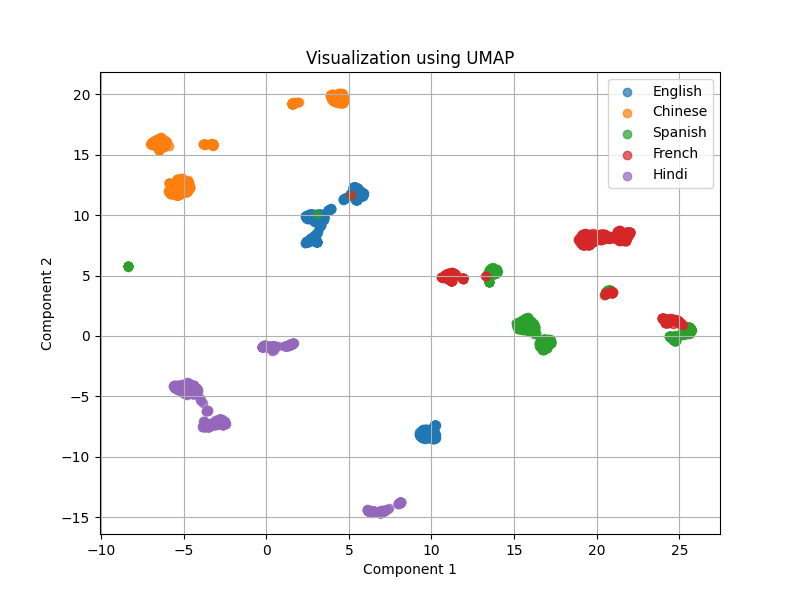}
\caption{UMAP, Layer 4}

\end{subfigure}
\hfill
\begin{subfigure}{0.18\textwidth}
\includegraphics[width=\textwidth]{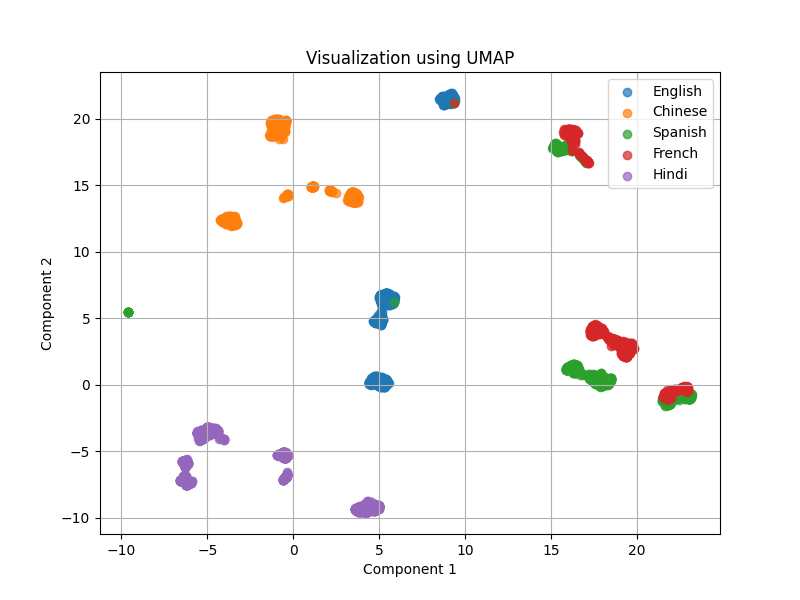}
\caption{UMAP, Layer 5}

\end{subfigure}
\vspace{0.2in} %
\begin{subfigure}{0.18\textwidth}      \includegraphics[width=\textwidth]{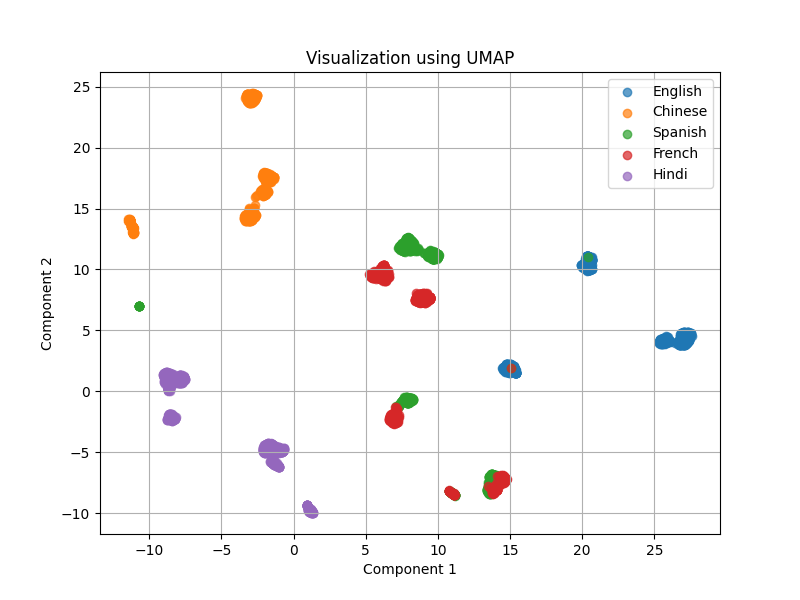}      \caption{UMAP, Layer 6}        \end{subfigure}  \hfill  \begin{subfigure}{0.18\textwidth}      \includegraphics[width=\textwidth]{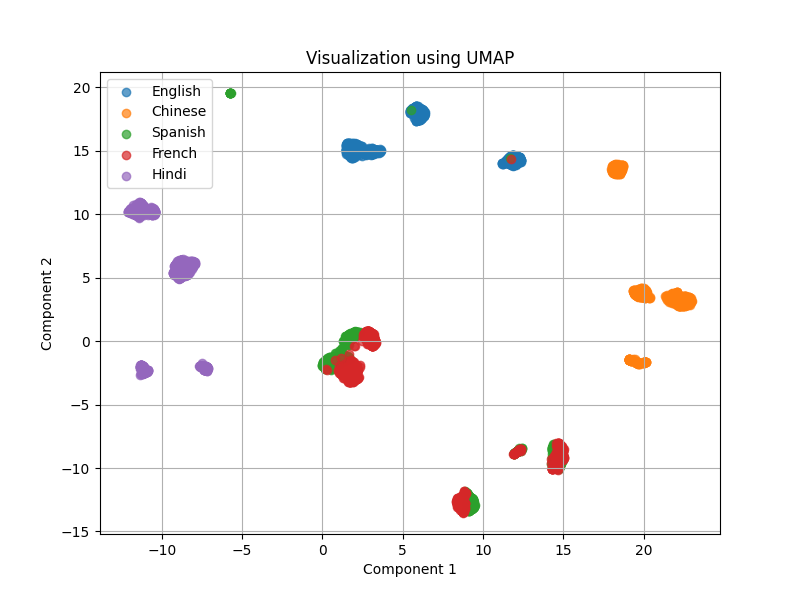}      \caption{UMAP, Layer 7}        \end{subfigure}  \hfill  \begin{subfigure}{0.18\textwidth}      \includegraphics[width=\textwidth]{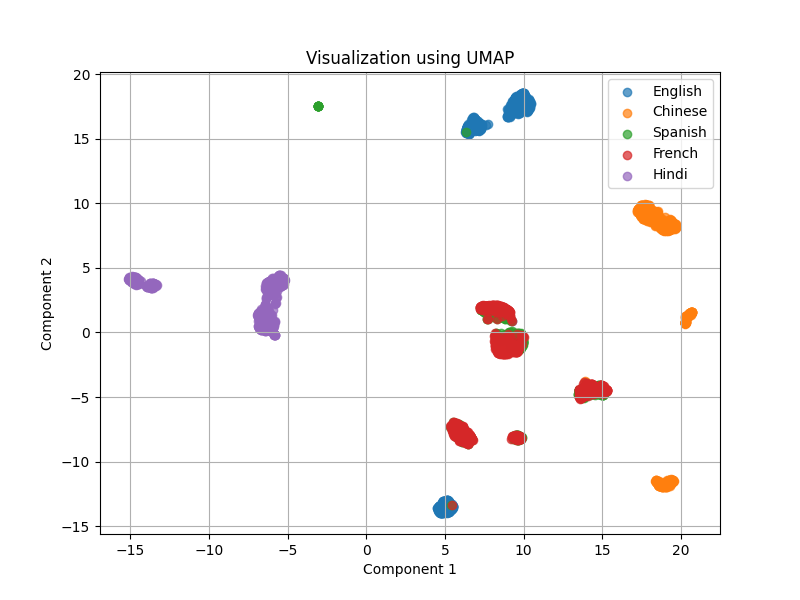}      \caption{UMAP, Layer 8}        \end{subfigure}  \hfill  \begin{subfigure}{0.18\textwidth}      \includegraphics[width=\textwidth]{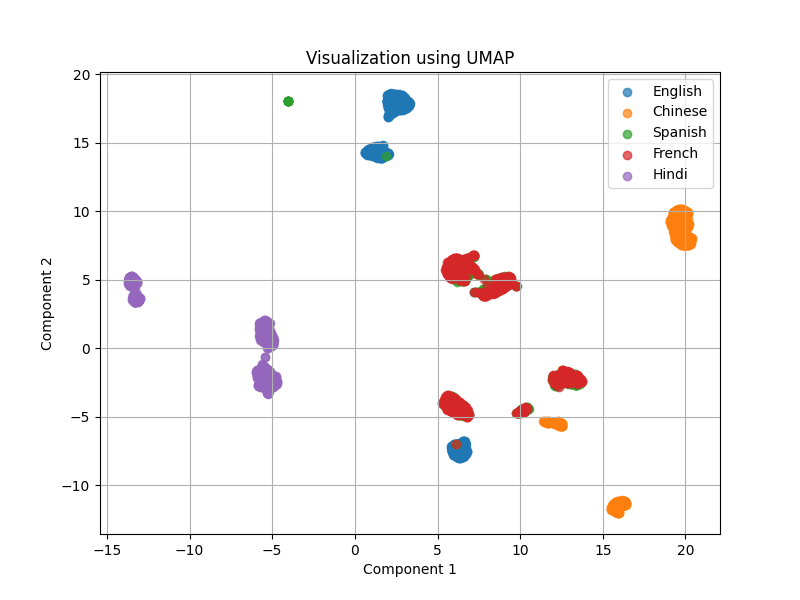}      \caption{UMAP, Layer 9}        \end{subfigure}  \hfill  \begin{subfigure}{0.18\textwidth}      \includegraphics[width=\textwidth]{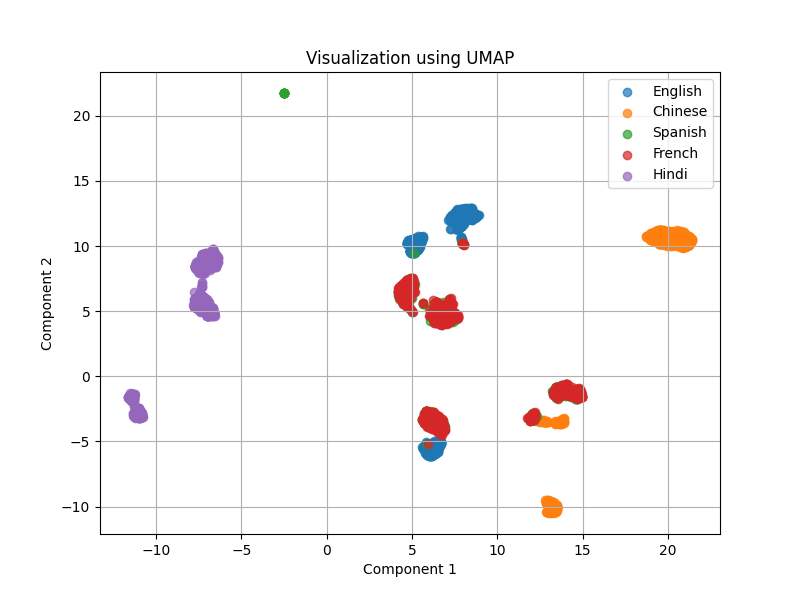}      \caption{UMAP, Layer 10}        \end{subfigure}    \vspace{0.2in}    %
\begin{subfigure}{0.18\textwidth}      \includegraphics[width=\textwidth]{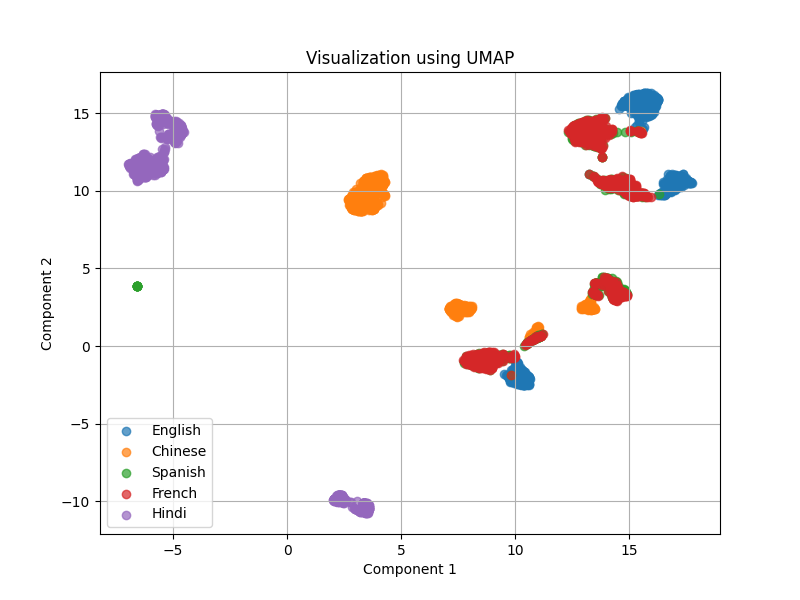}      \caption{UMAP, Layer 11}        \end{subfigure}  \hfill  \begin{subfigure}{0.18\textwidth}      \includegraphics[width=\textwidth]{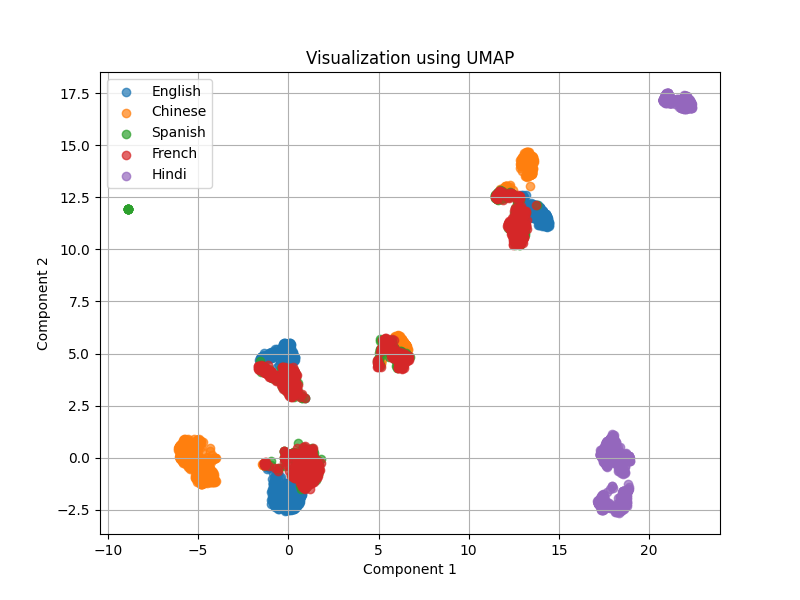}      \caption{UMAP, Layer 12}        \end{subfigure}  \hfill  \begin{subfigure}{0.18\textwidth}      \includegraphics[width=\textwidth]{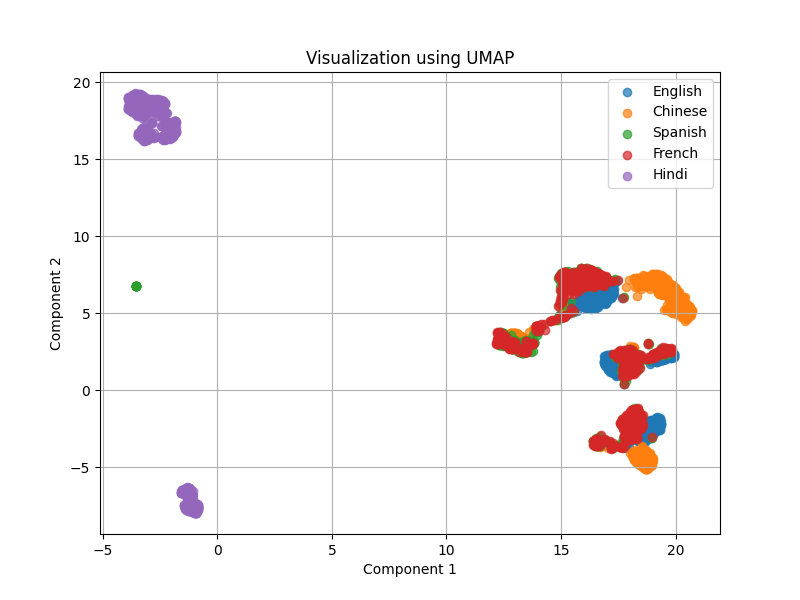}      \caption{UMAP, Layer 13}        \end{subfigure}  \hfill  \begin{subfigure}{0.18\textwidth}      \includegraphics[width=\textwidth]{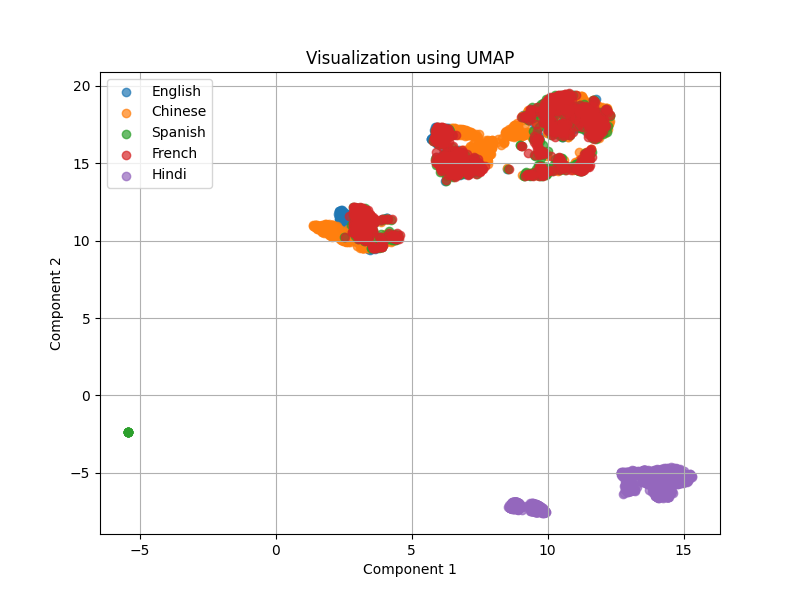}      \caption{UMAP, Layer 14}        \end{subfigure}  \hfill  \begin{subfigure}{0.18\textwidth}      \includegraphics[width=\textwidth]{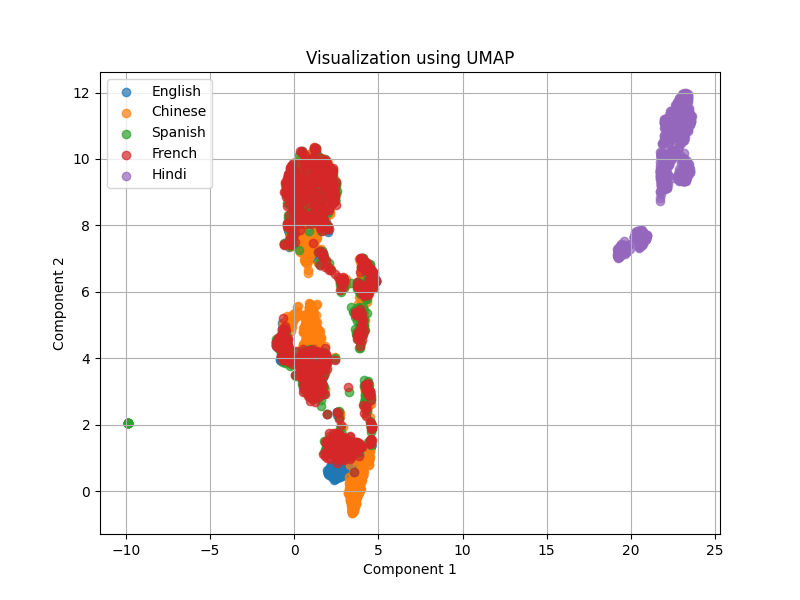}      \caption{UMAP, Layer 15}        \end{subfigure}    \vspace{0.2in}    %
\begin{subfigure}{0.18\textwidth}      \includegraphics[width=\textwidth]{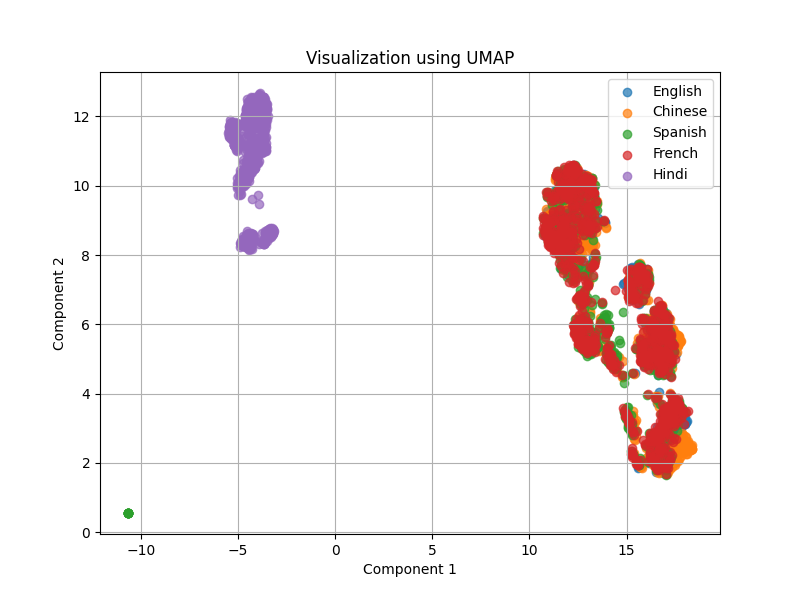}      \caption{UMAP, Layer 16}        \end{subfigure}  \hfill  \begin{subfigure}{0.18\textwidth}      \includegraphics[width=\textwidth]{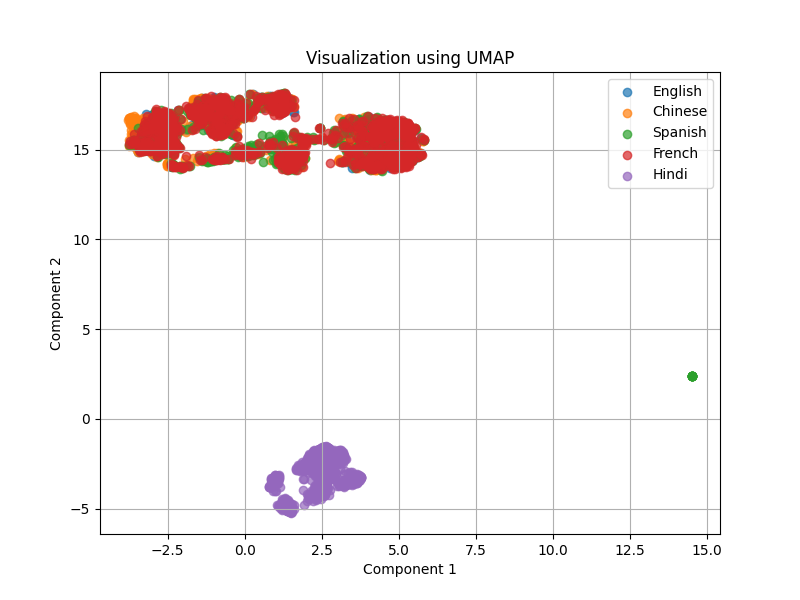}      \caption{UMAP, Layer 17}        \end{subfigure}  \hfill  \begin{subfigure}{0.18\textwidth}      \includegraphics[width=\textwidth]{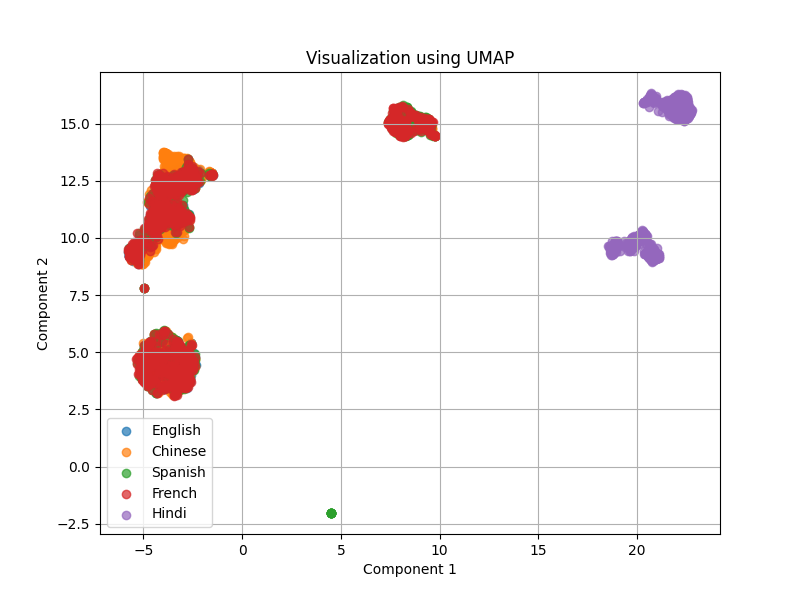}      \caption{UMAP, Layer 18}        \end{subfigure}  \hfill  \begin{subfigure}{0.18\textwidth}      \includegraphics[width=\textwidth]{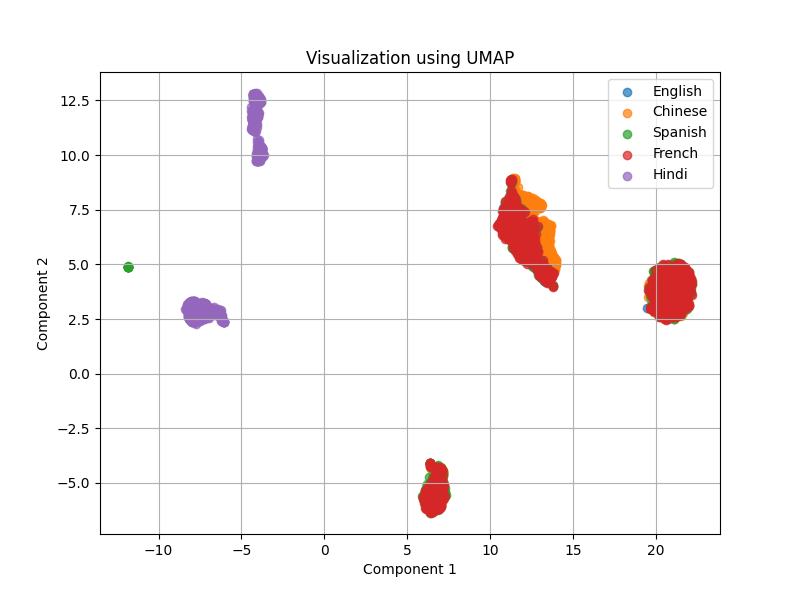}      \caption{UMAP, Layer 19}        \end{subfigure}  \hfill  \begin{subfigure}{0.18\textwidth}      \includegraphics[width=\textwidth]{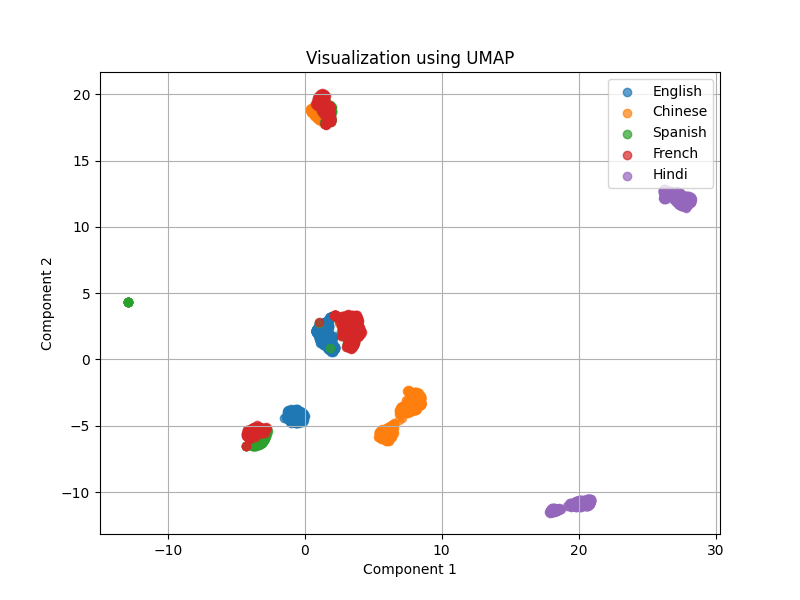}      \caption{UMAP, Layer 20}        \end{subfigure}    \vspace{0.2in}    %
\begin{subfigure}{0.18\textwidth}      \includegraphics[width=\textwidth]{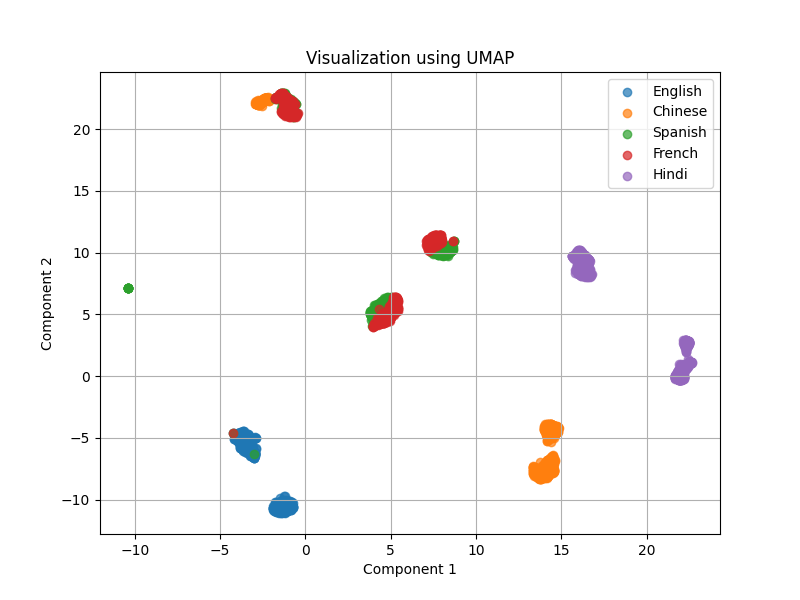}      \caption{UMAP, Layer 21}        \end{subfigure}  \hfill  \begin{subfigure}{0.18\textwidth}      \includegraphics[width=\textwidth]{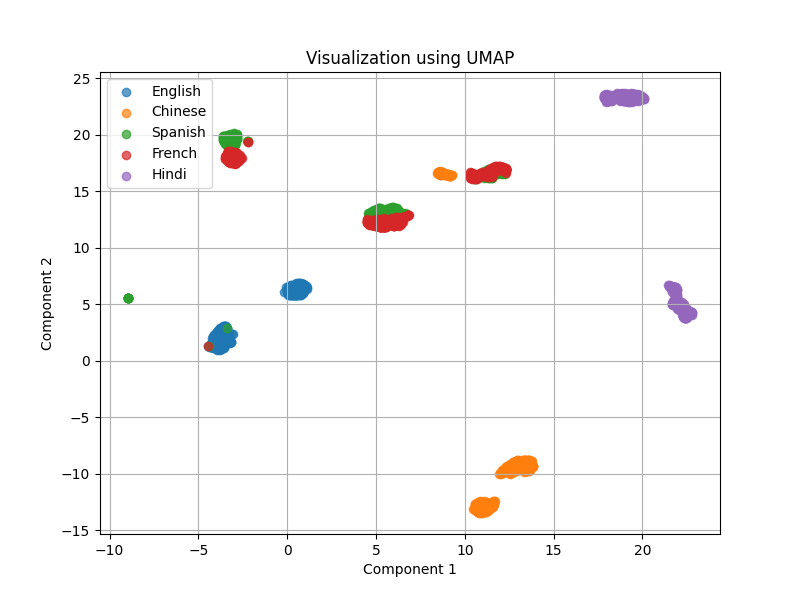}      \caption{UMAP, Layer 22}        \end{subfigure}  \hfill  \begin{subfigure}{0.18\textwidth}      \includegraphics[width=\textwidth]{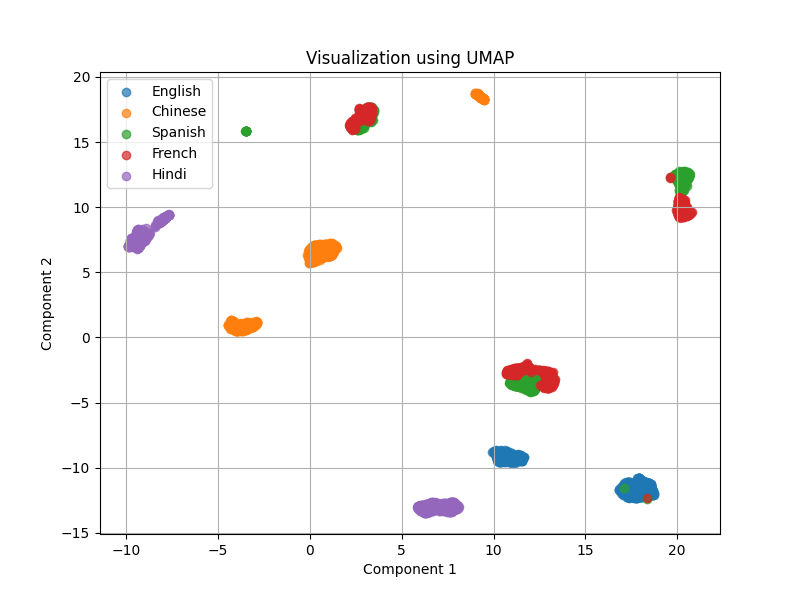}      \caption{UMAP, Layer 23}        \end{subfigure}  \hfill  \begin{subfigure}{0.18\textwidth}      \includegraphics[width=\textwidth]{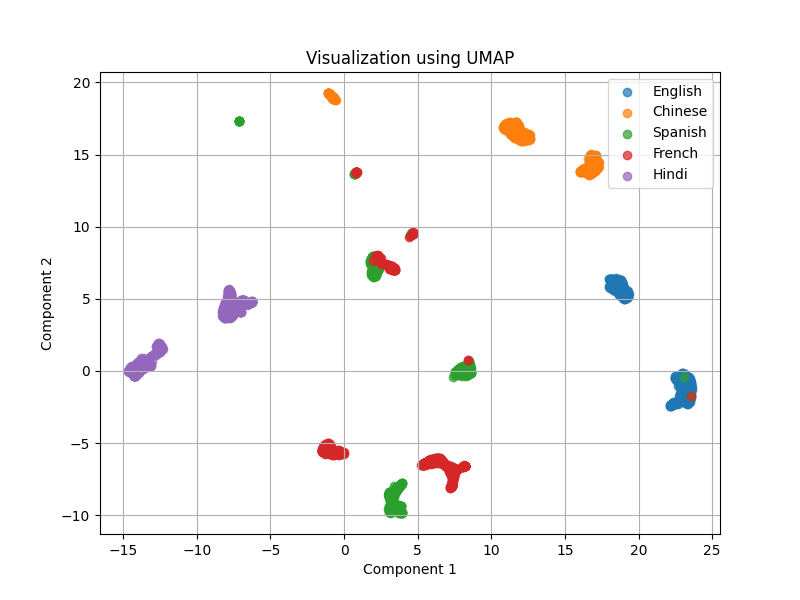}      \caption{UMAP, Layer 24}        \end{subfigure}  \hfill  \begin{subfigure}{0.18\textwidth}      \includegraphics[width=\textwidth]{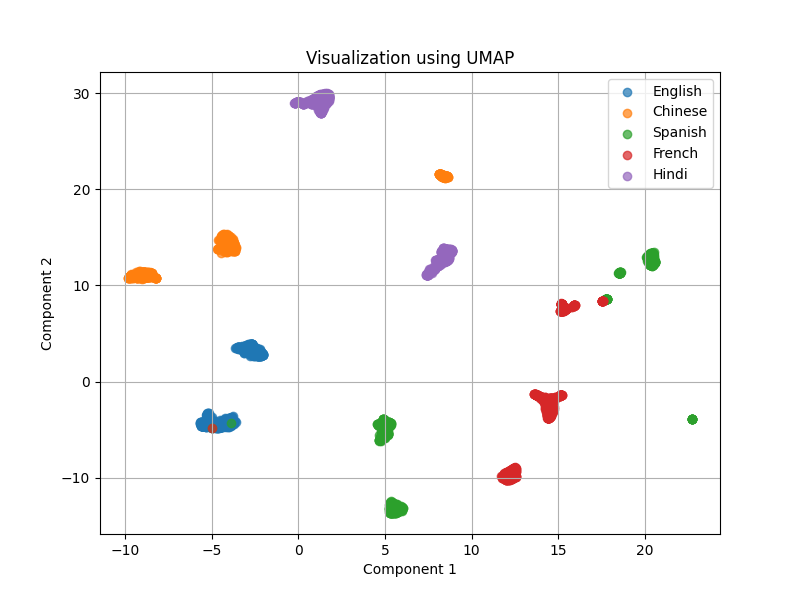}      \caption{UMAP, Layer 25}        \end{subfigure}    \vspace{0.2in}    %
\begin{subfigure}{0.18\textwidth}      \includegraphics[width=\textwidth]{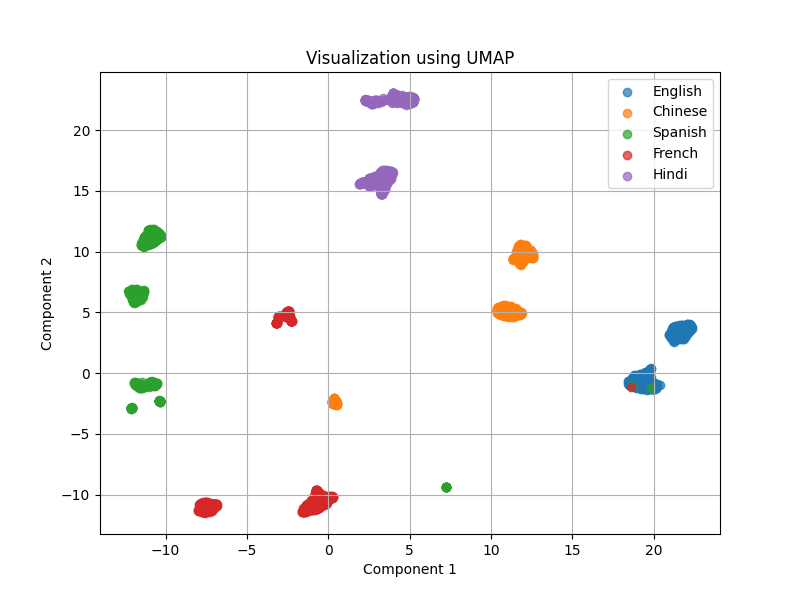}      \caption{UMAP, Layer 26}        \end{subfigure}  \hfill  \begin{subfigure}{0.18\textwidth}      \includegraphics[width=\textwidth]{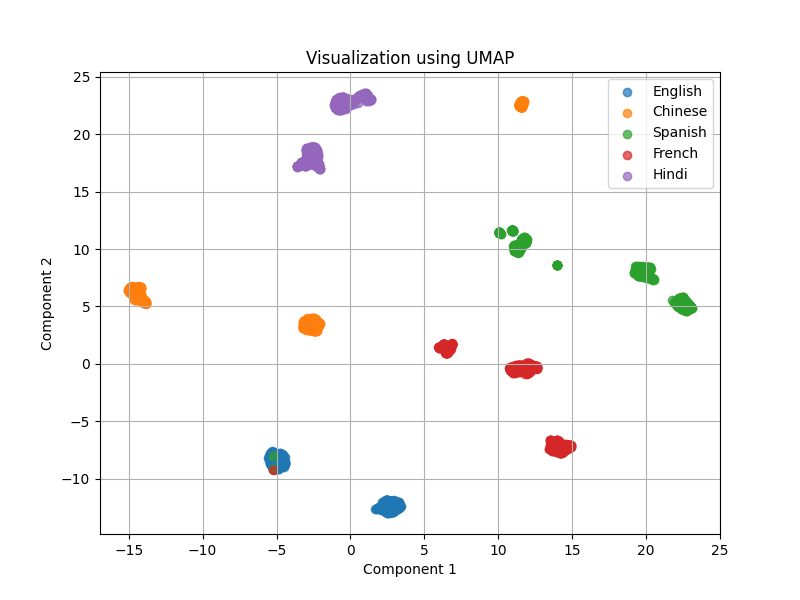}      \caption{UMAP, Layer 27}        \end{subfigure}  \hfill  \begin{subfigure}{0.18\textwidth}      \includegraphics[width=\textwidth]{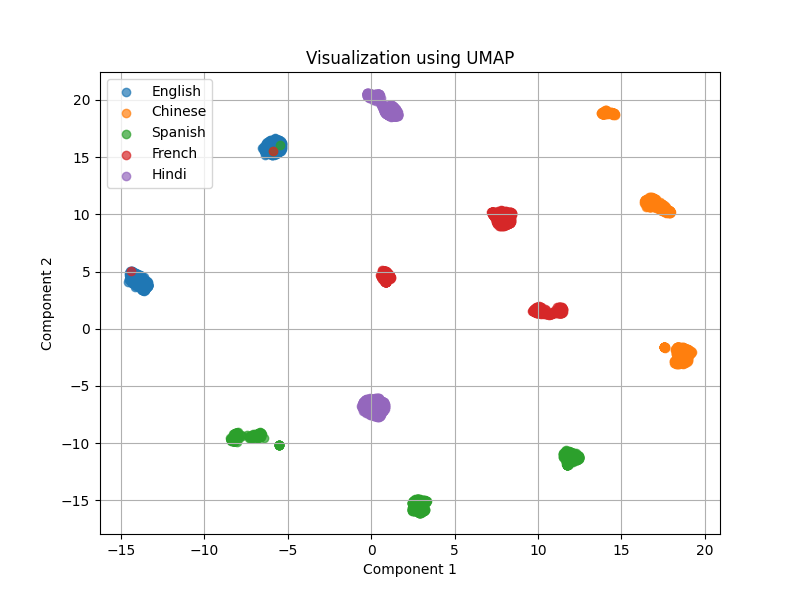}      \caption{UMAP, Layer 28}        \end{subfigure}      \caption{UMAP visualizations for layers 1-28 of Qwen2-7B-Instruct on the ProofWriter dataset.}  
\end{figure*}

\end{document}

%% file: math_commands.tex
\usepackage{amsmath,amsfonts,bm}

\def\eqref#1{equation~\ref{#1}}

\def\1{\bm{1}}

\DeclareMathAlphabet{\mathsfit}{\encodingdefault}{\sfdefault}{m}{sl}
\SetMathAlphabet{\mathsfit}{bold}{\encodingdefault}{\sfdefault}{bx}{n}



%% file: iclr2027_conference.bib
@article{wei2022emergent,
  title={Emergent abilities of large language models},
  author={Wei, Jason and Tay, Yi and Bommasani, Rishi and Raffel, Colin and Zoph, Barret and Borgeaud, Sebastian and Yogatama, Dani and Bosma, Maarten and Zhou, Denny and Metzler, Donald and others},
  journal={arXiv preprint arXiv:2206.07682},
  year={2022}
}

@inproceedings{pan2023logic,
  author       = {Liangming Pan and
                  Alon Albalak and
                  Xinyi Wang and
                  William Yang Wang},
  editor       = {Houda Bouamor and
                  Juan Pino and
                  Kalika Bali},
  title        = {Logic-LM: Empowering Large Language Models with Symbolic Solvers for
                  Faithful Logical Reasoning},
  booktitle    = {Findings of the Association for Computational Linguistics: {EMNLP}
                  2023, Singapore, December 6-10, 2023},
  series       = {Findings of {ACL}},
  volume       = {{EMNLP} 2023},
  pages        = {3806--3824},
  publisher    = {Association for Computational Linguistics},
  year         = {2023},
  url          = {https://doi.org/10.18653/v1/2023.findings-emnlp.248},
  doi          = {10.18653/V1/2023.FINDINGS-EMNLP.248},
  bibsource    = {dblp computer science bibliography, https://dblp.org}
}

@article{meta2024introducing,
  title={Introducing meta llama 3: The most capable openly available llm to date},
  author={Meta, AI},
  journal={Meta AI},
  year={2024}
}

@article{bai2023qwen,
  title={Qwen technical report},
  author={Bai, Jinze and Bai, Shuai and Chu, Yunfei and Cui, Zeyu and Dang, Kai and Deng, Xiaodong and Fan, Yang and Ge, Wenbin and Han, Yu and Huang, Fei and others},
  journal={arXiv preprint arXiv:2309.16609},
  year={2023}
}

@article{abdin2024phi,
  title={Phi-3 technical report: A highly capable language model locally on your phone},
  author={Abdin, Marah and Aneja, Jyoti and Awadalla, Hany and Awadallah, Ahmed and Awan, Ammar Ahmad and Bach, Nguyen and Bahree, Amit and Bakhtiari, Arash and Bao, Jianmin and Behl, Harkirat and others},
  journal={arXiv preprint arXiv:2404.14219},
  year={2024}
}

@article{achiam2023gpt,
  title={Gpt-4 technical report},
  author={Achiam, Josh and Adler, Steven and Agarwal, Sandhini and Ahmad, Lama and Akkaya, Ilge and Aleman, Florencia Leoni and Almeida, Diogo and Altenschmidt, Janko and Altman, Sam and Anadkat, Shyamal and others},
  journal={arXiv preprint arXiv:2303.08774},
  year={2023}
}

@article{guo2025deepseek,
  title={Deepseek-r1: Incentivizing reasoning capability in llms via reinforcement learning},
  author={Guo, Daya and Yang, Dejian and Zhang, Haowei and Song, Junxiao and Zhang, Ruoyu and Xu, Runxin and Zhu, Qihao and Ma, Shirong and Wang, Peiyi and Bi, Xiao and others},
  journal={arXiv preprint arXiv:2501.12948},
  year={2025}
}

@article{wendler2024llamas,
  title={Do llamas work in english? on the latent language of multilingual transformers},
  author={Wendler, Chris and Veselovsky, Veniamin and Monea, Giovanni and West, Robert},
  journal={arXiv preprint arXiv:2402.10588},
  year={2024}
}

@inproceedings{li2024safety,
  author       = {Shen Li and
                  Liuyi Yao and
                  Lan Zhang and
                  Yaliang Li},
  title        = {Safety Layers in Aligned Large Language Models: The Key to {LLM} Security},
  booktitle    = {The Thirteenth International Conference on Learning Representations,
                  {ICLR} 2025, Singapore, April 24-28, 2025},
  publisher    = {OpenReview.net},
  year         = {2025},
  url          = {https://openreview.net/forum?id=kUH1yPMAn7},
  bibsource    = {dblp computer science bibliography, https://dblp.org}
}

@inproceedings{jin2024exploring,
    title = "Exploring Concept Depth: How Large Language Models Acquire Knowledge and Concept at Different Layers?",
    author = "Jin, Mingyu  and Yu, Qinkai  and
      Huang, Jingyuan  and
      Zeng, Qingcheng  and
      Wang, Zhenting  and
      Hua, Wenyue  and
      Zhao, Haiyan  and
      Mei, Kai  and
      Meng, Yanda  and
      Ding, Kaize  and
      Yang, Fan  and
      Du, Mengnan  and
      Zhang, Yongfeng",
    booktitle = "Proceedings of the 31st International Conference on Computational Linguistics",
    month = jan,
    year = "2025",
    address = "Abu Dhabi, UAE",
    publisher = "Association for Computational Linguistics",
    url = "https://aclanthology.org/2025.coling-main.37/",
    pages = "558--573",
}

@misc{OpenAI2024,
  author       = "{OpenAI}",
  title        = "{Learning to reason with LLMs}",
  year         = {2024},
  url          = {https://openai.com/index/learning-to-reason-with-llms/},
  note         = {Accessed: 2024-10-14}
}

@misc{LlamaWebsite2024,
  author       = "{LlamaWebsite}",
  title        = "{Introducing Llama 3.2}",
  year         = {2024},
  url          = {https://www.llama.com/},
  note         = {Accessed: 14-Oct-2024}
}

@article{guo2024deepseek,
  title={DeepSeek-Coder: When the Large Language Model Meets Programming--The Rise of Code Intelligence},
  author={Guo, Daya and Zhu, Qihao and Yang, Dejian and Xie, Zhenda and Dong, Kai and Zhang, Wentao and Chen, Guanting and Bi, Xiao and Wu, Yu and Li, YK and others},
  journal={arXiv preprint arXiv:2401.14196},
  year={2024}
}

@inproceedings{waswani2017attention,
  title={Attention is All you Need},
  author={Ashish Vaswani and Noam Shazeer and Niki Parmar and Jakob Uszkoreit and Llion Jones and Aidan N. Gomez and Lukasz Kaiser and Illia Polosukhin},
  booktitle={Neural Information Processing Systems},
  year={2017},
  url={https://api.semanticscholar.org/CorpusID:13756489}
}

@misc{nostalgebraist2020logit,
  author = {Nostalgebraist},
  title = {Interpreting GPT: The Logit Lens},
  year = {2020},
  howpublished = {LessWrong},
  url = {https://www.lesswrong.com/posts/AcKRB8wDpdaN6v6ru/interpreting-gpt-the-logit-lens},
  note = {Accessed: 2025-02-12}
}

@inproceedings{hewitt2019structural,
  title={A structural probe for finding syntax in word representations},
  author={Hewitt, John and Manning, Christopher D},
  booktitle={Proceedings of the 2019 Conference of the North American Chapter of the Association for Computational Linguistics: Human Language Technologies, Volume 1 (Long and Short Papers)},
  pages={4129--4138},
  year={2019}
}

@inproceedings{houlsby2019parameter,
  title={Parameter-efficient transfer learning for NLP},
  author={Houlsby, Neil and Giurgiu, Andrei and Jastrzebski, Stanislaw and Morrone, Bruna and De Laroussilhe, Quentin and Gesmundo, Andrea and Attariyan, Mona and Gelly, Sylvain},
  booktitle={International conference on machine learning},
  pages={2790--2799},
  year={2019},
  organization={PMLR}
}

@inproceedings{pfeiffer2020adapterfusion,
  author       = {Jonas Pfeiffer and
                  Aishwarya Kamath and
                  Andreas R{\"{u}}ckl{\'{e}} and
                  Kyunghyun Cho and
                  Iryna Gurevych},
  editor       = {Paola Merlo and
                  J{\"{o}}rg Tiedemann and
                  Reut Tsarfaty},
  title        = {AdapterFusion: Non-Destructive Task Composition for Transfer Learning},
  booktitle    = {Proceedings of the 16th Conference of the European Chapter of the
                  Association for Computational Linguistics: Main Volume, {EACL} 2021,
                  Online, April 19 - 23, 2021},
  pages        = {487--503},
  publisher    = {Association for Computational Linguistics},
  year         = {2021},
  url          = {https://doi.org/10.18653/v1/2021.eacl-main.39},
  doi          = {10.18653/V1/2021.EACL-MAIN.39},
  bibsource    = {dblp computer science bibliography, https://dblp.org}
}

@article{karimi2021compacter,
  title={Compacter: Efficient low-rank hypercomplex adapter layers},
  author={Karimi Mahabadi, Rabeeh and Henderson, James and Ruder, Sebastian},
  journal={Advances in Neural Information Processing Systems},
  volume={34},
  pages={1022--1035},
  year={2021}
}

@inproceedings{li2023prompt,
  title={Prompt distillation for efficient llm-based recommendation},
  author={Li, Lei and Zhang, Yongfeng and Chen, Li},
  booktitle={Proceedings of the 32nd ACM International Conference on Information and Knowledge Management},
  pages={1348--1357},
  year={2023}
}

@inproceedings{hu2021lora,
  author       = {Edward J. Hu and
                  Yelong Shen and
                  Phillip Wallis and
                  Zeyuan Allen{-}Zhu and
                  Yuanzhi Li and
                  Shean Wang and
                  Lu Wang and
                  Weizhu Chen},
  title        = {LoRA: Low-Rank Adaptation of Large Language Models},
  booktitle    = {The Tenth International Conference on Learning Representations, {ICLR}
                  2022, Virtual Event, April 25-29, 2022},
  publisher    = {OpenReview.net},
  year         = {2022},
  url          = {https://openreview.net/forum?id=nZeVKeeFYf9},
  bibsource    = {dblp computer science bibliography, https://dblp.org}
}

@inproceedings{tafjord2020proofwriter,
  title={ProofWriter: Generating Implications, Proofs, and Abductive Statements over Natural Language},
  author={Oyvind Tafjord and Bhavana Dalvi and Peter Clark},
  booktitle={Findings},
  year={2020},
  url={https://api.semanticscholar.org/CorpusID:229371222}
}

@article{han2022folio,
  title={Folio: Natural language reasoning with first-order logic},
  author={Han, Simeng and Schoelkopf, Hailey and Zhao, Yilun and Qi, Zhenting and Riddell, Martin and Zhou, Wenfei and Coady, James and Peng, David and Qiao, Yujie and Benson, Luke and others},
  journal={arXiv preprint arXiv:2209.00840},
  year={2022}
}

@article{srivastava2022beyond,
  title={Beyond the Imitation Game: Quantifying and extrapolating the capabilities of language models},
  author={BIG-bench authors},
  journal={Transactions on Machine Learning Research},
  issn={2835-8856},
  year={2023},
  url={https://openreview.net/forum?id=uyTL5Bvosj},
  note={}
}

@article{meng2022locating,
  title={Locating and editing factual associations in gpt},
  author={Meng, Kevin and Bau, David and Andonian, Alex and Belinkov, Yonatan},
  journal={Advances in neural information processing systems},
  volume={35},
  pages={17359--17372},
  year={2022}
}

@inproceedings{fan2024not,
  author       = {Siqi Fan and
                  Xin Jiang and
                  Xiang Li and
                  Xuying Meng and
                  Peng Han and
                  Shuo Shang and
                  Aixin Sun and
                  Yequan Wang},
  title        = {Not All Layers of LLMs Are Necessary During Inference},
  booktitle    = {Proceedings of the Thirty-Fourth International Joint Conference on
                  Artificial Intelligence, {IJCAI} 2025, Montreal, Canada, August 16-22,
                  2025},
  pages        = {5083--5091},
  publisher    = {ijcai.org},
  year         = {2025},
  url          = {https://doi.org/10.24963/ijcai.2025/566},
  doi          = {10.24963/IJCAI.2025/566},
  bibsource    = {dblp computer science bibliography, https://dblp.org}
}

@inproceedings{gromov2024unreasonable,
  author       = {Andrey Gromov and
                  Kushal Tirumala and
                  Hassan Shapourian and
                  Paolo Glorioso and
                  Daniel A. Roberts},
  title        = {The Unreasonable Ineffectiveness of the Deeper Layers},
  booktitle    = {The Thirteenth International Conference on Learning Representations,
                  {ICLR} 2025, Singapore, April 24-28, 2025},
  publisher    = {OpenReview.net},
  year         = {2025},
  url          = {https://openreview.net/forum?id=ngmEcEer8a},
  bibsource    = {dblp computer science bibliography, https://dblp.org}
}

@inproceedings{men2024shortgpt,
  author       = {Xin Men and
                  Mingyu Xu and
                  Qingyu Zhang and
                  Qianhao Yuan and
                  Bingning Wang and
                  Hongyu Lin and
                  Yaojie Lu and
                  Xianpei Han and
                  Weipeng Chen},
  editor       = {Wanxiang Che and
                  Joyce Nabende and
                  Ekaterina Shutova and
                  Mohammad Taher Pilehvar},
  title        = {ShortGPT: Layers in Large Language Models are More Redundant Than
                  You Expect},
  booktitle    = {Findings of the Association for Computational Linguistics, {ACL} 2025,
                  Vienna, Austria, July 27 - August 1, 2025},
  series       = {Findings of {ACL}},
  volume       = {{ACL} 2025},
  pages        = {20192--20204},
  publisher    = {Association for Computational Linguistics},
  year         = {2025},
  url          = {https://aclanthology.org/2025.findings-acl.1035/},
  bibsource    = {dblp computer science bibliography, https://dblp.org}
}

@article{cobbe2021gsm8k,
  author       = {Karl Cobbe and
                  Vineet Kosaraju and
                  Mohammad Bavarian and
                  Mark Chen and
                  Heewoo Jun and
                  Lukasz Kaiser and
                  Matthias Plappert and
                  Jerry Tworek and
                  Jacob Hilton and
                  Reiichiro Nakano and
                  Christopher Hesse and
                  John Schulman},
  title        = {Training Verifiers to Solve Math Word Problems},
  journal      = {CoRR},
  volume       = {abs/2110.14168},
  year         = {2021},
  url          = {https://arxiv.org/abs/2110.14168},
  eprinttype    = {arXiv},
  eprint       = {2110.14168},
  bibsource    = {dblp computer science bibliography, https://dblp.org}
}

@inproceedings{timkey2021all,
  title={All Bark and No Bite: Rogue Dimensions in Transformer Language Models Obscure Representational Quality},
  author={William Timkey and Marten van Schijndel},
  booktitle={Conference on Empirical Methods in Natural Language Processing},
  year={2021},
  url={https://api.semanticscholar.org/CorpusID:237453326}
}

@article{hotelling1933analysis,
  title={Analysis of a complex of statistical variables into principal components.},
  author={Hotelling, Harold},
  journal={Journal of educational psychology},
  volume={24},
  number={6},
  pages={417},
  year={1933},
  publisher={Warwick \& York}
}

@article{van2008visualizing,
  title={Visualizing data using t-SNE.},
  author={Van der Maaten, Laurens and Hinton, Geoffrey},
  journal={Journal of machine learning research},
  volume={9},
  number={11},
  year={2008}
}

@article{mcinnes2018umap,
  title={Umap: Uniform manifold approximation and projection for dimension reduction},
  author={McInnes, Leland and Healy, John and Melville, James},
  journal={arXiv preprint arXiv:1802.03426},
  year={2018}
}

@misc{weissteinInflectionPoint,
  author       = {Weisstein, Eric W.},
  title        = {Inflection Point},
  howpublished = {\url{https://mathworld.wolfram.com/InflectionPoint.html}},
  note         = {A Wolfram Web Resource. Accessed: 2025-09-23}
}

@article{qwen2,
    title   = {Qwen2 Technical Report}, 
    author  = {An Yang and Baosong Yang and Binyuan Hui and Bo Zheng and Bowen Yu and Chang Zhou and Chengpeng Li and Chengyuan Li and Dayiheng Liu and Fei Huang and Guanting Dong and Haoran Wei and Huan Lin and Jialong Tang and Jialin Wang and Jian Yang and Jianhong Tu and Jianwei Zhang and Jianxin Ma and Jin Xu and Jingren Zhou and Jinze Bai and Jinzheng He and Junyang Lin and Kai Dang and Keming Lu and Keqin Chen and Kexin Yang and Mei Li and Mingfeng Xue and Na Ni and Pei Zhang and Peng Wang and Ru Peng and Rui Men and Ruize Gao and Runji Lin and Shijie Wang and Shuai Bai and Sinan Tan and Tianhang Zhu and Tianhao Li and Tianyu Liu and Wenbin Ge and Xiaodong Deng and Xiaohuan Zhou and Xingzhang Ren and Xinyu Zhang and Xipin Wei and Xuancheng Ren and Yang Fan and Yang Yao and Yichang Zhang and Yu Wan and Yunfei Chu and Yuqiong Liu and Zeyu Cui and Zhenru Zhang and Zhihao Fan},
    journal = {arXiv preprint arXiv:2407.10671},
    year    = {2024}
}

@article{zhong2024beyond,
  title={Beyond English-centric LLMs: What language do multilingual language models think in?},
  author={Zhong, Chengzhi and Cheng, Fei and Liu, Qianying and Jiang, Junfeng and Wan, Zhen and Chu, Chenhui and Murawaki, Yugo and Kurohashi, Sadao},
  journal={arXiv preprint arXiv:2408.10811},
  year={2024}
}

@inproceedings{sun2025transformer,
  title={Transformer layers as painters},
  author={Sun, Qi and Pickett, Marc and Nain, Aakash Kumar and Jones, Llion},
  booktitle={Proceedings of the AAAI Conference on Artificial Intelligence},
  volume={39},
  number={24},
  pages={25219--25227},
  year={2025}
}

@article{zhao2024how,
  title={How do large language models handle multilingualism?},
  author={Zhao, Yiran and Zhang, Wenxuan and Chen, Guizhen and Kawaguchi, Kenji and Bing, Lidong},
  journal={Advances in Neural Information Processing Systems},
  volume={37},
  pages={15296--15319},
  year={2024}
}

@article{lad2024remarkable,
  title={The remarkable robustness of llms: Stages of inference?},
  author={Lad, Vedang and Lee, Jin Hwa and Gurnee, Wes and Tegmark, Max},
  journal={arXiv preprint arXiv:2406.19384},
  year={2024}
}

@inproceedings{alkhamissi2024llm,
  title={The llm language network: A neuroscientific approach for identifying causally task-relevant units},
  author={AlKhamissi, Badr and Tuckute, Greta and Bosselut, Antoine and Schrimpf, Martin},
  booktitle={Proceedings of the 2025 Conference of the Nations of the Americas Chapter of the Association for Computational Linguistics: Human Language Technologies (Volume 1: Long Papers)},
  pages={10887--10911},
  year={2025}
}

@inproceedings{skean2025layer,
  author       = {Oscar Skean and
                  Md Rifat Arefin and
                  Dan Zhao and
                  Niket Patel and
                  Jalal Naghiyev and
                  Yann LeCun and
                  Ravid Shwartz{-}Ziv},
  title        = {Layer by Layer: Uncovering Hidden Representations in Language Models},
  booktitle    = {Forty-second International Conference on Machine Learning, {ICML}
                  2025, Vancouver, BC, Canada, July 13-19, 2025},
  publisher    = {OpenReview.net},
  year         = {2025},
  url          = {https://openreview.net/forum?id=WGXb7UdvTX},
  bibsource    = {dblp computer science bibliography, https://dblp.org}
}

@article{wu2024semantic,
  title={The semantic hub hypothesis: Language models share semantic representations across languages and modalities},
  author={Wu, Zhaofeng and Yu, Xinyan Velocity and Yogatama, Dani and Lu, Jiasen and Kim, Yoon},
  journal={arXiv preprint arXiv:2411.04986},
  year={2024}
}

@inproceedings{cheng2024emergence,
  author       = {Emily Cheng and
                  Diego Doimo and
                  Corentin Kervadec and
                  Iuri Macocco and
                  Lei Yu and
                  Alessandro Laio and
                  Marco Baroni},
  title        = {Emergence of a High-Dimensional Abstraction Phase in Language Transformers},
  booktitle    = {The Thirteenth International Conference on Learning Representations,
                  {ICLR} 2025, Singapore, April 24-28, 2025},
  publisher    = {OpenReview.net},
  year         = {2025},
  url          = {https://openreview.net/forum?id=0fD3iIBhlV},
  bibsource    = {dblp computer science bibliography, https://dblp.org}
}

@inproceedings{
   zhang2023adaptive,
   title={Adaptive Budget Allocation for Parameter-Efficient Fine-Tuning },
   author={Qingru Zhang and Minshuo Chen and Alexander Bukharin and Pengcheng He and Yu Cheng and Weizhu Chen and Tuo Zhao},
   booktitle={The Eleventh International Conference on Learning Representations },
   year={2023},
   url={https://openreview.net/forum?id=lq62uWRJjiY}
}

@inproceedings{lin2023awq,
  title={AWQ: Activation-aware Weight Quantization for LLM Compression and Acceleration},
  author={Lin, Ji and Tang, Jiaming and Tang, Haotian and Yang, Shang and Chen, Wei-Ming and Wang, Wei-Chen and Xiao, Guangxuan and Dang, Xingyu and Gan, Chuang and Han, Song},
  booktitle={MLSys},
  year={2024}
}

@book{von1841organic,
  title={Organic chemistry in its applications to agriculture and physiology},
  author={von Liebig, Justus Freiherr and Playfair, Lyon Playfair Baron},
  year={1841},
  publisher={J. Owen}
}

@inproceedings{jawahar2019does,
  title={What does BERT learn about the structure of language?},
  author={Jawahar, Ganesh and Sagot, Beno{\^\i}t and Seddah, Djam{\'e}},
  booktitle={Proceedings of the 57th annual meeting of the association for computational linguistics},
  pages={3651--3657},
  year={2019}
}

@inproceedings{tenney2019bert,
  title={BERT rediscovers the classical NLP pipeline},
  author={Tenney, Ian and Das, Dipanjan and Pavlick, Ellie},
  booktitle={Proceedings of the 57th annual meeting of the association for computational linguistics},
  pages={4593--4601},
  year={2019}
}

@inproceedings{geva2021transformer,
  title={Transformer feed-forward layers are key-value memories},
  author={Geva, Mor and Schuster, Roei and Berant, Jonathan and Levy, Omer},
  booktitle={Proceedings of the 2021 Conference on Empirical Methods in Natural Language Processing},
  pages={5484--5495},
  year={2021}
}

@article{todd2023function,
  title={Function vectors in large language models},
  author={Todd, Eric and Li, Millicent L and Sharma, Arnab Sen and Mueller, Aaron and Wallace, Byron C and Bau, David},
  journal={arXiv preprint arXiv:2310.15213},
  year={2023}
}

@article{sia2024does,
  title={Where does in-context learning happen in large language models?},
  author={Sia, Suzanna and Mueller, David and Duh, Kevin},
  journal={Advances in Neural Information Processing Systems},
  volume={37},
  pages={32761--32786},
  year={2024}
}

@article{merullo2024talking,
  title={Talking heads: Understanding inter-layer communication in transformer language models},
  author={Merullo, Jack and Eickhoff, Carsten and Pavlick, Ellie},
  journal={Advances in Neural Information Processing Systems},
  volume={37},
  pages={61372--61418},
  year={2024}
}

@inproceedings{lepori2025racing,
  title={Racing thoughts: Explaining contextualization errors in large language models},
  author={Lepori, Michael A and Mozer, Michael Curtis and Ghandeharioun, Asma},
  booktitle={Proceedings of the 2025 Conference of the Nations of the Americas Chapter of the Association for Computational Linguistics: Human Language Technologies (Volume 1: Long Papers)},
  pages={3020--3036},
  year={2025}
}

@inproceedings{lu2025paths,
  title={Paths not taken: Understanding and mending the multilingual factual recall pipeline},
  author={Lu, Meng and Zhang, Ruochen and Eickhoff, Carsten and Pavlick, Ellie},
  booktitle={Proceedings of the 2025 Conference on Empirical Methods in Natural Language Processing},
  pages={15077--15107},
  year={2025}
}

@article{khandelwal2025language,
  title={How Do Language Models Compose Functions?},
  author={Khandelwal, Apoorv and Pavlick, Ellie},
  journal={arXiv preprint arXiv:2510.01685},
  year={2025}
}
